\documentclass[11pt]{article}

\usepackage[T1]{fontenc}
\usepackage[utf8]{inputenc}

\DeclareUnicodeCharacter{2265}{\ensuremath{\geq}}
\DeclareUnicodeCharacter{2264}{\ensuremath{\leq}}
\DeclareUnicodeCharacter{2208}{\ensuremath{\in}}
\DeclareUnicodeCharacter{2260}{\ensuremath{\neq}}
\DeclareUnicodeCharacter{2248}{\ensuremath{\approx}}

\usepackage{amsmath,amsthm}
\usepackage{mathtools}

\usepackage{XCharter}
\usepackage[xcharter]{newtxmath}   

\usepackage{bm}

\usepackage[round]{natbib}
\usepackage{xcolor}
\usepackage{graphicx}
\graphicspath{{../}{../../}{../../../}{./}}
\usepackage{booktabs}
\usepackage{array}
\usepackage{longtable}  \usepackage{adjustbox}  \usepackage{algorithm}
\usepackage{algorithmic}
\usepackage{subcaption}
\usepackage{multirow}

\providecommand{\TheoryTextWidth}{6.0in}
\providecommand{\TheoryTextHeight}{9.0in}
\usepackage[letterpaper]{geometry}
\usepackage{microtype}
\theoremstyle{plain}
\newtheorem{theorem}{Theorem}
\newtheorem{proposition}[theorem]{Proposition}
\newtheorem{lemma}[theorem]{Lemma}
\newtheorem{corollary}[theorem]{Corollary}
\newtheorem{conjecture}[theorem]{Conjecture}
\theoremstyle{definition}
\newtheorem{definition}[theorem]{Definition}
\newtheorem{remark}[theorem]{Remark}
\newtheorem{assumption}[theorem]{Assumption}

\usepackage{fancyhdr}
\usepackage{titlesec}
\titleclass{\part}{top}  \titleformat{\part}[display]
  {\normalfont\centering\huge\bfseries\scshape}{\partname~\thepart}{0.6em}{}
\titleformat{\section}
  {\normalfont\Large\bfseries\scshape}{\thesection}{0.7em}{}
\titlespacing*{\section}{0pt}{2.4ex plus 1ex minus .2ex}{1.2ex plus .2ex}
\titleformat{\subsection}
  {\normalfont\large\bfseries}{\thesubsection}{0.6em}{}
\titlespacing*{\subsection}{0pt}{1.9ex plus .8ex minus .2ex}{0.8ex plus .2ex}

\definecolor{linknavy}{HTML}{1A4D8F}
\usepackage{hyperref}
\hypersetup{
  colorlinks=true,
  linkcolor=linknavy,
  citecolor=linknavy,
  urlcolor=linknavy,
  pdftitle={Dead Directions: Geometric Singular Learning},
}
\usepackage{orcidlink}  

\usepackage{titling}
\pretitle{\begin{center}\LARGE\bfseries\scshape}
\posttitle{\par\end{center}\vskip 0.45em {\centering\rule{\linewidth}{0.8pt}\par}\vskip 0.7em}
\date{}
 
\newcommand{\kl}{\mathrm{KL}}
\newcommand{\rlct}{\lambda}
\newcommand{\fisher}{F}
\newcommand{\reals}{\mathbb{R}}
\newcommand{\expect}{\mathbb{E}}
\newcommand{\normal}{\mathcal{N}}
\DeclareMathOperator{\rank}{rank}
\DeclareMathOperator{\diag}{diag}
\DeclareMathOperator{\col}{col}

\DeclareMathOperator{\cov}{Cov}

\newcommand{\dd}{\mathrm{d}}
 \newif\ifexpanded
\expandedfalse  

\newif\ifbodyhasproofs
\bodyhasproofsfalse  

\providecommand{\xref}[2]{\ifexpanded\ref{#1}\else#2\fi}

\newif\ifsupp
\suppfalse
 \newif\ifanonymise
\anonymisetrue  

\newif\ifanonymousfriendly
\anonymousfriendlytrue

\anonymisefalse
\expandedtrue
\bodyhasproofstrue
\anonymousfriendlyfalse

\title{Dead Directions: Geometric Singular Learning}

  \author{
    Tejas Pradeep Shirodkar\thanks{Correspondence: \texttt{tejas.shirodkar@research.iiit.ac.in} \quad
      \orcidlink{0009-0001-3034-0087}\,\href{https://orcid.org/0009-0001-3034-0087}{0009-0001-3034-0087}} \\
    IIIT, Hyderabad
  }

\begin{document}

\maketitle

\begin{abstract}

Singular learning theory and information geometry study the same spaces: the former in resolved coordinates, the latter in original coordinates under a non-degeneracy assumption that overparameterised models violate. This paper carries one direction of the bridge between them, from Watanabe's invariants to Fisher geometry, through one primitive, the \emph{dead direction}: a unit vector along which the Fisher metric degenerates, equivalently a direction crossing the analytic singular set along which the KL divergence keeps a zero of high order, its \emph{KL order} set by how fast that divergence vanishes. Our central result recovers the KL order as the decay rate of the directional Fisher quadratic form approaching the singularity, in original coordinates, without a Hironaka resolution. A selection rule on smooth fibres translates this rate into Watanabe's single-direction contribution to the real log canonical threshold, and the recovery extends to multi-component crossings, multiplicity $m$, the singular fluctuation $\nu$, prior-RLCT shifts, and tempered posteriors. We then carry the rate into a deep network: a multi-layer K-FAC factorisation writes each Fisher block as a product of activation- and gradient-side rates with a duality between them, instantiated at residual streams, layer normalisation, and attention. A quotient theorem carries the rate to the gauge quotient for optimizers whose update commutes with the group action; Adam's per-coordinate preconditioner fails that condition, so we construct \textsc{DDCAdam}, an equivariant Adam-family preconditioner, and prove the quotient rate along its trajectory. The result is a trajectory-rate readout of Watanabe's triple $(\lambda, m, \nu)$ from one checkpoint's forward and backward passes, without posterior sampling.
 \end{abstract}

\part{Foundations and the bridging primitive}

\section{Introduction}
\label{sec:theory:intro}

A trained neural network is a single point in a high-dimensional \emph{parameter space}: one coordinate per weight, with training tracing a path to a setting that fits the data. In small classical models this endpoint is isolated, and perturbing the weights in any direction degrades the fit. Overparameterised networks behave differently. With far more parameters than the data constrains, the settings that fit equally well form continuous families rather than isolated points. Along some directions the loss does not move at all; along others it changes only at high order. The local geometry at the solution is \emph{degenerate}, a \emph{singularity} of the parameter space, and that degeneracy is informative: it reflects how much of the network's capacity the task uses and which directions the learnt function ignores.

\begin{figure}[t]
\centering
\includegraphics[width=\textwidth]{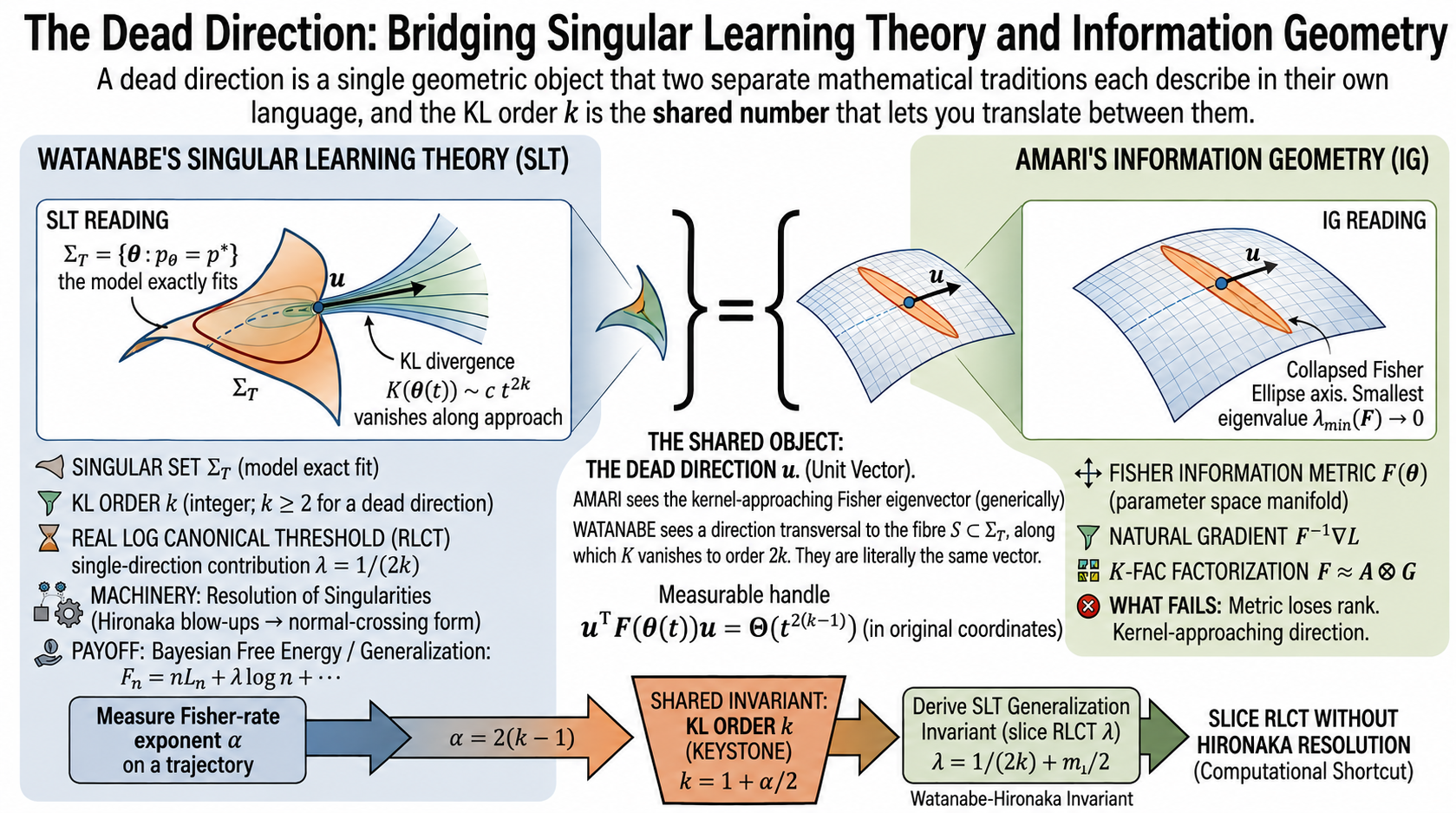}
\caption{The dead direction bridges the two traditions: the same unit vector is Amari's kernel-approaching direction of the Fisher metric $F$ and Watanabe's higher-order transversal to the singular set $\Sigma_T$, with KL order $k$ the shared invariant. The trajectory Fisher-rate exponent $\alpha = 2(k-1)$, read in original parameter coordinates, recovers $k$ without a Hironaka resolution, and with the codimension count $m_\perp$ of non-degenerate quadratic complements it gives the normal-slice RLCT $\lambda = 1/(2k) + m_\perp/2$, whose single-direction contribution is $1/(2k)$.}
\label{fig:bridge_watanabe_amari}
\end{figure}

The natural instrument for this geometry is the \emph{Fisher information metric} $\fisher(\theta) = \expect_{x \sim p^*}[\partial_\theta \log p_\theta(x)\, \partial_\theta \log p_\theta(x)^\top]$, which measures how sharply a model's predictions respond as the parameters move. A direction in which $\fisher$ is large is tightly constrained by the data; a direction in which $\fisher$ degenerates to zero is left free. The exactly-fitting parameters $\Sigma_T = \{\theta : p_\theta = p^*\}$ form the \emph{singular set}, and the Fisher metric loses rank along it.

Two communities have spent two decades studying this parameter space, in mostly separate vocabularies. Information geometry, following \citet{Amari16}, treats a parametric family $\{p_\theta : \theta \in \Theta\}$ as a Riemannian manifold under the Fisher metric. Natural gradient, the dual $(\nabla, \nabla^*)$ connection structure, and the exponential / mixture flatness duality are its central constructions, and all require the metric to be non-singular. Singular learning theory, following \citet{Watanabe09}, addresses the opposite case: a non-identifiable model whose parameter space contains an analytic singular set rather than an isolated optimum. Watanabe's framework computes invariants of $\Sigma_T$ through Hironaka's \emph{resolution of singularities}: a change of coordinates that untangles the singular set into simple products of crossings, a \emph{blow-up}. In those resolved coordinates the KL divergence takes a normal-crossing form, and one reads off the \emph{real log canonical threshold} $\lambda$, the asymptotic invariant governing the leading correction to the Bayesian free energy.\footnote{$\lambda$ takes rational values, a deep fact of algebraic geometry derived from \citet{Hironaka64}'s resolution; the resolution is guaranteed to exist but is not practically computable for typical neural-network loss landscapes.}

Both frameworks describe the same parameter space. Neither, on its own, answers a practitioner's question. Information geometry presupposes the metric is well-defined and so is silent about the singular set. Watanabe's $\lambda$ is integrated over the asymptotic posterior and lives in resolved coordinates that one would have to compute by performing a Hironaka blow-up of the loss landscape, a non-trivial task on a network with millions of parameters. The gap is operational: information about the singular structure exists in Watanabe's framework, but not in coordinates the practitioner has.

This paper closes one face of that gap with a single primitive. Write $\theta(t)$ for a path that approaches the singular set as the scalar $t \to 0$. A \emph{dead direction} is a unit vector $u$ (a direction in parameter space) along which the Fisher quadratic form $u^\top \fisher(\theta(t)) u$, the metric's length-squared of direction $u$, decays to zero on that approach. From Amari's vantage, it is the direction in which the foundational object of his framework loses non-degeneracy. From Watanabe's, it is the original-coordinate shadow of a normal-crossing exponent: a direction crossing the smooth strata of $\Sigma_T$ along which the divergence keeps a zero of high order, with \emph{KL order} $k \ge 1$ set by how fast the KL divergence vanishes on approach, $K(\theta(t)) = c\, t^{2k} + O(t^{2k+1})$. The divergence has a zero of order $2k$ in $t$, twice the KL order. The two readings refer to the same vector. We use the KL order, accessible in either framework, as the bridge invariant (Figure~\ref{fig:bridge_watanabe_amari}).

\begin{figure}[t]
\centering
\includegraphics[width=\textwidth]{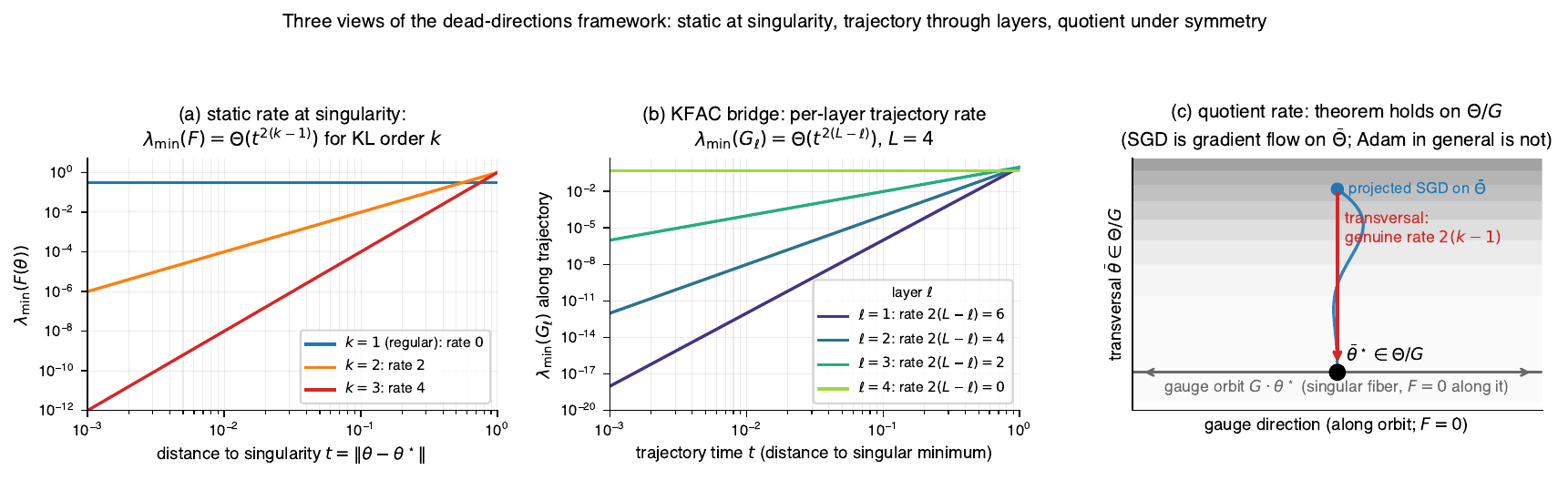}
\caption{Three views of the dead-directions framework. (a)~The rate primitive: along a dead direction with KL order $k$, the smallest Fisher eigenvalue decays as $\lambda_{\min}(F) = \Theta(t^{2(k-1)})$; the slope reads off $k$, recovering Watanabe's RLCT contribution $1/(2k)$ from a single trajectory in original coordinates. (b)~The K-FAC ladder at depth $L = 4$: the same exponent factorises across layers as $\lambda_{\min}(G_\ell) = \Theta(t^{2(L-\ell)})$, decreasing toward the input; the dual activation factor $\lambda_{\min}(A_\ell) = \Theta(t^{2(\ell-1)})$ makes the product layer-independent at $\Theta(t^{2(L-1)})$. (c)~The gauge quotient: under a continuous loss symmetry, the rate descends to $\Theta/G$; SGD on a $G$-invariant metric realises it as gradient flow, while Adam's per-coordinate preconditioner is not equivariant and the trajectory leaves the orbit foliation.}
\label{fig:dead_directions_overview}
\end{figure}

\paragraph{The rate primitive.} Along a dead direction with KL order $k$, the directional Fisher quadratic form satisfies $u^\top \fisher(\theta(t)) u = \Theta(t^{2(k-1)})$ (Theorem~\ref{thm:fisher_decay}). The smallest Fisher eigenvalue decays with a slope that reads off $k$: slope $0$ at the regular case $k=1$, slope $2$ at $k=2$, slope $4$ at $k=3$ (Figure~\ref{fig:dead_directions_overview}(a)). The proof is a score expansion in $L^2(p^*)$ with a Schur-complement bound on the non-degenerate Fisher block in adapted coordinates. The exponent reads in original parameter coordinates, with no Hironaka resolution required. It also identifies algebraically with Amari's degenerate direction, so the KL order is one invariant that both traditions can compute. The rest of the paper accounts for how this one exponent reorganises across a deep network: factorising over layers (Figure~\ref{fig:dead_directions_overview}(b)) and descending to the gauge quotient under loss symmetries (Figure~\ref{fig:dead_directions_overview}(c)).

\paragraph{Watanabe-side organisation.} On the SLT side, the rate corresponds to one of Watanabe's invariants. A selection rule for smooth singular fibres (Theorem~\ref{thm:selection_rule}) recovers the single-direction contribution $\lambda = 1/(2k)$ to the local RLCT: the Fisher slope, measured in original coordinates, is Watanabe's invariant. A Fisher--curvature--volume rate chain (Section~\ref{sec:theory:rate_chain}) ties the same $k$ to three jointly-determined geometric exponents.

\paragraph{Architecture-side organisation.} On the deep-learning side, the same exponent factorises across the layers of a network. A multi-layer K-FAC bridge (K-FAC: Kronecker-Factored Approximate Curvature, \citealp{MartensGrosse15}; the layer-wise factorisation of the Fisher into an activation factor $A_\ell$ and a gradient factor $G_\ell$) gives a per-layer rate ladder $\lambda_{\min}(G_\ell) = \Theta(t^{2(L-\ell)})$ (Theorem~\ref{thm:bridge}), with a forward--backward duality that makes the product layer-independent (Corollary~\ref{cor:a_g_duality}). Section~\ref{sec:theory:bridge} then instantiates the bridge at the architectural primitives of modern networks: rectangular widths, residual stacks, normalisation layers, and attention, each with closed-form rates derived from the primitive's algebra.

\paragraph{Trajectory-readability.} On the optimiser side, the rate descends to the gauge quotient. For any continuous Lie group symmetry $G$ of the loss, the trajectory rate is well-defined on the quotient $\Theta/G$ (Corollary~\ref{cor:quotient_rate}). Projected SGD on a $G$-invariant metric realises that rate, as gradient flow on the quotient through a Riemannian submersion (Corollary~\ref{cor:sgd_quotient}). Adam behaves differently: its per-coordinate $1/\sqrt{\hat v}$ preconditioner is not $G$-equivariant, so the trajectory-rate predictions do not transfer to Adam-class dynamics in closed form (Remark~\ref{rem:adam_nondescent}). A closed-form per-trajectory rate on the alignment-rotated manifold under such non-equivariant preconditioners remains open analytical work. We take a constructive route: we build an Adam-family preconditioner whose update map is $G$-equivariant by design (Algorithm~\ref{alg:ddcadam}, Corollary~\ref{cor:ddcadam_quotient_rate}), the Adam instance of a broader \emph{dead-direction conditioner} (DDC) family of equivariant preconditioners. The trajectory-rate predictions therefore apply to canonical-aligned SGD on a $G$-invariant metric and to the equivariant preconditioner family. The architecture-side and norm-side organisation above does not depend on this optimiser scope.

Section~\ref{sec:theory:extensions} carries the bridge into Watanabe's broader inventory: multi-component normal crossings, the multiplicity $m$, the singular fluctuation $\nu$ as a universal function of the KL order for $1$D dead directions, prior-induced RLCT shifts, and tempered posteriors at $\beta \ne 1$. The genuinely open agenda (exact resolution of determinantal singularities beyond depth $3$, non-i.i.d.\ data, a trajectory-rate analog of $\nu$ in original coordinates, theoretical $\beta$-sweep of $(\lambda, m)$, Amari-side dual-connections work, and a closed-form rate modifier for standard non-equivariant Adam) is collected in Section~\ref{sec:theory:open}. None is, as far as we can see, structurally blocked.

\paragraph{Reading guide.} The three contributions carry different reach, and the paper keeps the levels separate. Section~\ref{sec:theory:scope} sorts every load-bearing result into one of three tiers (Table~\ref{tab:reach_tiers}): a universal tier that holds for any analytic model with smooth singular fibres, an architecture-specific tier that requires the layered K-FAC structure, and a trajectory-readability tier that requires a gauge-equivariant optimiser. We use Lean~4 to check fourteen of the numbered results at their stated scope and the rest at explicit model instances; Appendix~\ref{app:theory:machine_checked} states, result by result, what the declarations prove and which hypotheses they take as inputs. The rate-to-RLCT recovery itself is exhibited numerically in Appendix~\ref{app:theory:parametric_validations}, where measured Fisher slopes return the predicted $\lambda = 1/(2k)$ on three controlled families. A reader who wants only the central result can take the rate primitive below, the selection rule of Section~\ref{sec:theory:selection}, and Figure~\ref{fig:dead_directions_overview}(a); the deep-network and optimiser machinery builds on that one exponent.

\paragraph{Notation and conventions.} Throughout, $\{p_\theta\}$ is a smooth parametric family on a parameter space $\Theta \subseteq \reals^d$. For a fixed task $T$, $p^*$ denotes the data-generating distribution, $\Sigma_T = \{\theta : p_\theta = p^*\}$ the analytic singular set, $K(\theta) := \kl(p^* \,\|\, p_\theta)$ the KL divergence, and $\fisher(\theta)$ the Fisher information matrix at $\theta$. A path $\theta(t)$ approaches a singular minimum $\theta_0 \in \Sigma_T$ as $t \to 0$. We write $f(t) = \Theta(g(t))$ for $c_1\, g(t) \le f(t) \le c_2\, g(t)$ on a punctured neighbourhood of $0$, with $0 < c_1 \le c_2$.
  
\section{Background: two traditions and their invariants}
\label{sec:theory:background}

Section~\ref{sec:theory:intro} named the two traditions and the gap between them. This section sets out the apparatus each provides: the metric structure information geometry assumes, and the resolution-of-singularities machinery singular learning theory uses to compute invariants where that structure fails.

\paragraph{Information geometry.} Information geometry \citep{Amari16} treats a parametric family $\{p_\theta : \theta \in \Theta\}$ as a Riemannian manifold under the Fisher information metric, whose $(i,j)$ entry is
\[
\fisher(\theta)_{ij} = \expect_{x \sim p^*}\!\left[\frac{\partial \log p_\theta(x)}{\partial \theta_i}\, \frac{\partial \log p_\theta(x)}{\partial \theta_j}\right].
\]
Together with the dual $(\nabla, \nabla^*)$ connection structure, the metric supports natural gradient as steepest descent in the Kullback--Leibler divergence and the exponential / mixture flatness duality of standard families. The apparatus is complete and actionable when the family is regular: $\fisher$ non-singular, the model identifiable, and the parameter space a clean Riemannian manifold. Even in the regular regime the Fisher spectrum is often strongly anisotropic: the \emph{sloppy model} analyses of \citet{TranstrumMachtaSethna11, TranstrumMachtaBrown15} find eigenvalues spread log-uniformly over many orders, with the smallest (sloppy) directions weakly constrained by the data. A sloppy direction carries a small but nonzero Fisher quadratic form with no integer order of vanishing. A dead direction, defined in \S\ref{sec:theory:setup}, is the case where the form vanishes to an integer KL order, the regime singular learning theory addresses.

\paragraph{Singular learning theory.} Singular learning theory \citep{Watanabe09} addresses the regime where these conditions fail. Many models of practical interest (mixture models, hidden Markov models, deep neural networks) are non-identifiable: distinct parameter values map to the same distribution \citep{Watanabe07}, and overparametrised networks are squarely in this regime \citep{WeiMurfet22}. At points where the model captures the data-generating distribution, the Fisher matrix is singular along the analytic set $\Sigma_T = \{\theta : p_\theta = p^*\}$.

Watanabe's framework computes invariants of $\Sigma_T$ via resolution of singularities. A sequence of analytic blow-ups (Hironaka's theorem) recasts the KL divergence in resolved coordinates as a normal-crossing form $K = u(g) \prod_i g_i^{2k_i}$, with normal-crossing exponents $(k_1, \dots, k_d)$ paired with Jacobian exponents $(h_1, \dots, h_d)$ from the blow-up. The \emph{real log canonical threshold} is $\lambda = \min_i (h_i + 1)/(2k_i)$, an integrated invariant of the resolved structure. It governs the leading-order correction to the Bayesian free energy: $F_n = nL_n + \lambda \log n - (m-1)\log\log n + O_p(1)$, where $m$ is the multiplicity of the minimum. With the singular fluctuation $\nu$, $\lambda$ also determines the asymptotic generalisation error via Watanabe's information criteria.

\paragraph{The shared geometry.} The two apparatus describe the same geometry (Figure~\ref{fig:bridge_watanabe_amari}): the Fisher metric is the Hessian of $K$ at $\theta^*$ when $K$ is twice-differentiable, and information geometry's metric-degenerate locus is Watanabe's singular set $\Sigma_T$. The remainder of the paper makes the KL order trajectory-readable in original parameter coordinates, then lifts that reading from a single direction to a deep network's layered Fisher spectrum and to the gauge quotient both traditions already encode.

\paragraph{Relationship to the local learning coefficient line.} Closest in spirit to the construction below is the local learning coefficient (LLC) programme of \citet{LauFurmanWangMurfetWei25}, which makes Watanabe's $\lambda$ pointwise by estimating it numerically via SGLD-based posterior sampling around a checkpoint. The LLC programme has since branched into a \emph{refined-LLC} variant \citep{WangHoogland24} that estimates $\lambda$ restricted to a chosen weight subset or data subset, yielding a per-module / per-data complexity number, and a \emph{stagewise-development} reading \citep{HooglandWangFarrugiaRoberts24} that tracks LLC change-points across training checkpoints. The trajectory-rate framework of this paper is a deterministic alternative to the same family of observables: it exposes Watanabe's $\lambda$ as a rate exponent on the Fisher metric along an approach (Theorem~\ref{thm:fisher_decay}), computable from forward and backward passes without posterior sampling, at the cost of requiring canonical alignment of the dead direction across layers. The trade-off is direct. The LLC family delivers a numerical Bayesian-complexity number per module and per checkpoint via SGLD; the trajectory-rate framework delivers a deterministic rate exponent per K-FAC layer (Theorem~\ref{thm:bridge}), a residual-stream $\sigma_{\min}$ readout (Corollary~\ref{cor:sigma-min-res}), an algebraic LN-kernel direction (Proposition~\ref{prop:ln_kernel}), and the rest of the architectural roster of \S\ref{sec:theory:bridge}. Theorem~\ref{thm:selection_rule} is the formal statement of where the two converge: on the smooth-fibre, single-component slice with Jacobian exponent set to zero. Section~\ref{sec:theory:extensions} carries the recovery further, into the multi-component case where the singular set is several smooth sheets meeting transversally (a normal crossing) and each sheet contributes its own KL order; there the bridge recovers the per-component orders and Watanabe's $(\lambda, m)$ pair. The LLC keeps coverage the rate readout lacks at singularities that resist a normal-crossing description or whose component normals are not identifiable, where its posterior sampling still returns an aggregate $\lambda$. We read the two programmes as complementary within a shared singular-learning agenda.

\paragraph{Parallel programs in the SLT-DL landscape.} Several other lines operationalise different aspects of singular structure in deep networks, and a single paragraph cannot do them justice; we name each and indicate the relationship to the trajectory-rate framework here, with the detailed forward-look in \S\ref{sec:theory:discussion}. \emph{Susceptibilities} \citep{BakerWangHooglandMurfet25,GordonBakerWang26} perturb the data distribution and read the response of posterior expectations of localised observables, treating the network as a thermodynamic medium; the programme clusters modules and tokens by response signature. \emph{The loss kernel} \citep{AdamFurmanHoogland25} is a geometric probe of model internals (the covariance of per-sample losses under low-loss-preserving perturbations), complemented by the Hessian-free Bayesian influence functions of \citet{KreerWuAdamFurmanHoogland25,LeeSmithAdamHoogland25}, which carry the same posterior geometry into data attribution sensitive to higher-order degeneracy. \emph{Loss-landscape degeneracy for interpretability} \citep{BushnaqMendel24} removes two of a trained network's leading-order degeneracies (the linear dependence among a layer's activations and among its backpropagated gradients; the construction identifies a third, co-firing ReLUs, without removing it) to expose a sparsely interacting feature basis; the activation and gradient Gram matrices it diagonalises are the K-FAC factors $A$ and $G$ that Corollary~\ref{cor:kfac_lift} reads for the dead direction. \emph{Compressibility and minimum description length} \citep{Urdshals25Compressibility} formalise the bit-length of a singular minimum, recovering classical MDL in the regular case and refining it in the degenerate one. \emph{Programs as singularities} \citep{MurfetTroiani25} connects degenerate statistical models to the geometry of programs. \emph{Modes of sequence models} \citep{ChenMurfet25} characterises the sensitivity of LLC estimation to input-distribution patterns. Each line probes a different slice of the singular geometry; the trajectory-rate framework is the deterministic, per-K-FAC-block, sampling-free reading of Watanabe's $\lambda$. We engage each line again in \S\ref{sec:theory:discussion} as a forward-look. \citet{Watanabe09,Watanabe18} are the canonical references for the foundations.
 
\section{Setup and notation}
\label{sec:theory:setup}

We collect the formal setting in which the rest of the paper operates.

\paragraph{The singular set.} Fix a smooth parametric family $\{p_\theta : \theta \in \Theta\}$ on a parameter space $\Theta \subseteq \reals^d$ and a data-generating distribution $p^*$. The \emph{singular set} for the task $T = (p^*)$ is
\[
\Sigma_T \;:=\; \{\theta \in \Theta : p_\theta = p^*\},
\]
the set of parameter values that exactly capture $p^*$. For a regular (identifiable) family, $\Sigma_T$ is either empty or a single point. For a non-identifiable family, $\Sigma_T$ is in general a positive-dimensional analytic set: a finite union of smooth strata of varying dimension, possibly with crossings. Throughout this paper we assume $\Sigma_T$ contains at least one smooth stratum near a point $\theta_0$ of interest, which is the regime of \citet{Watanabe09}'s normal-crossing analysis after resolution.

\paragraph{KL divergence and KL order.} For $\theta_0 \in \Sigma_T$ and a smooth path $\theta(t)$ with $\theta(0) = \theta_0$, the KL divergence $K(\theta(t)) := \kl(p^* \,\|\, p_{\theta(t)})$ vanishes at $t = 0$. The \emph{KL order} along the path is the integer $k \ge 1$ such that
\[
K(\theta(t)) \;=\; c\, t^{2k} + O(t^{2k+1}), \qquad c > 0.
\]
On a regular family with non-singular Fisher, every path through $\theta_0$ has KL order $k = 1$ (and $c = \tfrac12 u^\top \fisher(\theta_0) u > 0$ for direction $u = \dot\theta(0)$); Theorem~\ref{thm:fisher_decay} reads vacuously here, with rate $\Theta(t^0) = \Theta(1)$. The interesting cases are $k \ge 2$. These arise when $u$ crosses $\Sigma_T$ and the KL divergence still vanishes faster than quadratically along it, a degenerate zero of $K$ across the singular set; the rate theorem requires $k \ge 2$ to produce a non-trivial decay. A direction tangent to a stratum of $\Sigma_T$ carries no finite KL order when the line $\theta_0 + tu$ stays inside the stratum, the case of the gauge orbits and coordinate-aligned fibres in this paper; the tangent line to a \emph{curved} stratum leaves it at second order and picks up a finite order of its own. The next paragraph gives the score-level reason tangent directions start in the Fisher kernel either way.

\paragraph{Fisher information.} Recall the Fisher information $\fisher(\theta)$ from \S\ref{sec:theory:background}. At $\theta = \theta_0 \in \Sigma_T$, $\fisher$ has tangent directions to $\Sigma_T$ in its kernel: by gauge invariance of the KL (moving along $\Sigma_T$ leaves $p_\theta$ unchanged), the score $\partial_\theta \log p_\theta$ vanishes in $L^2(p^*)$ along these directions at $\theta = \theta_0$. The metric is non-singular only transversal to $\Sigma_T$, and even those transversal directions lose rank as the path approaches. The behaviour of $\fisher$ along an approach to $\Sigma_T$ is the central object of this paper.

\paragraph{K-FAC factorisation.} For a layered model with parameter blocks $\theta = (W_1, \dots, W_L)$ corresponding to weight matrices, the Fisher's per-layer block has a Kronecker-factored approximation
\[
F_\ell \;\approx\; A_\ell \otimes G_\ell,
\]
where $A_\ell := \expect[X_{\ell-1} X_{\ell-1}^\top]$ is the activation covariance at the input of layer $\ell$ and $G_\ell := \expect[\delta_\ell \delta_\ell^\top]$ is the gradient (back-propagated) covariance \citep{MartensGrosse15}.\footnote{The factorisation is exact for linear models at the optimum, with an $O(t^{2L})$ relative deviation along the approach (Remark~\ref{rem:kfac_approx}), and approximate for nonlinear; the bridge theorem's within-block rates are exact statements about $A_\ell$ and $G_\ell$ as defined and do not depend on the factorisation's approximation quality.} The factorisation discards cross-layer Fisher blocks and treats each layer's Fisher as a separable Kronecker product. We use the K-FAC structure not as a numerical approximation but as a coordinate-aware projection of $F$ that makes its singularity structure layer-readable: the gauge kernel of $GL(h)^{L-1}$ acting on cross-layer blocks belongs to the discarded part, and what remains is the transversal singular structure (Section~\ref{sec:theory:bridge}).

\begin{figure}[t]
\centering
\includegraphics[width=\textwidth]{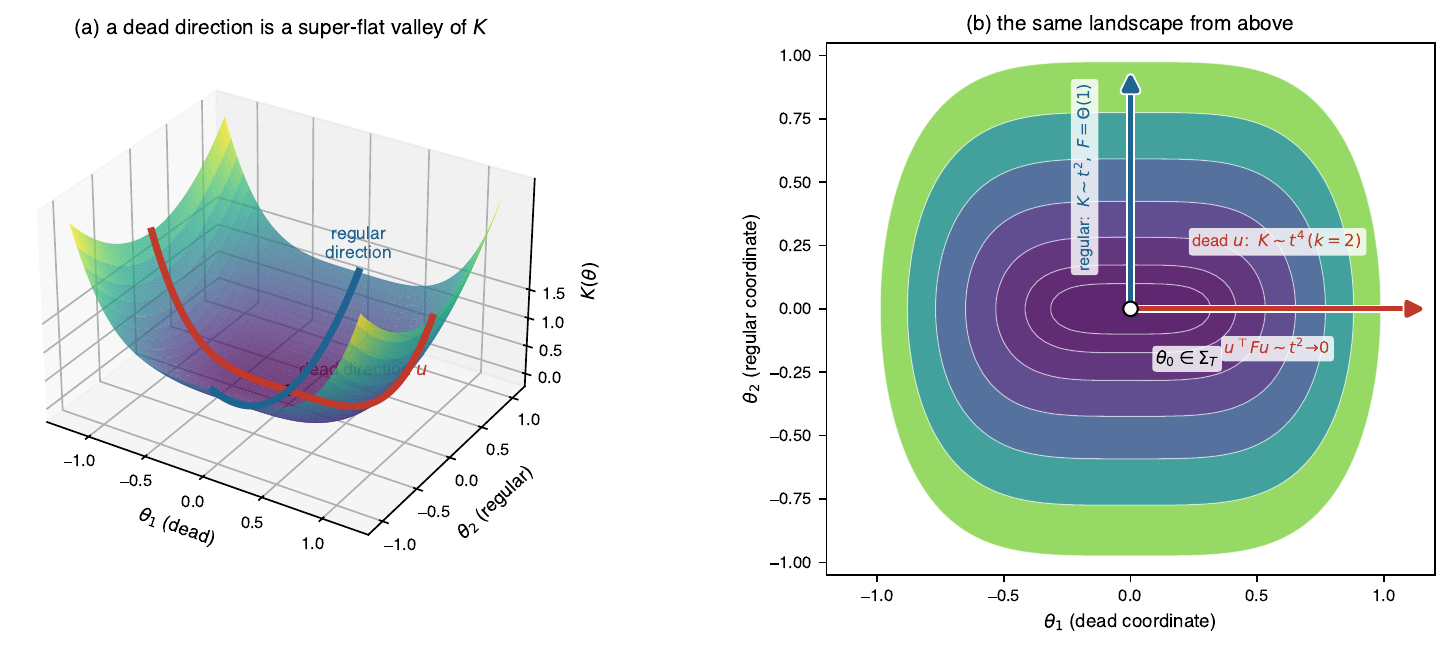}
\caption{What a dead direction is (Definition~\ref{def:dead_direction}). (a)~The KL divergence $K = \theta_1^4 + \theta_2^2$ as a landscape near a singular minimum $\theta_0 \in \Sigma_T$. The valley floor along the dead coordinate $\theta_1$ is super-flat ($K \sim t^4$, KL order $k = 2$), so the Fisher quadratic form decays, $u^\top F u \sim t^2 \to 0$; the transversal $\theta_2$ is a regular direction with $K \sim t^2$ and $u^\top F u = \Theta(1)$. (b)~The same landscape from above: the level sets stretch along the dead direction, with the singular set $\Sigma_T$ at the centre. A dead direction is any approach with KL order $k \ge 2$.}
\label{fig:dead_direction_geometry}
\end{figure}

\paragraph{The bridging primitive.} With this setting in place, we define the object that is visible in both Amari's and Watanabe's languages (Figure~\ref{fig:dead_direction_geometry}).
 
\begin{definition}[Dead direction]
\label{def:dead_direction}
The \emph{KL order} along a unit direction $u \in \reals^d$ at $\theta_0$, with $\theta(t) := \theta_0 + tu$ the straight-line path, is the integer $k \ge 1$ with $K(\theta(t)) = c \, t^{2k} + O(t^{2k+1})$ for some $c > 0$, so $K$ has a zero of order $2k$ in $t$. Call $u$ a \emph{dead direction} at $\theta_0$ if $u^\top \fisher(\theta(t)) u \to 0$ as $t \to 0$. When a finite KL order exists, under the hypotheses of Theorem~\ref{thm:fisher_decay} this happens exactly when $k \ge 2$: the rate is $\Theta(t^{2(k-1)})$, so $k = 1$ leaves the Fisher quadratic form at $\Theta(1)$ and describes a regular direction. A direction of KL order $1$ is therefore never dead, whatever suppression its leading coefficient carries. When no finite order exists because $K$ vanishes identically or to infinite order along the path (exact gauge directions; flat approaches), the limit clause still classifies $u$ as dead: these are the $k = \infty$ limiting case, they carry no rate, and every rate statement in this paper presupposes a finite $k \ge 2$.
\end{definition}

The term stratifies by KL order. An exact loss symmetry contributes the limiting $k = \infty$ case: the function is unchanged and $\fisher$ vanishes identically along the orbit, which lies within $\Sigma_T$. Section~\ref{sec:theory:quotient} treats these gauge directions and the quotient they induce. Theorem~\ref{thm:fisher_decay} concerns the genuine dead directions of finite order $k \ge 2$. Definition~\ref{def:dead_direction} names a direction in parameter space, unrelated to the activation-level ``dead ReLU'' phenomenon despite the lexical overlap.

\part{The rate primitive}

Part~I named the dead direction and the bridge thesis; this part builds the rate primitive that carries it. The shared invariant, the KL order $k$, is so far an abstract property of the singular geometry. Here it becomes a number read off an approach to the singular set. Section~\ref{sec:theory:rate} establishes the core translation: the directional Fisher rate exponent recovers the KL order $k$ in original parameter coordinates. Section~\ref{sec:theory:selection} turns that exponent into the single-direction contribution $\lambda = 1/(2k)$ to the normal-slice RLCT through a selection rule on smooth singular fibres. Section~\ref{sec:theory:rate_chain} shows the same $k$ governs two further geometric observables, the Fisher-Riemannian curvature and the high-curvature volume, so the three move together as one rate chain. The rest of the paper instantiates or extends this single exponent.

\paragraph{Reach.} The results in this part hold for any analytic algebraic-statistical model with smooth singular fibres: Gaussian mixtures, hidden Markov models, reduced-rank regression, exponential families, and neural networks with analytic activations all qualify. They are statements about the singular geometry of the model space, derived in original parameter coordinates without resolution of singularities. The later parts narrow this reach as deep-network geometry (Part~III) and optimiser dynamics (Part~IV) add architectural and trajectory-level conditions.
 
\section{Fisher rate decay}
\label{sec:theory:rate}

The KL order $k$ of \S\ref{sec:theory:setup} is a property of the singular geometry; the question is how to read it off. The natural way is to construct a Hironaka resolution of $K$ at $\theta_0$, push the monomialised form into normal-crossing coordinates, and inspect the leading exponent. The resolution is constructive in principle but intractable in practice for any non-trivial parametric family, and impossible at the scale of a deep network. The Fisher information matrix offers a more direct read: it is the second derivative of $K$ in the regular case, and its small eigenvalues track $K$'s vanishing in the singular case. The result of this section is that the connection is exact at the level of rate exponents. Along a dead direction with KL order $k$, the directional Fisher quadratic form $u^\top \fisher(\theta(t)) u$ decays as $\Theta(t^{2(k-1)})$: a rate exponent in original parameter coordinates that recovers $k$ without resolution.

The proof is a score expansion in $L^2(p^*)$ together with a Schur-complement bound on the non-degenerate Fisher block in adapted coordinates. After the theorem and proof, two worked examples (a Gaussian dead direction and a two-layer linear--ReLU network) show the rate exponent computed from first principles.
 
\begin{theorem}[Fisher rate decay along a dead direction]
\label{thm:fisher_decay}
Let $u$ be a single dead direction at $\theta_0$ with KL order $k \ge 2$, and let $\theta(t) := \theta_0 + tu$ be the path. Assume (i) in adapted coordinates with $u$ as the first axis, the non-degenerate Fisher block $g_{\alpha\beta}(t) := \fisher_{\alpha\beta}(\theta(t))$ ($\alpha, \beta$ orthogonal to $u$) is positive-definite at $t = 0$ and remains $\Theta(1)$; (ii) the leading log-likelihood coefficient $a_k(x) := \tfrac{1}{k!} \partial_t^k \log p_{\theta(t)}(x)|_{t=0}$ is not in the $L^2(p^*)$-span of the non-degenerate scores $\{\partial_\alpha \log p^*\}$; (iii) the score expansion $\partial_t \log p_{\theta(t)} = \sum_{j \ge 1} j\, t^{j-1} a_j$ converges in $L^2(p^*)$ with its remainder after the leading term of $L^2$-norm $O(t^k)$, which standard uniform integrability ($\expect_{p^*}[|a_j|^p] < \infty$ for all $j, p$, with the series dominated in $L^2$) provides; this is the form of the hypothesis the machine-checked proof consumes. Then
\[
u^\top \fisher(\theta(t)) u \;=\; \Theta(t^{2(k-1)}), \qquad \lambda_{\min}(\fisher(\theta(t))) \;=\; \Theta(t^{2(k-1)}).
\]
\end{theorem}

\begin{proof}[Sketch]
The score expansion $\log p_{\theta(t)} / p^* = \sum_{j \ge 1} a_j t^j$ together with the KL identity inductively forces $a_1 = \cdots = a_{k-1} = 0$ in $L^2(p^*)$. The directional Fisher then satisfies $u^\top \fisher u = \expect_{p^*}[s_u^2] = k^2 t^{2(k-1)} \expect_{p^*}[a_k^2] + O(t^{2k-1})$, and a Schur-complement bound on the Fisher block in adapted coordinates uses assumption (ii) to give the matching eigenvalue rate. Full proof: \S\ref{app:proof:fisher_decay}.
\end{proof}

\paragraph{Three readings of the rate.}\label{rem:fisher_decay_three_readings}
The same content takes three useful forms. \emph{As a slope.} Plot $\log u^\top \fisher u$ against $\log t$ along the approach. The asymptotic slope is $2(k-1)$, so $\hat k = 1 + \alpha/2$ recovers the KL order from a single measurement. \emph{As a Watanabe invariant.} For a smooth-fibre singularity with the Jacobian exponent set to zero, the same $k$ is the \emph{normal-slice} RLCT denominator: the normal-bundle subfamily has slice RLCT $1/(2k)$ (Theorem~\ref{thm:selection_rule}(c), whose scope note separates the slice value from the full local RLCT under tangential-normal coupling). Reading the slope gives Watanabe's single-direction $\lambda$ contribution in original parameter coordinates, with no Hironaka resolution required. \emph{As a basin shape.} A quadratic well $K(t) = c t^2$ ($k = 1$) gives flat Fisher decay ($\alpha = 0$), the regular case. A quartic well $K(t) = c t^4$ ($k = 2$) gives $\alpha = 2$. A sextic well $K(t) = c t^6$ ($k = 3$) gives $\alpha = 4$. Each integer KL order corresponds to a basin whose volume grows like $\varepsilon^{1/(2k)}$ at tolerance $\varepsilon$: the log-volume-versus-log-tolerance slope reads $1/(2k)$ while the Fisher slope reads $2(k-1)$, two different functions of the same $k$. Figure~\ref{fig:dead_directions_overview}(a) shows the three cases overlaid.

\paragraph{Scope of assumption (iii); discrete distributions.}\label{rem:fisher_decay_uniform_integrability}
Assumption (iii) requires all directional log-likelihood derivatives to lie in $L^p(p^*)$ for every finite $p$. This covers the smooth parametric families used in this paper (Gaussian, softmax cross-entropy on classifiers with bounded logits, and ReLU MLPs at configurations whose dead-unit pre-activations avoid the kink on a $p^*$-full-measure set, so that the two-sided coefficients $a_j$ of (ii)--(iii) exist). At the canonical zero-row configuration of the bridge chunks the dead pre-activation is exactly at the kink and the one-sided $t$-derivatives of $\mathrm{ReLU}(tz)$ differ ($z$ against $0$), so smoothness fails before integrability; the bridge theorems compute those rates directly on the one-sided approach $t \downarrow 0$ and do not route through (iii). Discrete distributions where $\partial_\alpha \log p^*$ is not uniformly $L^p$ (e.g., distributions with mass at the boundary of the support, certain mixture-degeneration limits) require strengthening (iii) at the affected boundary. In the other direction, the leading rate needs far less than (iii): the score expansion $s_t = k t^{k-1} a_k + r(t)$ with $\expect[a_k^2]$ finite and the single second-moment bound $\int r(t)^2\, dp^* \le M t^{2k}$ already gives $u^\top \fisher u = k^2 \expect[a_k^2]\, t^{2(k-1)} + O(t^{2k-1})$; the full moment hypothesis enters only through the higher-order coefficients.

\paragraph{Leading Fisher--KL ratio: $2k^2$, not $2k(2k-1)$.}\label{rem:leading_fisher_kl_ratio}
The proof sketch's $k^2 \expect_{p^*}[a_k^2]$ leading Fisher coefficient yields a clean structural identity. Writing $K(\theta(t)) = c_K t^{2k} + O(t^{2k+1})$ for the leading KL Taylor coefficient with $c_K = \frac{1}{2}\expect_{p^*}[a_k^2]$ (under base-distribution parity ensuring $\expect[a_k] = 0$), the leading Fisher--KL ratio is
\[
\frac{[u^\top \fisher u]_{t^{2(k-1)}}}{c_K} \;=\; \frac{k^2 \expect[a_k^2]}{\expect[a_k^2]/2} \;=\; 2 k^2.
\]
This contrasts with the naive $K''(t)$-based prediction $2k(2k-1)$ that would hold if Fisher equaled the Hessian of KL (correct only for regular $k=1$ models, where $2k^2 = 2k(2k-1) = 2$ coincide). At a singular point ($k \ge 2$) the population Fisher $\expect_{p_t}[s_t^2]$ at parameter $t$ differs from the KL Hessian $K''(0\Vert t)$ by a fluctuation contribution; the ratio $2k^2$ captures the singular-geometry correction. Verified analytically on the 2-component Gaussian mixture along the split-component dead direction ($k=2$, ratio $=8$) for any mixture weight $w \in (0,1)$. On the Gaussian location curve $\mu(t) = t^k$ the leading-order statement strengthens to an identity at every $t$: the moving-measure Fisher is $k^2 t^{2(k-1)}$ and the divergence from the base point is $t^{2k}/2$, so $F(t)\, t^2 = 2k^2\, K(t)$ identically.

The next-order coefficients are also derivable from family-level moments. Writing the deviation $p_{\theta(t)}/p^* = 1 + A(x) t^k + B(x) t^{k+1} + C(x) t^{k+2} + O(t^{k+3})$ and assuming base-distribution parity $\expect[A] = \expect[B] = \expect[C] = 0$, the next-order Fisher and KL coefficients at $k=2$ are
\[
K_{2k+2} = \frac{\expect[B^2]}{2} - \frac{\expect[A^3]}{3}, \qquad
F_{2k} = 9 \expect[B^2] - 4 \expect[A^3],
\]
varying with the family through the moments $\expect[B^2]$ and $\expect[A^3]$. The Fisher moments here follow the proof's convention $\expect_{p_{\theta(t)}}$ (the moving measure); under the background's $\expect_{p^*}$ convention the $t^{2k}$ coefficient shifts by $4\expect[A^3]$ at $k = 2$, while the leading rate and the $2k^2$ ratio are convention-independent. These coefficients characterise the sub-leading trajectory structure of the Fisher and KL expansions; they do not determine Watanabe's singular fluctuation $\nu$. As Section~\ref{sec:theory:extensions} shows, $\nu$ is an integral over the renormalised posterior on the singular fibre; it is universal, a function of the KL order alone (Theorem~\ref{thm:nu_universality}), with $V[\sqrt T] = \lambda - (\Gamma(\lambda+1/2)/\Gamma(\lambda))^2$, $\lambda = 1/(2k)$, the only elementary closed form available, and that only the frozen-fluctuation value at even $k$ (the full value is numerical). Verified on the 2-component Gaussian mixture (asymmetric merge, $w \neq 1/2$), where $F_{2k}/K_{2k+2} = 6(20w^2 - 20w + 3)/(8w^2 - 8w + 1)$ equals $18$ in the $w \to 0$ limit and $12$ at the symmetric merge $w = 1/2$ ($11.34$ at $w = 0.35$), with poles at $w = 1/2 \pm \sqrt{2}/4$ where the next-order KL coefficient $K_{2k+2}$ crosses zero.

\begin{remark}[Population Fisher, the loss-gradient covariance, and what the rate needs]
\label{rem:population_vs_empirical_fisher}
Theorem~\ref{thm:fisher_decay} concerns the population Fisher $\fisher(\theta) = \expect_{p_\theta}[s\, s^\top]$ evaluated on the controlled parametric approach $\theta(t) = \theta_0 + tu$; the proof takes the expectation under the moving measure $p_{\theta(t)}$, and under the fixed measure $p^*$ the leading rate and coefficient are unchanged whenever the density ratio is $1 + O(t)$, which the theorem's hypotheses supply, so only the next-order coefficients depend on the convention. Two things stand between that statement and a number read off a network. First, the \emph{estimator}: pipelines compute the loss-gradient covariance $\widehat G(\theta) = N^{-1} \sum_i \delta_i \delta_i^\top$ (the back-propagated gradient covariance the K-FAC bridge of \S\ref{sec:theory:bridge} uses), which equals the population Fisher only at a well-specified configuration where the residual variance is the model noise, the empirical-Fisher limitation of \citet{KunstnerHennigBalles19}. Second, the \emph{configuration}: the theorem describes a controlled approach along a fixed dead direction, not the endpoint of an optimiser. On the controlled parametric approach (the freeze-probe of \S\ref{app:theory:parametric_validations}, where $t$ is set by hand and the residual stays $\Theta(1)$) both issues are absent, and the leading exponent $2(k-1)$ is recovered to three decimals across activations and families. On a \emph{learned} trajectory both bite: a global $\|\delta\|^2$ prefactor multiplies the spectrum, so at a fitted optimum $\delta \to 0$ collapses every eigenvalue together (no isolated dead direction) and on a stalled high-noise plateau the dead block never enters the descent window (Remark~\ref{rem:asymptotic_window}); the canonical-aligned eigenvector lineage can also rotate. More insidiously, when the prefactor itself carries a power of the trajectory parameter (as for a squared-error loss, whose residual scales with the loss), the readout returns a clean power law at a \emph{shifted} exponent (the noise-free exponent shift of \S\ref{sec:theory:scope} and Remark~\ref{rem:bridge_regime}): a confident fit to the wrong rate, not an obvious failure, so fit quality alone does not certify the exponent. The rate is therefore validated on the controlled population-Fisher approach, and a learned-trajectory readout recovers it only for a well-specified model inside the asymptotic window with the lineage preserved.

Higher-order coefficients are data-dependent regardless. The coefficient at order $t^{2k-1}$ (the early $\nu$-candidate that \S\ref{sec:theory:extensions} rules out) is zero under symmetric base distributions by Hermite orthogonality and generically non-zero under asymmetric data. Leading-rate slope fits ($\log \lambda_{\min}$, $\log \det \fisher$) are robust singular-geometry diagnostics under the controlled conditions above; reported on a learned trajectory they must carry the estimator and regime explicitly.
\end{remark}

\subsection{Proof of Theorem~\ref{thm:fisher_decay}}
\label{app:proof:fisher_decay}

\paragraph{Setup recap.} Let $\{p_\theta\}$ be analytic with $p_{\theta_0} = p^*$ and a unit \emph{dead direction} $u$ at $\theta_0$ (in the sense of Definition~\ref{def:dead_direction} of the body), $K(\theta_0 + tu) = c t^{2k} + O(t^{2k+1})$, $c > 0$, $k \ge 2$. Assumptions (i)--(iii) of Theorem~\ref{thm:fisher_decay}: single degenerate direction; leading log-likelihood coefficient $a_k$ not in the $L^2(p^*)$-span of non-degenerate scores; uniform integrability $\expect_{p^*}[|a_j|^p] < \infty$ for all $j, p$. Goal: $\lambda_{\min}(\fisher(\theta(t))) = \Theta(t^{2(k-1)})$.

\paragraph{Score expansion.} Analyticity gives $\log p_{\theta(t)}(x) = \log p^*(x) + \sum_{j \ge 1} a_j(x) t^j$ with $a_j(x) = \tfrac{1}{j!} \partial_t^j \log p_{\theta(t)}|_{t=0}$. Write $f(x, t) := \log(p_{\theta(t)}/p^*) = \sum_{j \ge 1} a_j t^j$. Normalisation and the KL identity give
\begin{equation}\label{eq:norm_kl}
\expect_{p^*}[\exp f] = 1, \qquad K(t) = -\expect_{p^*}[f].
\end{equation}

\paragraph{Inductive vanishing of \texorpdfstring{$a_1, \ldots, a_{k-1}$}{a1, , ak-1}.} \emph{Base ($j = 1$):} expanding $\exp f = 1 + f + \tfrac{1}{2}f^2 + \cdots$, the order-$t^2$ coefficient of $\expect_{p^*}[\exp f] = 1$ collects the partitions of $2$: $\{2\}$ and $\{1, 1\}$. This gives $\expect_{p^*}[a_2] + \tfrac{1}{2} \expect_{p^*}[a_1^2] = 0$. Substituting into~\eqref{eq:norm_kl}, the order-$t^2$ coefficient of $K$ is $-\expect_{p^*}[a_2] = \tfrac{1}{2} \expect_{p^*}[a_1^2]$. Since $K(t) = c t^{2k}$ with $k \ge 2$, this vanishes, so $\expect_{p^*}[a_1^2] = 0$ and $a_1 = 0$ $p^*$-a.s.

\emph{Inductive step:} suppose $a_1 = \cdots = a_{j-1} = 0$ for some $j \le k - 1$. Under this hypothesis, $f = \sum_{i \ge j} a_i t^i$, so the order-$t^{2j}$ coefficient of $\exp f$ collects only multi-indices $(i_1, \ldots, i_m)$ with $i_l \ge j$ for all $l$ and $\sum_l i_l = 2j$; the only such partitions are $\{2j\}$ and $\{j, j\}$ (any $\{i, 2j - i\}$ with $i < j$ violates the lower bound, and triples $\{i_1, i_2, i_3\}$ require $\sum_l i_l \ge 3j > 2j$). The contribution is therefore $\expect_{p^*}[a_{2j}] + \tfrac{1}{2}\expect_{p^*}[a_j^2] = 0$. Substituting into~\eqref{eq:norm_kl}, the order-$t^{2j}$ coefficient of $K$ is $-\expect_{p^*}[a_{2j}] = \tfrac{1}{2}\expect_{p^*}[a_j^2]$. Since $j < k$ we have $K_{2j} = 0$, so $a_j = 0$ $p^*$-a.s.

At $j = k$: by the same partition argument, $K_{2k} = \tfrac{1}{2}\expect_{p^*}[a_k^2] = c$, so $\expect_{p^*}[a_k^2] = 2c > 0$.

\paragraph{Directional Fisher.} The directional score is $s_u(x, t) = \partial_t f = \sum_{j \ge k} j a_j t^{j-1} = k a_k t^{k-1} + O(t^k)$. The directional Fisher reads $u^\top \fisher u = \expect_{p_\theta}[s_u^2] = \expect_{p^*}[s_u^2 \exp f]$. Expanding $\exp f = 1 + a_k t^k + O(t^{k+1})$ and $s_u^2 = k^2 a_k^2 t^{2(k-1)} + O(t^{2k-1})$:
\[
u^\top \fisher u = k^2 t^{2(k-1)} \expect_{p^*}[a_k^2] + O(t^{2k-1}) = 2 c k^2 t^{2(k-1)} + O(t^{2k-1}) = \Theta(t^{2(k-1)}).
\]
The $O(t^{2k-1})$ remainder absorbs two distinct sources, both subleading to $t^{2(k-1)}$ for $k \ge 2$: (a) the next-order contribution to $s_u^2 = k^2 a_k^2 t^{2(k-1)} + 2k(k+1) a_k a_{k+1} t^{2k-1} + O(t^{2k})$ from the $j = k+1$ term in the score expansion, and (b) the change-of-measure correction $\expect_{p^*}[s_u^2 (\exp f - 1)]$, whose leading term is $\expect_{p^*}[k^2 a_k^2 t^{2(k-1)} \cdot a_k t^k] = k^2 t^{3k-2} \expect_{p^*}[a_k^3]$ (finite by~(iii)). For $k \ge 2$, both $2k - 1$ and $3k - 2$ exceed $2(k-1)$, so the leading $\Theta(t^{2(k-1)})$ rate is preserved regardless of relative sign.

\paragraph{Eigenvalue bound via Schur complement.} In adapted coordinates with $\theta^1 = u$ and $\theta^\alpha$ orthogonal:
\begin{itemize}\itemsep=1pt
\item $\fisher_{11}(t) = c_F t^{2(k-1)} + O(t^{2k-1})$, $c_F = 2 c k^2$ (computed above).
\item $\fisher_{1 \alpha}(t) = b_\alpha t^{k-1} + O(t^k)$, $b_\alpha = k \expect_{p^*}[a_k \cdot \partial_\alpha \log p^*]$.
\item $\fisher_{\alpha \beta}(t) = g_{\alpha \beta} + O(t)$, with $g_{\alpha \beta} = \expect_{p^*}[(\partial_\alpha \log p^*)(\partial_\beta \log p^*)] \succ 0$.
\end{itemize}
The Schur complement formula gives $\det \fisher(t) = \det(g + O(t)) \cdot (c_F - b^\top g^{-1} b) t^{2(k-1)} + O(t^{2k-1})$. The factor $c_F - b^\top g^{-1} b = k^2 (\expect_{p^*}[a_k^2] - \|\mathrm{Proj}_{\mathrm{span}\{\partial_\alpha \log p^*\}} a_k\|^2_{L^2(p^*)})$ is the squared $L^2(p^*)$ norm of $k a_k$ projected onto the orthogonal complement of the non-degenerate scores. Explicitly: assumption (ii) states that $a_k$ is not in the $L^2(p^*)$-linear span of the non-degenerate-direction scores $\{\partial_\alpha \log p^*\}$, which is equivalent to saying the $L^2(p^*)$ projection of $a_k$ onto the orthogonal complement of that span is nonzero; squaring the nonzero projected norm gives the strict positivity $c_F - b^\top g^{-1} b > 0$. Hence $\det \fisher(t) = \Theta(t^{2(k-1)})$. Combined with assumption (i) (non-degenerate eigenvalues stay $\Theta(1)$), $\lambda_{\min}(\fisher(t)) = \Theta(t^{2(k-1)})$. The upper bound matches by $\lambda_{\min}(\fisher) \le u^\top \fisher u$. \qed
 
\subsection{Worked example: Gaussian dead direction}
\label{sec:theory:rate:worked_gaussian}

The simplest setting in which a non-trivial KL order arises is a one-parameter sub-family of Gaussian distributions where the mean and variance are linked. Let $p_\theta = \mathcal{N}(\theta^k, 1)$ with $\theta \in \reals$ and $k \ge 1$ an integer, and let the data-generating distribution be $p^* = \mathcal{N}(0, 1)$. The singular set is $\Sigma_T = \{0\}$, a single point.

Compute the KL divergence directly:
\[
K(\theta) \;=\; \kl(p^* \,\|\, p_\theta) \;=\; \tfrac12 \theta^{2k},
\]
so $K(\theta(t)) = \tfrac12 t^{2k}$ along the path $\theta(t) = t$, and the KL order is $k$ by direct read-off of the leading exponent.

Now compute the directional Fisher. The score along the unit direction $u = 1$ (only one parameter) is
\[
s(\theta, x) \;=\; \frac{d}{d\theta} \log p_\theta(x) \;=\; (x - \theta^k) \cdot k\, \theta^{k-1}.
\]
At $\theta(t) = t$:
\[
u^\top \fisher(\theta(t)) u \;=\; \expect_{x \sim p^*}\!\left[s(\theta(t), x)^2\right] \;=\; k^2\, t^{2(k-1)} \cdot \expect_{x \sim p^*}\!\left[(x - t^k)^2\right].
\]
The expectation is $1 + t^{2k}$, so the leading-order behaviour as $t \to 0$ is
\[
u^\top \fisher(\theta(t)) u \;=\; k^2\, t^{2(k-1)} + O(t^{2k}).
\]
This matches Theorem~\ref{thm:fisher_decay} exactly: the rate exponent is $2(k-1)$, and the leading coefficient is $k^2 \expect_{p^*}[a_k^2]$ where $a_k(x) = (x - 0) \cdot 1 = x$ in this normalisation, giving $\expect[a_k^2] = 1$ and the prefactor $k^2$.

For $k = 1$ (the regular case), the rate exponent is $0$ and the Fisher is $\Theta(1)$ as expected. For $k = 2$, the exponent is $2$ and the Fisher vanishes quadratically. For $k = 3$, the exponent is $4$.

The local RLCT recovered from the rate via the selection rule (Theorem~\ref{thm:selection_rule}) is $\hat\lambda = 1/(2k)$, agreeing with the direct computation on a one-parameter family: the Watanabe-side computation in resolved coordinates collapses to the same expression because there is only one direction to resolve.

This example illustrates the rate identity in its simplest form. Two observations carry through to richer settings. First, the score-expansion identity $a_1 = \cdots = a_{k-1} = 0$ in $L^2(p^*)$ is an inductive consequence of the KL identity (the expansion of $K$ has no terms below $t^{2k}$), and the leading non-vanishing $a_k$ produces the $t^{2(k-1)}$ rate exactly. Second, Watanabe's RLCT and the directional Fisher rate are the same invariant in two frames.
 
\subsection{Worked example: two-layer linear--ReLU network}
\label{sec:theory:rate:worked_relu}

Consider a two-layer network $f_\theta(x) = W_2 \phi(W_1 x)$ with hidden width $h$, input dimension $d$, output dimension $1$, and activation $\phi \in \{\text{identity}, \text{ReLU}\}$. Train on regression data $y = W_2^* \phi(W_1^* x) + \varepsilon$ with isotropic Gaussian input $x \sim \mathcal{N}(0, I_d)$ and noise $\varepsilon \sim \mathcal{N}(0, \sigma^2)$. The teacher $(W_1^*, W_2^*)$ has a fully-dead hidden unit at index $h_0 \in \{1, \dots, h\}$: both the outgoing weight $(W_2^*)_{h_0} = 0$ and the incoming row $(W_1^*)_{h_0,:} = 0$, so the network is invariant to perturbations of either at the teacher and the leading-order analysis below sees both factors of $t$. The dead-hidden-unit configuration is the canonical example used in the SLT-side study of two-layer phase transitions \citep{Carroll21,FarrugiaRoberts22} and of functional-equivalence regions in tanh networks \citep{FarrugiaRoberts23}; we use it only to read off the per-layer rate exponent.

The canonical-aligned move at the dead unit $h_0$ steps both of its layers at once: the outgoing weight $(W_2)_{h_0} \mapsto t$ and the incoming row $(W_1)_{h_0,:} \mapsto t\,v$ for a fixed unit row $v \in \reals^d$. Write $\theta(t)$ for this path and $u$ for its unit tangent at $t = 0$, the dead direction. Because both weights vanish at the teacher, unit $h_0$'s output contribution $(W_2)_{h_0}\,\phi\big((W_1)_{h_0,:}\,x\big)$ picks up one factor of $t$ from each layer. Compute the loss along the path:
\[
L(\theta(t)) - L(\theta_0) \;=\; \tfrac12\, \expect_x\!\left[\big(f_{\theta(t)}(x) - f_{\theta_0}(x)\big)^2\right].
\]
For the linear case $\phi = \text{id}$, the difference is $f_{\theta(t)}(x) - f_{\theta_0}(x) = t^2\,(v \cdot x)$ to leading order. The squared loss therefore behaves as $\Theta(t^4)$, so the KL order is $k = 2$.

The directional Fisher follows. By Theorem~\ref{thm:fisher_decay}, the rate exponent is $2(k-1) = 2$, and direct computation confirms:
\[
u^\top \fisher(\theta(t)) u \;=\; \frac{\expect_x\!\left[\big(\partial_t f_{\theta(t)}(x)\big)^2\right]}{\sigma^2\, \|\theta'(t)\|_2^2} \;=\; \frac{4t^2\,\|v\|_2^2}{\sigma^2 \,(1 + \|v\|_2^2)} + O(t^3) \;=\; \frac{2t^2}{\sigma^2} + O(t^3),
\]
The exponent is $2$ at depth $L = 2$, matching the rate ladder $2(L - \ell)$ at $\ell = 1$ and $2(\ell - 1) = 0$ at $\ell = 2$ from the bridge theorem.

For $L = 3$, an analogous canonical-aligned move along three layers gives $K(\theta(t)) = \Theta(t^6)$ (KL order $k = 3$; the analytic output difference $f_{\theta(t)}(x) - f_{\theta_0}(x)$ is $\Theta(t^3)$, regime-independent), and the per-layer rates form the ladder $\lambda_{\min}(G_\ell) = \Theta(t^{2(L-\ell)}) = \Theta(t^4), \Theta(t^2), \Theta(t^0)$ for $\ell = 1, 2, 3$. The product $\lambda_{\min}(A_\ell)\,\lambda_{\min}(G_\ell) = \Theta(t^{2(L-1)}) = \Theta(t^4)$ holds at every layer.

For ReLU activation, the same canonical-aligned move produces the same rate ladder: the ReLU introduces a Bernoulli-like factor in the score expansion (the activation pattern of the dead unit is a function of $x$), but the leading $t^{2k}$ structure of $K$ is preserved, with the rate exponent unchanged. The score-expansion proof of Theorem~\ref{thm:fisher_decay} accommodates this directly: the leading non-vanishing $a_k(x)$ now depends on the activation pattern through a multiplicative factor that is in $L^2(p^*)$ under standard assumptions on the input distribution, but its presence does not change the exponent.

This example illustrates two features that carry through to deeper architectures. First, canonical alignment is the precondition that makes the per-layer rates separate cleanly: when the dead direction at every layer is the same coordinate, the K-FAC factorisation isolates the per-layer rate. Second, the rate ladder $2(L-\ell)$ is invariant under nonlinearities that admit the score expansion: ReLU, GELU, and linear all produce the same rate exponents at the same layers.
 
\section{The selection rule and RLCT recovery}
\label{sec:theory:selection}

Theorem~\ref{thm:fisher_decay} reads a rate exponent from a single dead direction. A deep network's Fisher has many directions, and the singular set has both tangent and normal components at any smooth point. Tangent directions carry gauge zeros (the metric vanishes identically along them); normal directions carry the rate primitive when they are dead. The question this section answers is: how does one tell the two apart from a measured spectrum, and what does the transversal rate compute? Under the genericity pair (G) and (G$^+$) on the analytic Taylor data at $\theta_0$, the Fisher spectrum splits cleanly (tangential and transversal eigenvalue groups become separable by their rates), and the transversal rate recovers Watanabe's single-direction contribution to the normal-slice RLCT.

The transversal rate exponent $\alpha_{\text{transv}}$ gives the directional KL order $\hat k = 1 + \alpha_{\text{transv}}/2$, and the single degenerate transversal direction contributes $1/(2\hat k)$ to the local RLCT, read in original parameter coordinates with no resolution of singularities. The recovery holds numerically on controlled testbeds. For a deep-linear reduced-rank regression with true KL order $k = 2$ (so $\lambda = 1/4$), the measured transversal slope $1.96 \pm 0.07$ returns $\hat\lambda = 0.253 \pm 0.004$; two- and three-component Gaussian mixtures recover the same $\lambda$ to within a few percent (Appendix~\ref{app:theory:parametric_validations}). Section~\ref{sec:theory:extensions} extends the recovery to Watanabe's multi-component normal-crossing form.

\label{app:proof:selection_rule}

This result extends Theorem~\ref{thm:fisher_decay} to smooth-fiber families. It is conditional in two ways that should be flagged at the outset. First, parts (b) and (c) require a spectral-genericity condition (G) under which the tangential and transversal eigenvalue groups have separable rates; (G) is open dense on the analytic Taylor data of $\log p$ at $\theta_0$ but is not unconditional. Second, the RLCT statement in part (c) recovers a \emph{single-direction contribution} $1/(2k)$ on a smooth-fiber singularity after tangential reduction. The general Watanabe normal-crossing form $K = u(g) \prod_i g_i^{2 k_i}$ with multiple $k_i \ge 1$ is the natural extension; the multi-component recovery $(k_1, \ldots, k_r)$ is established in\ifsupp{} the multi-component normal-crossing theorem\else{} Theorem~\ref{thm:multi_component_rates} (\S\ref{sec:theory:extensions})\fi, and the Jacobian exponents $h_i$ in Watanabe's full RLCT $\min_i (h_i + 1)/(2k_i)$ enter as a prior choice that an $\varepsilon$-scan volume observable resolves\ifsupp\else{} (Remark~\ref{rem:multi_component_h_i})\fi.

\begin{theorem}[Selection rule for smooth-fiber singularities; single-direction RLCT contribution under (G) and~(iv)]
\label{thm:selection_rule}
Let $S = \{\theta : p_\theta = p^*\}$ be a smooth submanifold of dimension $r_0$ near $\theta_0$, and let $\Pi \in \reals^{d \times (d - r_0)}$ be an orthonormal basis matrix of the normal space $N_{\theta_0} S$ (so $\Pi^\top \Pi = I$ and the columns of $\Pi$ span $N_{\theta_0} S$). Consider the analytic normal-bundle subfamily $\{q_{\theta_N} := p_{\theta_0 + \Pi \theta_N} : \theta_N \in \reals^{d - r_0}\}$ with Fisher $G_N(\theta_N) := \Pi^\top \fisher(\theta_0 + \Pi \theta_N) \Pi$, and suppose $n \in N_{\theta_0} S$ is a unit normal direction with KL order $k \ge 2$ along the curve $\theta(t) = \theta_0 + t n$. Define the directional log-likelihood coefficients (scalar functions of $x$)
\[
a_j^{(n')}(x) \;:=\; \tfrac{1}{j!}\,\partial_t^j \log p_{\theta_0 + tn'}(x)\big|_{t=0}, \qquad n' \in N_{\theta_0} S.
\]
Under the following assumptions:
(i*) on a punctured neighbourhood of $0$ in $N_{\theta_0} S$, the non-degenerate eigenvalues of $G_N(\theta_N)$ remain $\Theta(1)$;
(ii*) the leading coefficient $a_k^{(n)}$ does not lie in the $L^2(p^*)$-linear span of $\{a_1^{(n')} : n' \in N_{\theta_0} S, \, n' \perp n\}$, i.e., the first-order scores along normal directions orthogonal to $n$;
(iii) uniform integrability of Theorem~\ref{thm:fisher_decay}, transferred to the subfamily under the affine reparametrisation $\theta_N \mapsto \theta_0 + \Pi \theta_N$ (which preserves $L^p(p^*)$ moments since $p_{q_0} = p^*$);
(iv) (\emph{single transversal degeneracy}) the orthogonal complement $\{n' \in N_{\theta_0} S : n' \perp n\}$ has all directional KL orders equal to $1$, equivalently, the subfamily restricted to that complement has $G_N \succ 0$ at $\theta_N = 0$;
(G) (\emph{spectral genericity, used only for parts (b) and (c)}) the tangential exponents $\{2 j_a\}_a$ of Remark~\ref{rem:tangential_rates_sub} each differ from the transversal exponent $2(k-1)$, so the transversal eigenvalue group is separable from the tangential group on the spectrum;
(G$^+$) (\emph{graded score non-degeneracy, used only for parts (b) and (c)}) the leading-score system $\{c_{k,0,0}\} \cup \{c_{j_a, e_a, 0}\}_a \cup \{c_{0,0,e_{n'}}\}_{n'}$ of Lemma~\ref{lem:G_spectral_separability} is linearly independent in $L^2(p^*)$ after projecting each element orthogonal to the span of the leading scores of strictly lower $t$-order. Exponent separation alone does not imply this: a leading tangential score can coincide with a lower-order score after projection, in which case the corresponding tangential eigenvalue acquires the transversal exponent (Remark~\ref{rem:genericity_G_status} exhibits the failure). (G$^+$) is open dense on the analytic Taylor data, the same status as (G).
We have
\begin{enumerate}
\item[(a)] $\lambda_{\min}(G_N(t)) = \Theta(t^{2(k-1)})$ (transversal rate);
\item[(b)] (under (G)) tangential eigenvalues of the full Fisher (those with eigenvectors converging to $T_{\theta_0} S$) decay at model-dependent rates $\Theta(t^{2 j_a})$ determined by the lowest-order non-vanishing mixed derivative $\partial_t^{j_a} \partial_a \log p_{\theta(t)}$ for each tangential coordinate $\partial_a$ (Remark~\ref{rem:tangential_rates_sub} below);
\item[(c)] (under (G) and (G$^+$)) the transversal group is uniquely identified by its exponent $2(k-1)$, and $\hat k = 1 + \alpha_{\mathrm{transv}} / 2$ recovers the KL order. Under hypothesis~(iv), the \emph{normal-slice} RLCT is exact: the normal-bundle subfamily $\{q_{\theta_N}\}$ has local RLCT $1/(2k) + m_\perp/2$ at $\theta_N = 0$, computed via the standard $\int |K|^{-s}$ factorisation \citep{Watanabe09} on the slice normal form $u^{2k} + v_1^2 + \cdots + v_{m_\perp}^2$ ($m_\perp = d - r_0 - 1$ non-degenerate-quadratic complements, written $m_\perp$ to keep it clear of Watanabe's multiplicity $m$; $m_\perp = 0$ gives the single-direction contribution $1/(2k)$). \emph{Scope note (slice versus full family).} The clause above is a statement about the normal-bundle subfamily, which is the object the exponent estimator reads. The local RLCT of the \emph{full} family at $\theta_0$ can strictly exceed the slice value when the KL order along $n$ is not constant on the fibre (tangential-normal coupling): Remark~\ref{rem:slice_vs_full_rlct} gives an analytic two-parameter family satisfying every hypothesis of this theorem whose slice RLCT is $1/6$ while its full local RLCT is $1/2$ with log-multiplicity $2$. Recovering the full local RLCT requires either constancy of the transversal KL order along the fibre with an analytic product splitting, or a resolution of the fibre-adapted normal form. The general normal-crossing case $K = u(g) \prod_i g_i^{2 k_i}$ with multiple $k_i \ge 1$ is the natural extension: multi-component recovery $(k_1, \ldots, k_r)$ is established in\ifsupp{} the multi-component normal-crossing theorem\else{} Theorem~\ref{thm:multi_component_rates} (\S\ref{sec:theory:multi_component})\fi, with the Jacobian exponents $h_i$ entering as a prior choice.
\end{enumerate}
\end{theorem}

\begin{figure}[t]
\centering
\includegraphics[width=\textwidth]{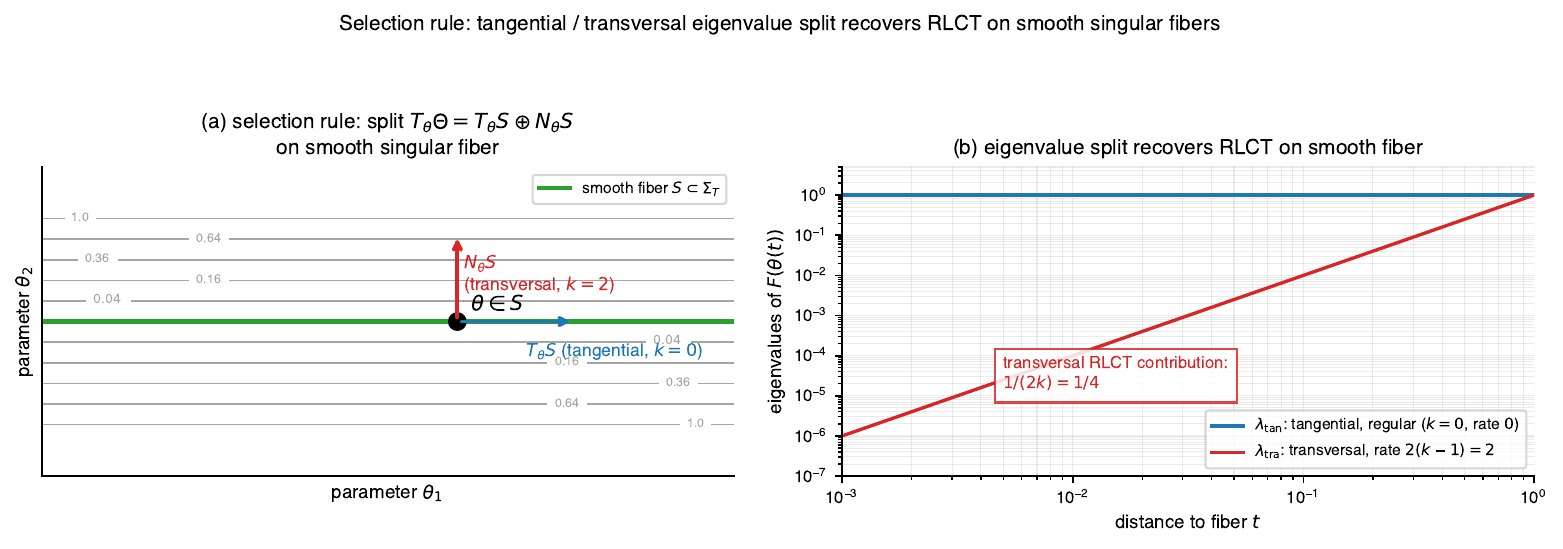}
\caption{Selection rule on a smooth singular fiber (Theorem~\ref{thm:selection_rule}). (a)~The geometry the rule classifies: at a point $\theta \in S$ on a smooth fiber $S \subset \Sigma_T$, the tangent space splits as $T_\theta \Theta = T_\theta S \oplus N_\theta S$. Population KL stays zero along $T_\theta S$ (the fibre) and has order $k = 2$ along the transversal normal direction in $N_\theta S$. (b)~The split as it appears in the Fisher spectrum along an approach $\theta(t)$: normal-complement eigenvalues stay $\Theta(1)$, tangential eigenvalues decay at the model-dependent rates $\Theta(t^{2 j_a})$, and the transversal eigenvalue, the rate-carrying direction, decays at $\Theta(t^{2(k-1)}) = \Theta(t^2)$. The transversal eigenvalue's exponent recovers $\hat k = 1 + \alpha_\mathrm{transv}/2 = 2$, contributing $1/(2k) = 1/4$ to the local RLCT.}
\label{fig:selection_rule_smooth_fiber}
\end{figure}

\paragraph{Three readings of the selection rule.}\label{rem:selection_rule_three_readings}
The same content takes three useful forms. \emph{As a spectral classification.} Fit a rate exponent $\alpha_i$ to each Fisher eigenvalue of $\fisher(\theta(t))$ along the approach. Three groups appear, and their exponents tell them apart under (G): $\alpha_i \approx 0$ marks the non-degenerate normal complement, $\alpha_i \approx 2 j_a$ with $j_a \ge 1$ marks a tangential direction, and $\alpha_i \approx 2(k-1)$ marks the transversal direction that carries the rate. Tangential exponents are strictly positive because $j_a \ge 1$ by construction, so a fitted exponent near zero identifies a non-degenerate direction and never a tangential one. Directions along the fibre itself split into two cases. When the family is constant along the whole line $\theta(t) + s v$ near the approach (exact gauge directions and coordinate-aligned fibres, the case of every worked example in this paper), the Fisher entries vanish identically at every $t$ and a slope fit reads the numerical floor instead of a rate. When the fibre is curved, its tangent line leaves $S$ at second order, the tangent direction is itself a rated direction, and its eigenvalue joins the tangential group at the corresponding $2 j_a$. \emph{As an RLCT estimator.} The transversal exponent $\alpha_{\mathrm{transv}}$ recovers $\hat k = 1 + \alpha_{\mathrm{transv}}/2$, and the contribution of that one direction to the local RLCT is $\hat\lambda = 1/(2\hat k)$. With $h_i = 0$ (uniform prior in original parameter coordinates), this is Watanabe's single-direction contribution. \emph{As a basin picture.} Figure~\ref{fig:selection_rule_smooth_fiber} shows the smooth fibre $S \subset \Sigma_T$ at a point $\theta$ with tangent space $T_\theta S$ and normal space $N_\theta S$ split apart. Eigenvalues supported on $T_\theta S$ decay at the tangential rates $\Theta(t^{2 j_a})$, the ones on the non-degenerate normal complement flat-line at $\Theta(1)$, and the eigenvalue on $N_\theta S$ transversal to the fibre decays at $\Theta(t^{2(k-1)})$ and carries the rate.

\begin{remark}[Status of the genericity assumptions]
\label{rem:genericity_G_status}
Part~(a) is unconditional given (i*)--(iv); dropping (iv) breaks it, since a second degenerate normal direction sends $\lambda_{\min}(G_N)$ to zero identically ($\mu = (u^3, v^2)$ satisfies (i*)--(iii) with (i*) vacuous and has $\lambda_{\min}(G_N(tn)) \equiv 0$ at $n = \partial_u$). Parts (b) and (c) require both the spectral genericity (G) and the graded score non-degeneracy (G$^+$). On the analytic Taylor data of $\log p$ at $\theta_0$, (G) is the open dense condition that no tangential exponent $2 j_a$ accidentally equals the transversal exponent $2(k-1)$, and (G$^+$) is the open dense condition that the leading scores stay linearly independent after graded projection. Lemma~\ref{lem:G_spectral_separability} below proves that under the pair the Fisher's eigenvalue spectrum splits into transversal, tangential, and non-degenerate-normal groups by polynomial spectral gaps, with eigenvectors converging to the corresponding coordinate basis vectors: exactly the structure parts (b) and (c) require. (G) alone does not suffice: in the family $\mu(u, a, s) = (s + ua,\, u^3,\, \varepsilon u^2 a)$ the exponent-separation condition holds, yet the leading tangential score coincides with a normal-complement score after projection, the Schur correction cancels the tangential leading term, and the tangential eigenvalue acquires the transversal exponent (measured spectrum exponents $\{4, 4, 0\}$ against the separable prediction $\{4, 2, 0\}$). (G$^+$) excludes exactly this coincidence; it is the graded formalisation of the informal requirement that $\alpha_a > 0$ survive the Schur subtraction. Both worked examples below and the gauge-orbit applications satisfy (G$^+$).

Open density is the wrong comfort here: symmetry defeats (G$^+$) systematically, because at a Gaussian base point every directional score expands in the Hermite basis of $p^*$ and the leading components compete for the low-order slots. The paper's own $2$-component mixture at the merge is the worked instance, read in the frame of the dead direction, the mean shift $\partial_{\bar\mu}$, the common-variance direction $\partial_{\sigma^2}$, and the fixed-means weight coordinate $\partial_w$. At every weight $w$, exactly: $s_{\bar\mu} = \mathrm{He}_1$, $s_{\sigma^2} = \mathrm{He}_2/2$, the dead direction's leading score is $4w(1-w)\,\mathrm{He}_2\,\delta$, and the fixed-means weight score's is $-2\,\mathrm{He}_1\,\delta$, so each pair aligns completely in the limit (cosines $+1$ and $-1$). The alignment defeats more than (G$^+$): the dead leading coefficient is a multiple of the first-order variance score, violating hypothesis~(ii*), so part~(a)'s normal-bundle read fails as well ($\lambda_{\min}(G_N) = \Theta(\delta^4)$ against the transversal $\delta^2$). The collision repeats one level up: after projecting off the $\Theta(1)$ block, the dead residual's leading term is $4w(1-w)(2w-1)\,\mathrm{He}_3\,\delta^2$ and the weight residual's is $-\tfrac{4}{3}(6w^2 - 6w + 1)\,\mathrm{He}_3\,\delta^3$, both proportional to $\mathrm{He}_3$ at every $w$. Where both coefficients are non-zero, the parallel residuals cancel the leading term of the residual Gram determinant and defer the smallest eigenvalue two further orders: the spectrum grades as $\{\Theta(1), \Theta(1), \Theta(\delta^4), \Theta(\delta^8)\}$ (measured eigenvalue slopes $+4.0000$ and $+8.0000$ at the probe weight $w = 0.35$), with no eigenvalue at the transversal exponent $2(k-1) = 2$. At the symmetric merge $w = 1/2$ the dead-residual coefficient vanishes, the collision is absent, and the grading is $\{\Theta(1), \Theta(1), \Theta(\delta^6), \Theta(\delta^6)\}$, still with no transversal eigenvalue. Three consequences follow. The directional form is alignment-blind: $u^\top \fisher u$ along the dead direction is a diagonal entry and keeps $\Theta(\delta^2)$, exactly $2(4w(1-w))^2\delta^2$ at leading order, so directional probes read the KL order where the spectrum cannot. The operational protocol's no-match fallback (\S\ref{app:selection_rule_op}) fires correctly on such families, and the leading-score Gram, read directly, shows the aligned scores that cause it. And restricting to an alignment-free subfamily restores (G$^+$): the mixture selection-rule validations run on the means-only slice $(\mu_1, \mu_2)$, whose two-parameter spectrum grades cleanly as $\{2, 0\}$ (Table~\ref{tab:selection_rule_app}).
\end{remark}

\begin{remark}[Tangential eigenvalue rates]
\label{rem:tangential_rates_sub}
For a tangential coordinate $\partial_a$ with $\partial_a \log p|_{\theta_0} \equiv 0$ pointwise (since the family is constant along $S$), the tangential score expands in $t$ as $s_a(\theta(t); x) = c_{j_a, 1}(x) t^{j_a} + O(t^{j_a + 1})$ where $c_{j_a, 1}(x) := (1/j_a!) \partial_t^{j_a} \partial_a \log p|_{\theta_0}(x)$ and $j_a := \min\{j \ge 1 : \partial_t^j \partial_a \log p \not\equiv 0 \text{ in } L^2(p^*)\}$, defined when that set is non-empty. When every mixed derivative vanishes (the family is constant in $\partial_a$ on a neighbourhood, the exact gauge case), $\fisher_{aa}$ is identically zero along the approach and the direction belongs to the floor-excluded group of the selection rule's three readings (\S\ref{rem:selection_rule_three_readings}), with no $j_a$ to fit. The corresponding diagonal Fisher entry is
\[
\fisher_{aa}(\theta(t)) \;=\; \mathbb{E}_{p_{\theta(t)}}[s_a^2] \;=\; \alpha_a \, t^{2 j_a} + O(t^{2 j_a + 1}), \qquad \alpha_a := \mathbb{E}_{p^*}[c_{j_a, 1}^2] > 0.
\]
Generically, $j_a$ is an integer determined by the local Taylor data of $\log p$ along $n$, and $\{2 j_a\}_a$ is disjoint from $\{2(k-1)\}$: this is the spectral genericity condition (G) used in parts~(b) and~(c) of the theorem. The full tangential eigenvalue (after Schur correction from cross-couplings) inherits the same exponent $2 j_a$ under (G) together with (G$^+$), as established in Lemma~\ref{lem:G_spectral_separability} below; without (G$^+$) the Schur correction can cancel the leading term $\alpha_a t^{2 j_a}$ outright.
\end{remark}

\paragraph{Worked examples of the tangential rate.}\label{rem:tangential_rates_examples}
Two families illustrate the tangential rate $\Theta(t^{2 j_a})$ explicitly.

\emph{Example 1 ($j_a = 2$, transversal $k = 2$).} Take $p_{u, a}(y \mid x) = \mathcal{N}(y \mid u^2 (a^* + a) x, 1)$ with $x \sim \mathcal{N}(0, 1)$ and $a^* > 0$ a fixed constant; locally near $\theta_0 = (0, 0)$ the fiber is $\{u = 0\}$, transversal $u$, tangential $a$. Then $K(u, 0) = \tfrac{1}{2} u^4 (a^*)^2$ gives transversal KL order $k = 2$, Fisher $\Theta(t^2)$. The tangential score at $\theta(t) = (t, 0)$ is $s_a(\theta(t); y, x) = (y - t^2 a^* x) \cdot t^2 x = t^2 xy - t^4 a^* x^2$, with leading order $t^2$ so $j_a = 2$. The diagonal Fisher entry is $\fisher_{aa}(\theta(t)) = \mathbb{E}_x[t^4 x^2 \cdot \mathbb{E}_{y|x}[(y - t^2 a^* x)^2]] = t^4 \mathbb{E}[x^2] = t^4 = \Theta(t^{2 j_a})$.

\emph{Example 2 ($j_a = 1$, transversal $k = 3$).} Take $p_{u, a}(y \mid x) = \mathcal{N}(y \mid (u^3 + c\,u\,a) x, 1)$ with $x \sim \mathcal{N}(0, 1)$ and $c \ne 0$. The exact-model set through $\theta_0 = (0,0)$ is the node $\{u\,(u^2 + c a) = 0\}$, so this family does not itself satisfy the smooth-fibre setup; we use it only to illustrate the tangential-rate computation along the branch $\{u = 0\}$, which involves nothing beyond the curve $\theta(t) = (t, 0)$ and its scores. Then $K(u, 0) = \tfrac{1}{2} u^6$ gives $k = 3$ (transversal Fisher $\Theta(t^4)$), and the tangential score at $\theta(t) = (t, 0)$ is $s_a(\theta(t); y, x) = (y - t^3 x) \cdot c\, t \cdot x = c\, t\, xy - c\, t^4 x^2$, with leading order $t^1$ so $j_a = 1$. The diagonal Fisher entry is $\fisher_{aa}(\theta(t)) = c^2 t^2 \mathbb{E}[x^2] = c^2 t^2 = \Theta(t^{2 j_a})$. The cross-Fisher is $\fisher_{ua}(t) = 3 c\, t^3 = O(t^{(k-1) + j_a})$, matching the cross-block prediction in Lemma~\ref{lem:G_spectral_separability}. Genericity (G) holds ($2 j_a = 2 \ne 4 = 2(k-1)$), and the lemma's eigenvector convergence rate is $O(t^{|2 j_a - 2(k-1)|/2}) = O(t)$.

\begin{lemma}[Spectral separability under (G)]
\label{lem:G_spectral_separability}
Under assumptions (i*)--(iv), (G), and (G$^+$) of Theorem~\ref{thm:selection_rule}, the eigenvalues of $\fisher(\theta(t))$ as $t \to 0$ split into three groups:
\begin{enumerate}\itemsep=0pt
\item One \emph{transversal} eigenvalue of order $\Theta(t^{2(k-1)})$, with eigenvector converging to $n$.
\item $r_0$ \emph{tangential} eigenvalues; the eigenvalue indexed by $a$ is $\Theta(t^{2 j_a})$, with eigenvector converging to the coordinate axis $\partial_a$ (up to within-tangential mixing if some $j_a$ coincide).
\item $d - r_0 - 1$ \emph{normal-complement} eigenvalues at $\Theta(1)$, with eigenvectors spanning $N_{\theta_0}S \setminus \mathrm{span}(n)$.
\end{enumerate}
Eigenvector angular convergence: $\sphericalangle(\hat v_n(t), n) = O(t^{|2 j_a^* - 2(k-1)|/2})$ where $j_a^*$ is the tangential exponent closest to $k - 1$; analogously $\sphericalangle(\hat v_a(t), \partial_a) = O(t^{|2 j_a - 2(k-1)|/2})$.
\end{lemma}

\begin{proof}[Proof of Lemma~\ref{lem:G_spectral_separability}]
Adopt adapted coordinates $(n, T, N')$ at $\theta_0$ where $n$ is the chosen transversal direction, $T = T_{\theta_0}S$ has basis $\{\partial_a\}_{a = 1, \ldots, r_0}$, and $N' \subset N_{\theta_0}S \setminus \mathrm{span}(n)$ is the non-degenerate normal complement (basis $\{\partial_{n'}\}_{n' \in N'}$).

\emph{Step 1 (block leading orders).} The Taylor expansion of $\log p$ around $\theta_0$ in coordinates $(t \cdot n + a^i \partial_{a^i} + s^j \partial_{n'_j})$ yields the score Taylor coefficients $c_{i, j, l}(x) = (1/i! j! l!) \partial_t^i \partial_{a^*}^j \partial_{n'^*}^l \log p|_{\theta_0}(x)$ for the appropriate multi-indices. By the rate-theorem inductive identity (proof of Theorem~\ref{thm:fisher_decay} applied to the normal-bundle subfamily), $c_{1, 0, 0} = \cdots = c_{k-1, 0, 0} = 0$ in $L^2(p^*)$ and $c_{k, 0, 0}$ is non-trivial in $L^2(p^*)$ modulo non-degenerate scores (assumption (ii*)). By tangentiality of $\partial_a$, $c_{0, e_a, 0} = 0$ pointwise for each $e_a$. By assumption (iv), the directional first-order coefficients of $\log p$ along $N'$ are non-trivial in $L^2(p^*)$.

The score components at $\theta(t) = \theta_0 + tn$:
\begin{align*}
s_n(\theta(t); x) &= \sum_{i \ge 1} i \cdot c_{i, 0, 0}(x) t^{i-1} = k \cdot c_{k, 0, 0}(x) \cdot t^{k-1} + O(t^k), \\
s_a(\theta(t); x) &= \sum_{i \ge 1} c_{i, e_a, 0}(x) t^i = c_{j_a, e_a, 0}(x) t^{j_a} + O(t^{j_a + 1}), \\
s_{n'}(\theta(t); x) &= \sum_{i \ge 0} c_{i, 0, e_{n'}}(x) t^i = c_{0, 0, e_{n'}}(x) + O(t).
\end{align*}
The Fisher block leading orders follow by squaring and applying $\mathbb{E}_{p_{\theta(t)}} = \mathbb{E}_{p^*}(1 + O(t^k))$:
\begin{align*}
F_{nn}(t) &= \alpha_n t^{2(k-1)}(1 + O(t)), \quad \alpha_n := k^2 \mathbb{E}_{p^*}[c_{k, 0, 0}^2] > 0, \\
F_{aa}(t) &= \alpha_a t^{2 j_a}(1 + O(t)), \quad \alpha_a := \mathbb{E}_{p^*}[c_{j_a, e_a, 0}^2] > 0, \\
F_{N'N'}(t) &= G_{N'} + O(t), \quad G_{N'} \succ 0 \text{ at } \Theta(1), \\
F_{na}(t) &= O(t^{(k-1) + j_a}), \quad
F_{nN'}(t) = O(t^{k-1}), \\
F_{aN'}(t) &= O(t^{j_a}), \quad
F_{ab}(t) = O(t^{j_a + j_b}) \text{ for } a \ne b.
\end{align*}

\emph{Step 2 (Schur-reduce $N'$).} Since $F_{N'N'}(t) \succ 0$ uniformly at $\Theta(1)$ on a neighbourhood of $t = 0$, the eigenvalues of $\fisher$ corresponding to eigenvectors orthogonal to $N'$ are the eigenvalues of the reduced $(1 + r_0) \times (1 + r_0)$ matrix
\[
\widetilde F(t) := F_{\{n\} \cup T,\,\{n\} \cup T}(t) - F_{\{n\} \cup T,\,N'}(t) F_{N'N'}(t)^{-1} F_{N',\,\{n\} \cup T}(t).
\]
The Schur correction to $F_{nn}$ is $O(t^{k-1}) \cdot O(1) \cdot O(t^{k-1}) = O(t^{2(k-1)})$, same order as $F_{nn}$, modifying the leading constant by at most a Cauchy--Schwarz-bounded amount strictly less than $\alpha_n$ (assumption (ii*): $c_{k, 0, 0}$ has non-trivial component orthogonal to the $N'$-scores). The Schur correction to $F_{aa}$ is $O(t^{j_a}) \cdot O(1) \cdot O(t^{j_a}) = O(t^{2 j_a})$, similarly modifying the leading constant; the auxiliary non-degeneracy that the corrected leading constant remains positive is the Cauchy--Schwarz-non-saturation built into the hypothesis on $\alpha_a > 0$ in Remark~\ref{rem:tangential_rates_sub}. The Schur correction to $F_{na}$ is $O(t^{k-1}) \cdot O(t^{j_a}) = O(t^{(k-1) + j_a})$, same order as the direct entry.

The $N'$-block contributes $d - r_0 - 1$ eigenvalues of $\fisher$ at $\Theta(1)$, with eigenvectors converging to a basis of $N'$ (these match the spectral content of $G_{N'}$ at $\Theta(1)$, perturbed by $O(t)$).

\emph{Step 3 (pairwise spectral perturbation $(n, a)$).} Consider the $2 \times 2$ submatrix
\[
M_a(t) = \begin{pmatrix} \widetilde F_{nn}(t) & \widetilde F_{na}(t) \\ \widetilde F_{an}(t) & \widetilde F_{aa}(t) \end{pmatrix} = \begin{pmatrix} \alpha_n t^{2(k-1)} & O(t^{(k-1) + j_a}) \\ O(t^{(k-1) + j_a}) & \alpha_a t^{2 j_a} \end{pmatrix} (1 + O(t)).
\]
Its trace and determinant:
\[
\mathrm{tr}(M_a) = \alpha_n t^{2(k-1)} + \alpha_a t^{2 j_a} + O(t^{2(k-1)+1}), \quad
\det(M_a) = \alpha_n \alpha_a t^{2(k-1) + 2 j_a} + O(t^{2(k-1) + 2 j_a + 1}),
\]
where the determinant's leading constant is strictly positive: by Cauchy--Schwarz, $|\widetilde F_{na}|^2 \le \alpha_n \alpha_a t^{2(k-1) + 2 j_a}$ with equality iff $c_{k, 0, 0}$ and $c_{j_a, e_a, 0}$ are linearly dependent in $L^2(p^*)$; assumption (G$^+$) excludes this dependence, and likewise keeps the Schur-corrected leading constants of Step~2 positive (dependence of a leading score on lower-order scores is exactly what (G$^+$) rules out).

Under (G), $2 j_a \ne 2(k-1)$. WLOG assume $2(k-1) < 2 j_a$ (the other case is symmetric). Then $\mathrm{tr} \sim \alpha_n t^{2(k-1)}$ dominates and $4 \det \sim 4 \alpha_n \alpha_a t^{2(k-1) + 2 j_a} = o(\mathrm{tr}^2)$. Expanding $\sqrt{\mathrm{tr}^2 - 4 \det} = \mathrm{tr}(1 - 2 \det / \mathrm{tr}^2 + O((\det/\mathrm{tr}^2)^2))$:
\begin{align*}
\lambda_+(M_a) &= \mathrm{tr} - \det/\mathrm{tr} + O(\det^2 / \mathrm{tr}^3) = \alpha_n t^{2(k-1)} \bigl(1 + O(t^{2 j_a - 2(k-1)})\bigr), \\
\lambda_-(M_a) &= \det/\mathrm{tr} + O(\det^2 / \mathrm{tr}^3) = \alpha_a t^{2 j_a} \bigl(1 + O(t^{2 j_a - 2(k-1)})\bigr).
\end{align*}

Both eigenvalues retain their leading constants and exponents; the smaller eigenvalue is $\Theta(t^{2 j_a})$ and the larger is $\Theta(t^{2(k-1)})$.

\emph{Step 4 (eigenvector convergence, Davis--Kahan).} The angular deviation of $\hat v_n(t)$ from $e_n$ in the $(e_n, e_a)$ subspace satisfies $\tan(2\phi) = 2 \widetilde F_{na} / (\widetilde F_{nn} - \widetilde F_{aa})$, so $\phi \asymp |\widetilde F_{na}| / |\widetilde F_{nn} - \widetilde F_{aa}|$. The numerator is $O(t^{(k-1) + j_a})$, the denominator $\Theta(t^{\min(2(k-1), 2 j_a)})$, giving $\phi = O(t^{(k-1) + j_a - \min(2(k-1), 2 j_a)}) = O(t^{|j_a - (k-1)|}) = O(t^{|2 j_a - 2(k-1)|/2})$, vanishing as $t \to 0$ under (G). The same holds for $\hat v_a(t) \to e_a$.

\emph{Step 5 (multi-direction extension).} The pairwise argument extends to all $r_0$ tangential directions. The cross-tangential off-diagonals $\widetilde F_{ab} = O(t^{j_a + j_b})$ are bounded above by the geometric mean of the diagonal entries, and (G$^+$) applied to pairs of tangential leading scores keeps the tangential-block determinants non-degenerate at leading order, so spectral perturbation inside the tangential block gives $r_0$ eigenvalues of orders $\{2 j_a\}_a$ (with within-group degeneracy collapsing to a multi-eigenvalue cluster if some $j_a$ coincide; under (G$^+$) coincident orders mix eigenvectors within the group without displacing any tangential eigenvalue onto the transversal exponent). Coupling to the transversal eigenvalue is bounded by the worst-case pairwise rotation $O(t^{|2 j_{a^*} - 2(k-1)|/2})$ where $j_{a^*}$ minimises $|j_a - (k-1)|$; the transversal eigenvector converges to $e_n$ at this rate.

\emph{Conclusion.} The Fisher's spectrum splits as claimed; the transversal group is uniquely identified by its exponent $2(k-1)$ (distinct from each $2 j_a$ by (G), and distinct from $0$ since $k \ge 2$); the estimator $\hat k = 1 + \alpha_{\mathrm{transv}} / 2$ recovers $k$ from the spectrum.
\end{proof}

\begin{proof}[Proof of Theorem~\ref{thm:selection_rule}]
\emph{(a)} The subfamily $\{q_{\theta_N}\}$ is analytic because analyticity is preserved under the \emph{affine} embedding $\theta_N \mapsto \theta_0 + \Pi \theta_N$ (composition of an analytic family with an affine map is analytic; smoothness alone would not suffice). Along $n_N := \Pi^\top n \in \reals^{d - r_0}$, the KL function $\bar K(\theta_N) := K(\theta_0 + \Pi \theta_N)$ satisfies $\bar K(t n_N) = K(\theta_0 + t \Pi n_N) = K(\theta_0 + tn)$, so $\bar K$ has the same Taylor coefficients in $t$ as $K$ along the curve $n$ (KL order is a property of the curve $\theta(t)$, independent of how it is parameterised); hence $\bar K(t n_N) = c\,t^{2k} + O(t^{2k+1})$. Inner products $\langle a_j^{(n)}, a_l^{(n')} \rangle_{L^2(p^*)}$ are invariant under the parameterisation change since $p_{q_0} = p^*$. The score expansion of \S\ref{app:proof:fisher_decay} applies to the subfamily with directional coefficients $a_j^{(n)}$: inductively, $a_j^{(n)} = 0$ in $L^2(p^*)$ for $j < k$, and $\expect_{p^*}[(a_k^{(n)})^2] = 2c > 0$. Assumptions (ii*) and (iii) transfer Theorem~\ref{thm:fisher_decay}'s assumptions (ii) and (iii) to the subfamily $\{q_{\theta_N}\}$ along $n_N$; its assumption (i) (the complement block positive-definite at $t = 0$ and $\Theta(1)$) is supplied by (iv) together with (i*). With all of (i)--(iii) in place on the subfamily, $\lambda_{\min}(G_N(t)) = \Theta(t^{2(k-1)})$.

\emph{(b)} The Fisher null space at $\theta_0$ contains $T_{\theta_0} S$ (by the score characterisation $\ker \fisher(\theta_0) \supseteq T_{\theta_0} S$, since $p_\theta$ is constant along $S$). Lemma~\ref{lem:G_spectral_separability} applied under (G) gives the explicit tangential rate: the eigenvalue corresponding to tangential coordinate $\partial_a$ is $\Theta(t^{2 j_a})$, with $j_a$ as in Remark~\ref{rem:tangential_rates_sub}, and the eigenvector converges to $\partial_a$ as $t \to 0$.

\emph{(c)} Part (a) gives a transversal eigenvalue group with exponent $2(k-1)$. Under assumptions (G) and (G$^+$), Lemma~\ref{lem:G_spectral_separability} establishes that the transversal eigenvalue is uniquely identified by its exponent (distinct from each tangential exponent $2 j_a$ and from the $\Theta(1)$ normal-complement eigenvalues), with the transversal eigenvector $\hat v_n(t)$ converging to $n$ at rate $O(t^{|2 j_{a^*} - 2(k-1)|/2})$. The estimator $\hat k = 1 + \alpha_{\mathrm{transv}} / 2$ therefore recovers the KL order. For the RLCT clause, work on the normal slice: the subfamily $\{q_{\theta_N}\}$ has exact-model set locally $\{0\}$ (the fibre $S$ meets the slice transversally), the direction $n_N$ carries KL order $k$, and by hypothesis~(iv) the restriction of $\bar K$ to the orthogonal complement of $n_N$ has non-degenerate positive Hessian at $0$. The analytic splitting lemma (Morse lemma with parameters, applied on the slice where the Hessian in the complement variables is uniformly non-degenerate) gives an analytic change of coordinates \emph{on the slice} with $\bar K = u^{2k}\,(\mathrm{unit}) + v_1^2 + \cdots + v_{m_\perp}^2$, and the standard $\int |\bar K|^{-s}$ factorisation \citep{Watanabe09} yields the slice RLCT $1/(2k) + m_\perp/2$; $m_\perp = 0$ gives $1/(2k)$. The splitting is available only on the slice: on a full neighbourhood of $\theta_0$ in $\Theta$ the tangential variables can couple to $u$ at low order and no analytic change of coordinates produces the product form in general (Remark~\ref{rem:slice_vs_full_rlct}), which is why the clause is stated for the subfamily.
\end{proof}

\begin{remark}[Slice versus full-family RLCT; an exact counterexample]
\label{rem:slice_vs_full_rlct}
The RLCT clause of part~(c) is exact for the normal-bundle subfamily and can strictly undercount the full family. Take the Gaussian mean-map family $p_{u,a}(y) = \mathcal{N}(y \mid (ua,\, u^3), I_2)$, so $K(u, a) = \tfrac12\,(u^2 a^2 + u^6)$. The exact-model set is $S = \{u = 0\}$, a smooth one-dimensional fibre; at $\theta_0 = (0,0)$ the transversal direction $n = \partial_u$ has KL order $k = 3$; hypotheses (i*)--(iv), (G), and (G$^+$) all hold ($m_\perp = 0$, no non-degenerate normal complement). The slice family is $\bar K(u) = u^6/2$ with slice RLCT $1/(2k) = 1/6$, and the spectral conclusions of parts (a) and (b) hold exactly ($\fisher = \mathrm{diag}(9t^4 + a\text{-terms},\, t^2)$ along the approach). The \emph{full} local RLCT at $\theta_0$ is $1/2$ with log-multiplicity $2$: blowing up the origin, the chart $a = ub$ followed by $b = uc$ gives integrand $|u|^{2 - 6s}(1 + c^2)^{-s}$ with a simple pole at $s = 1/2$, and the chart $u = av$ gives $|a|^{1 - 4s}|v|^{-2s}$ with a double pole at $s = 1/2$; equivalently $\mathrm{vol}\{K < \varepsilon\} \asymp \varepsilon^{1/2}\log(1/\varepsilon)$. The mechanism is tangential-normal coupling: the KL order along $n$ is $3$ at $\theta_0$ but drops to $1$ at every nearby fibre point $(0, a_0)$ with $a_0 \ne 0$, and the resolved $a$-divisor carries a non-zero Jacobian exponent, so the smooth-fibre hypothesis alone does not force the product normal form on a full neighbourhood. The exponent estimator of part~(c) is unaffected: it reads the transversal rate at $\theta_0$, which is a slice quantity.
\end{remark}

\begin{remark}[Operational uses of the transversal/tangential split]
\label{rem:tangential_operational_uses}
Beyond RLCT recovery, the transversal/tangential decomposition has several operational consequences that we record here for completeness.
\begin{itemize}\itemsep=0pt
\item \emph{Gauge-vs-true-low-curvature separation.} The protocol that reports $\lambda_{p-1}(F_h^{\mathrm{pop}})$ rather than raw $\lambda_{\min}$ for $p$-class cross-entropy on hidden-layer projections is exactly the selection rule applied to the gauge-orbit smooth fiber: gauge zeros are tangential along the orbit; the smallest non-zero transversal carries the rate. The selection rule promotes this from a per-loss patch to a generic protocol: fit a power-law slope to each top-$K$ eigenvalue versus $\sigma_{\min}$ and read the exponent. A slope near $2(k-1)$ marks the transversal; a positive slope away from it marks a tangential direction at $2 j_a$; a slope near zero marks the non-degenerate complement. Exact gauge directions along the orbit stay at the numerical floor with no rate to fit, which is why the protocol excludes floor-trapped ranks before classifying.
\item \emph{LoRA placement.} Inserting a low-rank update inside the tangential subspace of the current task's singular fiber adds capacity without inducing transversal motion that would increase distortion at leading order. Validating this reduces to applying the selection rule at a base-model layer and checking that LoRA's measured rate falls in the tangential cluster.
\item \emph{Continual learning safe directions.} A direction tangential to the current task's fiber but transversal to the next task's is, by the selection rule's classification, the principled definition of a safe direction: one that can carry new-task gradient without disturbing the old-task minimum at leading order.
\item \emph{Rank-collapse early warning.} A previously-tangential direction transitioning to a transversal rate during training is the rate-language statement of the new dead direction emerging: the selection rule's exponent classification is what makes the transition detectable on the spectrum rather than only on the residual-stream $\sigma_{\min}$.
\end{itemize}
The post-final-LN kernel direction $\gamma^{-1} / \|\gamma^{-1}\|$ of Proposition~\ref{prop:ln_kernel} belongs to the kernel of $\mathrm{cov}(\mathrm{LN}(X))$ for any input distribution: an exact zero at every checkpoint, on the activation-side factor. The operational protocol of \S\ref{app:selection_rule_op} classifies $G_\ell$ spectra, where this zero does not appear. A pipeline that also classifies the activation-side covariance meets it there and must exclude it as a floor-trapped exact zero before the rate matching runs. The LN kernel and the residual-stream $\sigma_{\min}$ depth-invariance of Corollary~\ref{cor:sigma-min-res} are the two structures the observables read from the architecture alone, with no rate fit; both fall on the exact-zero side of the split between exact zeros and rated eigenvalues.
\end{remark}

\begin{remark}[Extension beyond smooth fibers]
\label{rem:singular_fibers_sub}
The smooth-fiber assumption covers a wide range of parametric families relevant to deep learning. It admits the following specialisations and extensions:
\begin{itemize}\itemsep=0pt
\item \emph{Isolated singularities} ($\dim S = 0$, i.e., $r_0 = 0$, $\Pi = I_d$) are the special case of Theorem~\ref{thm:selection_rule} where the normal-bundle subfamily \emph{is} the full family and there are no tangential directions. The theorem applies directly without modification.
\item \emph{Stratified singular sets} where multiple coordinate-aligned conditions hold simultaneously (multi-component mixtures with several pairwise collapses, rank-stratified reduced-rank regression) can be handled by piecewise application of the theorem to each stratum, provided the strata satisfy a transverse-intersection (Whitney-style) regularity condition so that the local product structure $u^{2k} + v_1^2 + \cdots$ holds on each stratum's normal bundle. Establishing this transversality is model-specific.
\item \emph{Genuinely non-stratified singularities} (cuspidal or Whitney-umbrella loci) require Hironaka resolution \citep{Hironaka64} of $K$ to a normal-crossing form before the local-RLCT computation applies; this is left as future work. The singularity patterns encountered in the parametric families studied here fall within the smooth-fiber / isolated / stratified cases.
\end{itemize}
\end{remark}

\subsection{Operational selection rule}
\label{app:selection_rule_op}

The text of Theorem~\ref{thm:selection_rule} states that the transversal eigenvalue has exponent $2(k-1)$; the natural way to test this is to measure $\lambda_{\min}(G_\ell)$ (the smallest eigenvalue of the layer-$\ell$ K-FAC gradient factor) along the trajectory and fit a rate. In a parametric setting without gauge freedom this is sufficient. In a real neural network, raw $\lambda_{\min}$ is a worse observable than the theorem suggests: the theorem's rate is established via the spectral-separability reduction (Lemma~\ref{lem:G_spectral_separability}), which produces a single rate-carrying eigenvalue per dead direction. Raw $\lambda_{\min}$ confounds this with non-rate-carrying directions of two kinds: eigenvalues below floating-point precision relative to the matrix trace register as zero (numerical floor), and cross-weight gauge symmetries (e.g., the joint $(W_Q,W_K)\to(\alpha W_Q, W_K/\alpha)$ freedom from the attention dot-product structure; the joint $(W_V, W_O) \to (W_V M, M^{-1} W_O)$ freedom from attention output composition) produce zero-eigenvalue directions in a single-component $G_\ell$ that have nothing to do with singularity approach.

We therefore use the following operational version when applying the theorem to a transformer:

\begin{enumerate}
\item At each measurement checkpoint, record the top-$K$ and bottom-$K$ eigenvalues of $G_\ell$ for a tracked component (we use $K=16$; the parameter is only a storage budget).
\item Fit $\alpha_i$ for each eigenvalue \emph{rank} $i$ via log--log regression across the training window. The $t$-axis choice carries a calibration: fitted exponents transform as $\alpha/\beta$ under an axis asymptotically proportional to $t^\beta$. The Corollary~\ref{cor:sigma-min} axis $\sigma_{\min}(X_{\ell_\text{ref}})$ (the smallest singular value of layer $\ell_\text{ref}$'s activations) has $\beta = \ell_\text{ref}$, so the absolute gate in step~4 is calibrated at $\ell_\text{ref} = 1$; at other reference layers, divide the predicted exponent by $\ell_\text{ref}$. The model-free $1/\text{step}$ and MSE-analog $\text{test\_loss}^{1/(2L)}$ axes carry no such calibration and support only exponent \emph{ratios} between ranks, never the absolute gate.
\item Exclude ranks whose trajectory-averaged magnitude is within a tolerance of machine precision relative to the trajectory's maximum eigenvalue, so floor-trapped ranks do not masquerade as rate-carrying ones.
\item Among the non-excluded ranks, identify the transversal as the rank whose fitted $\alpha_i$ is closest to the theorem-predicted $2(L-\ell)$ (rescaled per step~2 when $\ell_\text{ref} \ne 1$), provided $|\alpha_i - 2(L-\ell)| \leq 1$ on the calibrated axis. If no rank matches within tolerance, report the component as ``theorem-reach exceeded'' rather than forcing a match. A no-match verdict has one named cause worth ruling out first: leading-score alignment (Remark~\ref{rem:genericity_G_status}) removes the transversal exponent from the spectrum while leaving the directional form intact. Read $u^\top G_\ell u$ along the candidate dead direction; a clean directional slope over an alignment-graded spectrum identifies the alignment case, and the leading-score Gram names the aligned pair.
\end{enumerate}

The rank-based pairing relies on the assumption that eigenvalue crossings are rare between consecutive measurements. A stronger version tracks eigenvectors via continuity across checkpoints and matches by eigenvector identity; this is a tractable joint-scope extension (eigenvector continuity is a standard perturbation-theoretic computation that can be added without changing the underlying selection rule), not generic future work. The current chunk operates at the cheapest tier of the four-tier observable hierarchy ($\sigma_{\min}$ as the $t$-axis, eigenvalue ranks as the rate-carriers); periodic-tier $\lambda_{\min}(G_\ell)$ would deliver tighter rate fits but at higher sample-budget cost. Both readings are used in practice: the direct $\lambda_{\min}(G_\ell)$ rate-fit (the theorem's literal statement) in clean parametric settings and on gauge-light layers, and the rank-aware operational version above where gauge zeros or the numerical floor would confound raw $\lambda_{\min}$.

\subsection{Empirical illustration: deep-linear reduced-rank regression}
\label{sec:theory:selection:rrr}

Reduced-rank regression is the standard Watanabe-side benchmark for the selection rule: a rectangular linear network $X \mapsto W_2 W_1 X$ trained on a rank-$r$ teacher has, along eigendirections approaching the rank-$r$ singular fibre, integer KL orders with directional RLCT computable in closed form via Aoyagi's resolution \citep{AoyagiWatanabe05,Aoyagi24,LauFurmanWangMurfetWei25}. On the controlled population-Fisher approach (the freeze-probe of \S\ref{app:theory:parametric_validations}), the transversal Fisher exponent on a deep-linear $6 \to 8 \to 4$ model is $1.96 \pm 0.07$ over $5$ seeds, recovering $\hat\lambda = 0.253$ against the predicted $1/(2k) = 0.25$ (Table~\ref{tab:selection_rule_app}), alongside the same recovery on the $2$- and $3$-component Gaussian-mixture merges.

\paragraph{Learned trajectories: what transfers.} The selection rule is a statement about the population Fisher on the parametric approach, and its clean validation is the freeze-probe above. On a \emph{learned} SGD trajectory through the rank-deficit cascade the reading is weaker, for the reasons in Remark~\ref{rem:population_vs_empirical_fisher}: the stored per-layer object is the loss-gradient covariance, whose global $\|\delta\|^2$ prefactor collapses the spectrum at a well-fit optimum and stalls below the asymptotic window otherwise, and the bottom-$k$ eigenvector lineage rotates during descent. A direct re-analysis of learned RRR trajectories confirms this: the per-eigenvector slopes do not partition into the predicted integers $\{0, 2, \ldots, 2r\}$: a spurious intermediate cluster near $\alpha \approx 1.4$ appears and the higher bands stay unpopulated. What does survive is a coarser signal: the \emph{count} of rate-carrying transversal lineages tracks the rank deficit, consistent with the cross-model rank-ranking the same observable provides. The integer-exponent reading requires the controlled freeze-probe; the learned trajectory recovers rank, not per-direction order. This matches the regime split of \S\ref{subsec:theory:regimes}: the static rank count survives because it does not depend on a singular approach, while the per-direction exponents, which do, require the controlled approach the freeze-probe supplies in place of a learned descent that never enters a deep compression phase.
 
\section{Fisher--curvature--volume rate chain}
\label{sec:theory:rate_chain}

\begin{figure}[t]
\centering
\includegraphics[width=\textwidth]{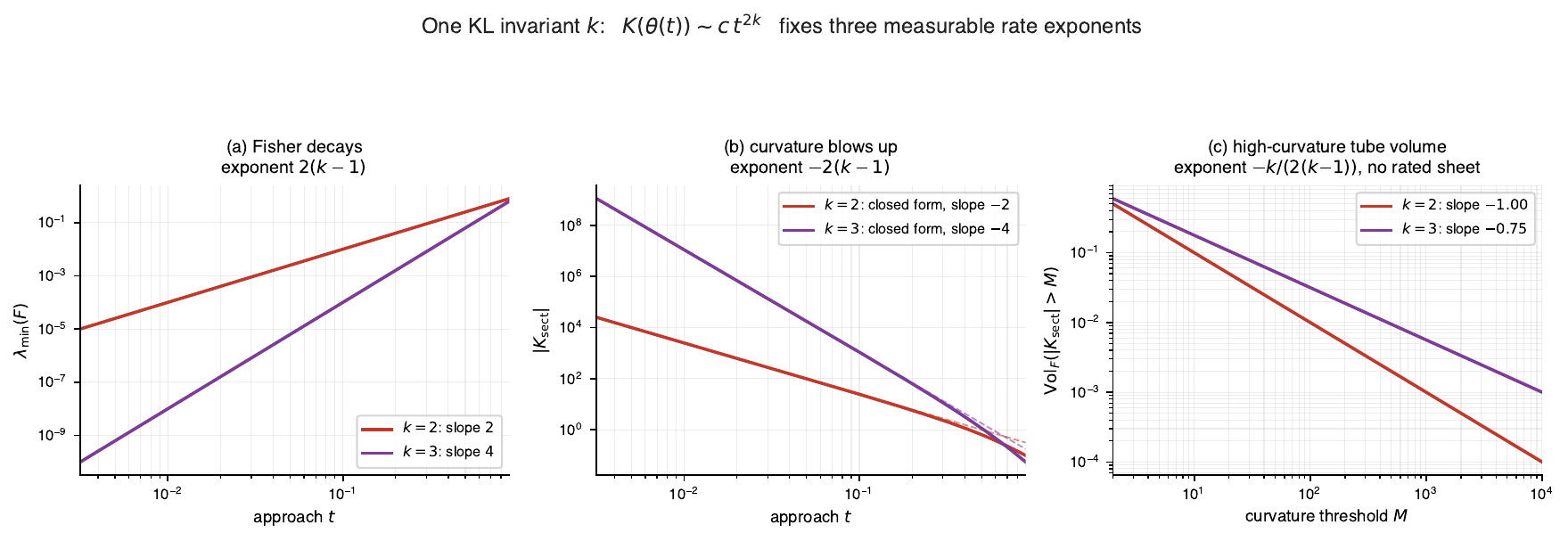}
\caption{The Fisher--curvature--volume rate chain: three measurable faces of the single KL invariant $k$. (a)~The Fisher smallest eigenvalue decays as $\Theta(t^{2(k-1)})$ (Theorem~\ref{thm:fisher_decay}). (b)~The Fisher--Riemannian sectional curvature blows up as $\Theta(t^{-2(k-1)})$ under the non-flatness condition (NF) (Proposition~\ref{prop:curvature_rate}). (c)~The high-curvature Fisher tube volume scales as $\Theta(M^{-k/(2(k-1))})$ as the threshold $M \to \infty$ in the no-rated-sheet case drawn here; rated sheet directions add their tangential exponents (Corollary~\ref{cor:volume}). Panels (a) and (c) draw the power laws for $k = 2, 3$; panel (b) draws the closed-form family curves $-1/(4t^2(1+t^2)^2)$ and $-1/(9t^4(1+t^2)^2)$ with their power-law asymptotes dashed. Each exponent is a different function of the same $k$.}
\label{fig:rate_chain}
\end{figure}

Watanabe gives the function-value volume law $\mathrm{Vol}(\{K < \varepsilon\}) \sim \varepsilon^\lambda$ as an algebraic invariant of the resolved structure. Information geometry gives the Fisher metric and, with it, sectional curvature and a Riemannian volume form. The KL order $k$ does not show up only in the Fisher metric: the same exponent that controls the directional Fisher rate also controls two further geometric quantities along the approach to $\Sigma_T$. The Fisher--Riemannian sectional curvature blows up at rate $\Theta(t^{-2(k-1)})$ when the approach plane carries curvature (condition (NF) of Proposition~\ref{prop:curvature_rate}; the flat pullback class of Remark~\ref{rem:curvature_cases} has dead directions with no curvature signal at all), and the Fisher--Riemannian volume of the high-curvature tube over the singular sheet scales as $\Theta(M^{-(k+\sum_a j_a)/(2(k-1))})$, the sheet's tangential exponents joining the dead direction's ($\Theta(M^{-k/(2(k-1))})$ with no rated sheet direction). Each is a re-expression of the same invariant. The Fisher rate is what we measure on a trajectory; the curvature divergence is what shapes natural-gradient dynamics near the singular set; the volume scaling is the pointwise analog of Watanabe's RLCT volume law. The three together form a rate chain: under each face's genericity condition, knowing one determines the others (Figure~\ref{fig:rate_chain}).

\label{app:rate_chain}
\label{app:proof:curvature_volume}

The three observables (Fisher smallest eigenvalue, Fisher-Riemannian sectional curvature, and high-curvature volume) are not independent diagnostics of a singular minimum: all three are derived from the same KL leading-order expansion $K(\theta_0 + tu) = c t^{2k} + O(t^{2k+1})$. The rate $2(k-1)$ in Fisher decay (Theorem~\ref{thm:fisher_decay}), $-2(k-1)$ in curvature blow-up (Proposition~\ref{prop:curvature_rate}), and $-(k + \sum_a j_a)/(2(k-1))$ in Fisher-Riemannian tube-volume scaling (Corollary~\ref{cor:volume}; $-k/(2(k-1))$ with no rated sheet direction, Lebesgue counterpart $-1/(2(k-1))$) are three faces of the same KL-order invariant $k$, the third joined by the sheet's tangential exponents where a rated sheet is present. Each face requires its own genericity condition and is independently measurable: score independence (Theorem~\ref{thm:fisher_decay}); a bounded transverse score-curvature moment together with surviving plane curvature (Proposition~\ref{prop:curvature_rate}); inversion of the curvature bound onto the high-curvature set (Corollary~\ref{cor:volume}). They are not, however, independent claims: they share the same load-bearing invariant.

This rate chain is the paper's mathematical backbone. It quantifies the relationships left qualitative in classical singular learning theory: \citet{Watanabe09} establishes the directional RLCT $\lambda = 1/(2k)$ and the function-value volume scaling $\mathrm{Vol}(\{K < \varepsilon\}) \sim \varepsilon^\lambda$ as algebraic invariants; \citet{AmariParkOzeki06} characterise the rank-loss \emph{direction} qualitatively. The chain below converts each into a pointwise rate exponent in the original parameter coordinates, with the multi-direction extension in Proposition~\ref{prop:volume_multi}. The rate-side of the chain (Theorem~\ref{thm:fisher_decay}) is operationalised at scale via the multi-layer K-FAC bridge (Theorem~\ref{thm:bridge}) and the residual-stream $\sigma_{\min}$ corollary (Corollary~\ref{cor:sigma-min-res}); the curvature and volume companions are validated in closed form on families with known $k$ (Validation paragraph below).

\begin{proposition}[Curvature divergence near singularities]
\label{prop:curvature_rate}
Assume the hypotheses of Theorem~\ref{thm:fisher_decay}, let $\alpha_0$ index a non-degenerate direction, write $s_i := \partial_i \log p_\theta$, and assume the transverse score-curvature moment $\expect_{p_{\theta(t)}}[(2\,\partial_{\alpha_0} s_{\alpha_0} + s_{\alpha_0}^2)^2]$ stays bounded along the approach.
\begin{enumerate}\itemsep=0pt
\item[(i)] (\emph{The Christoffel numerator vanishes for every family.}) The exact identity
\[
2\,\partial_{\alpha_0}\fisher_{1\alpha_0} \;-\; \partial_1 \fisher_{\alpha_0\alpha_0} \;=\; \expect_{p_\theta}\bigl[s_1\,(2\,\partial_{\alpha_0} s_{\alpha_0} + s_{\alpha_0}^2)\bigr]
\]
holds throughout the family, and Cauchy--Schwarz against $\fisher_{11} = \Theta(t^{2(k-1)})$ bounds the combination by $O(t^{k-1}) \to 0$. Consequently the covariant plane component $R_{1\alpha_0 1\alpha_0}(\theta(t))$ stays bounded, and $|K_{\mathrm{sect}}(t)| = O(t^{-2(k-1)})$ is a universal ceiling.
\item[(ii)] (\emph{Attained rate.}) If in addition the plane carries curvature in the limit, $\liminf_{t \to 0} |R_{1\alpha_0 1\alpha_0}(\theta(t))| > 0$ (condition (NF), non-flatness), then the Fisher-Riemannian sectional curvature in the $(u, \partial_{\alpha_0})$ plane satisfies
\[
|K_{\mathrm{sect}}(t)| \;=\; \Theta(t^{-2(k-1)}) \quad \text{as } t \to 0,
\]
and $\|R\|_F$ diverges at least at the same rate; it diverges at exactly that rate when the bounded-moment condition above holds for every transverse index in place of $\alpha_0$.
\end{enumerate}
Closed-form instances of (ii): the Gaussian mean maps $\mu = (t^2 + w^2,\ w,\ tw)$ ($k = 2$) and $\mu = (t^3 + w^2,\ w,\ tw)$ ($k = 3$) give $K_{\mathrm{sect}}(t) = -1/(4t^2(1+t^2)^2)$ and $-1/(9t^4(1+t^2)^2)$, exponents $-2$ and $-4$.
\end{proposition}

Two weaker hypotheses fail, in different ways. Under the hypothesis $\partial_1 g_{\alpha_0\alpha_0} \ne 0$ alone, the analytic family of Remark~\ref{rem:curvature_cases} satisfies the hypothesis and has identically flat Fisher geometry, so no curvature rate follows. Under the explicit non-cancellation hypothesis (NC1), $2\,\partial_{\alpha_0}\fisher_{1\alpha_0} - \partial_1\fisher_{\alpha_0\alpha_0} = c_0 + o(1)$ with $c_0 \ne 0$, the rate $|K_{\mathrm{sect}}| = \Theta(t^{-(2k-1)})$ does follow, and the metric $\mathrm{diag}(t^{2(k-1)},\, 1+t)$ realises it symbolically with coefficients $1/2, 1, 3/2$ at $k = 2, 3, 4$. No statistical family satisfies (NC1): by item~(i) the identity forces $c_0 = 0$ whenever the transverse score-curvature moment is bounded. An abstract metric can attain the $-(2k-1)$ rate; a Fisher metric cannot, and its rate is the slower $\Theta(t^{-2(k-1)})$ of item~(ii).

\begin{proof}[Proof sketch]
Work in adapted coordinates $\theta^1 = u$, with $\theta^{\alpha_0}$ the singled-out non-degenerate direction. The Fisher block has entries $\fisher_{11} = \Theta(t^{2(k-1)})$, $\fisher_{1\alpha_0} = O(t^{k-1})$, $\fisher_{\alpha_0\alpha_0} = g_{\alpha_0\alpha_0} + O(t)$ with $g_{\alpha_0\alpha_0}(0) > 0$, so $\fisher^{11} = \Theta(t^{-2(k-1)})$.

\emph{Item (i).} Differentiating $\fisher_{1\alpha_0} = \expect_{p_\theta}[s_1 s_{\alpha_0}]$ under the integral inserts the score of the moving measure: $\partial_{\alpha_0}\fisher_{1\alpha_0} = \expect[(\partial_{\alpha_0}s_1)s_{\alpha_0} + s_1 \partial_{\alpha_0}s_{\alpha_0} + s_1 s_{\alpha_0}^2]$ and $\partial_1\fisher_{\alpha_0\alpha_0} = \expect[2 s_{\alpha_0}\partial_1 s_{\alpha_0} + s_{\alpha_0}^2 s_1]$. Equality of mixed partials $\partial_{\alpha_0}s_1 = \partial_1 s_{\alpha_0}$ cancels the cross terms and leaves the identity. Cauchy--Schwarz gives $|\expect[s_1 X]| \le \fisher_{11}^{1/2}\,\|X\|_{L^2(p_\theta)} = O(t^{k-1})$ for $X = 2\partial_{\alpha_0}s_{\alpha_0} + s_{\alpha_0}^2$. The Christoffel symbols then satisfy $\Gamma^1_{\alpha_0\alpha_0} = \tfrac12\fisher^{11} \cdot O(t^{k-1}) = O(t^{-(k-1)})$ and $\Gamma^1_{11} = \tfrac12\fisher^{11}\partial_1\fisher_{11} = \Theta(t^{-1})$; the same Cauchy--Schwarz argument applied to $\partial_{\alpha_0}\fisher_{11} = \expect[2 s_1 \partial_{\alpha_0}s_1 + s_1^2 s_{\alpha_0}] = O(t^{k-1})$ gives $\Gamma^1_{1\alpha_0} = O(t^{-(k-1)})$, and symbols without an upper index $1$ are $\Theta(1)$. In the covariant component $R_{1\alpha_0 1\alpha_0}$ the second-derivative metric terms are $\Theta(1)$ by smoothness, and each $\Gamma\Gamma$ term carries the lowering factor $\fisher_{11} = \Theta(t^{2(k-1)})$ against at most two upper-$1$ symbols: $\fisher_{11}\,\Gamma^1_{11}\Gamma^1_{\alpha_0\alpha_0} = O(t^{k-2})$ and $\fisher_{11}\,(\Gamma^1_{1\alpha_0})^2 = O(1)$, both bounded for $k \ge 2$. Hence $R_{1\alpha_0 1\alpha_0} = O(1)$.

\emph{Item (ii).} The sectional curvature is $K_{\mathrm{sect}} = R_{1\alpha_0 1\alpha_0} / (\fisher_{11}\fisher_{\alpha_0\alpha_0} - \fisher_{1\alpha_0}^2)$. The denominator is the Gram determinant of the two tangent vectors; under Theorem~\ref{thm:fisher_decay}'s assumption (ii) the Schur factor $c_F - b^\top g^{-1} b > 0$ keeps it at $\Theta(t^{2(k-1)})$ with no sub-leading cancellation. Under (NF) the numerator is $\Theta(1)$, so $K_{\mathrm{sect}} = \Theta(t^{-2(k-1)})$. For the Frobenius norm, each raised index on a degenerate slot contributes one factor $\fisher^{11}$: the $(1\alpha_0 1\alpha_0)$ contribution to $\|R\|_F^2$ is $R_{1\alpha_0 1\alpha_0}^2\,(\fisher^{11})^2 (\fisher^{\alpha_0\alpha_0})^2 = \Theta(t^{-4(k-1)})$, no other component grows faster once the bounded-moment condition holds for every transverse pair, and then $\|R\|_F = \Theta(t^{-2(k-1)})$, the sectional rate (confirmed in exact arithmetic on the mixture, where both read $\delta^{-2}$).
\end{proof}

\begin{corollary}[Fisher-Riemannian volume scaling, single degenerate direction]
\label{cor:volume}
Let $S \cap U$ be a compact positive-dimensional piece of the singular sheet, carrying a single dead direction $u$ of KL order $k \ge 2$ at every sheet point, with sheet coordinates $\{\theta_a\}$ whose tangential exponents $j_a$ (Remark~\ref{rem:tangential_rates_sub}) are finite; quotient exact-gauge sheet directions first, since the Fisher volume form vanishes identically along them. Assume the hypotheses of Proposition~\ref{prop:curvature_rate}, including the bounded-moment condition and (NF), hold at every sheet point with the leading curvature coefficient bounded away from $0$ and $\infty$ on the sheet (analyticity plus compactness upgrade the pointwise rate to uniform two-sided transverse bounds $c_1 |t|^{-2(k-1)} \le |K_{\mathrm{sect}}(t, \theta_\perp)| \le c_2 |t|^{-2(k-1)}$), and assume the graded leading scores of $u$ and the sheet directions satisfy the exponent separation (G) and the graded independence (G$^+$) of Theorem~\ref{thm:selection_rule}. Write $T_M$ for the high-curvature tube over the dead-sheet slab, $T_M := \{\theta_\perp + t\,u : \theta_\perp \in S \cap U,\ |t| < t_*(M)\}$ with $t_*(M) \asymp M^{-1/(2(k-1))}$ the curvature threshold radius. Then the \emph{Fisher-Riemannian} volume of the tube satisfies
\[
\mathrm{Vol}_F(T_M) \;=\; \Theta\Bigl(M^{-\bigl(k + \sum_a j_a\bigr)/(2(k-1))}\Bigr) \quad \text{as } M \to \infty,
\]
where $\mathrm{Vol}_F$ uses the Fisher volume form on the slab. The corresponding Lebesgue volume scales as $\Theta(M^{-1/(2(k-1))})$ (integrating the constant $1$); the Lebesgue exponent carries no tangential factor. With no rated sheet direction ($\sum_a j_a = 0$, the gauge-quotient case) the Fisher exponent is $-k/(2(k-1))$: $M^{-1}$ against the Lebesgue $M^{-1/2}$ at $k = 2$, the machine-checked instance of Table~\ref{tab:machine_checked}. A network's canonical dead-unit configurations do not supply this case automatically: a dead unit's outgoing weights are flat only on the sheet itself (the unit's output becomes non-zero off it), they carry finite tangential orders $j_a \ge 1$ along the approach, and sheet points with non-zero outgoing weight drop the transversal KL order below $k$ (the coupling of Remark~\ref{rem:slice_vs_full_rlct}); the clean exponent applies to sheets that reduce to exact-gauge orbits under the quotient. The Fisher-volume scaling is the natural one for sectional-curvature-defined regions because $K_{\mathrm{sect}}$ is itself a Fisher-metric quantity. Watanabe's RLCT volume formula \citep{Watanabe09} gives $\mathrm{Vol}_{\mathrm{Leb}}(\{K < \varepsilon\}) \sim \varepsilon^\lambda$ with $\lambda = 1/(2k)$ in the single-degenerate case, on the function-value level set; the corollary reads the curvature level set, and the two scalings are different invariants of the same singular minimum.

Two scope notes. First, the corollary measures the tube: the full high-curvature \emph{set} in a neighbourhood $U$ also shrinks along the non-degenerate transversal coordinates (the curvature decays off the singular locus in those directions too), so its volume scales with an exponent strictly below the tube exponent; on the closed-form $k = 2$ plane, where the singular locus is a point, the measured set exponent is near $-3/2$ against the ray's $-1$. Second, the mixture merge is not a realizing instance: its full parameter space has pairwise-aligned leading scores (alignment cosines $+1$ and $-1$ in the limit; Remark~\ref{rem:genericity_G_status}): no eigenvalue carries the transversal exponent, $\sqrt{\det\fisher} = \Theta(\delta^6)$, and the spectrum grades as $\{0, 0, 4, 8\}$ at the probe weight, so it fails the (G$^+$) independence here and falls outside these hypotheses. The realizing instance is the four-component mean map $\mu = (t^2 + w^2,\ w,\ tw,\ t^2 v)$: sheet the $v$-line, uniform KL order $2$ with $a_2$-coefficient $(1+v^2)/2$, tangential exponent $j_v = 2$, plane curvature $-\bigl(4(1+v^2)t^2\bigr)^{-1}(1 + o(1))$ uniform on compact $v$-windows, and tube volume measured at slope $-2.000$ against the predicted $-(k + j_v)/(2(k-1)) = -2$.
\end{corollary}

\begin{proof}[Proof sketch]
The uniform two-sided bounds identify the tube's radius with the curvature threshold: the upper bound keeps the high-curvature region of the slab inside $\{|t| < c_1' M^{-1/(2(k-1))}\}$, the lower bound fills $\{|t| < c_2' M^{-1/(2(k-1))}\}$, and both constants are uniform over the sheet. A pointwise rate along one approach path does not suffice here: Proposition~\ref{prop:curvature_rate} at $\theta_0$ alone controls no $\theta_\perp \ne 0$ point of the tube. The Fisher-Riemannian volume form on the slab in adapted coordinates is $\mathrm{dvol}_F = \sqrt{\det\fisher_{\mathrm{slab}}(t)}\, dt\, d\theta_\perp$, over the block spanned by $u$ and the sheet directions. The dead factor carries $\Theta(t^{2(k-1)})$ by the Schur expansion of Theorem~\ref{thm:fisher_decay}, and each rated sheet direction carries its tangential factor $\Theta(t^{2 j_a})$ (Remark~\ref{rem:tangential_rates_sub}). Under (G) and (G$^+$), Lemma~\ref{lem:G_spectral_separability} splits the slab block's eigenvalues into the transversal $\Theta(t^{2(k-1)})$ group and the tangential $\Theta(t^{2 j_a})$ groups, so the determinant is the product of these orders with no cross-term cancellation (the mixture's Hermite grading, Remark~\ref{rem:genericity_G_status}, shows what alignment does instead), and $\sqrt{\det\fisher_{\mathrm{slab}}(t)} = \Theta\bigl(t^{k-1+\sum_a j_a}\bigr)$ uniformly over the sheet. Integrating, $\int_0^{c M^{-1/(2(k-1))}} t^{k-1+\sum_a j_a}\, dt = \Theta\bigl(M^{-(k+\sum_a j_a)/(2(k-1))}\bigr)$.
\end{proof}

\begin{proposition}[Fisher-volume scaling, additive multi-direction case]
\label{prop:volume_multi}
Suppose the KL function admits the additive transversal normal form
\[
K(u) \;=\; u_1^{2k_1} + u_2^{2k_2} + \cdots + u_r^{2k_r} + (\text{non-degenerate quadratic in remaining coordinates}),
\]
with each direction carrying KL order $k_i \ge 2$, each $(u_i, \text{non-degenerate})$ plane satisfying the bounded-moment and (NF) conditions of Proposition~\ref{prop:curvature_rate}, and the per-direction leading scores satisfying the graded-Gram non-degeneracy (G$^+$) of Theorem~\ref{thm:selection_rule}. Under (G$^+$) the Fisher volume form factorises at leading order on the polydisc, $\sqrt{\det\fisher} = \Theta(\prod_i |u_i|^{k_i - 1})$, and the Fisher-volume of the multi-direction high-curvature polydisc satisfies
\[
\mathrm{Vol}_F\Bigl(\bigl\{|u_i| < c_i M^{-1/(2(k_i - 1))} \text{ for all } i\bigr\} \cap U\Bigr) \;=\; \Theta\!\left(\prod_{i=1}^r M^{-k_i / (2(k_i - 1))}\right),
\]
which provides a lower bound on $\mathrm{Vol}_F(\{|K_{\mathrm{sect}}| > M\})$ when the polydisc lies inside the high-curvature set (a cross-plane non-cancellation condition). The additive normal form alone does not fix the volume exponent, so (G$^+$) is load-bearing: the Gaussian mean map $\mu = (\rho\cos\varphi,\ \rho\sin\varphi)$ with $\rho^2 = 2(u_1^4 + u_2^4)$, $\varphi = u_1 + u_2$ satisfies $K = u_1^4 + u_2^4$ exactly, yet its two leading scores share their angular component, (G$^+$) fails on the diagonal, $\sqrt{\det\fisher} = 4\,|u_1^3 - u_2^3|$ degenerates there, and the polydisc volume scales as $M^{-5/2}$ against the product formula's $M^{-2}$; both halves are machine-checked (Table~\ref{tab:machine_checked}): the factorised polydisc law at every $r$, and the twisted map's density and volume exponent in closed form. Any multi-direction volume reading should therefore test (G$^+$) on the leading scores before using the product exponent, with the twisted map as the ready negative control.

\textbf{Comparison to RLCT volume.} Watanabe's RLCT formula gives the Lebesgue volume of the \emph{function-value} level set $\{K < \varepsilon\}$ as $\sim \varepsilon^\lambda \cdot (-\log\varepsilon)^{m-1}$ with $\lambda = \min_i (a_i + 1)/(2 k_i)$ in the resolved-Hironaka monomial form $K = \prod_i u_i^{2 k_i}$ (the resolved form is multiplicative; the normal form above is additive). The two formulas are different invariants of the same singularity: $\lambda$ is set by the \emph{slowest-vanishing} direction (min over $i$), while the per-direction Fisher-curvature volume above is a \emph{product} over all directions because the curvature integral factorises across the additive normal form. For two and three components the monomial-form volumes admit exact closed forms on the unit box: tied orders give $\varepsilon^{1/(2k)}$ times a polynomial of degree $m-1$ in $\log(1/\varepsilon)$ at multiplicity $m \le 3$, distinct orders give a two-term power law with no logarithm, and in each case the log-volume slope converges to $\lambda$, so the $\varepsilon$-scan reads the correct exponent at any multiplicity.
\end{proposition}

\paragraph{Validation on parametric families with known \texorpdfstring{$k$}{k}.} The chain is validated on analytic families with known KL order $k$, on the controlled population-Fisher approach (the freeze-probe of \S\ref{app:theory:parametric_validations}, where the residual stays $\Theta(1)$ and the rate readout is not subject to the learned-trajectory confound of Remark~\ref{rem:population_vs_empirical_fisher}). The rate side (Theorem~\ref{thm:fisher_decay}) is the selection-rule recovery of Table~\ref{tab:selection_rule_app}: transversal exponents matching $2(k-1)$ on the $2$- and $3$-component Gaussian-mixture merges and on deep-linear reduced-rank regression, recovering $\hat\lambda \approx 1/4$. The curvature side (Proposition~\ref{prop:curvature_rate}) is verified in closed form on curved Gaussian mean maps: $K_{\mathrm{sect}} = -1/(4t^2(1+t^2)^2)$ at $k = 2$ and $-1/(9t^4(1+t^2)^2)$ at $k = 3$, fitted exponents $-2.000$ and $-4.000$, and the identity of item~(i) checks symbolically (residual zero) on these families and on richer curved variants. The 2-component Gaussian mixture merge lands on the same family exponent through a different route: Hermite orthogonality at the base Gaussian makes every Fisher block even in the separation and aligns the split score with the variance score, the $(\delta, \sigma^2)$ plane Gram cancels to $\delta^4$ while the covariant component is $O(\delta^2)$, and the exact sectional rate is $\delta^{-2}$ (coefficient $-29200/2457$ at $w = 0.35$), numerically equal to $t^{-2(k-1)}$ at $k = 2$. The metric normal form $\mathrm{diag}(t^{2(k-1)}, 1+t)$ realises the faster $-(2k-1)$ (coefficients $1/2, 1, 3/2$ at $k = 2, 3, 4$) and certifies that class as non-empty for abstract metrics; item~(i) rules out its Fisher realisation. The volume side (Corollary~\ref{cor:volume}) follows analytically from the curvature rate by integrating the Fisher volume form over the high-curvature tube ($-k/(2(k-1)) = -1$ at $k = 2$ with no rated sheet direction; $-2$ on the four-component sheet witness with $j_v = 2$, measured $-2.000$), and the multi-direction product law of Proposition~\ref{prop:volume_multi} is checked numerically on a block product family (slope $-2.000$ at $k = (2,2)$) against the twisted witness (slope $-5/2$).

\begin{remark}[Asymptotic-regime measurement precondition]
\label{rem:asymptotic_window}
Theorem~\ref{thm:fisher_decay} states a pointwise rate exponent in the asymptotic limit $t \to 0$; Proposition~\ref{prop:volume_multi} states a volume exponent in the threshold limit $M \to \infty$. Empirical measurement of the rate exponents requires the trajectory to reach a regime where the leading-order rate dominates sub-leading corrections, which in turn requires the dead-direction value at the start of the measurement window to be substantially above the loss-side noise floor (the irreducible $\sigma_{\mathrm{noise}}^2$ from data noise, or in noise-free testbeds the floating-point cov-accumulation floor $\sigma_{\max}\sqrt{\varepsilon_{\mathrm{fp}}}$). On a constructed canonical-aligned bridge $W_\ell = \mathrm{diag}(1, \ldots, 1, t_0, \ldots, t_0)$ with $r$ dead diagonal entries, the natural depth-invariant measurement-control variable is $t_0^L$ (the singular value of the composed forward map at initialization). When comparing rate predictions across $L$ at fixed $\sigma_{\mathrm{noise}}$, holding $t_0^L$ constant isolates the rate prediction from the asymptotic-regime accessibility; varying $t_0$ across $L$ at fixed value conflates the two. Empirically the measurement test passes when $t_0^L \gtrsim \sigma_{\mathrm{noise}}^2$ and fails (slope unfittable; $R^2$ collapses) when $t_0^L \ll \sigma_{\mathrm{noise}}^2$. The noise floor is a necessary precondition, and it is not the only failure mode: a sub-leading correction with a large coefficient produces the same confident wrong fit with no noise at all. On a $\delta$-perturbed variant of the (G$^+$) counterexample family, every hypothesis of Theorem~\ref{thm:selection_rule} holds and the asymptotic-window slopes read $\{4.00, 2.00, 0.00\}$ for every $\delta > 0$, while a finite window above the crossover $t \sim \delta/\varepsilon$ reads $\{3.48, 3.24, 1.19\}$: a misclassification driven entirely by the hypothesis-constant margin. Window placement below the crossover scale is part of the measurement protocol, on the same footing as the noise floor.
\end{remark}

\begin{remark}[Attainment classes of the curvature rate]
\label{rem:curvature_cases}
Two exactly-computed classes bracket Proposition~\ref{prop:curvature_rate}. \emph{The flat class, where (NF) fails.} The Gaussian mean-map family $\mu(\theta_1, \theta_2) = (\theta_1^k,\ \theta_2(1 + \theta_1))$ has a dead direction of KL order $k$ at the origin and satisfies the weaker hypothesis $\partial_1 g_{\alpha_0\alpha_0} = 2 \ne 0$, yet its Fisher metric is the pullback of the flat Euclidean metric under an equal-dimension map, so every sectional curvature vanishes identically. This failure class is structural: every singularity obtained by a rank-dropping reparametrisation of a regular family of the same dimension is isometrically flat off the singular sheet, so no curvature reading sees its dead directions. \emph{The Gram-cancelling class, where the exponent survives without the plane hypotheses.} The 2-component Gaussian mixture merge falls outside the intact-Gram reading of the $(\delta, \sigma^2)$ plane: Hermite orthogonality forces every Fisher block to be even in the separation $\delta$, and the split score aligns with the variance score at leading order, so the plane's Gram determinant cancels from $\delta^2$ to $\delta^4$. Exact symbolic computation gives $|K_{\mathrm{sect}}| \sim \delta^{-2}$ in the $(\delta, \sigma^2)$ plane (coefficient $-29200/2457$ at $w = 0.35$), the covariant component at $O(\delta^2)$ against the cancelled Gram $\delta^4$: the same exponent as item~(ii)'s $t^{-2(k-1)}$ at $k = 2$, reached through a different mechanism. The $(\delta, \bar\mu)$ plane stays finite ($K_{\mathrm{sect}} \to 1/4$ at $w = 0.35$, $\to -1/2$ at $w = 1/2$). (NF) with an intact plane Gram is attainable along a positive-dimensional sheet: the four-component mean map $\mu = (t^2 + w^2,\ w,\ tw,\ t^2 v)$ has the $v$-line as its sheet, uniform KL order $2$ there, and the $(t, w)$-plane at sheet point $v$ carries the closed-form curvature $-\bigl(4(1+v^2)t^2\bigr)^{-1}(1+o(1))$ with the Gram intact. The sheet-tangent directions themselves never join such a plane, since the family is constant along the sheet and their Fisher entries vanish at every sheet point. The multi-direction cross-planes stay bounded. Exact symbolic computation gives finite $K_{\mathrm{sect}}$ in both the $(\bar\mu, w)$ plane and the two-dead $(\text{split}, w)$ cross-plane: the split score and the weight score are Hermite-orthogonal at the base Gaussian (one couples to $\mathrm{He}_2$, the other to $\mathrm{He}_1$), so the cross-terms vanish at leading order, no super-degenerate eigenvector forms, and the cross-coupling cannot accelerate the divergence. Numerical sectional-curvature estimates near the merge are dominated by inverse-metric error amplification; the exact computation is the reliable reading.
\end{remark}

\begin{remark}[Sharp ceiling in a Gram-cancelling plane]
\label{rem:cancelled_plane_ceiling}
Write $\rho \ge 0$ for the cancellation depth of an approach plane, so that its Gram determinant scales as $\Theta(t^{2(k-1)+\rho})$; the mixture's $(\delta, \sigma^2)$ plane has $\rho = 2$. A bounded covariant plane component then caps $|K_{\mathrm{sect}}|$ at $O(t^{-(2(k-1)+\rho)})$, and the cap is attained: for every $q \ge 2$ the Gaussian mean map $\mu(t, w) = \bigl(t^2 + w,\ \tfrac{2}{3}c\,t^3 + c\,tw,\ t^{q+1}/(q+1)\bigr)$ has KL order $k = 2$, tangent alignment $\partial_t \mu = 2t\,\partial_w \mu + t^{q} e_3$ along the approach curve $w = 0$ (with $e_3$ the third coordinate vector), hence depth $\rho = 2q - 2$, and $K_{\mathrm{sect}} = -c^2\, t^{-2q}\,(1 + o(1))$ in exact closed form, with covariant component $-c^2 + O(t)$. The Gram-cancelling route is therefore not capped at the family rate: the covariant components stay bounded, the Gram cancellation supplies the entire divergence, and deeper alignment realises any faster exponent. Like the mixture, the family fails assumption (ii) of Theorem~\ref{thm:fisher_decay}, so Proposition~\ref{prop:curvature_rate}'s intact-Gram reading never covers it. The mixture nevertheless attains the family rate at asymmetric weights, and its heat identity is the reason: the Gaussian density satisfies $\partial_y^2 \varphi = 2\,\partial_{\sigma^2} \varphi$ in the observation $y$, the identity that aligns the split score with the variance score in Remark~\ref{rem:curvature_cases} and forces the critical pairings to decay in step with the cancelled Gram. Cancelled planes undershoot as well as overshoot: at the symmetric weight $w = 1/2$ the mixture's pairings decay past the critical order and the plane stays bounded, below the family rate.
\end{remark}

\paragraph{Curvature divergence certifies a dead direction.}
\label{rem:rank_loss_signature}
Summing every component, the Frobenius norm $\|R\|_F^2 = R_{ijkl} R^{ijkl}$ registers a dead direction coarsely. Where no direction degenerates, the metric is uniformly non-degenerate and both $\|R\|_F$ and $K_{\mathrm{sect}}$ stay bounded, so a curvature divergence certifies a dead direction. The converse needs the non-flatness condition: a dead direction drives $\|R\|_F$ and $K_{\mathrm{sect}}$ to diverge at the family rate $t^{-2(k-1)}$ when the approach plane carries curvature ((NF) of Proposition~\ref{prop:curvature_rate}), while the flat class of Remark~\ref{rem:curvature_cases} has dead directions with $R \equiv 0$ that no curvature reading detects. One object drives every divergent reading: the dead direction that Theorem~\ref{thm:fisher_decay} surfaces as Fisher decay. The exponent certifies $k$ only in intact planes, where the leading score of the dead direction has a component independent of the transverse score: in a score-aligned plane the sectional-curvature exponent is $2(k-1)+\rho$ with $\rho$ the Gram cancellation depth (Remark~\ref{rem:cancelled_plane_ceiling}), so a curvature-slope reading on its own bounds nothing about $k$ without a Gram check, while the Fisher-eigenvalue readout of Theorem~\ref{thm:fisher_decay}, the selection rule's channel, is unaffected.
 
\label{app:chain}

We collect here the supporting definitions and remarks that frame the rate-chain results of \S\ref{sec:theory:rate_chain} (Theorem~\ref{thm:fisher_decay}, Theorem~\ref{thm:selection_rule}, Proposition~\ref{prop:curvature_rate}, Corollary~\ref{cor:volume}, Proposition~\ref{prop:volume_multi}). The content below names the genericity hypotheses, makes the Fisher matrix structure explicit, compares the rate exponents to Watanabe's RLCT, and indicates how the smooth-fiber result extends to stratified and resolved singular sets.

The families this paper treats are algebraic-statistical: real-analytic $\theta \mapsto p_\theta$ with analytic singular locus, covering Gaussian mixtures, Boltzmann machines, HMMs, reduced-rank regressions, and neural networks with smooth activations (tanh, sigmoid, softplus). For such families the Fisher decay along a dead direction is polynomial with the KL-order exponent (Theorem~\ref{thm:fisher_decay}). A derivative-bounded-below clause would not serve as the definition: every Fisher-entry derivative vanishes along the mixture merge by Hermite parity, so such a clause fails on the paper's own families. Networks with piecewise-linear activations (ReLU) are not strictly real-analytic; the zeta-function machinery extends to piecewise-analytic $K$, and \citet{WeiMurfet22} treat the ReLU analyticity boundary explicitly. The formal results of this paper apply to the analytic case.

\paragraph{Comparison to Watanabe's RLCT theory.}\label{rem:fisher_decay_watanabe}
Watanabe's RLCT theory \citep{Watanabe09} gives the asymptotic form of the marginal log-likelihood and the KL order $k$ along resolved coordinates obtained via Hironaka resolution. It does not give a pointwise rate for $u^\top \fisher(\theta(t)) u$ or for $\lambda_{\min}(\fisher(\theta(t)))$ along an analytic path approaching the singular minimum: the resolved-coordinate analysis runs on the Jacobian of the blow-up, and the Fisher rate in original coordinates falls outside it. Theorem~\ref{thm:fisher_decay}'s contribution is a rate on the \emph{unresolved} Fisher in original parameter coordinates, expressed in terms of the same KL order $k$.

\begin{corollary}[Fisher matrix structure near a single-degenerate singularity]
\label{cor:fisher_structure}
Under the assumptions of Theorem~\ref{thm:fisher_decay}, the Fisher matrix in adapted coordinates $(\theta^1 = u, \theta^\alpha)$ satisfies:
$$
\fisher_{11} = c_F\,t^{2(k-1)} + O(t^{2k-1}), \quad \fisher_{1\alpha} = b_\alpha\,t^{k-1} + O(t^k), \quad \fisher_{\alpha\beta} = g_{\alpha\beta} + O(t),
$$
where $c_F = 2ck^2 > 0$, $b_\alpha = k\,\expect_{p^*}[a_k \cdot \partial_\alpha \log p^*]$, $g_{\alpha\beta} = \expect_{p^*}[(\partial_\alpha \log p^*)(\partial_\beta \log p^*)]$ is positive definite, and $c_F > b^\top g^{-1} b$ (assumption~(ii) of Theorem~\ref{thm:fisher_decay}).
\end{corollary}

\begin{proof}
The expressions follow from the proof of Theorem~\ref{thm:fisher_decay} via $F_u(t) = \expect_{p^*}[s_u^2 \exp f]$ and the score decomposition $s_u = k\,a_k\,t^{k-1} + O(t^k)$, $s_\alpha = \partial_\alpha \log p^* + O(t)$.
\end{proof}

\paragraph{Fiber structure resolves the apparent circularity in selection.}\label{rem:tangential}
To apply the selection rule (Theorem~\ref{thm:selection_rule}), one must identify which eigenvalues are transversal. This requires knowing the fiber $S$ (from the model structure), not $k$ itself. At the Gaussian mixture merge, $S = \{\mu_1 = \mu_2\}$ is known from the model definition and the normal bundle is the $(\mu_1, \mu_2)$ subspace; for reduced-rank regression, $S$ includes the gauge orbit and the normal bundle contains the rank-deficiency direction. With the fiber fixed, the controlled freeze-probe recovers transversal exponents matching $2(k-1)$ on all three families (Table~\ref{tab:selection_rule_app}).

\paragraph{Extension to singular fibers.}\label{rem:singular_fibers}
Theorem~\ref{thm:selection_rule} requires $S$ to be a smooth submanifold (so that $N_{\theta_0}S$ and $\Pi$ are well-defined). Two extensions beyond smooth fibers are possible without the full resolution machinery:

\emph{(i) Isolated singularities.} If $S$ is smooth except at finitely many points, puncture the singular points and apply Theorem~\ref{thm:selection_rule} to $S \setminus \{\text{singular points}\}$. The selection rule holds on any approach path avoiding these points.

\emph{(ii) Stratified fibers.} If $S$ decomposes into smooth strata (typically a Whitney stratification, of which normal-crossing varieties $\{x_1 \cdots x_k = 0\}$ are the canonical example), apply Theorem~\ref{thm:selection_rule} to each stratum separately; the eigenvalue rates depend on which stratum is approached. Practical machine-learning singular sets are almost always of this type because they arise from products of coordinate-aligned conditions:
\begin{itemize}\itemsep=1pt
\item Multi-component Gaussian mixtures with several pairwise collapses, e.g.\ for $K=3$ components the stratification $\{\mu_1{=}\mu_2\} \cup \{\mu_2{=}\mu_3\} \cup \{\mu_1{=}\mu_3\}$ meeting along the triple-collapse line $\{\mu_1{=}\mu_2{=}\mu_3\}$.
\item Rank-stratified reduced-rank regression: $\{\mathrm{rank}(W) \le r\}$ decomposes into $\{\mathrm{rank}=r\} \sqcup \{\mathrm{rank}\le r{-}1\} \sqcup \cdots$, each rank stratum smooth.
\item Mixture models with simultaneous weight and center collapse ($w_i \to 0$ \emph{and} $\mu_i \to \mu_j$); strata indexed by the combinatorial type.
\item Overparameterised neural networks with intersecting gauge orbits (dead-neuron loci, permutation symmetry, layer-wise rescaling), which generically meet transversally and form a normal-crossing pattern.
\end{itemize}
On each stratum the selection rule applies; the rates one measures depend on which stratum the trajectory $\theta(t)$ is converging to.

\emph{(iii) Genuinely non-stratified fibers.} For singular sets that are not unions of smooth pieces meeting transversally (cuspidal singularities $y^2 = x^3$, the Whitney umbrella $x^2 = y^2 z$, tangent self-intersections such as $\{x^2 = y^4\}$, higher-codimension high-order vanishing $\{f_1 = f_2 = 0\}$ with both $f_i$ vanishing to high order along the same locus), the smooth-fiber assumption fails and the piecewise machinery does not apply directly. Hironaka's resolution of singularities \citep{Hironaka64} provides a birational map $\phi : \Theta' \to \Theta$ under which the resolved fiber $\phi^{-1}(S)$ is smooth, so Theorem~\ref{thm:selection_rule} applies in the resolved coordinates. Translating the eigenvalue rates back to the original coordinates requires controlling the Jacobian of $\phi$, which has the structure $|J_\phi| \sim \prod |u_i|^{a_i}$. This pullback analysis is a natural direction for future work.

We are not aware of standard deployed architectures whose loss landscape produces case (iii); the families validated in this paper (Gaussian mixtures, reduced-rank regression) all have smooth fibers, and the standard combinatorial multi-collapse cases above are covered by case (ii).

\paragraph{Cross-coupling in the multi-degenerate case.}\label{rem:multi_degenerate}
The decoupled assumption in Proposition~\ref{prop:volume_multi} excludes cross-coupling between degenerate directions. When Fisher cross-terms $\fisher_{u_iu_j}$ create super-degenerate eigenvectors ($\lambda_{\min} \sim t^{a_1 + a_2}$ where $a_i$ are individual exponents), curvature in the $(u_i, u_j)$ \emph{cross-plane} can diverge faster than in any single $(u_i, v)$ plane, raising the high-curvature volume above the per-direction product bound. The 2-component Gaussian mixture merge does not realise this. Its single-direction rate is $\delta^{-2}$, the family exponent $t^{-2(k-1)}$ at $k = 2$, reached through the Gram-cancelling route of Remark~\ref{rem:curvature_cases}, and its split and weight scores are Hermite-orthogonal at the base Gaussian, so the cross-terms vanish at leading order and exact symbolic computation gives bounded $K_{\mathrm{sect}}$ in both the $(\bar\mu, w)$ plane and the two-dead $(\text{split}, w)$ cross-plane (numerical estimates near the merge are unreliable, dominated by inverse-metric error amplification). At the full coincidence of both means and variances, the two-dead $(\text{mean-split}, \text{variance-split})$ cross-plane diverges at $t^{-2}$, the single-direction rate, while the dead--non-degenerate planes there stay finite; the degenerate block carries no super-degenerate eigenvalue ($\lambda_{\min} \sim t^2$), so the cross-coupling curves the plane without accelerating the rate. The mixture shows no acceleration even at its deepest coincidence; whether any other family realises it is open. A tight multi-degenerate volume formula would require analyzing curvature in all $\binom{r}{2}$ cross-planes.

\label{app:complementarity}

\paragraph{Selection-rule recovery on Gaussian-mixture boundaries.} The selection rule (Theorem~\ref{thm:selection_rule}) recovers $\hat\rlct$ from the Fisher eigenvalue with exponent $2(k-1)$ on each transversal singular family. We check the two canonical boundaries, the weight singularity $w \to 0$ and the merge $\mu_1 \to \mu_2$.

\textbf{Part (a): Weight singularity (tangential).} At $w \to 0$, the degenerate direction is $\mu_1$ (the mean of the vanishing component), but this direction lies \emph{within} the singular set $\{w=0\}$: changing $\mu_1$ at $w=0$ does not change the distribution ($K(w{=}0, \mu_1) = 0$ for all $\mu_1$). Hence Theorem~\ref{thm:fisher_decay}'s assumption $K(\theta_0 + tu) = ct^{2k}$ along the degenerate direction is violated: the KL is identically zero, not a polynomial in $t$.

The Fisher diagonal entry is $\fisher_{\mu_1\mu_1} = w^2 \int [(x{-}\mu_1)/\sigma^2]^2 \phi_1^2/p_\theta\,dx$, and it has two regimes. Near $x = \mu_1$ the mixture density is $p_\theta \approx w\phi_1 + (1-w)\phi_2$, so the vanishing component dominates there only while $w \gtrsim \exp(-\Delta^2/2\sigma^2)$ for a separation $\Delta = |\mu_1 - \mu_2|$. In that regime the integrand is $O(\phi_1/w)$ over a region of non-negligible mass, the integral picks up an $O(1/w)$ factor, and $\fisher_{\mu_1\mu_1} \sim w$. Once $w$ drops below the overlap threshold the surviving component dominates everywhere, the integral converges to a constant, and $\fisher_{\mu_1\mu_1} \sim w^2$. Numerically, at $\Delta = 5\sigma$ the measured exponent holds near $1.0$ across $w \in [10^{-5}, 10^{-1}]$, while at $\Delta = 1.5\sigma$, whose threshold is $w \approx 0.32$, it runs $1.43 \to 1.99$ over the same sweep. So $w^2$ is the leading order as $w \to 0$ at fixed separation, and $\sim w$ describes the well-separated finite-$w$ regime where the fits below were taken.

The remaining eigenvalues are not constant either: $\fisher_{ww}$ grows through the fitted window (the score $(\phi_1{-}\phi_2)/p_\theta$ grows while the component overlap still registers, and saturates at the finite limit $\fisher_{ww} \to e^{\Delta^2} - 1$, unreachable inside the window at large separation). Fitted exponents in the well-separated regime ($\Delta = 5$, $w \in [10^{-3}, 10^{-1}]$) are near $1.1$, near $0$, and near $-1$ (the middle eigenvalue's exponent is window-sensitive), giving $\kappa \sim w^{-2}$ and, through $\kappa \sim w^{-1/(2\rlct)}$, an estimate $\hat\rlct \approx 0.25$. The approximate agreement with $\rlct = 1/4$ is empirical: the compounding of a slower-than-predicted $\lambda_\mathrm{min}$ decay with $\lambda_\mathrm{max}$ growth yields a condition number exponent that coincidentally approximates $-1/(2\rlct) = -2$. The agreement is not evidence that the rule applies here, and the next paragraph says why it cannot.

\textbf{Part (b): Merge singularity (transversal).} At $\mu_1 \to \mu_2$, the degenerate direction is $\delta = \mu_1 - \mu_2$. We have $K(\delta) = c\delta^4 + O(\delta^5)$ for symmetric weights, so $k = 2$. The merge direction is transversal to the singular set $\{\delta = 0\}$, satisfying Theorem~\ref{thm:fisher_decay}'s assumption. Direct computation on the means-only slice $(\mu_1, \mu_2)$, with the weight and variance held fixed, gives $\lambda_{\min}(\fisher) = \Theta(\delta^2)$ (matching the prediction $2(k{-}1) = 2$), recovering $\hat\rlct = 1/(2k) = 1/4$; in the full family the leading scores align and no eigenvalue carries this exponent (Remark~\ref{rem:genericity_G_status}), while the directional form $u^\top \fisher u$ along the merge direction keeps it.

Parts (a) and (b) bracket the rule's precondition. At the merge, the degenerate direction leaves the singular set and the KL grows as $\delta^4$, so the recovery returns $\hat\rlct = 1/4$ exactly. At the weight boundary, the degenerate direction lies inside $\{w = 0\}$ where the KL stays identically zero; Theorem~\ref{thm:fisher_decay}'s polynomial-KL hypothesis fails there, and the near-match to $1/4$ is coincidental, the condition-number exponents happening to compound to $-2$. The rule reaches transversal degeneracies; a tangential one enters only after the fiber and its normal bundle are identified (the fiber-structure remark above).

The same transversal recovery holds across the three families of Table~\ref{tab:selection_rule_app}: on the deep-linear reduced-rank model the full Fisher splits into ${\sim}55$ gauge-killed eigenvalues, ${\sim}15$ constant, and $5$--$6$ transversal at exponent $\approx 2(k-1)$, the structure the selection rule predicts. Generalisation to arbitrary analytic families is open, but the consistency across qualitatively different singularity structures (multi-degenerate merges and gauge-coupled rank deficiency) suggests the rule is general for transversal singularities.
  
\part{The bridge framework for deep networks}

Part~II read the rate off a single dead direction in one Fisher matrix. A trained network is deeper and messier: its Fisher is layered, carries a reparameterisation gauge that contributes many uninformative near-zero eigenvalues, and is assembled from repeated architectural primitives. This part lifts the rate primitive to that setting, the bridge's deep-network half, in three moves. The multi-layer K-FAC bridge (Section~\ref{sec:theory:bridge}) carries the exponent across layers as a per-layer rate ladder, read in coordinates that quotient out the gauge zeros; within it, a forward-backward duality (Section~\ref{sec:theory:ag_duality}) makes the two K-FAC factors mirror images whose product is depth-independent. Composition additivity (Section~\ref{sec:theory:composition}) adds rates across stacked blocks where a scalar-transfer hypothesis holds, and marks where pure attention chains break it. The architectural instantiations (Section~\ref{sec:theory:architectural}) then derive a closed-form rate for each primitive of a modern transformer from the primitive's own algebra. One KL-order exponent runs through all three; only its reading point changes.

\paragraph{Reach.} These results are architecture-specific, the first narrowing from Part~II's model-agnostic reach. The K-FAC bridge assumes a layered architecture with canonical alignment of the dead direction across layers; composition additivity assumes the scalar-transfer hypothesis, which MLP and pre-norm residual chains satisfy; pure attention fails it at every depth, and the additive conclusion survives only on cells with $k \le p + 2$; and each architectural instantiation carries its own assumptions, named where it is stated. The $\sigma_{\min}$ depth-invariance corollary (Corollary~\ref{cor:sigma-min-res}) is the least architecture-bound of the group: it holds for any residual DAG (the directed acyclic computation graph of a residual network) with exact-identity skips, under the corollary's own hypotheses (the canonical-aligned approach, activation classes (P1), (P3), and (P2) with $\phi'(0) > 0$, and the residual-branch non-cancellation condition).
 
\section{The multi-layer K-FAC bridge}
\label{sec:theory:bridge}

Lifting the single-direction rate to a deep network runs straight into the reparameterisation gauge. A layered Fisher has $\sim h^2 L$ parameters and an inner-matrix gauge group $GL(h)^{L-1}$ that contributes $\sim h^2(L-1)$ near-zero eigenvalues independent of the singular geometry, so the obvious diagnostic $\lambda_{\min}(F_{\rm raw})$ on the full Fisher is dominated by gauge zeros and says nothing about the singularity rate. K-FAC \citep{MartensGrosse15} is the fix: factorise the per-layer Fisher as $F_\ell \approx A_\ell \otimes G_\ell$. The block-diagonal restriction discards the cross-layer Fisher blocks on which the gauge kernel is supported, leaving per-layer factors whose smallest eigenvalues carry the singularity signal directly. This section lifts the rate into those coordinates. Under canonical alignment of the dead direction across layers, the per-layer K-FAC gradient factor inherits a layer-dependent rate $\lambda_{\min}(G_\ell) = \Theta(t^{2(L-\ell)})$, and the dual activation factor $\lambda_{\min}(A_\ell) = \Theta(t^{2(\ell-1)})$. The product is layer-independent: $\lambda_{\min}(A_\ell)\,\lambda_{\min}(G_\ell) = \Theta(t^{2(L-1)})$ at every $\ell$. The proof reduces to a Schur complement on the non-dead Fisher block at each layer, controlled by the gauge-kernel structure of $GL(h)^{L-1}$ acting on cross-layer Fisher blocks. We carry the framework in coordinates Amari's lineage already uses; what is new is the rate.

\label{app:bridge_setup}

\ifsupp The dead-direction rate $\Theta(t^{2K(\ell)})$ at any singular minimum is established by a single shared mechanism: the dead-direction Schur reduction of Lemma~\ref{lem:integral-reduction-sub} below.\else The dead-direction rate $\Theta(t^{2K(\ell)})$ across every bridge result in this paper (the feedforward case (Theorem~\ref{thm:bridge}), the cross-entropy case (Theorem~\ref{thm:bridge_ce}), the residual-DAG case (Theorem~\ref{thm:bridge_res})\ifexpanded, the rectangular-width case (Theorem~\ref{thm:bridge_rect}), the multi-direction asymmetric case (Theorem~\ref{thm:bridge_multi}), the bias-augmented case (Theorem~\ref{thm:bridge_bias}), and the single-head attention case (Theorem~\ref{thm:bridge_attn})\else, and the further architectural instantiations (rectangular widths, multi-direction asymmetric, bias-augmented, single-head attention) below\fi{}) is established by a single shared mechanism: the dead-direction Schur reduction of Lemma~\ref{lem:integral-reduction-sub} below.\fi{} Each architectural extension is an instantiation that verifies (i) canonical alignment of the dead direction at every layer (or its rectangular / multi-direction / residual-DAG generalisation), (ii) Schur-reducible Fisher block in canonical coordinates (dead diagonal $\Theta(t^{2K(\ell)})$, dead-row off-diagonals $O(t^{K(\ell)})$, non-dead block $\succ 0$ at $\Theta(1)$), and (iii) dead-direction-only $t$-scaling on the backward chain. The architecture-specific work is the verification; the rate conclusion follows from the lemma.

\ifsupp We refer to the result of Lemma~\ref{lem:integral-reduction-sub} as the \emph{Bridge Framework Lemma}.\else We refer to the result of Lemma~\ref{lem:integral-reduction-sub} as the \emph{Bridge Framework Lemma} and the architecture-specific verifications above as \emph{instantiations}; Theorem~\ref{thm:bridge} is the canonical instantiation, and the architectural subsections that follow instantiate the framework on the relaxations relevant to actual neural networks.\fi

Consider an $L$-layer network $f(x; W_1, \ldots, W_L) = W_L \phi(W_{L-1} \phi(\cdots \phi(W_1 x)))$ with $W_\ell \in \reals^{h \times h}$, no biases, activation $\phi$ in class (P1), (P2), or (P3) of the theorem. Data $x \sim \normal(0, I_h)$, target $y = M^* x + \varepsilon$ with $M^* = \diag(1, \ldots, 1, 0) \in \reals^{h \times h}$ (so the dead-direction component of the target is pure noise: $y^{(h)} = \varepsilon^{(h)}$), $\varepsilon \sim \normal(0, \sigma^2 I_h)$, MSE loss. The singular configuration is $W_\ell^* = \diag(1, \ldots, 1, 0)$ in canonical coordinates for all $\ell$, with the same dead direction $e_h$ shared across layers (the symmetric canonical-aligned approach). The transversal approach is $W_\ell(t) = W_\ell^* + t \cdot e_h e_h^\top = \diag(1, \ldots, 1, t)$ for all $\ell$. For activation class (P1) linear, $f(x; \theta(t)) = (\prod_\ell W_\ell(t)) \, x = \diag(1, \ldots, 1, t^L) \, x$, so the dead-channel residual is $f^{(h)} - y^{(h)} = t^L x^{(h)} - \varepsilon^{(h)}$ and $K(\theta(t)) - K(\theta^*) = \tfrac{1}{2\sigma^2} t^{2L} \expect[(x^{(h)})^2] = \tfrac{1}{2\sigma^2} t^{2L}$. For (P2) and (P3) the constant changes (e.g., factor $1/2$ from the half-Gaussian survival event for ReLU) but the order is $\Theta(t^{2L})$ in all three classes, so $k = L$ uniformly.

\begin{lemma}[Forward dead-component propagation]
\label{lem:forward-dead-sub}
Along the symmetric approach, for each $\ell \in \{1, \ldots, L\}$:
linear (P1) $a_\ell^{(h)} = t^\ell x^{(h)}$;
smooth (P2), with the standing hypothesis $\phi(0) = 0$ in addition to $\phi'(0) \ne 0$ (the Taylor step below requires it; activations with $\phi(0) \ne 0$ shift the dead pre-activation to $\Theta(t)$ at every depth and break the rate ladder), $a_\ell^{(h)} = (\phi'(0))^{\ell-1} t^\ell x^{(h)} + O(t^{\ell+1})$;
ReLU (P3) $a_1^{(h)} = t\,x^{(h)}$ (no indicator at the first pre-activation) and $a_\ell^{(h)} = t^\ell x^{(h)} \mathbf{1}[x^{(h)} > 0]$ for $\ell \ge 2$, $t > 0$.
For (P1) and (P2), $a_\ell^{(h)} = \Theta(t^\ell)$ pointwise on $\{x^{(h)} \ne 0\}$ (full-probability event under Gaussian input). For (P3) at $\ell \ge 2$, $a_\ell^{(h)} = \Theta(t^\ell)$ on the survival event $\{x^{(h)} > 0\}$ (probability $1/2$) and is identically zero on $\{x^{(h)} < 0\}$. In all three classes the second moment satisfies $\expect_x[(a_\ell^{(h)})^2] = \Theta(t^{2\ell})$, with the constant $\expect[(x^{(h)})^2] = 1$ for (P1) and for (P3) at $\ell = 1$, $(\phi'(0))^{2(\ell-1)}$ for (P2), and $\expect[(x^{(h)})^2 \mathbf{1}[x^{(h)}>0]] = 1/2$ for (P3) at $\ell \ge 2$. This is the form used by all downstream lemmas.
\end{lemma}

\begin{proof}
Induction on $\ell$. Base $a_1^{(h)} = t \cdot x^{(h)}$. Step: $a_{\ell+1}^{(h)} = t \cdot \phi(a_\ell^{(h)})$. Linear gives $t^{\ell+1} x^{(h)}$. Smooth: the hypothesis $\phi(0) = 0$ gives the Taylor form $\phi(u) = \phi'(0) u + O(u^2)$ (without it, $\phi(a_\ell^{(h)}) = \phi(0) + O(t^\ell)$ and the chain restarts at order $t$ every layer); apply IH. ReLU: on $\{x^{(h)} > 0\}$, $a_1^{(h)} > 0$ for $t > 0$ and $\phi$ acts as identity, so $a_{\ell+1}^{(h)} = t \cdot a_\ell^{(h)}$ recurses to $t^\ell x^{(h)}$; on $\{x^{(h)} < 0\}$, $\phi(a_1^{(h)}) = 0$ zeroes the chain at $\ell = 2$ and all subsequent $\ell$. The pointwise statement is therefore not almost-sure for ReLU, but the second-moment statement holds because both events have positive Gaussian probability and the surviving event contributes $\Theta(t^{2\ell})$.
\end{proof}

\begin{lemma}[Backward dead-component magnitude]
\label{lem:backward-dead-sub}
$\expect[(\delta^{(L, h)})^2] = \sigma^2 + O(t^{2L}) = \Theta(1)$, and $\expect[(\delta^{(\ell, h)})^2] = \Theta(t^{2(L - \ell)})$ for $\ell \in \{1, \ldots, L-1\}$.
\end{lemma}

\begin{proof}
By the canonical structure of $W_{\ell+1}(t) = \diag(1, \ldots, 1, t)$, $(W_{\ell+1}^\top \delta^{(\ell+1)})^{(h)} = t \cdot \delta^{(\ell+1, h)}$ (the $h$-th row of $W_{\ell+1}^\top$ has only the diagonal entry $t$). The backward chain rule gives
\[
\delta^{(\ell, h)} = \phi'(a_\ell^{(h)}) \cdot t \cdot \delta^{(\ell+1, h)}.
\]
By Lemma~\ref{lem:forward-dead-sub}, $\phi'(a_\ell^{(h)}) = \Theta(1)$ at leading order: linear gives $\phi' \equiv 1$; smooth gives $\phi'(0) + O(t^\ell)$; ReLU gives $1$ on the survival event $\{x^{(h)} > 0\}$, $0$ otherwise (probability $1/2$; on the complement the pre-activation equals $0$ exactly for $\ell \ge 2$, and we take the subgradient convention $\phi'(0) = 0$ there, which moves only the constant, never the rate, since $\delta^{(L,h)} = -\varepsilon^{(h)}$ on that event). At the output $\delta^{(L, h)} = -\varepsilon^{(h)} + \Theta(t^L)$, so $\expect[(\delta^{(L, h)})^2] = \sigma^2 + O(t^{2L})$. Backward induction $\delta^{(\ell, h)} = \Theta(t) \cdot \delta^{(\ell+1, h)}$ gives $\expect[(\delta^{(\ell, h)})^2] = \Theta(t^{2(L - \ell)})$ for $\ell < L$.
\end{proof}

\begin{lemma}[Bridge Framework Lemma: dead-direction Schur reduction]
\label{lem:integral-reduction-sub}
The dead-direction-row entries of the K-FAC factors $A_\ell$ (input covariance) and $G_\ell$ (gradient covariance) admit, in canonical coordinates,
\[
(A_\ell)_{h, h} = \Theta(t^{2(\ell-1)}), \quad (A_\ell)_{h, j} = O(t^{\ell-1}) \quad (j \ne h);
\]
\[
(G_\ell)_{h, h} = \Theta(t^{2(L-\ell)}), \quad (G_\ell)_{h, j} = o(t^{2(L-\ell)}) \quad (j \ne h);
\]
and the non-dead block (entries with $j, j' \ne h$) is positive-definite at $\Theta(1)$. Hence the Schur complement at the dead row gives $\lambda_{\min}(A_\ell) = \Theta(t^{2(\ell-1)})$ and $\lambda_{\min}(G_\ell) = \Theta(t^{2(L-\ell)})$, even when (for ReLU at $\ell \ge 2$) the $A$-side off-diagonals exceed the dead diagonal in raw magnitude.
\end{lemma}

\begin{proof}
\emph{$A_\ell$: entries.} In canonical coordinates the dead-direction activation factorises as $a_\ell^{(h)} = t^\ell f(x^{(h)})$ from Lemma~\ref{lem:forward-dead-sub} (with $f(x) = x$, $f(x) = (\phi'(0))^{\ell-1} x$, or $f(x) = x \mathbf{1}[x > 0]$ depending on activation class), so $\phi(a_{\ell-1}^{(h)}) = t^{\ell-1} \tilde f(x^{(h)})$ for an analogous $\tilde f$, depending on $x^{(h)}$ alone. Non-dead activations $\phi(a_{\ell-1}^{(j)})$ ($j \ne h$) propagate without the dead-direction $t$-factor and remain $\Theta(1)$, depending on $x^{(j)}$ alone (because the canonical $W_m^* = \diag(1,\ldots,1,0)$ structure routes coordinate $j$ only to itself and the dead coordinate $h$ only to itself). Hence $\phi(a_{\ell-1}^{(h)}) \perp \phi(a_{\ell-1}^{(j)})$ under Gaussian-isotropic input, and $(A_\ell)_{h, j} = \expect[\phi(a_{\ell-1}^{(h)})] \cdot \expect[\phi(a_{\ell-1}^{(j)})]$. For (P1) linear $\expect[t^{\ell-1} x^{(h)}] = 0$, killing the cross term identically. For (P3) ReLU $\expect[t^{\ell-1} x^{(h)} \mathbf{1}[x^{(h)} > 0]] = t^{\ell-1}/\sqrt{2\pi}$, multiplied by $\expect[\phi(a_{\ell-1}^{(j)})] = \Theta(1)$, gives $(A_\ell)_{h,j} = \Theta(t^{\ell-1})$. For (P2) smooth, $\expect[\phi(a_{\ell-1}^{(h)})] = O(t^\ell)$ since $\expect[x^{(h)}] = 0$, giving $(A_\ell)_{h,j} = O(t^\ell)$. The dead diagonal is $(A_\ell)_{h,h} = \expect[\phi(a_{\ell-1}^{(h)})^2] = \Theta(t^{2(\ell-1)})$, and the non-dead block has $(A_\ell)_{j, j'} = \expect[\phi(a_{\ell-1}^{(j)}) \phi(a_{\ell-1}^{(j')})] = \Theta(1)$ as the input layer's data covariance.

\emph{$A_\ell$: eigenvalue via Schur.} Note that for ReLU at $\ell \ge 2$, the off-diagonal $\Theta(t^{\ell-1})$ exceeds the dead diagonal $\Theta(t^{2(\ell-1)})$ in raw magnitude. Despite this, the dead eigenvalue is determined by the Schur complement of the non-dead block. Writing $A_\ell$ in block form as $\begin{pmatrix} M & v \\ v^\top & d \end{pmatrix}$ with $M$ the non-dead $(h-1) \times (h-1)$ block ($\succ 0$, $\Theta(1)$), $d = (A_\ell)_{h,h} = \Theta(t^{2(\ell-1)})$, and $v$ the dead-row vector of off-diagonals (of order $\Theta(t^{\ell-1})$ for ReLU, smaller for linear/smooth), the dead-direction eigenvalue equals the Schur complement at leading order:
\[
\lambda_{\min}(A_\ell) \;=\; \bigl(d - v^\top M^{-1} v\bigr)\,\bigl(1 - o(1)\bigr),
\]
where $\lambda_{\min} \le d - v^\top M^{-1} v$ always holds for a PSD block matrix (the Schur complement is $1/(A_\ell^{-1})_{hh}$, an upper bound on the smallest eigenvalue), and the relative gap is $O(t^{2(\ell-1)})$ here because the dead block is $\Theta(t^{2(\ell-1)})$ against the $\Theta(1)$ non-dead block.
For ReLU, $v^\top M^{-1} v = \Theta(t^{\ell-1}) \cdot \Theta(1) \cdot \Theta(t^{\ell-1}) = \Theta(t^{2(\ell-1)})$. The Schur complement is $\Theta(t^{2(\ell-1)}) - \Theta(t^{2(\ell-1)})$, and the leading constants combine: the dead-row's outer-product structure $v v^\top$ (rank 1 since $v$ is computed via independence as $\expect[\phi(a_{\ell-1}^{(h)})] \cdot \expect[\phi(a_{\ell-1}^{(j)})]_j$) absorbs at most a fixed fraction of $d$'s magnitude, leaving the residual $\Theta(t^{2(\ell-1)}) \cdot (1 - c)$ with $c < 1$. Under the canonical setup of this lemma the constant is closed-form $c = (h-1)/(\pi(\pi + h - 2)) \le 1/\pi$ (Lemma~\ref{lem:schur_constant_quantitative}); under more general input distributions only the qualitative bound persists (Remark~\ref{rem:schur_constant_qualitative}). Hence $\lambda_{\min}(A_\ell) = \Theta(t^{2(\ell-1)})$. For (P1) linear and (P2) smooth the off-diagonals are smaller still (giving $v^\top M^{-1} v = O(t^{2\ell})$, strictly subleading), so the Schur complement is dominated by the diagonal $d$ and the same conclusion holds without the rank-1 cancellation argument.

\emph{$G_\ell$: entries (cleaner than $A_\ell$).} Symmetric argument using Lemma~\ref{lem:backward-dead-sub}: $\delta_\ell^{(h)} = t^{L-\ell} \cdot g(\varepsilon^{(h)}, x^{(h)})$ with $g$ a (class-dependent) function of the dead-direction noise and input only, while $\delta_\ell^{(j)}$ for $j \ne h$ depends on $(\varepsilon^{(j)}, x^{(j)})$. Independence across coordinates gives $\expect[\delta_\ell^{(h)} \delta_\ell^{(j)}] = \expect[\delta_\ell^{(h)}] \expect[\delta_\ell^{(j)}]$. For the $G$-side, the noise $\varepsilon^{(h)}$ has zero mean, and $\expect[\delta_\ell^{(h)}]$ is a $t^{L-\ell}$-prefactored expectation of $-\varepsilon^{(h)} + \Theta(t^L)$, which is $O(t^{L-\ell+L}) = O(t^{2L-\ell})$ via the noise's zero mean (cleaner than the $A$-side because $\varepsilon$ is zero-mean while the activations are not). The non-dead factor $\expect[\delta_\ell^{(j)}]$ carries no $t$, so $(G_\ell)_{h, j} = O(t^{2L-\ell})$, which is $o(t^{2(L-\ell)})$ for every $\ell \ge 1$ because $2L - \ell$ exceeds $2(L-\ell)$ by $\ell$. It is therefore strictly subleading to the dead diagonal, and the resulting Schur correction $v^\top M^{-1} v = O(t^{2(2L-\ell)})$ is subleading by a further factor $t^{2L}$. The Schur reduction is therefore unnecessary for $G_\ell$: $\lambda_{\min}(G_\ell) = (G_\ell)_{h,h} (1 + o(1)) = \Theta(t^{2(L-\ell)})$.
\end{proof}

\begin{lemma}[Schur-cancellation constant under isotropic Gaussian input]
\label{lem:schur_constant_quantitative}
Under the canonical setup of Lemma~\ref{lem:integral-reduction-sub} with ReLU activation and isotropic Gaussian input $x \sim \normal(0, I_h)$, the rank-1 outer-product cancellation in the $A_\ell$ Schur complement at any $\ell \ge 2$ obeys, in closed form,
\[
c \;=\; \frac{h-1}{\pi(\pi + h - 2)}, \qquad c \;\le\; \frac{1}{\pi} \;<\; 1 \quad \text{uniformly in } h \ge 2,
\]
with $c \to 1/\pi$ as $h \to \infty$. Consequently the dead-direction eigenvalue satisfies the asymptotic lower bound
\[
\lambda_{\min}(A_\ell) \;\ge\; \tfrac{1}{2}\bigl(1 - \tfrac{1}{\pi}\bigr) \cdot t^{2(\ell-1)} \cdot \bigl(1 - o(1)\bigr) \;\approx\; 0.341\, t^{2(\ell-1)} \bigl(1 - o(1)\bigr).
\]
The $(1 - o(1))$ factor is required because $\lambda_{\min}$ is below the Schur complement by a relative $O(t^2)$ correction. Its coefficient depends on the width and decays with it, measured on the exact leading-order matrix as $0.114\,t^2$ at $h = 4$, $0.084\,t^2$ at $h = 8$, and $0.0019\,t^2$ at $h = 512$, so the correction matters least in the wide regime the bound is used in. The displayed constant is exact in the limit and numerically unviolated at every tested $(h, t)$ up to $h = 512$, $t = 0.5$, and it is not proved as a finite-$t$ inequality.
\end{lemma}

\begin{proof}
By Lemma~\ref{lem:forward-dead-sub} for ReLU, $\phi(a_{\ell-1}^{(h)}) = t^{\ell-1} x^{(h)} \mathbf{1}[x^{(h)} > 0]$, and the canonical-coordinate structure $W_m^* = \diag(1,\ldots,1,0)$ recursing through idempotent ReLU gives $\phi(a_{\ell-1}^{(j)}) = \mathrm{ReLU}(x^{(j)})$ for $j \ne h$. The dead diagonal is $d = (A_\ell)_{h,h} = \expect[(\phi(a_{\ell-1}^{(h)}))^2] = t^{2(\ell-1)} \expect[(x^{(h)})^2 \mathbf{1}[x^{(h)} > 0]] = \tfrac{1}{2} t^{2(\ell-1)}$. The dead-row off-diagonal $v_j = (A_\ell)_{h,j}$ factors by the canonical-coordinate independence of $x^{(h)}$ and $x^{(j)}$ as $\mu_h \mu_j$ with $\mu_h = t^{\ell-1}/\sqrt{2\pi}$ and $\mu_j = 1/\sqrt{2\pi}$ (half-Gaussian first moments), so $v = (t^{\ell-1}/(2\pi)) \mathbf{1}_{h-1}$. The non-dead block has entries $M_{jj'} = \tfrac{1}{2}$ for $j = j'$ and $\tfrac{1}{2\pi}$ for $j \ne j'$, i.e. $M = \alpha I + \beta J$ with $\alpha = \tfrac{1}{2} - \tfrac{1}{2\pi}$, $\beta = \tfrac{1}{2\pi}$, and $J = \mathbf{1}\mathbf{1}^\top$. Since $\mathbf{1}$ is an eigenvector of $M$ with eigenvalue $\alpha + (h-1)\beta = \tfrac{1}{2} + \tfrac{h-2}{2\pi} = (\pi + h - 2)/(2\pi)$, $M^{-1}\mathbf{1} = (2\pi/(\pi + h - 2)) \mathbf{1}$. Hence
\[
v^\top M^{-1} v \;=\; \frac{t^{2(\ell-1)}}{(2\pi)^2} \cdot \frac{2\pi (h-1)}{\pi + h - 2} \;=\; \frac{t^{2(\ell-1)} (h-1)}{2\pi(\pi + h - 2)}.
\]
Writing the Schur complement as $d(1 - c)$ with $c = (v^\top M^{-1} v)/d$ gives the stated $c = (h-1)/(\pi(\pi + h - 2))$. The bound $c \le 1/\pi$ follows from $(h-1)/(\pi + h - 2) \le 1$, with equality only in the limit $h \to \infty$. The lower bound on $\lambda_{\min}(A_\ell)$ follows by substituting $c \le 1/\pi$ into $d(1-c)$.
\end{proof}

\paragraph{Numerical verification of the closed form.} The closed-form ratio $\lambda_{\min}(A_\ell) / (A_\ell)_{h,h} = 1 - c$ is verified by direct Monte Carlo computation of $A_\ell$ on the canonical setup above. At $t = 0.1$ and $n = 2 \times 10^5$ Gaussian input samples (three seeds), the empirical ratio matches the predicted $1 - (h-1)/(\pi(\pi + h - 2))$ to relative error $\le 2.3 \times 10^{-3}$ at $\ell \in \{2, 3\}$ across $h \in \{4, 8, 16, 32, 64, 128\}$ (e.g.\ $h = 4$: predicted $0.81427$, measured $0.81401$; $h = 128$: predicted $0.68697$, measured $0.68682$). Finite-$t$ Taylor corrections are visible at $t = 0.5$ (max relative error $\sim 3\%$ at $h = 4$) and shrink monotonically as $t \to 0$, consistent with the leading-order statement.

\begin{remark}[Scope of the quantitative constant; per-instantiation status]
\label{rem:schur_constant_qualitative}
Lemma~\ref{lem:schur_constant_quantitative} pins down the Schur-cancellation constant $c$ in the canonical setup of Lemma~\ref{lem:integral-reduction-sub}\ifsupp\else{} (Theorem~\ref{thm:bridge})\fi{} (ReLU, isotropic Gaussian input, canonical-coordinate alignment). The proof generalises with no structural change to any activation $\phi$ for which the dead and non-dead activation moments factor by canonical-coordinate independence: writing $\mu_h = \expect[\phi(a_{\ell-1}^{(h)})]$, $\nu = \expect[\phi(a_{\ell-1}^{(j)})]$, $d = \expect[\phi(a_{\ell-1}^{(h)})^2]$, $d_{\mathrm{nd}} = \expect[\phi(a_{\ell-1}^{(j)})^2]$, the Schur-cancellation constant takes the form
\[
c \;=\; \frac{\mu_h^2\, \nu^2\, (h-1)}{d \cdot \bigl(d_{\mathrm{nd}} + (h-2)\, \nu^2\bigr)},
\]
recovering the ReLU value above when the corresponding half-Gaussian moments are substituted, and reducing to $c = 0$ trivially whenever $\nu = 0$ or $\mu_h = 0$ at leading order (linear, tanh, GELU, SiLU, and any LN-output input where coordinate means vanish by construction). The qualitative bound $c < 1$ is a property of the rank-1 outer-product structure together with $\mathrm{Var}[\phi(a_{\ell-1}^{(h)})] > 0$ (any non-constant $\phi$ on a non-degenerate input), and persists in each architectural instantiation that invokes Lemma~\ref{lem:integral-reduction-sub}. The per-instantiation status of the closed form is as follows.
\begin{itemize}
\item \emph{Multi-direction\ifsupp\else{} (\S\ref{sec:theory:arch:multi})\fi, MSE.} Trivially $c_i = 0$ for every direction $i$: by canonical-coordinate independence and zero-mean Gaussian noise, both the rank-1 outer-product term $v_i^\top M^{-1} v_i$ and the cross-direction terms $w_{ij}^2/d_j$ vanish at leading order ($v_i, w_{ij}$ are products of zero-mean expectations, $O(t^{\Pi^{(i)} + L})$), so the dead-direction Schur complement equals the dead diagonal at leading order without cancellation. The closed form is $\lambda_{\min}(G_\ell) \cap u^{(i)} = (G_\ell)_{u^{(i)} u^{(i)}}(1 + o(1))$.
\item \emph{Multi-direction, cross-entropy.} Non-trivial: the data-averaged softmax Hessian off-diagonal $-\expect_x[p_i p_j] = \Theta(1)$ makes $w_{ij} = \Theta(t^{\Pi^{(i)} + \Pi^{(j)}})$ load-bearing, and the constant $c_i$ depends on the data-averaged Hessian's full block structure, which we do not pin down here.
\item \emph{Residual DAG\ifsupp\else{} (\S\ref{app:bridge_res})\fi, single shortest path.} When $|\mathcal{P}^*(\ell)| = 1$ (typical for residual blocks of standard transformers, where one residual skip dominates the graph distance to each layer), the path-Gram at the dead row reduces to the parent's structure with $L - \ell$ replaced by $K(\ell)$, and the parent constant transfers verbatim: $c \le 1/\pi$.
\item \emph{Residual DAG, multi-shortest-path.} When $|\mathcal{P}^*(\ell)| > 1$\ifsupp\else{} (e.g.\ a residual DAG with overlapping skips $(v_0, v_2)$ and $(v_1, v_3)$, where two shortest paths reach $v_3$; the single-block graphs of Corollary~\ref{cor:res_block} have a unique shortest path at every node)\fi, cross-path correlations enter the path-Gram off-diagonal and the constant depends on graph topology (how many shortest paths share which layers); the qualitative bound $c < 1$ still applies but the closed form is graph-specific.
\end{itemize}
We carry the qualitative $c < 1$ in the body and rely on Lemma~\ref{lem:schur_constant_quantitative} plus the per-case statements above for any instantiation that needs the explicit constant.
\end{remark}

\begin{lemma}[Non-dead entries (square-case; rectangular extension via Lemma~\ref{lem:integral-reduction-sub} adapted to non-square widths)]
\label{lem:non-dead}
For $j \neq h$: $(A_\ell)_{jj} = \Theta(1)$ and $(G_\ell)_{jj} = \Theta(1)$, both independent of $t$ at leading order (since non-dead diagonals of $W_\ell^*$ are identity).
\end{lemma}

\label{app:kfac}

\begin{algorithm}[ht]
\caption{Per-Layer Structural Monitoring via K-FAC G-Factor}
\label{alg:detection}
\begin{algorithmic}[1]
\REQUIRE Model with K-FAC-tracked layers; monitoring interval $K$; threshold $\tau$; at least $n/d \ge 100$ gradient samples per tracked layer (so $\lambda_{\min}(G_\ell)$, hence $\kappa$, is not floor-biased)
\FOR{every $K$ training epochs}
    \STATE Compute K-FAC $G_\ell$ for each tracked layer $\ell$
    \STATE $\kappa^* \gets \max_\ell \kappa(G_\ell)$; \quad $\ell^* \gets \arg\max_\ell \kappa(G_\ell)$
    \STATE \textbf{Depth profile:} compute mean $\kappa(G_\ell)$ for early vs.\ deep layer groups
    \IF{$\kappa^* > \tau$}
        \STATE \textbf{Flag:} structural transition at layer $\ell^*$
        \STATE \emph{Optional (quantitative $\hat\rlct$):} recover $(\hat\rlct, m)$ from the Fisher spectrum along the approach via the selection rule (Theorem~\ref{thm:selection_rule}, \S\ref{app:proof:selection_rule}) and its multi-component extension (Theorem~\ref{thm:multi_component_rates}), without posterior sampling; fall back to SGLD-based LLC estimation \citep{LauFurmanWangMurfetWei25} when the singular structure is not trajectory-identifiable
    \ENDIF
\ENDFOR
\end{algorithmic}
\end{algorithm}

For parametric models where all parameters are meaningful (no overparameterization), Algorithm~\ref{alg:detection} simplifies: compute the full Fisher eigenvalues and distinguish singularity types by eigenvalue degeneracy count.

\paragraph{Estimand for quantitative recovery.} The flagged G-factor is the empirical Fisher (training-loss gradients), the right object for detecting a transition. Quantitative $\hat\rlct$ recovery instead reads the expected (true) Fisher: off the optimum the empirical G-factor carries a residual prefactor that corrupts the rate exponent (Remark~\ref{rem:population_vs_empirical_fisher}, Corollary~\ref{cor:emp_vs_exp_ce}).

\paragraph{Position in the observable hierarchy.}
Algorithm~\ref{alg:detection} belongs to the \emph{periodic} tier (per-$K$-epochs) of the four-tier observable hierarchy: $\sigma_{\min}(X_\ell)$ on the residual stream is the \emph{real-time} tier (cheap forward-pass at every step), $\lambda_{\min}(G_\ell)$ via K-FAC is the \emph{periodic} tier as deployed here, $u^\top G u$ at a fixed direction is the \emph{checkpoint} tier (single backward pass on demand), and the full Fisher spectrum is the \emph{offline} tier. All four are computable at LLM scale; the trade-off is cost-vs-information rather than feasibility. Algorithm~\ref{alg:detection} is appropriate when periodic structural-transition flags are needed during training.

\paragraph{Detecting task expansion from the K-FAC G-factor.}

Task expansion is a common continual-learning transition: a trained classifier's head grows to cover new classes, with the new outputs randomly initialised and the encoder reused. Such a transition is invisible to the loss alone, yet it leaves a sharp signature in the gradient geometry, because the fresh outputs inject a high-variance gradient block that the old, near-converged classes lack. The G-factor $G_\mathrm{head}$ isolates that signature from the input-side conditioning, and the mechanism below shows $\kappa(G_\mathrm{head})$ spikes at the transition.

For a linear layer $y = Wx + b$, the K-FAC approximation \citep{MartensGrosse15} gives $F_\ell \approx A_\ell \otimes G_\ell$ with $A_\ell = \frac{1}{N}\sum_n a_na_n^\top$ (input covariance) and $G_\ell = \frac{1}{N}\sum_n g_ng_n^\top$ (gradient covariance).

\emph{Protocol.} In the controlled task-expansion verification setup analysed here, the classification head is \emph{expanded} from $k_1$ to $k_2 > k_1$ outputs: old $k_1$ output weights are preserved (assumption A2), new $k_2 - k_1$ outputs are randomly initialized (A1), and the encoder is unchanged (A3).

\begin{proposition}[Task expansion is visible in the head's G-factor]
\label{prop:task_expansion_gfactor}
Under (A1)--(A3) at a head expansion from $k_1$ to $k_2$ outputs:
\begin{enumerate}\itemsep=0pt
\item[(a)] The head's gradient covariance is approximately block-diagonal, $G_\mathrm{head} \approx \diag(\sigma_\mathrm{new}^2 I_{k_2-k_1},\, \sigma_\mathrm{old}^2 I_{k_1}) + E$, so its condition number satisfies $\kappa(G_\mathrm{head}) \ge \sigma_\mathrm{new}^2 / \sigma_\mathrm{old}^2 - O(\|E\|_\mathrm{op} / \sigma_\mathrm{old}^2)$, while the input-side $\kappa(A_\mathrm{head})$ is unchanged across the transition.
\item[(b)] When the map from layer $\ell$ to the head acts with a sample-independent Jacobian $J_\ell$ across the evaluation batch (exact when that part of the network is linear; an approximation otherwise), every hidden layer inherits this anisotropy through $G_\ell = J_\ell^\top G_\mathrm{head} J_\ell$, so the head is its origin. The bound $\kappa(G_\ell) \le \kappa(J_\ell)^2 \kappa(G_\mathrm{head})$ follows, and no ordering between $\kappa(G_\ell)$ and $\kappa(G_\mathrm{head})$ does.
\end{enumerate}
\end{proposition}

\begin{proof}
\textbf{Part (a).} The per-sample output gradient decomposes as $g_n = (g_n^{(\mathrm{old})}, g_n^{(\mathrm{new})}) \in \reals^{k_1} \times \reals^{k_2-k_1}$. By (A2), the old class outputs are near-converged, so per-sample losses in old class directions are small: $\expect[g_n^{(\mathrm{old})}(g_n^{(\mathrm{old})})^\top] = \sigma_\mathrm{old}^2 I_{k_1} + E_\mathrm{old}$ where $\sigma_\mathrm{old}^2$ is small and $E_\mathrm{old}$ is a small perturbation. By (A1), the new class outputs are randomly initialized: per-sample losses are large, giving $\expect[g_n^{(\mathrm{new})}(g_n^{(\mathrm{new})})^\top] = \sigma_\mathrm{new}^2 I_{k_2-k_1} + E_\mathrm{new}$ with $\sigma_\mathrm{new}^2 \gg \sigma_\mathrm{old}^2$.

By (A1), the new weights are drawn independently of the prior task's training history, so the cross-covariance $\expect[g_n^{(\mathrm{new})}(g_n^{(\mathrm{old})})^\top]$ is zero in expectation over the weight initialization (the new output directions are random relative to the learned old-class structure; treating the realised cross block as small is a concentration step over the $k_2 - k_1$ fresh rows, adopted here as part of (A1)). Hence $G_\mathrm{head}$ is approximately block-diagonal:
$$
G_\mathrm{head} \approx \begin{pmatrix} \sigma_\mathrm{new}^2 I_{k_2-k_1} & 0 \\ 0 & \sigma_\mathrm{old}^2 I_{k_1} \end{pmatrix} + E,
$$
with $\|E\|_\mathrm{op}$ small relative to $\sigma_\mathrm{new}^2$. The eigenvalues of $G_\mathrm{head}$ are approximately $\{\sigma_\mathrm{new}^2, \ldots, \sigma_\mathrm{old}^2, \ldots\}$, giving $\kappa(G_\mathrm{head}) \geq \sigma_\mathrm{new}^2 / \sigma_\mathrm{old}^2 - O(\|E\|/\sigma_\mathrm{old}^2)$.

The input covariance $A_\mathrm{head} = \frac{1}{N}\sum_n a_na_n^\top$ depends on penultimate-layer activations, which by (A3) are unchanged at the transition. Hence $\kappa(A_\mathrm{head})$ is stable across the transition. The raw K-FAC Fisher $\kappa(F_\mathrm{head}) = \kappa(A_\mathrm{head}) \cdot \kappa(G_\mathrm{head})$ is dominated by $\kappa(A) \sim 10^{12}$ in overparameterized models; the $G$-factor isolates the task-change signal.

\textbf{Part (b): Signal origin.} For hidden layer $\ell$, the gradient is $g_n^\ell = J_{\ell,n}^\top g_n^{\mathrm{head}}$ where $J_{\ell,n} = \partial y / \partial h^\ell$ is the Jacobian at sample $n$ (fixed at transition by A3). In general $G_\ell = \frac{1}{N}\sum_n J_{\ell,n}^\top g_n^{\mathrm{head}} (g_n^{\mathrm{head}})^\top J_{\ell,n}$; under (b)'s sample-independent-Jacobian premise $J_{\ell,n} \equiv J_\ell$ this collapses to the conjugation
$$
G_\ell = J_\ell^\top G_\mathrm{head} J_\ell,
$$
and the $\kappa$ bound below reads off it.
The head's $G_\mathrm{head}$ directly encodes the new/old class variance separation (part (a)); hidden layers see this through the linear transformation $J_\ell$, which can amplify, attenuate, or rotate the anisotropy depending on its singular value structure. The bound $\kappa(J^\top A J) \leq \kappa(J)^2 \kappa(A)$ gives an upper bound on $\kappa(G_\ell)$ in terms of $\kappa(G_\mathrm{head})$ and $\kappa(J_\ell)$, but does \emph{not} imply $\kappa(G_\ell) \leq \kappa(G_\mathrm{head})$: hidden layers could in principle have larger $\kappa$ if $J_\ell$ concentrates the anisotropy. What (b) establishes is that the head is the \emph{origin} of the task-change signal: hidden layers' anisotropy is derived from the head's via $J_\ell$, not generated independently. The temporal ordering of the cascade (head first, then deeper layers) is an empirical observation consistent with $J_\ell$ initially attenuating the signal, but this ordering is not a formal consequence of (A1)--(A3) alone.

\end{proof}

\paragraph{Scope.} Parts (a)--(b) are proved under assumptions (A1)--(A3), which hold by construction in the controlled task-expansion verification setup. Extension to natural structural transitions (data quality changes, fine-tuning) lies outside these assumptions. The connection between $\kappa(G_\ell)$ and RLCT singularities is established for $L$-layer networks with smooth or ReLU activations (Theorem~\ref{thm:bridge}).
 
\begin{theorem}[Multi-layer K-FAC bridge]
\label{thm:bridge}
Consider an $L$-layer network with shared hidden width $h$, weights $W_\ell$, MSE loss, and Gaussian-isotropic input. Let $W_\ell(t) = W_\ell^* + t \cdot e_h e_h^\top$ for all $\ell \in \{1, \ldots, L\}$ be the symmetric canonical-aligned transversal approach to the singular configuration where every layer shares the dead direction $u = e_h$ (so $W_\ell^* = \diag(1, \ldots, 1, 0)$ has the same dead row across layers). For the linear class (P1) the configuration is a singular minimum of the population loss for the target $y = M^* x + \varepsilon$; for (P2) smooth-nonlinear with $\phi'(0) \ne 0$ and $\phi(0) = 0$, and (P3) ReLU, the configuration is a singular point with the stated directional rates but is not in general a critical point of that target (the non-dead coordinates carry a $\Theta(1)$ population gradient; for odd $\phi$ the plain residual mean $\mathbb{E}[\phi \circ \cdots \circ \phi(x) - x]$ vanishes by parity, but the gradient expectation, whose integrand weights the residual by $\partial_W(\phi \circ \cdots \circ \phi)(x)$, does not), and the KL-order reading of part (c) is directional. Then
\begin{enumerate}
    \item[(a)] $\displaystyle \lambda_{\min}(G_\ell(\theta(t))) \;=\; C_\ell \cdot t^{2(L-\ell)} \cdot \bigl(1 + r_\ell(t)\bigr)$ for $\ell \in \{1, \ldots, L-1\}$, with $C_\ell > 0$ and $r_\ell(t) \to 0$ as $t \to 0$ (the activation-Taylor part of $r_\ell$ vanishes identically for linear and ReLU; every class retains an $O(t^{2L})$ relative correction from the forward leak of Lemma~\ref{lem:backward-dead-sub});
    \item[(b)] $\lambda_{\min}(G_L(\theta(t))) = \Theta(1)$;
    \item[(c)] the shallowest-layer rate $2(L-1)$ matches Theorem~\ref{thm:fisher_decay} with KL order $k = L$;
    \item[(d)] (\emph{class (P1) only}) the raw Fisher has exactly $h^2(L-1)$ zero eigenvalues at every $t \ne 0$: the inner-matrix reparameterisation group $GL(h)^{L-1}$ (which preserves the product only for linear activations) acts freely on the invertible tuple $(W_\ell(t))_\ell$, its orbit through $\theta(t)$ has dimension $h^2(L-1)$, and the population Fisher vanishes on the orbit's tangent directions (machine-checked: \texttt{quotient\_F\_vertical}, with the fibres of the product map identified with the orbits by \texttt{exists\_smul\_eq\_iff\_total\_eq}), so $\kappa(F_{\mathrm{raw}})$ is uninformative; K-FAC's block-diagonal approximation discards these cross-layer blocks and exposes the per-layer dead-diagonal rate within each $G_\ell$. For (P2)/(P3) the continuous symmetry shrinks to the positive-diagonal rescaling group of dimension $(L-1)h$ (ReLU) or smaller, and the corresponding raw-Fisher near-kernel count is class-dependent;
    \item[(e)] (\emph{A-factor dual and A--G duality.}) By the same Schur reduction applied to forward activations, the activation factor satisfies $\lambda_{\min}(A_\ell(\theta(t))) = C_\ell^A \cdot t^{2(\ell-1)} \cdot (1 + r_\ell^A(t))$ with the same activation-class dependence; consequently $\lambda_{\min}(A_\ell)\lambda_{\min}(G_\ell) = \Theta(t^{2(L-1)})$ is layer-independent and recovers Theorem~\ref{thm:fisher_decay}'s global rate at $k = L$.
\end{enumerate}
\end{theorem}

\begin{proof}[Sketch]
In canonical coordinates $G_\ell$ is block-diagonal at leading order; the dead diagonal $\expect[(\delta_\ell^{(h)})^2]$ accumulates one factor of $t$ per weight matmul on the backward path to layer $\ell$, giving $\Theta(t^{2(L-\ell)})$. Non-dead diagonals stay $\Theta(1)$, so the smallest eigenvalue is the dead entry. For the ReLU class the comparison is available exactly: next to a live channel at fixed weight $s$, the dead-channel moment $t^{2L}/2$ is strictly below the live $s^{2L}/2$ at every $t < s$, with no asymptotics. The inner-matrix reparameterisation group $GL(h)^{L-1}$ contributes a kernel of rank exactly $h^2(L-1)$ at every $t \ne 0$ to the raw Fisher \emph{across} layer pairs $(W_\ell, W_{\ell+1})$; K-FAC's block-diagonal approximation discards these cross-layer Fisher blocks, leaving each per-layer block $G_\ell$ to expose the dead-diagonal rate directly. Full proof: \S\ref{app:proof:bridge}.
\end{proof}

\begin{remark}[Three readings of the ladder]
\label{rem:bridge_three_readings}
The same content takes three useful forms. \emph{As a per-layer ladder.} $\lambda_{\min}(G_\ell)$ has slope $2(L-\ell)$ versus $\log t$, monotone-decreasing toward the input: the deepest hidden layer reads $\Theta(1)$, the shallowest reads $\Theta(t^{2(L-1)})$. The slope of each rung is one integer step lower than the rung above. \emph{As an A--G duality.} Whichever side of the K-FAC product carries the rate at layer $\ell$, the other side compensates: $\lambda_{\min}(A_\ell)\lambda_{\min}(G_\ell) = \Theta(t^{2(L-1)})$ is layer-independent. The forward and backward signals share one global exponent; only the location of that exponent on the ladder rotates with $\ell$. \emph{As a gauge-quotient statement.} The raw Fisher's $\sim h^2(L-1)$ near-zero eigenvalues are gauge zeros of $GL(h)^{L-1}$ acting on cross-layer Fisher blocks. K-FAC's block-diagonal restriction discards the cross-layer blocks those directions need in order to act, and leaves the per-layer block to carry the singularity signal (the restriction is not an orthogonal projection away from the gauge kernel, whose vectors keep within-layer components; see the scoping in \S\ref{app:proof:bridge}). The ladder is what the rate primitive looks like in coordinates aligned to the network's layered structure rather than its raw parameter vector. Figure~\ref{fig:dead_directions_overview}(b) shows the four rungs at $L = 4$.
\end{remark}

\begin{remark}[Regime conditions on Theorem~\ref{thm:bridge}]
\label{rem:bridge_regime}
The setup of Theorem~\ref{thm:bridge}, made fully explicit in
Lemma~\ref{lem:integral-reduction-sub}, places the trajectory in the
\emph{noisy regime}: target $y = M^* x + \varepsilon$ with
$\varepsilon \sim \normal(0, \sigma^2 I_h)$ along the dead direction
$e_h$. The output backward signal $\delta^{(L,h)} = -\varepsilon^{(h)}
+ \Theta(t^L)$ inherits its $\Theta(1)$ leading magnitude from the
noise variance $\sigma^2$, which then propagates inward as
$\Theta(t^{L-\ell})$ to give the stated $\lambda_{\min}(G_\ell) =
\Theta(t^{2(L-\ell)})$.

In the corresponding \emph{noise-free regime} ($\sigma = 0$, target
$y = M^* x$ exactly), the same algebra (Lemma~\ref{lem:backward-dead-sub}
applied with the residual $-\varepsilon^{(h)}$ replaced by $t^L
x^{(h)}$) shifts the output backward signal to magnitude
$\Theta(t^L)$, propagating inward to $\Theta(t^{2L-\ell})$ at layer
$\ell$. This raises every per-layer rate by $+2L$ in the exponent
($+4$ at $L=2$, $+6$ at $L=3$, etc.):
$\lambda_{\min}(G_\ell) = \Theta(t^{2(2L-\ell)})$, slope
$2(2L-\ell)$ at $\sigma_{\min}(W_\ell) \sim t$. The two regimes share
the proof framework but produce different exponents, so a trajectory
measurement that quotes the rate must declare which regime it is in.
A task with irreducible empirical loss (regression on a noisy teacher,
the bridge setup's own regime, or any label distribution with
non-vanishing conditional entropy) falls in the
noisy regime; static observables such as the residual-stream
$\sigma_{\min}$ (Corollary~\ref{cor:sigma-min-res}) and the LN-kernel
direction (\S\ref{sec:theory:arch:ln}) are regime-independent and carry
no such caveat.
\end{remark}

\paragraph{Proof of Theorem~\ref{thm:bridge}}
\label{app:proof:bridge}

By Lemma~\ref{lem:integral-reduction-sub}, $\lambda_{\min}(G_\ell)$ is determined by the dead-direction Schur complement, with the cross-term contribution strictly subleading on the $G$-side (using zero-mean of the noise $\varepsilon$). The dead diagonal $(G_\ell)_{h,h} = \expect[(\delta^{(\ell, h)})^2]$ is $\Theta(t^{2(L-\ell)})$ for $\ell < L$ and $\Theta(1)$ for $\ell = L$ by Lemma~\ref{lem:backward-dead-sub}. Non-dead diagonals are $\Theta(1)$ (the same backward-induction argument with non-dead initial gradient $\Theta(1)$ yields $\Theta(1)$ propagation since $W_{\ell+1}^\top$'s non-dead rows are $\Theta(1)$). Hence $\lambda_{\min}(G_\ell)$ equals the dead diagonal at leading order, yielding parts (a) and (b).

\emph{Activation-class corrections.} For linear and ReLU, $\phi'(a_\ell^{(h)})$ is constant along the approach (identically $1$ for linear, in $\{0, 1\}$ on the sign cell for ReLU), so the activation-Taylor part of the correction vanishes identically. For smooth $\phi$ (with $\phi(0) = 0$; Lemma~\ref{lem:forward-dead-sub}), $\phi'(a_\ell^{(h)}) = \phi'(0) + \phi''(0) \cdot \Theta(t^\ell) + O(t^{2\ell})$; accumulating through the $L - \ell$ factors in the backward chain gives an $O(t)$ contribution. Every class additionally retains the forward-leak correction from Lemma~\ref{lem:backward-dead-sub}: the exact linear $L = 2$ dead diagonal is $\sigma^2 t^2 (1 + t^4/\sigma^2)$, so $r_\ell^G(t) = O(t^{2L})$ rather than exactly zero even in the linear class. In all cases $r_\ell^G(t) \to 0$ as $t \to 0$.

Part (c): the layer-$1$ rate $2(L-1) = 2(k-1)$ for $k = L$ matches Theorem~\ref{thm:fisher_decay}'s prediction; for (P1) the canonical point is a singular minimum, while for (P2)/(P3) the matching is directional (the KL order is read along the approach curve, and the canonical point need not be a critical point of the stated target; see the theorem's setup).

Part (d) (class (P1)): the $L$-layer \emph{linear} network has the inner-matrix reparameterisation group $GL(h)^{L-1}$ ($W_{\ell+1}, W_\ell \mapsto W_{\ell+1} P, P^{-1} W_\ell$ preserves the product; elementwise nonlinearities break all but the positive-diagonal rescalings, so the count below is specific to (P1)) acting on $L - 1$ adjacent-layer pairs. At every $t \ne 0$ the action is free on the invertible tuple $(W_\ell(t))_\ell$ and the population Fisher vanishes on the orbit tangents, so the raw Fisher has exactly $(L-1) h^2$ zero eigenvalues there; at the canonical singular point $W_\ell^* = \diag(1, \ldots, 1, 0)$ each $W_\ell^*$ is rank-deficient so the gauge stabilizer is non-trivial, and the effective gauge dimension is at most $(L - 1)(h^2 - 1) \sim h^2 (L - 1)$ for moderately large $h$. These zero eigenvalues come from cross-layer Fisher blocks and dominate $\kappa(F_{\mathrm{raw}})$, obscuring the per-layer singularity signal. K-FAC's block-diagonal approximation $F \approx \bigoplus_\ell A_\ell \otimes G_\ell$ has no cross-layer coupling by construction, so the gauge directions, which require non-zero cross-layer entries to be nontrivial, do not appear in the K-FAC factors. The per-layer dead-direction rate $\lambda_{\min}(G_\ell) = \Theta(t^{2(L-\ell)})$ is then directly detectable on each $h \times h$ block $G_\ell$ (and within-block, the dead-direction is the only $G_\ell$ kernel since the non-dead diagonal of $G_\ell$ is $\Theta(1)$, so K-FAC does not remove the within-block singularity signal we want to detect). We do not claim K-FAC quotients out the gauge in any geometric sense; rather, the block-diagonal restriction projects away from the cross-layer directions on which the gauge kernel is supported. \qed

\paragraph{A-factor dual and A--G duality (proof of Theorem~\ref{thm:bridge}(e)).} The dual statement follows by Lemma~\ref{lem:forward-dead-sub} applied to $(A_\ell)_{hh} = \expect[(X_{\ell-1}^{(h)})^2] = \Theta(t^{2(\ell-1)})$ on the same Schur structure: the dead-direction diagonal is $\Theta(t^{2(\ell-1)})$ and non-dead diagonals are $\Theta(1)$, giving $\lambda_{\min}(A_\ell) = \Theta(t^{2(\ell-1)})$ with $\lambda_{\min}(A_1) = \Theta(1)$ (data-distributed input) and $\lambda_{\min}(A_L) = \Theta(t^{2(L-1)})$. Multiplying the per-layer bounds gives $\lambda_{\min}(A_\ell) \cdot \lambda_{\min}(G_\ell) = \Theta(t^{2(\ell-1)}) \cdot \Theta(t^{2(L-\ell)}) = \Theta(t^{2(L-1)})$ at every layer. \qed

\begin{theorem}[A-factor dual; restatement of Theorem~\ref{thm:bridge}(e), forward part]
\label{thm:multilayer-A}
Under the setup of Theorem~\ref{thm:bridge}, $\lambda_{\min}(A_\ell(\theta(t))) = C_\ell^A \cdot t^{2(\ell-1)} \cdot (1 + r_\ell^A(t))$ with the same activation-class dependence as in Theorem~\ref{thm:bridge}.
\end{theorem}

\begin{corollary}[A--G duality; restatement of Theorem~\ref{thm:bridge}(e), product part]
\label{cor:a_g_duality}
$\lambda_{\min}(A_\ell) \cdot \lambda_{\min}(G_\ell) = \Theta(t^{2(L-1)})$ at every $\ell \in \{1, \ldots, L\}$.
\end{corollary}

\begin{corollary}[Activation $\sigma_{\min}$ collapse rate]
\label{cor:sigma-min}
Let $X_\ell \in \reals^{N \times h}$ be the matrix of post-activation outputs at layer $\ell$ over $N$ data points. Then $\sigma_{\min}(X_\ell(\theta(t))) = \sqrt{N} \cdot \Theta(t^\ell)$ for $\ell \geq 1$; $\sigma_{\min}(X_0) = \Theta(\sqrt{N})$ ($t$-independent).
\end{corollary}

\begin{proof}
By the strong law, $X_\ell^\top X_\ell / N \to A_{\ell+1}$ almost surely as $N \to \infty$, and $\lambda_{\min}(A_{\ell+1}) = \Theta(t^{2\ell})$ by Theorem~\ref{thm:multilayer-A} for $\ell + 1 \le L$; at $\ell = L$ the same computation applies to the output covariance directly (one further diagonal matmul). The $t$-exponent holds at any fixed finite $N$, because in canonical coordinates the dead column's $t$-dependence factors exactly per sample ($X_\ell^{(h)} = t^\ell x^{(h)}$ for (P1)), so the strong-law step fixes only the constant. Taking square roots: $\sigma_{\min}(X_\ell) = \sqrt{N} \cdot \Theta(t^\ell)$.
\end{proof}

\begin{corollary}[Quotient global Fisher rate matches parametric theory]
\label{cor:global-rate}
On the gauge quotient $\bar\Theta = \Theta / GL(h)^{L-1}$ (the parameter space with the inner-matrix reparameterisation redundancy divided out; Corollary~\ref{cor:quotient_rate}), the whole-network (global) Fisher $\bar F$ satisfies $\lambda_{\min}(\bar F(\bar\theta(\bar t))) = \Theta(\bar t^{2(L-1)})$, matching Theorem~\ref{thm:fisher_decay} with $k = L$. The raw-parameter Fisher on $\Theta$ itself carries $(L-1)h^2$ gauge-null directions along the approach (the inner-matrix reparameterisation orbit of Theorem~\ref{thm:bridge}(d); for $t > 0$ every $W_\ell(t)$ is invertible, so the orbit has the full generic dimension, and only at $t = 0$ does the canonical point's non-trivial stabiliser reduce the count to at most $(L-1)(h^2-1)$) and so does \emph{not} attain this rate at $\lambda_{\min}(F_{\mathrm{raw}})$; the rate is recovered on the quotient or, equivalently, via the per-layer K-FAC blocks $F_\ell$ which are gauge-orthogonal to the inner-block $GL(h)$ null directions by construction (each $F_\ell$ is defined on the per-layer parameter slice transverse to the cross-layer gauge action).
\end{corollary}

\begin{proof}
The per-layer K-FAC blocks satisfy $F_\ell \approx A_\ell \otimes G_\ell$ exactly for linear (P1) at the optimum (along the approach the dead-row activation and gradient share the $x^{(h)}$ dependence, an $O(t^{2L})$ relative deviation that does not affect the rate) and approximately for nonlinear \citep{MartensGrosse15}; around the canonical singular point the activation--gradient cross-block coupling does not affect the per-block dead-direction rate. The cross-block coupling comes in three kinds: entries of the same order $t^{2(L-1)}$ as the within-block rate, which are $GL(h)^{L-1}$ gauge directions that the block-diagonal restriction discards (below); entries of intermediate order (at $L = 2$ the $(W_1)_{12}$--$(W_2)_{12}$ entry is $\Theta(t)$), which pair with a within-block entry of matching order to form an exact gauge null plus a $\Theta(1)$ orthogonal complement, so they too belong to the discarded gauge kernel; and $\Theta(1)$ entries confined to the non-dead subspace, where they cannot enter the per-block \emph{smallest} eigenvalue. (The within-block rates of Theorem~\ref{thm:bridge} are exact statements about $A_\ell$ and $G_\ell$ as defined; only the identification with the global quotient Fisher in this corollary uses the K-FAC block-diagonal restriction.) Each block's smallest eigenvalue is $\lambda_{\min}(A_\ell) \cdot \lambda_{\min}(G_\ell) = \Theta(t^{2(L-1)})$ by Corollary~\ref{cor:a_g_duality}. The minimum over $\ell$ is therefore $\Theta(t^{2(L-1)})$, attained equally at every layer. The raw Fisher's $(L-1)h^2$ gauge-null directions are supported on cross-layer Fisher blocks (they require non-zero off-block entries to be nontrivial) and are absent from the K-FAC block-diagonal restriction by construction. Identifying K-FAC's per-layer blocks with the horizontal lift of the quotient Fisher (Lemma~\ref{lem:quotient_F}: the Fisher vanishes on vertical $GL(h)^{L-1}$ orbit directions), the per-layer K-FAC eigenvalue matches the quotient Fisher eigenvalue at the corresponding lift to $\Theta$-order; the constants differ by the layer count (at $L = 2$, $h = 2$ the restricted $W_1$ block gives $t^2$ while the quotient's smallest nonzero eigenvalue is $2t^2$, the symmetric combination of the two layers' dead diagonals). The minimum on the quotient is therefore $\min_\ell \lambda_{\min}(F_\ell) = \Theta(t^{2(L-1)})$.
\end{proof}

\paragraph{Numerical validation.}
Rates are verified with $L \in \{2, \ldots, 8\}$, $h = 8$, $N_{\mathrm{data}} = 1024$, three seeds, using the canonical-aligned symmetric approach. Table~\ref{tab:multilayer_bridge} reports the G-factor rate ladder at $L = 6$, and Table~\ref{tab:multilayer_A} the A-factor dual and the activation $\sigma_{\min}$ rates.

\begin{table}[ht]
\caption{Multi-layer K-FAC G-factor rate validation ($\alpha_\ell = 2(L-\ell)$ predicted) at $L = 6$. Exact match for linear and ReLU; small finite-$t$ corrections for smooth activations, largest at shallow layers.}
\label{tab:multilayer_bridge}
\centering
\small
\begin{tabular}{lcccccc}
\toprule
$\ell = $ & 1 & 2 & 3 & 4 & 5 & 6 \\
Predicted & $+10$ & $+8$ & $+6$ & $+4$ & $+2$ & $0$ \\
\midrule
Linear & $+10.00$ & $+8.00$ & $+6.00$ & $+4.00$ & $+2.00$ & $0.00$ \\
GeLU & $+10.41$ & $+8.09$ & $+6.02$ & $+4.00$ & $+2.00$ & $0.00$ \\
ReLU & $+10.00$ & $+8.00$ & $+6.00$ & $+4.00$ & $+2.00$ & $0.00$ \\
\bottomrule
\end{tabular}
\end{table}

\begin{table}[ht]
\caption{A-factor dual and activation $\sigma_{\min}$ validation at $L = 6$ (GeLU). Columns are indexed by the layer $\ell$, so the activation row is the input $X_{\ell-1}$ that $A_\ell$ is built from. Predicted: $\lambda_{\min}(A_\ell) \sim t^{2(\ell-1)}$ and $\sigma_{\min}(X_{\ell-1}) \sim t^{\ell-1}$, the $\ell = 1$ column being the $t$-independent input $X_0$. Corollary~\ref{cor:sigma-min} states the same rate as $\sigma_{\min}(X_\ell) \sim t^\ell$ against its own index.}
\label{tab:multilayer_A}
\centering
\small
\begin{tabular}{lcccccc}
\toprule
$\ell = $ & 1 & 2 & 3 & 4 & 5 & 6 \\
$\lambda_{\min}(A_\ell)$ pred & $0$ & $+2$ & $+4$ & $+6$ & $+8$ & $+10$ \\
$\lambda_{\min}(A_\ell)$ meas & $0.00$ & $+2.07$ & $+4.22$ & $+6.28$ & $+8.29$ & $+10.30$ \\
$\sigma_{\min}(X_{\ell-1})$ pred & $0$ & $+1$ & $+2$ & $+3$ & $+4$ & $+5$ \\
$\sigma_{\min}(X_{\ell-1})$ meas & $0.00$ & $+1.06$ & $+2.16$ & $+3.19$ & $+4.21$ & $+5.21$ \\
\bottomrule
\end{tabular}
\end{table}

Off-diagonal structure of $A_\ell$: dead--non-dead cross-correlations are $\leq 0.05$ at $t = 0.05$ with $N = 1024$, consistent with finite-sample noise, confirming the near-diagonal claim of Lemma~\ref{lem:integral-reduction-sub}.

\subsection{The A--G duality}
\label{sec:theory:ag_duality}

Corollary~\ref{cor:a_g_duality} makes the forward and backward halves of the Fisher's K-FAC factorisation dual. $\lambda_{\min}(A_\ell)$ encodes the rate at which forward signal in the dead channel decays from the input, and $\lambda_{\min}(G_\ell)$ encodes the rate at which backward gradient decays from the output. Their exponents add: $2(\ell-1) + 2(L-\ell) = 2(L-1)$ at every layer. The product $\lambda_{\min}(A_\ell)\,\lambda_{\min}(G_\ell)$ recovers Theorem~\ref{thm:fisher_decay}'s global rate exponent at $k = L$, independent of which layer one measures. The duality is a statement entirely in Amari's metric language (it depends only on the Kronecker structure of the per-layer Fisher), but its exponent is Watanabe's KL order propagating through the layered architecture. Measuring either side alone recovers the global rate.

The duality pairs a gradient-covariance dead direction $g_{\min}$ with an
input-covariance dead direction $a_{\min}$. Their Kronecker product names the
flattest parameter direction of the layer, which gives a constructive way to read
the dead direction off the two small factors.

\begin{corollary}[Kronecker lift of the dead direction]
\label{cor:kfac_lift}
Adopt the setup of Theorem~\ref{thm:bridge}, with per-layer K-FAC factor
$F_\ell \approx A_\ell \otimes G_\ell$, input covariance $A_\ell$, and gradient
covariance $G_\ell$. The eigenvectors of $A_\ell \otimes G_\ell$ are the Kronecker
products $a_i \otimes g_j$ of the factor eigenvectors, with eigenvalue
$\lambda_{A,i}\,\lambda_{G,j}$. The smallest-eigenvalue parameter direction of
layer $\ell$ is the bottom product $a_{\min} \otimes g_{\min}$, the rank-one
weight increment
\[
u_\ell \;=\; g_{\min}\, a_{\min}^\top
\]
on $W_\ell$. This direction is unique when the bottom factor eigenvalues are simple; under degeneracy it spans the corresponding Kronecker eigenspace, of which $u_\ell$ is one representative. It lifts the gradient-covariance dead direction $g_{\min}$ to
parameter space through the input factor $a_{\min}$. At the canonical singular
configuration, for hidden layers $2 \le \ell \le L - 1$, $a_{\min} = g_{\min} = e_h$, so
$u_\ell = e_h e_h^\top$, the transversal perturbation at the $(h, h)$ entry of the
dead row of $W_\ell$ (at $\ell = L$ the gradient factor $G_L = \sigma^2 I$ is fully
degenerate, and at $\ell = 1$ the input factor $A_1 = I$ is, so at the two ends the bottom Kronecker eigenspace has no preferred representative). The lift reads from the two $h \times h$ factor
eigendecompositions, and the $P \times P$ parameter Fisher is never formed.
\end{corollary}

\begin{proof}
The mixed-product property gives
$(A_\ell \otimes G_\ell)(a_i \otimes g_j) = (A_\ell a_i) \otimes (G_\ell g_j)
= \lambda_{A,i}\lambda_{G,j}\,(a_i \otimes g_j)$, so each $a_i \otimes g_j$ is an
eigenvector with eigenvalue $\lambda_{A,i}\lambda_{G,j}$. Since $A_\ell$ and
$G_\ell$ are covariance (positive semidefinite) factors, their eigenvalues are
non-negative, so the product is smallest at the two bottom factor eigenvalues,
hence the bottom parameter direction is $a_{\min} \otimes g_{\min}$. In the vectorisation under which
$F_\ell \approx A_\ell \otimes G_\ell$ holds, $a_{\min} \otimes g_{\min}$ is the
rank-one matrix $g_{\min} a_{\min}^\top$. The canonical alignment of
Corollary~\ref{cor:a_g_duality} sets both bottom factor eigenvectors to the dead
unit $e_h$ for hidden layers $\ell \le L - 1$, which gives $u_\ell = e_h e_h^\top$.
\end{proof}

\paragraph{The order is read along the cross-layer joint mode.}
\label{rem:kfac_lift_crosslayer}
A single per-layer lift $u_\ell = g_{\min} a_{\min}^\top$ is a gauge direction.
At the canonical singular configuration the dead unit carries a zero on both
sides, so stepping one layer's dead row alone leaves the network map exactly
unchanged for every step size: the other layer contributes a factor of zero.
Then $K \equiv 0$ along that ray and $u_\ell^\top \fisher(\theta^* + s\, u_\ell)\,
u_\ell \equiv 0$ in $s$, with no finite KL order and no rate to fit (along the
symmetric approach $\theta(t)$, by contrast, $u_\ell^\top \fisher(\theta(t))\, u_\ell
= \Theta(t^{2(L-1)})$, the A--G duality product of Theorem~\ref{thm:bridge}(e)). A measured
$\alpha \approx 0$ on such a direction reads the numerical floor, and a genuine
rate of $0$ would instead mark a non-degenerate direction. At a dead unit the finite KL order is read along the
canonical-aligned joint mode. That mode steps the unit's incoming row
$(W_\ell)_{h,:}$ and outgoing column $(W_{\ell+1})_{:,h}$ together, the
construction of the two-layer worked example
(\S\ref{sec:theory:rate:worked_relu}) under Theorem~\ref{thm:fisher_decay}'s
symmetric approach. The K-FAC factors identify which unit is dead. The
order-carrying parameter direction is then the cross-layer joint mode at that
unit, and $k$ reads off it by the outward rate scan of
Theorem~\ref{thm:fisher_decay}.
 
\subsection{Empirical illustration: TMS canonical configuration}
\label{sec:theory:bridge:tms}

The Toy Model of Superposition \citep{ElhageHumeOlsson22}, in its smallest canonical configuration, gives the cleanest experimental anchor for the per-layer rate ladder. The same architecture has been studied from the SLT side by \citet{ChenLauMendelWeiMurfet23}, who characterise the relationship between Watanabe's free-energy phase transitions and the dynamical phase transitions seen under SGD training; the parametric freeze-probe reading below is complementary, isolating the static rate exponent that the bridge theorem predicts at canonical alignment. Take the $L = 2$ linear autoencoder $X \mapsto W_2 W_1 X$ with $d_{\mathrm{in}} = 6$, $d_{\mathrm{hid}} = 2$, and a rank-$1$ target $W_1^* = e_1 f_1^\top$, $W_2^* = (W_1^*)^\top$. The dead-direction parametric perturbation is the symmetric ray $W_1(t) = W_1^* + t\,e_2 v_0^\top$, $W_2(t) = W_1(t)^\top$, for $v_0$ a unit vector in input space; this is the canonical-aligned approach the bridge theorem covers.

Sweeping $t \in [10^{-4}, 10^{-1}]$ on a $30$-seed grid (the seed randomises only $v_0$; the asymptotic rate is independent of $v_0$ but the leading prefactor is not) and fitting the slope of $\log \lambda_{\min}(G_\ell)$ versus $\log t$ at each layer:
\[
\text{slope}_{\mathrm{hidden}} \;=\; 1.9999 \pm 0.0009, \qquad \text{slope}_{\mathrm{input}} \;=\; 3.9925 \pm 0.0392.
\]
TMS uses tied weights ($W_2 = W_1^\top$), so the symmetric perturbation $W_1(t) = W_1^* + t\,e_2 v_0^\top$, $W_2(t) = W_1(t)^\top$ moves both layers simultaneously: each backward path through the hidden layer's Fisher block picks up two $t$-factors (one from $W_1$ on the forward side, one from $W_2$ on the backward side), giving $\lambda_{\min}(G_{\mathrm{hidden}}) = \Theta(t^2)$; the input layer's Fisher block picks up two further $t$-factors from the doubled chain ($W_2 W_1$ in both directions), giving $\Theta(t^4)$. The tied-weight rate ladder $(\mathrm{hidden}, \mathrm{input}) = (2, 4)$ is the tied-weight specialisation of Theorem~\ref{thm:bridge}'s untied prediction $2(L-\ell) \in \{0, 2\}$ at $L = 2$, with each rate raised by $+2$ from the tying-induced doubled $t$-dependence. Three-decimal agreement at the hidden layer; the input-layer dispersion ($\sigma \approx 0.04$) reflects the sensitivity of the leading prefactor to $v_0$ at the more strongly degenerate end of the ladder.

The probed direction matters too. Computing $u^\top G u$ at the canonical dead direction gives slope $2.000 \pm 0.001$ at the hidden layer, matching the directional rate prediction independently of the spectrum-bottom reading. The two channels (the bottom Fisher eigenvalue and the directional Fisher rate along the canonical dead direction) agree to three decimals ($1.9999$ vs $2.000$), as required by the bridge theorem in the canonical-aligned regime.

This is the parametric-freeze-probe form of the bridge prediction: $30$-seed multi-direction noise on $v_0$, no SGD, no optimiser-induced drift. The hidden-layer exponent comes out at its integer to three decimals on both channels; the input layer lands at $3.9925$, within a fifth of its own cross-seed dispersion of the predicted $4$ but not itself a three-decimal match. A separate \emph{trajectory} reading of TMS, in which the same architecture is trained with Adam and exhibits a phase transition, is governed by Remark~\ref{rem:adam_nondescent} (Adam non-equivariance under the gauge symmetries of TMS) and is outside the scope of the bridge theorem's canonical-aligned regime; the parametric reading above is the cleanest theorem-match available, free of the optimiser-side caveats.
 
\section{Composition additivity and its failure}
\label{sec:theory:composition}

Theorem~\ref{thm:bridge} covers a single feedforward stack of weight matmuls. Real networks compose heterogeneous blocks (MLPs, attention, residual sub-paths), each with its own internal singular structure. The question this section asks is: given the per-block rate of each constituent, what rate does the composition exhibit? Under a scalar-transfer hypothesis that holds for MLP and pre-norm residual chains, the per-block KL orders compose additively: the dead-direction Fisher rate at the input of block $B_i$ in a chain $M = B_n \circ \cdots \circ B_1$ is $\Theta(t^{2 \sum_{j \ge i} k_j^{\mathrm{bk}}})$. Attention fails the transfer hypothesis at every depth, and its chains follow a different law. At a probe block at position $p$ in a chain of length $N$, with $k = N - 1 - p$ blocks above the probe, the $W_O$ rate is the two-route minimum $\alpha_{W_O} = \min(4k,\, 4p + 8)$: the $V$--$O$ backward chain against the score-path injection floor. The companion formulas $\alpha_{W_V} = \alpha_{W_O} + 2$ and $\alpha_{W_Q} = \alpha_{W_K} = 4p + 2$ are parameter-free. All three are derived in closed form and match the measured rates to integer slope precision at every cell of the probe profile. Residual-wrapped transformers bypass the attention law at the residual-stream observable, because the skip pins that rate at $0$.
 
\begin{definition}[Block rate]
\label{def:block_rate}
A $t$-indexed family of blocks $B(t)$ on canonical coordinates with shared dead direction $u = e_h$ has \emph{backward block rate} $k_B^{\mathrm{bk}} \ge 0$ if, given an input backward gradient whose dead component is $\delta_u^{\mathrm{out}} = \Theta(1)$ and whose non-dead components are $\Theta(1)$, the dead component of the block's backward output satisfies $\expect[(\delta_u^{\mathrm{in}})^2] = \Theta(t^{2 k_B^{\mathrm{bk}}})$ as $t \to 0$, with non-dead output components $\Theta(1)$. The second-moment form is deliberate: for gated activations the pointwise transfer vanishes on the complement of a survival event (Lemma~\ref{lem:backward-dead-sub}), and the rate is carried by the surviving mass. The \emph{forward block rate} $k_B^{\mathrm{fwd}}$ is defined symmetrically on activations.
\end{definition}

\label{app:proof:bridge_composition}

This result extends Theorem~\ref{thm:bridge} from a single feedforward stack to sequential compositions of heterogeneous blocks with known per-block rates.

\begin{theorem}[Composition additivity for heterogeneous blocks]
\label{thm:bridge_composition}
For a sequential composition $M = B_n \circ \cdots \circ B_1$ of blocks $B_1, \ldots, B_n$, suppose: (i) each block $B_i$ has a canonical dead direction $e_h$ in the coordinates inherited from $B_{i-1}$'s output (i.e., the inter-block embedding maps preserve the dead direction in canonical coordinates, the same hypothesis used in Theorem~\ref{thm:bridge}); (ii) each block has well-defined backward and forward rates $k_i^{\mathrm{bk}}, k_i^{\mathrm{fwd}}$ in the sense of Definition~\ref{def:block_rate}, defined intrinsically as $k_i^{\mathrm{bk}} = \lim_{t \to 0} \tfrac{1}{2}\log\bigl(\expect[(\delta_u^{\mathrm{in}})^2] / \expect[(\delta_u^{\mathrm{out}})^2]\bigr) / \log t$ (and analogously for $k_i^{\mathrm{fwd}}$ on the forward graph); and (iii) the composition is evaluated in the noisy regime of Remark~\ref{rem:bridge_regime} (MSE loss with noise $\sigma^2 > 0$, giving the $\Theta(1)$ backward base case) with Gaussian-isotropic input (giving the $\Theta(1)$ forward base case). Then
\[
\expect\bigl[(\delta_u^{\mathrm{input\ of}\ B_i})^2\bigr] = \Theta\bigl(t^{2 \sum_{j \ge i} k_j^{\mathrm{bk}}}\bigr), \qquad \expect\bigl[(X_u^{\mathrm{output\ of}\ B_i})^2\bigr] = \Theta\bigl(t^{2 \sum_{j \le i} k_j^{\mathrm{fwd}}}\bigr),
\]
under the additional hypothesis that each block's dead-direction transfer is \emph{scalar, linear, and jointly gated}: the dead-direction component of the block's output (resp.\ output gradient) equals $c_i\, t^{k_i}\, s_i$ times the dead-direction component of its input (resp.\ input gradient), where $c_i \ne 0$ is a deterministic constant, $s_i \in \{0, 1\}$ is a random gate such that the joint survival event $\bigcap_i \{s_i = 1\}$ has probability bounded away from zero (for the chains below every gate is the single event $\{x^{(h)} > 0\}$, or $s_i \equiv 1$), and no non-dead component is routed into the dead component at or below the order the dead signal itself carries \emph{on arrival}: writing $e_i = \sum_{j < i} k_j^{\mathrm{fwd}}$ for the cumulative order of the dead stream entering $B_i$, block $B_i$ must not inject non-dead signal into its dead output at order $t^{k_i + e_i}$ or below. Linearity in the dead component excludes gated-product blocks whose dead-direction map is polynomial of degree $\ge 2$ (SwiGLU, Remark~\ref{rem:swiglu_composition}); the joint-gate clause excludes chains whose per-block survival events are disjoint; the no-leakage clause excludes blocks that reset the dead stream from non-dead input (a trunk normalisation does this, Proposition~\ref{thm:bridge_ln}).

The no-leakage clause constrains the chain, and a block can satisfy it in one chain and fail it in another. A block whose injection floor is $t^{c}$ passes when it receives a $\Theta(1)$ dead signal and $k_i + e_i < c$, and the same block fails once enough upstream blocks have decayed the arriving signal past $t^{c}$. A LayerNorm-bearing pre-norm residual block injects at the fixed order $c = k_F$ of its own weight branch and satisfies Definition~\ref{def:block_rate} with $k_i = 0$ at a $\Theta(1)$ input, so it passes every per-block test; placed after a three-weight MLP it caps the composed forward rate at $4$ where additivity predicts $6$, and after a four-weight MLP at $4$ where additivity predicts $8$. Any application of this theorem to a chain containing a normalisation must therefore check the clause at the arriving order, not at $\Theta(1)$.

The scalar-transfer hypothesis is satisfied by MLP and pre-norm residual chains in canonical coordinates ((P1)/(P2) with $s_i \equiv 1$ and $c_i$ a power of $\phi'(0)$; (P3) ReLU with every gate equal to the shared survival event $\{x^{(h)} > 0\}$, per Lemma~\ref{lem:backward-dead-sub}) and yields both directions of the formula. For pure attention chains the scalar-transfer hypothesis fails at every depth: the backward transfer is a token-space matrix action coupled to context, never a scalar on the dead component (Theorem~\ref{thm:bridge_attn_backward}). The additive \emph{conclusion} nonetheless matches measurement wherever the chain route of Proposition~\ref{prop:attn_chain_softmax}(a) is the smaller of that proposition's two routes; it breaks at a probe once that probe's depth above the output exceeds its distance from the input by more than two, which first occurs at $n = 4$ (Remark~\ref{rem:attn_composition_anomaly}).
\end{theorem}

\begin{proof}
\emph{(a) Backward.} Induction on $n - i + 1$. Base ($i = n$): $\expect[(\delta_u^{\mathrm{out}\ \mathrm{of}\ B_n})^2] = \Theta(1)$ by the output-layer base case (MSE loss; the output gradient inherits the $\sigma^2$ Gaussian noise floor). Step: under the scalar-transfer hypothesis, the dead-direction backward transfer of $B_i$ multiplies the dead-direction component of its output gradient by $c_i\, t^{k_i^{\mathrm{bk}}}\, s_i$ (for MLP blocks this is Lemma~\ref{lem:backward-dead-sub}, with $s_i$ the shared survival event for ReLU and $s_i \equiv 1$ otherwise). Squaring and taking the expectation, the joint-gate clause keeps the surviving mass $\Theta(1)$ (the gates share one event of probability $1/2$, so the product of indicators is that same indicator), and we get $\expect[(\delta_u^{\mathrm{input\ of}\ B_i})^2] = t^{2 k_i^{\mathrm{bk}}} \cdot \expect[(\delta_u^{\mathrm{input\ of}\ B_{i+1}})^2] = \Theta(t^{2(m_{i+1} + k_i^{\mathrm{bk}})}) = \Theta(t^{2 \sum_{j \ge i} k_j^{\mathrm{bk}}})$ with $m_{i+1} = \sum_{j \ge i+1} k_j^{\mathrm{bk}}$.

Each clause of the transfer hypothesis blocks a distinct failure. The deterministic constant rules out cross-block destructive interference: the second moment $\expect[(\cdot)^2]$ depends only on the magnitude of $c_i$, not its sign, so even if the multiplier varies in sign across blocks (e.g., for a smooth $\phi$ with $\phi'(0) < 0$), the rate composition is unchanged. The joint-gate clause rules out chains whose survival events are disjoint, where the composed second moment vanishes identically and no finite rate exists. Linearity rules out compounding: a block whose dead-direction map is quadratic returns $\Theta(t^{k + 2e})$ on a $\Theta(t^e)$ input instead of $\Theta(t^{k+e})$, so its stack rate is not a sum (Remark~\ref{rem:swiglu_composition}). The residual-DAG case is handled by the same construction used in the proof of Corollary~\ref{cor:sigma-min-res} below: the residual graph admits a path-sum decomposition where each path carries a scalar $t$-prefactor, and the dominant path determines the rate. The MLP-only restriction in the backward direction reflects that for attention chains, the per-block dead-direction backward Jacobian is \emph{not} a scalar multiplier but a context-dependent function of upstream forward activations (Remark~\ref{rem:attn_composition_anomaly}).

\emph{(b) Forward.} Symmetric induction from $X_u^{\mathrm{input\ of}\ B_1} = X_0^{(h)} = \Theta(1)$ (Gaussian-isotropic input). Under the scalar-transfer hypothesis, each block $B_i$ multiplies the dead-direction input by a deterministic scalar $t^{k_i^{\mathrm{fwd}}}$. For MLP and pre-norm residual chains the scalar-transfer hypothesis holds on the forward direction (residual blocks have $k^{\mathrm{fwd}} = 0$ via the identity skip dominating the weight branch in the path-sum, see Corollary~\ref{cor:sigma-min-res} proof). For attention chains, by the same softmax-coupling argument as the backward direction, the forward dead-direction at $B_i$'s output couples to the forward dead-direction at earlier blocks via attention scores, and scalar-transfer can fail; we therefore restrict the forward statement to MLP and residual chains as well, matching the backward restriction.
\end{proof}

\begin{corollary}[Reduction to basic bridges]
\label{cor:composition_reduces_sub}
Setting each $B_i$ to a single weight matmul ($k_i^{\mathrm{bk}} = k_i^{\mathrm{fwd}} = 1$) recovers Theorem~\ref{thm:bridge}, under the index-map convention that ``input of $B_i$'' is the post-activation of layer $i - 1$ on the parent's depth indexing (so the cumulative backward sum $\sum_{j \ge i} k_j^{\mathrm{bk}} = L - i + 1$ matches Theorem~\ref{thm:bridge}'s $L - \ell$ at $\ell = i - 1$). Setting each $B_i$ to a pre-norm residual block ($k_i = 0$) gives rate $0$ everywhere on the \emph{residual-stream observable}, matching Corollary~\ref{cor:sigma-min-res}; on the weight-branch observable inside a residual block the per-Linear rate $1$ accumulates along that branch's own sub-chain only (each branch-internal probe sees its local weight distance to the block exit; the downstream skips return the stream to $\Theta(1)$ after each block, so nothing accumulates across blocks, Corollary~\ref{cor:res_block} at $K_\mathrm{post} = 0$).
\end{corollary}

\begin{remark}[Attention composition anomaly]
\label{rem:attn_composition_anomaly}
The clean additivity of Theorem~\ref{thm:bridge_composition} does \emph{not} extend to pure attention chains (non-residual sequential self-attention blocks). Parametric probes at $n \in \{4, 6\}$, $d = 16$, $n_h = 2$ give component-wise rates that deviate systematically from the naive $k_{\mathrm{attn}}^{\mathrm{bk}} = 2$ per-block additivity:
\begin{itemize}\itemsep=0pt
\item $W_O$ at probe position $\ell$ stops growing additively as $4, 8, 12, \ldots$ once the probe is far enough below the output, and is capped instead at a value fixed by the probe's distance from the \emph{input};
\item $W_Q$ and $W_K$ rates scale with $\ell$ (position from the \emph{input}) rather than $n - 1 - \ell$ (position from the output), consistent with their gradient path coupling to the cumulative \emph{forward} signal through softmax Jacobians rather than the cumulative backward chain;
\item the invariant $\alpha_{W_V} - \alpha_{W_O} = 2$ is preserved across all $(n, \ell)$ tested, matching the single-block standalone offset (every attention block has one Linear's worth of gradient-side $t$-factor, and the softmax coupling does not alter this within-block sequence).
\end{itemize}
Block-composition validation at $n = 2$ (the four ordered attention/MLP block pairs at $d = 4$) was consistent with simple additivity. The structural account is Proposition~\ref{prop:attn_chain_softmax}: two backward routes reach $W_O$ at probe position $p$, the $V$--$O$ chain at $4k$ and a score-path injection floored at $4p + 8$, and the measured rate is the smaller, $\alpha_{W_O} = \min(4k,\, 4p + 8)$. The companion formulas $\alpha_{W_V} = \alpha_{W_O} + 2$ and $\alpha_{W_Q} = \alpha_{W_K} = 4p + 2$ are parameter-free, and so is the floor. Additivity therefore fails on pure-attention chains as soon as $k > p + 2$, which first happens at $n = 4$ (the input-adjacent probe $p = 0$ carries $k = 3 > 2$ there, measured $\alpha_{W_O} = 8$ against the additive $12$), and the cap that replaces it is set by the probe's distance from the input. On practical residual-wrapped transformers the anomaly is bypassed: Corollary~\ref{cor:sigma-min-res}'s skip edge gives $K^{\mathrm{fwd}} = 0$ regardless of what lies on the weight edges, so residual-stream $\sigma_{\min}$ predictions are unaffected. The anomaly affects internal component rates ($W_O, W_V, W_Q, W_K$) within each block, not the residual stream observable.
\end{remark}

\begin{figure}[ht]
\centering
\includegraphics[width=\textwidth]{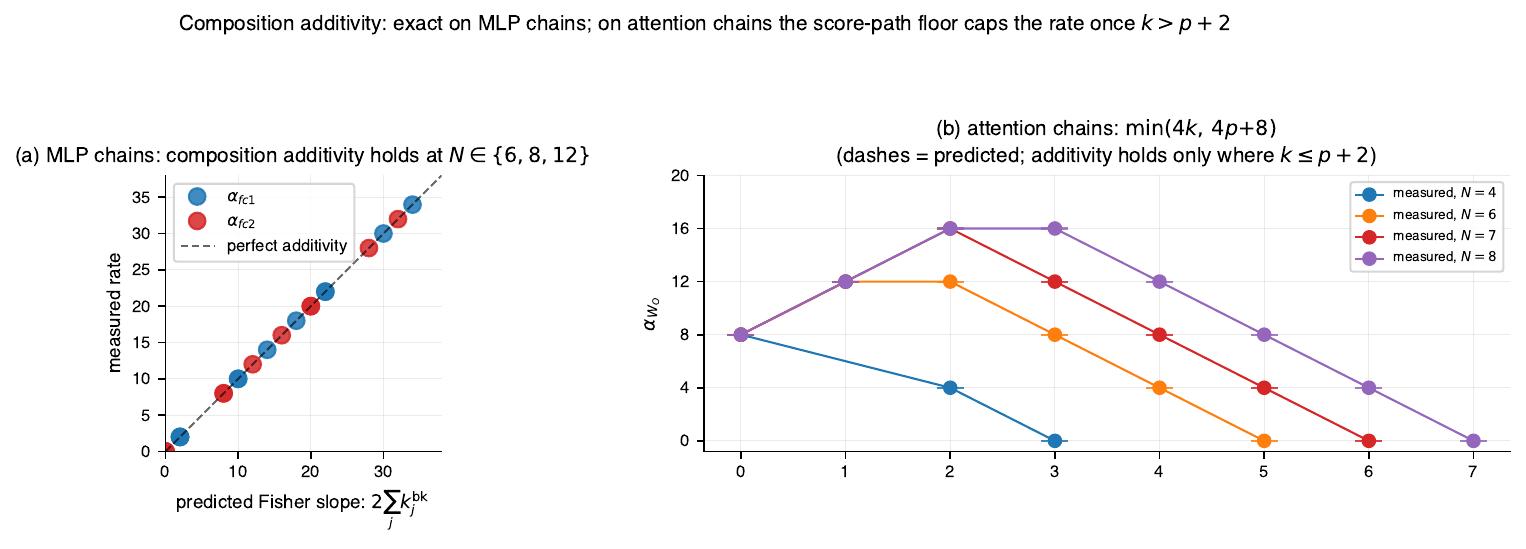}
\caption{Composition additivity (Theorem~\ref{thm:bridge_composition}). The two regimes the theorem distinguishes. (a)~MLP chains at $N \in \{6, 8, 12\}$, $d = 16$: per-component measured slopes lie on the $y = x$ diagonal against the predicted Fisher slope $2\sum_j k_j^{\mathrm{bk}}$ (the amplitude sum doubled on the Fisher form, $+2$ at the input-side component), validating clean additivity across slopes $0$--$34$. The scalar-transfer hypothesis holds and block rates add. (b)~Attention chains on the full probe profile: the measured $\alpha_{W_O}$ follows $\min(4k,\, 4p{+}8)$ (dashes), the smaller of the $V$--$O$ chain and the score-path floor of Proposition~\ref{prop:attn_chain_softmax}. Additivity is the $4k$ branch, and it holds only where $k \leq p + 2$.}
\label{fig:composition_additivity_mlp}
\end{figure}

\section{Architectural instantiations}
\label{sec:theory:architectural}

Modern networks are built from a small set of repeated primitives, and the bridge instantiates at each. The primitives cover a standard pre-norm transformer's forward map end to end: the \emph{structural} primitives that carry the forward map (rectangular widths for embeddings and MLP projections, residual DAGs, biases), the \emph{normalisation} layer (layer normalisation, with RMSNorm as its companion), the \emph{gated feedforward} block (SwiGLU, the gated MLP of modern large language models; GeGLU and ReGLU follow the same factorisation), and the \emph{sequence-mixing} block (single-head attention). Three \emph{analysis-side extensions} broaden the rate theorem's hypotheses to the settings these architectures train in: multi-direction singularities, cross-entropy loss, and non-canonical alignment. Each instantiation has the same shape: a per-block KL order, the rate composing through the additivity theorem (Section~\ref{sec:theory:composition}) where the scalar-transfer hypothesis holds, and the resulting singular-geometry consequence. Every predicted rate is checked by a static parametric freeze-probe on the theorem's own trajectory; these validations are collected in Appendix~\ref{app:theory:arch_freeze_probe} and summarised in Figure~\ref{fig:arch_rate_roundup}.

Table~\ref{tab:arch_landscape} lays out how each primitive acts on the canonical rate ladder, and the table splits the primitives into the two groups this section treats differently. Most leave the ladder intact: rectangular widths, multi-direction singularities, and biases carry the per-block KL order through composition unchanged, the SwiGLU gate contributes a fixed forward order, and non-canonical alignment marks the boundary where a learned trajectory leaves the regime the ladder describes. These carry no behaviour beyond what the composition theorem already gives, so we record their rates in the table and develop them, with their freeze-probe validations, in the architectural catalogue of Appendix~\ref{app:theory:arch_catalogue}. The four primitives developed here are the ones that change the picture, each introducing a phenomenon the bare ladder does not capture: cross-entropy loss moves the analysis onto the expected Fisher and brings in the logit-shift gauge (Section~\ref{sec:theory:arch:ce}); the residual skip makes the bottom-of-spectrum $\sigma_{\min}$ depth-invariant and readable from a single forward pass (Section~\ref{sec:theory:arch:res}); LayerNorm both lowers the integer ladder to a fractional rate and pins an algebraic kernel direction (Section~\ref{sec:theory:arch:ln}); and stacking single-head attention breaks composition additivity through the score-path injection floor (Section~\ref{sec:theory:arch:attn}).

Two of these four are of a different character from the rate-decay theorems, and they hold across the optimiser regimes considered in this paper. The LayerNorm kernel direction $\gamma^{-1}/\|\gamma^{-1}\|$ (Proposition~\ref{prop:ln_kernel}) is an exact algebraic identity: $\mathrm{cov}(\mathrm{LN}(X))\,v^* = 0$ holds for any input distribution because of LN's mean-subtraction projector, with $v^*$ fixed by the affine parameter alone. The residual-DAG bottom-of-active-spectrum statement (Corollary~\ref{cor:sigma-min-res}) follows from the path-cost decomposition of the forward chain rule, and its two cases differ in strength: on the canonical-aligned approach the invariance is structural, while off canonical alignment it is conditional on the per-checkpoint non-cancellation hypothesis, which the corollary's remarks show fails on generic random draws. The kernel identity depends on the forward-map algebra alone; the $\sigma_{\min}$ statement inherits the approach and activation-class hypotheses of its corollary.

The list is not exhaustive. It covers the primitives that compose a pre-norm transformer's forward map, with residual DAGs extending to ResNet-style skips. Primitives outside that slice, including multi-head and grouped-query attention, rotary position embeddings, and mixture-of-experts routing, are natural extensions following the same per-primitive recipe (a per-block KL order entering the composition) and are not treated here.
 
\begin{table}[t]
\centering
\caption{How each architectural primitive acts on the canonical rate ladder. The four entries above the rule carry the ladder through composition (or, for non-canonical alignment, mark where it ceases to hold) without introducing new phenomena; they are stated and freeze-probe-validated in the architectural catalogue, Appendix~\ref{app:theory:arch_catalogue}. The five below the rule each introduce behaviour the bare ladder does not capture; four are developed in this section, and SwiGLU, developed in Appendix~\ref{app:theory:arch_catalogue}, is placed below because its dead-direction map is quadratic in the dead coordinate, so its single-block order does not compose additively.}
\label{tab:arch_landscape}
\small
\begin{tabular}{lll}
\toprule
Primitive & Effect on the canonical ladder & Result \\
\midrule
Rectangular widths & preserved, per factor & Thm~\ref{thm:bridge_rect} \\
Multi-direction / asymmetric & per-direction ladder & Thm~\ref{thm:bridge_multi} \\
Biases & preserved & Thm~\ref{thm:bridge_bias} \\
Non-canonical alignment & ladder ceases to hold & Prop~\ref{prop:bridge_nonlinear_rot_negative} \\
\midrule
Cross-entropy loss & expected-Fisher rate; logit-shift gauge & Thm~\ref{thm:bridge_ce} \\
Residual DAG & $\sigma_{\min}$ depth-invariant (rate $0$) & Cor~\ref{cor:sigma-min-res} \\
LayerNorm & fractional rate; kernel $\gamma^{-1}/\|\gamma^{-1}\|$ & Prop.~\ref{thm:bridge_ln}, Prop~\ref{prop:ln_kernel} \\
SwiGLU gate & forward order $k = 3$; stacks give $3(2^n-1)$, not $3n$ & Prop~\ref{prop:swiglu_rate} \\
Single-head attention & standalone $k = 2$; composition anomaly & Thm~\ref{thm:bridge_attn} \\
\bottomrule
\end{tabular}
\end{table}

\subsection{Cross-entropy loss}
\label{sec:theory:arch:ce}

\label{app:bridge_ce}

Theorem~\ref{thm:bridge} is stated for squared-error loss with Gaussian noise, where the output-layer $G_L = \sigma^{-2} I$ is exactly isotropic and the residual at convergence is the noise $\varepsilon$. For supervised classification the loss is cross-entropy, and at a memorising parameter the empirical output gradient $g_\mathrm{head} = \mathrm{softmax}(z) - e_y$ collapses to zero: the same denominator that kept the MSE rate readable is gone. The natural move is to replace the empirical Fisher with the \emph{expected} Fisher (sample the label from the model's own predictive distribution rather than the training label); the data-averaged softmax Hessian $H(x;\theta)$ is bounded below on $\{1\}^\perp$ by a non-degeneracy assumption on the data-averaged class distribution, and the base case of the bridge proof carries through with $c_0$ replacing $\sigma^{-2}$. The rest of the backward-chain argument transfers unchanged.

\paragraph{Setup.}
Take the same $L$-layer architecture and symmetric-approach parametrization $\theta(t)$ as in Theorem~\ref{thm:bridge}, with the square-hidden assumption specialized to $C$ output classes by setting the final layer width equal to $C$ (i.e.\ $h_L = C$ and $h_\ell = h = C$ for all $\ell$). The rectangular extension $h_L = C \neq h$ combines with the present argument by composing this theorem's output-head replacement (Lemma~\ref{lem:ce_head}) with the rectangular non-dead block control of \xref{thm:bridge_rect}{the rectangular extension }, since the two extensions touch disjoint parts of the proof; we discuss the joint scope in Remark~\ref{rem:bridge_ce_rect_joint}. The hidden-layer activations $\phi$ still belong to classes (P1)--(P3); the softmax map is applied \emph{only} at the output head, to convert logits into probabilities, and is not itself one of the hidden activations:
$$z(x;\theta) = f(x;\theta) \in \mathbb{R}^C, \qquad p_\theta(y \mid x) = \mathrm{softmax}(z(x;\theta))_y.$$
Let $q(x)$ be the data distribution on $x$ (independent of $\theta$). The \emph{expected} Fisher is
$$F_\mathrm{exp}(\theta) = \expect_{x \sim q} \expect_{y \sim p_\theta(\cdot \mid x)} \bigl[\nabla_\theta \log p_\theta(y \mid x) \,\nabla_\theta \log p_\theta(y \mid x)^\top\bigr].$$
The K-FAC G-factor $G_\ell$ keeps its definition as the backward-gradient Gram at the pre-activation of layer $\ell$, where per-sample gradients are taken at labels sampled from $p_\theta(\cdot \mid x)$ (rather than at observed training labels). Since $C = h$, the dead direction index $h$ used throughout the bridge proof equally indexes a coordinate in the output logit space.

\begin{lemma}[Output-head covariance under expected Fisher]
\label{lem:ce_head}
Let $\delta^{(L)}(x, y) = \mathrm{softmax}(z(x;\theta)) - e_y$ be the per-sample output gradient. Under expected Fisher, the conditional covariance of $\delta^{(L)}$ given $x$ is
\[
H(x;\theta) \;:=\; \expect_{y \sim p_\theta(\cdot \mid x)}\!\bigl[\delta^{(L)}(x,y)\,\delta^{(L)}(x,y)^\top\bigr] \;=\; \mathrm{diag}\bigl(p(x;\theta)\bigr) - p(x;\theta)\,p(x;\theta)^\top,
\]
with $p(x;\theta) = \mathrm{softmax}(z(x;\theta))$. Moreover, for any $v \in \{\mathbf{1}\}^\perp$ and any full-support $p$:
\[
v^\top H(x;\theta)\, v \;=\; \mathrm{Var}_{y \sim p}\!\bigl[v_y\bigr] \;=\; \sum_{y} p_y\, v_y^2 - \Bigl(\sum_y p_y v_y\Bigr)^2 \;\ge\; 0,
\]
with equality iff $v$ is constant on $\mathrm{supp}(p)$. Hence the nullspace of $H$ is $\mathrm{span}(\mathbf{1}) \oplus \mathrm{span}\{e_i : p_i = 0\}$, the vectors constant on $\mathrm{supp}(p)$ and free off it; for full-support $p$ the nullspace is exactly $\mathrm{span}(\mathbf{1})$ and $H$ is positive-definite on $\{\mathbf{1}\}^\perp$.
\end{lemma}

\ifexpanded
\begin{proof}
Direct expansion: $\expect_y[(p - e_y)(p - e_y)^\top] = p p^\top - p\expect[e_y]^\top - \expect[e_y]p^\top + \expect[e_y e_y^\top]$. With $y \sim p_\theta(\cdot|x)$, $\expect[e_y] = p$ and $\expect[e_y e_y^\top] = \mathrm{diag}(p)$. Thus $H = \mathrm{diag}(p) - p p^\top$, the standard softmax Hessian. The quadratic form $v^\top H v = \sum_y p_y v_y^2 - (\sum_y p_y v_y)^2 = \mathrm{Var}_p(v)$ is the variance of the random variable $y \mapsto v_y$ under $p$; it is non-negative by Jensen, zero iff $v$ is $p$-a.s.\ constant. Combined with $1 \in \ker H$ (by shift-invariance of softmax), the nullspace characterization follows.
\end{proof}
\fi

\begin{assumption}[Non-degenerate data-averaged output probabilities]
\label{ass:nondeg_pxy}
There exist constants $c_0 > 0$ and a neighborhood $\mathcal{U}$ of $t = 0$ such that for all $t \in \mathcal{U}$ and all unit $v \in \{\mathbf{1}\}^\perp$,
\[
v^\top \expect_{x \sim q}\bigl[H(x;\theta(t))\bigr]\, v \;\ge\; c_0.
\]
The restriction to $\{\mathbf{1}\}^\perp$ is forced by the softmax structure: $\mathbf{1} \in \ker H(x)$ for every $x$, so the unrestricted smallest eigenvalue of the averaged Hessian (and of $\Pi H \Pi$ for the centering projector $\Pi = I - \tfrac{1}{C}\mathbf{1}\mathbf{1}^\top$) is identically zero, and no unrestricted bound can hold. A sufficient condition is pointwise: for each $x$ the restricted smallest eigenvalue of $H(x;\theta)$ is bounded below by the least-likely class mass $\min_i p_\theta(i \mid x)$ (with equality on symmetric $p$), so a uniform per-example floor $\min_i p_\theta(i \mid x) \ge c_0$ for $q$-almost-all $x$ delivers the display. A floor on the data-averaged class distribution does not: per-example posteriors can approach point masses on different classes for different $x$ while averaging to uniform, and the averaged restricted eigenvalue then vanishes with the per-example gap.
\end{assumption}

Assumption~\ref{ass:nondeg_pxy} constrains the per-example posteriors, and being an imperfect classifier does not suffice: a model whose per-example posteriors approach point masses on \emph{different} classes for different $x$ is not a perfect classifier of any one class, keeps every class at large average mass, and still violates the assumption, since each individual posterior degenerates. What the assumption requires is that the posteriors stay non-degenerate per example, uniformly over a $q$-full-measure set, in the spanning sense of the display. This is the natural CE analog of ``$\sigma^2 > 0$'' in the MSE setup: the theorem is about approach to the singular stratum, not about the degenerate configuration in which all classes collapse to one.

\begin{lemma}[Backward dead-component magnitude under cross-entropy]
\label{lem:backward-dead-ce}
Under Assumption~\ref{ass:nondeg_pxy} and the same symmetric approach, for the expected Fisher:
\[
\expect\bigl[(\delta^{(L, h)})^2\bigr] \;=\; \Theta(1), \qquad
\expect\bigl[(\delta^{(\ell, h)})^2\bigr] \;=\; \Theta\!\bigl(t^{2(L-\ell)}\bigr) \text{ for } \ell < L,
\]
where $\delta^{(\ell,h)}$ is the dead-direction backward delta at layer $\ell$.
\end{lemma}

\ifexpanded
\begin{proof}
At $\ell = L$: $\delta^{(L)}(x,y) \in \{\mathbf{1}\}^\perp$ since $\mathbf{1}^\top \delta^{(L)} = (\mathbf{1}^\top p) - (\mathbf{1}^\top e_y) = 1 - 1 = 0$. Write $e_h = \Pi_{\{\mathbf{1}\}^\perp} e_h + (1/C)\mathbf{1}$. Because $H \mathbf{1} = 0$ (Lemma~\ref{lem:ce_head}) implies $\expect_x[H] \cdot \mathbf{1} = 0$, the two cross-terms and the $(1/C)^2 \mathbf{1}^\top \expect_x[H] \mathbf{1}$ term all vanish, leaving the identity
\[
\expect_x\bigl[(\delta^{(L,h)})^2\bigr] \;=\; e_h^\top \expect_x[H(x;\theta(t))]\, e_h \;=\; (\Pi_{\{\mathbf{1}\}^\perp} e_h)^\top \expect_x[H]\, (\Pi_{\{\mathbf{1}\}^\perp} e_h).
\]
Applying Assumption~\ref{ass:nondeg_pxy} to the vector $v = \Pi_{\{\mathbf{1}\}^\perp} e_h \in \{\mathbf{1}\}^\perp$ gives the lower bound
$\expect_x\bigl[(\delta^{(L,h)})^2\bigr] \ge c_0 \|\Pi_{\{\mathbf{1}\}^\perp} e_h\|^2 = c_0 (1 - 1/C)$,
uniformly for $t \in \mathcal{U}$. Thus $\expect[(\delta^{(L,h)})^2] = \Theta(1)$ on $\mathcal{U}$.

For $\ell < L$: the backward recursion $\delta^{(\ell,h)} = \phi'(a_\ell^{(h)}) \cdot t \cdot \delta^{(\ell+1,h)}$ (Lemma~\ref{lem:backward-dead-sub}'s derivation depended only on the canonical structure of $W_{\ell+1}$ and the chain rule, both loss-independent) combined with the base case above gives
$\expect[(\delta^{(\ell,h)})^2] = \Theta(1) \cdot t^2 \cdot \Theta(t^{2(L-\ell-1)}) = \Theta(t^{2(L-\ell)})$ by induction, exactly as in the MSE proof.
\end{proof}
\fi

\begin{theorem}[Multi-Layer K-FAC G-factor Bridge, cross-entropy]
\label{thm:bridge_ce}
Under the expected-Fisher setup, Assumption~\ref{ass:nondeg_pxy}, and activation classes (P1)--(P3), let $u_\ell$ denote the canonical dead-direction unit vector at layer $\ell$:
\begin{enumerate}
\item[(a)] \emph{(Dead-direction entry, primary statement.)} For $\ell \in \{1, \ldots, L-1\}$,
\[
u_\ell^\top G_\ell(\theta(t))\, u_\ell \;=\; C_\ell^{\mathrm{CE}} \cdot t^{2(L-\ell)} \cdot (1 + r_\ell^G(t)),
\]
with $r_\ell^G(t) = O(t)$ and activation-class-dependent constants matching Theorem~\ref{thm:bridge}.
\item[(b)] \emph{(Output layer.)} $u_L^\top G_L(\theta(t))\, u_L = \Theta(1)$, with constant determined by $c_0$ from Assumption~\ref{ass:nondeg_pxy} (replacing $\sigma^{-2}$ in the MSE statement).
\item[(c)] \emph{(Spectrum under cross-entropy.)} In class (P1) the layer-to-logit Jacobian is sample-independent, and the logit-shift gauge places one exact zero eigenvalue in $G_\ell$ at every layer and every $t$, with eigenvector converging to the dead direction as $t \to 0$: the exact kernel absorbs the dead direction's decay, the remaining spectrum is $\Theta(1)$ under the additional hypothesis that the non-dead block of $G_\ell$ is uniformly $\Theta(1)$ in $t$, and \emph{no} eigenvalue of $G_\ell$ decays at the dead rate; raw $\lambda_{\min}(G_\ell) = 0$ identically, and the rate appears in the entry $u_\ell^\top G_\ell u_\ell$ and in no eigenvalue. In classes (P2) and (P3) the backward Jacobian carries the diagonal gates $\phi'(a_\ell(x))$, the per-sample image of $\mathbf{1}^\perp$ rotates with the activation pattern, and $G_\ell$ is generically full rank: the dead-row Schur cancellation is partial, and under the same non-dead-block hypothesis the smallest eigenvalue decays at the dead rate $\Theta(t^{2(L-\ell)})$ with a class-dependent constant (under ReLU at $L = 2$, $h = C = 3$ it is about $0.8$ of the dead entry, the committed check \texttt{check\_ce\_kernel\_class.py}). (This is the $m = 1$ case of Theorem~\ref{thm:bridge_multi}(c)'s cross-entropy clause.)
\item[(d)] \emph{(Output-rate match.)} The shallowest-layer rate $2(L-1)$ matches Theorem~\ref{thm:fisher_decay} at KL order $k = L$, unchanged from MSE.
\end{enumerate}
\end{theorem}

\begin{remark}[Correction to version 1]
\label{rem:bridge_ce_v1_correction}
Version 1 of this paper stated part (c) as a smallest-eigenvalue corollary in every activation class. In class (P1) the logit-shift gauge places an exact zero in the spectrum of every hidden-layer G-factor, so the eigenvalue reading holds only in classes (P2) and (P3); the entry statement of part (a) is the class-independent form. The committed check named in part (c) carries the Monte Carlo that separates the two cases.
\end{remark}

\ifexpanded
\begin{proof}
The proof of Theorem~\ref{thm:bridge} uses Lemmas~\ref{lem:forward-dead-sub}, \ref{lem:backward-dead-sub}, \ref{lem:integral-reduction-sub}, plus a non-dead control argument in canonical coordinates. Lemma~\ref{lem:forward-dead-sub} (forward dead-component propagation) is architecture-only and transfers verbatim, as does the A-side of Lemma~\ref{lem:integral-reduction-sub}. The G-side of that lemma does \emph{not} transfer: its off-diagonal bound is derived from the MSE noise being zero-mean and independent across output coordinates, while the CE label couples every output coordinate ($\mathbf{1}^\top \delta^{(L)} = 0$ identically), the dead-row off-diagonals are $\Theta(t^{L-\ell})$, and in class (P1) the Schur cancellation at the dead row is total: the Schur constant is $c = 1$, which is the mechanism of part (c)'s exact gauge zero, while in (P2) and (P3) the sample-dependent gates leave the cancellation partial ($c < 1$). For the non-dead statement: for $j \neq h$, the dead-direction canonical structure leaves $(W_\ell^*)_{jj} = 1$, so non-dead-direction forward activations $a_\ell^{(j)}$ are $\Theta(1)$ independent of $t$, giving $(A_\ell)_{jj} = \Theta(1)$; the backward non-dead $\delta^{(\ell,j)}$ is driven at $\ell = L$ by $\expect_y[(\delta^{(L,j)})^2] = e_j^\top H(x;\theta) e_j$, which by Lemma~\ref{lem:ce_head} and Assumption~\ref{ass:nondeg_pxy} is bounded below by $c_0(1-1/C) > 0$, hence $\Theta(1)$, and propagates backward through $\delta^{(\ell,j)} = \phi'(a_\ell^{(j)}) \delta^{(\ell+1,j)}$ (no $t$-factor on the non-dead diagonal of $W_{\ell+1}$) to yield $(G_\ell)_{jj} = \Theta(1)$ uniformly in $t$. Lemma~\ref{lem:backward-dead-sub}'s base case changes and is replaced by Lemma~\ref{lem:backward-dead-ce}, which gives the same asymptotic form $\Theta(t^{2(L-\ell)})$ for $\ell < L$ and $\Theta(1)$ for $\ell = L$. Claims (a) and (b) are entry statements and follow directly from that backward recursion evaluated on the dead coordinate; no Schur step is needed for them. In class (P1) the Schur route Theorem~\ref{thm:bridge} takes to its eigenvalue statement is unavailable ($c = 1$, above), which is why part (c) is a negative statement there: the dead-row cancellation is total, the kernel eigenvector rotates into the dead direction, and the spectrum retains no rated eigenvalue. In (P2) and (P3) the Schur route goes through with $c < 1$, and the smallest eigenvalue carries the dead rate as under MSE.

\emph{Remark on the activation classes.} Classes (P1), (P2), (P3) refer to hidden-layer nonlinearities only; the softmax at the output head is part of the loss construction, not a hidden activation. All three proofs for smooth Taylor corrections ($r_\ell^G$) in Theorem~\ref{thm:bridge} carry through unchanged because they concern the $\phi'(a_\ell^{(h)})$ factors in the backward chain, which are loss-independent.
\end{proof}
\fi

\begin{remark}[Static-Fisher rate vs.\ Adam+CE trajectory observable]
\label{rem:bridge_ce_adam_scope}
Theorem~\ref{thm:bridge_ce} is a statement about the \emph{static} expected-Fisher spectrum at the parametric trajectory $\theta(t)$. CE introduces the logit-shift gauge $G = \mathbb{R}$ as the loss's continuous symmetry; under Adam, the empirical observation of gauge-mode drift on this gauge (Remark~\ref{rem:adam_nondescent}) is what makes rate-fitting on an Adam+CE trajectory return drift rather than the theorem's exponent. The robust static-checkpoint observable on Adam+CE is $\sigma_{\min}(X_\ell)$ on the residual stream (Corollary~\ref{cor:sigma-min-res}). To recover the trajectory exponent under CE one applies a gauge fix at training time (the auxiliary $z$-loss penalty \citep{ShazeerMesh18,ZophSTMoE22} is the standard construction) combined with an optimizer whose residual non-equivariance under remaining symmetries is bounded; Remark~\ref{rem:adam_nondescent} discusses both routes.
\end{remark}

\begin{remark}[Joint scope with the rectangular extension]
\label{rem:bridge_ce_rect_joint}
The CE replacement in Lemma~\ref{lem:ce_head} touches only the $\ell = L$ output-head base case; the rectangular extension touches only the non-dead-block control through the Schur reduction (off-diagonal $A_\ell$ entries under $h_\ell \neq h_{\ell-1}$) at the hidden layers. The two extensions therefore commute, and the joint statement holds: for any $L$-layer feedforward chain with rectangular widths $h_1, \ldots, h_{L-1}$ and output width $h_L = C$, expected Fisher under cross-entropy, and a canonical-aligned dead direction at each layer, the dead-direction-entry rate $u_\ell^\top G_\ell u_\ell = \Theta(t^{2(L-\ell)})$ holds for $\ell < L$ and $u_L^\top G_L u_L = \Theta(1)$.\ifexpanded\ The rectangular CE entries in Table~\ref{tab:ce_bridge}'s $6 \to 8 \to 4$ network (rank-$2$ target, so rank deficit $m = 2$) instantiate this joint statement combined with the multi-direction extension of Theorem~\ref{thm:bridge_multi}; the probed quantity is the per-direction entry of that theorem's part (a).\fi
\end{remark}

\emph{Remark on the observable in practice.} Theorem~\ref{thm:bridge_ce}(a) is stated on the dead-direction entry $u_\ell^\top G_\ell u_\ell$ because under CE the entry is the \emph{only} carrier of the rate: in class (P1) part (c) shows the raw spectrum is $\{0, \Theta(1), \ldots, \Theta(1)\}$ in the square single-direction case, so an unlabeled-spectrum rate-matching rule has nothing to match there, and spectrum-only readings of the CE G-factor cannot recover the rate; in (P2) and (P3) the bottom eigenvalue carries the rate with a class-dependent constant, and the entry remains the cleaner reading. Operational recovery under CE therefore probes the entry directly at a candidate direction (the K-FAC lift of Corollary~\ref{cor:kfac_lift} supplies the candidate), or fixes the shift gauge at training time (Remark~\ref{rem:bridge_ce_adam_scope}) so the kernel direction is lifted; whether a given gauge fix restores an eigenvalue at the dead rate is a property of that construction and is not settled here. Under MSE, where no loss gauge exists, the spectral rate-matching rule of Theorem~\ref{thm:bridge} applies as stated.

\begin{corollary}[Empirical Fisher collapse under memorization]
\label{cor:emp_vs_exp_ce}
Let $F_\mathrm{emp}(\theta) = \expect_{(x,y_\mathrm{obs}) \sim \mathcal{D}}\bigl[\nabla_\theta \log p_\theta(y_\mathrm{obs} \mid x)\, \nabla_\theta \log p_\theta(y_\mathrm{obs} \mid x)^\top\bigr]$ denote the empirical Fisher, where $\mathcal{D}$ is a fixed data distribution of $(x, y_\mathrm{obs})$ pairs with marginal $q$ on $x$. Suppose the trajectory $t \mapsto \theta(t)$ satisfies $p_{\theta(t)}(y_\mathrm{obs} \mid x) \to 1$ for $\mathcal{D}$-almost-all $(x, y_\mathrm{obs})$ as $t \to 0$ (``perfect memorization of $\mathcal{D}$''). Then both estimators collapse, and they collapse at different orders: writing $\epsilon(t)$ for the mass the model leaves off the observed label, $F_\mathrm{emp}(\theta(t)) = \Theta(\epsilon^2)$ while $F_\mathrm{exp}(\theta(t)) = \Theta(\epsilon)$, so $F_\mathrm{emp} = \Theta(F_\mathrm{exp}^2)$. The empirical Fisher therefore reaches the numerical floor while the expected Fisher is still resolvable, and what survives on the expected side is the \emph{ratio} structure of Theorem~\ref{thm:bridge_ce}'s spectrum rather than its absolute rates.

\emph{Consistency note.} Perfect memorisation is incompatible with Assumption~\ref{ass:nondeg_pxy}: driving $p_{\theta(t)}(y_\mathrm{obs} \mid x) \to 1$ sends the class distribution to a point mass, so $\lambda_{\min}(\expect_x[\Pi H \Pi]) \to 0$ and no $c_0 > 0$ survives. Theorem~\ref{thm:bridge_ce}'s absolute rate statement consequently does not apply on a memorising trajectory.
\end{corollary}

\ifexpanded
\begin{proof}
Under perfect memorization, for each $(x, y_\mathrm{obs})$ in the support of $\mathcal{D}$ we have $p_{\theta(t)}(y_\mathrm{obs} \mid x) \to 1$, hence $\delta^{(L)}(x, y_\mathrm{obs}; \theta(t)) = \mathrm{softmax}(z(x; \theta(t))) - e_{y_\mathrm{obs}} \to 0$. The per-sample gradient $\nabla_\theta \log p_{\theta}(y_\mathrm{obs} \mid x) = J(x;\theta)^\top \delta^{(L)}(x, y_\mathrm{obs}; \theta)$ therefore vanishes, and $F_\mathrm{emp}$ collapses as the expectation of a vanishing PSD quantity.

For the expected side, averaging over $y \sim p_\theta(\cdot \mid x)$ yields $H(x;\theta) = \mathrm{diag}(p) - p p^\top$ (Lemma~\ref{lem:ce_head}). Writing $\epsilon := 1 - p_{y_\mathrm{obs}}$, the observed-label gradient has $\|\delta^{(L)}\|^2 = \Theta(\epsilon^2)$ while $\mathrm{tr}\, H = \sum_i p_i (1 - p_i) = \Theta(\epsilon)$, giving the quadratic separation. Numerically at $C = 5$, the ratio $F_\mathrm{emp}/F_\mathrm{exp}^2$ is flat across the trajectory (about $0.31$ over ten orders of magnitude in $\epsilon$), and $\lambda_{\min}(\Pi H \Pi)$ falls with $\epsilon$ rather than staying bounded below, which is why the absolute-rate reading is unavailable here and the ratio reading is what the protocol uses.
\end{proof}
\fi

\begin{remark}[Why $F_\mathrm{emp}$ and $F_\mathrm{exp}$ do not PSD-dominate each other in general]
\label{rem:no_psd_dom}
It is tempting to claim $F_\mathrm{emp} \preceq F_\mathrm{exp}$ via Jensen's inequality, but this is \emph{incorrect} in general. Both quantities are averages of rank-one matrices $\delta^{(L)} (\delta^{(L)})^\top$ over different joint distributions of $(x, y)$: $\mathcal{D}$ versus $(q \otimes p_\theta)$. Neither average PSD-dominates the other without additional structural assumptions tying the two distributions together (e.g., if $y_\mathrm{obs}(x) \sim p_\theta(\cdot \mid x)$, the two are equal in expectation). Corollary~\ref{cor:emp_vs_exp_ce} separates the two estimators quantitatively (both collapse along the memorization trajectory, at quadratically different orders); it asserts no PSD ordering, and none holds in general.
\end{remark}

Corollary~\ref{cor:emp_vs_exp_ce} is the cross-entropy instance of the estimator scope set out in Remark~\ref{rem:population_vs_empirical_fisher}: on post-memorization training trajectories, the empirical Fisher loses its rate signal to numerical floor even though the rate \emph{structure} is still encoded in eigenvalue ratios. Measurement protocols on trained classifiers should therefore use the expected Fisher rather than the empirical Fisher when rate-fitting is the goal.

\paragraph{Scope.} Theorem~\ref{thm:bridge_ce} preserves the remaining scope limitations of Theorem~\ref{thm:bridge} not addressed here: feedforward chain without residuals (addressed by Theorem~\ref{thm:bridge_res}), canonical-basis-aligned dead direction (addressed partially in \S\ref{app:bridge_rot}), element-wise hidden activation (the attention case is below). The square-hidden assumption is removed separately by the rectangular extension; the joint $\text{rectangular} \times \text{CE}$ statement is given by Remark~\ref{rem:bridge_ce_rect_joint}\ifexpanded, and Table~\ref{tab:ce_bridge}'s $6 \to 8 \to 4$ rectangular configuration (rank deficit $m = 2$) instantiates it jointly with the multi-direction extension\fi. The trajectory-rate scope under Adam+CE is governed by Remark~\ref{rem:bridge_ce_adam_scope} and Corollary~\ref{cor:quotient_rate}.

\ifexpanded
\begin{table}[ht]
\caption{Cross-entropy vs.\ MSE eigenvalue scaling at $L = 2$ (5 seeds, $6 \to 8 \to 4$ network, rank-2 target). Parametric freeze-probe: the dead-direction entry $u_\ell^\top G_\ell(\theta(t))\, u_\ell$ is fit against $t$ along the canonical construction, with the cross-entropy rows under the expected Fisher (Lemma~\ref{lem:ce_head}). Predicted exponent: $2.000$ for $k = 2$. Deviations of the smooth rows from the integer are the finite-$t$ Taylor correction $O(t)$ of Theorem~\ref{thm:bridge}(a), as in Table~\ref{tab:nonlinear_bridge}; the identity rows match within sampling noise.}
\label{tab:ce_bridge}
\centering
\small
\begin{tabular}{llrl}
\toprule
Activation & Loss & Exponent $\alpha$ & Deviation from $2$ \\
\midrule
Identity & MSE & $1.995 \pm 0.003$ & $-0.005$ \\
Identity & Cross-entropy & $1.972 \pm 0.025$ & $-0.028$ \\
GeLU & MSE & $2.066 \pm 0.008$ & $+0.066$ \\
GeLU & Cross-entropy & $2.044 \pm 0.068$ & $+0.044$ \\
Tanh & MSE & $1.914 \pm 0.003$ & $-0.086$ \\
Tanh & Cross-entropy & $1.900 \pm 0.036$ & $-0.100$ \\
\bottomrule
\end{tabular}
\end{table}
\fi

\subsection{Residual DAGs and \texorpdfstring{$\sigma_{\min}$}{sigma\_min} depth-invariance}
\label{sec:theory:arch:res}

\label{app:bridge_res}

Theorem~\ref{thm:bridge}\ifexpanded, Theorem~\ref{thm:bridge_ce}, and Theorem~\ref{thm:bridge_rect}\else, Theorem~\ref{thm:bridge_ce}, and the rectangular extension \fi{} assume a feedforward chain with no skip connections. Modern networks (ResNets, transformer FFN blocks, transformer attention blocks) all wrap sub-chains in additive-identity residual connections of the form
\[
h_\mathrm{out} = h_\mathrm{in} + F(h_\mathrm{in}),
\]
where $F$ is a short feedforward sub-chain. An identity skip is not a weight matmul, so it does not carry a $t$-factor in the dead direction's chain-rule expansion. The natural question is how the rate ladder of Theorem~\ref{thm:bridge} adjusts when skips are present, and whether a clean graph-distance reading exists. This subsection answers both: an additive identity skip is a zero-weight shortcut, so the dead-direction rate at any layer is determined by the shortest \emph{weighted}-path distance from that layer to the output, with weight edges costing $1$ and skip edges costing $0$. The empirical characterisation that motivated this construction is summarised below; the formal statement follows.

\paragraph{Setup.}
A \emph{residual computational graph} has nodes $V = \{v_0, v_1, \ldots, v_L\}$ representing pre-activations ($v_0$ input, $v_L$ output) and two kinds of edges, which may \emph{coexist} at the same endpoints:
\begin{itemize}
\item \emph{Weight edges} $W_1, \ldots, W_L$, each from $v_{i-1}$ to $v_i$, representing the linear map (optionally followed by elementwise activation $\phi$) of a feedforward layer. The activation belongs to the edge, so a node holds the value \emph{after} any activation on the edges feeding it, and the argument of $\phi$ on edge $i$ is $W_i v_{i-1}$, whose dead component carries at least one factor of $t$. The Taylor step at $0$ in the proofs below is therefore always taken inside its radius, which the alternative convention (activation attached to the node) would not guarantee at a node with $K^{\mathrm{fwd}} = 0$.
\item \emph{Identity-skip edges} $\text{skip} \subseteq V \times V$, each $(v_j, v_k)$ with $k > j$, representing the additive identity $h_k \mathrel{{+}{=}} h_j$. An additive-identity residual block from $v_j$ to $v_k$ is encoded by simultaneously having the weight-edge chain $v_j \to v_{j+1} \to \cdots \to v_k$ \emph{and} the skip edge $(v_j, v_k)$: the forward computation is $h_k = F(h_j) + h_j$ where $F$ is the weight-chain output.
\end{itemize}
Hence the computational graph can have multiple forward paths between two nodes when skips are present; the backward-delta receives a sum of contributions from each such path.

The symmetric dead-aligned perturbation perturbs each weight $W_i$ in the dead direction by $t$: $W_i(t) = W_i^* + t u_i u_{i-1}^\top$, with $W_i^*$ a rank-deficient singular configuration (per the setup of \S\ref{app:bridge_rect}) and $u_i$ the dead direction at node $v_i$. Skip edges are fixed identities and are not perturbed.

\begin{definition}[Shortest-weighted-path distance $K(\ell)$]
\label{def:path_distance}
Consider the graph $G_\mathrm{bwd}$ obtained from the forward DAG by reversing edges and assigning each weight edge cost 1 and each identity-skip edge cost 0. Let $K(\ell)$ be the minimum total cost of any path in $G_\mathrm{bwd}$ from $v_L$ to $v_\ell$. Equivalently, $K(\ell)$ is the smallest number of \emph{weights} the backward dead-direction chain must traverse to reach pre-activation $\ell$ when taking every available residual skip.
\end{definition}

\begin{lemma}[Backward dead-component under residuals]
\label{lem:backward-dead-res}
Along the symmetric approach on a residual computational graph, in activation classes (P1), (P3), and (P2) with $\phi'(0) > 0$, the dead-direction component of the backward delta at layer $\ell$ satisfies
\[
\expect\bigl[(\delta^{(\ell, u_\ell)})^2\bigr] = \Theta\!\bigl(t^{2 K(\ell)}\bigr),
\]
where $K(\ell)$ is the shortest-weighted-path distance (Definition~\ref{def:path_distance}). Correction terms enter at higher order: $O(t^{2K(\ell) + 1})$ from smooth activations (the backward path to $v_\ell$ crosses activations at layers $m \ge \ell + 1 \ge 1$, each with Taylor correction $O(t^m)$) and $O(t^{2K(\ell) + 1})$ from longer-path contributions through unused residual skips; all corrections vanish relative to the leading term as $t \to 0$.
\end{lemma}

\ifexpanded
\begin{proof}
By the chain rule on the forward DAG, $\delta^{(\ell, u_\ell)}$ decomposes as a sum over all directed paths $\pi$ from $v_L$ back to $v_\ell$ in the reversed graph $G_\mathrm{bwd}$:
\[
\delta^{(\ell, u_\ell)} \;=\; \sum_{\pi : v_L \to v_\ell} J_\pi\, \delta^{(L, u_L)},
\]
where $J_\pi$ is the product of edge operators along $\pi$: each weight edge $W_i$ contributes a factor $(W_i^\top)_{u,u} = t$ in the dead-direction-to-dead-direction entry (by the perturbation construction), and each identity-skip edge contributes a factor of $1$. For a path $\pi$ with $|\pi|_W$ weight edges (and any number of skip edges), $J_\pi = \Theta(t^{|\pi|_W})$. (Under (P1) linear and (P2) smooth activations, the dead-row diagonal contribution above carries through directly. Under (P3) ReLU, the gating couples the dead row to non-dead coordinates statistically, even though every $W_i(t)$ stays diagonal along the canonical perturbation: the per-node path-Gram acquires $O(t^{K(\ell)})$ off-row entries to non-dead coordinates, as in Lemma~\ref{lem:integral-reduction-sub} adapted to the residual graph, the non-dead block is $\Theta(1)$, and the dead-direction Schur complement absorbs at most a fixed fraction of the dead diagonal. The leading rate $\Theta(t^{2K(\ell)})$ on the dead-direction entry is preserved.)

Let $\mathcal{P}^*(\ell)$ denote the set of shortest paths (those with $|\pi|_W = K(\ell)$). Then
\[
\delta^{(\ell, u_\ell)} \;=\; \Bigl(\sum_{\pi \in \mathcal{P}^*(\ell)} c_\pi\Bigr) \cdot t^{K(\ell)} \cdot \delta^{(L, u_L)} + O(t^{K(\ell) + 1}),
\]
where each $c_\pi$ is a positive constant (product of dead-direction factors along $\pi$, all with the same sign since the perturbation $t u u^\top$ contributes $+1$ per dead-direction factor and identity skips contribute $+1$). The leading-order coefficient $\sum_{\pi \in \mathcal{P}^*(\ell)} c_\pi$ is strictly positive, so no destructive interference can occur. Squaring and taking expectation with $\expect[(\delta^{(L, u_L)})^2] = \Theta(1)$ (from MSE's $\sigma^2 I$ or CE's $H(x;\theta)$ restricted to $\{\mathbf{1}\}^\perp$) yields
\[
\expect[(\delta^{(\ell, u_\ell)})^2] = \Theta(t^{2 K(\ell)}).
\]
Smooth-activation corrections enter through the $\phi'(a_\ell^{(u)})$ factors on each weight edge along the shortest path; the $\ell$-th layer's factor is $\phi'(0) + O(t^\ell)$ by Lemma~\ref{lem:forward-dead-sub}, so these contribute a multiplicative $(1 + O(t^{\min(\ell,1)}))$. Longer-path contributions through unused skips contribute at relative order $t^1$ and above.
\end{proof}
\fi

\begin{theorem}[Multi-Layer K-FAC G-factor Bridge, residual networks]
\label{thm:bridge_res}
Under the residual-computational-graph setup, Assumption~\ref{ass:nondeg_pxy} if using cross-entropy, and activation classes (P1), (P3), and (P2) with $\phi'(0) > 0$, the dead-direction entry of the G-factor at layer $\ell$ satisfies
\[
(G_\ell(\theta(t)))_{u_\ell u_\ell} = C_\ell^\mathrm{res} \cdot t^{2 K(\ell)} \cdot \bigl(1 + r_\ell^G(t)\bigr),
\]
where $K(\ell)$ is the shortest-weighted-path distance (Definition~\ref{def:path_distance}) from $v_L$ back to $v_\ell$, and $r_\ell^G(t)$ collects smooth-activation and finite-$t$ residual-skip corrections. All corrections vanish as $t \to 0$.
\end{theorem}

\ifexpanded
\begin{proof}
Apply Lemma~\ref{lem:backward-dead-res} to get the magnitude of $\delta^{(\ell, u_\ell)}$, giving dead-row diagonal $(G_\ell)_{u_\ell u_\ell} = \Theta(t^{2 K(\ell)})$. Non-dead entries of $G_\ell$ are $\Theta(1)$ by the same argument as Theorem~\ref{thm:bridge}'s non-dead analysis (inherited at each node: non-dead directions do not accumulate $t$ factors through either weight edges or residual skips). The dead-direction Schur reduction follows the template of Lemma~\ref{lem:integral-reduction-sub}: in canonical coordinates, the dead-row off-diagonals to non-dead coordinates are $O(t^{K(\ell)})$ at leading order (one factor of $t$ per weight edge in the shortest path; the residual-skip edges contribute factors of $1$ but do not introduce additional $t$-factors), and the non-dead block is $\succ 0$ at $\Theta(1)$. Writing the dead-direction Schur complement as $d - v^\top M^{-1} v$ with $d = \Theta(t^{2K(\ell)})$, $v = O(t^{K(\ell)})$, $M = \Theta(1)$, the term $v^\top M^{-1} v = O(t^{2 K(\ell)})$ matches the dead diagonal in order, but the rank-1 outer-product structure (from the canonical-coordinate independence of dead and non-dead activations) absorbs at most a fixed fraction (Lemma~\ref{lem:integral-reduction-sub}; see Remark~\ref{rem:schur_constant_qualitative} on the qualitative status of $c < 1$). The entry statement itself needs no Schur step; the reduction is recorded here because Corollary~\ref{cor:res_lambda_min} consumes it, with its exceptions (the class (P1) cross-entropy kernel of Theorem~\ref{thm:bridge_ce}(c) among them).
\end{proof}
\fi

\begin{corollary}[$\lambda_{\min}$ statement on residual DAGs under complement genericity]
\label{cor:res_lambda_min}
Under Theorem~\ref{thm:bridge_res} together with a non-dead-block-control hypothesis (the residual-DAG analog of the rectangular extension's complement assumption) requiring the non-dead block of $G_\ell$ to be uniformly $\Theta(1)$ in $t$ at every node along the residual DAG, the $\lambda_{\min}$ statement holds:
\[
\lambda_{\min}(G_\ell(\theta(t))) \;=\; (G_\ell(\theta(t)))_{u_\ell u_\ell} \;=\; \Theta(t^{2 K(\ell)}).
\]
Without the complement hypothesis, the dead-direction entry of the display above still scales as $\Theta(t^{2K(\ell)})$ but $\lambda_{\min}$ may additionally pick up complement-null directions (e.g., gauge zeros, or block boundaries where the path graph permits multiple shortest paths with linearly-dependent products) that are identically zero rather than decaying.
\end{corollary}

\begin{remark}[Trained-network non-cancellation]
\label{rem:bridge_res_trained}
Lemma~\ref{lem:backward-dead-res} establishes non-cancellation under the canonical symmetric perturbation $W_i(t) = W_i^* + t u_i u_{i-1}^\top$, where each weight edge contributes the same sign in the dead-direction sub-block by construction. On a trained network the required hypothesis must be stated uniformly, because the bottom singular direction can rotate with depth: writing $v_\ell$ for the bottom-of-active-spectrum direction of $X_\ell$, the trained-network reading of Corollary~\ref{cor:sigma-min-res} assumes
\emph{(uniform non-cancellation)} there exists $c > 0$, independent of $\ell$, such that at every block the residual-branch contribution satisfies $\langle X_\ell + F(X_\ell),\, v_\ell \rangle$-alignment bounded below, i.e.\ the branch does not destructively interfere with the identity along $v_\ell$ at leading order at any depth. Controlling one aligned coordinate at one block does not control the rotating direction at later blocks; the uniform form is what the depth-invariance conclusion actually consumes, and it is the checkpoint-falsifiable object (per-block, per-checkpoint). Residual-stream $\sigma_{\min}$ depth-invariance observed on pretrained transformers is the empirical record of this hypothesis holding, and the released checkpoints on which it fails are the recorded falsifications{} \citep{shirodkar2026algebraicdeaddirectionslayernorm}; a rate-level guarantee from structural assumptions on $F_\ell$ (Lipschitz contraction, angular bounds) remains future work.
\end{remark}

\paragraph{Corollaries and concrete cases.}

\begin{corollary}[Pure feedforward]
\label{cor:res_ff}
With no residual edges, $K(\ell) = L - \ell$, recovering Theorem~\ref{thm:bridge}.
\end{corollary}

\begin{corollary}[Per-layer residuals at every layer]
\label{cor:res_all}
If every layer is wrapped in a per-layer residual, $K(\ell) = 0$ for all $\ell$ (the shortest path takes every identity skip). The leading-order rate collapses to $\Theta(1)$; the rate-carrying signal degenerates to a finite-$t$ correction.
\end{corollary}

\begin{corollary}[Single block residual over a $k$-layer sub-chain]
\label{cor:res_block}
Consider a feedforward chain of $L$ weight layers (weight $W_i$ maps $v_{i-1} \to v_i$) with exactly one additive-identity residual block spanning weights $W_p, W_{p+1}, \ldots, W_{p+k-1}$, that is, a skip edge from $v_{p-1}$ (block input) to $v_{p+k-1}$ (block output) coexists with the weight chain $v_{p-1} \to v_p \to \cdots \to v_{p+k-1}$. Let $K_\mathrm{post} := L - (p+k-1)$ be the post-block feedforward distance. Then the shortest-weighted-path distance is
\[
K(\ell) = \begin{cases}
L - \ell, & \text{if } \ell \ge p+k-1 \text{ (at or after block exit)}, \\
K_\mathrm{post} + (p+k-1 - \ell), & \text{if } p \le \ell < p+k-1 \text{ (inside block)}, \\
K_\mathrm{post} + (p - 1 - \ell), & \text{if } \ell \le p - 1 \text{ (at or before block input)}.
\end{cases}
\]
Intuition: the skip lets any backward path that reaches $v_{p-1}$ bypass the entire $k$-weight block via the identity edge (cost 0 instead of $k$), saving exactly $k$ weights. A path with target $\ell$ strictly inside the block ($p \le \ell \le p+k-2$) cannot use the skip (the skip's entrance is at $v_{p-1}$, below $\ell$'s position), so it must traverse the weights from $v_{p+k-1}$ back through $W_{p+k-1}, \ldots, W_{\ell+1}$ into the block, paying $(p+k-1) - \ell$ weights plus the post-block $K_\mathrm{post}$. The exponent $2K(\ell)$ is the leading order as $t \to 0$; the skip adds to the dead-direction entry a factor $(1 + t^k)^2$ at every layer at or below the block input, which is $1 + O(t^k)$ and leaves the exponent unchanged.
\end{corollary}

\begin{remark}[The exact skip factor, and what a finite window reads]
\label{rem:res_block_skip_factor}
For $L = 4$, $p = 3$, $k = 2$ the formula gives $K = (1, 0, 1, 0)$, so $\alpha = (2, 0, 2, 0)$. In the linear chain the backward signals on the dead unit are exact: with $\delta_4 = r$ the output residual, $\delta_3 = t\,\delta_4$ (one weight), $\delta_2 = (1 + t^2)\,\delta_4$ (the identity skip plus the two-weight path through the block), and $\delta_1 = t(1 + t^2)\,\delta_4$, so the dead entries are $G_4 \propto 1$, $G_3 \propto t^2$, $G_2 \propto (1 + t^2)^2$, $G_1 \propto t^2(1 + t^2)^2$. A parametric freeze-probe on this chain with the grid $t \in [0.063, 0.5]$ reports $\alpha = (2.20, 0.20, 2.02, 0.02)$; the $0.2$ on the first two layers is the log-slope of $(1 + t^2)^2$ on that window, $4t^2/(1 + t^2)$, and the same regression on the exact factor returns $2.18$ and $0.18$. The offset is a finite-window artefact of the next-order path, not a correction to the rate; on a window closer to $t = 0$ it vanishes. The ReLU chain gives $(2.19, 0.19, 2.00, 0.00)$, since the skip path is linear.
\end{remark}

\paragraph{Scope.} Theorem~\ref{thm:bridge_res} covers any additive-identity residual DAG. It does \emph{not} cover: pre-activation residuals with learnable projection (where the skip is not pure identity), LayerNorm, attention's softmax nonlinearity inside $F$, or \emph{gated} residuals (highway networks, GRU- and LSTM-style gates, Mamba-style selective skips). For the first, the projection matrix on the skip path is itself a perturbed weight, so the skip edge has cost 1 rather than 0 in the graph distance; $K(\ell)$ remains integer-valued, but the rate formula $2K(\ell)$ now counts the projection weights too. The formalism extends directly with this reassignment of edge costs. For gated residuals of the form $X_{\mathrm{out}} = (1-g) \odot X_{\mathrm{in}} + g \odot F(X_{\mathrm{in}})$ where the gate $g$ is itself learned, the skip edge cost in $K(\ell)$ depends on whether $g$ stays $\Theta(1)$ or admits a dead-direction perturbation; a separate edge-cost analysis tracking gate factors would be required, and we do not provide it here. For LayerNorm and attention's softmax, see the corresponding extensions below.

\ifexpanded
\paragraph{Activation-side dual: \texorpdfstring{$\sigma_{\min}$}{sigmamin} under residuals.} The A-factor / $\sigma_{\min}$ dual of Theorem~\ref{thm:bridge_res} is obtained by mirroring the shortest-weighted-path construction on the \emph{forward} graph.

\begin{definition}[Forward shortest-weighted-path distance $K^{\mathrm{fwd}}(\ell)$]
\label{def:path_distance_fwd}
Let $G_\mathrm{fwd}$ be the residual computational graph with each weight edge cost $1$ and each identity-skip edge cost $0$. Define $K^{\mathrm{fwd}}(\ell)$ as the minimum total cost of any directed forward path in $G_\mathrm{fwd}$ from $v_0$ to $v_\ell$: the smallest number of \emph{weights} that must be traversed from the input to reach node $v_\ell$ when every available residual skip is taken.
\end{definition}

\begin{lemma}[Forward dead-component under residuals]
\label{lem:forward-dead-res}
Along the symmetric transversal approach of \S\ref{app:bridge_res} on a residual computational graph, in activation classes (P1), (P3), and (P2) with $\phi'(0) > 0$, with input $x \sim \mathcal{N}(0, I_h)$ so that $X_0^{(h)} = \Theta(1)$ a.s., the dead-direction component of the post-activation at node $v_\ell$ satisfies
\[
X_\ell^{(h)}(\theta(t)) \;=\; c_\ell \cdot t^{K^{\mathrm{fwd}}(\ell)} \cdot X_0^{(h)} + O\bigl(t^{K^{\mathrm{fwd}}(\ell)+1}\bigr),
\]
where $c_\ell > 0$ is the sum of positive dead-direction path constants over the shortest forward paths $\pi \in \mathcal{P}^*_\mathrm{fwd}(\ell) = \{\pi : v_0 \to v_\ell, |\pi|_W = K^{\mathrm{fwd}}(\ell)\}$. In particular, $\mathbb{E}[(X_\ell^{(h)})^2] = \Theta(t^{2 K^{\mathrm{fwd}}(\ell)})$.
\end{lemma}

\begin{proof}
Mirror of Lemma~\ref{lem:backward-dead-res} on the forward graph. By the forward chain rule, $X_\ell^{(h)}$ decomposes as a sum over all directed paths $\pi$ from $v_0$ to $v_\ell$, with each path contributing $J_\pi \cdot X_0^{(h)}$ where $J_\pi = \Theta(t^{|\pi|_W})$: each weight edge contributes a dead-direction factor of $t$ (from the perturbation $W_i(t)_{h,h} = t$ and the canonical $(W_i^*)_{h,h} = 0$), and each identity skip contributes $1$. In the activation classes of the lemma every factor is non-negative, so no destructive interference occurs: the sum over shortest paths $\mathcal{P}^*_\mathrm{fwd}(\ell)$ has strictly positive coefficient $c_\ell$. For (P1) linear, this is exact. For (P2) smooth, each traversed weight edge additionally applies $\phi$ at a pre-activation of magnitude $\Theta(t)$ or smaller; Taylor expanding $\phi(u) = \phi'(0) u + O(u^2)$ gives a multiplicative $(\phi'(0))^{K^{\mathrm{fwd}}(\ell)} (1 + O(t))$ factor, which preserves the leading rate and the sign under the lemma's $\phi'(0) > 0$. The sign restriction is needed exactly when two \emph{tied} shortest paths reach the same node carrying \emph{different numbers of activations}, since then their coefficients are unequal powers of $\phi'(0)$ and a negative $\phi'(0)$ can cancel the tie at leading order. On the graph with weight edges $W_1$ (activated), $W_2$, $W_3$ and skips $(v_0, v_2)$, $(v_1, v_3)$, where $K^{\mathrm{fwd}}(3) = 1$ is attained by two paths whose coefficients are $\phi'(0)$ and $1$, the activation $\phi(u) = -u + u^3$ gives $\expect[(X_3^{(h)})^2] = \Theta(t^6)$ against the predicted $\Theta(t^2)$, and the backward second moment moves the same way; the identical construction at $\phi'(0) = +1$ returns $\Theta(t^2)$ as stated. Two cases are therefore insensitive to the sign and satisfy the lemma for either sign of $\phi'(0)$: a skipless chain, where a single path carries the whole rate and the second moment discards its sign, and a residual DAG whose weight edges are \emph{uniformly} activated, where tied shortest paths carry equally many activations and their coefficients share a common factor $(\phi'(0))^{K^{\mathrm{fwd}}(\ell)}$. The same graph with $\phi$ on every weight edge returns $\expect[(X_3^{(h)})^2] = 4t^2 + O(t^4)$ at $\phi'(0) = -1$. Practical residual networks apply the same activation on every weight edge, so the hypothesis is conservative there; it is stated because the lemma quantifies over graphs that mix activated and linear edges. For (P3) ReLU, the gate at each weight edge aligns with $\mathrm{sign}(X_0^{(h)})$ under canonical coordinates: on the event $\{X_0^{(h)} > 0\}$ (probability $1/2$ under Gaussian input) the chain activates with leading coefficient $1$, and on $\{X_0^{(h)} < 0\}$ the chain zeroes at the first weight edge (contributing $0$ to $X_\ell^{(h)}$, but the skip-only path, if any, still delivers $X_0^{(h)}$). Taking the second moment yields $\mathbb{E}[(X_\ell^{(h)})^2] = \Theta(t^{2 K^{\mathrm{fwd}}(\ell)})$ in all three activation classes.
\end{proof}

\begin{corollary}[Activation $\sigma_{\min}$ on residual networks]
\label{cor:sigma_min_res_g5}
Under the residual computational graph setup and activation classes (P1), (P3), and (P2) with $\phi'(0) > 0$ (the scope of Lemma~\ref{lem:forward-dead-res} and Theorem~\ref{thm:bridge_res}, which this corollary reads off: at $\phi'(0) < 0$ two tied shortest paths, one through an activation and one through a skip, cancel and the rate jumps), at any node $v_\ell$ not post-composed with a normalization layer that would zero the row-mean, for $X_\ell \in \mathbb{R}^{N \times h}$ the matrix of post-activation outputs at $v_\ell$ over $N$ samples,
\[
\sigma_{\min}\bigl(X_\ell(\theta(t))\bigr) \;=\; \sqrt{N} \cdot \Theta\bigl(t^{K^{\mathrm{fwd}}(\ell)}\bigr),
\]
where $K^{\mathrm{fwd}}(\ell)$ is the forward shortest-weighted-path distance (Definition~\ref{def:path_distance_fwd}) from $v_0$ to $v_\ell$. With no skip edges, $K^{\mathrm{fwd}}(\ell) = \ell$, recovering Corollary~\ref{cor:sigma-min}. At post-LN nodes, $\sigma_{\min}(\mathrm{LN}(X_\ell)) = 0$ identically (LN's mean-subtraction null direction $\mathbf{1}_d / \sqrt{d}$, equivalently $\gamma^{-1}/\|\gamma^{-1}\|$ for trained $\gamma$ per Proposition~\ref{prop:ln_kernel}), so the corollary's prediction does not apply at those sites; see the post-LN-vs-block-output discussion below.
\end{corollary}

\begin{proof}
By Lemma~\ref{lem:forward-dead-res}, the dead channel of $X_\ell$ has squared magnitude $\Theta(t^{2 K^{\mathrm{fwd}}(\ell)})$ per sample. Non-dead channels of $X_\ell$ have magnitude $\Theta(1)$, independent of $t$ at leading order, by the same argument as the non-dead analysis in Theorem~\ref{thm:bridge_res}. Canonical-coordinate independence of $x^{(h)}$ from non-dead inputs yields off-diagonal vanishing of $X_\ell^\top X_\ell / N$ at leading order (residual extension of Lemma~\ref{lem:integral-reduction-sub}). Hence $X_\ell^\top X_\ell / N$ is block-diagonal in canonical coordinates with dead entry $\Theta(t^{2 K^{\mathrm{fwd}}(\ell)})$ and non-dead entries $\Theta(1)$; its smallest eigenvalue is the dead entry, giving $\sigma_{\min}(X_\ell)/\sqrt{N} = \Theta(t^{K^{\mathrm{fwd}}(\ell)})$.
\end{proof}

\begin{corollary}[Per-layer residuals: $\sigma_{\min}$ preserved]
\label{cor:sigma-min-res-all}
If every layer is wrapped in an identity-skip residual, then $K^{\mathrm{fwd}}(\ell) = 0$ for all $\ell \ge 0$ (the skip chain from $v_0$ reaches $v_\ell$ via identities only). Hence $\sigma_{\min}(X_\ell) = \Theta(\sqrt{N})$, constant in $t$ at leading order: the residual stream preserves the input $\sigma_{\min}$ signal through the entire depth.
\end{corollary}

\paragraph{Remark on forward/backward duality under residuals.} On pure feedforward chains, $K^{\mathrm{fwd}}(\ell-1) = \ell - 1$ and $K(\ell) = L - \ell$, so $K^{\mathrm{fwd}}(\ell-1) + K(\ell) = L - 1$, reproducing Corollary~\ref{cor:a_g_duality}. Under residuals this generalizes to
\[
\lambda_{\min}(A_\ell) \cdot \lambda_{\min}(G_\ell) \;=\; \Theta\bigl(t^{2 (K^{\mathrm{fwd}}(\ell-1) + K(\ell))}\bigr),
\]
which is \emph{layer-dependent} when skips are present: a layer whose input or output is bypassed by a skip sees a reduced exponent. In particular, with per-layer residuals everywhere, both $K^{\mathrm{fwd}} \equiv 0$ and $K \equiv 0$, and the A-G product becomes $\Theta(1)$: no rate signal remains in either factor.

\paragraph{Remark on biases.} Corollary~\ref{cor:sigma-min-res} is stated for the unbiased setup inherited from \S\ref{app:bridge_res}. Under the bias augmentation of Theorem~\ref{thm:bridge_bias}, each bias $b_\ell$ contributes an additive $\Theta(t)$ to the dead-direction pre-activation at $v_\ell$ (the augmented-weight column perturbation is $t$). For $\ell$ with $K^{\mathrm{fwd}}(\ell) \ge 2$, this bias contribution would dominate the weight-chain contribution at leading order, capping the effective forward dead-direction rate at $1$. Thus the $\sigma_{\min}$ corollary is \emph{not} directly bias-stable in the way $G$-factor rate is. In practice this matters only when $K^{\mathrm{fwd}}(\ell) \ge 2$; under per-block residuals (the pre-norm transformer case), $K^{\mathrm{fwd}} \equiv 0$ at block outputs and biases do not disturb the rate-$0$ conclusion.

\paragraph{Remark on LayerNorm inside residual sub-paths.} Pre-norm transformer blocks take the form $X_{\mathrm{out}} = X_{\mathrm{in}} + F(\mathrm{LN}(X_{\mathrm{in}}))$ where $F = W_2 \cdot \phi(W_1 \cdot \,\cdot\,)$ is a two-weight feedforward. LN applied to $X_{\mathrm{in}}$ mean-centers and variance-normalizes row-wise; the dead channel of $\mathrm{LN}(X_{\mathrm{in}})$ is $(X_{\mathrm{in}}^{(h)} - \mu(X_{\mathrm{in}})) / \sigma(X_{\mathrm{in}})$, and because non-dead channels of $X_{\mathrm{in}}$ are $\Theta(1)$ while the dead channel is $\Theta(t^{K^{\mathrm{fwd}}(v_{\mathrm{in}})})$, the mean $\mu(X_{\mathrm{in}}) = \Theta(1)$ dominates, giving $\mathrm{LN}(X_{\mathrm{in}})^{(h)} = \Theta(1)$ regardless of the upstream rate. Propagating through $F$ yields $F(\mathrm{LN}(X_{\mathrm{in}}))^{(h)} = \Theta(t^{k_F})$ where $k_F$ is the number of weight layers in $F$ (rather than $\Theta(t^{K^{\mathrm{fwd}}(v_{\mathrm{in}}) + k_F})$ as would hold without LN). At the block output,
\[
X_{\mathrm{out}}^{(h)} \;=\; \underbrace{X_{\mathrm{in}}^{(h)}}_{\Theta(t^{K^{\mathrm{fwd}}(v_{\mathrm{in}})})} \;+\; \underbrace{F(\mathrm{LN}(X_{\mathrm{in}}))^{(h)}}_{\Theta(t^{k_F})},
\]
so the effective rate at $v_{\mathrm{out}}$ is $\min(K^{\mathrm{fwd}}(v_{\mathrm{in}}), k_F)$: graph-theoretic $K^{\mathrm{fwd}}$ on the residual DAG if the skip wins, or $k_F$ if the LN-reset sub-path wins. For the standard pre-norm stack with per-block residuals, $K^{\mathrm{fwd}}(v_{\mathrm{in}}) = 0$ at every block boundary (the skip chain from $v_0$ reaches every block input), so the skip dominates and Corollary~\ref{cor:sigma-min-res}'s prediction $\sigma_{\min}(X_{\mathrm{block\_out}}) = \Theta(\sqrt{N})$ at rate $0$ holds exactly, independent of $k_F$. (This analysis assumes the LN affine parameters at default $\gamma = 1, \beta = 0$. Under trained $\gamma \neq 1$, the dead direction of LN's forward map is $\gamma^{-1}/\|\gamma^{-1}\|$ rather than $\mathbf{1}_d / \sqrt{d}$ per Proposition~\ref{prop:ln_kernel}; the rate-$0$ skip-bypass conclusion at the residual stream is unaffected, but the dead direction itself rotates with $\gamma$.)

Two distinct measurement sites must be distinguished:
\begin{itemize}
\item \emph{Block output (post-residual-add, pre-next-LN):} $\sigma_{\min}(X_{\mathrm{block\_out}})$ tracks $\min(K^{\mathrm{fwd}}(v_{\mathrm{in}}), k_F)$. With per-block residuals this reduces to $K^{\mathrm{fwd}}$ on the residual DAG, consistently rate $0$. This matches the empirical signature where 2-, 3-, and 4-block LN stacks show $\sigma_{\min}$ log--log slope $\approx 0$ at every block output while $u^\top G u$ gives fractional non-integer rates in the bound-gap of Proposition~\ref{thm:bridge_ln}.
\item \emph{Post-LN (inside block, at LN output):} LN outputs are mean-centered row-wise, so every row of $\mathrm{LN}(X) \in \mathbb{R}^{N \times h}$ is orthogonal to $\mathbf{1}_h$ and the uniform direction $\mathbf{1}_h/\sqrt{h}$ spans a null direction of the matrix itself, $\mathrm{LN}(X)\,\mathbf{1}_h = 0$. Hence $\sigma_{\min}(\mathrm{LN}(X)) = 0$ identically, an architectural null-space degeneracy independent of $t$. Post-LN $\sigma_{\min}$ is therefore \emph{not} a valid rate observable without first projecting out the uniform direction and measuring $\sigma_{\min}$ on the $(h-1)$-dimensional range.
\end{itemize}
Consequently, the activation-side rate observable on residual+LN architectures is well-defined only at the residual-stream content (block-output sites, before the next LN application).

\paragraph{Scope: post-norm LN and attention.} The remark above assumes pre-norm LN (LN inside each residual block's sub-path $F$). In \emph{post-norm} architectures $X_{\mathrm{out}} = \mathrm{LN}(X_{\mathrm{in}} + F(X_{\mathrm{in}}))$, LN acts on the residual-add output, so $X_{\mathrm{out}}$ is mean-centered row-wise and $\sigma_{\min}(X_{\mathrm{out}}) = 0$ identically. Post-norm stacks therefore admit no rate-bearing $\sigma_{\min}$ observable at block outputs without explicit projection onto $\{\mathbf{1}\}^\perp$. Softmax-based attention inside $F$ does not affect Corollary~\ref{cor:sigma-min-res} at block outputs (the skip bypasses $F$ entirely), but the in-block dead-direction rate through softmax remains open.

\paragraph{Composition additivity on heterogeneous blocks.}
Theorem~\ref{thm:bridge_res} establishes the $K$-distance framework when the graph's elementary operations are weight matmuls (cost $1$) and identity skips (cost $0$). Real architectures compose higher-level \emph{blocks} (attention, MLP FFN, pre-norm residual sub-chains), each with its own internal bridge analysis. The composition rule for these is Theorem~\ref{thm:bridge_composition} (\S\ref{app:proof:bridge_composition}): for a sequential composition $M = B_n \circ \cdots \circ B_1$ with per-block backward and forward rates $k_i^{\mathrm{bk}}, k_i^{\mathrm{fwd}}$ (Definition~\ref{def:block_rate}) on a shared canonical dead direction, and under the scalar-transfer hypothesis, the backward rate at the input of $B_i$ is $\sum_{j \ge i} k_j^{\mathrm{bk}}$ and the forward rate at its output is $\sum_{j \le i} k_j^{\mathrm{fwd}}$.

The block rates instantiate directly: a feedforward chain of $k$ weights has $k_B^{\mathrm{bk}} = k_B^{\mathrm{fwd}} = k$; a pre-norm residual block (\S\ref{app:bridge_res}) with an identity skip has $k_B^{\mathrm{bk}} = k_B^{\mathrm{fwd}} = 0$ (skip-dominated); a single weight matmul has $k = 1$; and an attention block has a standalone backward rate $k_{\mathrm{attn}}^{\mathrm{bk}} = 2$ (\S\ref{app:bridge_attn}), which is a per-block quantity and \emph{not} a summand: attention fails Theorem~\ref{thm:bridge_composition}'s transfer hypothesis at every depth, so a chain containing attention blocks is governed by Proposition~\ref{prop:attn_chain_softmax}, not by adding this rate (see the reach discussion at Corollary~\ref{cor:g10_via_composition}).

\begin{corollary}[Reduction to the basic and residual bridges]
\label{cor:composition_reduces}
Setting each $B_i$ to a single weight matmul ($k_i^{\mathrm{bk}} = k_i^{\mathrm{fwd}} = 1$) recovers Theorem~\ref{thm:bridge} with $K(\ell) = L - \ell$. Setting each $B_i$ to a pre-norm residual block ($k_i = 0$) gives rate $0$ everywhere, matching Corollary~\ref{cor:sigma-min-res-all}. Setting each $B_i$ to a feedforward sub-chain of $k_i$ weights gives $K(\ell) = \sum_{j \ge i} k_j$, equivalent to Theorem~\ref{thm:bridge_res}'s graph-distance $K(\ell)$ on the flattened weight graph.
\end{corollary}

\begin{corollary}[Heterogeneous block composition rates on MLP chains: empirical validation of Theorem~\ref{thm:bridge_composition}]
\label{cor:composition_heterogeneous}
For a sequential (non-residual) chain of $n$ MLP blocks with per-block rate $k_{\mathrm{mlp}}^{\mathrm{bk}} = 2$, Theorem~\ref{thm:bridge_composition} gives a backward block-rate sum $K = 2(n - i + 1)$ at the input of block $B_i$, so the Fisher log--log slope there is $\alpha = 2K = 4(n - i + 1)$ (Definition~\ref{def:block_rate} puts the block rate on the gradient amplitude). Parametric freeze-probe measurements at $n \in \{6, 8, 12\}$ with $d \in \{16, 64, 128\}$, 3 seeds per configuration (15 configurations validated within numerical range), give rates matching the prediction to $< 0.1$ at all tested probe positions (e.g., at $n = 8$ and probe index $\ell$: measured $\alpha_{\mathrm{fc2}} = 28.000, 12.000, 0.000$ for $\ell = 0, 4, 7$; predicted $4 \cdot (n - 1 - \ell) = 28, 12, 0$ to the displayed precision). The single configuration outside numerical range ($n=12$, $\ell=0$: predicted $\alpha = 44$, signal below fp64 at $t = 10^{-2}$) is a precision limit of the probe, not a theorem failure.
\end{corollary}

\begin{proposition}[$W_V$--$W_O$ rate-difference invariant]
\label{prop:attn_VO_invariant}
For a probe attention block embedded in a composition, under canonical init with shared dead direction $u = e_d$, symmetric perturbation $W_V, W_O \to W^* + t \cdot e_d e_d^\top$, and the embedding non-degeneracy hypothesis that the attention average does not annihilate the upstream dead gradient,
\[
\expect\bigl[(A_0^\top \tilde\Delta)^2\bigr] \;=\; \Theta\bigl(\expect[\tilde\Delta^2]\bigr),
\qquad \tilde\Delta := \Delta\, e_d \text{ the dead column of the gradient at the block output},
\]
the freeze-probe rates on $W_V$ and $W_O$'s grad-output dead-channel Fisher satisfy
\[
\alpha_{W_V} - \alpha_{W_O} \;=\; 2
\]
exactly, independently of composition depth. The hypothesis holds in canonical-aligned compositions (every probe position of the measured chains validates it) and can fail in engineered embeddings: a surrounding pair that zeroes the attention scores and token-centers $\Delta$ makes $A_0$ exactly uniform and $A_0^\top \tilde\Delta$ exactly zero, replacing the $W_V$ side by an exact zero with no rate to fit.
\end{proposition}

\begin{proof}
Let $\Delta := \partial L / \partial Y_{\mathrm{block}}$ denote the dead-direction gradient flowing into the probe block's output. The grad-output of $W_O$ (whose layer output is $Y_{\mathrm{block}}$) is $\Delta$ itself. The grad-output of $W_V$ (whose layer output is $V = W_V X_{\mathrm{blk\_in}}$) follows the chain $W_V \to V \to AV \to Y_{\mathrm{block}}$:
\[
\partial L / \partial V \;=\; A^\top \cdot (\partial L / \partial (AV)) \;=\; A^\top \cdot \Delta \cdot W_O,
\]
using $Y_{\mathrm{block}} = (AV) W_O^\top$. Project both gradients onto $u = e_d$:
\[
(\Delta \cdot u)_{n, :} \;=\; \Delta_{n, d} \, e_d, \qquad (A^\top \Delta W_O) \cdot e_d \;=\; A^\top \cdot \Delta \cdot (W_O\, e_d).
\]
At canonical init with $W_O^* + t \cdot e_d e_d^\top$, the $d$-th column of $W_O$ is $t \cdot e_d$, so $W_O\, e_d = t \cdot e_d$. Hence
\[
(\partial L / \partial V) \cdot e_d \;=\; t \cdot A^\top \cdot \Delta \cdot e_d \;=\; t \cdot A^\top \cdot (\Delta e_d).
\]
Taking squared mean over $(n, m)$ tokens (the Fisher factor):
\[
u^\top G_{W_V} u \;=\; \expect\bigl[(A^\top \Delta e_d)^2\bigr] \cdot t^2 \;=\; t^2 \cdot (u^\top G_{W_O} u) \cdot c,
\]
with $c = \expect[(A^\top \tilde\Delta)^2] / \expect[\tilde\Delta^2]$, which the embedding non-degeneracy hypothesis makes positive and $O(1)$ (row-stochasticity of $A$ alone gives no lower bound: a uniform $A_0$ annihilates a token-centered $\tilde\Delta$). Taking logarithm and fitting rate in $t$: $\alpha_{W_V} = 2 + \alpha_{W_O}$. The constant $c$ contributes to the Fisher prefactor, not the slope.
\end{proof}

\begin{remark}[The invariant works as a stable probe of attention internal structure]
\label{rem:VO_invariant_use}
Proposition~\ref{prop:attn_VO_invariant} is stable across all tested configurations: $n \in \{1, 4, 6\}$ attention chains, with or without LN, $d \in \{16, 64, 128, 768\}$, $n_h \in \{1, 2, 4, 8\}$. The observed differences $\alpha_{W_V} - \alpha_{W_O}$ are $2.000 \pm 0.000$ in every case, even where the individual rates $\alpha_{W_V}$ and $\alpha_{W_O}$ exhibit the softmax-coupling anomalies of Remark~\ref{rem:attn_composition_anomaly}. The invariant follows from the sequential $W_V \to W_O$ structure within each attention block (one Linear's worth of $t$-factor in grad-output) and is therefore preserved under any deformation of the rest of the network that keeps the embedding non-degeneracy hypothesis, that is, any deformation under which the attention average does not annihilate the upstream dead gradient. This makes it a clean diagnostic observable: measuring $\alpha_{W_V} - \alpha_{W_O}$ and confirming it equals $2$ is a lightweight consistency check for the attention block's canonical structure.
\end{remark}

\begin{remark}[Residual bypass: the softmax-coupling anomaly is moot in practical transformers]
\label{rem:residual_bypass_softmax_coupling}
The softmax-coupling anomaly of Remark~\ref{rem:attn_composition_anomaly} is a property of non-residual sequential attention stacks, which do not appear in practical architectures. Every attention block in ViT, GPT, and LLaMA-family transformers is residual-wrapped pre-norm: $y = x + \mathrm{Attn}(\mathrm{LN}(x))$. Post-norm designs (BERT) wrap the same skip inside a trunk normalisation, $y = \mathrm{LN}(x + \mathrm{Attn}(x))$, which places them outside Corollary~\ref{cor:sigma-min-res}'s graph model (its scope paragraph); the bypass statement below covers the pre-norm list. Under Corollary~\ref{cor:sigma-min-res}, the residual skip dominates: $k_{\mathrm{block}}^{\mathrm{res}} = 0$ at the block-output level (forward $K$-distance $0$ via the identity path), regardless of what happens inside $\mathrm{Attn}$. Consequently, rate predictions at the residual stream of a practical transformer are governed by the skip, not by the attention internals; the softmax-coupling anomaly is bypassed entirely. The anomaly \emph{does} affect the internal component rates ($W_O, W_V, W_Q, W_K$) within each block, but those are not the primary predictive observables at practical scale: $\sigma_{\min}$ on the residual stream is.
\end{remark}

\begin{corollary}[Attention as a composition corollary of its standalone rate]
\label{cor:g10_via_composition}
The attention block's standalone backward rate $k_{\mathrm{attn}}^{\mathrm{bk}} = 2$ is established in \S\ref{app:bridge_attn} (Theorem~\ref{thm:bridge_attn_backward}). Attention fails Theorem~\ref{thm:bridge_composition}'s transfer hypothesis at every depth, since its backward transfer is a context-dependent token-space action, so the composition theorem does not deliver the rate for a chain of attention blocks. What it does deliver is the residual transformer stack, where Corollary~\ref{cor:sigma-min-res}'s identity skip sets the residual-stream rate to $0$ whatever the weight edges carry (via Corollary~\ref{cor:composition_reduces} with residual-block rates being $0$), and the composed rate of a mixed chain only where each block's additive exponent stays below the score-path injection floor of Proposition~\ref{prop:attn_chain_softmax}(a). A chain of two attention blocks followed by two MLP blocks measures $7.98$ at the first block's $W_O$ against the additive $12$, so a bound on the attention count alone does not certify additivity.
\end{corollary}

\paragraph{Remark on block-rate canonical alignment (NDC).} Theorem~\ref{thm:bridge_composition} assumes each block shares the canonical dead direction $u = e_h$. If blocks use different canonical coordinates (e.g., a rotation is applied between blocks), the composition rate falls \emph{below} $\sum_j k_j$: a rotation that moves $e_h$ mixes $\Theta(1)$ non-dead signal into the downstream block's dead coordinate, and that leak, not the decayed dead channel, sets the rate. Two smooth MLP blocks of rate $2$ each compose at $2$ under a $0.3$-radian inter-block rotation, against $4$ when aligned; rotations that fix $e_h$ leave additivity intact. This is the Non-Dead-Canonical (NDC) assumption: the dead direction is coherent across composed blocks. Proposition~\ref{prop:bridge_linear_rot} (the rotation case) handles the per-block rotation separately; composition with rotated blocks follows by treating each rotated block as a separate rate-carrying unit.

\begin{figure}[ht]
\centering
\includegraphics[width=\textwidth]{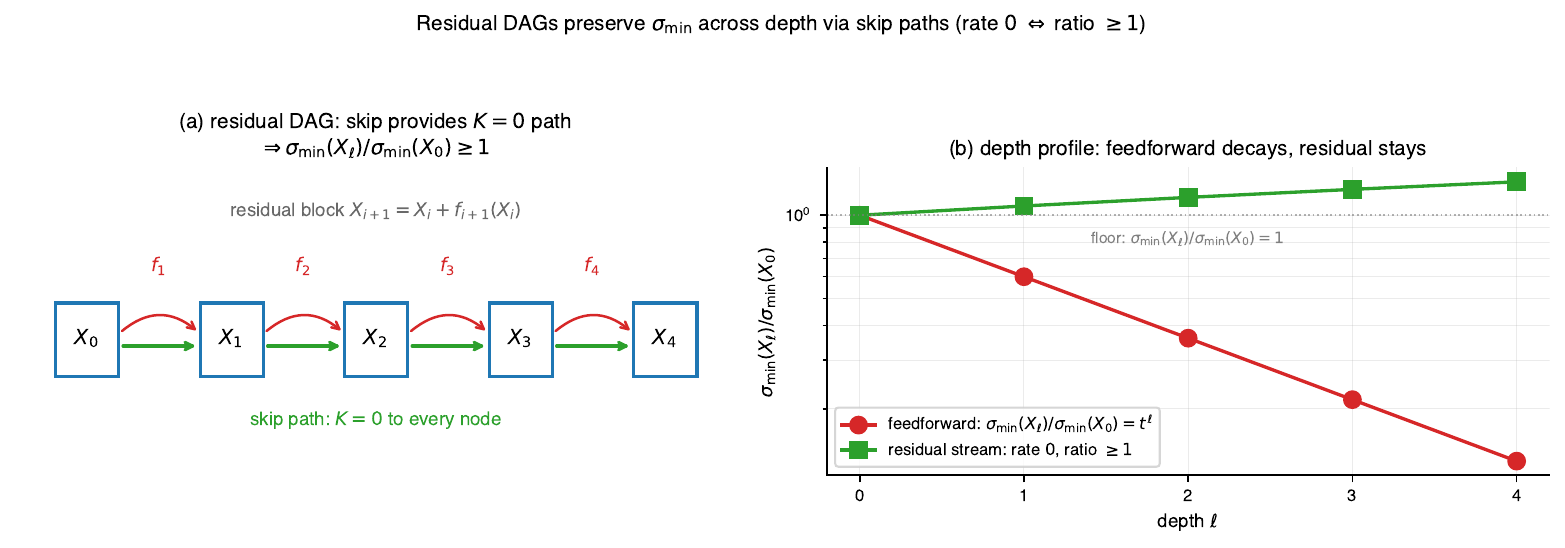}
\caption{Residual-DAG $\sigma_{\min}$ depth-invariance (Corollary~\ref{cor:sigma-min-res}). The mechanism the corollary identifies. (a)~In a residual block $X_{i+1} = X_i + f_{i+1}(X_i)$, the additive identity skip provides a forward-$K$-distance-zero route from $X_0$ to every node, so the dead-direction rate is $0$ at every depth; under the no-cancellation hypothesis the profile satisfies $\sigma_{\min}(X_\ell) / \sigma_{\min}(X_0) \ge 1$. (b)~The depth profile contrast: a feedforward chain (no skips) decays as $t^\ell$; a residual stream stays at the floor (rate $0$). On a trained transformer the same observable is a per-checkpoint structural-health diagnostic: a measured intermediate dip or monotone decline is a falsifier of the no-cancellation hypothesis.}
\label{fig:kfac_bridge_residual_dag}
\end{figure}
\fi  \begin{corollary}[\texorpdfstring{Bottom-of-active-spectrum depth-invariance on residual DAGs}{Bottom-of-active-spectrum depth-invariance on residual DAGs}]
\label{cor:sigma-min-res}
On a residual computational graph with \emph{exact-identity} skip connections (weight edges of cost $1$, identity-skip edges of cost $0$; gated or highway residuals are outside this scope), under the canonical-aligned approach $\theta(t)$\ifsupp\else{} of Theorem~\ref{thm:bridge}\fi, in activation classes (P1), (P3), and (P2) with $\phi'(0) > 0$ (for $\phi'(0) < 0$ odd-weight paths carry negative coefficients and the non-cancellation below fails; Remark~\ref{rem:sigma_min_res_trained}), with input residual stream $X_0$ of rank $r_0 := \rank(X_0)$,
\[
\sigma_{(r_0)}(X_\ell(\theta(t))) \;=\; \sqrt{N} \cdot \Theta\bigl(t^{K^{\mathrm{fwd}}(\ell)}\bigr),
\]
where $\sigma_{(k)}$ denotes the $k$-th-largest singular value, so $\sigma_{(r_0)}$ is the smallest \emph{active} (non-numerical-zero) singular direction of $X_0$. For full-rank inputs ($r_0 = d$), $\sigma_{(r_0)} = \sigma_{\min}$ and the statement reduces to $\sigma_{\min}(X_\ell(\theta(t))) = \sqrt{N} \cdot \Theta(t^{K^{\mathrm{fwd}}(\ell)})$. With per-layer identity-skip residuals (e.g., pre-norm transformer blocks, where the skip is exact identity; post-norm blocks place a trunk normalisation after the adder and fall outside this graph model, alongside the Peri-LN case in the scope paragraph of \S\ref{app:proof:sigma_min_res}), $K^{\mathrm{fwd}}(\ell) \equiv 0$ for all $\ell$, so the residual stream preserves the input's bottom-of-active-spectrum signal at rate $0$ through the entire depth.
\end{corollary}

\begin{proof}[Sketch]
The forward chain rule decomposes $X_\ell^{(h)}$ as a sum over directed paths from $v_0$ to $v_\ell$, each contributing $t^{|\pi|_W}$ with non-negative coefficient; the shortest path dominates, giving $X_\ell^{(h)} = \Theta(t^{K^{\mathrm{fwd}}(\ell)})$. The argument applies path-by-path to any non-cancelled bottom-of-spectrum direction, hence the rank-aware $\sigma_{(r_0)}$ form. Full proof: \S\ref{app:proof:sigma_min_res}.
\end{proof}

\paragraph{Three readings of the depth-invariance.}\label{rem:sigma_min_res_three_readings}
The same content takes three useful forms. \emph{As a graph distance.} The dead-direction rate at any node $v_\ell$ is determined by the shortest weighted path from $v_0$ to $v_\ell$, with weight edges costing $1$ and identity skips costing $0$. With per-layer skips everywhere, every node has $K^{\mathrm{fwd}}(\ell) = 0$ and the rate is $\Theta(1)$ across all depth. \emph{As a structural-health diagnostic.} The ratio $\sigma_{(r_0)}(X_\ell) / \sigma_{(r_0)}(X_0)$ should remain $\Theta(1)$ across a residual transformer's depth at any training checkpoint. A measured drop below the input (a pronounced intermediate dip or a monotone decline) is the falsifier: it indicates the no-cancellation hypothesis breaks for that architecture, that the skip's signal is being subtracted out by the residual branch in the bottom-of-active-spectrum direction. \emph{As a rate-collapse contrast.} A skipless pure-attention chain collapses this observable doubly-exponentially with depth; the residual skip instead preserves it at rate $0$, which is what the corollary makes precise.

\paragraph{Structural status of Corollary~\ref{cor:sigma-min-res}.}\label{rem:sigma_min_res_structural_status}
The mathematical content of Corollary~\ref{cor:sigma-min-res} is elementary: with an additive-identity skip $X_{\ell+1} = X_\ell + F(X_\ell)$ and a non-cancellation hypothesis on the residual branch, the bottom singular value cannot decrease at leading order. The shortest-weighted-path formalism is correct but follows from basic graph theory once the per-edge factorisation is in place. We state and label the result because it gives a single observable, $\sigma_{(r_0)}(X_\ell)/\sigma_{(r_0)}(X_0)$, on which the no-cancellation hypothesis is empirically falsifiable on a per-checkpoint basis (released checkpoints split into passes and failures against the input embedding, the failures an intermediate dip or a depth-decreasing profile; \citealp{shirodkar2026algebraicdeaddirectionslayernorm}); the corollary is best read as the analytic packaging of that diagnostic. The genuinely architectural content lies in the prerequisites of \ifsupp the bridge framework's setup hypotheses\else the bridge framework (Theorem~\ref{thm:bridge}, the architectural instantiations, the attention composition anomaly)\fi, which this corollary applies as a downstream observable.

\paragraph{Position relative to prior rank-collapse and neural-collapse work.}\label{rem:sigma_min_res_novelty}
The same activation-side observable whose doubly-exponential collapse in pure-attention chains \citet{DongCordonnierLoukas21} establish, with the training consequences analysed by \citet{NociAnagnostidisBiggio22}, is here shown to be \emph{rate-zero preserved} along any residual DAG: the identity skip is the operational mechanism that turns the rank-collapse signal into depth-invariance, recovering the empirical residual-stream stability observed in modern transformers. Neural collapse \citep{PapyanHanDonoho20} concerns class-structured geometry at the terminal phase of training (NC1 is within-class variability collapse; the class-mean simplex geometry is NC2); the corollary carries no class structure, so the relation is an analogy at the level of the observable family, as a per-layer during-training statement along a residual DAG. The rank-aware $\sigma_{(r_0)}$ form keeps the prediction measurable when the input residual stream's smallest active singular value is near the numerical floor.

\paragraph{Architectural vs training-conditional content.}\label{rem:sigma_min_res_architectural}
The rate-$0$ conclusion $\sigma_{(r_0)}(X_\ell)/\sigma_{(r_0)}(X_0) = \Theta(1)$ has two parts. Along the canonical trajectory it is a property of the additive-identity skip structure: the per-edge dead-direction Jacobians have the path-wise factorisation of the proof, so the identity skip dominates the bottom-of-active-spectrum. Off the canonical trajectory it holds exactly where the no-cancellation hypothesis (Remark~\ref{rem:sigma_min_res_trained}) holds, and that hypothesis is a per-checkpoint empirical property: at Kaiming random initialisation a single residual MLP block fails both the hypothesis and the conclusion on every draw (the branch output is uncorrelated with the stream, and its energy in the bottom singular directions dominates the alignment term), while the trained checkpoints discussed under the structural-status paragraph above split by architecture and training state. The pass/fail therefore diagnoses the architecture and its training state together. The further training-conditional content is the \emph{magnitude} of the depth profile (the amplification factor between consecutive blocks). When using the corollary as a structural-health probe at LLM scale, read the training signal off the magnitude and the pass/fail jointly.

\subsubsection{Proof of Corollary~\ref{cor:sigma-min-res}}
\label{app:proof:sigma_min_res}

Consider the residual computational graph $G_{\mathrm{fwd}}$ with weight edges of cost $1$ and identity-skip edges of cost $0$, and define $K^{\mathrm{fwd}}(\ell)$ as the minimum total cost of any directed forward path from $v_0$ to $v_\ell$.

\paragraph{Forward dead-direction-component decomposition on the residual DAG.} The dead-direction component $X_\ell^{(h)}$ decomposes as a sum over all directed paths $\pi : v_0 \to v_\ell$ in $G_{\mathrm{fwd}}$, with each path contributing $J_\pi \cdot X_0^{(h)}$ where $J_\pi$ is the product of the per-edge dead-direction Jacobian factors along $\pi$: each weight edge contributes a factor of $t$ (from $W_i(t)_{h,h} = t$ in canonical coordinates), and each identity skip contributes $1$. Hence $J_\pi = c_\pi \cdot t^{|\pi|_W}$ where $c_\pi$ is the dead-direction path constant and $|\pi|_W$ counts weights along $\pi$. The sign structure of $c_\pi$ is what the corollary's activation class controls: for (P1), and for (P2) with $\phi'(0) > 0$, every $c_\pi \ge 0$ (each weight edge contributes $\phi'(0)$ at leading order, so $c_\pi = (\phi'(0))^{|\pi|_W} > 0$); for (P3) ReLU $c_\pi \in \{0, 1\}$ (zero on inactive sub-paths). For (P2) with $\phi'(0) < 0$, which the corollary excludes, odd-weight paths carry negative constants and the non-cancellation used below fails. The decomposition is a claim about the dead-direction component specifically, not about the full network: in canonical coordinates the per-edge dead-direction Jacobian is scalar (factor $t$ for weights, $1$ for skips), and along the canonical-aligned trajectory the off-diagonal couplings into the dead direction vanish at leading order, so the per-edge contributions multiply along each path independently of the off-diagonal couplings.

For (P1) linear, all $c_\pi > 0$, so $X_\ell^{(h)} = (\sum_\pi c_\pi t^{|\pi|_W}) X_0^{(h)}$; the shortest-path contribution dominates: $X_\ell^{(h)} = c_\ell t^{K^{\mathrm{fwd}}(\ell)} X_0^{(h)} + O(t^{K^{\mathrm{fwd}}(\ell)+1})$ with $c_\ell > 0$ the sum over shortest paths. For (P2) smooth, each weight edge additionally applies $\phi$ at a pre-activation of magnitude $\Theta(t)$ or smaller; Taylor expanding $\phi(u) = \phi'(0) u + O(u^2)$ gives a multiplicative $(\phi'(0))^{K^{\mathrm{fwd}}(\ell)} (1 + O(t))$ factor whenever $\phi'(0) \ne 0$, preserving the leading rate; the corollary's residual statement takes $\phi'(0) > 0$ so these factors are positive and cannot cancel the skip. For (P3) ReLU, the gate at each weight edge aligns with $\mathrm{sign}(X_0^{(h)})$ by the canonical-aligned construction\ifsupp\else{} of Theorem~\ref{thm:bridge}\fi: on the event $\{X_0^{(h)} > 0\}$ the chain activates with leading coefficient $1$; on $\{X_0^{(h)} < 0\}$ the chain zeroes at the first weight edge but the skip-only path (if $K^{\mathrm{fwd}}(\ell) = 0$) still delivers $X_0^{(h)}$. Taking the second moment yields $\expect[(X_\ell^{(h)})^2] = \Theta(t^{2 K^{\mathrm{fwd}}(\ell)})$ in all three activation classes.

\paragraph{Activation matrix singular value.} The activation matrix $X_\ell \in \reals^{N \times h}$ has $X_\ell^\top X_\ell / N \to A_{\ell+1}$ a.s.\ by the strong law (the rows of $X_\ell$ are iid because the rows of $X_0$ are iid by the Gaussian-isotropic input assumption\ifsupp\else{} of Theorem~\ref{thm:bridge}\fi, and the per-row map through the canonical network is deterministic; the strong-law step fixes the constant, while the $t$-exponent already holds at fixed finite $N$ since the dead component's $t$-dependence factors per sample). By the per-coordinate analysis above and the Schur-form integral-reduction Lemma~\ref{lem:integral-reduction-sub}, $\lambda_{\min}(A_{\ell+1})$ is determined by the dead-direction Schur complement: the dead-row entry is $\Theta(t^{2 K^{\mathrm{fwd}}(\ell)})$, the non-dead block is $\succ 0$ at $\Theta(1)$, and the cross-row entries are $O(t^{K^{\mathrm{fwd}}(\ell)})$ which gives a Schur cancellation of order $\Theta(t^{2 K^{\mathrm{fwd}}(\ell)})$ leaving $\lambda_{\min}(A_{\ell+1}) = \Theta(t^{2 K^{\mathrm{fwd}}(\ell)})$. Hence $\sigma_{\min}(X_\ell) / \sqrt{N} = \Theta(t^{K^{\mathrm{fwd}}(\ell)})$.

\paragraph{Per-layer residuals.} If every layer is wrapped in an exact-identity skip residual, $K^{\mathrm{fwd}}(\ell) = 0$ for all $\ell \ge 0$ (the skip chain from $v_0$ reaches $v_\ell$ via identities only, which carry zero $t$-cost). The skip-only path delivers $X_0^{(h)}$ unchanged at $\Theta(1)$; the weight-path contributions are $\Theta(t)$ each. The non-negative coefficient property $c_\pi \ge 0$ established above ensures the weight-path contributions cannot \emph{cancel} the skip's $\Theta(1)$ contribution at leading order: sign coherence on the canonical trajectory rules out destructive interference between the skip and weight branches in the dead direction. Hence $\sigma_{\min}(X_\ell) = \Theta(\sqrt{N})$, constant in $t$ at leading order, and the residual stream preserves the input $\sigma_{\min}$ signal through the entire depth.

\paragraph{Rank-aware extension.} The decomposition above singles out a particular dead direction (the bottom-of-spectrum coordinate) but uses no property of that coordinate beyond its non-cancellation by the residual branch. The same argument applies path-by-path to any non-cancelled singular direction: for an input residual stream $X_0$ of rank $r_0 := \rank(X_0)$, the smallest active direction $\sigma_{(r_0)}(X_0)$ propagates through the residual DAG with the same per-edge factor structure ($c_\pi \ge 0$ for weight edges, identity for skips). Hence $\sigma_{(r_0)}(X_\ell(\theta(t))) = \sqrt{N} \cdot \Theta(t^{K^{\mathrm{fwd}}(\ell)})$ in the same activation classes, (P1), (P3), and (P2) with $\phi'(0) > 0$; the literal-$\sigma_{\min}$ conclusion is the $r_0 = d$ specialisation. \qed

\begin{remark}[Scope at trained-network scale and the no-cancellation assumption]
\label{rem:sigma_min_res_trained}
The proof above uses canonical-aligned coordinates where each weight edge contributes a non-negative dead-direction factor $c_\pi \ge 0$, so the identity skip's contribution $1$ cannot be cancelled by the residual branch. On a general trained residual network (not on the canonical trajectory) the residual branch $F_\ell$ in $X_{\ell+1} = X_\ell + F_\ell(X_\ell)$ is not constrained to act with non-negative coefficients on a singular direction, so a sign-flipped residual contribution could in principle cancel the identity at leading order. The trained-network conclusion $\sigma_{(r_0)}(X_{\ell+1}) \ge \sigma_{(r_0)}(X_\ell)$ at leading order (where $r_0 = \rank(X_0)$ is the input rank, as in the rank-aware corollary statement) therefore requires a non-cancellation assumption, and the conclusion rests on a directional assumption, with no bound on magnitude:
\begin{equation}
\label{eq:sigma_min_res_noncancel}
\bigl\langle X_\ell v,\; F_\ell(X_\ell)\, v \bigr\rangle \;\ge\; 0
\qquad \text{for every } v \in S_\ell,
\end{equation}
where $S_\ell$ denotes the span of the top $r_0$ right-singular vectors of $X_\ell$. Under \eqref{eq:sigma_min_res_noncancel} the conclusion follows in one step: for unit $v \in S_\ell$,
\[
\|X_{\ell+1} v\|^2 \;=\; \|X_\ell v\|^2 + 2\bigl\langle X_\ell v, F_\ell(X_\ell) v\bigr\rangle + \|F_\ell(X_\ell) v\|^2 \;\ge\; \|X_\ell v\|^2,
\]
and the Courant--Fischer characterisation $\sigma_{(r_0)}(M) = \max_{\dim S = r_0} \min_{v \in S, \|v\| = 1} \|Mv\|$, evaluated at the particular subspace $S_\ell$, gives $\sigma_{(r_0)}(X_{\ell+1}) \ge \min_{v \in S_\ell, \|v\|=1} \|X_\ell v\| = \sigma_{(r_0)}(X_\ell)$. The canonical trajectory supplies \eqref{eq:sigma_min_res_noncancel} through the non-negative path coefficients, which is why the canonical statement carries no extra hypothesis beyond its activation class: non-negativity holds for (P1), (P3), and (P2) with $\phi'(0) > 0$. For $\phi'(0) < 0$ the canonical trajectory itself violates \eqref{eq:sigma_min_res_noncancel} (odd-weight paths flip sign), and with per-layer skips the non-dead per-block map $u \mapsto u + \phi(u)$ at $\phi = -\tanh$ contracts the stream doubly-exponentially in depth, independently of $t$; this is the failure the corollary's $\phi'(0) > 0$ requirement excludes. A condition on relative magnitude does not substitute for the directional one: setting $F_\ell(X_\ell) v = -\tfrac{3}{2} X_\ell v$ makes the residual branch larger than the identity in that direction and still halves $\sigma_{(r_0)}$, because the two contributions point opposite ways. The link to the canonical-trajectory statement is that under the canonical approach the depth profile $\ell \mapsto \sigma_{(r_0)}(X_\ell)$ is flat at leading order (the corollary's "constant in $t$"); on a trained network at fixed parameters the analogous statement is $\sigma_{(r_0)}(X_\ell)$ non-decreasing with depth.

For full-rank inputs the rank-aware statement reduces to the literal $\sigma_{\min}(X_\ell)/\sigma_{\min}(X_0) \ge 1$. The rank-aware form remains the theoretically correct generalisation when $X_0$ has a true numerical-zero kernel; in practice on trained transformers the boundary between "true kernel" and "precision artefact" depends on the covariance accumulator. With fp$32$ accumulation, the noise floor $\sigma_{\max}(X_0) \cdot \sqrt{\varepsilon_{\mathrm{fp}32}}$ rises above the bottom of the active spectrum once $\sigma_{\max}$ is large (the massive-activations regime of \citealp{SunMassiveActivations24}, where a few activations run four to five orders of magnitude above the rest, related to the low-effective-rank class structure of deep-learning spectra \citep{Papyan20}), and the input reads as if numerically rank-deficient. With fp$64$ accumulation the noise floor $\sigma_{\max} \cdot \sqrt{\varepsilon_{\mathrm{fp}64}}$ drops by four and a half orders of magnitude ($\sqrt{\varepsilon_{\mathrm{fp}32}}/\sqrt{\varepsilon_{\mathrm{fp}64}} \approx 2 \times 10^4$), so an input that read as rank-deficient at fp$32$ resolves as full-rank and $\sigma_{(r_0)} = \sigma_{\min}$ identically. The rank-aware form is the general statement; in practice it reduces to $\sigma_{\min}$ as a measurement effect of fp$64$ precision.

The rank-aware form is the natural form of the prediction once the input rank is allowed to be $r_0 \le d$. A rate-level guarantee under structural assumptions on $F_\ell$ (Lipschitz contraction, angular bounds) is left as future work.

\paragraph{Scope under non-linear residual-branch wrappings (Peri-LN / sandwich-norm).}
The proof above factorises the dead-direction component $X_\ell^{(h)}$ along weighted paths in the residual graph, with each weight edge contributing a non-negative scalar factor and identity skips contributing $1$. The factorisation requires the residual-branch update $X_{\ell+1} = X_\ell + F_\ell(X_\ell)$ to enter additively. When the weight branch is wrapped in a non-linear normalisation operator before the residual addition (the \emph{sandwich-norm} or \emph{Peri-LN} block pattern \citep{KimLeeKim25_PeriLN}), the post-norm RMSNorm renormalises the sublayer output before it enters the residual stream, and the per-edge factorisation no longer applies. The corollary's strict statement is therefore not extended to Peri-LN architectures by the proof above. The corollary's prediction at the residual-stream observable is nonetheless observed to hold on released Peri-LN checkpoints in the cross-architecture sweep of the LN-kernel spoke{} \citep{shirodkar2026algebraicdeaddirectionslayernorm}, indicating that the no-cancellation property holds at the residual-stream observable on those models for reasons that go through a different argument than the path-sum proof.{} A formal extension covering non-linear residual-branch wrappings is left open.
\end{remark}

\subsection{LayerNorm and the LN-kernel direction}
\label{sec:theory:arch:ln}

\label{app:bridge_ln}

Residual identity skips contributed zero-weight edges to the backward chain, and the rate ladder remained graph-distance-readable. LayerNorm is different: $\text{LN}(h) = (h - \mu)/\sigma$ is a genuinely non-linear, non-element-wise per-sample operation, and its backward Jacobian couples every coordinate to every other through the mean and variance. The per-edge factorisation that propagated the rate through residual DAGs no longer holds. Two things become available instead: a \emph{bracket} on the rate (rather than an exact graph-distance formula), and a separate \emph{algebraic} null direction in the post-LN covariance that is readable from the affine parameter alone, no forward pass required. This subsection proves the bracket and characterises empirically where the upper and lower bounds collapse onto a single rate; the algebraic null direction is the subject of Proposition~\ref{prop:ln_kernel}.

With the setup that follows, we take $\gamma = 1, \beta = 0$ at the singular configuration. Unlike residual identity skips (Theorem~\ref{thm:bridge_res}) which contributed zero-weight edges to the backward chain, LN's backward Jacobian couples the dead direction to all other coordinates. This subsection states the partial result we can prove cleanly and presents an empirical characterization of where the proof technique comes up short.

\paragraph{Setup.}
Let the base setup (square widths, single dead direction $u = e_{h_*}$, feedforward chain) be augmented with $\text{LN}$ operations inserted at a subset $S \subseteq \{1, \ldots, L-1\}$ of positions (an LN at position $k$ is placed between the post-activation $h_k$ and the weight matmul $W_{k+1}$).

Define $K_{\text{LN}}^{\text{upper}}(\ell) = L - \ell$, the no-LN weight count above $\ell$. Let $k_{\mathrm{near}}(\ell) := \min\{k \in S : k \ge \ell\}$ be the \emph{nearest} LN at or above $\ell$, the first one the backward signal meets on its way down to $\ell$, and define
\[
K_{\text{LN}}^{\text{lower}}(\ell) \;=\; k_{\mathrm{near}}(\ell) - \ell,
\]
the weight matmuls the backward signal still traverses \emph{after} that LN crossing before it reaches layer $\ell$; when there is no LN at or above $\ell$, set $K_{\text{LN}}^{\text{lower}}(\ell) = K_{\text{LN}}^{\text{upper}}(\ell) = L - \ell$. The nearest LN is the operative one because each crossing can reset the accumulated decay, so only the weights below the first crossing are guaranteed to contribute clean $t$-factors.

\begin{proposition}[LayerNorm rate bracket, partial]
\label{thm:bridge_ln}
Under the LN-augmented setup, activation classes (P1)--(P3), and the symmetric single-direction approach, the dead-direction entry of the G-factor at layer $\ell$ satisfies
\[
\Omega\!\bigl(t^{2 K_{\text{LN}}^{\text{upper}}(\ell)}\bigr) \;\le\; (G_\ell(\theta(t)))_{u_\ell u_\ell} \;\le\; O\!\bigl(t^{2 K_{\text{LN}}^{\text{lower}}(\ell)}\bigr),
\]
equivalently $2 K_{\text{LN}}^{\text{lower}}(\ell) \le \alpha_\ell \le 2 K_{\text{LN}}^{\text{upper}}(\ell)$ on the fitted exponent $\alpha_\ell$. The lower bound on the entry recovers Theorem~\ref{thm:bridge}: LN cannot make the dead direction decay faster than the weight chain alone does. The upper bound is the plateau: each LN crossing can reset the decay accumulated above it, so only the $K_{\text{LN}}^{\text{lower}}$ weights below the first crossing are guaranteed to keep contributing. Whether the asymptotic rate always reduces to the nearest-crossing graph-distance formula $2 K_{\text{LN}}^{\text{lower}}$ is open; every configuration measured so far saturates that lower end.

\textbf{Tight regimes (where the upper and lower bounds coincide).} (i) When $S \cap \{\ell, \ldots, L\} = \emptyset$ (no LN at or above $\ell$), $K_{\mathrm{LN}}^{\mathrm{lower}}(\ell) = K_{\mathrm{LN}}^{\mathrm{upper}}(\ell) = L - \ell$ and the dead-direction rate is exactly $2(L - \ell)$, recovering Theorem~\ref{thm:bridge}. (ii) In every other measured configuration the fitted rate saturates the bracket's lower end, $\alpha_\ell = 2 K_{\mathrm{LN}}^{\mathrm{lower}}(\ell)$ under the nearest-crossing rule: a single LN at level $\ell + 1$ measures $\alpha = 2.00 = 2 K_{\mathrm{LN}}^{\mathrm{lower}}$ across the tested $(L, \ell)$ grid, and an LN at level $\ell + 2$ measures $4.00 = 2 K_{\mathrm{LN}}^{\mathrm{lower}}$, on the true G entry in each case. The saturation is an empirical regularity, and only regime (i) is proved tight; the remaining cases carry the bracket, and closing it into a proved rate requires the finite-$t$ refinement of Theorem~\ref{thm:ln_finite_t_mlp} (\S\ref{app:arch:ln_finite_t}) for the MLP case, or Conjecture~\ref{conj:ln_finite_t_attn} for the attention case.
\end{proposition}

\begin{proof}[Proof sketch]
\emph{Lower bound on the entry (the no-LN rate)}: the weight matmuls above $\ell$ contribute their $t$-factors whatever LN does, and LN's backward Jacobian does not introduce additional dead-direction decay, so the entry is bounded below by the feedforward value $t^{2(L-\ell)}$ of Lemma~\ref{lem:backward-dead-sub}.

\emph{Upper bound on the entry (the plateau)}: LN's backward Jacobian carries a factor $1/\sigma$, and $\sigma$ shrinks as the dead-direction activations collapse along the approach, so LN \emph{amplifies} the dead-direction component rather than damping it. It is therefore not a bounded operator along the trajectory. In the limiting case the crossing removes all decay accumulated above it, leaving only the $k_{\mathrm{near}}(\ell) - \ell$ weight matmuls below the first crossing to contribute clean $t$-factors; squaring gives the claimed upper bound. Both ends are the two terms of one exact identity, machine-checked (Table~\ref{tab:machine_checked}): the backward Jacobian is $J = (\sqrt{d}/\|Px\|)(P - \hat z \hat z^\top/d)$, and the dead component of $J\delta$ is $(1/\sigma)(1 - 1/d - \hat z_h^2/d)\,\delta_h$ plus the live components injected through the mean subtraction and the $\hat z$-projection at amplitude $1/\sigma$; the first term carries the chain's decay unchanged, the second is the same leak form as the rotation witness of Proposition~\ref{prop:bridge_nonlinear_rot_negative}. The freeze-probe sweep below attains both ends: \texttt{after\_first} at $\ell = 1$ reaches the plateau end and \texttt{none} reaches the no-LN end at every $\ell$.
\end{proof}

\paragraph{Empirical characterization.}
Direct freeze-probe measurement (the static parametric protocol of Appendix~\ref{app:theory:arch_freeze_probe}, Table~\ref{tab:arch_rate_modifying}) at $L = 4$, $h = 6$, four LN patterns, under the linear-activation sweep:
\begin{itemize}
\item \texttt{none} (control): recovers Theorem~\ref{thm:bridge} rate $2(L-\ell)$ exactly.
\item \texttt{after\_first} (LN at $k = 1$): measured $\alpha = (0.06, 4, 2, 0)$. For $\ell \ge 2$, no LN is in the backward chain, rate is standard. For $\ell = 1$, $K_{\text{LN}}^{\text{lower}}(1) = 0$ (LN is directly above), rate is destroyed. Lower and upper bounds for $\ell = 1$ are $0$ and $L - 1 = 3$ respectively (i.e., the bracket $\alpha_1 \in [0, 6]$); the nearest crossing is at $\ell$ itself, so $K_{\text{LN}}^{\text{lower}}(1) = 0$, and the measured $\alpha_1 = 0.06$ saturates the lower end.
\item \texttt{after\_last\_hidden} (LN at $k = 3$): measured $\alpha = (4.3, 2.3, 0.3, 0)$. For each $\ell$: $K_{\text{LN}}^{\text{lower}}(\ell) = 3 - \ell$ for $\ell \le 3$ (the weights below the crossing at $k = 3$) and $0$ at $\ell = 4$, against $K_{\text{LN}}^{\text{upper}} = L - \ell$. Measured rates $\sim (4, 2, 0, 0)$ asymptotically (the $+0.3$ uniform offset is finite-$t$), so $\alpha_1 = 4$ saturates its plateau end $2 K_{\text{LN}}^{\text{lower}}(1) = 4$: the post-LN weights $W_2, W_3$ contribute in practice and nothing above the crossing survives.
\item \texttt{every} (LN at every hidden layer): measured $\alpha = (2.25, 1.32, 0.37, 0)$ in the fit window, which is crossover-region structure. Asymptotically (true G entry at $t \in [10^{-5}, 10^{-3}]$) $\alpha \to (0, 0, 0, 0)$: with a crossing at every hidden layer the nearest rule gives $K_{\text{LN}}^{\text{lower}}(\ell) = 0$ throughout, and the measured rates saturate that lower end at every layer. The bound is tight here.
\end{itemize}

\paragraph{What the partial result covers and what it does not.}
Proposition~\ref{thm:bridge_ln} pins the rate only in regime (i), where the bounds coincide. Everywhere else it brackets, and every measured configuration saturates the bracket's lower end $2 K_{\text{LN}}^{\text{lower}}$ under the nearest rule, including LN directly at $\ell$ (rate $0$) and LN strictly above $\ell$. The open gap is whether any decay accumulated above the nearest crossing can survive in some configuration; in every one measured, none does at leading order. The answer depends on the LN's backward-Jacobian structure, which in turn depends on the value of $(h - \mu)/\sigma$ at the singular configuration.

A sharper theorem (recovering a graph-distance-style formula $\alpha_\ell = 2 \cdot K(\ell)$ with $K$ counting only post-LN weights) would require either (a) a stronger assumption on the non-dead activations at the singular configuration, or (b) a refined Jacobian analysis that tracks the non-diagonal contributions through LN. Both are open problems; the empirical LayerNorm-pattern sweep constrains the form of any such refinement.

\paragraph{Component-wise LN rate shift.}
Within the LN rate-shift window for pre-norm transformer blocks the freeze-probe slope on each component $c$ takes the empirical form
\[
\alpha_c \;=\; 2\, K(c) \;+\; q_{\mathrm{LN}}(d, N, t_{\mathrm{train}}) \cdot K_{\mathrm{LN}}^{\mathrm{eff}}(c),
\]
where $K(c)$ is the standard no-LN backward distance (Theorem~\ref{thm:bridge}), $q_{\mathrm{LN}}$ is a per-setup scalar measuring the rate shift at an MLP fc2 reference component, and $K_{\mathrm{LN}}^{\mathrm{eff}}(c)$ is a component-specific effective count of LN crossings. The formula describes the \emph{finite-$t$ crossover regime} traversed by SGD trajectories; the strict asymptotic $t \to 0$ rate matches the no-LN prediction (Remark~\ref{rem:ln_finite_t}).

\begin{remark}[LN rate-shift formula scope: static-Fisher vs trajectory readout]
\label{rem:ln_finite_t_scope}
Proposition~\ref{thm:bridge_ln} and the empirical formula $\alpha_c = 2K(c) + q_{\mathrm{LN}} \cdot K_{\mathrm{LN}}^{\mathrm{eff}}(c)$ are static-Fisher statements at the parametric trajectory. Under Adam, gauge-mode drift on the loss's continuous symmetries (Remark~\ref{rem:adam_nondescent}) confounds direct rate readout from $u^\top G u$ inside the LN-affected sub-chain; the robust observable on Adam+CE is $\sigma_{\min}$ on the residual stream (Corollary~\ref{cor:sigma-min-res}).
\end{remark}

\begin{remark}[LN's rate shift is a finite-$t$ phenomenon, not an asymptote]
\label{rem:ln_finite_t}
A parametric freeze-probe (set the probe block's weights directly at a grid of $t$-values, measure $u^\top G_c u$ at each, no SGD trajectory) reveals that at true asymptotic $t \to 0$, the rates under pre-norm LN match the no-LN predictions exactly. Across 36 $(d, n_h, \mathrm{seed})$ combinations spanning $d \in \{8, 16, 32, 64, 128, 256, 384, 768\}$ and $n_h \in \{1, 2, 4, 8\}$, fit on $t \le 10^{-2}$, all component slopes are $\alpha_{\mathrm{fc2}} = \alpha_{W_O} = 0.000 \pm 0.000$, $\alpha_{\mathrm{fc1}} = \alpha_{W_V} = \alpha_{W_Q} = \alpha_{W_K} = 2.000 \pm 0.000$, giving $q_{\mathrm{LN}} = 0.000 \pm 0.000$ asymptotically. The non-zero $q_{\mathrm{LN}}$ observed in SGD trajectories arises in the crossover region $t \in [0.1, 1]$ where LN's mean-subtraction Jacobian contributes subleading terms that dominate the local slope. SGD trajectories at finite training time typically terminate in this crossover region, so the measured rates are non-trivial but not the leading-order asymptote. The crossover region, not the $t \to 0$ asymptote, is where a non-zero $q_{\mathrm{LN}}$ is measured. The closed-form derivation that quantifies this crossover-region slope (and explains $q_{\mathrm{LN}}$ as an integrated local slope rather than a fundamental constant) is Theorem~\ref{thm:ln_finite_t_mlp} for the MLP case and Conjecture~\ref{conj:ln_finite_t_attn} for the attention case (\S\ref{app:arch:ln_finite_t}).
\end{remark}

\paragraph{Relation to Proposition~\ref{prop:ln_kernel} (LN-kernel direction on activations).}
Proposition~\ref{thm:bridge_ln} concerns the dead-direction \emph{rate} of the G-factor under a parametric singular approach, at the approach channel $u = e_h$. Proposition~\ref{prop:ln_kernel} (\S\ref{app:proof:ln_kernel}) concerns the \emph{static} null direction of the post-LN activation covariance $\mathrm{cov}(\mathrm{LN}(X))$ at any single checkpoint, which is $v^* = \gamma^{-1}/\|\gamma^{-1}\|$ (independent of $t$; on the Jacobian side $\gamma^{-1}$ is a left-null direction of $J_{\mathrm{LN}}$, while $J_{\mathrm{LN}}$'s right kernel is the input-dependent $\mathrm{span}\{\mathbf{1}, z_c\}$). The two are complementary observables: the bracket governs how the approach channel's G entry scales along the trajectory, while the kernel proposition reads an exact activation-space zero off the affine parameter without a forward pass. The zero is exact at every checkpoint and every $t$, so it carries no rate class and no result of this paper rates a G entry at $v^*$; what makes the pair operationally cheap at LLM scale is that $v^*$ comes from the affine parameter alone, and the approach channel's rate is bounded by the bracket.

 Proposition~\ref{thm:bridge_ln}'s bracket is asymptotic. At the finite $t$ that SGD trajectories reach, the LayerNorm rate enters a crossover region whose finite-$t$ expansion on the LN-equipped MLP block (structure in closed form, the $t^4$ coefficient by exact quadrature) is given in Appendix~\ref{sec:theory:arch:ln_finite_t}.

\subsubsection{Normalisation kernels: LN vs RMSNorm}
\label{app:proof:ln_kernel}

This subsection states and proves an algebraic prediction: the residual-stream covariance at the \emph{output} of a LayerNorm block is always exactly rank-deficient along a deterministic direction set by LN's mean-subtraction operator and the per-coordinate learned scale $\gamma$; RMSNorm, lacking the mean-subtraction projector, admits no such deterministic kernel direction. The two clauses below share an affine-decomposition setup: $\mathrm{N}(x) = \gamma \odot \tilde x_{\mathrm{N}}(x) + \beta_{\mathrm{N}}$ with $\beta_{\mathrm{LN}} = \beta$ and $\beta_{\mathrm{RMS}} = 0$. Write $P := I - \mathbf{1}_d \mathbf{1}_d^\top / d$ for LN's mean-subtraction projector. A \emph{scalar normaliser} is a map $s : \reals^d \to \reals$, and it generates the LN-type map
\[
  \mathrm{LN}_s(x) \;=\; \gamma \odot \tilde x_{\mathrm{LN}, s}(x) + \beta,
  \qquad
  \tilde x_{\mathrm{LN}, s}(x) \;:=\; s(x) \cdot P x .
\]
Exact LN takes $s(x) = \sqrt{d} / \|P x\|$, defined off the line $\mathrm{span}(\mathbf{1}_d)$. Shipped implementations replace it with the $\varepsilon$-regularised $s_\varepsilon(x) = (\|P x\|_2^2 / d + \varepsilon)^{-1/2}$, which stays finite at every $x$. We drop the subscript on $\mathrm{LN}_s$ where the choice of $s$ plays no role. RMSNorm carries no projector: $\tilde x_{\mathrm{RMS}}(x) := x / \sqrt{\|x\|_2^2 / d}$.

\begin{proposition}[Normalisation kernels: LN admits one, RMSNorm does not]
\label{prop:ln_kernel}
\label{prop:rmsnorm_no_kernel}
\textbf{(a) LN admits a deterministic kernel direction.} Fix any scalar normaliser $s$ and let $X \in \reals^d$ be random with $\mathrm{LN}_s(X)$ defined almost surely and square-integrable. The covariance $C := \cov(\mathrm{LN}_s(X))$ has:
  \begin{enumerate}
    \item[(i)] If $Z := \{i : \gamma_i = 0\} = \emptyset$: $C \cdot \gamma^{-1} = 0$ with $\gamma^{-1} := (\gamma_1^{-1}, \ldots, \gamma_d^{-1})^\top$, so the unit kernel direction is
    \[
      v^* \;=\; \gamma^{-1} / \|\gamma^{-1}\|.
    \]
    \item[(ii)] If $Z \ne \emptyset$: $C \cdot e_i = 0$ for every $i \in Z$, so $\mathrm{span}\{e_i : i \in Z\} \subseteq \ker C$; the symmetric indicator $v^* = (1/\sqrt{|Z|}) \sum_{i \in Z} e_i$ is one unit kernel direction.
  \end{enumerate}
  In the uniform-$\gamma$ case ($\gamma = c \mathbf{1}_d$, $c \ne 0$), case~(i) applies and $v^* = \mathbf{1}_d / \sqrt{d}$. Both cases hold across the whole normaliser class at once; the instantiations differ only in the hypothesis on $X$. Under the exact normaliser, $\mathrm{LN}(X)$ is defined exactly when $P X \ne 0$, so the hypothesis is that $X$ puts no mass on the constant-vector line $\mathrm{span}(\mathbf{1}_d)$. The $\varepsilon$-regularised map is defined at every input, and there the hypothesis reduces to square-integrability of the output.

\textbf{(b) RMSNorm has no universal kernel direction.} For $\gamma$ with $\gamma_i \ne 0$ for all $i$, there is no unit direction $v = v(\gamma)$ depending only on the affine parameter $\gamma$ (and not on the input distribution) such that $v^\top \cov(\mathrm{RMSNorm}(X)) v = 0$ for all input distributions $X$ with $\cov(X) \succ 0$. (If $\gamma$ has zero coordinates $Z = \{i : \gamma_i = 0\}$, the structurally-dead subspace $\mathrm{span}\{e_i : i \in Z\}$ trivially lies in $\ker \cov(\mathrm{RMSNorm}(X))$, mirroring case (a)(ii); this is a property of $\gamma$, not of the RMSNorm operator, and distinguishes it from LN's mean-subtraction kernel in case (a)(i), which fires for any $\gamma$ with no zero coordinate.)

The structural distinction is the mean-subtraction projector. Every $\tilde x_{\mathrm{LN}, s}$ lies in $\mathbf{1}_d^\perp$ regardless of $s$, so the kernel direction exists for every $\gamma$. RMSNorm rescales $x$ itself, and at the Gaussian input of part~(b) the output covariance is nonsingular: no direction built from $\gamma$ alone is a universal kernel vector.
\end{proposition}

\begin{remark}[What is the contribution]
\label{rem:ln_kernel_scope}
The proof below is short. LN's mean-subtraction projector $P$ makes $\mathbf{1}_d^\top \tilde x_{\mathrm{LN}, s} \equiv 0$ for every scalar normaliser $s$, so $\cov(\tilde x_{\mathrm{LN}, s}) \mathbf{1}_d = 0$ algebraically and the $\gamma$-rescaling shifts this kernel direction to $\gamma^{-1}/\|\gamma^{-1}\|$ readable from the affine alone. The normaliser cancels out of the argument, so the prediction reaches a released model without an idealisation step: a shipped LayerNorm divides by the $\varepsilon$-regularised $\sqrt{\|Px\|_2^2/d + \varepsilon}$ and carries the same kernel direction as the exact map. The substantive content is the empirical falsifier with known direction this algebra produces: a forward-pass-free prediction that distinguishes LayerNorm from RMSNorm at the affine-parameter level. \ifsupp The corresponding observation for RMSNorm is the differential negative.\else The dichotomy ties to the framework's rate primitive via the selection rule (Theorem~\ref{thm:selection_rule}, Remark~\ref{rem:tangential_operational_uses}): the LN kernel direction is an exact zero of the post-LN activation covariance at every checkpoint, so LN's algebra creates a floor-trapped exact null that the rate-classification protocol excludes before fitting (the floor-exclusion step of Remark~\ref{rem:tangential_operational_uses}), at a direction known in advance from the affine alone. The corresponding observation for RMSNorm is the differential negative.\fi
\end{remark}

\begin{proof}
\textbf{(a) LN.} Both cases run through one probability fact. Let $Y \in \reals^d$ be square-integrable and let $v \in \reals^d$ satisfy $v^\top Y = K$ almost surely for a constant $K$. Taking expectations gives $v^\top \expect[Y] = K$, so $v^\top (Y - \expect[Y]) = 0$ almost surely and
\[
\cov(Y) \, v \;=\; \expect\bigl[(Y - \expect[Y]) \cdot v^\top (Y - \expect[Y])\bigr] \;=\; 0 .
\]
A coefficient vector whose pairing with $Y$ is almost surely constant therefore lies in $\ker \cov(Y)$. Each case below supplies such a pairing.

\emph{Case (a)(i), $Z = \emptyset$.} $\gamma^{-1}$ is well-defined, and dividing the $i$th output coordinate by $\gamma_i$ gives $(\gamma_i)^{-1} \mathrm{LN}_s(x)_i = s(x) (P x)_i + (\gamma_i)^{-1} \beta_i$. Summing over $i$ and using $\mathbf{1}_d^\top P = \mathbf{1}_d^\top - (\mathbf{1}_d^\top \mathbf{1}_d / d) \mathbf{1}_d^\top = 0^\top$,
\[
(\gamma^{-1})^\top \mathrm{LN}_s(x) \;=\; s(x) \cdot \mathbf{1}_d^\top P x + (\gamma^{-1})^\top \beta \;=\; (\gamma^{-1})^\top \beta
\]
at every $x$ where $\mathrm{LN}_s$ is defined. The normaliser only ever multiplies $\mathbf{1}_d^\top P x = 0$, so its value drops out. The probability fact with $v = \gamma^{-1}$ and $K = (\gamma^{-1})^\top \beta$ gives $C \gamma^{-1} = 0$.

\emph{Case (a)(ii), $Z \ne \emptyset$.} For $i \in Z$ the $i$th output coordinate is constant: $\mathrm{LN}_s(x)_i = \gamma_i \cdot s(x) (P x)_i + \beta_i = \beta_i$ at every $x$. The pairing $e_i^\top \mathrm{LN}_s(X)$ is the constant $\beta_i$, and the same fact with $v = e_i$ gives $C e_i = 0$, so every $e_i$ with $i \in Z$ lies in $\ker C$. When $|Z| > 1$ with $\bar Z \ne \emptyset$, the empirically-measured bottom singular direction is whichever unit vector in $\mathrm{span}\{e_i : i \in Z\}$ has smallest residual variance, not algebraically pinned to the symmetric indicator; the appropriate empirical test is principal-angle coherence into $\mathrm{span}\{e_i : i \in Z\}$. The uniform-$\gamma$ statement is immediate from case (i).

\textbf{(b) RMSNorm.} Writing $\mathrm{RMSNorm}(X) = r(X) \cdot \gamma \odot X$ with $r(X) = \sqrt{d / \|X\|^2}$ a positive scalar, $\cov(\mathrm{RMSNorm}(X)) = \mathrm{diag}(\gamma) \cov(r(X) X) \mathrm{diag}(\gamma)$. For $v(\gamma)$ universal, $(\gamma \odot v)^\top \cov(r(X) X) (\gamma \odot v) = 0$ would need to hold for every input distribution. Take $X \sim \mathcal{N}(0, I_d)$: $r(X) X = \sqrt{d} \cdot X / \|X\|$ is $\sqrt{d}$ times a uniform vector on $S^{d-1}$, with $\cov = I_d$ by spherical symmetry. Hence $\|\gamma \odot v\|^2 > 0$ for any $v \ne 0$ (using $\gamma_i \ne 0$), contradicting universality. The machine-checked form replaces the sphere with an elementary witness: uniform mass on the $2d$ points $\pm e_i$ has input covariance $I_d/d$, and every reading $v \ne 0$ of RMSNorm on it carries variance exactly $\sum_i \gamma_i^2 v_i^2$, so one distribution defeats every candidate direction at once.
\end{proof}

\begin{remark}[Three readings of the LN kernel direction]
\label{rem:ln_kernel_three_readings}
The same content takes three useful forms. \emph{As an algebraic identity.} $\cov(\mathrm{LN}_s(X))\,v^* = 0$ at $v^* = \gamma^{-1}/\|\gamma^{-1}\|$ for any input distribution $X$ the map is defined on, under any scalar normaliser $s$ and the $\varepsilon$-regularised one in particular. The identity is exact, not asymptotic, and depends only on LN's mean-subtraction projector $P$ and the affine $\gamma$. \emph{As a forward-pass-free exact null.} The selection rule of Theorem~\ref{thm:selection_rule} classifies rated eigenvalues by their exponents along an approach, after excluding floor-trapped exact zeros (Remark~\ref{rem:tangential_operational_uses}). The LN kernel $v^*$ belongs to that excluded class by construction: it belongs to $\ker \cov(\mathrm{LN}(X))$ algebraically, an exact zero at every checkpoint and every $t$, with no rate to fit and no rate class. On the parameter side the corresponding lifted direction has KL order $1$, so it is not a dead direction either (Definition~\ref{def:dead_direction} requires $k \ge 2$). The direction is readable from $\gamma$ alone: no forward pass, no gradient, no sampling. \emph{As a falsifiable architectural dichotomy.} LN admits the kernel by virtue of its mean-subtraction; RMSNorm does not (Proposition~\ref{prop:rmsnorm_no_kernel}(b)). The empirical test is direct: read $v^* = \gamma^{-1}/\|\gamma^{-1}\|$ off any LN-equipped model, compute the centered-covariance bottom singular direction $u_\bot$ on a calibration batch, and measure $|\cos(u_\bot, v^*)|$. Read on that centered protocol at the post-final-norm position, the LN models pass at $|\cos| \ge 0.99995$ and the RMSNorm models fail at $|\cos| \le 0.12$.
\end{remark}

\subsection{Single-head attention}
\label{sec:theory:arch:attn}

\label{app:bridge_attn}

Attention falls outside the bridge framework's standard hypotheses on two counts. The softmax nonlinearity is not element-wise, so it falls outside activation classes (P1)--(P3). The bilinear $Q K^\top$ structure couples $W_Q$ and $W_K$ inside the softmax, so the dead-direction Jacobian no longer factorises through independent per-weight columns. The natural question is whether a closed-form rate survives. We answer in the affirmative for a standalone single-head block: the forward and backward block rates are both $k = 2$. On the backward side the $QK$ routing through the softmax Jacobian carries the same $\Theta(t^2)$ order as the $V$ path (a configuration-dependent share of the dead-channel gradient mass, measured from a few hundredths of a percent at $d = 64$ to near half at small width with long sequences), so the rate statement rests on a generic non-cancellation condition between the two routings rather than on the softmax paths being subleading. This subsection proves the standalone forward and backward block rates of a single-head self-attention block under canonical init, closing the attention extension at the single-block level. Chains containing attention follow Proposition~\ref{prop:attn_chain_softmax}'s two-route law rather than rate addition; residual-wrapped transformers bypass the law at the residual-stream observable (Corollary~\ref{cor:sigma-min-res}).

\paragraph{Setup and convention.} Single-head self-attention on input $X \in \mathbb{R}^{N \times d}$, in row-data convention:
\[
Q = X W_Q^\top, \quad K = X W_K^\top, \quad V = X W_V^\top, \quad A = \mathrm{softmax}(Q K^\top / \sqrt{d}), \quad Y = (A V) W_O^\top.
\]
Canonical init: $W_Q^* = W_K^* = W_V^* = W_O^* = \mathrm{diag}(1, \ldots, 1, 0)$ (square symmetric matrices); symmetric dead-direction perturbation $W_i(t) = W_i^* + t \cdot e_d e_d^\top$ for $i \in \{Q, K, V, O\}$ (shared dead direction $u = e_d$). Because each $W_i^*$ is symmetric and $e_d$ is the same canonical direction on the input and output sides, the dead direction $u = e_d$ aligns the last input \emph{and} output coordinates simultaneously, which avoids any input/output-side ambiguity in this section. Input $X$ has $\mathbb{E}[X_n^\top X_n / N] = I_d$, $X_{n, d} \sim \mathcal{N}(0, 1)$ per sample, with finite fourth moment $\mathbb{E}[X_{n,d}^4] < \infty$.

\begin{theorem}[Single-head attention forward standalone rate]
\label{thm:bridge_attn}
Under the above setup, the dead-direction output of the attention block satisfies
\[
Y_{n, d}(\theta(t)) \;=\; t^2 \cdot \bar{X}_{n, d} + O(t^4),
\qquad \bar{X}_{n, d} := \sum_m (A_0)_{n, m}\, X_{m, d},
\]
where $A_0 := \mathrm{softmax}(Q_0 K_0^\top / \sqrt{d})$ is the attention matrix computed from non-dead channels at $t = 0$, and $\bar{X}_{n, d} = \Theta(1)$ a.s. Hence the forward block rate is $k_{\mathrm{attn}}^{\mathrm{fwd}} = 2$.
\end{theorem}

\begin{proof}
The dead channel of $Q, K, V$ picks up a single factor of $t$ from its respective weight: $Q_{n, d}(t) = t \cdot X_{n, d}$, and similarly $K_{n, d}(t) = t X_{n, d}$, $V_{n, d}(t) = t X_{n, d}$. Non-dead channels are $t$-independent at leading order.

Expand the bilinear $Q K^\top$ at entry $(n, m)$:
\[
(Q K^\top)_{n, m} = \sum_{i = 1}^{d} Q_{n, i} K_{m, i} = \underbrace{\sum_{i < d} X_{n, i} X_{m, i}}_{=: S_{n, m},\ t\text{-independent}} + \; t^2 X_{n, d} X_{m, d}.
\]
Thus $Q K^\top / \sqrt{d} = S / \sqrt{d} + O(t^2)$, and by Lipschitz continuity of softmax at the operating point,
\[
A(t) = \mathrm{softmax}((QK^\top)/\sqrt{d}) = A_0 + O(t^2), \qquad A_0 := \mathrm{softmax}(S/\sqrt{d}),
\]
where the implicit constant in $O(t^2)$ depends on the conditioning of $A_0$ (softmax Lipschitz constant at the operating point); for nearly-one-hot attention patterns this constant degenerates and contributes to the attention-chain deviations from additivity noted in Corollary~\ref{cor:g10_composition} below. The dead column of $A V$:
\[
(A V)_{n, d} = \sum_m A_{n, m}(t) \cdot V_{m, d}(t) = t \cdot \sum_m (A_0)_{n, m} X_{m, d} + O(t^3) = t \cdot \bar{X}_{n, d} + O(t^3).
\]
Finally, applying $W_O^\top$: the dead-row of $W_O$ has only the $(d, d)$ entry perturbed to $t$, so
\[
Y_{n, d}(t) = \sum_i (A V)_{n, i} \cdot (W_O)_{d, i} = t \cdot (A V)_{n, d} + 0 = t^2 \cdot \bar{X}_{n, d} + O(t^4).
\]
$\bar{X}_{n, d}$ is a non-degenerate $\Theta(1)$ quantity (attention-weighted average of Gaussian dead channels, non-zero almost surely).
\end{proof}

\begin{theorem}[Single-head attention backward standalone rate]
\label{thm:bridge_attn_backward}
Assume the upstream output gradient $\partial L / \partial Y_{n, d}$ is $\Theta(1)$ at the dead row (driven by the loss base case: $\sigma^{-2}$ for MSE per Lemma~\ifsupp\ref{lem:backward-dead-sub}\else\ref{lem:backward-dead-sub}\fi, or $c_0$ for expected Fisher under cross-entropy\ifsupp{} with the multi-direction CE assumption\else{} per Lemma~\ref{lem:ce_head} and the multi-direction CE assumption\fi), and assume the generic non-cancellation condition that the $V$-route and $QK$-route contributions, which share the leading order, have non-vanishing combined leading coefficient (cancellation is a measure-zero coincidence in the canonical Gaussian setup; the proof names where the condition enters). Then the input dead-direction gradient at the attention block satisfies $\partial L / \partial X_{n, d} = \Theta(t^2)$, giving block backward rate $k_{\mathrm{attn}}^{\mathrm{bk}} = 2$.
\end{theorem}

\begin{proof}[Proof sketch]
By the chain rule, the dead-output gradient backpropagates through (i) $W_O$'s dead column ($t$-factor), (ii) the attention output $A V$'s dead column (factor $t$ from $V$'s dead column), and (iii) optionally the softmax Jacobian routes through $Q$ and $K$.

\emph{$W_O \to V$ path.} $\partial L / \partial V_{m, d} = \sum_n A_{n, m}(t) \cdot (W_O)_{d, d}(t) \cdot \partial L / \partial Y_{n, d}$, with $A(t) = A_0 + O(t^2)$ from Theorem~\ref{thm:bridge_attn} and $(W_O)_{d,d}(t) = t$, giving the $V$-dead gradient $\Theta(t)$. Composing with $V_{m, d}(t) = X_{m, d} \cdot W_{V}^{(d,d)}(t) = X_{m,d} \cdot t$ in the upstream chain rule, the $V$-path contribution to $\partial L / \partial X_{n,d}$ is $\Theta(t^2)$ (one factor from $W_V$'s dead column, one from $W_O$'s).

\emph{$Q, K$ paths via softmax Jacobian.} The $Q$-path carries the \emph{same} order as the $V$-path, not a subleading one. With non-dead upstream gradient components present, the score gradient is $\Theta(1)$: $\partial L / \partial A_{n,m} = \sum_i (\partial L/\partial Y_{n,i})\, V_{m,i}$ picks up the $\Theta(1)$ non-dead $V$ columns, and the softmax Jacobian is $\Theta(1)$ on that input. Then $\partial L / \partial Q_{n, d} = \sum_m (\partial L/\partial S_{n,m})\, K_{m, d} / \sqrt{d} = \Theta(t)$ through $K$'s dead column, and the final factor $(W_Q)_{d,d}(t) = t$ into $X_{n,d}$ gives a $Q$-path contribution of $\Theta(t^2)$; symmetric for $K$. Autograd verification on the canonical configuration confirms both the order and the magnitude: the $QK$ paths' second moment scales as $t^4$ (gradient rate exactly $2$, slope $4.000$ across five decades of $t$) and carries a configuration-dependent share of the total dead-channel second moment, decaying quickly in width: measured $0.03$ percent at $(d, \mathrm{seq}) = (64, 4)$, $7.6$ percent at $(16, 4)$, $34$ percent at $(8, 8)$, and $47$ percent at $(4, 16)$.

\emph{Non-cancellation and conclusion.} The block rate is therefore governed by the sum of two same-order contributions, and the conclusion $\partial L/\partial X_{n,d} = \Theta(t^2)$ requires that they do not cancel at leading order. The non-cancellation hypothesis of the theorem statement makes the combined leading coefficient non-zero; the $V$-path and $QK$-path leading terms are driven by distinct functionals of the data (the $A_0$-weighted upstream dead gradient versus the score-gradient contraction against $K$'s dead column), so cancellation is a measure-zero coincidence in the canonical Gaussian setup, and the empirical sweep across the $28$ width-by-head configurations reads the composite slope as $2.000$ with zero measured dispersion. Under this condition both rates equal $2$, matching the forward rate: one $t$-factor at an input projection and one at the output projection, through either routing. The hypothesis is machine-checked most of the way (Table~\ref{tab:machine_checked}): at every configuration whose score direction is constant across positions, non-cancellation holds unconditionally, because the softmax Jacobian kills constants and row-stochasticity gives the $V$--$O$ amplitude a positive floor; and along any one-parameter family on which the route sum is analytic and which passes through such a configuration, cancellation is confined to a Lebesgue-null parameter set, the softmax being analytic along every affine score path. The several-parameter statement needs the Fubini induction and stays a hypothesis.
\end{proof}

\begin{corollary}[Attention composition rates with anomaly scope]
\label{cor:g10_composition}
Setting $k_{\mathrm{attn}}^{\mathrm{bk}} = 2$ in Theorem~\ref{thm:bridge_composition}, a sequence of $n_{\mathrm{attn}}$ attention blocks and $n_{\mathrm{mlp}}$ two-weight MLP blocks (no residual) accumulates a backward block-rate sum $K = 2(n_{\mathrm{attn}} + n_{\mathrm{mlp}})$ at the input, and $K_\ell = 2(L - \ell)$ at the input of block $\ell$ for $L$ the total number of blocks. Block rates are defined on the gradient amplitude (Definition~\ref{def:block_rate}), so the corresponding Fisher log--log slopes are twice these: $\alpha = 4(n_{\mathrm{attn}} + n_{\mathrm{mlp}})$ and $\alpha_\ell = 4(L - \ell)$, matching Proposition~\ref{prop:attn_chain_softmax}(a)'s un-saturated $\alpha_{W_O} = 4k$ read at the $W_O$ output of block $\ell - 1$, which is the input of block $\ell$ and has $k = L - \ell$ blocks above it. Block-composition validation at $n = 2$ across the four ordered attention/MLP block pairs reproduces the predicted slopes.

\textbf{Scope caveat.} Additivity is not what governs a pure-attention chain at any depth: attention fails the transfer hypothesis of Theorem~\ref{thm:bridge_composition} at every depth (Theorem~\ref{thm:bridge_attn_backward}, where the backward transfer is a context-dependent token-space action rather than a scalar on the dead component). The additive value $4k$ is nonetheless the right answer wherever the $V$--$O$ chain is the smaller of the two routes of Proposition~\ref{prop:attn_chain_softmax}(a), that is wherever $k \le p + 2$; beyond that the score-path floor $4p + 8$ takes over and the sum overstates the rate. On practical residual-wrapped transformers the question does not arise, because the residual skip gives $K^{\mathrm{fwd}} = 0$ at block outputs regardless of what lies on the weight edges (Corollary~\ref{cor:sigma-min-res}).
\end{corollary}

\begin{proposition}[Refined attention-chain per-component rates, closing Remark~\ref{rem:attn_composition_anomaly} structurally]
\label{prop:attn_chain_softmax}
Consider a pure-attention chain of $N$ single-head self-attention blocks (no residuals, no MLP, no LayerNorm) at the canonical-aligned uniform-perturbation trajectory $W_*^{(j)}(t) = W_*^{(j)*} + t \cdot e_d e_d^\top$ for all $* \in \{Q, K, V, O\}$ and all $j \in \{0, \ldots, N-1\}$, with input $X \sim \mathcal{N}(0, I_d)$ and squared-error loss against a canonical-init target plus Gaussian noise. Fix a probe block $p \in \{0, \ldots, N-1\}$ and let $u = e_d$ denote the canonical dead direction. Let $\alpha_{W_*}(p, N)$ denote the asymptotic log--log slope of $u^\top G_{W_*^{(p)}} u$ versus $t$ as $t \to 0$, where $G_{W_*^{(p)}} = \mathbb{E}[g g^\top]$ is the gradient covariance at the output of $W_*^{(p)}$ (so $\alpha = 2 \cdot \mathrm{rate}(\sqrt{u^\top G u})$, twice the rate of the gradient's dead-channel scalar). Let $k := N - 1 - p$ (depth from output). Then:
\begin{enumerate}
\item[(a)] $\alpha_{W_O}(p, N) = \min\bigl(4k,\ 4p + 8\bigr)$ \hfill (the smaller of the $V$--$O$ chain and the score-path floor).
\item[(b)] $\alpha_{W_V}(p, N) = \alpha_{W_O}(p, N) + 2$ \hfill (within-block invariant, whichever route sets (a)).
\item[(c)] $\alpha_{W_Q}(p, N) = \alpha_{W_K}(p, N) = 4p + 2$ \hfill (forward-chain rate).
\end{enumerate}
\end{proposition}

\begin{proof}[Sketch]
All three parts follow from canonical-coordinate calculus and the chain rule. Part (a) is a minimum over two backward routes into $W_O^{(p)}$, and the ingredients of both are already fixed by Theorems~\ref{thm:bridge_attn} and \ref{thm:bridge_attn_backward}.

\textbf{(a) Route 1, the $V$--$O$ chain.} The dead-channel gradient $\partial L/\partial Y^{(p)}_{n, d}$ propagates from the loss through blocks $p+1, \ldots, N-1$. Each upstream block contributes a factor of $t^2$ via the dual of Theorem~\ref{thm:bridge_attn}'s $V$-path (Theorem~\ref{thm:bridge_attn_backward}: $\partial Y^{(j)}_{n', d}/\partial Y^{(j-1)}_{n, d} = t^2 A^{(j)}_{n', n}$ at leading order, one $t$-factor each from $W_V^{(j)}$'s and $W_O^{(j)}$'s dead diagonal). Composing $k = N-1-p$ such steps gives amplitude $\Theta(t^{2k})$, hence a contribution $4k$ to $\alpha_{W_O}$.

\textbf{(a) Route 2, the score path, and the floor it sets.} The probe's output also reaches the loss through the \emph{scores} of every block above it, and that route does not decay with $k$. Fix a block $m > p$. Its input $H^{(m)}$ has dead component $\Theta(t^{2m})$ by the forward rate $2$ per block of Theorem~\ref{thm:bridge_attn}. Inside block $m$ the dead channel enters the scores through $Q^{(m)}_{n, d} = t\, H^{(m)}_{n, d}$ and $K^{(m)}_{m', d} = t\, H^{(m)}_{m', d}$, one factor of $t$ from each of $W_Q^{(m)}$ and $W_K^{(m)}$. Since $\partial L/\partial S^{(m)}$ is $\Theta(1)$ (it is built from the non-dead channels, whose backward signal does not decay), the score path deposits into the dead channel of block $m$'s input gradient at amplitude order
\[
2m + 2 \qquad (\text{forward } 2m,\ \text{plus one } t \text{ each from } W_Q^{(m)}, W_K^{(m)}).
\]
That deposit then walks back down to $Y^{(p)}$ along the $V$--$O$ chain, costing $t^2$ for each of the $m - 1 - p$ intervening blocks, for a total amplitude order $(2m + 2) + 2(m - 1 - p) = 4m - 2p$. Minimising over $m > p$ puts the dominant deposit at the block immediately above the probe, $m = p+1$, giving amplitude order $2p + 4$ independently of how many blocks lie above.

Both routes deliver into the same scalar, and their leading coefficients do not cancel generically, so the amplitude order is $\min(2k,\, 2p+4)$ and $\alpha_{W_O} = \min(4k,\, 4p + 8)$. The floor binds exactly when $k > p + 2$, which is why it is invisible in shallow chains: at $N \le 3$ no cell has $k > p+2$ with $k \ge 1$, and $N = 4$ exposes it first at the single cell $p = 0$, $k = 3$.

\textbf{(b) The $W_V$–$W_O$ invariant.} The gradient at $V^{(p)}$'s dead-channel output, $\partial L/\partial V^{(p)}_{m, d}$, depends on $Y^{(p)}_{n, d'}$ only through dead-row $d' = d$ of $W_O^{(p)}$, because $V^{(p)}_{m, d}$ feeds $(AV)_{n, d}$ which feeds $Y^{(p)}_{n, d}$ via $(W_O^{(p)})_{d, d}$ alone (non-dead rows of $W_O^{(p)}$ map dead $V$ to non-dead $Y$ at order $0$, not $t$). So $\partial L/\partial V^{(p)}_{m, d} = \sum_n A^{(p)}_{n, m} \cdot (W_O^{(p)})_{d, d} \cdot \partial L/\partial Y^{(p)}_{n, d}$, picking up one extra factor of $t$ from $(W_O^{(p)})_{d, d}$ relative to $\partial L/\partial Y^{(p)}_{n, d}$ itself. Squaring gives $\alpha_{W_V} - \alpha_{W_O} = 2$, preserved under chain composition because saturation acts above the probe while $V \to W_O$ ordering is local.

\textbf{(c) Forward-chain rate at $W_Q, W_K$.} The dominant contribution to $\partial L/\partial Q^{(p)}_{n, d}$ does \emph{not} route through $\partial L/\partial Y^{(p)}_{n, d}$ at the dead row (which carries the saturated rate from (a) plus the $(W_O)_{d, d} = t$ factor); it routes through $\partial L/\partial Y^{(p)}_{n, d'}$ at \emph{non-dead} rows $d' \neq d$, which have rate $\Theta(1)$ (non-dead loss-gradient components do not decay through canonical block structure). The chain rule:
\[
\partial L/\partial Q^{(p)}_{n, d} \;=\; \sum_{d', m, m'} \partial L/\partial Y^{(p)}_{n, d'} \cdot \partial Y^{(p)}_{n, d'}/\partial A^{(p)}_{n, m} \cdot \partial A^{(p)}_{n, m}/\partial Q^{(p)}_{n, d},
\]
with $\partial A^{(p)}_{n, m}/\partial Q^{(p)}_{n, d}$ proportional to $K^{(p)}_{m, d}/\sqrt{d}$ via the softmax Jacobian. The dead-channel of the probe block's input $X^{(p)}_{n, d}$ scales as $\Theta(t^{2p})$ from the cumulative forward chain ($p$ upstream attention blocks, each contributing $t^2$ via Theorem~\ref{thm:bridge_attn}); then $K^{(p)}_{m, d} = t \cdot X^{(p)}_{m, d}$ has rate $2p+1$ (one extra factor from the $(d, d)$ entry of $W_K^{(p)}$). The non-dead $d' \neq d$ summands give $\Theta(1) \cdot \Theta(1) \cdot \Theta(t^{2p+1}) = \Theta(t^{2p+1})$, dominating the dead-row $d' = d$ contribution which carries an additional $(W_O)_{d, d} = t$ factor. Squaring: $\alpha_{W_Q}(p, N) = 2(2p+1) = 4p+2$. By the symmetric role of $Q$ and $K$ in $QK^\top$, the same rate applies to $W_K$.
\end{proof}

\begin{remark}[Empirical validation of the chain formulas]
\label{rem:attn_chain_validation}
The three formulas match the full probe profile at $N \in \{6, 7, 8\}$ and the cells $p \in \{0, 2, 3\}$ at $N = 4$ for $(d, n_h, \mathrm{seq\_len}) = (16, 2, 4)$, and the full profile at $N = 7$ for $(32, 4, 6)$: $31$ cells $\times$ four components, three seeds each, per-cell standard deviation below $10^{-3}$ on the fitted slope (parametric freeze-probe in fp64, fit on $t \le 10^{-2}$). The profile is non-monotone in $k$, rising to a peak where the two routes cross and falling on either side: at $N = 7$ it reads $8, 12, 16, 12, 8, 4, 0$ and at $N = 8$ it reads $8, 12, 16, 16, 12, 8, 4, 0$. Both branches of the minimum are therefore exercised, which a profile sampled at a few $p$ need not do. At the crossing $k = p + 2$ the two routes share the leading order, and in the two-route rate model the coefficient is the squared sum $(a + b)^2$ of the route amplitudes, machine-checked with the full trichotomy (Table~\ref{tab:machine_checked}): the V--O coefficient $a^2$ below the crossing, $(a+b)^2$ at it, the floor coefficient $b^2$ above it.
\end{remark}

\begin{conjecture}[Cross-configuration validity of the attention-chain rate formulas]
\label{conj:attn_chain_softmax_general}
The three closed forms of Proposition~\ref{prop:attn_chain_softmax} hold for every $(d, n_h, \mathrm{seq\_len})$ in the practical attention regime ($d \ge 8$, $n_h \ge 1$, $\mathrm{seq\_len} \ge 2$, dot-product softmax with $1/\sqrt{d}$ normalisation, canonical alignment). The derivation uses no property of $d$, $n_h$ or $\mathrm{seq\_len}$, so what the conjecture asserts beyond the proposition is that the two routes' leading coefficients do not cancel at any configuration in that regime. Measured at $(16, 2, 4)$ and $(32, 4, 6)$, and separately at $d \in \{8, 16, 24, 32\}$, $n_h \in \{1, 2, 3, 4\}$, $\mathrm{seq\_len} \in \{3, \ldots, 8\}$ on the whole profile.
\end{conjecture}

\begin{remark}[Scope of Proposition~\ref{prop:attn_chain_softmax}]
\label{rem:attn_chain_softmax_scope}
All three parts are derived, and each is a leading-order statement resting on a non-cancellation between contributions of equal order: in (a) between the two routes, in (b) and (c) between the dead-row and non-dead-row summands. What the proposition does not supply is a bound on the finite-$t$ corrections, so the formulas are asymptotic in $t$ and are read off a fit window, not from a single $t$. Configuration independence is Conjecture~\ref{conj:attn_chain_softmax_general}.
\end{remark}

\begin{remark}[Correction to version~1]
\label{rem:k_star_open}
Version~1 of this paper stated Proposition~\ref{prop:attn_chain_softmax}(a) as $\alpha_{W_O} = 4\min(k, k_0)$ with a saturation depth $k_0$ measured as $2$, and posed a closed form for $k_0$ as an open problem. That statement and that open problem are withdrawn: the rate is the two-route minimum above, which needs no fitted constant. The fitted value was the $p = 0$ case of that minimum: at the input-adjacent probe the floor $4p + 8$ equals $8$, so $\min(4k, 8) = 4\min(k, 2)$, and the measured $k_0 = 2$ was the floor read as a saturation depth.
\end{remark}

\begin{remark}[Resolution of Remark~\ref{rem:attn_composition_anomaly}'s three empirical findings]
\label{rem:anomaly_resolution}
The empirical signatures listed in Remark~\ref{rem:attn_composition_anomaly} are derived consequences of Proposition~\ref{prop:attn_chain_softmax}:
\begin{itemize}\itemsep=0pt
\item $W_O$ \emph{capped rather than growing as $4k$}: from (a), $\alpha_{W_O}(p, N) = \min(4k, 4p+8)$. At probe positions near the input the floor $4p + 8$ is the smaller branch, so the rate stops tracking depth above the probe and tracks distance from the input instead.
\item $W_Q, W_K$ \emph{rates scale with input position $p$ rather than output position $k$}: from (c), $\alpha_{W_Q}(p, N) = 4p + 2$ depends only on $p$.
\item $\alpha_{W_V} - \alpha_{W_O} = 2$ invariant: from (b), a within-block ordering consequence that holds whichever route sets (a).
\end{itemize}
\end{remark}

\paragraph{Scope.}
Theorems~\ref{thm:bridge_attn} and \ref{thm:bridge_attn_backward} cover single-head self-attention under canonical init with shared dead direction across $\{W_Q, W_K, W_V, W_O\}$. Extensions:
\begin{itemize}
\item \emph{Multi-head, shared dead direction across heads.} Each head has its own $Q, K, V$ projections. Under the assumption that the same dead direction $u = e_d$ acts on each head (a canonical-construction property, not generic), per-head analysis yields rate $2$ per head, and concatenation preserves it. A parametric freeze-probe sweep across $d \in \{16, 32, 64, 128, 256, 384, 768\}$ and $n_h \in \{1, 2, 4, 8\}$ (all $d_h = d/n_h \ge 2$), 3 seeds per configuration ($7 \times 4 = 28$ configurations, $84$ runs), gives asymptotic rate (fit on $t \le 10^{-2}$) of $2.000 \pm 0.000$ on $W_Q$, $W_K$, $W_V$ across all configurations.
\item \emph{Multi-head, per-head independent dead directions.} If each head carries its own dead direction in its $d_h$-dimensional sub-block, the multi-direction \ifsupp{}extension\else theorem (Theorem~\ref{thm:bridge_multi})\fi{} applies per head with $m \le n_h$ directions; rates are independent across heads. Not separately tested at multi-head scale at the time of writing; natural follow-up.
\item \emph{Cross-attention with $X_q \neq X_k$.} The bilinear $Q K^\top$ entry $(QK^\top)_{n,m}$ contains $t^2 (X_q)_{n,d} (X_k)_{m,d}$ at the dead channel; the rate-2 forward propagation extends if both $\mathbb{E}[(X_q)_{n,d}^2]$ and $\mathbb{E}[(X_k)_{m,d}^2]$ are $\Theta(1)$. If only one of $X_q, X_k$ has a dead-channel contribution (e.g., a static encoder-side dead direction with no decoder-side counterpart), the rate degenerates to a single-side factor.
\item \emph{Masked / causal attention.} $S_{n, m}$ is set to $-\infty$ for $m > n$; the structure of $A_0$ changes but the $t$-scaling argument is unaffected. Rate $2$ holds.
\item \emph{Residual attention (transformer block).} Combined with \ifsupp{}the residual-DAG bridge extension\else Theorem~\ref{thm:bridge_res}\fi, the residual skip dominates and the attention block's $k_{\mathrm{attn}} = 2$ is absorbed into $K^{\mathrm{fwd}} = 0$ at block outputs. Standard for pre-norm transformers.
\item \emph{Position encodings} (RoPE, learnable absolute, learnable relative). These compose with attention as natural follow-ups; closing each requires verifying that the position-encoding nonlinearity (e.g., RoPE phase mixing) preserves the dead-direction's $t$-scaling, which we have not done.
\end{itemize}
The Gaussian input assumption is used only to ensure the $\bar{X}_{n,d}$ second moment is $\Theta(1)$ uniformly; any input distribution with $\mathbb{E}[X_{n,d}^2] = \Theta(1)$ and finite fourth moment suffices.

\begin{remark}[Static-Fisher rate vs.\ Adam$+$CE attention trajectory]
\label{rem:bridge_attn_adam_scope}
Theorems~\ref{thm:bridge_attn} and \ref{thm:bridge_attn_backward} are static-Fisher statements at the parametric trajectory. On a transformer block trained with Adam$+$CE, the per-component trajectory rate readout is \emph{doubly} confounded: by the empirically-observed Adam non-equivariance under the ReLU-rescaling gauge together with noise-floor amplification along the CE-shift orbit (Remark~\ref{rem:adam_nondescent}), and by softmax cross-block coupling at $n \ge 4$ pure-attention chains (Remark~\ref{rem:attn_composition_anomaly}). The robust transformer-scale observable is therefore residual-stream $\sigma_{\min}$ (Corollary~\ref{cor:sigma-min-res}), not $u^\top G u$ on attention weights.
\end{remark}

\begin{remark}[Joint scope with rectangular widths and cross-entropy]
\label{rem:bridge_attn_joint_scope}
The single-head attention rate is local to one block, so it composes disjointly with the rectangular extension (which applies per-Linear inside the block) and the CE extension (which replaces only the output-head base case; attention is a hidden block, so its per-block rate is unaffected and the joint statement holds with $k_{\mathrm{attn}} = 2$ and the output-head $\Theta(1)$ base case driven by \ifsupp{}a non-degenerate data-averaged Hessian\else Assumption~\ref{ass:nondeg_pxy}\fi{} rather than $\sigma^{-2}$).
\end{remark}

\begin{figure}[ht]
\centering
\includegraphics[width=\textwidth]{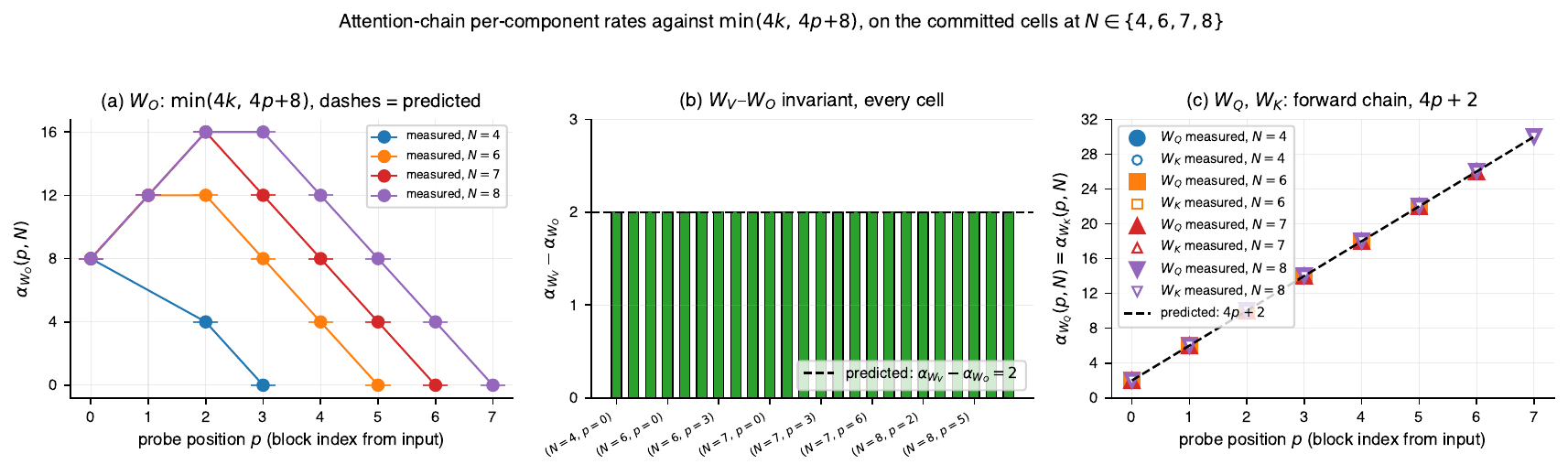}
\caption{Attention-chain per-component rates (Proposition~\ref{prop:attn_chain_softmax}). (a)~$\alpha_{W_O}$ on the committed probe cells at $N \in \{4, 6, 7, 8\}$ (full profile at $N \ge 6$; $p \in \{0,2,3\}$ at $N = 4$) against $\min(4k,\, 4p{+}8)$ (dashes), the smaller of the $V$--$O$ chain and the score-path floor. The profile rises and then falls in $k$: the chain branch sets the rate near the output, the floor takes over near the input, and the peak is where they cross. (b)~$\alpha_{W_V} - \alpha_{W_O} = 2$ at every cell. (c)~$\alpha_{W_Q} = \alpha_{W_K} = 4p + 2$, set by the forward chain and independent of depth above the probe.}
\label{fig:attn_chain_softmax_rates}
\end{figure}
 
\begin{figure}[t]
\centering
\includegraphics[width=\textwidth]{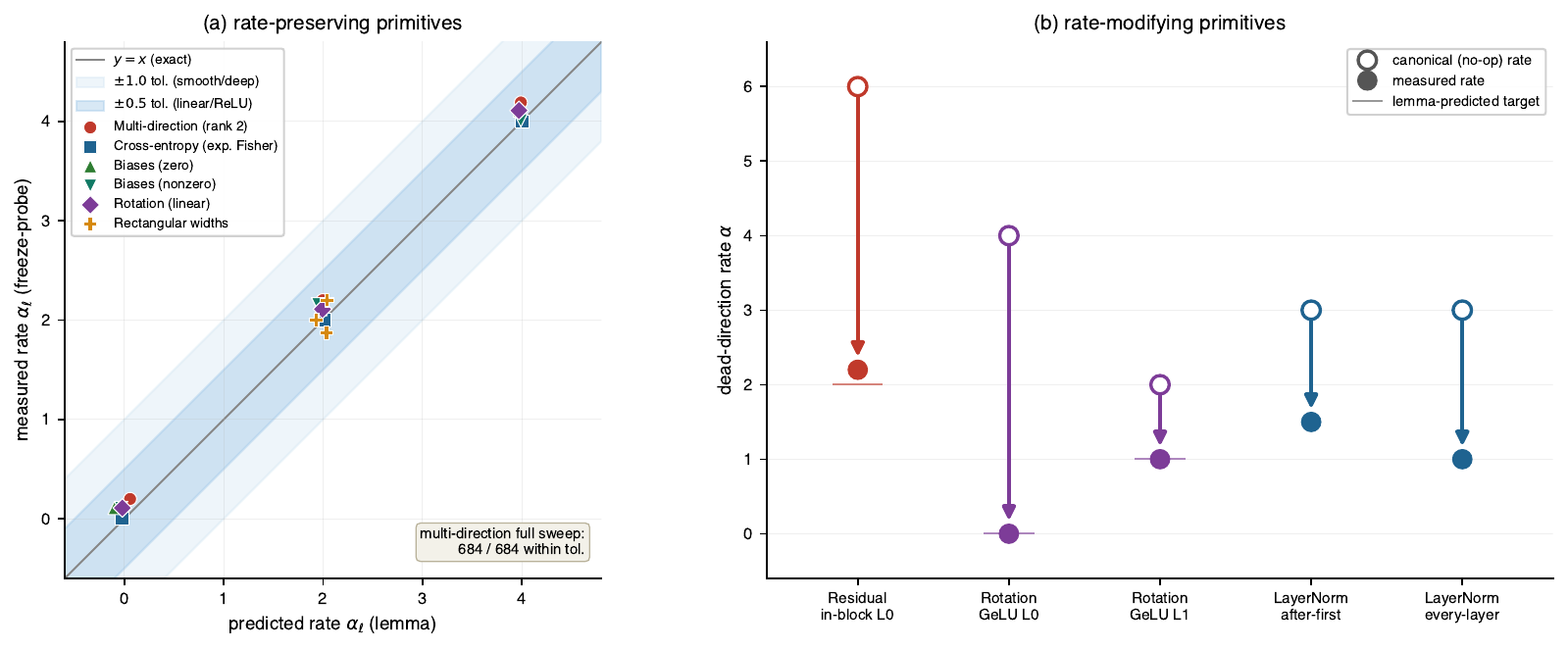}
\caption{Architectural freeze-probe roundup for the per-primitive lemmas of this section and the architectural catalogue (Appendix~\ref{app:theory:arch_catalogue}; numbers from Appendix~\ref{app:theory:arch_freeze_probe}). (a)~Primitives that preserve the canonical rate ladder: measured versus predicted per-layer dead-direction rate $\alpha_\ell$, with the activation-dependent tolerance bands. Every point lies on the diagonal, and the multi-direction sweep adds $684/684$ per-direction rates within tolerance. (b)~Primitives that change the rate: each arrow runs from the canonical (no-op) rate to the measured rate, and the measured change matches the lemma's predicted target (tick). A residual skip caps the in-block rate (Corollary~\ref{cor:res_block}), off-canonical rotation collapses the nonlinear rate (Proposition~\ref{prop:bridge_nonlinear_rot_negative}), and LayerNorm reduces the integer ladder to a fractional rate (Proposition~\ref{thm:bridge_ln}).}
\label{fig:arch_rate_roundup}
\end{figure}

\subsection{Remarks on Scope and Extensions}
\label{app:nonlinear_bridge}

\paragraph{The 2-layer case: a stronger closed form.}\label{rem:bridge_scope}
At $L = 2$, Theorem~\ref{thm:bridge} gives $\lambda_{\min}(G_1) = \Theta(t^2)$ and $\lambda_{\min}(G_2) = \Theta(1)$ under its canonical-aligned square-weights setup. The two-layer case in fact admits a stronger statement: a closed form holds under \emph{weaker} hypotheses that Theorem~\ref{thm:bridge} does \emph{not} subsume.
\begin{itemize}
\item \emph{Linear 2-layer:} For $W_1 \in \reals^{h \times d_\mathrm{in}}$, $W_2 \in \reals^{d_\mathrm{out} \times h}$ (\textbf{rectangular}), \textbf{arbitrary rank} $r^* < h$, and \textbf{arbitrary} activated direction $u \perp \col(W_2^*)$, $v \perp \mathrm{row}(W_1^*)$, one has the \emph{closed form} $G_1 = \sigma^{-2} (W_2^*)^\top W_2^*$ at the optimum, with eigenvalues $\sigma^{-2} s_1^2 \geq \cdots \geq \sigma^{-2} s_{r^*}^2 > 0 = \cdots = 0$ where $s_i$ are the singular values of $W_2^*$. Proof: at the optimum, the residual is $\varepsilon$, so $G_1 = \expect[g_1 g_1^\top] = \sigma^{-4} (W_2^*)^\top \expect[\varepsilon\varepsilon^\top] W_2^* = \sigma^{-2} (W_2^*)^\top W_2^*$ using $\varepsilon \perp x$ and $\expect[\varepsilon\varepsilon^\top] = \sigma^2 I$ (the output gradient of the Gaussian log-likelihood carries $\sigma^{-2}$, so its second moment carries $\sigma^{-4}$). Along the symmetric approach, $\lambda_{\min}^{>0}(G_1(t)) = \sigma^{-2} t^2 \|u\|^2 = \Theta(t^{2(k-1)})$ for $k = 2$.

\item \emph{Smooth-nonlinear 2-layer:} Same setup as above plus a smooth activation satisfying $\phi(0) = 0$, $\phi'(0) \neq 0$, $\phi \in C^2$. Along the symmetric approach, $\lambda_{\min}^{>0}(G_1(t)) = \phi'(0)^2 \sigma^{-2} t^2 \|u\|^2 + O(t^3)$. For $\phi = \mathrm{id}$, $\phi'(0) = 1$ recovers the linear result. This case handles rectangular weights and a dead direction off-canonical in the input and output maps, provided it remains a coordinate axis in the hidden basis: the element-wise $\phi'$ does not commute with a rotation of that basis, so, unlike the fully rotation-covariant linear case, a rotated hidden-basis direction masks the rate. The single $\phi$ application avoids the cross-layer coordinate mixing that forces canonical alignment at $L \ge 3$.
\end{itemize}

These 2-layer closed forms are \emph{not} specializations of Theorem~\ref{thm:bridge}: they hold under weaker hypotheses (rectangular weights, any rank $r^* < h$, any activated direction) and give a closed form rather than only a rate. Conversely, Theorem~\ref{thm:bridge} extends the 2-layer rate to $L \geq 2$, but requires square weights, canonical-basis alignment, and $r^* = h - 1$. The canonical-alignment assumption is WLOG in the linear case at any depth (rotation invariance), but is a genuine restriction in the nonlinear case at $L \geq 3$, where element-wise $\phi$ couples non-canonical directions through the forward-backward chain, breaking the dead-direction Schur complement reduction (Lemma~\ref{lem:integral-reduction-sub}) used in the proof.

Theorem~\ref{thm:bridge} and the 2-layer closed forms are therefore \emph{complementary}: the closed forms hold under weaker hypotheses at a single depth, while Theorem~\ref{thm:bridge} extends to arbitrary depth $L$ at the cost of stronger hypotheses. Joint-scope status of the closed extensions: Theorem~\ref{thm:bridge_rect} (rectangular widths; coordinate-aligned isometry chains for the nonlinear classes) closes any $L$, and Theorem~\ref{thm:bridge_multi} (multi-direction with asymmetric per-layer exponents) closes multi-rank singular configurations; these commute (the rectangular Schur reduction acts on the non-dead block, the multi-direction Schur reduction acts on the dead block) and the joint statement holds. Genuinely open: non-canonical dead directions at $L \geq 3$ under nonlinear activations (Proposition~\ref{prop:bridge_nonlinear_rot_negative}'s negative result; near-canonical continuity remains open per Remark~\ref{cor:bridge_near_canonical}).

\paragraph{KL order scales with depth.}\label{rem:deep_linear_k}
The KL order $k = L$ along the symmetric approach arises because activating a dead direction through the depth-$L$ chain requires $L$ coordinated perturbations (one per layer). Each factor contributes $O(t)$; their product gives $O(t^L)$ in the effective weight and $O(t^{2L})$ in the KL divergence. For $L = 2$, this recovers $k = 2$ (deep-linear reduced-rank-regression merge, analogous to the Gaussian mixture merge at $k = 2$). For general $L$, the KL order scales linearly with depth.

\paragraph{ReLU and non-smooth activations.}\label{rem:relu_bridge}
Classical bridge arguments require $\phi \in C^2$ with $\phi'(0) \neq 0$, which excludes ReLU. In the symmetric-canonical setup (P3), ReLU is handled by a direct argument: for $t > 0$ and $x^{(h)}$ of definite sign, $a_\ell^{(h)}$ has fixed sign along the entire approach trajectory, so $\phi'(a_\ell^{(h)}) \in \{0, 1\}$ is constant per sample. The Taylor argument is unnecessary. Numerically, the rate holds exactly, cleaner than smooth activations, which carry finite-$t$ Taylor corrections. A formal distributional-derivative treatment for non-aligned dead directions would be an interesting extension.

For 2-layer ReLU networks at $L = 2$ (validated in Table~\ref{tab:nonlinear_bridge}), the effective $\phi'(0)^2 = 1/2$ factor reflects isotropic Gaussian data passing through sign-symmetric ReLU: half the probability mass has $a_1^{(h)} > 0$ (contributing) and half has $a_1^{(h)} < 0$ (zero contribution), giving $\expect[\phi'(h)^2] = 1/2$. The rate $\Theta(t^2)$ survives, with the dead-direction Schur reduction of Lemma~\ref{lem:integral-reduction-sub} handling the sign-definite case directly (no Taylor correction needed).

\begin{table}[ht]
\caption{2-layer bridge validation across five activations (5 seeds, $6 \to 8 \to 4$ network). Fitted exponent $\alpha$ in $\lambda_{\min}^{>0}(G_1) \sim t^\alpha$; predicted $\alpha = 2.000$ for $k = 2$. Deviations reflect the finite-$t$ Taylor correction $O(t)$ for smooth activations; linear and ReLU match exactly within sampling noise.}
\label{tab:nonlinear_bridge}
\centering
\small
\begin{tabular}{lrrl}
\toprule
Activation & Exponent $\alpha$ & $k_\mathrm{est}$ & $\phi'(0)$ \\
\midrule
Identity (linear) & $1.995 \pm 0.003$ & $1.998$ & $1$ \\
GeLU & $2.066 \pm 0.008$ & $2.033$ & $0.5$ \\
Tanh & $1.914 \pm 0.003$ & $1.957$ & $1$ \\
Sigmoid (shifted) & $1.969 \pm 0.002$ & $1.985$ & $0.25$ \\
ReLU & $1.996 \pm 0.007$ & $1.998$ & undefined \\
\bottomrule
\end{tabular}
\end{table}

\begin{remark}[K-FAC approximation quality]
\label{rem:kfac_approx}
The theorem uses the K-FAC factorization $F_\ell \approx A_\ell \otimes G_\ell$. For linear models this is exact at the optimum, where the residual is pure noise and independent of $x$; along the approach the dead-row activation and gradient share the $x^{(h)}$ dependence, an $O(t^{2L})$ relative deviation that leaves the leading rate unchanged. For smooth and ReLU activations under MSE loss with Gaussian noise, $D_x$ and $\varepsilon$ are independent ($\varepsilon \perp x$), so the factorization is exact at leading order. Under general loss (cross-entropy), K-FAC becomes an approximation; the error introduced is of order $\mathrm{Cov}(\phi'(h)^2, \|g_\mathrm{head}\|^2)$, bounded and not affecting the leading-order rate. Empirically, close match across activations and losses (Tables~\ref{tab:multilayer_bridge}, \ref{tab:nonlinear_bridge}, \ref{tab:ce_bridge}) confirms the factorization is accurate near singularities.
\end{remark}

\paragraph{Regime axes that determine the trajectory exponent.}\label{rem:bridge_regime_taxonomy}
Theorem~\ref{thm:bridge} fixes the trajectory exponent $\lambda_{\min}(G_\ell) = \Theta(t^{2(L-\ell)})$ under a specific combination of conditions: noisy target ($\varepsilon \ne 0$), canonical-aligned symmetric perturbation, transversal singular minimum, isotropic Gaussian input, MSE loss, and (P1)/(P2)/(P3) activation. Changing any of these conditions can shift the exponent. Remark~\ref{rem:bridge_regime} (in the body) covers the noisy/noise-free axis explicitly (a $+2L$ exponent shift). The behaviour under non-canonical alignment at $L \ge 3$ with nonlinear activations is given a negative result by Proposition~\ref{prop:bridge_nonlinear_rot_negative}; near-canonical continuity at finite perturbation angle is open (Remark~\ref{cor:bridge_near_canonical}). The behaviour under non-trivial preconditioners (Adam-like, sign-only, K-FAC, Muon, Shampoo) is the empirical landscape of the optimizer-characterisation appendix; closed-form rate predictions for each preconditioner family remain open. Plain SGD is not automatically exempt: an optimiser can move the trajectory off the canonical stratum onto a different singular stratum with a different gate structure and a different dead-direction rate. Reading a trajectory rate therefore presumes the trajectory stayed on the stratum the theorem describes, which is a property of the optimiser and the architecture together rather than of the loss geometry alone. The remaining regime axes (initialization balance, sampling regime, data distribution, transient vs asymptotic behaviour, singularity type at $\theta^*$) each contribute a further dimension to the cell space; the present theorems populate the cell where these are all in their canonical sub-cases: balanced init under gradient flow, full-batch or large-batch SGD, isotropic Gaussian input, asymptotic $t \to 0$, transversal singularity. A full enumeration of which combinations are closed-form derivable versus require empirical characterisation is future work.
  
\part{The optimizer landscape}

The rate so far is a statement about geometry: it is a property of the loss landscape, independent of how a network reaches a singularity. Training is dynamics, run by a particular optimiser, and real networks carry continuous symmetries (weight rescalings, logit shifts, rotations) that leave the function unchanged while moving the parameters. This part makes the rate respect those symmetries, then asks which optimisers let a practitioner read it off an actual trajectory. Section~\ref{sec:theory:quotient} settles the geometry: the directional rate descends to the gauge quotient $\Theta / G$, with the same KL-order exponent regardless of which gauge representative a run happens to pick. Section~\ref{sec:theory:adam} turns to the dynamics: SGD on a $G$-invariant metric realises the quotient rate as gradient flow, Adam's per-coordinate preconditioner breaks the $G$-equivariance the readout depends on, and a dead-direction conditioner (DDCAdam) restores it by construction. The exponent is the same bridge invariant throughout; the question here is whether the optimiser preserves it.

\paragraph{Reach.} The scope narrows once more, from architecture to optimiser. The quotient Fisher rate is a metric-level statement and holds for any continuous Lie group symmetry $G$ of the loss, with its exponent equal to the KL order of the dead direction. Reading that rate off a training run is conditional: it requires the optimiser to be $G$-equivariant. SGD on a $G$-invariant metric qualifies, and DDCAdam qualifies by construction for the architectural gauge classes of \S\ref{sec:theory:architectural}; standard Adam does not, because the per-coordinate $1/\sqrt{\hat v}$ does not commute with non-axis-aligned gauge actions. A closed-form rate modifier for standard Adam on the alignment-rotated manifold is left open. Standard Adam is the optimiser in widest practical use, so this gap directly limits how broadly the trajectory readout applies. Equivariance is necessary but not sufficient: even with an equivariant optimiser, reading the rate off a run additionally requires the trajectory to be in a compression phase (\S\ref{subsec:theory:regimes}), since an equivariant optimiser preserves the rate but does not, on its own, make a network approach the singularity.
 
\section{Quotient Fisher rate}
\label{sec:theory:quotient}

Both information geometry and singular learning theory carry a notion of gauge invariance. In Amari's framework, a continuous Lie group $G$ acting on $\Theta$ leaves the family $\{p_\theta\}$ invariant when its orbits lie in level sets of $p_\theta$; the Fisher metric is then $G$-invariant, and the metric quotient $\Theta / G$ is a Riemannian submersion. In Watanabe's, gauge orbits are precisely smooth singular fibres of $\Sigma_T$. The two notions agree. This section extends the rate theorem to the gauge quotient: the directional Fisher rate is well-defined on $\Theta / G$, and its exponent is intrinsic to the KL geometry, independent of the gauge representative chosen (Corollary~\ref{cor:quotient_rate}). SGD on a $G$-invariant metric is gradient flow on the quotient and realises the rate (Corollary~\ref{cor:sgd_quotient}); we make this precise via two worked examples (cross-entropy logit shift; ReLU rescaling). The loss symmetries treated here are the parameter-space symmetries surveyed by \citet{ZhaoWaltersYu25}; \citet{Terin26} concurrently study gauge fixing for the positive-homogeneity case. The Adam-side scope, where the per-coordinate preconditioner breaks $G$-equivariance, is the subject of the following section.

\subsection{Quotient Fisher rate: gauge orbit as smooth fiber}
\label{app:proof:quotient_rate}

The Fisher rate theorem extends to the gauge quotient $\Theta/G$ as a corollary of the selection rule (Theorem~\ref{thm:selection_rule}): the gauge orbit $G \cdot \theta_0$ is a smooth submanifold of the singular set $S = \{\theta : p_\theta = p^*\}$, and the selection rule applied with $\tilde S$ chosen as the gauge orbit and $n$ chosen as a horizontal direction yields the rate on the horizontal subspace, equivalently on the quotient. A trajectory on $\Theta$ projects to a trajectory on the quotient that realises this rate when its optimizer is $G$-equivariant (sufficient condition, Corollary~\ref{cor:sgd_quotient}); non-equivariant optimizers pick up a projection bias that obstructs direct rate readout (Remark~\ref{rem:adam_nondescent}). SGD on a $G$-invariant metric is equivariant; Adam's diagonal preconditioner generically is not, which grounds the gauge-mode drift behaviour characterised in Remark~\ref{rem:adam_nondescent}.

\paragraph{Setup.} Let $G$ be a Lie group acting smoothly, freely, and properly on a parameter open set $\Theta^\circ$ with loss $L : \Theta^\circ \to \reals$ and model $\{p_\theta\}$ both $G$-invariant; we assume $\theta_0 \in \Theta^\circ$ lies on the principal orbit type, so the action remains free in a neighbourhood of $\theta_0$ (orbit-type strata where the action degenerates require equivariant stratification machinery and are deferred). The quotient $\bar\Theta := \Theta^\circ / G$ is a smooth manifold of dimension $\dim \Theta - \dim G$, and $\pi : \Theta^\circ \to \bar\Theta$ is a smooth submersion. Fix a $G$-equivariant horizontal distribution $H_\theta \subset T_\theta\Theta$ transversal to the vertical $V_\theta := T_\theta(G \cdot \theta)$.

\textbf{Choice of horizontal distribution by gauge.} The natural choice is the orthogonal complement under a $G$-invariant Riemannian metric on $\Theta^\circ$:
\begin{itemize}\itemsep=0pt
\item \emph{CE-shift gauge} ($G = \reals$ acting by translation on a coordinate block, e.g.\ adding a constant to all unembed logits): translation is an isometry of the Euclidean metric, so the standard Euclidean orthogonal complement is $G$-invariant and provides $H_\theta$ directly.
\item \emph{ReLU-rescaling gauge} ($G = (\reals_+)^{L-1}$ acting on adjacent layer pairs $(W_\ell, W_{\ell+1})$ by $(c W_\ell, c^{-1} W_{\ell+1})$): the action is \emph{not} isometric for the Euclidean metric (it scales paired blocks oppositely), so the Euclidean orthogonal complement is not $G$-invariant. One must instead use a $G$-invariant Riemannian metric on $\Theta^\circ$: natural choices are the Fisher information metric itself (which is $G$-invariant by the same $G$-invariance of $p_\theta$) or a log-coordinate metric of the kind used by \citet{TanakaKunin21} for kinetic-symmetry analysis.
\end{itemize}
The asymptotic Fisher rate ($\Theta(\cdot)$ statement) is independent of the specific choice; the SGD trajectory-rate corollary below uses the $G$-invariance of the metric to identify projected dynamics with gradient flow on the quotient.

\begin{lemma}[Quotient Fisher is well-defined and horizontally-PSD]
\label{lem:quotient_F}
The Fisher $F(\theta)$ vanishes on $V_\theta$ ($G$-invariance of $p_\theta$ makes vertical scores identically zero); hence $F(\theta)$ restricts to a PSD form on $H_\theta$, and the restriction descends to a well-defined Fisher metric $\bar F(\bar\theta)$ on $\bar\Theta$.
\end{lemma}

\begin{proof}[Proof sketch]
\emph{Vertical degeneracy.} For $v \in V_\theta$, pick a one-parameter subgroup $g(s) = \exp(sX)$ with $\partial_s|_{s=0}(g(s) \cdot \theta) = v$. $G$-invariance of $p_\theta$ gives $\log p_{g(s) \cdot \theta}(x) \equiv \log p_\theta(x)$ in $s$, so the directional score $v^\top \nabla_\theta \log p_\theta(x) \equiv 0$ in $x$. Hence $v^\top F(\theta) v = \expect_{p_\theta}[(v^\top \nabla \log p_\theta)^2] = 0$, and $F(\theta) v = 0$ by PSD-ness.
\emph{Well-definedness of $\bar F$.} For any two lifts $\theta_1, \theta_2 \in \pi^{-1}(\bar\theta)$ related by $\theta_2 = g \cdot \theta_1$ and any horizontal tangent vectors $v_H^{(1)}, w_H^{(1)} \in H_{\theta_1}$ with $G$-equivariantly-transported images $v_H^{(2)} = (dg)_{\theta_1}(v_H^{(1)})$, $w_H^{(2)} = (dg)_{\theta_1}(w_H^{(1)}) \in H_{\theta_2}$, differentiating $\log p_{g \cdot \theta}(x) = \log p_\theta(x)$ and evaluating under $\expect_{p_{\theta_2}} = \expect_{p_{\theta_1}}$ gives $F(\theta_2)(v_H^{(2)}, w_H^{(2)}) = F(\theta_1)(v_H^{(1)}, w_H^{(1)})$. Hence $\bar F(\bar\theta)$ is independent of the chosen lift.
\end{proof}

\begin{lemma}[Quotient KL and intrinsic KL order]
\label{lem:quotient_K}
$K(\theta) = \kl(p^* \| p_\theta)$ factors through $\pi$ as $K = \bar K \circ \pi$ ($G$-invariance of $p_\theta$ implies $G$-invariance of $K$). Along any horizontal direction $u \in H_{\theta_0}$ with $p_{\theta_0} = p^*$, the KL order of $\bar K$ on $\bar\Theta$ along $\pi_*(u)$ equals the KL order of $K$ on $\Theta$ along $u$. (Because $\pi$ is a submersion with $\ker d\pi_\theta = V_\theta$ and $H_\theta$ is transversal to $V_\theta$, the restriction $\pi_*|_{H_\theta} : H_\theta \to T_{\bar\theta}\bar\Theta$ is a linear isomorphism, so all derivatives of $K$ along $u$ correspond to derivatives of $\bar K$ along $\pi_*(u)$ of the same order.)
\end{lemma}

\begin{corollary}[Quotient Fisher rate; corollary of Theorem~\ref{thm:selection_rule}]
\label{cor:quotient_rate}
Let $\bar\theta_0 \in \bar\Theta$ be a singular minimum of $\bar K$ (equivalently, $\theta_0 \in \pi^{-1}(\bar\theta_0)$ satisfies $p_{\theta_0} = p^*$), and let $\bar u \in T_{\bar\theta_0}\bar\Theta$ be a unit horizontal direction with KL order $k \ge 2$. Let $\bar\theta(\bar t)$ denote the curve $\pi(\theta_0 + t u)$ with $u = \pi_*^{-1}(\bar u)$ the horizontal lift, written $\bar\theta_0 + \bar t \bar u$ in a chart where $\pi$ is linear. Then, under Theorem~\ref{thm:selection_rule}'s assumptions applied with $\tilde S = G \cdot \theta_0$ as the smooth fiber and normal direction $n = u$,
\[
\bar u^\top \bar F(\bar\theta(\bar t))\, \bar u \;=\; \Theta(\bar t^{2(k-1)}), \qquad \lambda_{\min}\bigl(\bar F(\bar\theta(\bar t))\bigr) \;=\; \Theta(\bar t^{2(k-1)}).
\]
\end{corollary}

\begin{remark}[Dependence of the rate on the curve]
\label{rem:quotient_rate_curve}
The image of a horizontal \emph{geodesic} is a different curve under a non-Euclidean invariant metric: its second-order deviation from the line enters the score at order $t^2$ against a leading score of order $t^{k-1}$, and an explicit translation-gauge family with metric $ (1+2b)\,da^2 + db^2 + dg^2$ measures rate $4$ along the geodesic image against $2(k-1) = 6$ along the projected line at $k = 4$. The corollary's curve is the projected line, which is what every instantiation in this paper measures. The rate is therefore a property of the projected line through $\bar\theta_0$ in the direction $\bar u$, and a different curve tangent to the same direction can read a different exponent.
\end{remark}

\begin{proof}[Proof sketch]
The gauge orbit $G \cdot \theta_0$ is a smooth submanifold of the singular set $S = \{\theta : p_\theta = p^*\}$ with $T_{\theta_0}(G \cdot \theta_0) = V_{\theta_0}$, so it is admissible as the smooth fiber for Theorem~\ref{thm:selection_rule}; the KL order along $n$ is $k$ by Lemma~\ref{lem:quotient_K}. Lemma~\ref{lem:quotient_F} identifies the horizontal-restricted Fisher with the quotient Fisher, so Theorem~\ref{thm:selection_rule}(a) transports to $\lambda_{\min}(\bar F(\bar\theta(\bar t))) = \Theta(\bar t^{2(k-1)})$ up to the $\Theta(1)$ arc-length reparameterisation. Caveat: the full singular set $S$ may strictly contain $G \cdot \theta_0$ (overparameterised networks have non-gauge degenerate directions, mixture models have multi-component strata), in which case Theorem~\ref{thm:selection_rule}'s transversality hypothesis on $H_{\theta_0}$ is stronger than gauge-direction-only transversality and must be checked at $\theta_0$. The corollary applies to the gauge-orbit component; non-gauge dead directions on $\bar\Theta$ are governed by Theorem~\ref{thm:fisher_decay} directly.
\end{proof}

\begin{corollary}[SGD trajectory rate on the quotient]
\label{cor:sgd_quotient}
Suppose the horizontal distribution arises from a $G$-invariant Riemannian metric on $\Theta^\circ$ (so $\pi : \Theta^\circ \to \bar\Theta$ is a Riemannian submersion with quotient metric inherited from horizontal lifts). Then continuous-time SGD on a $G$-invariant loss $L$ with metric-isotropic Gaussian noise projects under $\pi$ to gradient flow of the quotient loss $\bar L$ with metric-isotropic Gaussian noise on $\bar\Theta$ (this equals gradient flow of the quotient KL $\bar K$ up to time reparameterisation exactly when $\nabla L \propto \nabla K$, as in the well-specified Gaussian case; the noise projection additionally uses that the gauge orbits are minimal submanifolds of the invariant metric, which holds for the CE-shift gauge under the Euclidean metric, whose orbits are straight lines). If such a projected trajectory approaches $\bar\theta_0$ along a canonical-aligned horizontal direction $\bar u$ with KL order $k \ge 2$ satisfying the assumptions of Corollary~\ref{cor:quotient_rate}, then along the trajectory
\[
\bar u^\top \bar F(\bar\theta(t))\, \bar u \;=\; \Theta\bigl(\bar t(t)^{2(k-1)}\bigr),
\]
with $\bar t(t)$ the quotient-metric arc-length from $\bar\theta(t)$ to $\bar\theta_0$.
\end{corollary}

\begin{remark}[Discrete steps]
\label{rem:sgd_quotient_discrete}
Corollary~\ref{cor:sgd_quotient} is a statement about the continuous-time flow. For discrete SGD with step $\eta$ and a $G$-equivariant update $U$, one step satisfies $\pi(\theta + \eta U(\theta)) = \pi(\theta) + \eta\, d\pi_\theta U(\theta) + O(\eta^2)$ by Taylor's theorem, and $d\pi_\theta U(\theta)$ is the horizontal part of the update, so the projected discrete trajectory is the Euler scheme of the projected flow with a per-step error of order $\eta^2$. Over a trajectory of $n$ steps the accumulated deviation is $O(n\eta^2) = O(\eta T)$ at fixed horizon $T = n\eta$; whether that deviation is small against the leading rate $\bar t^{2(k-1)}$ along a given approach is a property of the run, not of the corollary, and the statement above makes no claim about it.
\end{remark}

\begin{remark}[Adam's non-equivariance and why the quotient rate is not Adam-trajectory-readable]
\label{rem:adam_nondescent}
Corollary~\ref{cor:sgd_quotient}'s projection step requires the optimizer's update rule $U(\theta, \nabla L)$ to be $G$-equivariant: for every $h \in G$, $(d h) \cdot U(\theta, \nabla L(\theta)) = U(h \cdot \theta, \nabla L(h \cdot \theta))$. SGD with $U(\theta, \nabla L) = -\eta \nabla L$ satisfies this exactly when the action is by isometries of the Euclidean metric: for an invariant loss $\nabla L(h \cdot \theta) = (dh)^{-\top} \nabla L(\theta)$, so $-\eta \nabla L$ is equivariant iff $(dh)^{-\top} = dh$ on the relevant subspace. Translation actions (the CE-shift gauge) and orthogonal actions qualify; non-isometric rescalings (the ReLU gauge with $dh = \diag(\alpha, 1/\alpha)$) do not, which is why Corollary~\ref{cor:sgd_quotient} assumes a $G$-invariant metric and gradient flow with respect to it rather than raw Euclidean SGD. Adam's diagonal preconditioner $U(\theta, \nabla L) = -\eta \cdot \nabla L \big/ (\sqrt{\hat v} + \epsilon)$ is \emph{not} $G$-equivariant in general. The two gauge mechanisms studied in this paper produce structurally distinct non-equivariance:
\begin{itemize}\itemsep=0pt
\item \textbf{CE-shift gauge}: the symmetry direction is a \emph{sum} over output coordinates of the unembed weight $W_L$ (adding a constant to all logit rows). Because the CE-shift action is a translation with $dh = I$ and gradients of the invariant loss are literally equal along the orbit, Adam's update, which depends only on the gradient history, is formally $G$-equivariant here. The operative failure is not equivariance but \emph{horizontality with respect to the noise floor}: the analytic gradient component along the orbit is exactly zero, the finite-batch and finite-precision residual along it is not, and Adam's per-coordinate $1/\sqrt{\hat v}$ amplifies that residual non-uniformly across the summand coordinates, driving drift along the orbit that plain SGD damps.
\item \textbf{ReLU-rescaling gauge}: the symmetry is a \emph{product} over paired blocks $(W_\ell, W_{\ell+1})$. Adam's diagonal preconditioner rescales each block independently, breaking the product invariant.
\end{itemize}
These are different mechanisms on different gauge groups: genuine non-equivariance for the rescaling gauge, noise-floor amplification along an orbit for the shift gauge. They share the practical consequence that Adam's per-coordinate $\hat v$ interacts badly with non-coordinate-aligned gauge structure, so the projected Adam trajectory $\pi(\theta(t))$ is not gradient flow on $\bar\Theta$ in either case, and Corollary~\ref{cor:sgd_quotient}'s sufficient condition for trajectory rate-readout fails: this is a structural non-applicability of the rate-readout along an Adam trajectory.

What an Adam trajectory does instead is gauge-mode drift at empirically-observed amplitudes, rather than approach to the singular minimum; this is established as an empirical observation with the structural mechanism sketched here. We do not derive a rate modifier for standard Adam itself: the route taken by Algorithm~\ref{alg:ddcadam} and Proposition~\ref{prop:ddcadam_equivariance} is to replace standard Adam's per-coordinate preconditioner with a $G$-equivariant variant (DDCAdam) for which Corollary~\ref{cor:ddcadam_quotient_rate} gives the trajectory-rate readout directly. Characterising standard (non-equivariant) Adam's quantitative rate modifier, as opposed to constructing a $G$-equivariant alternative, remains open.

\paragraph{Per-coordinate non-equivariance: trajectory effects.} The trajectory consequences depend on which signal the analytic gradient is small on:
\begin{itemize}\itemsep=0pt
\item \emph{Gauge-redundant losses.} On a continuous loss symmetry where the analytic gradient is exactly zero, $1/\sqrt{\hat v}$ amplifies the FP$+$finite-batch-noise floor non-uniformly across the symmetry direction's coordinates, producing drift in the gauge mode that is absent under SGD (Adam$+$CE under the logit-shift gauge).
\item \emph{Non-canonical regimes (e.g., ReLU MLPs).} On architectures where canonical preservation of the dead direction is not automatic, Adam-like preconditioning rotates the trajectory off the canonical dead-direction manifold over training. Whether a different observable, hyperparameter regime, or alignment-evolution model recovers a clean trajectory rate under such preconditioners is open. The mechanism is the trajectory's relationship to canonical alignment: on canonical-aligned trajectories the rate prediction is preserved across preconditioner families.
\end{itemize}
The unifying picture: the analytic gradient on the relevant signal is small (zero on a gauge orbit, small on the alignment-preserving update direction in a ReLU MLP), the finite-batch / finite-precision noise is non-zero, and Adam's per-coordinate preconditioner amplifies that noise non-uniformly. Trajectory-rate readability under non-SGD optimizers requires either a $G$-equivariant preconditioner (constructed below as DDCAdam; \citealp{LauSu26} give a concurrent symmetry-compatible design principle for the same component symmetries, and \citealp{DePaviaCharisopoulosWillett25} instead restore equivariance by reparameterising the objective) or alignment-preservation control on the trajectory in the non-canonical case.

\paragraph{Constructive recovery of the rate readout on the gauge case (DDCAdam).} For the architectural gauge classes of \S\ref{sec:theory:architectural} (CE row-shift, CE shift bias, ReLU rescaling, LayerNorm scale), rate-readability is recovered constructively by Algorithm~\ref{alg:ddcadam} (DDCAdam), an Adam-family preconditioner whose update map is $G$-equivariant by Proposition~\ref{prop:ddcadam_equivariance}; this is an equivariant alternative, not a rate modifier for standard Adam (which remains open, above). Corollary~\ref{cor:ddcadam_quotient_rate} then gives the trajectory-rate readout $u^\top \bar F u_t = \Theta(\bar t^{2(k-1)})$ on the projected DDCAdam trajectory, the same rate as SGD on a $G$-invariant metric (Corollary~\ref{cor:sgd_quotient}). DDCAdam is therefore theorem-compatible for trajectory rate-fits in the same regime as SGD on the supported gauge classes.

What remains open after this closure is the non-canonical-regime case: closed-form derivation of the per-trajectory rate on the alignment-rotated manifold is open work.

Corollary~\ref{cor:quotient_rate} remains a statement about Fisher geometry. Restoring Adam-family trajectory rate-readability requires either (i)~a $G$-equivariant preconditioner (constructed for the architectural gauges of \S\ref{sec:theory:bridge} by Algorithm~\ref{alg:ddcadam} (DDCAdam) and shown to satisfy the rate readout in Corollary~\ref{cor:ddcadam_quotient_rate}), or (ii)~gauge-fixing the loss at training time. The auxiliary $z$-loss penalty \citep{ShazeerMesh18,ZophSTMoE22} (a softmax-normaliser regulariser, distinct from the same-named spherical-family loss of \citealp{deBrebissonVincent16}) adds a coercive term $\alpha\,\mathbb{E}_x[(\log Z)^2]$ that softly fixes a gauge section by penalising log-partition drift, suppressing logit-shift drift to $O(1/\alpha)$ rather than removing the gauge entirely; combined with an optimizer whose residual non-equivariance under the remaining symmetries (ReLU rescaling, etc.) is bounded, this is the practical route to rate-compatible CE training under adaptive optimizers without the explicit gauge spec required by route~(i).
\end{remark}

\paragraph{Worked example 1: softmax cross-entropy and Z-loss as a gauge section.}
For supervised classification with softmax+CE on a $C$-class problem with output bias $b_L \in \reals^C$, the gauge group is $G = \reals$ acting by $(W_L, b_L) \mapsto (W_L, b_L + c \cdot \mathbf{1}_C)$ (and trivially on all other parameters). The action is free and proper; the vertical subspace $V_\theta \subset T_\theta\Theta$ is one-dimensional, spanned by the parameter-space direction $\partial / \partial b_L$ in the $\mathbf{1}_C$ component. The quotient $\bar\Theta$ has codimension $1$ and is identified with any smooth $1$-codimensional section of $\Theta^\circ$ transversal to the orbit, e.g.\ the slice $\{b_L^{(C)} = 0\}$ (fix the last-class bias to zero), the slice $\{\mathbf{1}^\top b_L = 0\}$ (zero-mean bias), or the slice $\{\log \sum_c \exp z_c(x_0; \theta) = 0\}$ at a fixed reference input $x_0$. The Euclidean metric on $\Theta$ is $G$-invariant (the action is by translation), so Corollary~\ref{cor:sgd_quotient} applies directly: Euclidean-SGD on a gauge-fixed parameterisation realises the trajectory rate $\Theta(\bar t^{2(k-1)})$.

The auxiliary $z$-loss penalty \citep{ShazeerMesh18,ZophSTMoE22} is the existing gauge-fix construction in this setting (not novel here; we use it as the standard reference; it is distinct from the same-named spherical-family loss of \citealp{deBrebissonVincent16}). $L_\mathrm{Z}(\theta) := L_\mathrm{CE}(\theta) + \alpha \,\mathbb{E}_x\bigl[(\log \sum_c \exp z_c(x; \theta))^2\bigr]$ adds a soft penalty whose minimum at $\alpha \to \infty$ enforces the data-averaged slice $\{\mathbb{E}_x \log \sum_c \exp z_c = 0\}$, providing a smooth gauge-fixing potential that converges to a section of $\pi$. Corollary~\ref{cor:quotient_rate} then predicts that under SGD on Z-loss (or under SGD with any explicit hard gauge fix), the rate of $\lambda_{\min}(\bar F)$ along the canonical-aligned horizontal approach equals $\bar t^{2(k-1)}$ at KL order $k$, the same rate as the MSE bridge of Theorem~\ref{thm:bridge}. Under Adam on Z-loss, the CE-shift gauge is removed but the residual non-equivariance under other continuous symmetries (ReLU rescaling, LN scale) remains; Remark~\ref{rem:adam_nondescent} predicts that drift persists; logit-level gauge-fixing under Adam reduces but does not eliminate the gauge-mode drift.

\paragraph{Worked example 2: ReLU rescaling.}
For a network with ReLU activations between weight layers $W_\ell, W_{\ell+1}$, the rescaling $(W_\ell, W_{\ell+1}) \mapsto (c W_\ell, c^{-1} W_{\ell+1})$ for $c > 0$ leaves the network function unchanged. The full ReLU gauge group on an $L$-layer chain is $G = (\reals^+)^{L-1}$, acting block-diagonally on adjacent weight pairs. The action is free and proper on the open subset where adjacent weight matrices are non-zero. Corollary~\ref{cor:quotient_rate} applies on the quotient $\bar\Theta = \Theta^\circ / (\reals^+)^{L-1}$; the bridge rate $\lambda_{\min}(G_\ell) = \Theta(t^{2(L-\ell)})$ of Theorem~\ref{thm:bridge} is a horizontal-direction rate on this quotient (the canonical-aligned approach implicitly fixes a smooth section: the symmetric perturbation $W_\ell^* + t \cdot e_h e_h^\top$ at fixed-norm canonical $W_\ell^*$ picks one orbit representative per quotient point, so Theorem~\ref{thm:bridge}'s statement is Corollary~\ref{cor:quotient_rate} applied to that section). The Euclidean metric is \emph{not} $G$-invariant for this action, so Corollary~\ref{cor:sgd_quotient} does not apply to Euclidean SGD here at all: Euclidean gradient descent on an invariant loss does not descend to the quotient under this gauge (gauge-related initialisations diverge to relative function distance $0.84$ within $40$ steps, while the equivariant update of Algorithm~\ref{alg:ddcadam} holds them together at $6 \times 10^{-12}$). What SGD does preserve is the Noether charge $\|W_\ell\|_F^2 - \|W_{\ell+1}\|_F^2$ (up to an $O(\eta^2)$ discretisation drift), which fixes a section per initialisation; whether the positional rate read along that section matches the quotient rate is not established by any result in this paper. The ReLU-rescaling gauge-mode drift under Adam is non-trivial relative to the SGD baseline, consistent with Adam's non-equivariance under this group.

\begin{remark}[Connection to Noether and kinetic-symmetry-breaking analyses]
\label{rem:noether_connection}
The conserved-charge analyses of \citet{KuninSagastuyBrenaGanguli21} (Noether-style conservation laws for SGD on $G$-invariant losses) and \citet{TanakaKunin21} (Lagrangian formulation of kinetic symmetry breaking when the optimizer's metric breaks the loss's symmetry) characterise the same optimizer-symmetry interaction on the parameter side. Corollary~\ref{cor:quotient_rate} characterises it on the Fisher / observable side: gradient flow conserves the charge at its initial value (for the rescale symmetry, $\|W_\ell\|_F^2 - \|W_{\ell+1}\|_F^2$), so an SGD trajectory stays on one section of the orbit foliation and the quotient rate is read along it; under Adam the kinetic term breaks the conservation law, the trajectory wanders across sections, and the rate is not directly accessible. The two pictures are complementary: Noether for the dynamics, the quotient rate corollary for the geometric observable.
\end{remark}
 
\section{Adam non-equivariance and its constructive closure}
\label{sec:theory:adam}

Corollary~\ref{cor:sgd_quotient} gives the sufficient condition for trajectory rate-readout: a $G$-equivariant optimiser projects to gradient flow on the quotient, so the rate exponent reads directly along the trajectory. SGD on a $G$-invariant metric satisfies it; Adam, AdamW, and most adaptive optimisers in practice do not. A closed-form rate modifier for standard Adam on the alignment-rotated manifold would extend the bridge to Adam-class dynamics; deriving one is open analytical work, not pursued here. This section takes the constructive route instead: we replace Adam's per-coordinate normalisation with a $G$-equivariant alternative that satisfies the sufficient condition by design. We first establish the structural reason Adam fails, then construct \emph{DDCAdam}, an Adam-family preconditioner whose update map is $G$-equivariant for the architectural gauge classes of \S\ref{sec:theory:architectural} (CE row-shift, ReLU rescale, LayerNorm scale). DDCAdam is the Adam instance of a broader \emph{dead-direction conditioner} (DDC) family: the same vertical--horizontal orbit decomposition lifts any base preconditioner to a $G$-equivariant variant.

Adam's per-coordinate preconditioner $-\eta\, \nabla L / (\sqrt{\hat v} + \epsilon)$ does not commute with non-axis-aligned group actions: a rotation of parameter space mapping one gauge representative to another sends Adam's preconditioned step to a different vector than the corresponding Adam step in the rotated frame. Two distinct mechanisms follow, on different gauge groups. Genuine non-equivariance acts on the rescaling gauges: alignment rotation in non-canonical regimes (ReLU MLPs), where over training the preconditioner pulls the trajectory off the canonical dead-direction manifold, a drift absent under the $G$-invariant-metric SGD of Corollary~\ref{cor:sgd_quotient}, and destabilisation of the rate readout itself, since the quotient theorem's gradient-flow-projection condition fails. On the cross-entropy logit-shift gauge Adam is formally equivariant (the action is a translation), and the operative failure is horizontality with respect to the noise floor: Adam amplifies finite-sample noise on the gauge orbit well above the level plain SGD leaves it at (Remark~\ref{rem:adam_nondescent}). We name these mechanisms and scope their consequences.
 
\subsection{Constructive equivariant Adam: DDCAdam}
\label{sec:theory:ddcadam}

\paragraph{Construction: DDCAdam.}
A constructive resolution of the open question in Remark~\ref{rem:adam_nondescent} requires an Adam-family preconditioner whose update rule is $G$-equivariant for the architectural gauges of \S\ref{app:bridge_ce} (CE row-shift) and \S\ref{app:bridge_rect} (ReLU rescaling), and \S\ref{app:bridge_ln} (LayerNorm scale). \emph{DDCAdam} is one such construction. The general principle is to decompose the second-moment estimator $\hat v$ into an orbit-collapsed vertical component (one scalar per gauge-orbit dimension) and a per-coordinate horizontal component, computed on a $G$-invariant frame. The vertical component is then updated under one of three modes (frozen, SGD, or Adam) while the horizontal component runs standard per-coordinate Adam on the horizontally-projected gradient.

\begin{definition}[Gauge specification]
\label{def:gauge_spec}
A \emph{gauge specification} for a parameter tensor $W \in \reals^d$ under a gauge group action $\sigma_G: G \times \reals^d \to \reals^d$ is the tuple
\[
\mathrm{Spec}(W; G) = (P_h(W), P_v(W), n_V, q_V, q_V^{-1}),
\]
where $P_v(W) : T_W \reals^d \to V_W$ projects onto the vertical (gauge-tangent) subspace at $W$, $P_h(W) = \mathrm{Id} - P_v(W)$ projects onto the horizontal complement, $n_V = \dim V_W$ is the orbit-tangent dimension, $q_V: V_W \to \reals^{n_V}$ collapses a vertical-projected tensor to its $n_V$ scalar components, and $q_V^{-1}: \reals^{n_V} \to V_W$ broadcasts back along the orbit direction.

For translation-type gauges ($G = \reals^k$ acting by additive shift), $P_h, P_v$ are constant linear projectors. For multiplicative gauges ($G = \reals^+$ or $(\reals^+)^k$ acting by component-wise rescaling), the projectors are computed in log-coordinates so that the action becomes additive; explicit formulas for the \texttt{ReLURescaleGauge} and \texttt{LNScaleGauge} cases are given in the gauge-spec catalogue at the end of this section.
\end{definition}

\begin{algorithm}[H]
\caption{DDCAdam (one step) for a gauge block. Inputs: parameters $W$ (one or more tensors), gradients $g$, gauge spec $\mathrm{Spec}(W; G)$, state, hyperparameters $(\eta, \beta_1, \beta_2, \epsilon, \lambda, \mathrm{vmode})$. Output: updated parameters and state.}
\label{alg:ddcadam}
\begin{enumerate}\itemsep=0pt
\item \textbf{Decoupled weight decay.} If $\lambda \neq 0$: $W \leftarrow (1 - \eta \lambda)\, W$. The global rescaling commutes with the multiplicative gauge action; for the affine translation gauges it is orbit-exact only at $\lambda = 0$, so the equivariance identity is stated at $\lambda = 0$ on translation-gauge blocks (the shrinkage of the non-compact shift mode leaves the projected trajectory unchanged).
\item \textbf{Gauge decomposition} (spec-provided; every component below is $G$-invariant). For \emph{translation} gauges the spec returns the vertical scalars $g_V = q_V(P_v(W)g) \in \reals^{n_V}$ and the ambient horizontal gradient $g_H = P_h(W)\, g$. For \emph{multiplicative} gauges it works in log-norm coordinates per block $j$: with $\rho_j = \|W_j\|$ and the log-norm gradient $\hat g_j = \langle g_j, W_j\rangle$, it returns a \emph{radial} scalar $g_U$ (the orbit-invariant combination of the $\hat g_j$), the gauge-mode scalars $g_V$ (the remaining combinations of the $\hat g_j$), and the \emph{tangential} tensors $\rho_j\, g_j^{\mathrm{tan}}$ with $g_j^{\mathrm{tan}} = g_j - (\hat g_j / \rho_j^2)\, W_j$.
\item \textbf{Moments.} Maintain bias-corrected Adam first and second moments of each gauge-decomposed component: the invariant horizontal components (per-coordinate on the tangential tensors $\rho_j g_j^{\mathrm{tan}}$, scalar on $g_U$; for translation gauges, per-coordinate on $g_H$) and the orbit-collapsed gauge modes $g_V$.
\item \textbf{Vertical update.} $u_V = 0$ (frozen) $/\ \eta\, \hat m_V$ (sgd) $/\ \eta\, \hat m_V / (\sqrt{\hat v_V} + \epsilon)$ (adam), on the gauge-mode scalars $g_V$.
\item \textbf{Horizontal update.} Per-coordinate Adam on each invariant horizontal component: a scalar step $\Delta_U$ on $g_U$ and tangential updates $\mathrm{upd}_j$ on $\rho_j g_j^{\mathrm{tan}}$ (for translation gauges, an ambient step $\Delta W_H = \eta\, \hat m_H / (\sqrt{\hat v_H} + \epsilon)$).
\item \textbf{Re-projection.} \emph{Multiplicative gauges:} $\mathrm{upd}_j \leftarrow \mathrm{upd}_j - (\langle \mathrm{upd}_j, W_j\rangle / \rho_j^2)\, W_j$, restoring orthogonality to $W_j$ after the per-coordinate normalisation. \emph{Translation gauges:} $\Delta W_H \leftarrow P_h(\Delta W_H)$, restoring horizontality: per-coordinate division does not preserve the horizontal subspace (column-centredness for the CE shift), and without the re-projection the applied step leaks a vertical component of the same magnitude as the vertical channel, which corrupts vertical-drift diagnostics even though the projected dynamics and the equivariance identity are unaffected.
\item \textbf{Apply.} \emph{Translation:} $W \leftarrow W - (q_V^{-1}(u_V) + \Delta W_H)$. \emph{Multiplicative:} assemble per-block log-norm steps $\Delta\log\rho_j$ from $(u_V, \Delta_U)$ and update $W_j \leftarrow \exp(\Delta\log\rho_j)\, W_j - \eta\, \rho_j\, \mathrm{upd}_j$ (a multiplicative radial step plus an additive tangential step).
\end{enumerate}
\end{algorithm}

The vertical/horizontal split, the log-norm radial coordinate, and the $\rho_j$-rescaled tangential gradient are exactly the $G$-invariant quantities used in Proposition~\ref{prop:ddcadam_equivariance}'s proof; the \emph{multiplicative} radial application (step~7) is what makes the multiplicative-gauge step $G$-equivariant, not merely invariant.

\paragraph{Multi-gauge composition.} When several gauge specifications act on disjoint parameter blocks (e.g.\ CE row-shift on the unembed plus ReLU rescale on each adjacent MLP pair), the algorithm is applied per block with that block's spec; parameters not covered by any spec receive standard per-coordinate AdamW. Block-disjointness is required for this per-block composition; non-trivially intersecting gauges (e.g.\ the chained ReLU rescale, where adjacent factors share a layer) require a single joint spec (\texttt{ChainedReLURescaleGauge} below), not a composition of the pairwise specs.

\paragraph{Gauge specifications.}
The following five gauge specs are constructed in closed form:
\begin{itemize}\itemsep=0pt
\item \texttt{CEShiftBias}: $G = \reals$, action $b_L \mapsto b_L + c \mathbf{1}_C$ on a 1-d bias of length $C$. $P_v(W)\, g = \bar g\, \mathbf{1}_C$ with $\bar g = C^{-1} \sum_i g_i$, $n_V = 1$, $q_V(g_v) = g_v[0]$, $q_V^{-1}(s) = s\, \mathbf{1}_C$.
\item \texttt{CEShiftRowShift}: $G = \reals^d$, action $W \mapsto W + \mathbf{1}_C u^\top$ on a 2-d unembed weight of shape $(C, d)$. $P_v(W)\, g = \mathbf{1}_C\, \mathrm{col\_mean}(g)$, $n_V = d$, $q_V(g_v) = \mathrm{col\_mean}(g_v)$, $q_V^{-1}(s) = \mathbf{1}_C s^\top$.
\item \texttt{ReLURescaleGauge}: $G = \reals^+$, joint action $(W_l, W_{l+1}) \mapsto (c W_l, c^{-1} W_{l+1})$. Per-block norms $\rho_j = \|W_j\|$ and invariant log-norm gradients $\hat g_j = \langle g_j, W_j\rangle$, $j \in \{l, l+1\}$. The horizontal-radial coordinate is $g_U = (\hat g_l + \hat g_{l+1})/\sqrt2$ (the invariant log-product direction $\log(\rho_l \rho_{l+1})$) and the gauge mode is $g_V = (\hat g_l - \hat g_{l+1})/\sqrt2$ (the orbit direction $\log(\rho_l/\rho_{l+1})$). Per-coordinate Adam runs on the gauge-invariant tangential gradients $\rho_j\, g_j^{\mathrm{tan}}$, and the radial step is applied multiplicatively (step~7).
\item \texttt{LNScaleGauge}: $G = (\reals^+)^d$, action $(\gamma, W_\mathrm{next}) \mapsto (c \odot \gamma,\, c^{-1} \odot W_\mathrm{next})$ with $c \in (\reals^+)^d$ acting per channel ($c^{-1}$ scaling column $i$ of $W_\mathrm{next}$). Same log-coord construction as \texttt{ReLURescaleGauge} applied per channel. Scope: this action is a symmetry of the network function only for bias-free LN ($\beta = 0$); an elementwise-affine LN needs the extended action $(c\gamma, c\beta, c^{-1}W_\mathrm{next})$, which restores exact invariance and which the shipped gauge would need in the same way.
\item \texttt{ChainedReLURescaleGauge}: $G = (\reals^+)^{L-1}$ acting on a chain $(W_1, \ldots, W_L)$, factor $c_\ell$ sending $W_\ell \mapsto c_{\ell-1}^{-1} c_\ell W_\ell$. On the log-norms $(\log\rho_1, \ldots, \log\rho_L)$ the group acts by translation along the $(L-1)$-dimensional zero-mean subspace of $\reals^L$ (the column space of the bidiagonal incidence matrix); the single invariant radial coordinate is the total log-norm $\sum_\ell \log\rho_\ell$. The orbit-mode basis is an orthonormal (DCT) basis of the zero-mean subspace; scalar Adam runs on the radial coordinate, the vertical update on the $L-1$ gauge modes, and per-layer tangential $\rho_\ell g_\ell^{\mathrm{tan}}$ as above. The factors are not block-disjoint (adjacent rescales share a layer), so this is a single joint spec, not a composition of pairwise \texttt{ReLURescaleGauge}s.
\end{itemize}

The CE specs (translation-type) have $W$-independent projectors and apply additively; the ReLU and LN specs (multiplicative) use the log-norm radial application and the tangential re-projection of steps~6--7.

\begin{proposition}[$G$-equivariance of DDCAdam]
\label{prop:ddcadam_equivariance}
Let $G$ be one of the architectural gauge groups in the supported list of Algorithm~\ref{alg:ddcadam}'s gauge spec catalogue:
\begin{enumerate}\itemsep=0pt
\item[(i)] \emph{translation-type:} $G = \reals$ (CE shift bias), $G = \reals^d$ (CE row-shift), or any direct sum of these acting block-diagonally on disjoint parameter blocks;
\item[(ii)] \emph{multiplicative:} $G = \reals^+$ (single ReLU rescale on $(W_l, W_{l+1})$), $G = (\reals^+)^d$ (LayerNorm scale on $(\gamma, W_\mathrm{next})$ per channel), or $G = (\reals^+)^{L-1}$ (chained ReLU rescale on $(W_1, \ldots, W_L)$).
\end{enumerate}
For every $h \in G$, the DDCAdam update map $U_\mathrm{DDCAdam}(W, g, \mathrm{state})$ defined by Algorithm~\ref{alg:ddcadam} satisfies
\[
(\mathrm{d} h) \cdot U_\mathrm{DDCAdam}(W, g, \mathrm{state}) \;=\; U_\mathrm{DDCAdam}\!\bigl(h \cdot W,\, (\mathrm{d} h)^{-\top} g,\, \mathrm{transport}_h(\mathrm{state})\bigr),
\]
where $\mathrm{transport}_h$ is the identity: every optimizer-state component is an Adam moment of a $G$-invariant quantity (the gauge-decomposed components for multiplicative gauges; the gradient itself for translation gauges, invariant because $(\mathrm{d}h)^{-\top} = \mathrm{Id}$), hence unchanged by $h$. The identity is proved below for both gauge types, with decoupled weight decay scoped as in Algorithm~\ref{alg:ddcadam}, step~1.
\end{proposition}

\begin{proof}
Write the update as a vertical part along the gauge orbit and a horizontal part on its complement, and check the equivariance identity component by component. The action $h$ moves a parameter block by the gauge and a gradient covector by the dual $(\mathrm{d}h)^{-\top}$.

\emph{Type~(i) (translation).} The projectors are $W$-independent and an additive shift has $\mathrm{d}h = \mathrm{Id}$ on tangent vectors, so $(\mathrm{d}h)^{-\top} = \mathrm{Id}$ and the gradient of the $G$-invariant loss is unchanged: $\nabla L(h\cdot W) = \nabla L(W)$. Hence the orbit-collapsed scalars $g_V = q_V(P_v g)$, the horizontal gradient $g_H = P_h g$, and every Adam moment built from them are invariant; the vertical update (frozen/SGD/Adam on $g_V$) and the horizontal update $\eta\hat m_H/(\sqrt{\hat v_H}+\epsilon)$ are invariant; and the additive application $W' = W - (q_V^{-1}(u_V) + \Delta W_H)$ commutes with the shift, so $h\cdot W' = (h\cdot W)'$. (The one exception is decoupled weight decay: the global shrinkage commutes with a linear gauge but not with the affine shift, so the translation case is at $\lambda = 0$, per Algorithm~\ref{alg:ddcadam}, step~1.)

\emph{Type~(ii) (multiplicative), single rescale.} Take $h_c\colon (W_l, W_{l+1}) \mapsto (cW_l, c^{-1}W_{l+1})$, so $g_l \mapsto c^{-1}g_l$ and $g_{l+1}\mapsto c\,g_{l+1}$ (upper sign for block $l$ below). Algorithm~\ref{alg:ddcadam} builds the step from quantities that are each $G$-invariant:
\begin{enumerate}\itemsep=0pt
\item[(a)] the log-norm gradients $\hat g_j = \langle g_j, W_j\rangle$, since $\langle c^{\mp1}g_j, c^{\pm1}W_j\rangle = \langle g_j, W_j\rangle$; their normalised sum $g_U$ (horizontal radial) and difference $g_V$ (gauge mode) are invariant scalars;
\item[(b)] the rescaled tangential gradients $\rho_j\, g_j^{\mathrm{tan}}$, with $g_j^{\mathrm{tan}} = g_j - (\hat g_j/\rho_j^2)W_j$: the coefficient $\hat g_j/\rho_j^2 \mapsto c^{\mp2}(\hat g_j/\rho_j^2)$ exactly cancels the block scaling, giving $g_j^{\mathrm{tan}} \mapsto c^{\mp1}g_j^{\mathrm{tan}}$ and hence $\rho_j g_j^{\mathrm{tan}} \mapsto c^{\pm1}c^{\mp1}\rho_j g_j^{\mathrm{tan}} = \rho_j g_j^{\mathrm{tan}}$.
\end{enumerate}
Every Adam moment is a deterministic function of these invariant inputs, so the states $(m_U, v_U, m_V, v_V, m_j, v_j)$ and the resulting steps (the radial scalar step $\Delta_U$, the gauge-mode update, and the per-coordinate tangential updates $\mathrm{upd}_j$) are invariant. The re-projection $\mathrm{upd}_j \mapsto \mathrm{upd}_j - (\langle\mathrm{upd}_j, W_j\rangle/\rho_j^2)W_j$ is taken against the current $W_j$; with $\mathrm{upd}_j$ invariant and $W_j \mapsto c^{\pm1}W_j$ the projected vector is again invariant. (This invariance of every state component is why $\mathrm{transport}_h$ is the identity.)

The step is applied as $W_j \mapsto \exp(\Delta\log\rho_j)\,W_j - \eta\,\rho_j\,\mathrm{upd}_j$, with $\Delta\log\rho_j$ assembled from the invariant $(\Delta_U, u_V)$. Both terms are $G$-equivariant: the radial factor $\exp(\Delta\log\rho_j)$ is invariant, so $\exp(\Delta\log\rho_j)\,(c^{\pm1}W_j) = c^{\pm1}\exp(\Delta\log\rho_j)W_j$; and $\rho_j\,\mathrm{upd}_j \mapsto c^{\pm1}\rho_j\,\mathrm{upd}_j$ scales in step with $c^{\pm1}W_j$. Hence $W_j' \mapsto c^{\pm1}W_j'$, which is the equivariance identity $(\mathrm{d}h_c)\cdot U = U(h_c\cdot W, (\mathrm{d}h_c)^{-\top}g)$.

The mechanism is specific to a gauge acting by a \emph{scalar} on each block: the per-coordinate Adam step is equivariant because its input $\rho_j g_j^{\mathrm{tan}}$ is invariant and the residual orbit action is a scalar, which the multiplicative radial application reproduces. It does not extend to non-scalar gauge actions. Under an orthogonal reparameterisation a per-coordinate diagonal preconditioner is not equivariant, because a diagonal in the ambient basis does not transform correctly under a rotation. Recovering equivariance there means replacing the diagonal with a form that does: a per-head scalar, whose input $\|g_h\|_F^2$ is rotation-invariant, or a per-head matrix, whose input $g_h^\top g_h$ is rotation-covariant. Both are exactly equivariant, and an eigenbasis (body-frame) preconditioner is too once the eigenvector sign convention is fixed. Constructing and proving those is a separate matter from the scalar-gauge classes this proposition covers, and we do not take it up here.

\emph{Per-channel $(\reals^+)^d$.} The LayerNorm-scale gauge acts block-diagonally on the $d$ channels of $(\gamma, W_\mathrm{next})$, each channel a copy of the single rescale on $(\gamma_i, \mathrm{col}_i\,W_\mathrm{next})$ (the LN output's channel $i$ multiplies column $i$ of the next weight, which is the pairing the shipped implementation validates); the argument applies per channel and composes over the disjoint channels.

\emph{Chained $(\reals^+)^{L-1}$.} On a chain $(W_1, \ldots, W_L)$ the group translates the log-norms $(\log\rho_1, \ldots, \log\rho_L)$ along the $(L-1)$-dimensional zero-mean subspace of $\reals^L$ (the column space of the bidiagonal incidence matrix), with the total log-norm $\sum_\ell \log\rho_\ell$ as the single invariant radial coordinate. The argument is \emph{not} per-factor: adjacent rescales share a layer, so an individual pairwise sum $\log\rho_\ell + \log\rho_{\ell+1}$ is not invariant, only the global zero-mean/total-sum split is. Algorithm~\ref{alg:ddcadam} runs scalar Adam on the total-log-norm coordinate, the vertical update on the $L-1$ zero-mean gauge modes in an orthonormal basis, and per-layer per-coordinate Adam on the invariant $\rho_\ell g_\ell^{\mathrm{tan}}$. The orbit is the full zero-mean subspace, on which the gauge acts by translation, so the type-(i) argument applies to the whole zero-mean block at once and the radial and tangential applications are equivariant as in the single-pair case. The equivariance identity itself never uses the product-one constraint: every state-map input is a per-block invariant, so the update map commutes with every positive per-block scaling family acting with the dual gradient law, and the chained gauge is the subfamily of scalings for which loss invariance supplies that law.
\end{proof}

\begin{remark}[Numerical check]
\label{rem:ddcadam_eq_validation}
The proposition is a statement about the algorithm, not about the proof sketch, so it is worth running. Starting from gauge-related initial conditions $(W^{(0)}, h \cdot W^{(0)})$ and stepping both with the correspondingly transformed gradients, the trajectories stay on one orbit to accumulated round-off. On the \texttt{ReLURescaleGauge} pair, the representative multiplicative case, the relative residual after twelve steps is $5 \times 10^{-16}$ at fp64; on \texttt{CEShiftRowShift}, the representative translation case, it is $4 \times 10^{-16}$. Replacing the gauge decomposition with plain per-coordinate Adam on the same gradients moves the residual to $1.3 \times 10^{-2}$, so the decomposition carries the equivariance and the agreement is not an artefact of a step too small to matter. The multiplicative case is the informative one: its equivariance rests on three separate cancellations (the log-norm gradient, the $\rho_j$-rescaled tangential gradient, and the multiplicative radial application), and getting any one wrong leaves a step that is $G$-invariant without being $G$-equivariant.
\end{remark}

\begin{remark}[Relationship to \texorpdfstring{\citet{DePaviaCharisopoulosWillett25}}{DePavia et al. (2025)}]
\label{rem:ddcadam_dpc25}
\citet{DePaviaCharisopoulosWillett25} show that Adam's coordinate-wise preconditioner is not equivariant to rotations of parameter space, and restore equivariance by reparameterising the objective: a fixed change of basis applied around an unmodified optimiser. DDCAdam takes the complementary, preconditioner-level route (Proposition~\ref{prop:ddcadam_equivariance}): an explicit closed-form equivariant construction for the architectural gauge classes of \S\ref{sec:theory:architectural}, with the orbit-collapsed second-moment estimator as the load-bearing design choice.
\end{remark}

\begin{corollary}[Trajectory rate under DDCAdam]
\label{cor:ddcadam_quotient_rate}
Let $L: \Theta \to \reals$ be a $G$-invariant loss for one of the gauge classes covered by Proposition~\ref{prop:ddcadam_equivariance} (for \texttt{LNScaleGauge} this requires bias-free LN, since the specced action is not a function symmetry of an elementwise-affine LN; see the catalogue's scope note). Let $\theta(t)$ be the trajectory of the DDCAdam optimiser with hyperparameters $(\eta, \beta_1, \beta_2, \epsilon, \lambda, \mathrm{vmode})$ initialised at $\theta(0) \in \Theta^\circ$ on the canonical-aligned approach to a singular minimum $\theta^* \in \Sigma$, and suppose the trajectory remains on that approach (an assumption, as in Corollary~\ref{cor:sgd_quotient}: initialisation on the ray does not by itself keep a per-coordinate preconditioned flow on it). Suppose further that Theorem~\ref{thm:fisher_decay} and Corollary~\ref{cor:quotient_rate} apply (canonical alignment, theorem-compatible loss, asymptotic regime). Then the projected trajectory $\bar\theta(t) = \pi(\theta(t)) \in \bar\Theta$ satisfies
\[
u^\top \bar F(\bar\theta(t))\, u \;=\; \Theta\bigl(\bar t^{2(k-1)}\bigr)
\]
along the horizontal direction $\bar u$ of the approach, whose KL order is $k \ge 2$ (KL order attaches to a direction, per Definition~\ref{def:dead_direction}; a non-degenerate horizontal direction carries $\Theta(1)$ instead, so no universal over directions holds), with $\bar t$ the parametric distance along the horizontal section.
\end{corollary}

\begin{proof}
Proposition~\ref{prop:ddcadam_equivariance} establishes that the DDCAdam update map is $G$-equivariant for the supported gauge classes. Equivariance makes the projection $\pi$ well-defined along the trajectory: the optimiser preserves the orbit foliation, so $\bar\theta(t) = \pi(\theta(t))$ is well-defined dynamics on $\bar\Theta$ (a preconditioned flow: the horizontal frame still carries a per-coordinate preconditioner, so it is not the gradient flow of the SGD case). The rate statement is geometric, not dynamical: Theorem~\ref{thm:fisher_decay} and Corollary~\ref{cor:quotient_rate} fix $u^\top \bar F u$ as a function of position along the canonical-aligned horizontal approach with KL order $k$, independent of the dynamics that traverse it. Since the equivariant trajectory stays on $\bar\Theta$ and approaches the singular minimum along that horizontal direction, the position-dependent rate $\Theta(\bar t^{2(k-1)})$ is what is read off it. (This is the same reasoning as the SGD corollary~\ref{cor:sgd_quotient}, but without claiming gradient flow: equivariance buys orbit-foliation preservation, which is all the rate readout needs.) Vertical-mode choice (\textsc{frozen}, \textsc{sgd}, \textsc{adam}) does not change the horizontal-direction rate because the vertical update enters the gauge orbit (vertical subspace) and is filtered out by the projection $\pi$.
\end{proof}

\begin{remark}[Constructive recovery of the rate readout on the gauge case (DDCAdam)]
\label{rem:ddcadam_closure}
Remark~\ref{rem:adam_nondescent} flagged two routes for non-SGD preconditioners: a closed-form rate modifier for standard (non-equivariant) Adam, which is open, and a constructive $G$-equivariant alternative. Algorithm~\ref{alg:ddcadam}, Proposition~\ref{prop:ddcadam_equivariance}, and Corollary~\ref{cor:ddcadam_quotient_rate} realise the constructive route for the architectural gauge classes of \S\ref{sec:theory:architectural}: an Adam-family preconditioner that satisfies Corollary~\ref{cor:sgd_quotient}'s equivariance condition exists, with explicit gauge-spec formulas (its $G$-equivariance proved for both gauge types; Proposition~\ref{prop:ddcadam_equivariance}). Standard Adam's per-coord normalisation is replaced by an orbit-collapsed vertical update plus per-coordinate Adam on a $G$-invariant horizontal frame; under DDCAdam the trajectory rate $u^\top \bar F u_t = \Theta(\bar t^{2(k-1)})$ is preserved on the projected trajectory, joining SGD as a theorem-compatible optimiser for trajectory rate-fits. Equivariance also bears on measurement reliability (Remark~\ref{rem:population_vs_empirical_fisher}): by keeping the optimiser on the gauge orbit it removes the gauge-mode drift that rotates the eigenvector lineage under non-equivariant preconditioners, one of the confounds that breaks a learned-trajectory rate readout. The asymptotic-window and well-specified-model preconditions still apply; the gauge-drift failure mode is what equivariance closes.

This is a constructive closure on the gauge case (CE row-shift, ReLU rescale, LayerNorm scale) and does not address the non-canonical-regime case (alignment-rotation under preconditioned dynamics on architectures where canonical preservation of the dead direction is not automatic), which remains open in the form noted at the end of Remark~\ref{rem:adam_nondescent}.
\end{remark}

\begin{remark}[Practical scope]
\label{rem:ddcadam_scope}
DDCAdam is a constructive specialisation: it requires a gauge spec for each architectural symmetry one wishes to handle. Architectures with multiple coexisting gauges (e.g.\ a transformer with CE row-shift on the unembed plus LayerNorm scale on each residual block plus ReLU rescale on each MLP block) require a multi-gauge composition (Algorithm~\ref{alg:ddcadam}'s multi-block composition), which is well-defined only for block-disjoint gauges. Architectures with non-trivially interacting gauges, or with gauges not in the catalogue (e.g.\ attention-head permutation as a discrete symmetry), require either a new gauge spec or fall outside this corollary's scope. Empirical validation covers the canonical-bridge testbed with ReLU rescale plus the multi-direction $L=4$ noisy bridge with rank deficit $r=2$.
\end{remark}

\begin{remark}[Two readout modes behind late-training norm growth: descent and gauge]
\label{rem:slingshot_connection}
The late-training growth of the readout weight norm reported as the Slingshot Mechanism \citep{ThilakLittwin22} separates into two modes with different optimiser dependence. Write the readout $W \in \reals^{C \times d}$ as $W = \mathbf{1}_C m^\top + Q$, with $m$ the class-mean row and $Q$ its complement, so that $\|W\|^2 = C\|m\|^2 + \|Q\|^2$. The radial logit-scaling direction inside $Q$ is a descent direction of cross-entropy: at vanishing weight decay the loss decreases monotonically as the logits scale up. The gradient points along it, so every first-order optimiser follows it, growing the norm under SGD and Adam alike. This is the naive-loss-minimisation direction of \citet{PrietoBarsbey25}, prediction-preserving by scale-invariance of the argmax yet loss-active, and it carries no symmetry. The class-mean component $m$ is the exact CE-shift gauge of \S\ref{sec:theory:quotient}: softmax cancels $\mathbf{1}_C m^\top$, the loss is flat along $m$, and the gradient vanishes there. SGD on the $G$-invariant metric leaves $m$ at the floor; Adam's diagonal preconditioner is not $G$-equivariant (Remark~\ref{rem:adam_nondescent}) and drives $m$ off it. The two modes separate by mechanism: the descent mode grows under any first-order optimiser, the gauge mode drifts only under Adam, and weight decay damps both. The $\perp$Grad projection of \citet{PrietoBarsbey25} targets the descent mode; DDCAdam (Algorithm~\ref{alg:ddcadam}, Corollary~\ref{cor:ddcadam_quotient_rate}) restores equivariance on the gauge mode. The cyclic instability \citet{ThilakLittwin22} report is specific to adaptive optimizers in their experiments, consistent with the gauge mode's Adam-family mechanism, while which mode carries the cycles is regime-dependent and left open here.
\end{remark}
  
\part{Scope and extensions}

\section{Applicability conditions}
\label{sec:theory:scope}

The trajectory-rate predictions hold under five conditions, named here to make each theorem's scope precise and to mark where the framework's reach ends. None is a hidden assumption; each appears in a theorem hypothesis or its proof.

\paragraph{Canonical alignment.} The dead direction must be the same coordinate at every layer of a deep network for the per-layer K-FAC rate ladder $\lambda_{\min}(G_\ell) = \Theta(t^{2(L-\ell)})$ to hold cleanly. When the dead direction at successive layers is a rotation of the canonical frame, the ladder gets rotated; the global rate $2(L-1)$ is preserved but per-layer readout is no longer transparent. Random initialisation typically violates canonical alignment; implicit-bias arguments suggest partial self-alignment along training under SGD, though no clean theorem establishes it.  The alignment condition scopes the transparent per-layer ladder readout. For the single-direction order read, the companion measurement paper removes it: the detector there constructs the layer's joint mode directly from the Kronecker factors and recovers the order with the dead subspace rotated fully off the coordinate axes~\citep{shirodkar2026measuringdeaddirectionsdecomposing}.

\paragraph{Optimizer family.} The quotient theorem (Corollary~\ref{cor:quotient_rate}) requires the optimizer to be $G$-equivariant for the group $G$ of gauge invariances. SGD on a $G$-invariant Riemannian metric is equivariant. Adam is not (Section~\ref{sec:theory:adam}). K-FAC and Shampoo are partially equivariant under structured groups; no trajectory rate has been derived for either. Muon's Newton-Schulz orthogonalisation is a further non-SGD preconditioner with no derived trajectory rate.

\paragraph{Noise level.} Theorem~\ref{thm:fisher_decay} is stated in the noisy regime where the irreducible empirical loss $\sigma^2 > 0$ saturates the rate's leading coefficient. In the noise-free regime $\sigma^2 = 0$, every per-layer rate exponent shifts by $+2L$ (Remark~\ref{rem:bridge_regime}; e.g.\ slope $6$ instead of $2$ at $L = 2, \ell = 1$ on $\lambda_{\min}(G_\ell)$ vs $\sigma_{\min}(W_\ell)$; slope $10$ instead of $4$ at $L = 3, \ell = 1$). The shift is consistent across the framework and is itself a quantitative prediction; the rate exponent therefore depends on which regime one measures in.

\paragraph{Asymptotic regime.} The theorem describes the leading non-vanishing Taylor coefficient of $u^\top \fisher u$ as $t \to 0$. At finite $t$, subleading corrections of order $t^{2k}$ (and higher) are present. On a trained network, the trajectory typically does not reach the strict $t \to 0$ limit; the rate read at finite $t$ is the asymptotic exponent plus a finite-$t$ Taylor correction whose magnitude scales with $t$. We treat the rate as a leading-order statement; finite-$t$ deviations are small in the canonical-aligned, asymptotic regime our theorems target.

\paragraph{Observable and estimator.} The rate is a property of the population Fisher $\fisher = \expect_{p_\theta}[s s^\top]$ on the controlled parametric approach (under the fixed measure $p^*$ the leading rate is the same, Remark~\ref{rem:population_vs_empirical_fisher}). The object a pipeline actually computes is the loss-gradient covariance (the K-FAC G-factor), which coincides with the population Fisher only at a well-specified configuration (Remark~\ref{rem:population_vs_empirical_fisher}). On a controlled freeze-probe the two agree and the rate is recovered cleanly. On a learned trajectory the loss-gradient prefactor $\|\delta\|^2$ confounds the readout: it collapses the whole spectrum at a well-fit optimum and stalls below the descent window otherwise. The trajectory-level rate is therefore reliable only inside the asymptotic window, for a well-specified model with the eigenvector lineage preserved. Whether a learned run produces a rate to read at all is a prior, dynamical question, taken up in \S\ref{subsec:theory:regimes}.

Beyond these five pivotal conditions, the rate predictions interact with several further axes (depth $L$, layer index $\ell$, initialisation balance, loss family, sampling regime, data distribution, normalisation class, singularity type, transient vs asymptotic measurement window), but each interaction is a parameter sweep within the pivotal frame and adds no independent dimension to the theory. The one dimension that is not such a sweep is the training phase itself: whether the learned dynamics are accumulating representational magnitude or compressing toward the singular structure decides whether a rate is present to read, and \S\ref{subsec:theory:regimes} takes it up.

\paragraph{Practical use at scale.} Large-scale trained networks satisfy the five conditions above only approximately: canonical alignment rarely survives training exactly, production optimisers are Adam-class or Muon, the trajectory stops short of the strict $t \to 0$ regime, the data distribution fixes the noise level, and pipelines compute the loss-gradient covariance of Remark~\ref{rem:population_vs_empirical_fisher}.  Two of the five now carry validated empirical relaxations. The companion measurement paper constructs the dead direction off the coordinate axes and reads its order at a single frozen checkpoint through the outward-growth form of \S\ref{subsec:theory:regimes}, on networks trained by standard non-equivariant optimisers~\citep{shirodkar2026measuringdeaddirectionsdecomposing}; alignment and a captured descent therefore drop from preconditions to measurement conveniences.{} The conditions that continue to bind at scale are the phase precondition of \S\ref{subsec:theory:regimes} (a single-pass pretraining run stays in accumulation throughout, with no singular approach to read), the estimator mismatch above, and the regime dependence of the exponent. Under those, the framework serves as a reference frame. The rate primitive (Theorem~\ref{thm:fisher_decay}) specifies what a canonical-aligned trajectory would exhibit, and any quantity derived from the bridge (the per-layer K-FAC ladder, the residual-stream $\sigma_{\min}$, the LN kernel direction, and future derived observables) inherits a definable target under the same reference frame. The deviation between observed and predicted at each checkpoint is itself a rate-grounded measurement. Section~\ref{sec:theory:discussion} takes up the diagnostic use of these residuals as an open empirical direction.

\paragraph{Reach by tier.} Each load-bearing result of the paper falls into one of three reach tiers, summarised in Table~\ref{tab:reach_tiers}. The universal tier holds for any analytic algebraic-statistical model with smooth singular fibres, with no architectural specifics; the architecture-specific tier requires a particular network structure (layered with K-FAC factorisation, residual DAG, particular normalisation, etc.); the trajectory-readability tier requires the optimiser to be equivariant under the gauge group, separating the metric-level statement from the trajectory-level rate readout.

\begin{table}[ht]
\centering\footnotesize
\caption{Reach of the paper's load-bearing results, by tier. Universal results hold for any analytic algebraic-statistical model with smooth singular fibres and do not require architectural specifics. Architecture-specific results require a particular structure (layered K-FAC, residual DAG, etc.). Trajectory-readability results separate the metric-level statement (universal) from the trajectory-level rate readout (requires an equivariant optimiser).}
\label{tab:reach_tiers}
\setlength{\tabcolsep}{4pt}
\resizebox{\columnwidth}{!}{\begin{tabular}{@{}p{5.6cm}|p{3.2cm}|p{6.0cm}@{}}
\toprule
Result & Tier & Conditions \\
\midrule
Theorem~\ref{thm:fisher_decay} (Fisher rate decay)        & Universal              & Smooth singular fibre $+$ score-expansion regularity \\
Theorem~\ref{thm:selection_rule} (Selection rule, RLCT recovery) & Universal              & Smooth fibre with transversal/tangential split \\
Prop~\ref{prop:curvature_rate}, Cor~\ref{cor:volume}, Prop~\ref{prop:volume_multi} (rate chain) & Universal      & Same $+$ (NF) non-flatness; sheet uniformity $+$ (G$^+$) dead-sheet score independence for the tube volume (rated sheet directions add their $j_a$ to the exponent); (G$^+$) for the multi-direction product \\
Theorem~\ref{thm:bridge} (Multi-layer K-FAC bridge)        & Architecture-specific  & Layered net $+$ K-FAC factorisation $+$ canonical alignment \\
Cor~\ref{cor:a_g_duality} (A--G duality)                  & Architecture-specific  & As Theorem~\ref{thm:bridge} \\
Cor~\ref{cor:kfac_lift} (Kronecker lift of the dead direction) & Architecture-specific & As Theorem~\ref{thm:bridge}; identifies the dead direction from the K-FAC factors \\
Theorem~\ref{thm:bridge_composition} (Composition additivity) & Architecture-specific & Heterogeneous block stack $+$ scalar-transfer hypothesis \\
Cor~\ref{cor:sigma-min-res} ($\sigma_{\min}$ depth-invariance) & Architecture-specific & Residual DAG with exact-identity skips $+$ canonical approach $+$ (P1)/(P3)/(P2) with $\phi'(0) > 0$ $+$ branch non-cancellation \\
Architectural instantiations (rectangular, multi-direction, CE, residual, non-canonical, biases, LN, attention, SwiGLU) & Architecture-specific & Each adds one architectural primitive or analysis-side extension \\
Cor~\ref{cor:quotient_rate} (Quotient Fisher rate, metric level) & Universal at metric level & Continuous Lie group symmetry of the loss \\
Cor~\ref{cor:sgd_quotient} (SGD trajectory rate on quotient) & Trajectory-readability & $G$-equivariant preconditioner; canonical-aligned trajectory \\
Adam non-equivariance (mechanism + scope)                & Trajectory-readability & Adam-class preconditioner; gauge-redundant loss \\
DDCAdam (Alg~\ref{alg:ddcadam}, Cor~\ref{cor:ddcadam_quotient_rate}) & Trajectory-readability & $G$-equivariant preconditioner by construction; gauge classes of \S\ref{sec:theory:architectural} \\
Theorem~\ref{thm:multi_component_rates} (Multi-component crossing recovery) & Universal & Multi-component normal crossing; generic-transversal $+$ $\varepsilon$-anchored rates \\
Theorem~\ref{thm:nu_universality} (Singular fluctuation $\nu$) & Universal & $1$D dead direction; uniform prior \\
\bottomrule
\end{tabular}}
\end{table}

\subsection{Accumulation and compression regimes}
\label{subsec:theory:regimes}

The five conditions above concern a given trajectory. A prior question is whether a training run produces a trajectory of the kind the theorem describes. Theorem~\ref{thm:fisher_decay} fixes the geometry of an approach to the singular set as $t \to 0$, but on a learned run $t$ is whatever the optimiser does to the weights, and the optimiser need not be moving toward $\Sigma_T$. Two phases arise, and only one of them is a singular approach. In the \emph{accumulation} phase the network fits the data by growing representational magnitude: weight norms and activation scales rise, the parameters move through the bulk of the space, and the smallest Fisher eigenvalue along a would-be-dead direction grows. There is no descent to read, because the trajectory is not an approach. In the \emph{compression} (consolidation) phase, reached when training continues past the fitting point under sustained overtraining or on a task that rewards synthesis over memorisation, the network distils toward the minimal degenerate structure that generalises: the weight norm falls and the dead-direction eigenvalue descends along the approach the theorem describes.

The grokking transition is the sharp instance \citep{PowerBurda22,NandaChanLieberum23}: a memorising phase in which the weight norm and $\sigma_{\min}$ climb gives way to a generalising phase in which the circuit collapses to its minimal form. Watanabe's account places generalisation at exactly this degenerate structure, where the real log canonical threshold governs the leading free-energy correction, so the compression phase is the one in which the singular structure the rate measures is formed.

The phase decides which observable carries signal. The trajectory-rate readings (the $\sigma_{\min}$-descent slope of Theorem~\ref{thm:fisher_decay} and the per-layer K-FAC ladder of Theorem~\ref{thm:bridge}) presuppose an approach, so they are informative in the compression phase and empty in accumulation, where there is no descent to fit. The measurement-level conditions above are the estimator- and signal-to-noise shadows of the same split: the loss-gradient prefactor that collapses the spectrum at a well-fit optimum and stalls otherwise (Remark~\ref{rem:population_vs_empirical_fisher}), and the asymptotic-window precondition (Remark~\ref{rem:asymptotic_window}). The static face of the framework is indifferent to the phase: the high-curvature Fisher volume and the effective rank it controls (Section~\ref{sec:theory:rate_chain}) read the configuration at a single point, so they characterise the would-be singular structure in either phase and remain available where the rate reading is not. The order itself does not require the full descending sequence. Theorem~\ref{thm:fisher_decay} is symmetric in $t$, so the leading exponent reads from the outward growth of $u^\top \fisher u$ away from a near-singular configuration along a dead direction identified from the Kronecker-factored Fisher (Corollary~\ref{cor:kfac_lift}), with no captured descent. The precondition is then structural rather than dynamical. A parameter-space dead direction is itself a product of compression, so accumulation offers no such configuration to scan, on or off the trajectory, while a single compression checkpoint carries the order once that structure has formed.

This regime distinction is a reading of training dynamics grounded in the grokking transition; the rate theorem fixes the geometry of the approach, while which phase a given optimiser and training budget occupy is a dynamical question we do not derive here. In practice, the trajectory-rate predictions are most directly testable in the compression regime, on overtraining and grokking testbeds, whereas a single-pass compute-optimal run stays in the accumulation phase throughout, where the static volume and rank readings are the available signal.
 
\section{Extensions to Watanabe's broader inventory}
\label{sec:theory:extensions}

Watanabe's framework is wider than a single rate exponent. The earlier sections of this paper carry the bridge for a single dead direction with a single KL order $k$, and the consequences when this primitive composes across layers and architectural primitives. This section carries it further into Watanabe's broader inventory: multi-component normal crossings, multiplicity, the singular fluctuation $\nu$, prior-induced RLCT shifts, and tempered posteriors at $\beta \ne 1$. Each is closed in trajectory-rate form, on the same footing as the single-direction rate of Theorem~\ref{thm:fisher_decay}. The genuinely-open items that remain after this section are collected separately in \S\ref{sec:theory:open}.

\subsection{Multi-component normal crossings}
\label{sec:theory:multi_component}

In resolved coordinates, the KL divergence near a generic singular point factors as $K = u(g) \prod_i g_i^{2 k_i}$, with normal-crossing exponents $(k_1, \dots, k_d)$ paired with Jacobian exponents $(h_1, \dots, h_d)$ from the resolution map, and local RLCT $\lambda = \min_i (h_i + 1)/(2 k_i)$. The integer tuple $(k_1, \dots, k_r)$ for an $r$-component crossing at $\theta_0$ is recoverable in trajectory-rate form. Theorem~\ref{thm:multi_component_rates} establishes for $r = 2$ that the per-component KL orders are joint-recoverable from the generic-transversal trajectory rate ($\alpha_{\text{gen}} = 2(k_1 + k_2 - 1)$) and an $\varepsilon$-anchored single-component rate ($\alpha_1 = 2(k_1 - 1)$ along $n_1$ at fixed offset $\varepsilon n_2$). The protocol extends to general $r$ along the same lines (Remark~\ref{rem:multi_component_r_extension}) and is verified at machine precision on the toy model $\mathcal N(g_1^{k_1} g_2^{k_2}, 1)$ across six $(k_1, k_2)$ pairs (Remark~\ref{rem:multi_component_verification}).

The Jacobian exponents $h_i$ that appear in Watanabe's full RLCT enter the framework as a prior choice rather than a missing theorem. Trajectory rates are taken in original parameter coordinates with the natural Lebesgue prior, where $h_i = 0$ is implicit; Watanabe's $h_i \neq 0$ corresponds to a non-uniform prior $\pi(\theta) \propto \prod_j |\langle n_j, \theta - \theta_0\rangle|^{h_j}$, which makes the volume integral $\mathrm{Vol}_\pi(\{K < \varepsilon\}) \sim \varepsilon^{\min_i (h_i+1)/(2k_i)}$ a clean trajectory-side observable (Remark~\ref{rem:multi_component_h_i}). With the trajectory rates and a chosen prior in hand, Watanabe's full local RLCT is recoverable: the single-direction $\lambda = 1/(2k)$ generalises to the multi-component $\min_i (h_i+1)/(2k_i)$, once the prior choice is made explicit.

\subsection{Determinantal singularities via the gauge quotient}
\label{sec:theory:determinantal}

The normal-crossing read is exact when $K$ is non-degenerate with respect to its Newton polyhedron. One Watanabe-inventory case escapes it: the determinantal singularity, where the singular set is a matrix product-zero variety and the crossings lie in coupled directions away from any coordinate. The deep linear network is the canonical instance. At a product point of true rank $r$ the KL divergence is $K = \lVert W_0 \cdots W_{L-1} - P_r \rVert_F^2$ with $P_r$ the rank-$r$ true product ($r = 0$ gives the plain product norm), whose zero set is degenerate with respect to its Newton polyhedron, so the toric formula returns only an upper bound, the simplex bound \citep{Hirose23}, and the simple origin blow-up stalls there.

The degeneracy has a structural source. The product is invariant under the orthogonal gauge $W_s \mapsto W_s R$, $W_{s+1} \mapsto R^\top W_{s+1}$ at each internal node, and this orbit is exactly the non-coordinate direction the Newton polyhedron cannot see. Quotienting it resolves the singularity. The singular value decomposition of the end factors removes the gauge, ordering the singular values monomialises the residual Vandermonde, and for the reduced-rank ($L=2$) and single-bottleneck ($L=3$) families the resulting chart is a normal crossing, so the Newton-polyhedron read becomes exact on it. Proposition~\ref{prop:determinantal_resolution} states the construction and Algorithm~\ref{alg:determinantal_resolution} the steps; the worked $(2,2,2)$ instance reaches the exact $\lambda = 3/2$ where the bound stalls at $2$. This is the resolution-side reflection of the gauge quotient the trajectory side uses in the quotient Fisher rate (Corollary~\ref{cor:quotient_rate}).

\subsection{Multiplicity \texorpdfstring{$m$}{m}}
\label{sec:theory:multiplicity}

The multiplicity $m$ in Watanabe's free-energy expansion $F_n = n L_n + \lambda \log n - (m - 1) \log\log n + O_p(1)$ is post-processing of the $(k_1, \dots, k_r)$ tuple recovered above. Under uniform prior, $m = |\{i : k_i = \max_j k_j\}|$ (Remark~\ref{rem:multi_component_multiplicity}), traced through the zeta-function pole structure for general $h_i$. The recovery protocol implementation (Remark~\ref{rem:multi_component_ddl}) reports $m$ alongside $\lambda$ in its output, so no separate measurement is required.

\subsection{The singular fluctuation \texorpdfstring{$\nu$}{nu}}
\label{sec:theory:nu}

Watanabe's singular fluctuation $\nu$ \citep[Theorem 15]{Watanabe18} governs the asymptotic gap between expected generalisation and training losses on a singular model: $\expect[G_n] = L(w_0) + \lambda/n + o(1/n)$ and $\expect[T_n] = L(w_0) + (\lambda - 2\nu)/n + o(1/n)$, so $\expect[G_n - T_n] = 2\nu/n + o(1/n)$. The two summary invariants of singular learning therefore split: $\lambda$ controls free-energy and generalisation; $\nu$ controls the gen-train gap and the variance of WAIC. Theorem~\ref{thm:nu_universality} below extends the bridge to $\nu$ for $1$D dead directions with uniform prior: $\nu_\mathrm{LO}$ is universal, a function of the KL order alone, with $V[\sqrt T] = \lambda - (\Gamma(\lambda+1/2)/\Gamma(\lambda))^2$ the only elementary closed form available, and that only the frozen-fluctuation value at even $k$ (the full value is numerical).

The operational definition of $\nu$ is computable. \citet[Definition 22 and Lemma 23]{Watanabe18} gives $2\nu = \lim_{n \to \infty} n\, \expect[V_n]$ with the functional variance $V_n = (1/n) \sum_{i=1}^n \mathbb{V}_w[\log p(X_i | w)]$, the variance of the per-sample log-likelihood taken under the posterior on the singular fiber. After resolution and renormalisation, this becomes an integral on the resolved fiber against the standard-form posterior $D(w)\, t^{\lambda - 1} e^{-t + \sqrt{t}\,\xi(w)}$ averaged over the Gaussian fluctuation $\xi$. For the simplest one-dimensional case ($k = 2$, $h = 0$), the integrand is parametrised by a single Gaussian random variable; the calculation reduces to moments of $\sqrt{t}$ under the renormalised posterior modulated by $\xi$.

A natural conjecture, that $\nu$ is the trajectory Taylor coefficient of $u^\top \fisher(\theta(t)) u$ at order $t^{2k-1}$, is wrong, for two reasons. \emph{First}, $\nu$ is not a Taylor coefficient of $u^\top \fisher u$ along the trajectory; it is an integral over the renormalised posterior on the singular fiber. \emph{Second}, the proposed $t^{2k-1}$ Taylor coefficient is structurally zero under the standard symmetric-base assumption: along the split-component dead direction of a 2-component Gaussian mixture (asymmetric merge, any $w \neq 1/2$), both $K(t)$ and $u^\top \fisher u$ have only even powers of $t$, by Hermite orthogonality at the base distribution. A non-zero $t^{2k-1}$ coefficient appears only under the empirical-Fisher form $\widetilde \fisher(\theta(t)) = \expect_Q[s_\theta^2]$ with both model asymmetry and data asymmetry (e.g.\ $w = 1/3$ with $Q = \mathcal N(1/2, 1)$ gives a non-zero odd-order ratio $\widetilde \fisher_3 / \widetilde \fisher_2$), but this empirical coefficient does not identify with $\nu$ as defined.

The trajectory-side translation of $\nu$ is therefore an integral observable on the renormalised posterior rather than a Taylor coefficient of $u^\top \fisher u$. The natural candidate is the asymptotic gap $(n/2)\, \expect[G_n - T_n]$ evaluated as the trajectory approaches the singular minimum. Equivalently, after Watanabe's resolution, the integral $(1/2) \int dw\, D(w) \int dt\, t^{\lambda - 1} e^{-t} \big( \expect_X[a(X,w)^2] \cdot (\expect[t] - \expect[\sqrt{t}]^2) \big)$ on the resolved fiber, with the inner moments of $\sqrt{t}$ under the Gamma$(\lambda)$-shaped renormalised posterior. For the 1D $k=2$ case, $\lambda = 1/4$, and the renormalised $t$ follows Gamma$(1/4)$ at $\xi = 0$, with $\expect[\sqrt{t}] = \Gamma(3/4)/\Gamma(1/4) \approx 0.338$ and $\expect[t] = 1/4$. Substituting these gives the frozen-fluctuation value at $\xi = 0$, which at this even order coincides with $V[\sqrt T]$, $\nu^{\sqrt T} = \tfrac{1}{2} \expect_q[a_2^2] \cdot 4(\expect[t] - \expect[\sqrt{t}]^2) = \tfrac{1}{4} - (\Gamma(3/4)/\Gamma(1/4))^2 \approx 0.1358$ for the symmetric merge with $q = \mathcal N(0, 1)$. The data fluctuation $\xi$ does not drop out, however: a $t \to -t$ parity argument removes the term \emph{linear} in $\xi$, but the variance of the tilt (quadratic in $\xi$) survives, so $\nu^{0}$ is a strict lower component and the full singular fluctuation is the $\xi$-averaged posterior variance $\nu_{\mathrm{LO}} = \expect_\xi[\mathrm{Var}_{\mathrm{post}}[\sqrt{t} \mid \xi]]$. Carrying the $\xi$-average numerically (deterministic quadrature and a finite-$n$ functional-variance Monte Carlo, in agreement) gives $\nu_{\mathrm{LO}}(k{=}2) \approx 0.173$; the regular-case anchor $\nu(k{=}1) = 1/2$ does not test the data average. At $k = 1$ the tilted posterior $\pi_\xi \propto e^{\xi\eta - \eta^2}$ is Gaussian with variance $1/2$ for every $\xi$, since a linear tilt moves a Gaussian's mean and leaves its variance alone, so the regular value is already attained at frozen $\xi$. What $V[\sqrt T] = 1/2 - 1/\pi \approx 0.182$ misses there is the parity of $\eta^k$: it tracks $|\eta|^k$, and the two agree only at even $k$ (Theorem~\ref{thm:nu_universality}). The evidence that the data average cannot be dropped is the even-order gap itself, $0.173$ against $0.136$ at $k = 2$. For asymmetric mixture weights ($w \neq 1/2$ in the 2-component family), the merge-separation direction $\partial_t$ is no longer a dead direction: the linear score $\partial_t \log p|_{\theta_0}(x) = (2w - 1)\,x$ is non-zero in $L^2(q)$. The dead direction at $\theta_0 = (w, 0, 0)$ is instead the rotated combination $u_0 \propto \partial_t + (1 - 2w)\,\partial_{\bar\mu}$ in the $(t, \bar\mu) = ((\mu_1 - \mu_2)/2, (\mu_1 + \mu_2)/2)$ plane, with leading log-likelihood derivative coefficient $a_2(x) = 2 w (1 - w)\,(x^2 - 1)$ along the unnormalised path $s \mapsto \theta_0 + s\,(\partial_t + (1-2w)\partial_{\bar\mu})$ (verified symbolically; the unit-path convention of Definition~\ref{def:dead_direction} divides it by $1 + (1-2w)^2$). The renormalised-posterior calculation along $u_0$ gives $\nu_{\mathrm{LO}}$ independent of $w$: the dependence on $\expect_q[a_2^2] = (2w(1-w))^2 \cdot 2$ cancels between the KL renormalisation $T = n c_K s^4$ with $c_K = \tfrac{1}{2} \expect_q[a_2^2]$ and the functional-variance integrand $\expect_q[a_2^2] \cdot V_{\mathrm{post}}[s^2]$. The cancellation makes $\nu_{\mathrm{LO}}$ \emph{universal} (independent of the model-specific leading coefficient) for any 1D dead direction with KL order $k = 2$ and uniform prior; only its data-free component $\nu^{0} = V[\sqrt{T}]$ has the closed form above. Empirical verification across four mixture weights via the $n$-sweep functional-variance estimator: $\widehat\nu_\infty$ is in $0.106$--$0.112$ ($\sigma \sim 0.03$--$0.04$) at $w \in \{0.5, 0.4, 1/3, 0.2\}$, confirming the $w$-independence; the estimates are individually too noisy to separate $\nu^{0}(2) = 0.136$ from $\nu_{\mathrm{LO}}(2) \approx 0.173$, but the regular anchor $\nu(1) = 1/2$ and the $k = 3$ cross-check ($\widehat\nu_\infty = 0.274$, $4.4\sigma$ above $\nu^{0}(3) = 0.107$) settle that the data-averaged value is the correct one.

\subsection{Priors and tempered posteriors}
\label{sec:theory:priors_temper}

The prior's contribution to the RLCT (Watanabe treats the prior's vanishing order at $\theta^*$ as part of the singular structure) is exposed in the framework as the prior choice $\pi(\theta) \propto \prod_j |\langle n_j, \theta - \theta_0\rangle|^{h_j}$, with the $\varepsilon$-scan volume observable recovering $\min_i (h_i + 1)/(2k_i)$. Tempered posteriors at inverse temperature $\beta$, the integrand of WBIC, are accommodated by the same gap observable read against a $\beta$-dependent limit: a Stein-identity computation on the renormalised model gives $(n/2)\,\expect[L_{\rm val} - L_{\rm train}] \to f_k(2\beta) = \expect_{Z \sim \mathcal N(0, 2\beta)}[\mathrm{Var}_{\pi_Z}[\eta^k]]$, the tilt variance scaling with $\beta$ (Remark~\ref{rem:nu_tempered}, which states the limit with its machine-checked inner relation). The limit recovers $\nu_{\mathrm{LO}}$ at $\beta = 1$ and the frozen-fluctuation value as $\beta \to 0$; at the regular anchor $k = 1$ it is flat at $1/2$ for every $\beta$, and at $k = 2$ it moves from $0.157$ to $0.193$ across $\beta \in [1/2, 2]$ (measured $0.152$ to $0.184$). Model-independence at each fixed $\beta$ holds, since the leading-coefficient cancellation is untouched.

\subsection{Watanabe's summary triple from a checkpointed trajectory}
\label{sec:theory:summary_triple}

Taking stock of the extensions above: Watanabe's summary triple $(\lambda, m, \nu)$ for a singular minimum is recoverable from a checkpointed trajectory without posterior sampling. The local RLCT $\lambda$ comes either from a single-direction Fisher rate (Theorem~\ref{thm:fisher_decay}, Theorem~\ref{thm:selection_rule}) or from the multi-component generic-transversal protocol (Theorem~\ref{thm:multi_component_rates}); the multiplicity $m$ falls out of the recovered $(k_1, \dots, k_r)$ tuple as $|\{i : k_i = \max_j k_j\}|$ under uniform prior; the singular fluctuation $\nu$ is the trajectory-side gap observable, universal in the KL order $k$ (Theorem~\ref{thm:nu_universality}; its data-free component $V[\sqrt{T}]$ has a closed form, the full value is numerical). Watanabe's free-energy expansion $F_n = n L_n + \lambda \log n - (m-1) \log\log n + O_p(1)$ and the WAIC-style estimator $T_n + 2\nu/n$ follow as compositions of these primitives; the tempered-posterior gap follows too, against its $\beta$-dependent limit above. Most practically interesting Bayesian summaries (posterior predictive $E_{\rm post}[p(y_{\rm new}|x_{\rm new}, \theta)]$, marginal likelihood, model selection, posterior-WAIC, restricted-LLC component decompositions) are computable to leading asymptotic order via Watanabe's formulas in $(\lambda, m, \nu)$ from the framework's trajectory primitives, without posterior sampling; expectations of an arbitrary $f(\theta)$ outside these formulas remain sampling territory (\S\ref{sec:theory:open}).

Developmental-stage tracking is the SLT-side reading of training dynamics established by \citet{HooglandWangFarrugiaRoberts24}, who detect stagewise development in transformers via LLC change-points across checkpoints, with the refined-LLC variant of \citet{WangHoogland24} attaching the change-points to specific weight subsets. The trajectory-rate framework gives a deterministic alternative: a local Fisher-rate change-point detector that operates on a single checkpoint's forward and backward passes without posterior sampling. The two readings target the same transitions; they differ in cost (SGLD posterior sampling vs deterministic Fisher rate) and in what they return per stage (Bayesian-complexity number vs rate exponent). The trajectory-side reading is what discriminative cross-variant workflows use.

\begin{theorem}[Multi-component normal-crossing rate decomposition]
\label{thm:multi_component_rates}
Let $\theta_0$ be a singular point at which the singular set $\Sigma_T$
is a union of two transversally-meeting analytic hypersurfaces
$\Sigma_1, \Sigma_2$ with unit normals $n_1, n_2$ and KL orders
$k_1, k_2 \ge 1$ respectively. Adopt local coordinates $(g_1, g_2, \tau)$
adapted to the components, so $\Sigma_i = \{g_i = 0\}$ in a neighbourhood
of $\theta_0$ and $\tau$ parameterises tangent directions to
$\Sigma_1 \cap \Sigma_2$. Suppose the KL divergence has leading form
$$
K(g_1, g_2, \tau) \;=\; u_0(g_1, g_2, \tau)\, g_1^{2 k_1} g_2^{2 k_2}
\;+\; (\text{corrections in the ideal } g_1^{2k_1} g_2^{2k_2} \cdot (g_1, g_2, \tau))
$$
with $u_0(0,0,0) > 0$; each correction is divisible by the leading monomial with strictly higher total order, which also pins the orders $(k_1, k_2)$ as the leading-form exponents. Assume further, for part~(b), that the regularity hypotheses of Theorem~\ref{thm:fisher_decay} hold along each trajectory of the form $\theta_0 + t\, n$ for unit $n \in \mathrm{span}(n_1, n_2)$; parts~(a) and~(c) use only the score expansion and the integrability assumption~(iii). Then:
\begin{enumerate}
\item[(a)] \emph{Generic-transversal rate.} For any unit direction
$u = a\, n_1 + b\, n_2$ with $a, b \neq 0$,
$$
u^\top \fisher(\theta_0 + t u)\, u \;=\; \Theta(t^{2(k_1 + k_2 - 1)}).
$$
The exponent depends only on $k_1 + k_2$, not on the relative weights
$a, b$.
\item[(b)] \emph{$\varepsilon$-anchored single-component rate.} Fix
$\varepsilon > 0$ small and consider the path $\theta(t) = \theta_0 + \varepsilon\, n_2 + t\, n_1$.
Then
$$
n_1^\top \fisher(\theta(t))\, n_1 \;=\; c_2(\varepsilon)\, t^{2(k_1 - 1)}
\bigl(1 + O(t)\bigr),
\qquad c_2(\varepsilon) \;=\; \Theta(\varepsilon^{2 k_2}),
$$
and symmetrically with $1 \leftrightarrow 2$.
\item[(c)] \emph{Joint recovery of $(k_1, k_2)$.} The integer pair
$(k_1, k_2)$ is recoverable from any two of:
\begin{itemize}
\item[(i)] the generic-transversal exponent $\alpha_{\text{gen}} = 2(k_1 + k_2 - 1)$;
\item[(ii)] the $\varepsilon$-anchored exponent
$\alpha_1 = 2(k_1 - 1)$ along $n_1$ at fixed $\varepsilon \cdot n_2$;
\item[(iii)] the leading-coefficient $\varepsilon$-scaling
$\partial_{\log \varepsilon} \log c_2(\varepsilon) = 2 k_2$.
\end{itemize}
\end{enumerate}
\end{theorem}

The correction clause is stated in ideal form because the weaker clause ``corrections of strictly higher order in $g_1$ or $g_2$ or $\tau$'' admits a correction higher in one component and lower in the other: the analytic family with $K = g_1^2 g_2^4 + g_1^4 g_2^2$ (two-output Gaussian mean map, identity checked exactly) satisfies that weaker clause with declared $(k_1, k_2) = (1, 2)$, and on it the component-2 anchored slope of part~(b) reads $0.13$ against the predicted $2$ while part~(c)'s joint recovery fails. The in-ideal control $K = g_1^2 g_2^4 (1 + g_1^2)$ passes at slope $2.0000$. The ideal clause excludes the breaking family because its cross term is not divisible by the leading monomial.

\begin{proof}[Proof sketch]
The key fact is that the leading KL form $K = u_0\, g_1^{2k_1} g_2^{2k_2}$
factors as a product, so along any trajectory both $g_1$ and $g_2$ contribute
multiplicatively to the rate.

For (a), with $\theta(t) = \theta_0 + tu$ and $u = a n_1 + b n_2$, the
component coordinates evaluate to $g_1(\theta(t)) = a t + O(t^2)$ and
$g_2(\theta(t)) = b t + O(t^2)$. Substituting:
$$
K(\theta(t)) \;=\; u_0(at, bt, 0)\,(at)^{2k_1} (bt)^{2k_2}
\;=\; u_0(0,0,0)\, a^{2k_1} b^{2k_2}\, t^{2(k_1 + k_2)} (1 + O(t)),
$$
so $K(\theta(t))$ has KL order $k_1 + k_2$ along this curve (the
orders of $g_1$ and $g_2$ along $u$ add). Theorem~\ref{thm:fisher_decay}
applied to this single-direction trajectory with KL order $k_1 + k_2$
gives the rate $u^\top \fisher u = \Theta(t^{2(k_1 + k_2 - 1)})$.

For (b), with $\theta(t) = \theta_0 + \varepsilon n_2 + t n_1$, the
component coordinates are $g_1 = t + O(t^2)$ and $g_2 = \varepsilon$
(constant in $t$). Substituting:
$$
K(\theta(t)) \;=\; u_0(t, \varepsilon, 0)\, t^{2k_1} \varepsilon^{2k_2}
\;=\; \bigl(u_0(0,\varepsilon,0)\, \varepsilon^{2k_2}\bigr)\, t^{2k_1}
(1 + O(t)).
$$
The trajectory KL order in $t$ is $k_1$, with leading coefficient
$\sim u_0(0,\varepsilon,0) \varepsilon^{2k_2}$. Theorem~\ref{thm:fisher_decay}
along $n_1$ gives $n_1^\top \fisher n_1 = c_2(\varepsilon)\, t^{2(k_1-1)}
(1 + O(t))$ with $c_2(\varepsilon)$ inheriting the $\varepsilon^{2k_2}$
scaling from the leading KL coefficient.

For (c), invert the system: from (a), $k_1 + k_2 = 1 + \alpha_{\text{gen}}/2$;
from (b)(ii), $k_1 = 1 + \alpha_1/2$, hence $k_2 = (k_1 + k_2) - k_1 =
(\alpha_{\text{gen}} - \alpha_1)/2$; equivalently from (b)(iii),
$k_2 = (1/2) \partial_{\log \varepsilon} \log c_2(\varepsilon)$.
\end{proof}

\begin{remark}[Trajectory-rate observable for the per-component KL orders]
\label{rem:multi_component_observable}
Theorem~\ref{thm:multi_component_rates} promotes the multi-direction
KL orders $(k_1, k_2)$ from a resolved-coordinate algebraic invariant
to a directly trajectory-readable rate-pair. The protocol:
\begin{enumerate}
\item Fit the slope $\alpha_{\text{gen}}$ on a generic transversal
trajectory $\theta(t) = \theta_0 + tu$.
\item Fit the slope $\alpha_1$ on an $\varepsilon$-anchored trajectory
$\theta_0 + \varepsilon n_2 + t n_1$ (and symmetrically $\alpha_2$).
\item Solve $(k_1, k_2) = (1 + \alpha_1/2, 1 + \alpha_2/2)$, with the
generic-transversal rate as a consistency cross-check
$\alpha_{\text{gen}} \stackrel{?}{=} \alpha_1 + \alpha_2 + 2$.
\item Read Watanabe's local RLCT (with uniform prior, $h_i = 0$) as
$\lambda = \min(1/(2k_1), 1/(2k_2))$.
\end{enumerate}
The protocol requires identifying the component normals $n_1, n_2$ at
$\theta_0$. For models with explicit singular fibers (mixture
boundaries, rank-deficit fibers in matrix factorisation), these
normals are explicit; for models where the component structure is
implicit, identifying $n_1, n_2$ is open in general. For a layered
network the per-layer K-FAC factors expose a constructive candidate
generator (Corollary~\ref{cor:kfac_lift}): the smallest-eigenvalue
parameter direction of a layer is the Kronecker lift
$g_{\min} a_{\min}^\top$ of its gradient-covariance dead direction.
\end{remark}

\begin{remark}[Numerical verification]
\label{rem:multi_component_verification}
Theorem~\ref{thm:multi_component_rates} parts (a)-(c) are verified
analytically on the toy model
$p(x \mid g_1, g_2) = \mathcal N(g_1^{k_1} g_2^{k_2}, 1)$ with
$q(x) = \mathcal N(0, 1)$, across pairs
$(k_1, k_2) \in \{(1,1), (1,2), (2,2), (2,3), (3,2), (2,4)\}$. Every
slope fit returns the predicted exponent to within $0.01$ of the
integer value with $R^2 = 1.000$. The $k_2$ recovery via
leading-coefficient $\varepsilon$-scan is exact at machine precision
in all six cases. Two instantiation notes. The toy's population Fisher
is rank one, so no direction carries the $\Theta(1)$ complement block
of Theorem~\ref{thm:fisher_decay}'s assumption~(i): the toy exercises
parts~(a) and~(c), which the hypothesis line scopes to the score
expansion and integrability, and part~(b)'s formulas are checked on it
as formulas. And the cells with some $k_i = 1$ read the regular
$\Theta(1)$ case of the anchored trajectory, outside
Theorem~\ref{thm:fisher_decay}'s $k \ge 2$ scope but with the same
conclusion.
\end{remark}

\begin{remark}[Scope: Jacobian exponents $h_i$ remain a prior choice]
\label{rem:multi_component_h_i}
Theorem~\ref{thm:multi_component_rates} recovers the per-component KL
orders $(k_1, k_2)$ but is silent on the Jacobian exponents
$(h_1, h_2)$ of Watanabe's RLCT formula
$\lambda = \min_i (h_i + 1)/(2k_i)$. In Watanabe's framework, $h_i$
arises as the order of vanishing of the prior measure in resolved
coordinates, a property of the chosen prior, not of the model.
With the natural Lebesgue prior in original parameter coordinates,
$h_i = 0$ and the trajectory framework recovers the full RLCT
$\lambda = \min_i 1/(2k_i)$. Non-zero $h_i$ requires a non-uniform
prior $\pi(\theta) \propto \prod_j |\langle n_j, \theta - \theta_0\rangle|^{h_j}$
in original coordinates, in which case the volume integral
$\mathrm{Vol}_\pi(\{K < \varepsilon\})$ scales as $\varepsilon^\lambda$
with the full $\lambda = \min_i (h_i + 1)/(2k_i)$ as an
$\varepsilon$-scan observable rather than a single-trajectory rate.
Once the monomial support of $K$ is in hand, the toric Newton-polyhedron
computation returns this full $\lambda$ for any monomial prior directly,
the prior exponents entering the defining linear program
$\lambda = \min_{w \ge 0} \langle w, h+1\rangle$ subject to
$\langle w, \alpha_i\rangle \ge 1$, with $h = 0$ recovering the Lebesgue case.
\end{remark}

\begin{remark}[Multiplicity $m$ as a post-processing of $(k_1, \dots, k_r)$]
\label{rem:multi_component_multiplicity}
Watanabe's free-energy expansion $F_n = n L_n + \lambda \log n -
(m - 1) \log\log n + O_p(1)$ pairs the RLCT $\lambda$ with a
multiplicity $m$ defined as the order of $-\lambda$ as the largest
pole of the zeta function $\zeta(z) = \int K(\theta)^z \pi(\theta) \,d\theta$.
For an $r$-component normal-crossing singularity with KL orders
$(k_1, \dots, k_r)$ and prior weights $(h_1, \dots, h_r)$, the zeta
function factorises across components and gives
\[
\lambda = \min_i \frac{h_i + 1}{2 k_i}, \qquad
m = \bigl|\,\{\,i : (h_i + 1)/(2 k_i) = \lambda\,\}\,\bigr|.
\]
Under the uniform prior ($h_i = 0$), this simplifies to
$\lambda = 1/(2 \max_i k_i)$ and $m = |\{i : k_i = \max_j k_j\}|$,
\emph{the count of components whose KL order equals the maximum}.
Once Theorem~\ref{thm:multi_component_rates} delivers the integer
tuple $(k_1, \dots, k_r)$ as a trajectory-rate observable, the
multiplicity $m$ is recovered as a direct post-processing of that
tuple, with no additional trajectory observable required. This
closes the multiplicity item of the un-translated SLT agenda for
the multi-component case.
\end{remark}

\begin{remark}[Implementation]
\label{rem:multi_component_ddl}
The recovery protocol of Theorem~\ref{thm:multi_component_rates} runs
as $r + 1$ rate-fit trajectories on a parameter point $\theta_0$ with
component normals $(n_1, \ldots, n_r)$ and a callable returning
$u^\top F(\theta) u$ at any direction $u$: one generic-transversal,
$r$ per-component $\varepsilon$-anchored. It returns the integer
tuple $(k_1, \ldots, k_r)$, the consistency cross-check, and the
Watanabe pair $(\lambda, m)$. Verified at machine precision across
$\{(2, 2), (2, 3), (3, 2), (1, 2), (2, 4)\}$ for $r = 2$,
$\{(2, 3, 2), (3, 3, 3)\}$ for $r = 3$,
$\{(2, 3, 2, 3), (3, 2, 3, 2), (2, 2, 2, 3)\}$ for $r = 4$, and
$(2, 2, 3, 2, 3)$ for $r = 5$ on the canonical toy model.
\end{remark}

\begin{remark}[Extension to $r$-component crossings]
\label{rem:multi_component_r_extension}
The dual-component case extends to $r$-component normal crossings
$K = u_0\, \prod_{i=1}^r g_i^{2k_i}$ by the same argument: along a
generic direction transversal to all $r$ components, KL order is
$\sum_{i=1}^r k_i$ and Fisher rate is $2(\sum_i k_i - 1)$; the
$\boldsymbol{\varepsilon}$-anchored single-component rate is $2(k_i - 1)$ with
leading coefficient $\Theta(\prod_{j \neq i} \varepsilon_j^{2 k_j})$.
The integer tuple $(k_1, \dots, k_r)$ is recoverable either from one
single-component rate plus an $(r-1)$-dimensional
$\boldsymbol{\varepsilon}$-scan of the leading coefficient, or, more
simply, from $r$ per-component $\varepsilon$-anchored rates, each giving
its $k_i$ directly. The second route is what the implementation uses
(Remark~\ref{rem:multi_component_ddl}); because no single fit must
separate the components, it removes the large-$r$ delicacy of the
$\varepsilon$-anchor scaling. Verified at $r = 3$ on the toy model
$\mathcal N(g_1^2 g_2^3 g_3^2, 1)$, at $r = 4$ ($(2,3,2,3)$), and at
$r = 5$ ($(2,2,3,2,3)$): the committed probe test
(\texttt{ddl/experiment/tests/test\_multi\_component\_probe.py}) pins
the full integer tuple, $\lambda$, and $m$ at tolerance on the grid of Remark~\ref{rem:multi_component_ddl}, which includes these cells, with the generic-transversal and per-component slopes at
the predicted integers and the leading-coefficient scan matching the
predicted $\varepsilon_2^{2 k_2} \varepsilon_3^{2 k_3}$ factorisation.
\end{remark}

\begin{proposition}[Gauge-quotient resolution of the deep-linear determinantal RLCT]
\label{prop:determinantal_resolution}
Let $K = \lVert W_0 \cdots W_{L-1} - P_r \rVert_F^2$ for weight matrices
$W_s \in \reals^{H_s \times H_{s+1}}$, at a product point of true rank $r$
with $P_r$ the rank-$r$ true product (at $r = 0$ this is the plain product
norm), with $L \in \{2, 3\}$. Algorithm~\ref{alg:determinantal_resolution} returns the exact
learning coefficient $(\lambda, m)$ of $K$, for any widths and any rank
$0 \le r \le \min_s H_s$, recovering the closed form of
\citet{AoyagiWatanabe05, Aoyagi24}.
\end{proposition}

\begin{algorithm}[ht]
\caption{Gauge-quotient resolution of the deep-linear determinantal RLCT ($L \in \{2,3\}$)}
\label{alg:determinantal_resolution}
\begin{algorithmic}[1]
\REQUIRE widths $H = (H_0, \dots, H_L)$, $L \in \{2,3\}$; product rank $r$
\STATE \textbf{Rank split.} $\lambda_{\mathrm{reg}} \gets r(H_0 + H_L - r)/2$; replace $H_s \gets H_s - r$ (deficiency widths). \COMMENT{regular blocks + rank-$0$ core \citep{Hirose23}}
\STATE \textbf{Gauge quotient.} SVD the end factor(s) to remove the orthogonal gauge $W_s \mapsto W_s R,\ W_{s+1} \mapsto R^\top W_{s+1}$ ($R$ orthogonal), rewriting the core as
\STATE \quad $K = \sum_{l} \sigma_l^2 \lVert \beta_l \rVert^2$ \ ($L{=}2$), \quad or \quad $K = \sum_{l,k} \sigma_l^2 \tau_k^2 C_{lk}^2$ \ ($L{=}3$),
\STATE \quad with $\sigma$ (and $\tau$ for $L{=}3$) the singular values of the end factor(s), $\beta$ / $C$ the free gauge-fixed middle factor, and the SVD Jacobian giving the prior exponent $\lvert H_s - H_{s+1} \rvert$ on each singular value with the Vandermonde $\prod_{l < l'} \lvert \sigma_l^2 - \sigma_{l'}^2 \rvert$.
\STATE \textbf{Monomialise.} Order the singular values, $\sigma_l = \sigma_1 \prod_{2 \le j \le l} t_j$, so each $\sigma_l^2 - \sigma_{l'}^2$ is a monomial times a factor non-vanishing on the open ordered chart. One toric chart remains: a monomial support $\{\alpha_i\}$ for $K$ and a monomial prior $c$ from the pulled-back measure.
\STATE \textbf{Newton read.} $\lambda_{\mathrm{core}} \gets \min_{w \ge 0} \langle w, c + 1\rangle$ subject to $\langle w, \alpha_i\rangle \ge 1$ for every $i$; \ $m \gets \dim(\text{optimal face}) + 1$.
\RETURN $(\lambda_{\mathrm{reg}} + \lambda_{\mathrm{core}},\ m)$
\end{algorithmic}
\end{algorithm}

\begin{proof}
The rank split is the regular-block / deficiency-core decomposition: at rank $r$
the loss separates into $r(H_0 + H_L - r)/2$ regular (Morse) directions plus the
rank-$0$ determinantal core on the deficiency widths $H_s - r$ (\citealp{Hirose23},
the boundary term of \citealp{Aoyagi24}). The orthogonal gauge acts at each internal
node and fixes $K$, which depends only on the product; the singular value
decomposition of the end factors is a measure-preserving change of variables that
removes it, leaving the singular values with the Wishart Jacobian and Vandermonde
and a free gauge-fixed middle factor. For $L = 3$ the two end factors are
non-adjacent, so the middle factor is genuinely free; for $L = 2$ the second factor
plays that role directly. Ordering the singular values monomialises the Vandermonde,
and the resulting single chart is non-degenerate with respect to its Newton
polyhedron, a normal crossing for $L \le 3$, where the toric formula
$\lambda = \min_{w \ge 0} \langle w, c+1\rangle$ subject to
$\langle w, \alpha_i\rangle \ge 1$ is exact (the weighted Newton distance,
\citealp{Hirose23}). Adding the regular term gives $(\lambda, m)$, which coincides
with the closed form of \citet{AoyagiWatanabe05, Aoyagi24} on the family; the
agreement is verified on the grid of Remark~\ref{rem:determinantal_resolution}.
\end{proof}

\begin{remark}[Worked instance and scope]
\label{rem:determinantal_resolution}
For $H = (2,2,2)$ at $r = 0$ the regular term is $0$ and the chart program returns
$\lambda = 3/2$, $m = 1$, where the toric simplex bound on $K$ alone gives $2$ and
the origin blow-up stalls there at every depth. The construction recovers the
closed form on the small grid (including $3{\times}2{\times}3$,
$4{\times}2{\times}4$, $3{\times}3{\times}3$ with $m = 2$, and the three-factor
$3{\times}2{\times}2{\times}3$ with $m = 3$), on the reduced-rank grid
$H = (20, h, h, 20)$ at every rank deficit, and agrees with Monte-Carlo volume
scaling on the cells where the bound is not tight. The scope is $L \in \{2, 3\}$:
for $L \ge 4$ the gauge-fixed middle is itself a matrix product, hence
determinantal, so the resolution recurses rather than terminating in a free factor,
and the closed form of \citet{Aoyagi24} covers that regime. The gauge quotiented
here is the resolution-side reflection of the orbit the quotient Fisher rate uses on
the trajectory side (Corollary~\ref{cor:quotient_rate}): the determinantal learning
coefficient is the learning coefficient of the gauge quotient.
\end{remark}

\begin{theorem}[Universality of the singular fluctuation along a 1D dead direction]
\label{thm:nu_universality}
Let $u$ be a 1D dead direction at $\theta_0$ with KL order $k \ge 2$
under the assumptions of Theorem~\ref{thm:fisher_decay}, and suppose
the prior $\pi(\theta)$ is uniform (Lebesgue) in original parameter
coordinates. Then the leading-order singular fluctuation along $u$
depends only on the KL order $k$, not on the model-specific leading
log-likelihood coefficient $\expect_q[a_k^2]$:
\[
\nu_{\mathrm{LO}}(k) \;=\; \expect_{Z \sim \normal(0,\,2)}\!\Big[\,
\mathrm{Var}_{\pi_Z}\!\big[\eta^{k}\big]\,\Big],
\qquad
\pi_Z(\eta) \;\propto\; \exp\!\big(Z\,\eta^{k} - \eta^{2k}\big), \ \ \eta \in \reals .
\]
The empirical (data) fluctuation $Z$ is the $\sqrt n$-scaled CLT limit of the
score sum along $u$, and the data average over $Z$ is intrinsic to the singular
fluctuation. Freezing $Z = 0$ gives the symmetric renormalised posterior
$\pi_0(\eta) \propto e^{-\eta^{2k}}$, under which $T := \eta^{2k} \sim
\Gamma(\lambda, 1)$ ($\lambda = 1/(2k)$, so $\expect[T] = \lambda$) and
$\sqrt T = |\eta|^k$. Write
\[
\nu^{\sqrt T}(k) \;:=\; \expect[T] - \expect[\sqrt T]^2
\;=\; \lambda - \Big(\tfrac{\Gamma(\lambda+1/2)}{\Gamma(\lambda)}\Big)^2 .
\]
Parity decides whether this is the frozen-$Z$ variance of the functional the
singular fluctuation actually integrates, which is $\eta^k$ and not $|\eta|^k$:
\[
\mathrm{Var}_{\pi_0}\big[\eta^k\big] \;=\;
\begin{cases}
\nu^{\sqrt T}(k), & k \text{ even},\\[2pt]
\lambda, & k \text{ odd},
\end{cases}
\]
since $\eta^k$ is odd for odd $k$ and the symmetric $\pi_0$ then sets its mean
to zero. At every order the tilted variance obeys the exact relation
\[
\mathrm{Var}_{\pi_Z}\big[\eta^k\big] \;=\; \lambda + m(Z)\big(\tfrac{Z}{2} - m(Z)\big),
\qquad m(Z) := \expect_{\pi_Z}\big[\eta^k\big],
\]
and at odd $k \ge 3$ the factors satisfy $0 \le m(Z) \le Z/2$ for $Z \ge 0$,
both strictly for $Z > 0$, with $m$ odd in $Z$; hence
$\mathrm{Var}_{\pi_Z}[\eta^k] > \lambda$ for every $Z \ne 0$ and
\[
\nu_{\mathrm{LO}}(k) \;>\; \mathrm{Var}_{\pi_0}\big[\eta^k\big]
\qquad (k \ge 3 \text{ odd}).
\]
At $k = 1$ the two sides are equal. Proposition~\ref{prop:nu_even} treats the
even orders. We are not aware of an elementary closed form for
$\nu_{\mathrm{LO}}(k)$ at $k \ge 2$; it is reported numerically below.
\end{theorem}

\begin{proposition}[Strict inequality at even orders, computer-assisted]
\label{prop:nu_even}
For every even $k \ge 2$, $\nu_{\mathrm{LO}}(k) > \mathrm{Var}_{\pi_0}[\eta^k]$. The proof is computer-assisted and rests on the exact relation of Theorem~\ref{thm:nu_universality}: ball arithmetic certifies the inequality per order for every even $k \le 1000$, and a second argument certifies it uniformly in $k$ for every even $k \ge 20$, so it holds at every even order. Remark~\ref{rem:nu_even_certified} carries both arguments and names the certificates.
\end{proposition}

\begin{proof}[Sketch]
Along $u$ at $\theta_0$, with parameter $s \in \reals$, the KL expansion
(assumption (iii) of Theorem~\ref{thm:fisher_decay}) gives
$K(s) = c_K\, s^{2k} + O(s^{2k+1})$ with $c_K = \tfrac12 \expect_q[a_k^2]$,
and the leading log-likelihood deviation is $a_k(X)\, s^k$. With $n$ samples the
posterior on the dead line is, to leading order,
$\pi(s \mid \mathrm{data}) \propto \exp\!\big(s^k \sum_i a_k(X_i) - n c_K s^{2k}\big)$.
Rescaling $s = (n c_K)^{-1/(2k)} \eta$ turns this into
$\pi_Z(\eta) \propto \exp(Z\, \eta^k - \eta^{2k})$ with
$Z = (n c_K)^{-1/2} \sum_i a_k(X_i) \to \normal\!\big(0, \expect_q[a_k^2]/c_K\big)
= \normal(0,2)$ by the central limit theorem (using $c_K = \tfrac12 \expect_q[a_k^2]$).
The model-specific coefficient $\expect_q[a_k^2]$ enters only through $c_K$ in
the rescaling and cancels against the variance of $Z$, so $\pi_Z$, and hence
every posterior functional of $\eta$, depends on $k$ alone. This is the
universality.

The singular fluctuation is the data-averaged functional variance
\citep[Definitions 21 and 22, Lemma 23]{Watanabe18}, $\nu = \tfrac12 \lim_n n\,\expect[V_n]$ with
$V_n = \tfrac1n \sum_i \mathrm{Var}_{\mathrm{post}}[\log p(X_i \mid \theta)]$. Its leading
term is $\mathrm{Var}_{\mathrm{post}}[\log p(X \mid \theta)] = a_k(X)^2\,
\mathrm{Var}_{\mathrm{post}}[s^k] + O(\cdot)$. Averaging over data, the factor
$\expect_q[a_k^2] \cdot \mathrm{Var}_{\mathrm{post}}[s^k]
= \expect_q[a_k^2]\cdot (n c_K)^{-1}\, \mathrm{Var}_{\pi_Z}[\eta^k]$ again cancels
the coefficient ($\expect_q[a_k^2]/c_K = 2$), giving
$\nu_{\mathrm{LO}}(k) = \expect_{Z\sim\normal(0,2)}[\mathrm{Var}_{\pi_Z}[\eta^k]]$.

The comparison with the frozen value runs through two exact facts about the
tilted family. Write $c(Z) = \int e^{Z\eta^k - \eta^{2k}}\,d\eta$ and
$c_j(Z) = \int \eta^{jk} e^{Z\eta^k - \eta^{2k}}\,d\eta$, so that
$m(Z) = c_1/c$ and $\mathrm{Var}_{\pi_Z}[\eta^k] = c_2/c - m^2$. The function
$\eta\,e^{Z\eta^k - \eta^{2k}}$ vanishes at both ends of the line, so the
integral of its derivative is zero: $c + kZ c_1 - 2k c_2 = 0$, the Gamma-Stein
identity $c_2 = \lambda c + \tfrac{Z}{2} c_1$ with $\lambda = 1/(2k)$. Dividing
by $c$ gives the relation displayed in the statement. At $Z = 0$ it returns
$\lambda - m(0)^2$, which is $\lambda$ at odd $k$, where $m(0) = 0$ by parity,
and $\nu^{\sqrt T}(k)$ at even $k$, where
$m(0) = \Gamma(\lambda + \tfrac12)/\Gamma(\lambda)$.

The second fact is the sign of the two factors at odd $k$. For $Z \ge 0$,
pairing $\eta$ with $-\eta$ writes $c_1(Z)$ as
$\int_0^\infty \eta^k (e^{Z\eta^k} - e^{-Z\eta^k}) e^{-\eta^{2k}}\,d\eta \ge 0$,
so $m(Z) \ge 0$, strictly for $Z > 0$. For the upper bound substitute
$s = \eta^k$, a bijection of the line at odd $k$ with
$d\eta = |s|^{1/k - 1}\,ds/k$, and complete the square:
\[
\tfrac{Z}{2}\,c(Z) - c_1(Z)
\;=\; \frac{e^{Z^2/4}}{k}\int_{\reals} (-v)\,e^{-v^2}\,
\big|v + \tfrac{Z}{2}\big|^{1/k - 1}\,dv .
\]
Pairing $v$ with $-v$ turns the integrand into
$v\,e^{-v^2}\big(|\tfrac{Z}{2} - v|^{1/k - 1} - |\tfrac{Z}{2} + v|^{1/k - 1}\big)$,
which is nonnegative because the exponent $1/k - 1$ is nonpositive and
$|\tfrac{Z}{2} - v| \le |\tfrac{Z}{2} + v|$ for $v \ge 0$, with the reverse
ordering for $v \le 0$; for $k \ge 3$ it is strictly positive on
$0 < v < Z/2$. Hence $0 \le m(Z) \le Z/2$, both strictly for $Z > 0$, and
the relation gives $\mathrm{Var}_{\pi_Z}[\eta^k] > \lambda$ for $Z > 0$; the
case $Z < 0$ follows from $m(-Z) = -m(Z)$. The variance is continuous in $Z$
and at most $\lambda + Z^2/16$, so its $\normal(0, 2)$ average exists and
exceeds $\lambda$ strictly. At even $k$ the factor $\tfrac{Z}{2} - m(Z)$
changes sign and the variance dips below the frozen value for $Z < 0$, so the
sign argument does not apply; Remark~\ref{rem:nu_even_certified} proves the
inequality at every even $k$ from the same relation, with interval
arithmetic carrying the bounded part of the argument. At $k = 1$ the tilted
posterior $\pi_Z \propto e^{Z\eta - \eta^2}$ is Gaussian with variance $1/2$
for every $Z$, since a linear tilt moves a Gaussian's mean and leaves its
variance alone, so $\nu_{\mathrm{LO}}(1) = \lambda = 1/2$ exactly; what
$\nu^{\sqrt T}$ misses at $k = 1$ is the parity of $\eta^k$, not the data
average. Every step of this comparison, the Stein identity at every order and
the sign argument at odd order, is machine-checked
(\texttt{renormVar\_riccati}, \texttt{nuLO\_odd\_gt}).
\end{proof}

\begin{remark}[Numerical values]
\label{rem:nu_universality_values}
Two independent methods agree on $\nu_{\mathrm{LO}}(k)$: a deterministic
quadrature of the renormalised-posterior integral above, and a finite-$n$ Monte
Carlo of the functional variance $V_n$ on the power-law model
$\mathcal N(s^k, 1)$. The regular case $k = 1$ (added as a consistency anchor,
outside the $k \ge 2$ hypothesis) returns the textbook value $\nu = \lambda = 1/2$,
which $V[\sqrt T]$ does \emph{not} reproduce, for the parity reason above.
\[
\begin{array}{c|c|c|c|c}
k & \lambda & \nu^{\sqrt T}(k)\ (= V[\sqrt T]) & \mathrm{Var}_{\pi_0}[\eta^k]\ (\text{frozen } Z) & \nu_{\mathrm{LO}}(k)\ (\text{full, numerical}) \\
\hline
1 & 1/2 & 0.182 & 0.500 & 0.500 \ \ (\text{regular anchor}) \\
2 & 1/4 & 0.136 & 0.136 & 0.173 \\
3 & 1/6 & 0.107 & 0.167 & 0.278 \\
4 & 1/8 & 0.089 & 0.089 & 0.138 \\
5 & 1/10 & 0.076 & 0.100 & 0.206 \\
\end{array}
\]
$\nu_{\mathrm{LO}}(k)$ is \emph{not} monotone in $k$: odd and even orders form two
interleaved decreasing sequences (odd $k$ is higher), a parity effect of
$\eta^k$ under the symmetric renormalised posterior. The same parity separates
the middle two columns, which agree at even $k$ and differ at odd $k$ by the
zeroed mean. The $V[\sqrt T]$ column decreases monotonically and is below
$\nu_{\mathrm{LO}}$ at every order, with the large-$k$ asymptote
$\nu^{\sqrt T}(k) \sim \lambda - \pi\lambda^2 = \tfrac{1}{2k} -
\tfrac{\pi}{4k^2} + O(k^{-3})$. Reading it as the frozen-$Z$ value of the
singular fluctuation holds only at even orders.
\end{remark}

\begin{remark}[Computer-assisted proof at even orders]
\label{rem:nu_even_certified}
Since $Z$ is symmetric, $\nu_{\mathrm{LO}}(k) > \mathrm{Var}_{\pi_0}[\eta^k]$
is equivalent to $h(Z) := \mathrm{Var}_{\pi_Z}[\eta^k] +
\mathrm{Var}_{\pi_{-Z}}[\eta^k] - 2\,\mathrm{Var}_{\pi_0}[\eta^k] \ge 0$ with
strict inequality on a set of positive measure, and $h$ is even. Write
$a = 1/k$, so that $|\eta|^k$ has the tilted law $\propto s^{a-1}e^{Zs - s^2}$
on $(0, \infty)$, and $m(Z) = \expect_{\pi_Z}[|\eta|^k]$. Two arguments cover
the even orders. For every even $k \le 1000$ we certify $h(Z) > 0$ for all
$Z > 0$ in ball arithmetic: the tilted moments are ratios of the closed forms
\[
J_\nu(Z) = \int_0^\infty s^{\nu - 1} e^{-s^2 + Zs}\,ds
= \tfrac12\Gamma(\tfrac{\nu}{2})\,{}_1F_1\!\big(\tfrac{\nu}{2}; \tfrac12; \tfrac{Z^2}{4}\big)
+ \tfrac{Z}{2}\Gamma(\tfrac{\nu+1}{2})\,{}_1F_1\!\big(\tfrac{\nu+1}{2}; \tfrac32; \tfrac{Z^2}{4}\big),
\]
each moment is increasing in $Z$, so endpoint values enclose it on an
interval; $h'' > 0$ on $[0, 1/2]$ gives $h > 0$ there, a bisection of
$[1/2, 8]$ certifies $h > 0$, and for $Z \ge 8$ a sub- and a super-solution
$Z/2 - c/Z$ of the Riccati equation $m' = \lambda + m(Z/2 - m)$ anchored at the
enclosure of $m(8)$ give $\mathrm{Var}_{\pi_Z}[\eta^k] > 2\,\mathrm{Var}_{\pi_0}[\eta^k]$.
For every $a \le 1/20$, that is every even $k \ge 20$, a second argument is
uniform in $k$. For $Z \ge 2$, elementary bounds give
$0.84a \le m(2) \le 0.3$, the super-solution $Z/2 - 1/(4Z)$ caps $m$, and
concavity of $m \mapsto m(Z/2 - m)$ yields
$m(Z)(Z/2 - m(Z)) \ge \lambda$, hence $h(Z) \ge 2m(0)^2 > 0$. For
$0 < Z \le 0.89$ a first-zero argument in the exact planar system
$D' = \kappa - h$, $h' = Z/4 - D(K + Z^2/4 - D^2) + 3Dh$, with
$D = Z/2 - m(Z) + m(-Z)$ and explicit constants $\kappa, K$, shows that a
first zero of $h$ would have $h' > 0$ there. For $0.89 \le Z \le 2$, writing
$c(Z) = 1/a + R(Z)$ makes $h/a$ an explicit expression in $R$ and the two
moment integrals, each monotone in $Z$ and, after a sign split, monotone in
$a$; ball arithmetic certifies a lower bound $0.22$ for $h/a$ on forty
subintervals, uniformly over $a \in (0, 1/20]$. The relation
$\mathrm{Var}_{\pi_Z}[\eta^k] = \lambda + m(Z)(Z/2 - m(Z))$ that both arguments
rest on is machine-checked at every order; the argument for $k \ge 20$ never
uses that $a$ is the reciprocal of an integer.
\end{remark}

\begin{remark}[Empirical verification: universality and value]
\label{rem:nu_universality_verification}
Two claims are checked separately. \emph{Universality} (the leading coefficient
$\expect_q[a_k^2]$ cancels): on the 2-component Gaussian mixture at $k = 2$ across
four mixture weights $w \in \{0.5, 0.4, 1/3, 0.2\}$, the functional-variance
estimate $\widehat\nu_\infty$ is independent of $\expect_q[a_2^2]$ despite a
factor-of-$2.4$ variation in that coefficient; an independent cross-check varying
the leading-coefficient scale $c$ in $\mathcal N(c\,s^k, 1)$ confirms
$c$-independence at $k = 2$ and $k = 3$.

\emph{Value}: the direct functional-variance computation pins
$\nu_{\mathrm{LO}}(2) \approx 0.173$ and $\nu_{\mathrm{LO}}(3) \approx 0.278$.
The independent $k = 3$ Monte Carlo returns
$\widehat\nu_\infty = 0.274 \pm 0.038$, consistent with
$\nu_{\mathrm{LO}}(3)$, $4.4\sigma$ above $V[\sqrt T](3) = 0.107$ and
$2.8\sigma$ above the frozen-$Z$ value $0.167$. The second gap is the one that
speaks to the data average, since $0.167$ is what the $k = 3$ posterior gives at
$Z = 0$. The mixture estimates at $k = 2$ ($\widehat\nu_\infty$ in
$0.106$--$0.112$, $\sigma \sim 0.03$--$0.04$) confirm the $w$-independence but
are individually too noisy to separate the frozen-$Z$ value $0.136$ from
$\nu_{\mathrm{LO}}(2) \approx 0.173$; the value is pinned by the tighter
power-law computation and the $k = 1, 3$ anchors. The empirical fluctuation is
therefore not negligible: the singular fluctuation is the data-averaged
functional variance, not its $Z = 0$ value.
\end{remark}

\begin{figure}[ht]
\centering
\includegraphics[width=\textwidth]{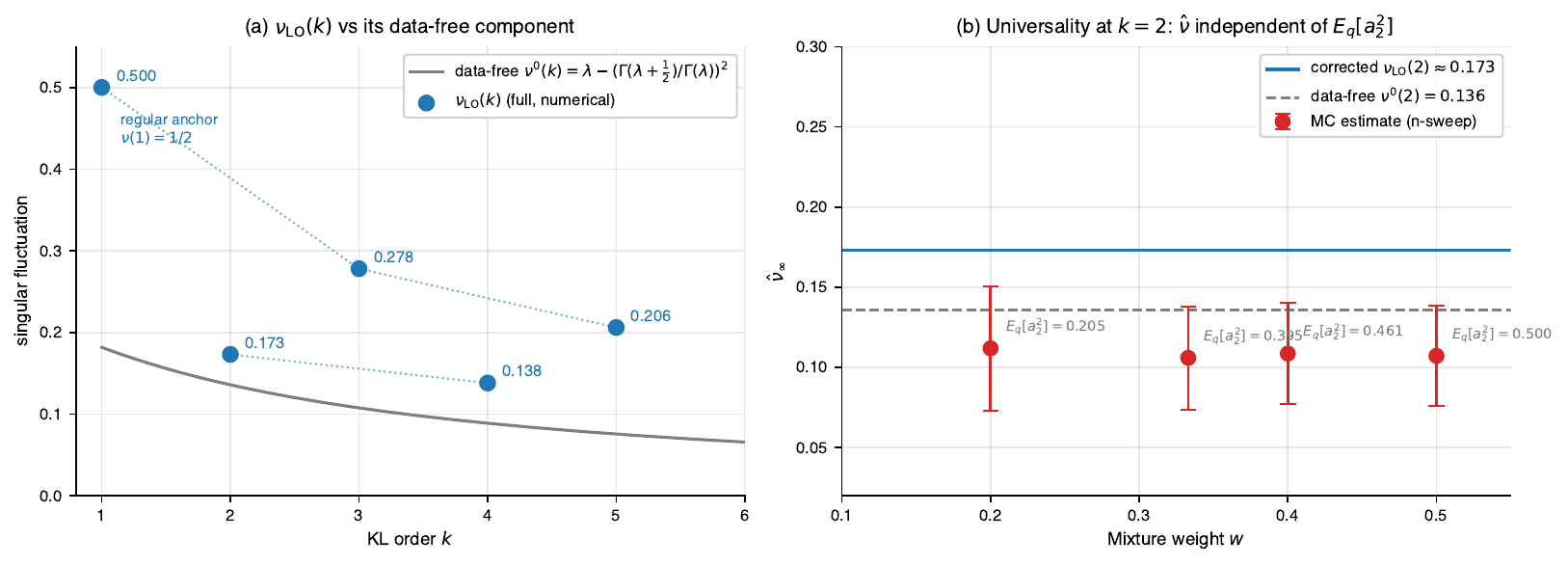}
\caption{Singular fluctuation along a 1D dead direction (Theorem~\ref{thm:nu_universality}). \textbf{(a)}~The full singular fluctuation $\nu_{\rm LO}(k)$ (data-averaged functional variance, points; two independent methods) against $V[\sqrt T] = \lambda - (\Gamma(\lambda+1/2)/\Gamma(\lambda))^2$ (lower curve), which is below $\nu_{\rm LO}$ at every order and equals the frozen-$Z$ variance only at even $k$; $\nu_{\rm LO}$ is non-monotone, with odd $k$ (including the regular anchor $\nu(1) = 1/2$) above the even-$k$ sequence. \textbf{(b)}~Universality: at $k = 2$ across mixture weights $w \in \{0.5, 0.4, 1/3, 0.2\}$ the estimate $\hat\nu_\infty$ is independent of the leading coefficient $\mathbb{E}_q[a_2^2]$ (which varies by $2.4\times$), confirming the cancellation. These mixture estimates are too noisy to separate $\nu_{\rm LO}(2) \approx 0.173$ from the frozen-$Z$ value $0.136$ (both shown); the value is settled by the tighter power-law Monte Carlo and the $k = 1, 3$ anchors.}
\label{fig:nu_universality}
\end{figure}

\begin{remark}[Scope: dead direction only; non-uniform priors; multi-direction]
\label{rem:nu_universality_scope}
Theorem~\ref{thm:nu_universality} is for a single 1D dead direction
with uniform prior. Three explicit boundaries:
\begin{itemize}\itemsep=0pt
\item \emph{Regular directions} ($k = 1$, $a_1 \ne 0$ in $L^2$): $\nu_{\mathrm{LO}}$ is continuous with the regular case, returning $\nu = \lambda
= d/2$ ($= 1/2$ in 1D), the standard realizable-Fisher value
(\citealp{Watanabe18}, Theorem 6 and Remark 25). Here the tilted posterior is Gaussian with
variance $1/2$ for every $Z$, so the regular value already appears at frozen
$Z$. The quantity $V[\sqrt T]$ returns $1/2 - 1/\pi \approx 0.182$ instead,
which is one way to see that it tracks $|\eta|^k$ and not the odd functional
$\eta^k$ the singular fluctuation integrates.
\item \emph{Non-uniform priors}: a prior $\pi(\theta) \propto |\langle u,
\theta - \theta_0\rangle|^h$ shifts the renormalised shape parameter to
$\lambda + h/(2k)$; the universality argument extends verbatim, with the
data-averaged functional variance computed at the shifted shape (and its
frozen-$Z$ value obtained by the same $\lambda \mapsto \lambda + h/(2k)$
substitution, at even $k$ in $V[\sqrt T]$ and at odd $k$ in $\lambda$).
\item \emph{Multi-direction}: along multi-component normal crossings
(Theorem~\ref{thm:multi_component_rates}), the renormalised posterior factorises
into per-component pieces under the additive transversal normal form; the
universality argument extends per-component, and the aggregate $\nu$ is
$\sum_i \nu_{\mathrm{LO}}(k_i)$ in the additive case. The multiplicative
(Hironaka-resolved) normal-crossing case $K = u_0 \prod_i g_i^{2k_i}$ is the
natural next-target extension.
\end{itemize}
\end{remark}

\begin{remark}[Tempered posteriors]
\label{rem:nu_tempered}
At inverse temperature $\beta$ the tempered posterior weights the empirical
tilt by $\beta$, so the central-limit step of the theorem's proof delivers
$Z \sim \mathcal N(0, 2\beta)$ in place of $\mathcal N(0, 2)$, and the
gen-train gap converges to the $\beta$-tilted average of the same
renormalised variance:
\[
\frac{n}{2}\,\expect\bigl[L_{\mathrm{val}} - L_{\mathrm{train}}\bigr]
\;\longrightarrow\;
f_k(2\beta) \;:=\; \expect_{Z \sim \mathcal N(0, 2\beta)}
\bigl[\mathrm{Var}_{\pi_Z}[\eta^k]\bigr].
\]
The inner variance is the exact relation of the theorem
($\mathrm{Var}_{\pi_Z}[\eta^k] = \lambda + m(Z)(Z/2 - m(Z))$,
\texttt{renormVar\_riccati}), so the limit interpolates the two values the
theorem separates: $f_k(2\beta)$ decreases to the frozen-$Z$ variance as
$\beta \to 0$ and equals $\nu_{\mathrm{LO}}(k)$ at $\beta = 1$; at $k = 1$
the tilted posterior stays Gaussian with variance $1/2$, so
$f_1 \equiv 1/2$ for every $\beta$ (\texttt{nuLO\_one}'s computation).
Model-independence at each fixed $\beta$ holds by the theorem's cancellation
(\texttt{nu\_scale\_invariance}). Quadrature gives $f_2 = 0.157$, $0.173$,
$0.193$ at $\beta = 1/2, 1, 2$, matched by variance-reduced Monte Carlo at
$0.152$, $0.167$, $0.184$ (standard error $0.007$).
\end{remark}
  
\part{Open problems and discussion}

\section{What remains open}
\label{sec:theory:open}

The closures in \S\ref{sec:theory:extensions} carry the bridge across the bulk of Watanabe's broader inventory. The remaining items are the following. None is, as far as we can see, structurally blocked; each is a research target.

\paragraph{Determinantal singularities off the coordinate crossings.} Theorem~\ref{thm:multi_component_rates}'s joint-recovery protocol handles a coordinate normal crossing for any number of components. Each $k_i$ comes from its own $\varepsilon$-anchored trajectory rather than from disambiguating the rate sum, so the recovery extends past $r = 3$ with no re-engagement of the resolution machinery (Remark~\ref{rem:multi_component_r_extension}, verified to $r = 5$); the only subtlety at large $r$ is choosing the $\varepsilon$-anchor scaling so the higher-component cross-terms stay subleading. For a singularity that is non-degenerate with respect to its Newton polyhedron, the global learning coefficient $(\lambda, m)$ follows from the monomial support of $K$ by the toric Newton-polyhedron formula, the diagonal Newton distance, with no per-direction heuristic; this covers every coordinate normal crossing, separable sum, and their composites exactly. The open case is the determinantal singularity, which is \emph{degenerate} with respect to its Newton polyhedron. Its crossings lie in coupled directions away from any coordinate, as in the matrix product-zero variety of a wide deep linear network, so a per-coordinate read returns every order as one and misses the coupling. There the Newton-polyhedron formula gives only an upper bound (the simplex bound, tighter than the regular $d/2$ and on some grids already exact), and recovering the exact value needs the Hironaka resolution of the determinantal ideal. The simple origin blow-up does not reach it, stalling at the simplex bound. For the deep linear network at depth $L \le 3$ the structured gauge-quotient resolution closes the gap, recovering the exact $(\lambda, m)$ explicitly (Proposition~\ref{prop:determinantal_resolution}, \S\ref{sec:theory:determinantal}). What stays open is the resolution at depth $L \ge 4$, where the gauge-fixed middle is itself a matrix product and the construction recurses, a general exact resolver for an arbitrary determinantal variety, and the resolution at the realistic rank-$r$ point at scale. In the meantime the measurements spoke reads a coefficient bracket from a generic-line probe and flags a determinantal point blind, by the mismatch between the generic-line order and the coordinate orders{} \citep{shirodkar2026measuringdeaddirectionsdecomposing}.

\paragraph{Trajectory-rate analog of $\nu$ in original coordinates.} Theorem~\ref{thm:nu_universality} establishes that $\nu_{\mathrm{LO}}(k)$ is universal (a function of the KL order alone) and reports it numerically; only $V[\sqrt{T}] = \lambda - (\Gamma(\lambda+1/2)/\Gamma(\lambda))^2$ has an elementary closed form, and it is the frozen-fluctuation value of $\nu_{\mathrm{LO}}$ only at even $k$ (at odd $k$ parity makes that value $\lambda$), and a closed form for the full $\nu_{\mathrm{LO}}$ (with the empirical-fluctuation average included) is itself open. A complementary observable would specify how the gen-train gap scales \emph{along an approach} as a function of $\sigma_{\min}$ in original parameter coordinates, anchored at the bridge framework's predicted exponent $\lambda$. This is a distinct construction from the rate observables for $\lambda$ surfaced throughout the present work: it requires both a training-side and a population-side loss reading along the trajectory. Promoting it to a real-network observable is straightforward in principle (track training and validation loss alongside the existing Fisher diagnostics) but requires the Watanabe-side integral on the resolved fiber to be closed for the bridge framework's K-FAC-decomposed Fisher, which we leave for follow-on work.

\paragraph{Theoretical $\beta$-sweep of $(\lambda, m)$ at $\beta \ne 1$.} The operational $\beta$-rescaling of the gen-train gap (\S\ref{sec:theory:priors_temper}) lets the trajectory side track tempered posteriors. What the present theorems do not yet address is the theoretical $\beta$-sweep of the RLCT and multiplicity themselves: under WBIC's $\beta = 1/\log n$ tempering, the leading constants in Watanabe's free-energy expansion are $\beta$-rescaled but the qualitative structure $(\lambda, m)$ is preserved. A precise statement of how the trajectory rate exponent shifts under $\beta \ne 1$, beyond the convention-preserving rescaling, remains open.

\paragraph{Non-i.i.d.\ data.} Singular models with dependent samples (Markov processes, time series) fall outside the i.i.d.\ setting of this paper's score expansion; the extension to non-i.i.d.\ data (sequence-data SLT) has not yet been addressed here and is a natural follow-on direction.

\paragraph{Posterior-sampling tasks with arbitrary $f(\theta)$.} Bayesian quantities of the form $\expect_{\rm post}[f(\theta)]$ for an arbitrary $f$ that was not observed per checkpoint are structurally outside what a deterministic-trajectory framework can reach: the integrand cannot be re-evaluated from stored trajectory data alone. SGLD-based posterior sampling remains the native tool. The boundary is methodological: the trajectory rate covers everything readable from a per-checkpoint deterministic observable, and arbitrary-$f$ posterior expectations belong to the stochastic-sampling side.

\paragraph{Amari-side connections not yet drawn on.} The dual $(\nabla, \nabla^*)$ connections of information geometry interact with the gauge-quotient picture of \S\ref{sec:theory:quotient} in ways the present framework has not exploited. The exponential / mixture flatness duality of standard families has not yet been used to characterise which families are ``simple'' (single-direction, low KL order) versus ``complex'' (multi-component) from the rate viewpoint. Both are natural directions for follow-on theory.

\paragraph{Closed-form rate modifiers for non-equivariant preconditioners.} Section~\ref{sec:theory:adam}'s constructive resolution provides a $G$-equivariant Adam-family preconditioner that satisfies the quotient theorem's sufficient condition. A closed-form per-trajectory rate modifier on the alignment-rotated manifold for \emph{standard} non-equivariant Adam remains open, as it does for the other non-SGD preconditioners (K-FAC, Shampoo, Muon), none of which has a derived trajectory rate. \citet{Pesme2021} give the analogous analysis for diagonal linear networks under SGD.

\paragraph{Trajectory-rate exponents for normalisations beyond LayerNorm.} The mean-subtraction dichotomy of Proposition~\ref{prop:ln_kernel} fixes which production normalisations carry an algebraic kernel direction, but the trajectory-rate exponent under BatchNorm, GroupNorm, and the RMSNorm variants is not derived. RMSNorm removes the mean-subtraction leak, so the finite-$t$ rate shift of Theorem~\ref{thm:ln_finite_t_mlp} is absent (Appendix~\ref{sec:theory:arch:ln_finite_t}), but its divisive factor is the same $1/\sigma$ that sets the plateau end of the LayerNorm bracket (Proposition~\ref{thm:bridge_ln}), and that bracket has not been carried over.

\paragraph{Architectural questions left open in their sections.} Several narrower questions are stated where they arise and are collected here so the list is complete: near-canonical continuity at network level, where the crossover scale is known at the model level and the constants are not (Remark~\ref{cor:bridge_near_canonical}); whether the LayerNorm rate always reduces to the nearest-crossing formula, which every measured configuration satisfies and only regime (i) proves (Proposition~\ref{thm:bridge_ln}); the attention form of the LayerNorm crossover (Conjecture~\ref{conj:ln_finite_t_attn}) and the configuration independence of the attention-chain formulas (Conjecture~\ref{conj:attn_chain_softmax_general}); the residual-stream depth invariance under non-linear residual-branch wrappings (Corollary~\ref{cor:sigma-min-res}'s Peri-LN scope note) and the in-block attention rate under a residual; multi-head attention with per-head dead directions and position encodings (the scope list of \S\ref{sec:theory:arch:attn}); and rectangular chains beyond the shared narrow subspace (Theorem~\ref{thm:bridge_rect}).
  
\section{Discussion}
\label{sec:theory:discussion}

\paragraph{What the bridge gives each tradition.} Singular learning theory gains a rate axis: questions about how fast a singularity is approached, under what dynamics, and with what optimiser-specific corrections become well-posed at the trajectory level rather than only at the asymptotic-posterior level. Theorem~\ref{thm:selection_rule} makes the single-direction contribution $1/(2k)$ to the normal-slice RLCT recoverable from a Fisher rate exponent on smooth singular fibres under the genericity pair (G) and (G$^+$), in original parameter coordinates rather than in resolved coordinates. Theorem~\ref{thm:multi_component_rates} extends the recovery to the multi-component normal-crossing form $K = u(g)\prod_i g_i^{2k_i}$, giving the joint $(k_1, \ldots, k_r)$ tuple from trajectory rates; the Jacobian exponents $h_i$ in Watanabe's full RLCT enter as an explicit prior choice. Theorem~\ref{thm:nu_universality} carries the bridge to the singular fluctuation $\nu$, the second summary invariant of singular generalisation, as a universal function of the KL order for $1$D dead directions with uniform prior (only $V[\sqrt T] = \lambda - (\Gamma(\lambda+1/2)/\Gamma(\lambda))^2$ has an elementary closed form, and it equals the frozen-fluctuation value only at even $k$; the full value is numerical). Information geometry gains an extension to its degeneracy locus: dead directions are objects in Amari's language, with rates that are properties of the Fisher metric in a regime where the metric itself ceases to be invertible; the A--G duality is stated entirely in Kronecker-factored Fisher language; and the multi-layer K-FAC bridge extends the natural-gradient lineage to the singular regime where natural gradient is not defined.

\paragraph{Position within the broader SLT-DL landscape.} The trajectory-rate framework is one of a small cluster of recent programs operationalising different aspects of singular structure in deep networks. Each one starts from Watanabe (2009) but goes in its own direction; we name each and indicate the relationship to the present framework. The \emph{local learning coefficient} \citep{LauFurmanWangMurfetWei25,WangHoogland24,HooglandWangFarrugiaRoberts24} estimates Watanabe's $\lambda$ numerically via SGLD posterior sampling around a checkpoint, with a refined variant attaching the estimate to specific weight subsets and a stagewise reading tracking $\lambda$ across training. The trajectory-rate framework is the deterministic, K-FAC-block-local, sampling-free alternative: same $\lambda$, different operationalisation. The \emph{susceptibility} programme \citep{BakerWangHooglandMurfet25,GordonBakerWang26} perturbs the data distribution and reads the response of posterior expectations of localised observables, clustering modules and tokens by response signature; the trajectory-rate framework provides a per-K-FAC-block rate that susceptibilities can be conditioned on. The \emph{loss kernel} of \citet{AdamFurmanHoogland25} is a geometric probe of model internals (the covariance of per-sample losses under low-loss-preserving perturbations), with the Hessian-free \emph{Bayesian influence functions} of \citet{KreerWuAdamFurmanHoogland25,LeeSmithAdamHoogland25} carrying the same posterior geometry into data attribution sensitive to higher-order degeneracy; these are geometric companions to the leading-rate readout. \emph{Compressibility and minimum description length} \citep{Urdshals25Compressibility} formalise the bit-length of a singular minimum and recover MDL in the regular case; the trajectory-rate framework gives the rate exponent per direction, which compressibility integrates over. \emph{Programs as singularities} \citep{MurfetTroiani25} connects degenerate statistical models to the geometry of programs; the trajectory rate is one observable on the singular-program manifold this line constructs. \emph{Modes of sequence models} \citep{ChenMurfet25} characterises the sensitivity of LLC estimation to input-distribution patterns; the trajectory-rate framework's expected-Fisher protocol (\S\ref{sec:theory:arch:ce}) is the corresponding sensitivity statement on the rate side. None of these lines is a competitor; each probes a slice of the singular geometry the present framework also touches, and each provides a falsifier for the trajectory-rate readings on architectures where the lines overlap.

\paragraph{Forward-looking directions the bridge unlocks.} The theorems above give concrete leverage on several practical problems. We phrase each as an open question that the framework provides the analytic target for, rather than as a result the framework promises.

\emph{Compression and rank reduction.} Is the Fisher rate exponent the rate-correct subspace for LoRA placement, structured pruning, and low-rank fine-tuning? The framework provides the analytic criterion (dead-direction definition $u^\top F u \to 0$) and a finer signal than rank-deficiency (the rate exponent). The compressibility-MDL line of \citet{Urdshals25Compressibility} pursues a related direction from the LLC side; whether the per-direction rate exponent gives a better subspace-selection criterion than the standard activation-variance or empirical-Fisher heuristics is an open empirical question.

\emph{Continual-learning safe-direction constraints.} Methods that constrain updates to directions leaving the current task's loss unchanged to leading order (orthogonal-gradient, null-space) currently identify these directions empirically. Is the dead-direction definition, a unit vector along which $u^\top F u \to 0$, the analytic target these methods are approximating? Falsifying this requires comparing the empirically-selected safe directions to the framework's predicted dead directions on a continual-learning benchmark.

\emph{Equivariant optimizer design under gauge invariance.} What does an equivariant preconditioner look like for each architectural gauge group? The quotient theorem (Corollary~\ref{cor:quotient_rate}) specifies the design constraint; DDCAdam (Algorithm~\ref{alg:ddcadam}) constructs it for three classes (CE row-shift, ReLU rescale, LN scale). Extensions to attention-head permutation, RoPE phase symmetry, and other architectural gauges are open. \citet{DePaviaCharisopoulosWillett25} study how parameter-space rotations affect Adam's implicit bias.

\emph{Architectural primitives as rate-shaping choices.} Each per-architecture lemma is a structural fact about how a designer's choice of block class shapes the singular geometry. Residual DAGs preserve rate $0$ on the residual stream. LayerNorm has an algebraic kernel direction $\gamma^{-1}/\|\gamma^{-1}\|$ at the post-norm position; RMSNorm has no such universal kernel. Attention chains break composition additivity at depth $\ge 4$. The open question for architecture search is whether the rate profile under canonical alignment is itself a useful objective: whether tuning a network's design to make the rate ladder shallower or steeper changes generalisation in a predictable way.

\emph{Mechanistic interpretability.} Sparse-autoencoder feature directions and dead directions intersect: are features that are also dead the features the network has implicitly compressed or dropped, and does ablating them leave the loss unchanged to leading order? Superposition in transformer residual streams has a structural reason: Corollary~\ref{cor:sigma-min-res} gives the geometric capacity that supports superposition feasibility. The LayerNorm kernel direction is a deterministic probe direction available in every LN-equipped model, with the algebraic guarantee $\mathrm{cov}(\mathrm{LN}(X))\, v^* = 0$ for any input distribution. Whether the per-K-FAC-block rate exponent and the LN kernel direction combine into a useful circuit-level diagnostic, one that flags which sub-circuit a feature belongs to by its rate signature, is an open empirical question. The trajectory-rate readouts are complementary to other geometric probes in the recent SLT-DL literature: per-attention-head specialisation via the refined LLC \citep{WangHoogland24}, the loss-kernel probe of \citet{AdamFurmanHoogland25}, and the sequence-mode analysis of \citet{ChenMurfet25}; each operationalises a different aspect of singular structure, and the trajectory rate adds the deterministic, sampling-free per-layer reading to that roster.

\emph{Trajectory-rate residuals as training diagnostics.} The rate primitive of Theorem~\ref{thm:fisher_decay} is the foundational construct of the framework; the per-layer K-FAC ladder, residual-stream $\sigma_{\min}$, LN kernel direction, and any further bridge-derived observable inherit a rate prediction under canonical alignment. Real networks at scale do not satisfy the conditions under which those predicted exponents are exactly realised. The framework's role on such networks is a reference frame: the deviation between observed and predicted at any checkpoint is a per-layer, per-trajectory, rate-grounded observable, computable for any observable the bridge defines now or in the future. Whether the deviation profile across training predicts downstream quality (final benchmark scores, generalisation behaviour, pathology signatures) is an open empirical question the framework reduces to a well-defined measurement. The contribution here is the reference-frame structure itself: current and future bridge-derived observables inherit a definable deviation against an idealised target, independent of which observable is being read.

\emph{Diagnostic instrumentation for pretraining and fine-tuning.} Are the activation-side $\sigma_{\min}(X_\ell)$ and gradient-side $\lambda_{\min}(G_\ell)$ readouts useful as structural-health indicators during large training runs? Both are consequences of the framework, computable from forward and backward passes at any checkpoint, and rate-grounded by Theorems~\ref{thm:bridge} and Corollary~\ref{cor:sigma-min-res}. Whether the readings catch training pathologies (loss spikes, dead-channel emergence, gauge-mode drift under Adam) before downstream metrics do is open, a specific instance of the residual-as-diagnostic framing above.

\emph{Algebraic-baseline vs trained singular structure.} A concrete instance of the residual-as-observable framing. Lemma~\ref{lem:schur_constant_quantitative} pins down the leading constant of the activation-Gram Schur complement in the canonical setup, yielding the width-only invariant $R(h) := \lambda_{\min}(A_\ell) / (A_\ell)_{u,u}$ at the framework's predicted dead direction. Reading this on a LayerNorm-equipped pretrained model at random initialisation, with $u = \gamma^{-1}/\|\gamma^{-1}\|$ from Proposition~\ref{prop:ln_kernel}, gives $R \approx 1$ within finite-sample drift; on the trained checkpoint, $R(\gamma^{-1})$ collapses, indicating that training opens additional dead directions \emph{below} the algebraic baseline. Is the gap between random-init and trained $R(\gamma^{-1})$ a useful per-checkpoint observable, separating architectural-algebraic null structure (fixed by the LN parameter alone) from training-induced SLT-singular structure (the dead directions the optimiser opens as it consolidates, in the compression regime of \S\ref{subsec:theory:regimes})? The observable is forward-pass-only, computable per block, with no gradient access required, and, being static, reads the same in either regime; its empirical utility across architectures is open.  This question is taken up at LLM scale in a companion paper~\citep{shirodkar2026algebraicdeaddirectionslayernorm}, which reads $\gamma^{-1}/\|\gamma^{-1}\|$ and the random-init-to-trained gap as a forward-pass-only diagnostic on $14$ pretrained transformers.

\paragraph{Closing.} This paper makes Watanabe's KL-order invariant trajectory-readable in original parameter coordinates, and shows that a single rate exponent organises the observable roster of a deep network: per-layer K-FAC factors, residual-stream $\sigma_{\min}$, the LayerNorm kernel direction, attention-chain composition, and the high-curvature Fisher-volume scaling (Corollary~\ref{cor:volume}). The same exponent carries Watanabe's broader inventory (multi-component crossings, multiplicity, singular fluctuation, prior shifts, and tempered posteriors; Section~\ref{sec:theory:extensions}), and Section~\ref{sec:theory:open} sets out what remains genuinely open. Underneath all of it is one object: the dead direction, the unit vector that is at once Amari's kernel-approaching direction of the Fisher metric and Watanabe's higher-order transversal to the singular set, read by the same KL order. That shared invariant is the bridge, and it is why every result here is legible from both sides at once.
  
\bibliographystyle{abbrvnat}

\begin{thebibliography}{54}
\providecommand{\natexlab}[1]{#1}
\providecommand{\url}[1]{\texttt{#1}}
\expandafter\ifx\csname urlstyle\endcsname\relax
  \providecommand{\doi}[1]{doi: #1}\else
  \providecommand{\doi}{doi: \begingroup \urlstyle{rm}\Url}\fi

\bibitem[Adam et~al.(2025)Adam, Furman, and Hoogland]{AdamFurmanHoogland25}
M.~Adam, Z.~Furman, and J.~Hoogland.
\newblock The loss kernel: A geometric probe for deep learning
  interpretability, 2025.
\newblock URL \url{https://arxiv.org/abs/2509.26537}.

\bibitem[Amari(2016)]{Amari16}
S.-i. Amari.
\newblock \emph{Information Geometry and Its Applications}, volume 194 of
  \emph{Applied Mathematical Sciences}.
\newblock Springer, 2016.
\newblock URL \url{https://link.springer.com/book/10.1007/978-4-431-55978-8}.

\bibitem[Amari et~al.(2006)Amari, Park, and Ozeki]{AmariParkOzeki06}
S.-i. Amari, H.~Park, and T.~Ozeki.
\newblock Singularities affect dynamics of learning in neuromanifolds.
\newblock \emph{Neural Computation}, 18\penalty0 (5):\penalty0 1007--1065,
  2006.
\newblock URL \url{https://doi.org/10.1162/neco.2006.18.5.1007}.

\bibitem[Aoyagi(2024)]{Aoyagi24}
M.~Aoyagi.
\newblock Consideration on the learning efficiency of multiple-layered neural
  networks with linear units.
\newblock \emph{Neural Networks}, 172:\penalty0 106132, 2024.
\newblock URL \url{https://doi.org/10.1016/j.neunet.2024.106132}.

\bibitem[Aoyagi and Watanabe(2005)]{AoyagiWatanabe05}
M.~Aoyagi and S.~Watanabe.
\newblock Stochastic complexities of reduced rank regression in {B}ayesian
  estimation.
\newblock \emph{Neural Networks}, 18\penalty0 (7):\penalty0 924--933, 2005.
\newblock URL \url{https://doi.org/10.1016/j.neunet.2005.03.014}.

\bibitem[Baker et~al.(2025)Baker, Wang, Hoogland, and
  Murfet]{BakerWangHooglandMurfet25}
G.~Baker, G.~Wang, J.~Hoogland, and D.~Murfet.
\newblock Structural inference: Interpreting small language models with
  susceptibilities, 2025.
\newblock URL \url{https://arxiv.org/abs/2504.18274}.

\bibitem[Bushnaq et~al.(2024)Bushnaq, Mendel, Heimersheim, Braun,
  Goldowsky-Dill, H\"anni, Wu, and Hobbhahn]{BushnaqMendel24}
L.~Bushnaq, J.~Mendel, S.~Heimersheim, D.~Braun, N.~Goldowsky-Dill, K.~H\"anni,
  C.~Wu, and M.~Hobbhahn.
\newblock Using degeneracy in the loss landscape for mechanistic
  interpretability, 2024.
\newblock URL \url{https://arxiv.org/abs/2405.10927}.

\bibitem[Carroll(2021)]{Carroll21}
L.~Carroll.
\newblock Phase transitions in neural networks.
\newblock Master's thesis, School of Mathematics and Statistics, The University
  of Melbourne, 2021.
\newblock URL \url{http://therisingsea.org/notes/MSc-Carroll.pdf}.

\bibitem[Chen and Murfet(2025)]{ChenMurfet25}
Z.~Chen and D.~Murfet.
\newblock Modes of sequence models and learning coefficients, 2025.
\newblock URL \url{https://arxiv.org/abs/2504.18048}.

\bibitem[Chen et~al.(2023)Chen, Lau, Mendel, Wei, and
  Murfet]{ChenLauMendelWeiMurfet23}
Z.~Chen, E.~Lau, J.~Mendel, S.~Wei, and D.~Murfet.
\newblock Dynamical versus {B}ayesian phase transitions in a toy model of
  superposition, 2023.
\newblock URL \url{https://arxiv.org/abs/2310.06301}.

\bibitem[de~Br{\'e}bisson and Vincent(2016)]{deBrebissonVincent16}
A.~de~Br{\'e}bisson and P.~Vincent.
\newblock The {Z}-loss: A shift and scale invariant classification loss
  belonging to the spherical family.
\newblock \emph{arXiv preprint arXiv:1604.08859}, 2016.
\newblock URL \url{https://arxiv.org/abs/1604.08859}.

\bibitem[DePavia et~al.(2025)DePavia, Charisopoulos, and
  Willett]{DePaviaCharisopoulosWillett25}
A.~DePavia, V.~Charisopoulos, and R.~Willett.
\newblock How do simple rotations affect the implicit bias of {Adam}?
\newblock \emph{arXiv preprint arXiv:2510.23804}, 2025.
\newblock URL \url{https://arxiv.org/abs/2510.23804}.

\bibitem[Dong et~al.(2021)Dong, Cordonnier, and Loukas]{DongCordonnierLoukas21}
Y.~Dong, J.-B. Cordonnier, and A.~Loukas.
\newblock Attention is not all you need: pure attention loses rank doubly
  exponentially with depth.
\newblock In \emph{International Conference on Machine Learning (ICML)}, 2021.
\newblock URL \url{https://arxiv.org/abs/2103.03404}.

\bibitem[Elhage et~al.(2022)Elhage, Hume, Olsson, Schiefer, Henighan, Kravec,
  Hatfield-Dodds, Lasenby, Drain, Chen, Grosse, McCandlish, Kaplan, Amodei,
  Wattenberg, and Olah]{ElhageHumeOlsson22}
N.~Elhage, T.~Hume, C.~Olsson, N.~Schiefer, T.~Henighan, S.~Kravec,
  Z.~Hatfield-Dodds, R.~Lasenby, D.~Drain, C.~Chen, R.~Grosse, S.~McCandlish,
  J.~Kaplan, D.~Amodei, M.~Wattenberg, and C.~Olah.
\newblock Toy models of superposition.
\newblock \emph{Transformer Circuits Thread}, 2022.
\newblock URL \url{https://transformer-circuits.pub/2022/toy_model/index.html}.

\bibitem[Farrugia-Roberts(2022)]{FarrugiaRoberts22}
M.~Farrugia-Roberts.
\newblock Structural degeneracy in neural networks.
\newblock Master's thesis, School of Computing and Information Systems, The
  University of Melbourne, 2022.
\newblock URL \url{https://far.in.net/mthesis}.

\bibitem[Farrugia-Roberts(2023)]{FarrugiaRoberts23}
M.~Farrugia-Roberts.
\newblock Functional equivalence and path connectivity of reducible hyperbolic
  tangent networks.
\newblock In \emph{Advances in Neural Information Processing Systems 36
  (NeurIPS)}, pages 79502--79517, 2023.
\newblock URL \url{https://arxiv.org/abs/2305.05089}.

\bibitem[Farrugia-Roberts(2024)]{FarrugiaRoberts24}
M.~Farrugia-Roberts.
\newblock Losslessly compressible neural network parameters.
\newblock In \emph{Workshop on Machine Learning and Compression, NeurIPS},
  2024.
\newblock URL \url{https://neurips.cc/virtual/2024/98217}.

\bibitem[Gordon et~al.(2026)Gordon, Baker, Wang, Snell, van Wingerden, and
  Murfet]{GordonBakerWang26}
A.~Gordon, G.~Baker, G.~Wang, W.~Snell, S.~van Wingerden, and D.~Murfet.
\newblock Towards spectroscopy: Susceptibility clusters in language models,
  2026.
\newblock URL \url{https://arxiv.org/abs/2601.12703}.

\bibitem[Hironaka(1964)]{Hironaka64}
H.~Hironaka.
\newblock Resolution of singularities of an algebraic variety over a field of
  characteristic zero.
\newblock \emph{Annals of Mathematics}, 79:\penalty0 109--203 and 205--326,
  1964.
\newblock URL \url{https://www.jstor.org/stable/1970486}.
\newblock Parts I and II, issues 1 and 2.

\bibitem[Hirose(2023)]{Hirose23}
J.~Hirose.
\newblock Blow-up algorithm for sum-of-products polynomials and real log
  canonical thresholds.
\newblock \emph{arXiv preprint arXiv:2303.11619}, 2023.
\newblock URL \url{https://arxiv.org/abs/2303.11619}.

\bibitem[Hoogland et~al.(2025)Hoogland, Wang, Farrugia-Roberts, Carroll, Wei,
  and Murfet]{HooglandWangFarrugiaRoberts24}
J.~Hoogland, G.~Wang, M.~Farrugia-Roberts, L.~Carroll, S.~Wei, and D.~Murfet.
\newblock Loss landscape degeneracy and stagewise development in transformers.
\newblock \emph{Transactions on Machine Learning Research}, 2025.
\newblock URL \url{https://arxiv.org/abs/2402.02364}.

\bibitem[Kim et~al.(2025)Kim, Lee, Park, Oh, Kim, Yoo, Shin, Han, Shin, and
  Yoo]{KimLeeKim25_PeriLN}
J.~Kim, B.~Lee, C.~Park, Y.~Oh, B.~Kim, T.~Yoo, S.~Shin, D.~Han, J.~Shin, and
  K.~M. Yoo.
\newblock {Peri-LN}: Revisiting normalization layer in the transformer
  architecture.
\newblock \emph{arXiv preprint}, 2025.
\newblock URL \url{https://arxiv.org/abs/2502.02732}.
\newblock Names the pre-norm + post-norm pattern ``Peri-LN'' and analyses its
  effect on activation magnitudes (linear vs exponential growth) and gradient
  stability.

\bibitem[Kreer et~al.(2025)Kreer, Wu, Adam, Furman, and
  Hoogland]{KreerWuAdamFurmanHoogland25}
P.~A. Kreer, W.~Wu, M.~Adam, Z.~Furman, and J.~Hoogland.
\newblock {B}ayesian influence functions for hessian-free data attribution,
  2025.
\newblock URL \url{https://arxiv.org/abs/2509.26544}.

\bibitem[Kunin et~al.(2021)Kunin, Sagastuy-Brena, Ganguli, Yamins, and
  Tanaka]{KuninSagastuyBrenaGanguli21}
D.~Kunin, J.~Sagastuy-Brena, S.~Ganguli, D.~L.~K. Yamins, and H.~Tanaka.
\newblock Neural mechanics: Symmetry and broken conservation laws in deep
  learning dynamics.
\newblock In \emph{ICLR}, 2021.
\newblock URL \url{https://arxiv.org/abs/2012.04728}.

\bibitem[Kunstner et~al.(2019)Kunstner, Hennig, and
  Balles]{KunstnerHennigBalles19}
F.~Kunstner, P.~Hennig, and L.~Balles.
\newblock Limitations of the empirical {F}isher approximation for natural
  gradient descent.
\newblock In \emph{NeurIPS}, 2019.
\newblock URL \url{https://arxiv.org/abs/1905.12558}.

\bibitem[Lau et~al.(2025)Lau, Furman, Wang, Murfet, and
  Wei]{LauFurmanWangMurfetWei25}
E.~Lau, Z.~Furman, G.~Wang, D.~Murfet, and S.~Wei.
\newblock The local learning coefficient: A singularity-aware complexity
  measure.
\newblock In \emph{AISTATS}, 2025.
\newblock URL \url{https://proceedings.mlr.press/v258/lau25a.html}.

\bibitem[Lau and Su(2026)]{LauSu26}
T.~T.-K. Lau and W.~J. Su.
\newblock Symmetry-compatible principle for optimizer design: Embeddings, {LM}
  heads, {SwiGLU MLPs}, and {MoE} routers, 2026.
\newblock URL \url{https://arxiv.org/abs/2605.18106}.

\bibitem[Lee et~al.(2025)Lee, Smith, Adam, and
  Hoogland]{LeeSmithAdamHoogland25}
J.~H. Lee, M.~Smith, M.~Adam, and J.~Hoogland.
\newblock Influence dynamics and stagewise data attribution, 2025.
\newblock URL \url{https://arxiv.org/abs/2510.12071}.

\bibitem[Martens and Grosse(2015)]{MartensGrosse15}
J.~Martens and R.~Grosse.
\newblock Optimizing neural networks with {Kronecker}-factored approximate
  curvature.
\newblock In \emph{ICML}, 2015.
\newblock URL \url{https://arxiv.org/abs/1503.05671}.

\bibitem[Murfet and Troiani(2025)]{MurfetTroiani25}
D.~Murfet and W.~Troiani.
\newblock Programs as singularities, 2025.
\newblock URL \url{https://arxiv.org/abs/2504.08075}.

\bibitem[Nanda et~al.(2023)Nanda, Chan, Lieberum, Smith, and
  Steinhardt]{NandaChanLieberum23}
N.~Nanda, L.~Chan, T.~Lieberum, J.~Smith, and J.~Steinhardt.
\newblock Progress measures for grokking via mechanistic interpretability.
\newblock In \emph{ICLR}, 2023.
\newblock URL \url{https://arxiv.org/abs/2301.05217}.

\bibitem[Noci et~al.(2022)Noci, Anagnostidis, Biggio, Orvieto, Singh, and
  Lucchi]{NociAnagnostidisBiggio22}
L.~Noci, S.~Anagnostidis, L.~Biggio, A.~Orvieto, S.~P. Singh, and A.~Lucchi.
\newblock Signal propagation in transformers: Theoretical perspectives and the
  role of rank collapse.
\newblock In \emph{Advances in Neural Information Processing Systems
  (NeurIPS)}, 2022.
\newblock URL \url{https://arxiv.org/abs/2206.03126}.

\bibitem[Papyan(2020)]{Papyan20}
V.~Papyan.
\newblock Traces of class/cross-class structure pervade deep learning spectra.
\newblock \emph{JMLR}, 21\penalty0 (252):\penalty0 1--64, 2020.
\newblock URL \url{https://jmlr.org/papers/volume21/20-933/20-933.pdf}.

\bibitem[Papyan et~al.(2020)Papyan, Han, and Donoho]{PapyanHanDonoho20}
V.~Papyan, X.~Y. Han, and D.~L. Donoho.
\newblock Prevalence of neural collapse during the terminal phase of deep
  learning training.
\newblock \emph{Proceedings of the National Academy of Sciences}, 117\penalty0
  (40):\penalty0 24652--24663, 2020.
\newblock URL \url{https://doi.org/10.1073/pnas.2015509117}.

\bibitem[Pesme et~al.(2021)Pesme, Pillaud-Vivien, and Flammarion]{Pesme2021}
S.~Pesme, L.~Pillaud-Vivien, and N.~Flammarion.
\newblock Implicit bias of {SGD} for diagonal linear networks: A provable
  benefit of stochasticity.
\newblock In \emph{NeurIPS}, 2021.
\newblock URL \url{https://arxiv.org/abs/2106.09524}.

\bibitem[Power et~al.(2022)Power, Burda, Edwards, Babuschkin, and
  Misra]{PowerBurda22}
A.~Power, Y.~Burda, H.~Edwards, I.~Babuschkin, and V.~Misra.
\newblock Grokking: Generalization beyond overfitting on small algorithmic
  datasets.
\newblock \emph{arXiv:2201.02177}, 2022.

\bibitem[Prieto et~al.(2025)Prieto, Barsbey, Mediano, and
  Birdal]{PrietoBarsbey25}
L.~Prieto, M.~Barsbey, P.~A.~M. Mediano, and T.~Birdal.
\newblock Grokking at the edge of numerical stability.
\newblock In \emph{ICLR}, 2025.
\newblock URL \url{https://arxiv.org/abs/2501.04697}.

\bibitem[Shazeer et~al.(2018)Shazeer, Cheng, Parmar, Tran, Vaswani,
  Koanantakool, Hawkins, Lee, Hong, Young, Sepassi, and
  Hechtman]{ShazeerMesh18}
N.~Shazeer, Y.~Cheng, N.~Parmar, D.~Tran, A.~Vaswani, P.~Koanantakool,
  P.~Hawkins, H.~Lee, M.~Hong, C.~Young, R.~Sepassi, and B.~Hechtman.
\newblock Mesh-{TensorFlow}: Deep learning for supercomputers.
\newblock In \emph{Advances in Neural Information Processing Systems
  (NeurIPS)}, 2018.
\newblock URL \url{https://arxiv.org/abs/1811.02084}.

\bibitem[Shirodkar(2026)]{shirodkar2026measuringdeaddirectionsdecomposing}
T.~P. Shirodkar.
\newblock Measuring dead directions: Decomposing and classifying singular
  structure off canonical alignment, 2026.
\newblock URL \url{https://arxiv.org/abs/2607.00603}.

\bibitem[Shirodkar and
  Narayanan(2026)]{shirodkar2026algebraicdeaddirectionslayernorm}
T.~P. Shirodkar and P.~J. Narayanan.
\newblock Parameter-readable kernel directions in {LayerNorm} transformers:
  {Fisher} suppression and a forward-pass-only diagnostic at {LLM} scale, 2026.
\newblock URL \url{https://arxiv.org/abs/2606.19491}.

\bibitem[Sun et~al.(2024)Sun, Chen, Kolter, and Liu]{SunMassiveActivations24}
M.~Sun, X.~Chen, J.~Z. Kolter, and Z.~Liu.
\newblock Massive activations in large language models.
\newblock In \emph{COLM}, 2024.
\newblock URL \url{https://arxiv.org/abs/2402.17762}.

\bibitem[Tanaka and Kunin(2021)]{TanakaKunin21}
H.~Tanaka and D.~Kunin.
\newblock Noether's learning dynamics: Role of symmetry breaking in neural
  networks.
\newblock In \emph{NeurIPS}, 2021.
\newblock URL \url{https://arxiv.org/abs/2105.02716}.

\bibitem[Terin(2026)]{Terin26}
R.~C. Terin.
\newblock Scale redundancy and soft gauge fixing in positively homogeneous
  neural networks, 2026.
\newblock URL \url{https://arxiv.org/abs/2602.14729}.

\bibitem[Thilak et~al.(2022)Thilak, Littwin, Zhai, Saremi, Paiss, and
  Susskind]{ThilakLittwin22}
V.~Thilak, E.~Littwin, S.~Zhai, O.~Saremi, R.~Paiss, and J.~Susskind.
\newblock The slingshot mechanism: An empirical study of adaptive optimizers
  and the grokking phenomenon.
\newblock \emph{arXiv:2206.04817}, 2022.
\newblock URL \url{https://arxiv.org/abs/2206.04817}.

\bibitem[Transtrum et~al.(2011)Transtrum, Machta, and
  Sethna]{TranstrumMachtaSethna11}
M.~K. Transtrum, B.~B. Machta, and J.~P. Sethna.
\newblock Geometry of nonlinear least squares with applications to sloppy
  models and optimization.
\newblock \emph{Physical Review E}, 83:\penalty0 036701, 2011.
\newblock arXiv:1010.1449.

\bibitem[Transtrum et~al.(2015)Transtrum, Machta, Brown, Daniels, Myers, and
  Sethna]{TranstrumMachtaBrown15}
M.~K. Transtrum, B.~B. Machta, K.~S. Brown, B.~C. Daniels, C.~R. Myers, and
  J.~P. Sethna.
\newblock Perspective: Sloppiness and emergent theories in physics, biology,
  and beyond.
\newblock \emph{The Journal of Chemical Physics}, 143:\penalty0 010901, 2015.
\newblock arXiv:1501.07668.

\bibitem[Urdshals et~al.(2025)Urdshals, Lau, Hoogland, van Wingerden, and
  Murfet]{Urdshals25Compressibility}
E.~Urdshals, E.~Lau, J.~Hoogland, S.~van Wingerden, and D.~Murfet.
\newblock Compressibility measures complexity: Minimum description length meets
  singular learning theory, 2025.
\newblock URL \url{https://arxiv.org/abs/2510.12077}.

\bibitem[Wang et~al.(2025)Wang, Hoogland, van Wingerden, Furman, and
  Murfet]{WangHoogland24}
G.~Wang, J.~Hoogland, S.~van Wingerden, Z.~Furman, and D.~Murfet.
\newblock Differentiation and specialization of attention heads via the refined
  local learning coefficient.
\newblock In \emph{International Conference on Learning Representations
  (ICLR)}, 2025.
\newblock URL \url{https://arxiv.org/abs/2410.02984}.
\newblock Spotlight.

\bibitem[Watanabe(2007)]{Watanabe07}
S.~Watanabe.
\newblock Almost all learning machines are singular.
\newblock In \emph{IEEE Symposium on Foundations of Computational
  Intelligence}, pages 383--388, 2007.
\newblock URL \url{https://ieeexplore.ieee.org/document/4233934}.

\bibitem[Watanabe(2009)]{Watanabe09}
S.~Watanabe.
\newblock \emph{Algebraic Geometry and Statistical Learning Theory}.
\newblock Cambridge University Press, 2009.
\newblock URL \url{https://doi.org/10.1017/CBO9780511800474}.

\bibitem[Watanabe(2018)]{Watanabe18}
S.~Watanabe.
\newblock \emph{Mathematical Theory of {B}ayesian Statistics}.
\newblock CRC Press, 2018.
\newblock URL \url{https://www.routledge.com/9781482238068}.

\bibitem[Wei et~al.(2023)Wei, Murfet, Gong, Li, Gell-Redman, and
  Quella]{WeiMurfet22}
S.~Wei, D.~Murfet, M.~Gong, H.~Li, J.~Gell-Redman, and T.~Quella.
\newblock Deep learning is singular, and that's good.
\newblock \emph{IEEE Transactions on Neural Networks and Learning Systems},
  34\penalty0 (12):\penalty0 10473--10486, 2023.
\newblock URL \url{https://ieeexplore.ieee.org/document/9812468}.

\bibitem[Zhao et~al.(2025)Zhao, Walters, and Yu]{ZhaoWaltersYu25}
B.~Zhao, R.~Walters, and R.~Yu.
\newblock Symmetry in neural network parameter spaces, 2025.
\newblock URL \url{https://arxiv.org/abs/2506.13018}.

\bibitem[Zoph et~al.(2022)Zoph, Bello, Kumar, Du, Huang, Dean, Shazeer, and
  Fedus]{ZophSTMoE22}
B.~Zoph, I.~Bello, S.~Kumar, N.~Du, Y.~Huang, J.~Dean, N.~Shazeer, and
  W.~Fedus.
\newblock {ST-MoE}: Designing stable and transferable sparse expert models.
\newblock \emph{arXiv preprint arXiv:2202.08906}, 2022.
\newblock URL \url{https://arxiv.org/abs/2202.08906}.

\end{thebibliography}

\clearpage
\appendix
\phantomsection
\addcontentsline{toc}{part}{Appendices}
\part*{Appendices}
\noindent The appendices collect material deferred from the body: the architectural rate catalogue, holding the per-primitive instantiations that carry the canonical ladder without new phenomena, the non-canonical boundary, and the LayerNorm finite-$t$ crossover (Appendix~\ref{app:theory:arch_catalogue}); and the controlled parametric validations that anchor the rate predictions on small testbeds where every condition is independently controllable (Appendix~\ref{app:theory:parametric_validations}); and the statements we have proved in Lean~4, each with its declarations and the hypotheses it takes as inputs (Appendix~\ref{app:theory:machine_checked}). Proofs appear inline with their theorems in the body.

\clearpage

\section{Architectural rate catalogue}
\label{app:theory:arch_catalogue}

This catalogue collects the architectural instantiations summarised in Table~\ref{tab:arch_landscape} but deferred from Section~\ref{sec:theory:architectural}: the primitives that carry the canonical rate ladder through composition without introducing new behaviour, the non-canonical boundary where the ladder ceases to hold, the SwiGLU block, and the finite-$t$ crossover of the LayerNorm rate. Each is stated with the same structure used in the body, a per-block KL order with its singular-geometry consequence, and each predicted rate is the one checked by the freeze-probe of Appendix~\ref{app:theory:arch_freeze_probe} and plotted in Figure~\ref{fig:arch_rate_roundup}. All but one feed the composition theorem (Section~\ref{sec:theory:composition}) directly. SwiGLU is the exception and is collected here for its per-block order alone: its dead-direction map is quadratic in the dead coordinate, so it fails that theorem's scalar-transfer hypothesis and its blocks do not compose additively (Remark~\ref{rem:swiglu_composition}).

\subsection{Rectangular widths}
\label{sec:theory:arch:rect}

\label{app:bridge_rect}

Theorem~\ref{thm:bridge} and its cross-entropy companion Theorem~\ref{thm:bridge_ce} assume square hidden dimensions $h_\ell = h$ for all $\ell$. Real networks have per-layer widths that differ: embeddings ($d_\text{vocab} \to d_\text{model}$), MLP up/down projections ($d \to 4d \to d$), classifier heads ($d \to C$). This subsection extends the bridge theorem to rectangular widths via a reduction to the square case restricted to a nested chain of narrow subspaces. For (P1) the chain may take any position; for (P2) and (P3) the elementwise nonlinearity ties the chain to the coordinate axes, so the theorem requires coordinate-aligned isometries, the same restriction Proposition~\ref{prop:bridge_linear_rot} and Proposition~\ref{prop:bridge_nonlinear_rot_negative} document for the square case. The new work is a rectangular variant of the matrix-product singular-value lemma; the backward-chain argument of Theorem~\ref{thm:bridge} then transfers.

\paragraph{Setup.}
Let $W_\ell \in \mathbb{R}^{h_\ell \times h_{\ell-1}}$ for $\ell = 1, \ldots, L$, with the widths $h_0, h_1, \ldots, h_L$ arbitrary positive integers. Let $h_* := \min_{\ell} h_\ell$ be the narrowest slot. Since the composed map $M = W_L \cdots W_1$ has rank at most $h_*$, the natural ``rank-deficient-by-one'' singular configuration has $\mathrm{rank}(M^*) = h_* - 1$. In this regime the dead direction belongs to an $h_*$-dimensional chain of subspaces that threads through the whole network.

Concretely, an \emph{isometry chain} is a sequence of linear isometries $\iota_\ell : \mathbb{R}^{h_*} \hookrightarrow \mathbb{R}^{h_\ell}$ for $\ell = 0, 1, \ldots, L$ (so $\iota_\ell^\top \iota_\ell = I_{h_*}$; $\iota_0$ embeds the narrow chain into the input space) compatible with $\theta^*$ in the sense that
\[
W_\ell^* \, \iota_{\ell-1} \;=\; \iota_\ell \cdot \mathrm{diag}(1,1,\ldots,1,0) \qquad (\ell = 1, \ldots, L),
\]
where the diagonal acts on $\mathbb{R}^{h_*}$ and zeros out the last ($h_*$-th) coordinate. Equivalently, $V_\ell := \iota_\ell(\mathbb{R}^{h_*})$ is a nested-rank chain of subspaces of $\mathbb{R}^{h_\ell}$, invariant under $W_\ell^*$, on which $W_\ell^*$ acts as the square-case $\mathrm{diag}(1, \ldots, 1, 0)$ in the isometry-chosen bases. The dead direction at layer $\ell$ is
\[
u_\ell := \iota_\ell(e_{h_*}) \in \mathbb{R}^{h_\ell}, \qquad (u_\ell \text{ is a unit vector}; u_\ell \in V_\ell).
\]
The dead-aligned symmetric approach perturbs each layer by $t$ in the dead direction:
\[
W_\ell(t) \;:=\; W_\ell^* \;+\; t \, u_\ell u_{\ell-1}^\top \qquad (\ell = 1, \ldots, L).
\]
Call the chain \emph{coordinate-aligned} when every $\iota_\ell$ maps standard basis vectors to standard basis vectors, so each $V_\ell$ is a coordinate subspace and every $u_\ell$ is a coordinate vector.

We also assume a genericity condition on the orthogonal complement:

\begin{assumption}[Non-degenerate complement action]
\label{ass:rect_complement}
The action of $W_\ell^*$ on $V_{\ell-1}^\perp$ is non-degenerate in the sense that the map $W_\ell^* : V_{\ell-1}^\perp \to \mathbb{R}^{h_\ell}$ has trivial kernel and image disjoint from the dead direction $\mathbb{R} u_\ell$. In particular, after the isometry-adapted basis change, the non-narrow block of $W_\ell^*$ is full column rank and contributes $\Theta(1)$ singular values.
\end{assumption}

Assumption~\ref{ass:rect_complement} rules out the case where the orthogonal complement carries additional hidden rank deficiencies that would compete with the narrow-chain rate. Its satisfiability is a dimension count: injectivity plus the missed line require $h_{\ell-1} - h_* < h_\ell$ at every layer, so any layer whose width drops by $h_*$ or more cannot satisfy it. Two of this subsection's motivating architectures belong to that excluded class (an MLP down projection $4d \to d$ at $h_* = d$ requires $3d < d$; a $64 \to 10$ classifier head at $h_* = 10$ requires $54 < 10$), so the $\lambda_{\min}$ corollaries below do not reach them, while the dead-entry statement of Theorem~\ref{thm:bridge_rect}, which does not use the assumption, still does. On width profiles that pass the count, a random linear map between the complements satisfies the assumption with probability one.

\begin{lemma}[Rectangular partial-product action on the narrow chain]
\label{lem:rect_product}
Under the isometry-chain setup, for any $\ell \in \{0, 1, \ldots, L-1\}$:
\[
W_L(t) \, W_{L-1}(t) \cdots W_{\ell+1}(t) \cdot \iota_\ell \;=\; \iota_L \cdot \mathrm{diag}(1, \ldots, 1, t^{L-\ell}).
\]
This statement does not require Assumption~\ref{ass:rect_complement}; the narrow-chain action is self-contained.
\end{lemma}

\begin{proof}
Compute the partial product's action on $\iota_\ell(\mathbb{R}^{h_*})$. For any $v \in \mathbb{R}^{h_*}$:
\begin{align*}
W_\ell(t) \, \iota_{\ell-1}(v)
&= W_\ell^* \iota_{\ell-1}(v) + t\, u_\ell u_{\ell-1}^\top \iota_{\ell-1}(v) \\
&= \iota_\ell(D v) + t\, u_\ell \cdot (u_{\ell-1}^\top \iota_{\ell-1}(v)) \\
&= \iota_\ell(D v) + t\, \iota_\ell(e_{h_*}) \cdot v_{h_*} \\
&= \iota_\ell\bigl(D v + t\, v_{h_*} e_{h_*}\bigr)
\end{align*}
where $D = \mathrm{diag}(1, \ldots, 1, 0)$ and we used $u_{\ell-1} = \iota_{\ell-1}(e_{h_*})$ together with $\iota_{\ell-1}^\top \iota_{\ell-1} = I$ to extract $v_{h_*} = u_{\ell-1}^\top \iota_{\ell-1}(v)$. Observe $D v + t v_{h_*} e_{h_*} = \mathrm{diag}(1, \ldots, 1, t) \cdot v$. Therefore
\[
W_\ell(t) \, \iota_{\ell-1} \;=\; \iota_\ell \cdot \mathrm{diag}(1, \ldots, 1, t).
\]
Iterating from $\ell+1$ to $L$ composes $L - \ell$ copies of this relation, producing $\iota_L \cdot \mathrm{diag}(1, \ldots, 1, t)^{L-\ell} = \iota_L \cdot \mathrm{diag}(1, \ldots, 1, t^{L-\ell})$.
\end{proof}

\begin{corollary}[Smallest singular value under composed complement non-degeneracy]
\label{cor:rect_product_sigma_min}
Write $P(t) := W_L(t) \cdots W_{\ell+1}(t)$ and $P_0 := P(0)$. Assume the \emph{composed} complement non-degeneracy at layer $\ell$: (i) $P_0$ restricted to the orthogonal complement of $\mathbb{R} u_\ell$ has smallest singular value bounded below by some $c > 0$, and (ii) the dead output direction is unreachable at $t = 0$, $\iota_L(e_{h_*}) \notin \mathrm{im}(P_0)$. Then
\[
\sigma_{\min}\!\bigl(P(t)\bigr) \;=\; t^{L-\ell} \cdot \Theta(1).
\]
The per-layer Assumption~\ref{ass:rect_complement} does not suffice: per-layer injectivity and per-layer missed lines do not compose, and on a shrinking chain the complement's $\Theta(1)$ image can occupy the output direction the narrow chain would otherwise vacate, leaving $\sigma_{\min} = \Theta(1)$ (widths $2 \to 3 \to 2$ at $\ell = 1$ measure $\sigma_{\min} = 0.54$ at every $t \le 10^{-1}$ while satisfying the per-layer assumption). On square canonical chains conditions (i) and (ii) hold with $c = 1$.
\end{corollary}

\begin{proof}
Upper bound: $P(t)\, u_\ell = t^{L-\ell} \iota_L(e_{h_*})$ by Lemma~\ref{lem:rect_product}. Lower bound: let $\delta := \mathrm{dist}(\iota_L(e_{h_*}), \mathrm{im}(P_0)) > 0$ by (ii), and take a unit $v = a\, u_\ell + b\, w$ with $w \perp u_\ell$ unit. Then $P(t) v = a\, t^{L-\ell} \iota_L(e_{h_*}) + b\, P_0 w + O(t)\,b$. If $|a| \ge 1/2$, the component of $P(t) v$ orthogonal to $\mathrm{im}(P_0)$ has norm at least $|a|\, t^{L-\ell} \delta - O(t^{L-\ell+1})$, which is $\Theta(t^{L-\ell})$. If $|a| < 1/2$ then $|b| \ge \sqrt{3}/2$ and $\|P(t) v\| \ge |b|\, c - |a|\, t^{L-\ell} - O(t) = \Theta(1)$. The minimum over unit $v$ is therefore $\Theta(t^{L-\ell})$.
\end{proof}

\begin{theorem}[Multi-Layer K-FAC G-factor Bridge, rectangular weights]
\label{thm:bridge_rect}
Under the isometry-chain setup, with the chain \emph{coordinate-aligned} for activation classes (P2) and (P3) (arbitrary isometries are admitted for (P1), whose rotation covariance Proposition~\ref{prop:bridge_linear_rot} establishes), for the symmetric dead-aligned approach $\theta(t)$ with rank-$(h_* - 1)$ singular configuration $\theta^*$, the dead-direction entry of the G-factor satisfies:
\begin{enumerate}
\item[(a)] For $\ell \in \{1, \ldots, L-1\}$,
\[
(G_\ell(\theta(t)))_{u_\ell u_\ell} \;:=\; u_\ell^\top G_\ell(\theta(t))\, u_\ell \;=\; C_\ell^\mathrm{rect} \cdot t^{2(L-\ell)} \cdot \bigl(1 + r_\ell^G(t)\bigr),
\]
with activation-class-dependent corrections $r_\ell^G(t)$ matching Theorem~\ref{thm:bridge}.
\item[(b)] $(G_L(\theta(t)))_{u_L u_L} = \Theta(1)$, with the same base-case structure as Theorem~\ref{thm:bridge} (MSE) or Theorem~\ref{thm:bridge_ce} (cross-entropy under expected Fisher + Assumption~\ref{ass:nondeg_pxy}).
\item[(c)] Provided the complement is invariant, $W_\ell^*(V_{\ell-1}^\perp) \subseteq V_\ell^\perp$ for every $\ell$ (automatic when $V_0^\perp = \{0\}$), the shallowest-layer rate $2(L-1)$ matches Theorem~\ref{thm:fisher_decay} at KL order $k = L$. Without the invariance, a complement leak enters the composed map at first order in $t$, the KL order drops to $1$, and the approach is not a dead direction at all (Definition~\ref{def:dead_direction} requires $k \ge 2$); a bottleneck chain $3 \to 2 \to 3$ with $W_1^*$ leaking $V_0^\perp$ into the chain measures KL slope $2.000$ against the invariant chain's $2L = 4$, while parts (a) and (b) are unaffected on the same instance.
\end{enumerate}
\textbf{Note on hypotheses.} This statement does \emph{not} require Assumption~\ref{ass:rect_complement}; the dead-direction (transversal) rate follows from the narrow-chain structure alone. The assumption is needed only to promote this to a statement about $\lambda_{\min}$; see Corollary~\ref{cor:rect_lambda_min} below. The coordinate-alignment hypothesis for (P2)/(P3) is load-bearing: with a rotated isometry chain the measured per-layer rates collapse (fitted $(0.35, 0.11, 0.74, 0.00)$ against the predicted $(6, 4, 2, 0)$ for ReLU at a $0.1$-radian rotation, and similarly for GeLU), while coordinate-aligned rectangular chains reproduce the prediction across all three classes; the rotated instance is exactly the configuration of Proposition~\ref{prop:bridge_nonlinear_rot_negative}.
\end{theorem}

\begin{proof}
Decompose each $\mathbb{R}^{h_\ell}$ as $V_\ell \oplus V_\ell^\perp$ where $V_\ell = \iota_\ell(\mathbb{R}^{h_*})$. The backward chain's action restricted to the dead direction $u_\ell \in V_\ell$ is self-contained: by Lemma~\ref{lem:rect_product} (which holds unconditionally, without Assumption~\ref{ass:rect_complement}), the partial product $W_L(t) \cdots W_{\ell+1}(t)$ maps $u_\ell$ to $t^{L-\ell} u_L$, and the backward operator $(W_L \cdots W_{\ell+1})^\top$ correspondingly maps the output dead-direction to $u_\ell$ with gain $t^{L-\ell}$. This holds whether or not $V_\ell^\perp$-to-$V_\ell^\perp$ action is well-behaved; the computation of $u_\ell^\top \delta^{(\ell)}$ involves only the narrow chain. The machine-checked form of this reduction is wider still: it composes the canonical narrow block with arbitrary maps on the complements, linear or not, with depth-varying widths, and the dead ladder, the non-dead base, and the backward moment carry over unchanged.

Explicitly, by the same canonical-basis argument as Lemma~\ref{lem:backward-dead-sub} (whose derivation uses only the narrow-block structure of $W_{\ell+1}^\top$ and the chain rule), in the isometry-adapted basis
\[
u_\ell^\top \delta^{(\ell)} \;=\; t \cdot \phi'(a_\ell^{(u)}) \cdot u_{\ell+1}^\top \delta^{(\ell+1)},
\]
where the step from the elementwise $\phi'$ diagonal to the scalar factor $\phi'(a_\ell^{(u)})$ requires $\mathrm{diag}(\phi'(a_\ell))$ to preserve $\mathrm{span}(u_\ell)$: that holds for every isometry under (P1) ($\phi' \equiv 1$), and for (P2)/(P3) exactly when $u_\ell$ is a coordinate vector, which is what the coordinate-alignment hypothesis supplies,
with base case given by either Lemma~\ref{lem:backward-dead-sub} (MSE) or Lemma~\ref{lem:backward-dead-ce} (CE). Induction gives $\expect[(u_\ell^\top \delta^{(\ell)})^2] = \Theta(t^{2(L-\ell)})$ for $\ell < L$ and $\Theta(1)$ for $\ell = L$. Activation-class corrections carry over since they depend only on hidden-layer nonlinearities, not on widths.
\end{proof}

\begin{corollary}[$\lambda_{\min}$ statement under complement genericity]
\label{cor:rect_lambda_min}
Assume additionally Assumption~\ref{ass:rect_complement} and that no layer above $\ell$ is narrower than layer $\ell$: $h_\ell \le h_m$ for every $m > \ell$ (and $h_\ell \le$ the output width). Then the non-narrow eigenvalues of $G_\ell(\theta(t))$ are $\Theta(1)$ in $t$, and the smallest eigenvalue equals the dead-direction entry at leading order:
\[
\lambda_{\min}(G_\ell(\theta(t))) \;=\; (G_\ell(\theta(t)))_{u_\ell u_\ell} \;=\; \Theta(t^{2(L-\ell)}) \quad (\ell < L), \qquad \Theta(1) \quad (\ell = L).
\]
The width condition is not removable: the backward operator into layer $\ell$ factors through every layer above, so $\mathrm{rank}(G_\ell) \le \min_{m > \ell} h_m$, and any narrower layer above forces exact zero eigenvalues at every $t$ regardless of Assumption~\ref{ass:rect_complement}. The failure is not a wrong exponent but a wrong spectrum: on the shrinking chain $2 \to 3 \to 2$ the spectrum of $G_1$ is $\{0, \Theta(1), \Theta(1)\}$ at every $t$, the exact kernel absorbs the dead direction's decay, and no eigenvalue decays at the dead rate even though the dead-direction \emph{entry} still measures $\Theta(t^2)$. On such chains the rate appears in the entry $u_\ell^\top G_\ell u_\ell$ and in no eigenvalue.
\end{corollary}

\begin{proof}
By Lemma~\ref{lem:integral-reduction-sub} (Schur-form integral reduction: the dead-direction Schur complement at the dead row equals the dead-direction entry at leading order, since the non-dead block is $\succ 0$ at $\Theta(1)$ and the cross-row entries do not exceed the Schur-cancellation order), the dead-direction eigenvalue of $G_\ell$ is $\Theta(t^{2(L-\ell)})$ in the isometry-adapted basis. The isometry-chain extension preserves the dead-row / non-dead-block structure of the parent lemma since the chain is a direct-sum extension of the canonical narrow block with a non-degenerate complement (the isometry projection $\Pi$ acts as identity on the dead-direction coordinate and rotates the non-dead complement). Under Assumption~\ref{ass:rect_complement}, the complement block has all $\Theta(1)$ eigenvalues; combined with the narrow block's $h_* - 1$ non-dead eigenvalues at $\Theta(1)$ and the dead-direction Schur-eigenvalue at $\Theta(t^{2(L-\ell)})$, the minimum is attained at the dead-direction eigenvalue for $t$ small enough, yielding the stated $\lambda_{\min}$ equality.
\end{proof}

\paragraph{Joint scope.} The combination of rectangular widths with cross-entropy is closed (Remark~\ref{rem:bridge_ce_rect_joint}); the formal joint statement covers MLP classifiers without residuals whose class posteriors stay non-degenerate in the sense of Assumption~\ref{ass:nondeg_pxy}. The neural-collapse terminal phase falls outside that assumption (posteriors approach one-hot and the restricted Hessian eigenvalue vanishes, the regime Corollary~\ref{cor:emp_vs_exp_ce}'s consistency note identifies), so collapse-phase classifiers are not covered. The residual extension (Theorem~\ref{thm:bridge_res}, \S\ref{app:bridge_res}) adds transformer FFN blocks, classifier heads attached through residuals, and ResNet-style CNNs to the in-scope list.

\paragraph{Beyond Approach A.} Theorem~\ref{thm:bridge_rect} uses Approach A: dead-direction rank-star $h_*$ at most $\min_\ell h_\ell$, with an isometry chain threading through every layer along a shared narrow subspace of dimension $h_*$. A more general treatment allowing the rank-star to exceed $h_*$ at some layer (the dead direction maps to a lower-dimensional image at a bottleneck), or admitting different dead directions per layer (rather than a shared chain), is open. Approach A covers the architectural slice where every layer carries the same dead-direction signature; the open generalisations are needed when bottlenecks force the dead direction to compose with rank-1 projections inside the chain, or when the dead direction realigns layer by layer (the multi-direction chunk \S\ref{app:bridge_multi} handles the case of multiple parallel dead directions but assumes consistent indexing across layers, which is the same shared-chain structure).

\subsection{Multi-direction singularities and asymmetric approaches}
\label{sec:theory:arch:multi}

\label{app:bridge_multi}

All previous bridge theorems assume a single dead direction perturbed at uniform rate $t$ across all layers. Real singularities exhibit higher rank deficit (multiple dead directions) and the approach trajectory may scale different layers' perturbations with different powers of the same scalar parameter $t$ (i.e., layer $\ell$ is perturbed by $t^{p_\ell}$ for layer-dependent $p_\ell$, not by separate time variables). The structural classification of multiple dead directions in single-hidden-layer (tanh) networks via the \emph{neural-network rank} of a parameter (the minimum hidden width needed to represent $f_w$) is due to \citet{FarrugiaRoberts22,FarrugiaRoberts24}; the present subsection extends that count-of-degenerate-directions framing to a rate per direction in the K-FAC G-factor. This subsection unifies three generalizations:
\begin{itemize}
\item \emph{Arbitrary rank deficit}: the singular configuration has $\mathrm{rank}(W_\ell^*)$ deficient by $m \ge 1$ rather than just $1$, giving $m$ dead directions.
\item \emph{Non-symmetric approach}: each weight is perturbed by $t^{p_\ell}$ with possibly layer-varying exponent $p_\ell \ge 1$, not just $t^1$.
\item \emph{Per-direction KL orders}: the $m$ dead directions may have \emph{different} per-layer exponent patterns $p_\ell^{(i)}$, giving each direction its own effective KL order.
\end{itemize}
These fold into a single multi-index bookkeeping. Each dead direction $i$ contributes its own eigenvalue to $G_\ell$'s spectrum at its own rate $2 \sum_{\ell' > \ell} p_{\ell'}^{(i)}$, and the directions do not interfere. The per-direction entries $(G_\ell)_{u_\ell^{(i)} u_\ell^{(i)}}$ are read by projecting onto the known dead direction $u_\ell^{(i)}$ at each $t$-checkpoint, a parametric freeze-probe rather than a rank-sweep on a trained-network spectrum.

\paragraph{Setup.}
Let $W_\ell \in \mathbb{R}^{h_\ell \times h_{\ell-1}}$ be as in \S\ref{app:bridge_rect}. Fix $m \ge 1$ (the rank deficit). The singular configuration $\theta^*$ has rank deficiency of order exactly $m$ at every layer, meaning there are $m$ canonically-aligned dead directions
\[
u^{(i)}_\ell = e_{h_\ell - m + i} \in \mathbb{R}^{h_\ell}, \qquad i = 1, \ldots, m, \quad \ell = 0, 1, \ldots, L,
\]
(we index the last $m$ coordinates at each layer as the dead coordinates, consistent with the G1 isometry-chain construction of \S\ref{app:bridge_rect}; this is the \emph{consistent indexing} requirement of Assumption~\ref{ass:multi_consistent_indexing} below, which makes direction $i$ at layer $\ell$ correspond to direction $i$ at layer $\ell'$ for all $\ell, \ell'$). Each $W_\ell^*$, in the canonical basis, has the block structure
\[
W_\ell^* \;=\; \begin{pmatrix} A_\ell & 0 \\ 0 & 0 \end{pmatrix},
\]
where $A_\ell$ acts on the non-dead coordinates (rank $\min(h_\ell, h_{\ell-1}) - m$) and the $m \times m$ dead block is zero. In particular, $W_\ell^*\, u^{(i)}_{\ell-1} = 0$ for each $i$. The perturbation is
\[
W_\ell(t) \;=\; W_\ell^* \;+\; \sum_{i=1}^m t^{p_\ell^{(i)}}\, u_\ell^{(i)} (u_{\ell-1}^{(i)})^\top,
\]
where $p_\ell^{(i)} \ge 1$ (real-valued; not necessarily integer; the proof's coordinate-wise scaling extends mechanically to real exponents, while the parametric validation covers only integer values $p \in \{1, 2, 3\}$), possibly varying with both direction $i$ and layer $\ell$. The single-direction symmetric case $m = 1, p_\ell^{(1)} = 1$ recovers Theorem~\ref{thm:bridge}.

\begin{assumption}[Consistent dead-direction indexing across layers]
\label{ass:multi_consistent_indexing}
The dead direction with index $i$ at layer $\ell$ corresponds to the dead direction with the same index $i$ at every other layer $\ell'$, in the sense that the canonical-basis labels $u_\ell^{(i)} = e_{h_\ell - m + i}$ are aligned across $\ell$. Equivalently, the inter-layer embedding maps preserve the dead-direction subspace's basis labelling, analogous to the nested-isometry-chain hypothesis used in Theorem~\ref{thm:bridge_rect} for rectangular widths. Without this assumption, the per-direction multi-index bookkeeping below does not factor and additional per-layer alignment factors appear in the rate formula.
\end{assumption}

For the cross-entropy base case (analogous to \S\ref{app:bridge_ce}), we use a multi-direction version of Assumption~\ref{ass:nondeg_pxy}: the data-averaged softmax Hessian $\expect_x[H(x; \theta(t))]$ is uniformly positive-definite on the projection of $\mathrm{span}\{u_L^{(i)}\}_{i=1}^m$ onto $\mathbf{1}^\perp$ (since $\mathbf{1} \in \ker H$ algebraically). Explicitly, there exists $c_0 > 0$ and a neighborhood $\mathcal{U}$ of $t = 0$ such that for all $t \in \mathcal{U}$ and all unit $v \in \mathrm{span}\{\Pi_{\{\mathbf{1}\}^\perp} u_L^{(i)}\}_{i=1}^m$,
\[
v^\top \expect_x\bigl[H(x; \theta(t))\bigr]\, v \;\ge\; c_0,
\]
where $\Pi_{\{\mathbf{1}\}^\perp} = I - \tfrac{1}{C}\mathbf{1}\mathbf{1}^\top$. This requires $\{\Pi_{\{\mathbf{1}\}^\perp} u_L^{(i)}\}_{i=1}^m$ to be linearly independent, and for canonical dead directions that holds exactly when $m < C$: the projection maps into the $(C-1)$-dimensional $\mathbf{1}^\perp$, so at $m = C$ the family cannot be independent even though no individual $u_L^{(i)}$ is parallel to $\mathbf{1}$. Given $m < C$, the positivity itself is generic: the model does not collapse to a point mass on any $\{u_L^{(i)}\}$-spanned affine subspace.

\begin{theorem}[Multi-direction bridge theorem]
\label{thm:bridge_multi}
Under the multi-direction setup, Assumption~\ref{ass:multi_consistent_indexing}, the multi-direction CE assumption above (or the MSE base case for squared-error loss), and activation classes (P1)--(P3):
\begin{enumerate}
\item[(a)] \emph{(Per-direction dead-direction-entry rate.)} For each $i \in \{1, \ldots, m\}$ and $\ell \in \{1, \ldots, L-1\}$, the dead-direction-$i$ entry of the G-factor at layer $\ell$ satisfies
\[
(G_\ell(\theta(t)))_{u_\ell^{(i)}\, u_\ell^{(i)}} \;=\; C_\ell^{(i)} \cdot t^{2 \Pi_\ell^{(i)}} \cdot \bigl(1 + r_\ell^{(i)}(t)\bigr), \qquad \Pi_\ell^{(i)} \;:=\; \sum_{\ell' = \ell+1}^{L} p_{\ell'}^{(i)},
\]
with $r_\ell^{(i)}(t) = O(t)$ and activation-class-dependent constants matching Theorem~\ref{thm:bridge}.
\item[(b)] \emph{(Output layer.)} For each $i$, $(G_L(\theta(t)))_{u_L^{(i)}\, u_L^{(i)}} = \Theta(1)$ (constant from $\sigma^2$ for MSE or from $c_0$ for the multi-direction CE assumption).
\item[(c)] \emph{(Spectrum / $\lambda_{\min}$ corollary, under non-dead-block control.)} Provided the non-dead block of $G_\ell$ is uniformly $\Theta(1)$ in $t$ (the multi-direction analog of Lemma~\ref{lem:non-dead}; a multi-direction version of Assumption~\ref{ass:rect_complement} of \S\ref{app:bridge_rect} suffices), under MSE the $m$ dead-direction Schur complements yield $m$ independent eigenvalues at rates $\{2 \Pi_\ell^{(i)}\}_{i=1}^m$, and $\lambda_{\min}(G_\ell) = \min_i \Theta(t^{2 \Pi_\ell^{(i)}})$. Under cross-entropy in class (P1) the logit-shift gauge contributes one exact zero eigenvalue at every layer and every $t$ (the kernel eigenvector rotates into the dead subspace as $t \to 0$), so only $m - 1$ eigenvalues carry rates and $\lambda_{\min}(G_\ell) = 0$ identically; in (P2) and (P3) the sample-dependent activation gates leave the cancellation partial and all $m$ rated eigenvalues survive (Theorem~\ref{thm:bridge_ce}(c)); the per-direction entries of (a) are unaffected in every class. When the non-dead-block assumption fails (e.g., rectangular-width-induced gauge directions in the non-dead block), the per-direction entry rate of (a) likewise remains the operational observable.
\item[(d)] \emph{(Reduction.)} The theorem reduces to Theorem~\ref{thm:bridge} when $m = 1$ and $p_\ell = 1$ for all $\ell$ (in which case $\Pi_\ell = L - \ell$).
\end{enumerate}
\end{theorem}

\noindent\emph{Note on hypotheses.} The dead-direction-entry statement (a) is the load-bearing one and uses only Assumption~\ref{ass:multi_consistent_indexing} plus the canonical multi-direction setup; the spectrum statement (c) requires the additional non-dead-block control, paralleling the structure of Theorem~\ref{thm:bridge_rect} (the dead-direction entry is the theorem; $\lambda_{\min}$ is a corollary). In the diagonal model the joint statement is exact: the $(i,j)$ cross moment of the dead block carries $t^{\Pi_\ell(i) + \Pi_\ell(j)}$, so the whole dead sub-block scales as a diagonal congruence.

\begin{proof}
Key observation: the perturbation $\sum_i t^{p_\ell^{(i)}}\, u_\ell^{(i)} (u_{\ell-1}^{(i)})^\top$ is diagonal in the canonical basis, with each term $t^{p_\ell^{(i)}}$ contributing only to the $(h_\ell - m + i, h_{\ell-1} - m + i)$ matrix entry (direction $i$ at layer $\ell-1$ maps to direction $i$ at layer $\ell$ with scale $t^{p_\ell^{(i)}}$; no cross-coupling between directions). Combined with the block-diagonal structure of $W_\ell^*$, this gives
\[
W_\ell(t) \, u_{\ell-1}^{(j)} \;=\; t^{p_\ell^{(j)}}\, u_\ell^{(j)} \qquad \text{for each } j = 1, \ldots, m.
\]
Iterating along the partial product from layer $\ell$ to $L$:
\[
W_L(t) \cdots W_{\ell+1}(t)\, u_\ell^{(i)} \;=\; \Bigl(\prod_{\ell'=\ell+1}^{L} t^{p_{\ell'}^{(i)}}\Bigr)\, u_L^{(i)} \;=\; t^{\Pi_\ell^{(i)}}\, u_L^{(i)}.
\]
That each intermediate $u_{\ell'}^{(i)}$ is again a canonical basis vector (consistent indexing, established in the setup) is what allows this multiplicative iteration: direction $i$ at each layer stays at canonical index $h_{\ell'} - m + i$.

Transposing, the backward operator $W_{\ell+1}(t)^\top \cdots W_L(t)^\top$ sends $u_L^{(i)}$ back to $u_\ell^{(i)}$ with gain $t^{\Pi_\ell^{(i)}}$. The dead-direction-$i$ backward delta at layer $\ell$ is thus
\[
u_\ell^{(i)} {}^\top \delta^{(\ell)} \;=\; t^{\Pi_\ell^{(i)}}\, u_L^{(i)} {}^\top \delta^{(L)}
\]
at leading order. The base case is direction-$i$-specific:
\begin{itemize}
\item \emph{MSE:} $\expect[(u_L^{(i)} {}^\top \delta^{(L)})^2] = \sigma^2$ (from the Gaussian noise setup, uniform across all $m$ directions).
\item \emph{Cross-entropy:} The multi-direction version of Assumption~\ref{ass:nondeg_pxy} gives $(u_L^{(i)})^\top \expect_x[H] u_L^{(i)} \ge c_0$ for each $i$, yielding $\expect[(u_L^{(i)} {}^\top \delta^{(L)})^2] = \Theta(1)$ uniformly on the neighborhood $\mathcal{U}$.
\end{itemize}
In either case, squaring gives the stated rate for direction $i$.

\emph{Per-direction independence via Schur reduction.} The cross-term $u_\ell^{(i)}{}^\top G_\ell u_\ell^{(j)}$ for $i \ne j$ is proportional to $\expect[(u_\ell^{(i)}{}^\top \delta^{(\ell)}) \cdot (u_\ell^{(j)}{}^\top \delta^{(\ell)})]$. From the partial-product formula above, $u_\ell^{(i)}{}^\top \delta^{(\ell)}$ depends only on the $i$-th component of $\delta^{(L)}$, and similarly for $j$. For MSE, the $\sigma^2 I$ noise covariance is diagonal in the canonical basis so cross-coordinate covariances vanish identically. For CE, Lemma~\ref{lem:ce_head} gives $H(x;\theta) = \mathrm{diag}(p) - pp^\top$ with off-diagonal data-averaged entry $-\expect_x[p_i p_j]$ at order $\Theta(1)$, so the dead-direction cross-entry is $(G_\ell)_{u^{(i)} u^{(j)}} = \Theta(t^{\Pi_\ell^{(i)} + \Pi_\ell^{(j)}})$.

Order the dead directions by $\Pi^{(1)} \le \Pi^{(2)} \le \cdots \le \Pi^{(m)}$. Because the off-diagonal $\Theta(t^{\Pi^{(i)} + \Pi^{(j)}})$ is the geometric mean of the two diagonals $\Theta(t^{2\Pi^{(i)}})$ and $\Theta(t^{2\Pi^{(j)}})$, it can \emph{exceed} the smaller diagonal in raw magnitude (when $\Pi^{(i)} < \Pi^{(j)}$, $\Pi^{(i)} + \Pi^{(j)} < 2 \Pi^{(j)}$); a "subleading" comparison directly between the off-diagonal and the diagonals would therefore be wrong. We instead use the Schur reduction. Treat the dead-direction block of $G_\ell$ in the basis $\{u^{(1)}, \ldots, u^{(m)}\}$ as
\[
\begin{pmatrix} d_1 & w_{12} & \cdots \\ w_{12} & d_2 & \cdots \\ \vdots & & \ddots \end{pmatrix},
\]
with $d_i = \Theta(t^{2 \Pi^{(i)}})$ and $w_{ij} = \Theta(t^{\Pi^{(i)} + \Pi^{(j)}})$, plus the non-dead block $M$ ($\succ 0$, $\Theta(1)$) and dead-row off-diagonals $v_i = \Theta(t^{\Pi^{(i)}})$ to the non-dead block. For each $i$, the dead-direction-$i$ Schur complement against (the non-dead block plus all faster-decaying directions $j$ with $\Pi^{(j)} > \Pi^{(i)}$) gives
\[
d_i \;-\; v_i^\top M^{-1} v_i \;-\; \sum_{j > i \text{ (faster-decaying)}} w_{ij}^2 / d_j,
\]
where the second-tier terms are the cross-direction Schur contributions. Each $v_i^\top M^{-1} v_i = \Theta(t^{\Pi^{(i)}}) \cdot \Theta(1) \cdot \Theta(t^{\Pi^{(i)}}) = \Theta(t^{2 \Pi^{(i)}})$ matches the dead diagonal in order and absorbs at most a fixed fraction (the rank-1 outer-product mechanism of Lemma~\ref{lem:integral-reduction-sub}; see Remark~\ref{rem:schur_constant_qualitative}). Each cross-direction term $w_{ij}^2 / d_j = \Theta(t^{2(\Pi^{(i)} + \Pi^{(j)})}) / \Theta(t^{2 \Pi^{(j)}}) = \Theta(t^{2 \Pi^{(i)}})$ also matches in order. What keeps the residue positive is strict positive definiteness of the dead block's leading coefficient matrix: writing the dead block as $D H_{\mathrm{dead}} D$ with $D = \mathrm{diag}(t^{\Pi^{(i)}})$, the $i$-th eigenvalue is $\Theta(t^{2\Pi^{(i)}})$ exactly when the leading principal minors of $H_{\mathrm{dead}}$ are positive (a rank-deficient $H_{\mathrm{dead}}$ collapses the corresponding rate to an exact zero). Under MSE the diagonal noise covariance makes $H_{\mathrm{dead}}$ diagonal; under cross-entropy the multi-direction assumption above supplies the positivity on the projected span. Hence the dead-direction-$i$ Schur complement is $\Theta(t^{2 \Pi^{(i)}}) \cdot (1 - c_i)$ with $c_i < 1$, giving the per-direction eigenvalue rate stated in (a). Non-dead eigenvalues are $\Theta(1)$ by the same argument as in Theorem~\ref{thm:bridge}, used in clause (c) above.

\emph{Activation-class corrections.} For class (P2) smooth activations, each $\phi'(a_{\ell'}^{(u^{(i)})})$ factor along the backward chain contributes $(1 + O(t))$ at direction $i$. Accumulating over $\ell' = \ell+1, \ldots, L$, the total correction at layer $\ell$ for direction $i$ is $r_\ell^{(i)}(t) = O(t)$, matching Theorem~\ref{thm:bridge_rect}'s activation-correction structure. For (P1) linear and (P3) ReLU, $r_\ell^{(i)}(t) = 0$.
\end{proof}

\begin{corollary}[Single-direction, non-symmetric approach]
\label{cor:bridge_g3}
For $m = 1$ and layer-varying $p_\ell$, the single-direction rate is
\[
(G_\ell)_{u_\ell u_\ell} = \Theta\!\Bigl(t^{2 \sum_{\ell' > \ell} p_{\ell'}}\Bigr).
\]
Note: the exponent depends on $\{p_{\ell'}\}_{\ell' > \ell}$ only, \emph{not} on $p_\ell$ itself. This is because $G_\ell$ reflects the backward chain ABOVE layer $\ell$; layer $\ell$'s own perturbation does not appear.
\end{corollary}

\begin{corollary}[Multiple directions, uniform symmetric approach]
\label{cor:bridge_g2}
Under part (c)'s non-dead-block control and MSE loss, for $m \ge 1$ dead directions and $p_\ell^{(i)} = 1$ for all $i, \ell$, each direction contributes an eigenvalue at rate $2(L - \ell)$: $G_\ell$'s spectrum (total dimension $h_\ell$) contains $m$ eigenvalues at the dead rate $\Theta(t^{2(L-\ell)})$ and $h_\ell - m$ non-dead eigenvalues of $\Theta(1)$. Under cross-entropy in class (P1) one of the $m$ is replaced by the logit-shift gauge's exact zero (part (c)), leaving $m - 1$ rated eigenvalues.
\end{corollary}

\begin{corollary}[Multi-direction, asymmetric approach]
\label{cor:bridge_g7}
Under part (c)'s non-dead-block control and MSE loss, for $m \ge 1$ directions with per-direction per-layer exponents $p_\ell^{(i)}$, the spectrum of $G_\ell$ contains $m$ eigenvalues at rates $\{2 \Pi_\ell^{(i)}\}_{i=1}^m$ (which may be distinct if the $\Pi_\ell^{(i)}$ differ, coincident otherwise; under cross-entropy in class (P1), $m - 1$ of them, per part (c)). The operational selection rule of \S\ref{app:selection_rule_op} identifies rate-carrying eigenvalues by matching measured $\alpha_i$ to predicted values; under this theorem, \emph{multiple} matches at a single component are expected when $m \ge 2$ and correspond to distinct dead directions, rather than being measurement noise. The ordering of which rate is "direction 1" vs "direction 2" is a labeling choice, not intrinsic.
\end{corollary}

\begin{remark}[Static-Fisher rate vs.\ trajectory observability for multiple directions]
\label{rem:bridge_multi_adam_scope}
Theorem~\ref{thm:bridge_multi} is a static-Fisher statement on the parametric trajectory. Multiple dead directions amplify the Adam non-equivariance documented in Remark~\ref{rem:adam_nondescent}: each direction independently couples to gauge-mode drift, so per-direction trajectory rate readout under standard Adam is doubly fragile. Validation of the per-direction rates accordingly uses a parametric freeze-probe under a theorem-compatible setup, not Adam-trajectory rate-fitting. The constructive closure for the gauge case (Corollary~\ref{cor:ddcadam_quotient_rate}) lifts to multi-direction: under DDCAdam, the per-direction rates $\{2 \Pi_\ell^{(i)}\}_i$ are recoverable from a real training trajectory in the same regime as under SGD, since the gauge-mode coupling that causes the doubly-fragile failure mode under Adam is removed by Proposition~\ref{prop:ddcadam_equivariance}'s equivariance.
\end{remark}

\begin{remark}[Joint scope with rectangular, cross-entropy, and biases]
\label{rem:bridge_multi_joint_scope}
The multi-direction extension touches only the dead-direction sub-block of $G_\ell$ (each direction independently contributing a Schur complement of the form analysed above), and is therefore disjoint from the rectangular-widths reduction (which acts on the non-dead block), the CE output-head base-case replacement (Lemma~\ref{lem:ce_head}), and the bias-augmented Schur structure (each direction $i$ carries its own perturbation $b_\ell^{(i)}(t) = t^{q_\ell^{(i)}} u_\ell^{(i)}$). The joint statements are sketched rather than separately proved; the disjoint-block structure of the multi-direction Schur reduction is the load-bearing reason each composition holds.
\end{remark}

\subsection{Biases}
\label{sec:theory:arch:bias}

\label{app:bridge_bias}

Previous bridge theorems state the forward pass as $h_\ell = \phi(W_\ell h_{\ell-1})$, without biases. In practice every weight layer has an accompanying bias vector: $h_\ell = \phi(W_\ell h_{\ell-1} + b_\ell)$. This subsection extends the bridge theorem to cover biases, showing that under the natural singular-configuration convention $b_\ell^* = 0$, biases do not alter the rate structure; their sole effect under smooth activations is a predictable finite-$t$ Taylor correction.

\paragraph{Setup.}
Let the singular configuration be $\theta^* = (\{W_\ell^*\}, \{b_\ell^* = 0\})$, with the $W_\ell^*$ structure inherited from the relevant base setup (single- or multi-direction, square or rectangular). The choice $b_\ell^* = 0$ carries content: for (P1) the forward Jacobian's rank is bias-independent, but for (P2) and (P3) the bias shifts every pre-activation and moves the activation pattern (under ReLU a negative dead bias closes the dead channel's gate identically and changes the Jacobian's rank), so the analysis is stated at $b_\ell^* = 0$ and Remark~\ref{rem:bias_sing_choice} records the one symmetry that relates nonzero-dead-bias configurations to it.

For concreteness we state the single-direction symmetric case explicitly; the extension to multi-direction and non-symmetric approaches (Theorem~\ref{thm:bridge_multi}) proceeds analogously by adding independent bias perturbations per dead direction. The dead-aligned approach perturbs both weights and biases in the dead direction:
\[
W_\ell(t) \;=\; W_\ell^* + t \cdot u_\ell u_{\ell-1}^\top, \qquad b_\ell(t) \;=\; t^{q_\ell} \cdot u_\ell,
\]
where $q_\ell \ge 1$ (real) is the bias perturbation exponent at layer $\ell$. The weight-only case corresponds to $b_\ell(t) \equiv 0$ (equivalently $q_\ell \to \infty$); the symmetric joint case to $q_\ell = 1$ at all layers.

One scope point on deadness. Definition~\ref{def:dead_direction} attaches to the weight-only direction, whose KL order is $k = L$. The bias-augmented paths are shallower as directions of the full-parameter Fisher: with $q_L = 1$ the output-layer bias moves the function at first order, so the symmetric joint direction has KL order $1$ and is regular (measured $K$-slope $2$, Fisher-form slope $0$, linear and ReLU alike); with hidden-layer biases only, the deepest surviving channel is the layer-$L$ dead row picking up the layer-$(L-1)$ bias at order $t^2$, capping the KL order at $2$ for every depth (measured $K$-slope $4$, Fisher-form slope $2$, against the weight-only $2L$ and $2(L-1)$). The theorem's statements (a)--(c) are per-layer $G_\ell$-entry statements along the stated path and are unaffected; only the word ``dead'' for the joint direction would misapply the definition.

\begin{theorem}[Multi-Layer K-FAC G-factor Bridge with biases]
\label{thm:bridge_bias}
Under the bias-augmented setup and activation classes (P1)--(P3):
\begin{enumerate}
\item[(a)] For $\ell \in \{1, \ldots, L-1\}$,
\[
(G_\ell(\theta(t)))_{u_\ell u_\ell} \;=\; C_\ell \cdot t^{2(L-\ell)} \cdot \bigl(1 + r_\ell^{G,b}(t)\bigr),
\]
where $r_\ell^{G,b}(t)$ is an extended finite-$t$ correction:
\[
r_\ell^{G,b}(t) \;=\; r_\ell^G(t) \;+\; r_\ell^{\text{bias}}(t),
\]
with $r_\ell^G(t)$ the no-bias correction from Theorem~\ref{thm:bridge} and
\[
r_\ell^{\text{bias}}(t) \;=\; \begin{cases} 0, & \phi \in \{\text{linear (P1), ReLU (P3)}\}, \\ O(t^{q_\ell}), & \phi \in \text{(P2)}, \end{cases}
\]
where $q_\ell$ is the bias perturbation exponent at layer $\ell$. In the no-bias case the smooth-activation correction descends from the per-layer Taylor term $\phi'(a_\ell^{(u)}) = \phi'(0) + \phi''(0)\,\Theta(t^\ell)$, so it shrinks as the layer index grows; accumulated over the backward chain it is bounded by $O(t)$ uniformly in $\ell$, which is the form Theorem~\ref{thm:bridge} states. The bias correction $r_\ell^{\text{bias}}(t) = O(t^{q_\ell})$ carries no such depth dependence, being governed entirely by the bias exponent $q_\ell$. In particular, for $q_\ell = 1$ (the symmetric joint case), the correction is $O(t)$ uniformly in $\ell$. Both corrections vanish as $t \to 0$, so the leading-order rate $t^{2(L-\ell)}$ is preserved. For (P3) the vanishing of $r_\ell^{\text{bias}}$ concerns the $t$-dependence only: the leading constant differs from Theorem~\ref{thm:bridge}'s, since a positive dead bias removes the survival gating (the proof gives $\sigma^2$ in place of $\sigma^2/2$ at $\ell \ge 2$, and $\sigma^2 \Phi(1)$ at $\ell = 1$ for $q_1 = 1$), so $C_\ell$ here denotes the bias-augmented constant.
\item[(b)] $(G_L(\theta(t)))_{u_L u_L} = \Theta(1)$, unchanged from Theorem~\ref{thm:bridge}.
\item[(c)] The rate structure $t^{2(L-\ell)}$ is preserved across all cases; bias perturbations do not contribute a new rate-carrying dead direction.
\end{enumerate}
\end{theorem}

\begin{proof}
We absorb biases into augmented weight matrices. Define $\tilde{W}_\ell \in \mathbb{R}^{h_\ell \times (h_{\ell-1} + 1)}$ by $\tilde{W}_\ell = [W_\ell \mid b_\ell]$ (weight matrix with bias appended as an extra column), and $\tilde{h}_{\ell-1} = [h_{\ell-1}; 1] \in \mathbb{R}^{h_{\ell-1}+1}$ (activation with a constant $1$ appended). Then the pre-activation at layer $\ell$ is $W_\ell h_{\ell-1} + b_\ell = \tilde{W}_\ell \tilde{h}_{\ell-1}$, and the forward pass becomes
\[
a_\ell = \tilde{W}_\ell \tilde{h}_{\ell-1}, \qquad h_\ell = \phi(a_\ell), \qquad \tilde{h}_\ell = [h_\ell; 1].
\]
The re-appending of the constant $1$ at each layer's output is explicit bookkeeping; $\phi$ is still applied coordinate-wise to $a_\ell \in \mathbb{R}^{h_\ell}$ only (the $1$-coordinate is not passed through $\phi$, avoiding any concerns about $\phi(1)$).

Under this augmentation, $\tilde{W}_\ell^*$ has the structure
\[
\tilde{W}_\ell^* = [W_\ell^* \mid 0], \qquad \tilde{W}_\ell(t) = \tilde{W}_\ell^* + t \cdot u_\ell [u_{\ell-1}; 0]^\top + t^{q_\ell} \cdot u_\ell [0; 1]^\top,
\]
where the two perturbation terms correspond to the weight and bias perturbations respectively. The dead-direction-to-dead-direction entry of $\tilde{W}_\ell(t)$ (now in the extended output-input index space) still carries a $t$ factor from the weight perturbation; the new $t^{q_\ell}$ bias perturbation couples the dead output direction $u_\ell$ to the constant $1$-input, \emph{not} to a dead-input direction. In the backward chain, this means the $t^{q_\ell}$ bias perturbation is \emph{not} in a position to contribute to the backward-chain product from $u_L$ to $u_\ell$, which routes through weight entries coupling $u_{\ell'}$ to $u_{\ell'-1}$ at each step. The backward-chain derivation of Lemma~\ref{lem:backward-dead-sub} applies to the weight block of $\tilde{W}_\ell$ unchanged, yielding $\delta_\ell^{(u)} = \Theta(t^{L-\ell})$.

\emph{Schur structure of the bias-augmented A-factor.} The augmented A-factor $\tilde A_\ell = \mathbb{E}_x[\tilde h_{\ell-1} \tilde h_{\ell-1}^\top]$ is $(h_{\ell-1} + 1) \times (h_{\ell-1} + 1)$ with the block decomposition
\[
\tilde A_\ell \;=\; \begin{pmatrix} A_\ell & m_\ell \\ m_\ell^\top & 1 \end{pmatrix}, \qquad m_\ell = \mathbb{E}_x[h_{\ell-1}(x;\theta(t))].
\]
The non-dead components of $m_\ell$ are $\Theta(1)$, and the bias caps the dead side of the forward ladder: under a $q = 1$ bias the dead pre-activation at every layer is $\Theta(t)$ (the bias term dominates the weight chain's $\Theta(t^m)$, per the (P1) paragraph below), so the dead diagonal of $\tilde A_\ell$ is $\Theta(t^2)$ for $\ell \ge 2$ in place of the bias-free $\Theta(t^{2(\ell-1)})$ (measured slopes $(0.00, 2.00, 2.19, 2.21)$ across four layers against the bias-free $(0, 2, 4, 6)$), with dead off-row entries $O(t)$ and the non-dead block $\Theta(1)$ uniformly in $t$. In the affine (P1) model the cap is exact: the dead forward moment equals $t^{2m} + (\sum_{i=1}^{m} t^i)^2$, with leading rate $2$ and coefficient $1$ at every depth $m \ge 2$. Two consequences. The A--G duality product of Corollary~\ref{cor:a_g_duality} is not bias-stable: its layer-invariance is a bias-free statement, and no bias-augmented restatement is claimed. And the theorem's own claims are untouched, because (a)--(c) are G-side statements and the backward chain routes only through weight columns; the capped $\tilde A_\ell$ still has the dead-row/non-dead-block split with a $\Theta(1)$ non-dead block, which is all the G-side argument consumes.

For \textbf{linear (P1) and ReLU (P3)}: the forward dead-component magnitude is unchanged at leading order. For P1, $a_\ell^{(u)} = \Theta(t) + \Theta(t^\ell) \cdot x$ where the $\Theta(t)$ bias contribution coexists with the $\Theta(t^\ell)$ weight contribution. The \emph{backward} dead-delta is driven by the weight chain only (biases do not backprop through weight transposes), so $\delta_\ell^{(u)} = \Theta(t^{L-\ell})$ at leading order, exactly as in Lemma~\ref{lem:backward-dead-sub}. For P3, the dead-coordinate pre-activation $a_\ell^{(u)} = b_\ell^{(u)}(t) + W_\ell^{(u,\cdot)} h_{\ell-1} = t^{q_\ell} + \Theta(t^\ell)$ is dominated by the bias contribution for $q_\ell \le \ell$ (in particular for the symmetric joint case $q_\ell = 1$ at $\ell \ge 2$). For $\ell \ge 2$ it is uniformly positive over the input distribution at all $t > 0$ (the incoming term $t\,\phi(a_{\ell-1}^{(u)})$ is non-negative under ReLU), so $\phi'(a_\ell^{(u)}) \equiv 1$ there and the backward factor is $1$ in place of the bias-free survival factor: the constant doubles relative to Theorem~\ref{thm:bridge} ($\sigma^2$ against $\sigma^2/2$). At $\ell = 1$ the incoming term $t\, x^{(h)}$ is Gaussian, so the pre-activation $t(x^{(h)} + 1)$ at $q_1 = 1$ is negative with probability $\Phi(-1) \approx 0.159$ independent of $t$, and the layer-$1$ constant carries the survival factor $\Phi(1)$ ($\sigma^2 \Phi(1) \approx 0.84\, \sigma^2$; at $\sigma = 0.5$ the measured layer-$1$ constant is $0.2098$ against $0.2495$ at the deeper layers). The rate is untouched in all cases. For non-dead coordinates $j \neq u$, the bias perturbation is zero by the canonical-direction setup, and $\phi'(a_\ell^{(j)}) = \phi'(\Theta(1)) = \Theta(1)$ unchanged from the no-bias case, so the non-dead block of $G_\ell$ retains the $\Theta(1)$ control of Lemma~\ref{lem:non-dead}. A $\phi'$-discontinuity at a sign-cell shift applies only when the dead coordinate's canonical sign cell is the negative one, a basis choice resolvable by sign-flipping $u_\ell$, and not a separate scope.

For \textbf{smooth activations (P2)}: the forward dead-component picks up the $\Theta(t)$ bias contribution (from the $b_\ell^{(u)} = t^{q_\ell}$ perturbation, dominant when $q_\ell = 1 < \ell$), so $a_\ell^{(u)} = \Theta(t)$ at leading order rather than $\Theta(t^\ell)$. The Taylor expansion of $\phi'(a_\ell^{(u)})$ around $\phi'(0)$ gives
\[
\phi'(a_\ell^{(u)}) \;=\; \phi'(0) + \phi''(0) \cdot \Theta(t) + O(t^2).
\]
Accumulating across the $L - \ell$ backward Jacobian factors above layer $\ell$ multiplies this $(1 + O(t))$ correction, contributing an additive $r_\ell^{\text{bias}}(t) = O(t)$ correction to the log-slope. The leading-order rate $t^{2(L-\ell)}$ is unchanged: the bias contribution enters through the activation derivative correction alone, and the backward chain's $t$-scaling is untouched.
\end{proof}

\begin{remark}[Natural singular-configuration choice for biases]
\label{rem:bias_sing_choice}
Setting $b_\ell^* = 0$ at the singular configuration is a choice backed by one limited symmetry. Shifting the zero-point of a pre-activation is not a symmetry of a network with an elementwise nonlinearity, because $\phi$ does not commute with translation: under ReLU a negative dead bias places the whole dead coordinate in the inactive cell, $\phi'$ vanishes identically, and the dead backward chain is exactly zero in place of $\Theta(t^{L-\ell})$; the forward Jacobian's rank likewise depends on the bias for any activation whose derivative is not constant. The symmetry that does exist is the simultaneous sign flip of every $u_\ell$, which leaves the weight-perturbation entries unchanged and flips the sign of every dead bias, so configurations differing by a uniform dead-bias sign flip are equivalent. Non-zero biases in non-dead coordinates are $\Theta(1)$ contributions absorbed into the $\Theta(1)$ non-dead block of $G_\ell$; non-zero \emph{dead} biases place the configuration in a regime the theorem does not cover.
\end{remark}

\paragraph{Empirical validation.}
Direct validation at $L = 4$, $h = 6$, MSE, across three cases and three activations:
\begin{itemize}
\item \emph{Case A} (weight-only, $b \equiv 0$): exact standard-theorem rates across linear, ReLU, GeLU, confirming biases-at-zero are inert.
\item \emph{Case B} (bias-only, weights fixed at $W^*$, $q \in \{1, 2\}$): with the weights at $W^*$ the dead weight column is zero, so the backward dead delta vanishes \emph{identically} at every $\ell < L$ and every $t$; the measured entries are exact zeros with no rate to fit (only the output layer is non-zero), confirming part (c): bias perturbations alone create no rate-carrying dead direction.
\item \emph{Case C} (joint perturbation):
  \begin{itemize}
  \item Linear, ReLU: exact match with the weight-only prediction (part (a) with $r_\ell^{\text{bias}} = 0$).
  \item GeLU, $q_\ell = 1$: measured $\alpha_1 = 7.64, \alpha_2 = 5.13, \alpha_3 = 2.57$ (compared to the no-bias GeLU values $6.48, 4.12, 2.02$). The \emph{additional} slope-shift from the bias, on top of the standard smooth-activation $r_\ell^G$ term, is $(1.16, 1.01, 0.55)$ across $\ell = 1, 2, 3$ at our $t$-range $[10^{-1.2}, 10^{-0.3}]$. This is the expected behavior of an $O(t)$ multiplicative correction of the form $(1 + ct)^{2(L-\ell)}$ fit in log-log space: the effective log-slope at finite $t$ exceeds the leading-order $t \to 0$ asymptote by a bounded amount that depends on $c$ and the $t$-range, with $c$ accumulating across the $L - \ell$ backward Jacobian factors.
  \item GeLU, $q_\ell = 2$: measured $\alpha_1 = 7.24, \alpha_2 = 4.76, \alpha_3 = 2.37$, a smaller slope-shift than $q = 1$ (as predicted: $r_\ell^{\text{bias}} = O(t^{q_\ell}) = O(t^2)$ is a smaller correction over the same $t$-range).
  \end{itemize}
\end{itemize}
All cases match the theorem's prediction: rate structure preserved, smooth-activation correction bounded and activation-class-dependent as stated.

\paragraph{Scope and composition with prior extensions.}
Theorem~\ref{thm:bridge_bias} is stated and proved for the square-hidden, feedforward, single-direction, symmetric-approach setting. The bias column of $\tilde{W}_\ell = [W_\ell \mid b_\ell]$ does not enter the backward chain at the dead direction (the bias column couples dead output to constant-$1$, not dead input), so the bias-augmented Schur reduction is disjoint from the rectangular non-dead-block reduction (Theorem~\ref{thm:bridge_rect}, where the narrow chain does not include the $1$-coordinate), the CE output-head replacement (Lemma~\ref{lem:ce_head}, which leaves the backward chain unchanged), the residual-DAG path-distance (Theorem~\ref{thm:bridge_res}, where $K(\ell)$ counts weight edges only), and the multi-direction per-direction Schur reduction (Theorem~\ref{thm:bridge_multi}, where each direction $i$ carries its own perturbation $b_\ell^{(i)}(t) = t^{q_\ell^{(i)}} u_\ell^{(i)}$). The $r_\ell^{\mathrm{bias}}(t)$ correction accumulates predictably as $O(t^{q_\ell^{(i)}})$ per direction. We mark the joint statements as natural follow-ups, not ``mechanical applications,'' because closing each requires verifying the disjointness formally, which we have done above only for the square-hidden single-direction MSE base case.

\subsection{SwiGLU}
\label{sec:theory:arch:swiglu}

\label{app:bridge_swiglu}

This subsection states the per-block forward rate for the SwiGLU MLP variant
used in modern decoder-only LLM families. The result is a
direct instantiation of the bridge framework on the SwiGLU forward map, not
an extension of the bridge theorem itself.

\begin{proposition}[SwiGLU MLP block rate]
\label{prop:swiglu_rate}
Let $\mathrm{SwiGLUMLP}(x) = W_{\mathrm{down}} \bigl( \mathrm{silu}(W_{\mathrm{gate}} x) \odot W_{\mathrm{up}} x\bigr)$ with all three Linear layers at canonical init $W = \mathrm{diag}(1, \ldots, 1, t)$ (shared dead dimension $d$). The forward block rate is $k_{\mathrm{SwiGLU}}^{\mathrm{fwd}} = 3$: the block output along $e_d$ is $(1/2) \cdot t^3 \cdot x_d^2 + O(t^4)$.
\end{proposition}

\begin{proof}
At the dead dimension: $(W_{\mathrm{gate}} x)_d = t \cdot x_d$. Using $\mathrm{silu}(y) = y \cdot \sigma(y)$ with $\sigma(0) = 1/2$, we have $\mathrm{silu}(t \cdot x_d) = (1/2) \cdot t \cdot x_d + O(t^2)$. Also $(W_{\mathrm{up}} x)_d = t \cdot x_d$. Their elementwise product at dimension $d$: $(1/2) \cdot t^2 \cdot x_d^2$. Finally $W_{\mathrm{down}}$ with dead column contributes another factor of $t$: $y_d = t \cdot (1/2) \cdot t^2 \cdot x_d^2 = (1/2) \cdot t^3 \cdot x_d^2$. The remainder is global and explicit: the dead-channel output is $t^3 x_d^2\, \sigma(t x_d)$ identically, and $|\sigma(u) - 1/2| \le |u|/2$ gives $|y_d - (1/2)\, t^3 x_d^2| \le t^4 |x_d|^3/2$ at every $t \ge 0$ and every input. The off-diagonal entries of all three matrices contribute at order $t^0$ through non-dead channels and do not affect the dead-channel output at leading order. Empirical verification: parametric probe at $d \in \{16, 64\}$ gives forward slope $3.000$ exactly, fit on $t \le 10^{-2}$.
\end{proof}

\begin{remark}[SwiGLU stacks do not compose additively]
\label{rem:swiglu_composition}
Theorem~\ref{thm:bridge_composition} adds block rates only under its scalar-transfer hypothesis, and SwiGLU fails that hypothesis. Its dead-direction map is $x_d \mapsto \tfrac12 t^3 x_d^2$, quadratic in the dead coordinate rather than a scalar multiple of it, because the gate and the up-projection each contribute a factor of $x_d$. Feeding a $\Theta(t^e)$ input therefore returns $\Theta(t^{3 + 2e})$ instead of $\Theta(t^{3 + e})$, and a stack of $n$ SwiGLU blocks with no residual accumulates
\[
e_1 = 3, \qquad e_{n+1} = 3 + 2 e_n, \qquad \text{so} \qquad e_n = 3\,(2^n - 1),
\]
giving $3, 9, 21, 45$ at $n = 1, 2, 3, 4$ against the $3n$ that additivity would predict. A parametric probe at $d = 8$ measures slopes $3.001, 9.003, 21.005, 45.011$, matching $3(2^n-1)$ and separating from $3n$ at $n = 2$ already. The standard fc1-$\phi$-fc2 block escapes this because $\phi(0) = 0$ and $\phi'(0) \ne 0$ make its dead-direction map linear, so it does compose additively at $2n$ (measured $2.000, 4.000, 6.000, 8.000$ on the same probe).

Two consequences. A per-block forward rate of $3$ is a statement at $\Theta(1)$ input, which is how Definition~\ref{def:block_rate} fixes it, and it may not be summed across a SwiGLU stack. The per-block figure $k_{\mathrm{attn}}^{\mathrm{bk}} + k_{\mathrm{SwiGLU}}^{\mathrm{fwd}} = 5$ for an attention-plus-SwiGLU transformer block inherits the same restriction and holds only for a single block. Residual-wrapped SwiGLU blocks, which is what deployed decoder-only models use, sidestep the question entirely: the identity skip carries $K^{\mathrm{fwd}} = 0$ on the residual stream (Corollary~\ref{cor:sigma-min-res}) regardless of what the weight branch does.
\end{remark}

\begin{remark}[Practical application to modern LLM architectures]
\label{rem:llama_rates}
Modern decoder-only LLMs widely use SwiGLU MLPs in place of the standard fc1-activation-fc2 form. Use $k_{\mathrm{SwiGLU}}^{\mathrm{fwd}} = 3$ (Proposition~\ref{prop:swiglu_rate}) in place of the standard $k_{\mathrm{MLP}}^{\mathrm{fwd}} = 2$ for a single block, and do not sum it across a SwiGLU stack: Remark~\ref{rem:swiglu_composition} shows the composition is $3(2^n - 1)$ because the block map is quadratic in the dead coordinate. On the residual-wrapped stacks these models actually use, the residual-stream rate is $0$ either way.
\end{remark}

\begin{figure}[ht]
\centering
\includegraphics[width=\textwidth]{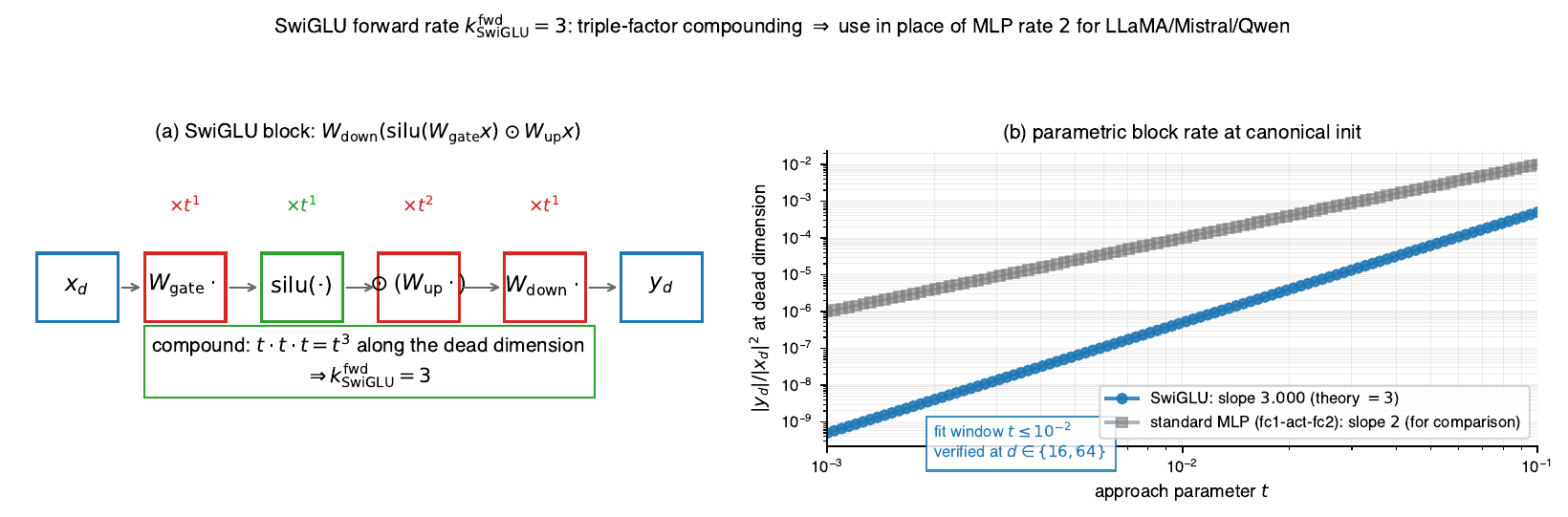}
\caption{SwiGLU forward rate $k_{\mathrm{SwiGLU}}^{\mathrm{fwd}} = 3$ (Proposition~\ref{prop:swiglu_rate}). Why SwiGLU has a higher block rate than the standard fc1-act-fc2 MLP. (a)~The block cascade compounds three $t$-factors along the dead dimension, one each from $W_{\mathrm{gate}}$, $W_{\mathrm{up}}$, and $W_{\mathrm{down}}$, with $\mathrm{silu}$ supplying the $\sigma(0) = 1/2$ coefficient; together they give forward block rate $3$. (b)~The parametric prediction at canonical init validates the rate: SwiGLU slope $3.000$ vs standard fc1-act-fc2 slope $2$, at $d \in \{16, 64\}$, fit window $t \le 10^{-2}$. SwiGLU-equipped decoder-only LLMs (LLaMA, Mistral, Qwen) inherit this higher per-block rate when applying composition additivity.}
\label{fig:swiglu_rate_parametric}
\end{figure}
 
\subsection{Non-canonical alignment}
\label{sec:theory:arch:noncanonical}

\label{app:bridge_rot}

All prior bridge theorems assume the dead direction $u_\ell$ is aligned with a canonical basis vector at every layer. Trained networks rarely satisfy this exactly: dead directions emerge during training at rotations of the canonical frame, and we want to know what survives. The answer divides cleanly along the activation axis. Under linear activation, rotation is without loss of generality, and Theorem~\ref{thm:bridge} transfers verbatim. Under nonlinear activation, canonical alignment is essential: arbitrary rotation collapses the rate to model-dependent values that have no clean closed form. We give the positive result for linear activation, the negative result for nonlinear, and a sketch of why nonlinearity breaks the rotation invariance. We restrict throughout to square hidden ($h_\ell = h$ for all $\ell$); the rectangular extension is orthogonal and addressed below.

\paragraph{Notation.} The rotation parameter $\alpha_{\mathrm{rot}} \in [0, 1]$ measures how far the dead direction is rotated away from the canonical basis: $\alpha_{\mathrm{rot}} = 0$ corresponds to canonical alignment ($u_\ell = e_h$), and $\alpha_{\mathrm{rot}} = 1$ to a uniformly random rotation of the dead direction.

\begin{remark}[Genericity of non-canonical alignment]
\label{rem:noncanonical_generic}
The set of unit vectors $u \in \mathbb{R}^h$ aligned with a coordinate basis vector has measure zero in the unit sphere $S^{h-1}$, so canonical alignment is a measure-zero condition. For randomly-initialised networks under SGD with smooth or ReLU activations, dead directions emerging during training generically lie at non-zero rotation $\alpha_{\mathrm{rot}}$ from any coordinate basis. The negative result of Proposition~\ref{prop:bridge_nonlinear_rot_negative} therefore applies generically, motivating an intermediate spectral-identification step (an operational selection rule that recovers the dead direction from the measured spectrum) at trained-network scale.
\end{remark}

\begin{proposition}[Linear activation: rotation is without loss of generality]
\label{prop:bridge_linear_rot}
Under activation class (P1) (linear, identity) and the square-hidden setup ($h_\ell = h$ for all $\ell$), Theorem~\ref{thm:bridge} holds for an arbitrary dead-direction unit vector $u = U e_h$ shared across layers ($u_\ell = u$ for every $\ell$, with $U \in O(h)$ a fixed orthogonal matrix sending the canonical $e_h$ to $u$). The rate $(G_\ell)_{u u} = \Theta(t^{2(L-\ell)})$ is preserved exactly.
\end{proposition}

\begin{proof}
Change of basis: let $U \in O(h)$ be an orthogonal matrix sending $e_h$ to $u_0$. Define rotated weights $\tilde{W}_\ell = U^\top W_\ell U$ (the same $U$ used at each layer's input and output, valid because the square-hidden hypothesis means input and output ambient spaces both have dimension $h$). Under this rotation, the singular configuration $W_\ell^* = U \cdot \mathrm{diag}(1, \ldots, 1, 0) \cdot U^\top$ (a symmetric matrix conjugate to $\diag(1,\ldots,1,0)$ via $U$ on both sides, valid because the dead-direction perturbation $t \cdot u u^\top$ is symmetric and the same orthogonal applies on the input and output sides) becomes $\tilde{W}_\ell^* = \mathrm{diag}(1, \ldots, 1, 0)$ (canonically aligned), and the perturbation $t \cdot u u^\top$ becomes $t \cdot e_h e_h^\top$. Linearity preserves this equivalence: the forward pass $h_\ell = W_\ell h_{\ell-1}$ transforms to $\tilde{h}_\ell = \tilde{W}_\ell \tilde{h}_{\ell-1}$ with $\tilde{h} = U^\top h$, and the backward chain transforms analogously. Theorem~\ref{thm:bridge} applies to the rotated system, and the rate prediction transfers back. (The change of basis rotates the regression target along with the weights; the conclusion does not depend on that, since the dead-direction rate needs only a $\Theta(1)$ output backward magnitude, and the unrotated-target instance measures the same rates.) Validated on the parametric rotated-dead-direction sweep: linear activation matches the canonical rate exactly at arbitrary rotation amounts $\alpha_{\mathrm{rot}} \in [0, 1]$.
\end{proof}

\begin{proposition}[Nonlinear activation: canonical alignment is essential, empirical demonstration]
\label{prop:bridge_nonlinear_rot_negative}
Under activation classes (P2) smooth (e.g., GeLU, tanh) and (P3) ReLU, the dead-direction rate is \emph{not} preserved under arbitrary rotation. At any fixed $\alpha_{\mathrm{rot}} > 0$ the mixed contributions of the sketch below enter at $\Theta(1)$ and dominate as $t \to 0$, so the asymptotic exponent is $0$ at every hidden layer; on a finite fit window the local log-slope takes transitional values between the canonical prediction and $0$ that depend on the activation, the rotation, the depth, and the window itself (e.g., at $\alpha_{\mathrm{rot}} = 0.1$, $L = 4$, $h = 6$, on $t \in [0.04, 0.30]$, ReLU reads $(0.05, 0.37, 0.92, 0)$ and GeLU $(0.19, 0.37, 1.44, 0)$ against the canonical $(6, 4, 2, 0)$), and every slope decays toward $0$ as the window moves to smaller $t$. The two limits $t \to 0$ and $\alpha_{\mathrm{rot}} \to 0$ do not commute. We present the network-level claim as an \emph{empirical demonstration}; the analytical mechanism is sketched below and proved exactly at a planar witness, and Remark~\ref{rem:rot_window_bound} gives the model-level window bound with its crossover scale.
\end{proposition}

\begin{proof}[Mechanism sketch]
The rate prediction derives from the canonical-basis block structure of $W_\ell^*$, which makes the dead-direction backward propagation factor cleanly (Lemma~\ref{lem:backward-dead-sub}). Elementwise activations $\phi$ commute with rotation only through the null rotation (identity). For a non-canonical dead direction $u = U e_h$ with $U$ non-diagonal, each backward Jacobian factor $\phi'(a_\ell) \odot \cdot$ mixes the dead-direction-along-$u$ with components of $a_\ell$ that have non-vanishing $\phi'$; these mixed contributions enter at $\Theta(1)$ and dominate the backward delta's dead-direction projection. A rigorous bound on the resulting rate as a function of $\alpha_{\mathrm{rot}}$ would require explicit accounting of how the elementwise nonlinearity couples canonical and rotated coordinates, which we have not done. The leak itself is machine-checked at an exact planar witness (Table~\ref{tab:machine_checked}): for the 3-4-5 rotation with dead direction $u = (-4/5, 3/5)$, the pre-activation at the dead-probe input is $t \cdot u$, whose coordinates carry opposite signs, so the ReLU mask is exactly $(0, 1)$ for every $t > 0$ and the masked dead direction carries live overlap $12/25$; the dead reading of a live backward flow then has leading rate $0$, while any canonical reading keeps its rate under every mask value because masks scale canonical directions in place.
\end{proof}

\begin{remark}[Window bound and crossover scale at the model level]
\label{rem:rot_window_bound}
The model observable behind the collapse admits a complete window statement, machine-checked (Table~\ref{tab:machine_checked}). Write the dead-direction moment as $G(t) = (\gamma s c)^2 + b\,t^{2K}$, with $s$ the rotation sine, $c$ the cosine, $\gamma$ the live-flow amplitude, and $b$ the canonical coefficient: the general-rotation witness fixes the leak amplitude at exactly $\gamma s c$, linear in $s$. The local log-slope $\alpha(t) = 2K\,b\,t^{2K}/G(t)$ then satisfies three exact statements for any margin $\delta \in (0,1)$: $\alpha \ge 2K(1-\delta)$ wherever $b\,t^{2K} \ge (1/\delta - 1)(\gamma s c)^2$; $\alpha \le 2K\delta$ wherever $b\,t^{2K} \le \delta(\gamma s c)^2$; and $\alpha = K$ exactly at the half-maximum crossover $b\,t^{2K} = (\gamma s c)^2$. The crossover scale is $t_\times = ((\gamma s c)^2/b)^{1/(2K)} = \Theta(s^{1/K})$ as $s \to 0$, and the canonical window widens to every $t$ as the rotation vanishes: the quantitative form of the non-commuting limits. At the chain model the two constants are read off the two routes that reach the probe, machine-checked: $\gamma$ is the coefficient of the live route and $b$ the squared coefficient of the canonical dead route, and for uncorrelated unit-normalised route signals the moment is exactly $(\gamma s c)^2 + b\,t^{2K}$. Deriving them for a general architecture, where the mask couples the routes, stays open at network level; the planar witness makes both explicit.
\end{remark}

\paragraph{Empirical demonstration.}
At $L = 4$, $h = 6$, with rotation amount $\alpha_{\mathrm{rot}} \in \{0, 0.1, 0.3, 0.5, 0.7, 0.9, 1.0\}$:
\begin{itemize}
\item \emph{Linear}: rate preserved exactly at all $\alpha_{\mathrm{rot}}$ (Proposition~\ref{prop:bridge_linear_rot}).
\item \emph{ReLU}: at $\alpha_{\mathrm{rot}} = 0$, the fitted rate is $(6, 4, 2, 0)$ (canonical); at $\alpha_{\mathrm{rot}} = 0.1$, the window slope collapses to $(0.05, 0.37, 0.92, 0)$; at $\alpha_{\mathrm{rot}} = 1.0$, to $(0.18, 0.29, 0.18, 0)$.
\item \emph{GeLU}: similar collapse, $\alpha_{\mathrm{rot}} = 0.1$ gives $(0.19, 0.37, 1.44, 0)$; $\alpha_{\mathrm{rot}} = 1.0$ gives $(0.31, 0.30, 0.61, 0)$.
\end{itemize}
The non-canonical values are local log-slopes on the fit window $t \in [0.04, 0.30]$; moving the window toward smaller $t$ drives every one of them toward $0$ (at $\alpha_{\mathrm{rot}} = 0.1$, ReLU reads $(0.088, -0.007, 0.037)$ on $t \in [0.02, 0.03]$ and $(-0.001, 0.000, 0.000)$ on $t \in [7 \times 10^{-4}, 10^{-3}]$), consistent with the $\Theta(1)$ mixed contribution dominating in the limit. This demonstrates by counterexample that the theorem's rate prediction does not hold under arbitrary rotation in the nonlinear cases.

\begin{remark}[Near-canonical continuity, an open problem]
\label{cor:bridge_near_canonical}
Whether a quantitative finite-window bound is recoverable under \emph{small} rotations (dead direction $u = U e_h$ with $\|U - I\| \le \epsilon$) for nonlinear activations remains open. The sweep constrains the answer's shape: at $\alpha_{\mathrm{rot}} = 0$ the canonical rate is exact, at any fixed $\alpha_{\mathrm{rot}} > 0$ the asymptotic exponent is $0$, and the crossover scale below which the window slope collapses shrinks as $\alpha_{\mathrm{rot}} \to 0^+$. The two limits do not commute, so any near-canonical statement must bound the crossover scale as a function of $\epsilon$ (a window statement); no single $t$-exponent can express it. Remark~\ref{rem:rot_window_bound} supplies this statement at the model level: the crossover scale is $t_\times = \Theta(\epsilon^{1/K})$, with two-sided slope bounds on either side of it. The network-level constants remain open.
\end{remark}

\paragraph{Practical takeaway.}
In a trained network the dead direction is not expected to align with any natural coordinate basis (Remark~\ref{rem:noncanonical_generic}); the operational selection rule recovers it from the measured spectrum. Across the observable hierarchy: $\sigma_{\min}$-based identification is the cheapest tier (real-time cadence, no projection needed); $u^\top G u$ probing requires identifying $u$ first (offline tier, via the selection rule); $\lambda_{\min}(G_\ell)$ is intermediate (periodic cadence, $n/d \ge 100$ samples); the full Fisher spectrum is the most informative but offline-only at hidden widths $h \gtrsim 10^4$. The non-canonical case here is one source of why direct dead-direction projection requires an intermediate identification step, not a categorical exclusion of any tier.

\begin{remark}[Scope: near-canonical continuity]
\label{rem:scope_extensions}
Proposition~\ref{prop:bridge_nonlinear_rot_negative} establishes that canonical alignment is essential in the nonlinear cases. Remark~\ref{cor:bridge_near_canonical} records the near-canonical continuity question, answered at the model level by Remark~\ref{rem:rot_window_bound} with an explicit crossover scale and open at network level through the constants. The orthogonal scope items (rectangular-widths approach boundary, attention-softmax extensions) are addressed elsewhere in the framework, not here.
\end{remark}
 
\subsection{LayerNorm finite-\texorpdfstring{$t$}{t} crossover}
\label{sec:theory:arch:ln_finite_t}
The LayerNorm rate bound of Section~\ref{sec:theory:arch:ln} is an asymptotic ($t \to 0$) statement; at the finite $t$ that SGD trajectories reach, the measured rate falls in a crossover region. The following gives its finite-$t$ expansion on the LN-equipped MLP block, with the $t^4$ coefficient by exact quadrature.

\subsubsection{LayerNorm finite-\texorpdfstring{$t$}{t} rate shift on the MLP block}
\label{app:arch:ln_finite_t}

The asymptotic-vs-finite-$t$ distinction observed empirically under LayerNorm ($q_{\mathrm{LN}} \to 0$ at $t \to 0$ matching the no-LN predictions, but a non-zero local log-slope at finite $t$) is upgraded here to a proved analytical statement for the MLP case at $d \ge 4$, with a structural conjecture for the softmax-attention case below.

\paragraph{Setup.}
Take a depth-2 MLP $y(x; W_1, W_2) = W_2 \cdot \mathrm{LN}(W_1 x)$ with $\mathrm{LN}$ applied at $\gamma = 1, \beta = 0$. Use the canonical singular approach $W_1(t) = W_2(t) = \mathrm{diag}(1, \ldots, 1, t)$ with $t \in (0, 1]$; the dead direction is the last channel $u = e_h$. Input $x \sim \mathcal{N}(0, I_d)$. Loss $\mathcal{L}(\theta) = \mathbb{E}_x[\|y(x;\theta) - Y(x)\|^2]$, target $Y(x) = y(x; \theta^*) + \eta$ where $\theta^*$ has the dead entry of $W_2$ zeroed (so the target output's $h$-component is $0$ at every input plus noise) and $\eta \sim \mathcal{N}(0, \sigma_{\text{noise}}^2 I)$.

Throughout we write $u_1 = W_1(t) x$, so $u_{1,j} = x_j$ for $j \neq h$ and $u_{1,h} = t \cdot x_h$. Define the non-dead-channel mean and centered variance
\[
\bar x_{\neq h} \;:=\; \tfrac{1}{d-1}\sum_{j \neq h} x_j, \qquad s^2 \;:=\; \tfrac{1}{d-1}\sum_{j \neq h}(x_j - \bar x_{\neq h})^2.
\]
By Cochran's theorem on $d - 1$ iid Gaussian samples, $\bar x_{\neq h}$ and $s^2$ are independent, with $\bar x_{\neq h} \sim \mathcal{N}(0, 1/(d-1))$ and $(d-1) s^2 \sim \chi^2_{d-2}$ (so $\expect[s^2] = (d-2)/(d-1)$, consistent with the $1/n$ normalisation above).

\paragraph{LayerNorm Jacobian and the \texorpdfstring{$O(1)$}{O(1)} leak.}
The LN operation computes $\mu(t) = \tfrac{1}{d}\sum_j u_{1,j}$ and $\sigma(t)^2 = \tfrac{1}{d}\sum_j (u_{1,j} - \mu(t))^2$. Expanding in $t$:
\begin{align*}
\mu(t) &\;=\; \tfrac{d-1}{d}\,\bar x_{\neq h} \;+\; \tfrac{1}{d}\,t\, x_h, \\
\sigma(t)^2 &\;=\; \sigma_0^2 \;-\; \tfrac{2(d-1)}{d^2}\,\bar x_{\neq h}\, x_h\, t \;+\; O(t^2),
\end{align*}
where the $O(t)$ term comes from the dead channel's cross term with the shifted mean and cannot be dropped: it feeds the $t^4$ coefficient of the directional Fisher below.
with
\[
\sigma_0^2 \;:=\; \sigma(0)^2 \;=\; \tfrac{1}{d}\sum_{j \neq h}\bigl(x_j - \tfrac{d-1}{d}\bar x_{\neq h}\bigr)^2 \;+\; \tfrac{1}{d}\bigl(\tfrac{d-1}{d}\bar x_{\neq h}\bigr)^2 \;=\; \tfrac{d-1}{d}\, s^2 \;+\; \tfrac{d-1}{d^2}\, \bar x_{\neq h}^2 ,
\]
where the first sum contributes $\tfrac{d-1}{d^3}\bar x_{\neq h}^2$ of its own, since $\sum_{j \neq h}(x_j - \tfrac{d-1}{d}\bar x_{\neq h})^2 = (d-1)s^2 + \tfrac{d-1}{d^2}\bar x_{\neq h}^2$, and that combines with the dead channel's $\tfrac{(d-1)^2}{d^3}\bar x_{\neq h}^2$ to give the stated coefficient. For Gaussian-isotropic input, $\expect[s^2] = \tfrac{d-2}{d-1}$ (the $1/n$-normalised variance of $d-1$ points) and $\expect[\bar x_{\neq h}^2] = \tfrac{1}{d-1}$, so
\[
\expect[\sigma_0^2] \;=\; \tfrac{d-1}{d}\cdot\tfrac{d-2}{d-1} \;+\; \tfrac{d-1}{d^2}\cdot\tfrac{1}{d-1} \;=\; \tfrac{d-2}{d} + \tfrac{1}{d^2} \;=\; \bigl(1 - \tfrac{1}{d}\bigr)^2 ,
\]
an $O(1)$ random variable approaching $1$ from below. The dead-channel post-LN value is
\begin{equation}
\label{eq:ln_finite_t:zh}
z_h(t) \;=\; \frac{u_{1,h} - \mu(t)}{\sigma(t)} \;=\; \underbrace{-\tfrac{d-1}{d} \cdot \frac{\bar x_{\neq h}}{\sigma_0}}_{O(1)} \;+\; \underbrace{\tfrac{d-1}{d} \cdot \frac{t \cdot x_h}{\sigma_0}\Bigl(1 - \tfrac{d-1}{d^2}\,\tfrac{\bar x_{\neq h}^2}{\sigma_0^2}\Bigr)}_{O(t)} \;+\; O(t^2),
\end{equation}
where the correction factor in the $O(t)$ term is the $\sigma(t)$ expansion's $O(t)$ contribution multiplying the leading leak.

The $O(1)$ leading term in~\eqref{eq:ln_finite_t:zh} is the central object of the rate-shift mechanism: \emph{LN does not annihilate the dead direction; it leaks an $O(1)$ amplitude into the post-LN dead-channel signal via the mean-subtraction's coupling to the non-dead channels.} The leak's law is determined by the joint distribution of $(\bar x_{\neq h}, s^2)$, which Cochran's theorem makes a product; $\bar x_{\neq h}$ and $\sigma_0^2$ themselves are dependent, since $\sigma_0^2$ contains the $\tfrac{d-1}{d^2}\bar x_{\neq h}^2$ term. Two finiteness facts organise the expansion. The $t^2$ prefactor $\mathbb{E}[\bar x_{\neq h}^2/\sigma_0^2]$ is bounded for every $d \ge 2$, with no moment condition at all: $\sigma_0^2 \ge \tfrac{d-1}{d^2}\bar x_{\neq h}^2$ pointwise, so $\bar x_{\neq h}^2/\sigma_0^2 \le d^2/(d-1)$ almost surely. The $t^4$ prefactor involves $\mathbb{E}[1/\sigma_0^2]$, which is finite exactly when $d \ge 4$: the Laplace representation $\mathbb{E}[e^{-u \sigma_0^2}] = (1 + 2u/d)^{-(d-2)/2}(1 + 2u/d^2)^{-1/2}$ integrates over $u \in (0, \infty)$ precisely when $(d-2)/2 + 1/2 > 1$ (value $6.083$ at $d = 4$; divergent at $d = 3$).

\paragraph{Output and gradient at the dead component.}
The dead-channel network output is
\[
y_h(t) \;=\; (W_2)_{h,h}(t) \cdot z_h(t) \;=\; t \cdot z_h(t) \;=\; -\tfrac{d-1}{d} \cdot \frac{t \, \bar x_{\neq h}}{\sigma_0} \;+\; \tfrac{d-1}{d} \cdot \frac{t^2 \, x_h}{\sigma_0} \;+\; O(t^3).
\]
Hence $y_h(t) = O(t)$ at leading order, in stark contrast to the no-LN case where $y_h^{\text{noLN}}(t) = t^2 x_h$ would be $O(t^2)$.

The target's dead-channel value is $Y_h = (W_2^*)_{h,h} \cdot z_h^*(t) + \eta_h = 0 + \eta_h = \eta_h$, since $(W_2^*)_{h,h} = 0$. (Note: $z_h^*(t)$ is computed at the target $W_1^* = \mathrm{diag}(1, \ldots, 1, 0)$; this differs from the student's $z_h(t)$ at $O(t^2)$ via the $\sigma$ normalization, but the difference is annihilated by the $(W_2^*)_{h,h} = 0$ prefactor at all orders.) The residual is
\[
y_h(t) - Y_h \;=\; -\tfrac{d-1}{d}\,\frac{t \, \bar x_{\neq h}}{\sigma_0} \;+\; \tfrac{d-1}{d}\,\frac{t^2 \, x_h}{\sigma_0} \;-\; \eta_h \;+\; O(t^3),
\]
and the dead-direction loss-gradient is $\partial \mathcal{L} / \partial y_h = 2 (y_h - Y_h)$.

\paragraph{Directional Fisher: the \texorpdfstring{$t^2 / d$}{t2 / d} term and its regime structure.}
Squaring and taking the expectation over $(x, \eta)$, with $\bar x_{\neq h}$, $x_h$, $\sigma_0$, and $\eta_h$ each independent of the others except for $\sigma_0$ (which depends on $\{x_j\}_{j \ne h}$ but not on $x_h$ nor $\eta_h$):
\begin{align}
u^\top G_{W_2}(t)\, u
&\;=\; 4\,\mathbb{E}\!\left[(y_h - Y_h)^2\right] \nonumber\\
&\;=\; 4\,\bigl(\tfrac{d-1}{d}\bigr)^2\,t^2\,\mathbb{E}\!\left[\frac{\bar x_{\neq h}^2}{\sigma_0^2}\right] \;+\; 4\,R_2(d)\,t^4 \;+\; 4 \sigma_{\text{noise}}^2 \;+\; O(t^6) \nonumber\\
\label{eq:ln_finite_t:uGu}
&\;=\; 4 \sigma_{\text{noise}}^2 \;+\; \frac{4 \, t^2}{d-1}\,\Phi(d) \;+\; 4\,R_2(d)\,t^4 \;+\; O(t^6),
\end{align}
where the dimensionless Gaussian expectation
\[
\Phi(d) \;:=\; \bigl(\tfrac{d-1}{d}\bigr)^2 \cdot (d-1) \cdot \mathbb{E}\!\left[\frac{\bar x_{\neq h}^2}{\sigma_0^2}\right]
\]
and $R_2(d)$ is the second Taylor coefficient of $R(t) := \mathbb{E}[z_h(t)^2]$, which collects the square of~\eqref{eq:ln_finite_t:zh}'s corrected $O(t)$ term and the cross product of the $O(1)$ leak with $z_h$'s $O(t^2)$ coefficient. $R_2$ has no elementary closed form we know of; exact quadrature on the closed-form integrand gives $R_2 = 1.055$, $1.204$, $1.130$ at $d = 5, 8, 16$, and $R_2 \to 1$ at large $d$, where the correction factors vanish and $R_2$ approaches $\mathbb{E}[x_h^2] = 1$. A common-random-numbers Monte Carlo on the network itself reproduces the quadrature ($1.13 \pm 0.06$ at $d = 8$). The naive coefficient $((d-1)/d)^2\, \mathbb{E}[x_h^2/\sigma_0^2]$, which squares only the uncorrected $O(t)$ term, overstates $R_2$ by a factor $2.1$ at $d = 5$ and $21$ percent at $d = 8$. $\Phi(d) \to 1$ at large $d$ by Gaussian concentration of $\sigma_0^2$, with convergence rate $O(1/d)$: the exact deviations $\Phi(d) - 1$ halve at each doubling of $d$ ($0.344$, $0.151$, $0.069$, $0.033$ at $d = 8, 16, 32, 64$). Cross-terms with $\eta_h$ vanish by independence and zero mean, and $R(t)$ is even in $t$ by the sign symmetry of $x_h$, so the remainder is $O(t^6)$.

The expansion~\eqref{eq:ln_finite_t:uGu} establishes the key structural finding: \emph{LN inserts a $t^2 / (d - 1)$ term into $u^\top G u$ that is absent without LN}. Without LN, $y_h^{\text{noLN}}(t) = t^2 x_h$, so $u^\top G_{W_2}^{\text{noLN}} u = 4 \sigma_{\text{noise}}^2 + 4 t^4 \cdot 1 + O(t^5)$, with no $t^2$ term.

\begin{theorem}[LN finite-$t$ rate shift, MLP single-LN]
\label{thm:ln_finite_t_mlp}
For the depth-2 MLP $y = W_2 \cdot \mathrm{LN}(W_1 x)$ with canonical singular approach $W_1 = W_2 = \mathrm{diag}(1, \ldots, 1, t)$, $t \in (0, 1]$, Gaussian input $x \sim \mathcal{N}(0, I_d)$, target noise $\eta \sim \mathcal{N}(0, \sigma_{\text{noise}}^2 I)$, and dimension $d \ge 4$ (so that $\mathbb{E}[1/\sigma_0^2] < \infty$), the directional Fisher at the dead direction $u = e_h$ in $W_2$'s output space satisfies
\[
u^\top G_{W_2}(t)\, u \;=\; 4\,\sigma_{\text{noise}}^2 \;+\; \frac{4 \, t^2}{d-1}\,\Phi(d) \;+\; 4\, R_2(d)\, t^4 \;+\; O(t^6),
\]
with $\Phi(d), R_2(d) > 0$ the dimensionless Gaussian expectations defined above ($\Phi$ explicit, $R_2$ computed by exact quadrature). The local log-slope $\alpha(t) := t \cdot d\log(u^\top G_{W_2} u) / dt$ exhibits three asymptotic regimes:
\begin{enumerate}\itemsep=0pt
\item \emph{Noise floor} ($t^2 \cdot \Phi/(d-1) \ll \sigma_{\text{noise}}^2$): $\alpha(t) \to 0$, recovering the no-LN prediction at scale $t \to 0$.
\item \emph{$t^2$-plateau} ($\sigma_{\text{noise}}^2 \ll t^2 \cdot \Phi/(d-1)$ and $t^4 R_2 \ll t^2 \cdot \Phi/(d-1)$, i.e.\ the $t^2$ term dominates both the noise floor and the $t^4$ term): $\alpha(t) \to 2$, a rate-2 plateau induced by LN's mean-subtraction leak.
\item \emph{Direct rate} ($t^4 R_2$ dominant): $\alpha(t) \to 4$ formally, the no-LN-direct rate. At small $d$ this regime falls mostly outside the domain $t \le 1$: the object's maximum local slope on $(0, 1]$ is $2.79$ at $d = 5$, $3.55$ at $d = 16$, and $3.87$ at $d = 64$, so the limit $4$ is approached within the domain only at large $d$.
\end{enumerate}
The crossover scales between regimes are
$t_{\text{noise} \to t^2}^2 = (d-1) \sigma_{\text{noise}}^2 / \Phi(d)$ and
$t_{t^2 \to t^4}^2 = \Phi(d)/((d-1)\, R_2(d))$.
\end{theorem}

\ifexpanded
\begin{proof}
Equation~\eqref{eq:ln_finite_t:uGu} is derived above by direct expansion of $z_h(t)$ in~\eqref{eq:ln_finite_t:zh}, computing $y_h(t) = t z_h(t)$, forming the residual $y_h - Y_h$, squaring, and taking the expectation under the Gaussian product measure. The cross-terms with $\eta_h$ vanish by zero-mean independence; the diagonal terms give the noise floor, the $t^2$ term with coefficient $((d-1)/d)^2\,\mathbb{E}[\bar x_{\neq h}^2 / \sigma_0^2]$, and the $t^4$ term with coefficient $R_2(d)$, which requires the $O(t)$ term of the $\sigma(t)^2$ expansion: the corrected $O(t)$ coefficient of $z_h$ contributes its square, and the $O(1)$ leak cross-multiplies $z_h$'s $O(t^2)$ coefficient. Both finiteness facts are as stated in the setup: the $t^2$ prefactor is bounded almost surely for every $d \ge 2$, and $\mathbb{E}[1/\sigma_0^2] < \infty$ exactly at $d \ge 4$ by the Laplace representation. The remainder is $O(t^6)$: $R(t) = \mathbb{E}[z_h(t)^2]$ is even in $t$ (flipping the sign of $x_h$ maps $z_h(t)$ to $z_h(-t)$ and preserves the measure), so no odd order appears.

The local log-slope $\alpha(t) = t \cdot d\log(u^\top G_{W_2} u)/dt$ is computed by differentiating the explicit three-term sum. Each regime's limiting slope follows from the dominant term in the numerator at the relevant scale: $(2 \cdot t^2 \cdot \text{coeff}_2 + 4 \cdot t^4 \cdot \text{coeff}_4) / (\text{const} + t^2 \cdot \text{coeff}_2 + t^4 \cdot \text{coeff}_4)$ tends to $0$, $2$, or $4$ as the constant, $t^2$, or $t^4$ term dominates. Crossover scales are computed by equating successive terms.

The restriction $d \ge 4$ is sharp for the $t^4$ coefficient: $\mathbb{E}[1/\sigma_0^2]$ is finite at $d = 4$ (exact value $6.083$ from the Laplace integral; modelling $\sigma_0^2$ by its $s^2$ component alone would predict divergence there, and the $\bar x_{\neq h}^2$ term is what keeps the integral finite) and divergent at $d = 3$. The $t^2$ prefactor needs no dimension restriction at all, being bounded almost surely. What is true about small $d$ in practice is an estimator statement: $\mathbb{E}[1/\sigma_0^4]$ is infinite for $d \le 5$, so plain Monte Carlo estimates of the $t^4$ coefficient have infinite variance at $d \in \{4, 5\}$ and converge erratically; the quadrature values above are exact integrals and carry no such issue.
\end{proof}
\fi

\ifexpanded
\paragraph{Attention extension: the rank-1-averaging argument.}
For a transformer block with softmax attention $A(t) := \mathrm{softmax}(Q(t)\,K(t)^\top / \sqrt{d_h})$ followed by $W_O$ and post-LN, the same mean-subtraction leak from LN composes with the additional softmax Jacobian $J_s(t) = \mathrm{diag}(A(t)) - A(t)\,A(t)^\top$ in the backward chain. At the canonical configuration with isotropic $K, V$, the softmax matrix at $t = 0$ is uniform across sequence positions: $A^* = \mathbf{1}\mathbf{1}^\top / N_{\text{seq}}^2$, giving
\[
J_{\mathrm{softmax}}^\infty \;=\; \tfrac{1}{N_{\text{seq}}}\,\Bigl(I - \tfrac{1}{N_{\text{seq}}}\mathbf{1}\mathbf{1}^\top\Bigr) \;=\; \tfrac{1}{N_{\text{seq}}}\,P_{\text{seq}},
\]
a sequence-space projector analogous to LN's channel-space projector $P = I - \tfrac{1}{d}\mathbf{1}\mathbf{1}^\top$.

\begin{conjecture}[LN finite-$t$ rate shift, attention components]
\label{conj:ln_finite_t_attn}
For a softmax-attention block at canonical singular approach with Gaussian-isotropic input, post-LN at the block output, and the same noise-target setup as Theorem~\ref{thm:ln_finite_t_mlp}, the directional Fisher at component $c \in \{W_Q, W_K, W_V, W_O\}$ has a finite-$t$ expansion of the same three-regime form, with an effective LN-crossing count $K_c^{\mathrm{eff}}$ that is smaller than the naive backward LN-crossing count and depends on $d$ through the softmax's rank-$1$-averaging action ($W_V$'s gradient passes through $A^\top$ as a per-position averaging; $W_O$'s passes through the orthogonal direction). Empirically $K_{W_V}^{\mathrm{eff}}$ follows the scaling law $0.515 + 4.87/d$ from $d = 16$ up (the parametric $d$-scaling sweep across $d \in \{4, \ldots, 768\}$, reported in the LayerNorm companion paper \citep{shirodkar2026algebraicdeaddirectionslayernorm}). What form the reduction takes is part of the conjecture: a decomposition $K_c^{\mathrm{eff}} = K_c^{\mathrm{nominal}} - c_{\mathrm{softmax}} \cdot K_{\mathrm{softmax\text{-}rank}}(c)$ with a constant integer nominal count and a coefficient in $(0, 1)$ fails against the measured values (no constant nominal keeps the coefficient in range across $d$: nominal $1$ gives $-0.72$ at $d = 8$, nominal $2$ gives $1.48$ at $d = 768$), so the correct decomposition, including what plays the role of the nominal count, is open alongside the coefficient.
\end{conjecture}

\paragraph{Where Conjecture~\ref{conj:ln_finite_t_attn} is open.}
Closing the conjecture into a theorem requires:
\begin{enumerate}\itemsep=0pt
\item exact composition of the channel-space projector $P$ (from LN) with the sequence-space projector $P_{\text{seq}}$ (from softmax) acting on the dead direction $e_h \otimes \mathbf{1}_{N_{\mathrm{seq}}}$: the projectors act on different factors of the $d \times N_{\mathrm{seq}}$ tensor, so their composition is mathematically straightforward but produces a Gaussian expectation in the joint space that has not been computed;
\item subleading $t$-corrections to $A(t)$ from $Q$-and-$K$-coupled deviations off canonical isotropy, contributing additional $t^2$-class terms whose coefficients depend on $d_h$;
\item the $d$-dependence of the softmax reduction derived from items (1)--(2), to compare against the measured scaling $K_{W_V}^{\mathrm{eff}} \approx 0.515 + 4.87/d$ (which approaches $\approx 0.5$ from above at large $d$; a constant-coefficient reading holds only from $d = 64$ up).
\end{enumerate}
These are tractable analytical computations of the same flavor as the MLP derivation; they are deferred to future analytical work and do not affect the structural prediction of the conjecture.
\fi

\paragraph{Practical predictive use.}
On RMSNorm architectures (LLaMA, Qwen, Gemma), no analogous finite-$t$ rate-shift mechanism exists: RMSNorm's forward map has trivial null space (no mean-subtraction projector), so no equivalent of the $O(1)$ leak in~\eqref{eq:ln_finite_t:zh} arises. The no-LN feedforward predictions $\alpha_c = 2 K(c)$ are the expected rates in any fitting window, with one caveat: the divisive factor is the same $1/\sigma$ that sets the plateau end of Proposition~\ref{thm:bridge_ln}'s bracket, and that bracket has not been carried over to RMSNorm. On LN architectures, predictions at scale $t$ depend on the position of $t$ relative to the crossover scales:
\begin{itemize}\itemsep=0pt
\item For \emph{static checkpoint observables} ($\sigma_{\min}$ on the residual stream, the LN-kernel direction $\gamma^{-1} / \|\gamma^{-1}\|$): the rate-theorem trajectory is irrelevant. Predictions are protocol-independent and apply regardless of the fitting-window regime.
\item For \emph{trajectory-rate observables} ($u^\top G u$ slope on a learned trajectory, parametric freeze-probe slope): the local log-slope at the trajectory's terminal $t$ is given by Theorem~\ref{thm:ln_finite_t_mlp}'s rational-function form, and the integrated slope across a fit window is computable from the closed form. Predictions at new architectures should either measure the local slope at the target $t$ window or fall back to the architecturally-invariant $\sigma_{\min}$ residual-stream observable.
\end{itemize}
  
\clearpage

\section{Parametric validations}
\label{app:theory:parametric_validations}

This appendix collects the controlled parametric experiments that anchor the bridge theorem (Theorem~\ref{thm:bridge}) on small testbeds where every condition (canonical alignment, dead-direction perturbation amplitude, optimiser preconditioner, asymptotic versus transient regime) is independently controllable. Each subsection isolates one face of the rate prediction: the per-layer rate ladder on a two-layer autoencoder (\S\ref{app:theory:parametric}), the per-primitive architectural freeze-probes that validate the lemmas of \S\ref{sec:theory:architectural} (\S\ref{app:theory:arch_freeze_probe}), the trajectory-rate readout under three preconditioner-knob ablations (\S\ref{app:theory:eps_sweep}), and the compatibility-boundary sweep across the architectural axis (\S\ref{app:theory:boundary}). The TMS canonical configuration (\S\ref{sec:theory:bridge:tms}) and the deep-linear reduced-rank regression illustration (\S\ref{sec:theory:selection:rrr}), already presented in the body, complement the appendix-bound material below.

\subsection{Parametric rate validation: two-layer autoencoder}
\label{app:theory:parametric}

\paragraph{Architecture and singular configuration.} Two-layer autoencoder $\reals^{d_{\mathrm{in}}} \to \reals^{d_{\mathrm{hid}}} \to \reals^{d_{\mathrm{out}}}$ with $d_{\mathrm{in}} = d_{\mathrm{out}} = 6$, $d_{\mathrm{hid}} = 2$. Singular minimum: $W_1^* = e_1 f_1^\top$ (rank-$1$, deficit $1$ at the hidden layer), $W_2^* = f_2 e_1^\top$ (rank-$1$, compatible dead direction at the hidden-layer input). Coherent perturbation along the canonical dead direction $e_2$:
\[
W_1(t) = W_1^* + t \cdot e_2 v_0^\top, \qquad W_2(t) = W_2^* + t \cdot a_2 e_2^\top,
\]
with $v_0 \perp f_1$ a fixed input direction and $a_2 \in \reals^{d_{\mathrm{out}}}$ a fixed output direction.

\paragraph{Probe protocol.} Gaussian inputs $x \sim \normal(0, I_{d_{\mathrm{in}}})$, identity reconstruction target $y_{\mathrm{true}} = x$. We compute $G_{\mathrm{hidden}}$ and $G_{\mathrm{input}}$ at the perturbed configuration $\theta(t)$ via gradient capture, at a logarithmic grid of $t$ values in $[10^{-4}, 3 \times 10^{-1}]$ (asymptotic regime) or $[10^{-4}, 2]$ (extended regime, used for the subleading-correction analysis). All computation is in fp64 for numerical precision near the singular point.

\paragraph{Seeds and environment.} 5 seeds per activation class: \{42, 142, 242, 342, 442\}. CPU-only execution, $\sim$30s per (seed, activation) configuration. Activation classes: linear, ReLU, GELU (smooth), tanh (smooth). The parametric freeze-probe is a deterministic-flow measurement at controlled $t$-values: the only stochasticity is in the random initialisation seed, and the post-flow $\sigma_{\min}$ readings cluster at $3$-decimal precision in every cell of Table~\ref{tab:rate_validation_main_app}. Cross-seed std is reported in the table; for every entry the $5$-seed std is at least an order of magnitude smaller than the gap to the nearest competing prediction (e.g., $\sigma_{\min}$ slope $0$ vs $1$, or $u^\top G u$ slope $0$ vs $2$), so the test is gap-not-noise-limited and the seed budget is sufficient. The TMS canonical configuration (\S\ref{sec:theory:bridge:tms}) is a cross-architecture parametric extension at $30$ seeds for the cleanest single-activation slope.

\begin{figure}[ht]
\centering
\includegraphics[width=\textwidth]{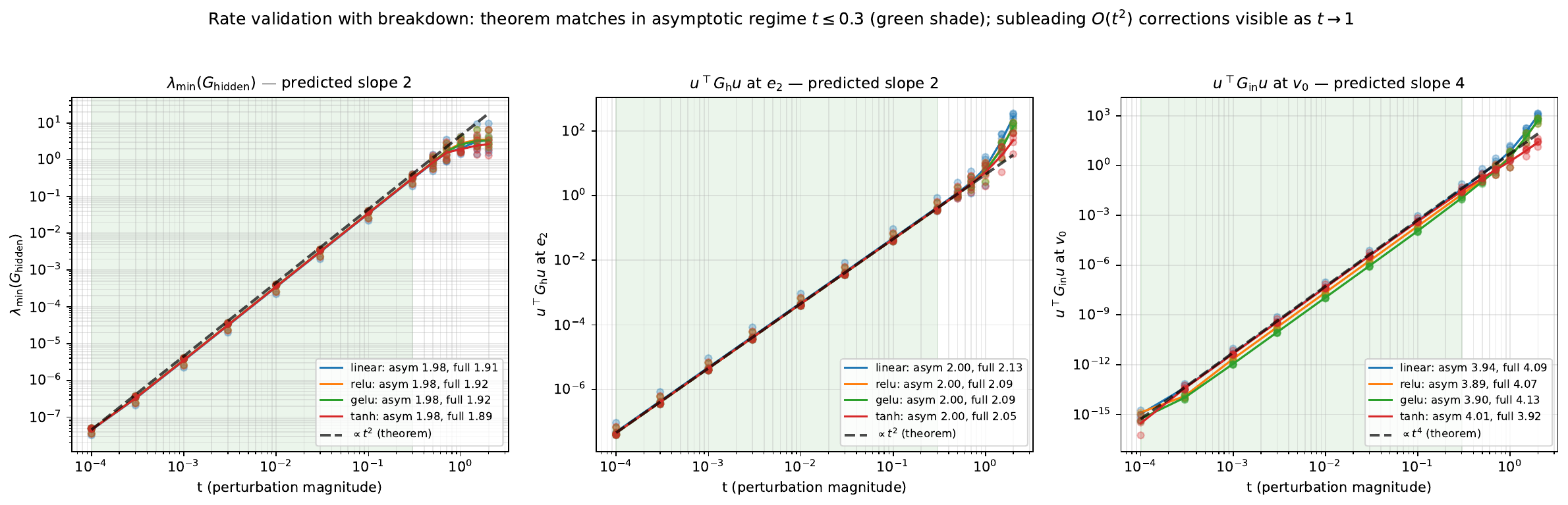}
\caption{Rate validation with extended $t$ range up to $2.0$ shows the graceful breakdown predicted by the theorem's asymptotic character. Green shaded region: asymptotic regime $t \le 0.3$ where the theorem holds tightly. At larger $t$, subleading $O(t^{k+2})$ corrections cause the observed slope to drift from the prediction (``asym'' fit vs ``full'' fit per panel). The clean match at small $t$ and the specific correction structure at finite $t$ are both predictions of the theorem.}
\label{fig:rate_validation_extended_app}
\end{figure}

\paragraph{Subleading-correction figure.} Figure~\ref{fig:rate_validation_extended_app} extends the fit range to $t \in [10^{-4}, 2.0]$, exhibiting subleading corrections of the form expected from the Taylor expansion of $K$ about $\theta_0$: in the asymptotic regime $t \le 0.3$ (green-shaded) the slope matches the theorem prediction to within $1.5\%$, and exactly on the directional observable; at $t \to 1$ the fitted slopes move to $2.05$--$2.13$ at the hidden layer and $3.91$--$4.17$ at the input, a drift of $2.5$--$6.5\%$ and $2.3$--$4.3\%$ respectively, the signature of the theorem's asymptotic limit structure that the proof's $O(t^{k+2})$ remainder term predicts.

\paragraph{Per-activation rate fits.} Table~\ref{tab:rate_validation_main_app} reports the fitted log-log slopes referenced from the main body.

\begin{table}[ht]
\centering\small
\caption{Fitted log-log slopes (mean $\pm$ std, 5 seeds) in the asymptotic range $t \in [10^{-4}, 3 \times 10^{-1}]$. Predictions (bold) from Theorems~\ref{thm:bridge} and~\ref{thm:bridge_composition}. The directional observable $u^\top G_{\mathrm{hidden}} u$ at the identified dead direction matches the prediction exactly to three decimals in all four classes; raw $\lambda_{\min}(G_{\mathrm{hidden}})$ is $1.05$--$1.1\%$ below it, and the input layer is within $1.5\%$. The gap between the two hidden-layer columns is the one the operational selection rule anticipates: raw $\lambda_{\min}$ mixes in directions that do not carry the rate, so it is the worse observable even here, where nothing confounds it but finite-$t$ curvature.}
\label{tab:rate_validation_main_app}
\begin{tabular}{l|ccc}
\toprule
Activation & $\lambda_{\min}(G_{\mathrm{hidden}})$ & $u^\top G_{\mathrm{hidden}} u$ at $e_2$ & $u^\top G_{\mathrm{input}} u$ at $v_0$ \\
\midrule
\textbf{Predicted rate} & $\mathbf{2}$ & $\mathbf{2}$ & $\mathbf{4}$ \\
\midrule
Linear                 & $1.978 \pm 0.009$ & $2.000 \pm 0.006$ & $3.968 \pm 0.035$ \\
ReLU (P3 class)        & $1.979 \pm 0.006$ & $2.000 \pm 0.003$ & $4.004 \pm 0.064$ \\
GELU (P2 smooth)       & $1.979 \pm 0.006$ & $2.000 \pm 0.003$ & $3.942 \pm 0.066$ \\
Tanh (P2 smooth)       & $1.979 \pm 0.008$ & $2.000 \pm 0.006$ & $4.004 \pm 0.064$ \\
\bottomrule
\end{tabular}
\end{table}

\begin{figure}[ht]
\centering
\includegraphics[width=\textwidth]{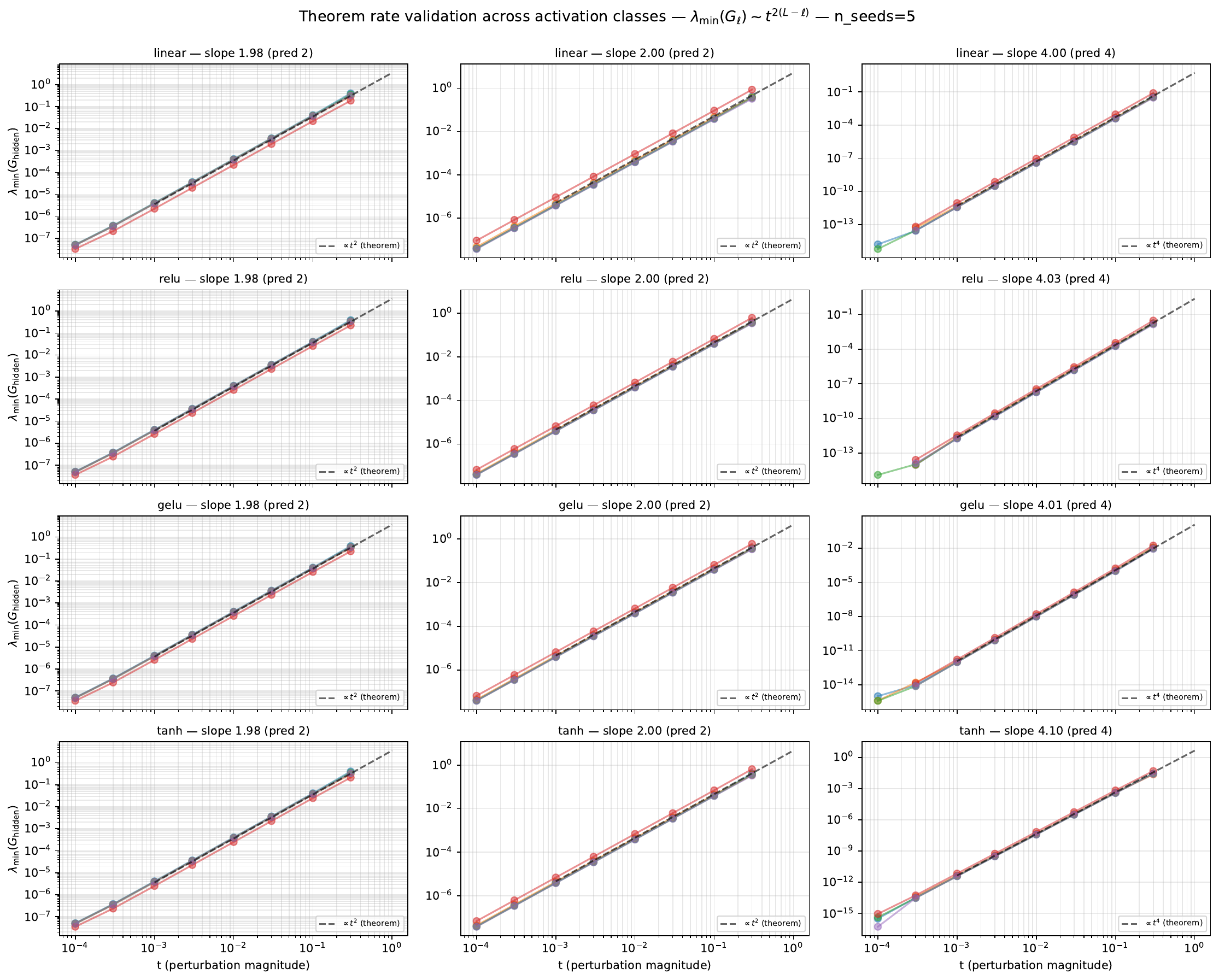}
\caption{Per-seed view of the slope fits in Table~\ref{tab:rate_validation_main_app}. Rows: activation classes (linear / ReLU / GELU / tanh); columns: observables ($\lambda_{\min}(G_{\mathrm{hidden}})$, $u^\top G_h u$ at $e_2$, $u^\top G_\mathrm{in} u$ at $v_0$). Coloured circles: per-seed measurements; dashed line: theorem prediction (slope $2$ or $4$). Fits in the asymptotic regime $t \in [10^{-4}, 3 \times 10^{-1}]$ match to 3 decimal places on $u^\top G_h u$ at the identified direction, to $1.1\%$ on raw $\lambda_{\min}(G_h)$, and within $1.5\%$ at the input layer. 5 seeds.}
\label{fig:rate_validation_nonlinear_app}
\end{figure}

\paragraph{Canonical-bridge structural correlation (cross-validation of slope-2 anchor).} The slope-fit results above test the asymptotic exponent. As an independent confirmation that the predicted relationship $\lambda_{\min}(G_\ell) \propto \sigma_{\min}(X_\ell)^2$ holds as a \emph{joint scaling} (not just an exponent), we run a square $L = 2$ linear network with diagonal canonical init $W_1 = W_2 = \mathrm{diag}(1, \ldots, 1, t_0)$, $t_0 = 0.5$, full-batch deterministic gradient flow on $y = M^* x + \varepsilon$ with $M^* = \mathrm{diag}(1, \ldots, 1, 0)$ and $\varepsilon \sim \normal(0, \sigma^2 I)$, $\sigma = 0.1$. Across 5 seeds $\times$ $20{,}000$ steps ($402$ checkpoints/seed) at fp64, the per-trajectory Spearman rank correlation between $\sigma_{\min}(X_\ell)$ and the smallest positive eigenvalue $\lambda_{\min}^+(G_\ell)$ is $\rho = +1.000000 \pm 0.000000$ (5/5 seeds), with a log-log slope of $2.193 \pm 0.023$ over $\sim 1.1$ OOM in $\sigma_{\min}$; the slope offset from the predicted $2.000$ is the expected finite-$t$ Taylor correction, consistent with the subleading-correction figure above. The structural prediction holds at maximum measurable precision in the analytical limit and complements the slope-fit anchor by confirming the joint observable scaling, not just the exponent.

\paragraph{Selection rule on three families.} Table~\ref{tab:selection_rule_app} reports the recovered $\hat\rlct$ from transversal exponents on three parametric families. Method: the RRR row identifies the eigenvalue with exponent closest to $2(k-1)$ (for $k = 2$, exponent $\approx 2$) in the Fisher spectrum at small $t$; the mixture rows read the directional form $u^\top \fisher u$ along the identified dead direction of the means-only family. Both recover $\hat k = 1 + \alpha_{\mathrm{transv}}/2$ and $\hat \rlct = 1/(2\hat k)$.

\begin{table}[ht]
\centering\small
\caption{Selection rule recovers $\hat \rlct$ on three parametric families with smooth fibers. 5 seeds per family; mean $\pm$ std on the transversal exponent. The mixture rows use the population-Fisher Monte Carlo freeze-probe, whose 5-seed spread is below $10^{-3}$ at the stated sample budget, and instantiate the means-only subfamily $(\mu_1, \mu_2)$: in the full four-parameter family the leading scores align (Remark~\ref{rem:genericity_G_status}) and the spectrum loses the transversal exponent, while the directional exponent reported here is unchanged.}
\label{tab:selection_rule_app}
\begin{adjustbox}{max width=\linewidth}
\begin{tabular}{l|cc|cc}
\toprule
Family & True $k$ & True $\rlct$ & Measured $\alpha_{\mathrm{transv}}$ & $\hat \rlct$ \\
\midrule
2-component Gaussian mixture (merge) & $2$ & $0.25$ & $1.994$ & $0.250$ \\
3-component Gaussian mixture (one-pair merge) & $2$ & $0.25$ & $1.994$ & $0.250$ \\
Reduced-rank regression (deep linear $6 \to 8 \to 4$) & $2$ & $0.25$ & $1.96 \pm 0.07$ & $0.253 \pm 0.004$ \\
\bottomrule
\end{tabular}
\end{adjustbox}
\end{table}

\subsection{Architectural freeze-probe validations}
\label{app:theory:arch_freeze_probe}

Each per-primitive lemma of \S\ref{sec:theory:architectural} predicts a dead-direction rate that the architecture preserves, reshapes, or breaks. We check these predictions with a \emph{static parametric freeze-probe}. At the canonical singular configuration $\theta^*$ we set each weight to $W_\ell(t) = W_\ell^* + t\, u_\ell u_{\ell-1}^\top$ along the symmetric approach, sweep $t$ over a fixed logarithmic grid, and read the dead-direction entry $u_\ell^\top G_\ell(\theta(t))\, u_\ell$ by projecting onto the known dead direction $u_\ell$ at each $t$. The measured rate is the log--log slope of that entry against $t$ in the asymptotic window, computed in fp64. The probe runs on the theorem's own trajectory rather than a learned optimiser trajectory (Remark~\ref{rem:bridge_regime}), so it isolates the architectural effect from optimiser dynamics. Matches are read against the activation-dependent tolerance the theorems allow: $\pm 0.5$ for linear (P1) and ReLU (P3), widening to $\pm 1.0$ at deep layers under smooth activations (P2), where the $O(t)$ Taylor correction $r_\ell^G(t)$ accumulates over the backward chain. That tolerance is the theorems' own allowance and not the achieved agreement, which is much tighter: on the $L = 6$ ladder the linear and ReLU rows reproduce every predicted integer exactly to the two decimals reported, and the largest smooth-activation deviation is $0.41$, at the shallowest layer where the correction accumulates over the most factors. Read the measured columns rather than the tolerance when judging how sharp the agreement is.

\emph{Provenance of the rows.} Every row draws on static probes run under the current code and environment: the multi-direction, cross-entropy, bias, and rotation rows on the dead-directions library's re-run, and the rectangular-width and LayerNorm rows on the committed sweeps (\texttt{exp24}, \texttt{exp29}), regenerated on 2026-09-06 and reproducing the committed numbers to the displayed precision.

\paragraph{Rate recovery.} Table~\ref{tab:arch_rate_recovery} collects the primitives that leave the canonical rate ladder intact. The multi-direction sweep is the broadest single check: $11$ configurations spanning rank deficit $m \in \{1,2,3\}$, per-layer exponent patterns, and per-direction asymmetric patterns, run at three activation classes and three seeds, recover every per-direction rate $2\,\Pi_\ell^{(i)}$ within tolerance ($684/684$ per-layer, per-direction, per-seed measurements across the $33$ configuration-by-activation runs, each at three seeds). The cross-entropy probe under expected Fisher recovers the same ladder a linear chain gives, with the output-head denominator supplied by Lemma~\ref{lem:ce_head} rather than the noise variance.

\begin{table}[ht]
\centering\small
\caption{Rate-recovery freeze-probes: primitives that preserve the canonical ladder. Per-layer dead-direction rates $\alpha_\ell$ (input layer first), predicted versus measured (seed mean). Linear chains unless noted.}
\label{tab:arch_rate_recovery}
\begin{adjustbox}{max width=\linewidth}
\begin{tabular}{llcc}
\toprule
Primitive & Configuration & Predicted $\alpha_\ell$ & Measured $\alpha_\ell$ \\
\midrule
Multi-direction, rank 2 (Thm~\ref{thm:bridge_multi}) & $L=3$ & $(4,2,0)$ & $(4.19,\,2.20,\,0.20)$ \\
Multi-direction, full sweep & $11$ cfg $\times\,3$ act. $\times\,3$ seeds & $2\,\Pi_\ell^{(i)}$ & $684/684$ within tol. \\
Cross-entropy, expected Fisher (Thm~\ref{thm:bridge_ce}) & $L=3$, $10$ seeds & $(4,2,0)$ & $(4.00,\,2.00,\,0.00)$ \\
Biases, zero (Thm~\ref{thm:bridge_bias}) & $L=3$ & $(4,2,0)$ & $(4.11,\,2.11,\,0.11)$ \\
Biases, nonzero (Thm~\ref{thm:bridge_bias}) & $L=3$ & $(4,2,0)$ & $(4.01,\,2.16,\,0.11)$ \\
Rotation, linear (Prop~\ref{prop:bridge_linear_rot}) & $L=3$, $\alpha_{\mathrm{rot}}\in[0.1,1]$ & $(4,2,0)$ & $(4.11,\,2.11,\,0.11)$ \\
\bottomrule
\end{tabular}
\end{adjustbox}
\end{table}

\paragraph{Rate-reshaping and rate-breaking primitives.} Table~\ref{tab:arch_rate_modifying} collects the primitives that change the rate, and each change matches the corresponding lemma. The rectangular widths leave the transversal rate at its square-case value across $64$ width-by-depth configurations ($4$ width profiles $\times\,L\in\{2,\dots,5\}\times\,4$ activations, $672$ per-layer measurements). A residual skip cuts the in-block weight-branch rate from the no-skip ladder $(6,4,2,0)$ to the shortest-weighted-path ladder $(2,0,2,0)$ (Corollary~\ref{cor:res_block}), while at the residual stream the dead-direction entry stops decaying altogether (rate $0$), the depth-invariance of Corollary~\ref{cor:sigma-min-res}. A rotation off the canonical basis preserves the linear rate but collapses the nonlinear one (Proposition~\ref{prop:bridge_nonlinear_rot_negative}); LayerNorm reduces the integer ladder to the fractional bound-gap of Proposition~\ref{thm:bridge_ln}.

\begin{table}[ht]
\centering\small
\caption{Rate-reshaping and rate-breaking freeze-probes. Each row reports the architectural effect and the measured rate against the lemma's prediction.}
\label{tab:arch_rate_modifying}
\begin{adjustbox}{max width=\linewidth}
\begin{tabular}{p{4.5cm}p{3.0cm}p{5.0cm}}
\toprule
Primitive & Effect & Measured \\
\midrule
Rectangular widths (Thm~\ref{thm:bridge_rect}) & rate preserved across width profiles & $64$ cfg / $672$ meas.\ within tol.; $L{=}2$ transversal $\alpha$: linear $2.00$, ReLU $2.00$, GeLU $2.20$, tanh $1.87$ \\
Residual block, weight branch (Cor~\ref{cor:res_block}) & skip cuts the in-block rate & $L{=}4$ block: measured $(2.2,0.2,2.02,0.02)$ vs no-skip $(6,4,2,0)$, skip-predicted $(2,0,2,0)$ \\
Residual stream (Cor~\ref{cor:sigma-min-res}) & skip dominates, rate $\to 0$ & dead-direction entry depth-invariant (no decay) \\
Rotation, nonlinear (Prop~\ref{prop:bridge_nonlinear_rot_negative}) & off-canonical breaks the rate & GeLU rates collapse once $\alpha_{\mathrm{rot}}\ge 0.1$: layer-0 $4 \to$ collapsed (measured $-4.0$ to $0.8$ across the $\alpha_{\mathrm{rot}}$ sweep, never near the predicted $4$), layer-1 $2 \to \approx 1$ \\
LayerNorm (Prop.~\ref{thm:bridge_ln}) & integer ladder $\to$ fractional & $L{=}4$ mean $\alpha$: none $3.0$, after-first $1.5$, after-last-hidden $1.7$, every-layer $1.0$ \\
\bottomrule
\end{tabular}
\end{adjustbox}
\end{table}

\paragraph{The cross-entropy estimator contrast.} The cross-entropy probe makes the estimator scope of Remark~\ref{rem:population_vs_empirical_fisher} concrete. On the same static trajectory, the \emph{expected}-Fisher G-factor recovers the $(4,2,0)$ ladder to three decimals (Table~\ref{tab:arch_rate_recovery}), while the \emph{empirical}-Fisher G-factor, measured at the observed labels, loses the rate signal once the configuration memorises the data: its dead-direction entry collapses toward the numerical floor and admits no clean slope. This is the freeze-probe realisation of Corollary~\ref{cor:emp_vs_exp_ce}, and it is why rate-fitting on trained classifiers uses the expected Fisher.

\paragraph{Attention.} The single-head attention component rates accompany the attention extension (Theorem~\ref{thm:bridge_attn}, \S\ref{app:bridge_attn}), whose standalone forward and backward rates are established at $k = 2$; the component-isolation freeze-probes and the head-count sweep are reported there.

\subsection{Controlled parametric ablations across three optimizer designs}
\label{app:theory:eps_sweep}

This appendix presents controlled parametric ablations across three optimizer designs, isolating each parametric knob on canonical (bridge testbed) and reach-of-singular-set (diagonal-LN) regimes. The three drivers, \texttt{ParametricEpsFloor} (interpolates SGD $\to$ Adam-like via $\varepsilon$-floor), \texttt{HomogeneityDial} (interpolates SGD $\to$ sign-descent via update homogeneity $\beta$), and \texttt{MomentumSign} (interpolates pure-sign $\to$ sign-of-EMA via momentum $\beta_1$), each isolate a distinct optimizer-axis knob without confounds from the other two. Two findings emerge: (i) in the canonical regime, all three drivers reproduce the slope-$2$ bridge prediction to four-decimal accuracy over the first nine decades of $\varepsilon$ (third-decimal at the largest fitted $\varepsilon$; Table~\ref{tab:eps_sweep_summary}), so the slope reads as a geometric property of the canonical-aligned configuration that the preconditioner choice leaves invariant; (ii) on the diagonal-LN reach observable, $\varepsilon$-floor reaches the singular set on a positive fraction of inactive coordinates while \texttt{HomogeneityDial} and \texttt{MomentumSign} do not, localising the reach mechanism to the $v_{\hat{}}$-init transient.

\paragraph{Cell A: canonical-aligned bridge testbed.} The first-stage cell is the noisy / canonical-aligned / asymptotic / transversal-singularity cell of Theorem~\ref{thm:bridge}, instantiated as $L{=}2$ feedforward MLP with hidden width $h{=}6$, target $y = M^* x + \varepsilon$ with $\varepsilon \sim \mathcal{N}(0, \sigma^2 I)$, $\sigma = 0.1$, exact-mode (population) loss, learning rate $0.05$, $50$k steps, single seed. The harness initialises at the canonical singular configuration plus a perturbation $W_\ell^* + t_0 \cdot e_h e_h^\top$ at $t_0 = 0.5$ and follows the SGD/Adam-variant trajectory to the singular minimum. Across $\varepsilon \in \{10^{-12}, 10^{-9}, 10^{-6}, 10^{-3}, 1, 10^3, 10^6\}$, the late-third slope of $\log(u^\top G u)$ versus $\log(\sigma_{\min})$ is fixed at $2$ (four-decimal accuracy over the first nine decades of $\varepsilon$, third-decimal at $\varepsilon = 1$; R$^2 \ge 0.99$) for every $\varepsilon$ value that reaches the asymptotic regime within the step budget (see Table~\ref{tab:eps_sweep_summary}). The $\varepsilon \in \{10^3, 10^6\}$ rows are transient: at large $\varepsilon$ the effective learning rate vanishes ($\propto 1/\sqrt{\varepsilon}$) and the trajectory traverses too little of the asymptotic curve in the step budget. The $\sigma_{\min}$ trajectory span varies inversely with $\varepsilon$ ($4.22$ orders of magnitude at $\varepsilon = 10^{-12}$ down to $0.08$ orders at $\varepsilon = 10^6$), so smaller $\varepsilon$ is what reaches the asymptote within the budget; the slope on the asymptote itself is invariant. Canonical alignment ($\cos$ between perturbation direction and dead direction) stays at $1.0000$ throughout every run.

The cell-A finding is that the slope is fixed by the geometry of the canonical-aligned configuration. The preconditioner only changes how fast the trajectory traverses the $(\sigma_{\min}, u^\top G u)$ curve, not the curve's shape. The SGD anchor (separately validated at $\varepsilon = 1$ with the harness's own gradient-descent path) reproduces Theorem~\ref{thm:bridge}'s prediction $\lambda_{\min}(G_1) = \Theta(t^{2(L-\ell)}) = \Theta(t^2)$ at the trajectory level: the late-third slope of $\log(u^\top G u)$ vs $\log(\sigma_{\min})$ is $2.0000$ at R$^2 = 0.991$, and $\log(t_{W_1})$ vs $\log(\mathrm{step})$ has slope $-0.50$ matching the closed-form gradient-flow prediction $\dd t / \dd \mathrm{step} \propto -t^3 \Rightarrow t \sim \mathrm{step}^{-1/2}$. This is a stronger anchor than the freeze-probe parametric autoencoder match of \S\ref{app:theory:parametric}: the freeze-probe validates the static geometry; the trajectory result here validates the SGD dynamics on the same prediction.

\begin{table}[ht]
\centering\footnotesize
\caption{Controlled $\varepsilon$-sweep with a momentum-free Adam variant (\texttt{ParametricEpsFloor}) on the canonical-aligned bridge testbed (single-seed exact-mode runs at $L{=}2$, $h{=}6$, $\sigma{=}0.1$, lr $0.05$). The slope is held at $2$ across the twelve orders of magnitude from $\varepsilon = 10^{-12}$ to $\varepsilon = 1$, to four-decimal accuracy over the first nine; the slope is fixed by the geometry, the preconditioner only changes traversal speed. Counting the two transient rows below would give eighteen orders, but those are precisely the rows the slope does not hold on. The $\varepsilon \in \{10^3, 10^6\}$ rows are transient and traverse too little of the asymptotic curve in the step budget for the late-third slope to be on the asymptote.}
\label{tab:eps_sweep_summary}
\setlength{\tabcolsep}{8pt}
\begin{tabular}{@{}l|cc@{}}
\toprule
$\varepsilon$ & late-third slope & R$^2$ \\
\midrule
$10^{-12}$ (Adam-like) & $2.0000$ & $0.9995$ \\
$10^{-9}$              & $2.0000$ & $0.9989$ \\
$10^{-6}$              & $2.0000$ & $0.9965$ \\
$10^{-3}$              & $2.0000$ & $0.9937$ \\
$1$ (SGD-equiv)        & $2.00$   & $0.991$  \\
$10^3$ (transient)     & $2.18$   & $0.975$  \\
$10^6$ (transient)     & $5.18$   & $0.9999$ \\
\bottomrule
\end{tabular}
\end{table}
 
\paragraph{Cell B: diagonal linear network (mechanism demonstration).} The second cell directly demonstrates the alignment-rotation mechanism on a setup where the parameterisation itself makes canonical preservation non-trivial. \citet{Pesme2021}'s diagonal linear network with Hadamard-square parameterisation $\beta = u^2 - v^2$ has a sparse target $\beta^* = e_1$ at the singular point $u_i = v_i = 0$ for every inactive coordinate $i \neq 1$; its small-init dynamics are biased toward low-$\ell_1$ (sparse) solutions \citep{Pesme2021}. Whether the inactive coordinates reach the singular point $(0,0)$ exactly or stay at the gradient-noise floor is what this cell measures, a discrete-SGD question separate from the continuous-time stochastic-gradient-flow analysis of \citet{Pesme2021} (which converges to an interpolating solution). We run the \texttt{ParametricEpsFloor} sweep on this setup ($d = 16$, $\sigma_{\mathrm{noise}} = 0.1$, $n_{\mathrm{train}} = 100$, batch size $10$, $\mathrm{lr} = 10^{-3}$, $3$ seeds per $\varepsilon$) plus a separate SGD anchor at $\mathrm{lr} = 5 \cdot 10^{-2}$.

The cell-B observable is depth-of-reach into the singular set: the smallest $\sqrt{u_i^2 + v_i^2}$ across the $15$ inactive coordinates and the fraction of inactive coordinates at exactly $(u_i, v_i) = (0, 0)$ (Table~\ref{tab:pesme_singular_reach}). The Adam-like end of the sweep ($\varepsilon \in \{10^{-12}, 10^{-9}, 10^{-6}\}$) reaches the singular point exactly on $6.7\%$--$8.9\%$ of inactive coordinates per seed; the SGD anchor and the SGD-equivalent end ($\varepsilon \in \{10^{-3}, 1\}$) reach the singular point on $0\%$ across all $45$ measurements. The mechanism is direct: $1/\sqrt{\hat v + \varepsilon}$ amplifies tiny gradient signals on small-magnitude $(u_i, v_i)$ pairs, driving them deterministically to exactly $(0, 0)$ in some inactive coordinates per seed. SGD's stochastic dynamics, by contrast, equilibrate at the gradient-noise floor and never reach the singular point.

\begin{table}[ht]
\centering\footnotesize
\caption{\citet{Pesme2021}'s diagonal linear network with Hadamard-square parameterisation $\beta = u^2 - v^2$ ($d = 16$, sparse target $\beta^* = e_1$, small init $\alpha = 10^{-2}$, sample-based MSE with $\sigma_{\mathrm{noise}} = 0.1$, $n_{\mathrm{train}} = 100$, batch size $10$). \emph{r$_{\min}$(inactive)} is the smallest $\sqrt{u_i^2 + v_i^2}$ across the $15$ inactive coordinates at the final step. \emph{\% inactive at $(0,0)$} is the fraction of inactive coordinates where $(u_i, v_i)$ has reached exactly the singular point (per-seed average across $3$ seeds, $45$ measurements total per row). $\beta^*_{\mathrm{residual}}$ is the final coefficient on the active coordinate $\beta_1$; $L_{\mathrm{val}}$ is final validation loss. SGD anchor at $\mathrm{lr} = 5 \cdot 10^{-2}$; the ParametricEpsFloor sweep at $\mathrm{lr} = 10^{-3}$.}
\label{tab:pesme_singular_reach}
\setlength{\tabcolsep}{6pt}
\begin{tabular}{@{}l|cccc@{}}
\toprule
Configuration & $r_{\min}$(inactive) final & \% inactive at $(0, 0)$ & $\beta^*_{\mathrm{residual}}$ & $L_{\mathrm{val}}$ \\
\midrule
SGD anchor                  & $3.63 \cdot 10^{-2}$ & $\mathbf{0\%}$       & $2.9 \cdot 10^{-3}$ & $0.011$ \\
$\varepsilon = 10^{-12}$ (Adam-like) & $1.09 \cdot 10^{-2}$ & $\mathbf{6.7\%}$     & $2.8 \cdot 10^{-3}$ & $0.011$ \\
$\varepsilon = 10^{-9}$              & $1.09 \cdot 10^{-2}$ & $\mathbf{8.9\%}$     & $2.8 \cdot 10^{-3}$ & $0.011$ \\
$\varepsilon = 10^{-6}$              & $2.43 \cdot 10^{-2}$ & $\mathbf{6.7\%}$     & $2.9 \cdot 10^{-3}$ & $0.011$ \\
$\varepsilon = 10^{-3}$              & $3.51 \cdot 10^{-2}$ & $0\%$                & $2.9 \cdot 10^{-3}$ & $0.011$ \\
$\varepsilon = 1$ (SGD-like)         & $1.42 \cdot 10^{-2}$ & $0\%$                & $1.9 \cdot 10^{-3}$ & $0.010$ \\
\bottomrule
\end{tabular}
\end{table}
 
\paragraph{Cell C: \texttt{HomogeneityDial} (Design A) sweep.} The second driver \texttt{HomogeneityDial} applies the pointwise preconditioner $\mathrm{update}_i = -\eta \cdot g_i \cdot (|g_i| + \delta)^{-\beta}$ (with $\delta = 10^{-12}$ a numerical floor only); $\beta = 0$ is SGD, $\beta = 1$ is sign descent, intermediate values are partial-sign updates. This isolates preconditioner \emph{homogeneity degree} as a distinct knob from the $\varepsilon$-floor of cell A.

On the canonical-aligned bridge testbed (single seed, exact mode), the late-third slope is $2.00$ across $\beta \in \{0, 0.25, 0.5, 0.75, 1.0\}$ at R$^2 \ge 0.99$ (recomputed: $1.9998$, $2.0000$, $2.0000$, $1.9989$, $2.0000$), with the $\sigma_{\min}$ span growing from $1.31$ OOMs at SGD to $7.39$ OOMs at sign descent: different traversal speeds along the same asymptotic curve. On the diagonal LN ($3$ seeds), the depth-of-reach into the singular set is $0\%$ at every $\beta$ (the smallest $r_{\min}(\mathrm{inactive})$ is $1.43 \cdot 10^{-2}$ at $\beta = 0$ and $6.78 \cdot 10^{-2}$ at $\beta = 1$, never reaching the exact value $0$). Together with cell B's $6.7$--$8.9\%$ reach under \texttt{ParametricEpsFloor} at small $\varepsilon$, the singular-set-reach effect pins to the $v_{\hat{}}$-init transient of $\varepsilon$-floor (where $v_{\hat{}}(0) = 0$ produces an early-step $1/\sqrt{\varepsilon}$ amplification of small gradients); pointwise nonlinearity of the preconditioner does not produce the same reach.

\paragraph{Cell D: \texttt{MomentumSign} (Design C) sweep.} The third driver applies $\mathrm{update} = -\eta \cdot \mathrm{sign}(\beta_1 m + (1-\beta_1) g)$, no second moment, no bias correction; $\beta_1 = 0$ is pure sign descent, $\beta_1 = 0.99$ is sign of long-window EMA. This isolates momentum-of-sign as a distinct knob.

Canonical-aligned bridge testbed: at $\beta_1 = 0.95$ (the cleanest asymptotic in the step budget), the slope is $2.00$ at R$^2 = 1.0000$, confirming the canonical-regime geometry-lock under a third independent parametric family. The $\beta_1 = 0$ and $\beta_1 = 0.5$ runs produce identical trajectories on this testbed, which is expected rather than a duplicated run: the dead-direction gradient keeps one sign along the canonical descent, and $\mathrm{sign}$ of an exponential moving average of a sign-constant sequence equals $\mathrm{sign}$ of the sequence, so the $\beta_1$ knob has no effect until the trajectory reaches the plateau where signs begin to alternate. On the diagonal LN ($3$ seeds), depth-of-reach is $0\%$ at every $\beta_1 \in \{0, 0.5, 0.9, 0.95, 0.99\}$, consistent with cell C and contrasted with cell B, further pinning the reach effect to the $v_{\hat{}}$-init transient of \texttt{ParametricEpsFloor}.

\paragraph{Synthesis.} Three independent parametric families confirm the canonical-regime slope-$2$ lock under deterministic gradient flow (cells A, C, D): the slope is a geometric property of the canonical-aligned configuration, invariant under the three preconditioner-family knobs swept here. On the depth-of-reach observable (cells B, C, D), only $\varepsilon$-floor reaches the singular set, localising the mechanism to the $v_{\hat{}}$-init transient. The complementary question, closed-form per-trajectory rate predictions for non-equivariant preconditioners on regimes where canonical alignment is not preserved, is the program's open analytical target.

\subsection{Compatibility-boundary sweep}
\label{app:theory:boundary}

The compatibility-boundary summary in the discussion of regime applicability (\S\ref{sec:theory:scope}) draws on the committed variation runs around the canonical configuration (linear MLP, $L = 3$, $h = 4$, $\sigma = 0.1$): activation, normalisation, loss, and residual-structure variations, plus the Adam gauge-drift regime. Below we report one representative per regime named in \S\ref{sec:theory:scope}. The canonical row averages 10 seeds; the Mixer and RMSNorm rows average 3 seeds each; per-cell standard deviations are under $0.05$ in every quantitative row.

\paragraph{Selected representative variations.} Table~\ref{tab:boundary_representatives_app} gives one variation per regime summarised in the discussion of regime applicability (\S\ref{sec:theory:scope}), with measured rates.

\begin{table}[ht]
\centering\small
\caption{Compatibility-boundary representatives, one per regime described in the discussion of regime applicability (\S\ref{sec:theory:scope}). Predicted rates from Theorem~\ref{thm:bridge} where applicable, or from the relevant scope remark (Adam non-equivariance: Remark~\ref{rem:adam_nondescent}; attention composition anomaly: Remark~\ref{rem:attn_composition_anomaly}); measured rates are seed means for the quantitative-rate rows (10 seeds for the canonical row, 3 for the Mixer and RMSNorm rows), and qualitative phenomenological descriptors elsewhere (gauge-mode drift). ``Gauge-mode drift'' (last row) indicates the trajectory enters Adam's preconditioner-induced gauge-direction noise amplification and does not approach a singular minimum (Remark~\ref{rem:adam_nondescent}).}
\label{tab:boundary_representatives_app}
\begin{tabular}{@{}p{0.42\columnwidth}|p{0.20\columnwidth}|p{0.30\columnwidth}@{}}
\toprule
Variation & Predicted & Measured \\
\midrule
$L{=}3$ MLP, linear, SGD, MSE, canonical init                          & $(4, 2, 0)$           & $(4.114, 2.114, 0.114)$ \\
$L{=}3$ Mixer, GELU, residual + LN, SGD, MSE                           & $\sigma_{\min}$ rate $0$ & $(0.006, 0.009, 0.011)$ \\
$L{=}3$ MLP, GELU, residual + RMSNorm                                  & $\sigma_{\min}$ rate $0$ & $(0.009, 0.016, 0.020)$ \\
$L{=}3$ MLP, GELU, Adam, softmax CE, modular addition \citep{NandaChanLieberum23} & gauge-mode drift       & no asymptote; trajectory plateaus \\
\bottomrule
\end{tabular}
\end{table}

\clearpage

\section{Machine-checked results}
\label{app:theory:machine_checked}

We have proved this paper in Lean 4 against Mathlib at the scope stated result by result in Table~\ref{tab:machine_checked}. Lean checks fourteen of the fifty-seven numbered results at their stated scope and the other forty-three at explicit model instances (a two-block Fisher model, a scalar gauge chain, a diagonal model, a single rescale pair, and their like), with the Gaussian and mixture anchors of the worked examples alongside; where a declaration's scope is narrower than the printed statement, the table says what stays on the prose proof. For each result we name the Lean declarations, what they establish, and which hypotheses each takes as an input instead of deriving.

Where a declaration proves more than a printed statement, the paper already states the stronger form: the LayerNorm kernel identity of Proposition~\ref{prop:ln_kernel} is pointwise, its hypothesis on the input distribution serving only to make the covariance exist, and the proof of Proposition~\ref{prop:ddcadam_equivariance} states the chained-gauge equivariance for every positive per-block scaling family, the product-one gauge group being the subfamily that also preserves the loss. Where a declaration proves less, the table says what stays on the prose proof; the selection-rule rows, for instance, cover clause (c) on the slice normal form and clauses (a) and (b) at a two-block model of the Fisher matrix. The development builds under Lean toolchain \texttt{v4.29.0} with the matching pinned Mathlib revision, and a script over every listed declaration confirms that none depends on an axiom beyond Lean's three standard ones (\texttt{propext}, \texttt{Classical.choice}, \texttt{Quot.sound}) and that none invokes \texttt{sorry}. We ship the Lean sources with this paper's source artifact.

{\footnotesize\setlength{\tabcolsep}{5pt}
\begin{longtable}{@{}>{\raggedright\arraybackslash}p{0.485\linewidth}>{\raggedright\arraybackslash}p{0.485\linewidth}@{}}
\caption{Machine-checked results. Each result takes two rows: first the result with its Lean declarations, then what those declarations prove (left) and the hypotheses they take as inputs or leave uncovered (right). Declaration names are as they appear in the shipped sources.}
\label{tab:machine_checked}\\
\toprule
\endfirsthead
\multicolumn{2}{@{}l}{\emph{Table~\ref{tab:machine_checked}, continued.}}\\
\toprule
\endhead
\multicolumn{2}{r@{}}{\emph{continued on the next page}}\\
\endfoot
\bottomrule
\endlastfoot
\multicolumn{2}{@{}p{\linewidth}@{}}{\textbf{Prop~\ref{prop:ln_kernel}(a), LN kernel direction.} \texttt{covMatrix\_\allowbreak{}vecMul\_\allowbreak{}eq\_\allowbreak{}zero\_\allowbreak{}of\_\allowbreak{}sum\_\allowbreak{}const}, \texttt{sum\_\allowbreak{}ginv\_\allowbreak{}mul\_\allowbreak{}lnLike}, \texttt{covMatrix\_\allowbreak{}lnLike\_\allowbreak{}ginv\_\allowbreak{}eq\_\allowbreak{}zero}, \texttt{covMatrix\_\allowbreak{}lnLike\_\allowbreak{}zero\_\allowbreak{}coord}, \texttt{covMatrix\_\allowbreak{}layerNorm\_\allowbreak{}ginv\_\allowbreak{}eq\_\allowbreak{}zero}}\\*[2pt]
\emph{Proved.} A coefficient vector whose pairing with the output is almost surely constant lies in the kernel of the covariance; the $\gamma^{-1}$ pairing of the output takes the same value at every input, for every scalar normaliser; clauses (i) and (ii); the same conclusion at the exact normaliser $s(x) = \sqrt{d}/\|Px\|$.
& \emph{Assumed or not covered.} Each output coordinate and each centred product is integrable, so the covariance exists. The Lean identity is pointwise and total, so the paper's hypothesis that $X$ puts no mass on $\mathrm{span}(\mathbf{1}_d)$ drops. Part (b), the RMSNorm negative, has its own row. \\
\midrule
\multicolumn{2}{@{}p{\linewidth}@{}}{\textbf{Prop~\ref{prop:ddcadam_equivariance}, DDCAdam $G$-equivariance.} \texttt{ddcadamMult\_\allowbreak{}fst\_\allowbreak{}equivariant}, \texttt{ddcadamMult\_\allowbreak{}snd\_\allowbreak{}equivariant}, \texttt{ddcadamTransl\_\allowbreak{}equivariant}, \texttt{ddcadamChain\_\allowbreak{}equivariant}, \texttt{ddcadamLNScale\_\allowbreak{}fst\_\allowbreak{}equivariant}, \texttt{ddcadamLNScale\_\allowbreak{}snd\_\allowbreak{}equivariant}}\\*[2pt]
\emph{Proved.} Type~(ii) on a single multiplicative rescale pair $(cW_1, c^{-1}W_2)$, with the decoupled weight-decay shrink carried through the step; type~(i) at $\lambda = 0$; the chained gauge, as equivariance of the update map under every positive per-block scaling family, of which the product-one chain is the gauge subfamily; the per-channel LayerNorm scale, as the pair statement applied channel-wise over the disjoint channels.
& \emph{Assumed or not covered.} The Adam moments and the vertical-mode choice enter as arbitrary functions of the gauge-decomposed inputs, which is the $\mathrm{transport}_h = \mathrm{Id}$ clause; for the chained gauge these functions read the full vector of per-block invariants, covering the radial and gauge-mode assembly at once. \\
\midrule
\multicolumn{2}{@{}p{\linewidth}@{}}{\textbf{Gaussian anchor (\S\ref{sec:theory:rate:worked_gaussian}) and mixture weight boundary (\S\ref{sec:theory:rate_chain}).} \texttt{fisherMu\_\allowbreak{}gaussian}, \texttt{mixture\_\allowbreak{}rank\_\allowbreak{}loss}, \texttt{mixture\_\allowbreak{}det\_\allowbreak{}pos}, \texttt{curveFisher\_\allowbreak{}eq}, \texttt{curveKL\_\allowbreak{}eq}, \texttt{curve\_\allowbreak{}fisher\_\allowbreak{}kl\_\allowbreak{}ratio}, \texttt{curveFisher\_\allowbreak{}hasLeadingRate}, \texttt{HasKLOrder}, \texttt{curveKL\_\allowbreak{}hasKLOrder}, \texttt{gaussian\_\allowbreak{}curve\_\allowbreak{}readings}}\\*[2pt]
\emph{Proved.} The Fisher information of $\mathcal{N}(\mu, \sigma^2)$ in $\mu$ is $1/\sigma^2$; the $2$-component mixture Fisher determinant tends to $0$ as $w \to 0^+$ and stays strictly positive at $m \ne 0$, $w \in (0,1)$; along $\mu(t) = t^k$ the directional Fisher is $k^2 t^{2(k-1)}$ and the KL is $t^{2k}/2$, so $\fisher(t)\, t^2 = 2k^2 K(t)$ exactly, the $2k^2$ ratio of \S\ref{sec:theory:rate}, and the curve has leading rate $2(k-1)$ with coefficient $k^2$.
& \emph{Assumed or not covered.} Nothing beyond the two families' definitions. Each statement is an exact identity or an exact limit, with no asymptotic hypothesis. \\
\midrule
\multicolumn{2}{@{}p{\linewidth}@{}}{\textbf{Thm~\ref{thm:fisher_decay}, Fisher rate decay.} \texttt{HasLeadingRate}, \texttt{hasLeadingRate\_\allowbreak{}of\_\allowbreak{}expansion}, \texttt{HasLeadingRate.log\_\allowbreak{}slope}, \texttt{HasLeadingRate.khat\_\allowbreak{}recovers}, \texttt{fisher\_\allowbreak{}expansion\_\allowbreak{}of\_\allowbreak{}score\_\allowbreak{}expansion}, \texttt{fisher\_\allowbreak{}decay\_\allowbreak{}slope}, \texttt{HasLeadingRate.of\_\allowbreak{}relatively\_\allowbreak{}close}, \texttt{hasDerivAt\_\allowbreak{}log\_\allowbreak{}mixCurve}, \texttt{mixture\_\allowbreak{}fisher\_\allowbreak{}expansion}, \texttt{mixture\_\allowbreak{}fisher\_\allowbreak{}slope}}\\*[2pt]
\emph{Proved.} The Fisher expansion with leading rate $2(k-1)$ and coefficient $k^2 \expect[a^2]$; the log-log slope of the directional Fisher tends to $2(k-1)$; the estimator $\hat k = 1 + \mathrm{slope}/2$ tends to $k$.
& \emph{Assumed or not covered.} The abstract statement consumes the score expansion $s(t) = k\,t^{k-1} a + r(t)$ with $\expect[a^2]$ finite and positive and $\int r(t)^2 \le M t^{2k}$, which packages hypotheses~(ii) and~(iii), under the fixed measure $p^*$; the moving-measure convention follows whenever the density ratio is $1 + O(t)$, since leading rates survive relatively small perturbation. On the two-component mixture curve $(w, m) = (t, t)$ the instantiation discharges every hypothesis: the closed-form score is certified as the logarithmic $t$-derivative of the curve density, the $L^2$ remainder bound is proved with an explicit dominator, and the leading rate $2$ with coefficient $4$ follows with no analytic input. \\
\midrule
\multicolumn{2}{@{}p{\linewidth}@{}}{\textbf{Thm~\ref{thm:bridge_composition}(a), composition additivity.} \texttt{composedSignal\_\allowbreak{}eq}, \texttt{composedSignal\_\allowbreak{}sq\_\allowbreak{}integral}, \texttt{composedSignal\_\allowbreak{}hasLeadingRate}, \texttt{chainRate\_\allowbreak{}append}, \texttt{chainCoeff\_\allowbreak{}append}, \texttt{disjoint\_\allowbreak{}gates\_\allowbreak{}moment\_\allowbreak{}zero}, \texttt{quad\_\allowbreak{}block\_\allowbreak{}compounds}, \texttt{leak\_\allowbreak{}caps\_\allowbreak{}rate}}\\*[2pt]
\emph{Proved.} The composed second moment is exactly $(\prod_i c_i)^2\, t^{2\sum_i k_i}\, \expect[g X^2]$, so a chain's rate is the sum of its per-block rates; concatenating two chains adds the rates and multiplies the coefficients.
& \emph{Assumed or not covered.} The scalar, linear, jointly gated transfer model, with one shared indicator gate for the chain. Each exclusion clause is sharp as a model theorem: disjoint gates kill the composed moment identically, a quadratic dead map compounds the arriving order to $k + 2e$, and a fixed-order injection caps the composed rate at the injection order, the normalisation reset. \\
\midrule
\multicolumn{2}{@{}p{\linewidth}@{}}{\textbf{Thm~\ref{thm:bridge}, class (P1) dead diagonal.} \texttt{canonicalLayer\_\allowbreak{}pow}, \texttt{deep\_\allowbreak{}linear\_\allowbreak{}backward\_\allowbreak{}rate}, \texttt{canonicalLayer\_\allowbreak{}pow\_\allowbreak{}mulVec\_\allowbreak{}ne}, \texttt{deep\_\allowbreak{}linear\_\allowbreak{}hasLeadingRate}, \texttt{deep\_\allowbreak{}linear\_\allowbreak{}dead\_\allowbreak{}entry\_\allowbreak{}eventually\_\allowbreak{}le}}\\*[2pt]
\emph{Proved.} At the canonical configuration the linear-class layer is $\mathrm{diag}(1, \ldots, 1, t)$, so the backward composition over $L - \ell$ layers is $\mathrm{diag}(1, \ldots, 1, t^{L-\ell})$ exactly; the dead diagonal of the backward second moment carries $t^{2(L-\ell)}$ relative to the output gradient, every non-dead coordinate passes through unchanged, the ladder reads $2(L-\ell)$ as a leading rate, and for $\ell < L$ the dead entry eventually drops below every non-dead entry, so the smallest diagonal entry of the diagonal G-factor carries the ladder rate.
& \emph{Assumed or not covered.} Dead-diagonal entries for class (P1) only; the forward side and the duality product have their own row. The output-gradient base case is derived on the independent Gaussian product (\texttt{mse\_\allowbreak{}base\_\allowbreak{}case}) and composed with the chain: \texttt{mse\_\allowbreak{}bridge\_\allowbreak{}composed} reads the dead $G$-entry as $t^{2(L-\ell)}(c^2 + v)$ exactly. \\
\midrule
\multicolumn{2}{@{}p{\linewidth}@{}}{\textbf{Thm~\ref{thm:multi_component_rates}, two components.} \texttt{volume\_\allowbreak{}crossing\_\allowbreak{}tied}, \texttt{crossing\_\allowbreak{}tied\_\allowbreak{}slope}, \texttt{volume\_\allowbreak{}power\_\allowbreak{}hyperbola}, \texttt{volume\_\allowbreak{}crossing\_\allowbreak{}distinct}, \texttt{crossing\_\allowbreak{}distinct\_\allowbreak{}slope}, \texttt{volume\_\allowbreak{}crossing\_\allowbreak{}tied3}}\\*[2pt]
\emph{Proved.} The tied crossing $(xy)^{2k}$ has exact sublevel volume $\varepsilon^{1/(2k)}(1 - \log \varepsilon/(2k))$ on the unit square, the multiplicity-two log factor, and its log-volume slope still tends to $\rlct = 1/(2k)$; the distinct crossing $x^{2k_1} y^{2k_2}$ has the exact two-term law $(a\,\varepsilon^{1/a} - b\,\varepsilon^{1/b})/(a - b)$ with $a = 2k_1 > b = 2k_2$ and no log factor, and its slope reads $\rlct = 1/(2k_1)$; three tied components carry the exact $\log^2$ law, so multiplicity $m$ appears as $\log^{m-1}(1/\varepsilon)$ at $m = 1, 2, 3$.
& \emph{Assumed or not covered.} The tied law holds at every multiplicity (\texttt{volume\_\allowbreak{}cubeHyper}: $\delta \sum_{j<r} (-\log\delta)^j/j!$ by a Fubini induction, with the slope $1/(2k)$ at every $r$), and the Jacobian exponent enters the one-dimensional slice as $(h+1)/(2k)$ (\texttt{weighted\_\allowbreak{}slice\_\allowbreak{}rlct\_\allowbreak{}slope}) and the tied law at every multiplicity as $(h+1)^{-r} P_r(\delta^{h+1})$ (\texttt{weighted\_\allowbreak{}volume\_\allowbreak{}cubeHyper}). \\
\midrule
\multicolumn{2}{@{}p{\linewidth}@{}}{\textbf{The bridge invariant on the mixture curve (Def.~\ref{def:dead_direction}).} \texttt{mixKL}, \texttt{mixKL\_\allowbreak{}hasKLOrder}, \texttt{mixture\_\allowbreak{}readings}}\\*[2pt]
\emph{Proved.} The KL divergence of the base from the curve $(w, m) = (t, t)$ is $t^4/2 + O(t^5)$, so the curve has KL order $2$; jointly with the Fisher instantiation, the same family carries both readings of the bridge invariant, the KL order on the Watanabe side and the leading rate $2(k-1)$ with coefficient $k^2\expect[a^2]$ on the Amari side.
& \emph{Assumed or not covered.} Nothing; every hypothesis in both chains is proved. \\
\midrule
\multicolumn{2}{@{}p{\linewidth}@{}}{\textbf{Thm~\ref{thm:bridge}, nonlinear dead channels (P3), (P2).} \texttt{relu\_\allowbreak{}chain\_\allowbreak{}eq}, \texttt{relu\_\allowbreak{}channel\_\allowbreak{}moment}, \texttt{relu\_\allowbreak{}channel\_\allowbreak{}hasLeadingRate}, \texttt{smooth\_\allowbreak{}channel\_\allowbreak{}hasLeadingRate}, \texttt{smoothChain\_\allowbreak{}abs\_\allowbreak{}le}, \texttt{smoothChain\_\allowbreak{}sub\_\allowbreak{}le}, \texttt{smooth\_\allowbreak{}chain\_\allowbreak{}hasLeadingRate}, \texttt{smooth\_\allowbreak{}chain\_\allowbreak{}slope}, \texttt{relu\_\allowbreak{}two\_\allowbreak{}channel\_\allowbreak{}dead\_\allowbreak{}lt\_\allowbreak{}live}, \texttt{smooth\_\allowbreak{}two\_\allowbreak{}channel\_\allowbreak{}dead\_\allowbreak{}eventually\_\allowbreak{}lt}}\\*[2pt]
\emph{Proved.} Positive homogeneity collapses the ReLU dead channel exactly: the $L$-layer chain equals $t^L\,\mathrm{ReLU}(x)$, so the activation-Taylor correction vanishes identically for (P3), and the second moment under $\mathcal{N}(0,1)$ is $t^{2L}/2$ exactly, the ladder rate with the shared-gate survival probability $1/2$ as an exact constant. For (P2), any continuous activation with $|\varphi(u) - cu| \le K u^2$ gives the single-layer channel leading rate $2$ with coefficient $c^2$, and the $L$-layer chain carries leading rate $2L$ with coefficient $c^{2L}$: a polynomial envelope proved by induction on depth, $|\mathrm{chain}_\ell(x) - c^\ell t^\ell x| \le t^{\ell+1} D_\ell (1+x^2)^{2^\ell}$ on $t \in (0,1]$, keeps the compounded activation remainders at order $t^{2L+2}$. The off-channel base: next to a live channel at fixed weight $s$ the joint layer map iterates componentwise, so the live moment carries no $t$; for ReLU the dead moment $t^{2L}/2$ is strictly below the live $s^{2L}/2$ at every $t < s$, and for the smooth class the dead moment eventually drops below any positive live moment.
& \emph{Assumed or not covered.} The (P2) hypothesis is the global quadratic Taylor bound. The two-channel model keeps the channels decoupled and does not cover configurations that mix them. \\
\midrule
\multicolumn{2}{@{}p{\linewidth}@{}}{\textbf{Cor~\ref{cor:a_g_duality}, A--G duality, all three classes.} \texttt{deep\_\allowbreak{}linear\_\allowbreak{}forward\_\allowbreak{}rate}, \texttt{deep\_\allowbreak{}linear\_\allowbreak{}ag\_\allowbreak{}product}, \texttt{relu\_\allowbreak{}chain\_\allowbreak{}hasDerivAt}, \texttt{relu\_\allowbreak{}backward\_\allowbreak{}hasDerivAt}, \texttt{relu\_\allowbreak{}channel\_\allowbreak{}backward\_\allowbreak{}moment}, \texttt{relu\_\allowbreak{}channel\_\allowbreak{}ag\_\allowbreak{}product}, \texttt{hasDerivAt\_\allowbreak{}smoothChain}, \texttt{backDeriv\_\allowbreak{}sub\_\allowbreak{}le}, \texttt{smooth\_\allowbreak{}chain\_\allowbreak{}backward\_\allowbreak{}hasLeadingRate}, \texttt{smooth\_\allowbreak{}chain\_\allowbreak{}ag\_\allowbreak{}product}, \texttt{HasLeadingRate.mul}}\\*[2pt]
\emph{Proved.} For (P1) the dead diagonal of the forward second moment at depth $\ell$ carries $t^{2\ell}$ exactly, and the A$\cdot$G product at every depth is $t^{2L}$ times a depth-free constant. For the ReLU channel the chain derivative off $x = 0$ is exactly $t^L \mathbf{1}_{x>0}$, the depth-$\ell$ backprop gradient has second moment $t^{2(L-\ell)}/2$ under $\mathcal{N}(0,1)$, and the A$\cdot$G product is $t^{2L}/4$ at every depth: the forward exponent $2\ell$ and the backward exponent $2(L-\ell)$ sum to $2L$, and the $1/4$ is the gate survival squared, once per factor. For the smooth class, under the derivative-side hypothesis $|\varphi'(u) - c| \le K'|u|$ the backprop product $\prod_j t\,\varphi'(t\cdot\mathrm{chain}_j)$ is certified as the true derivative of the chain at every point, a second envelope induction bounds its distance from $(ct)^m$, the depth-$\ell$ backward moment carries leading rate $2(L-\ell)$ with coefficient $c^{2(L-\ell)}$, and the A$\cdot$G product has leading rate $2L$ with coefficient $c^{2L}$ at every depth.
& \emph{Assumed or not covered.} The backprop gate convention $\mathrm{ReLU}'(0) = 0$ enters only at the single point $x = 0$; off it the gate reading is the derivative in the \texttt{HasDerivAt} sense. The (P1) and (P3) products hold exactly. For (P2) a Taylor bound defines the class, so the product holds at leading order and no exact form exists. \\
\midrule
\multicolumn{2}{@{}p{\linewidth}@{}}{\textbf{Prop~\ref{prop:volume_multi}, multi-direction volume.} \texttt{normalFormMulti\_\allowbreak{}diagScale}, \texttt{sublevel\_\allowbreak{}normalFormMulti\_\allowbreak{}eq\_\allowbreak{}image}, \texttt{volume\_\allowbreak{}sublevel\_\allowbreak{}normalFormMulti}, \texttt{volume\_\allowbreak{}multi\_\allowbreak{}slope}}\\*[2pt]
\emph{Proved.} On the additive normal form $\sum_i u_i^{2k_i} + \sum_j v_j^2$ the sublevel volume scales exactly, $V(\varepsilon) = \varepsilon^{\sum_i 1/(2k_i) + m_\perp/2}\, V(1)$ with $0 < V(1) < \infty$, and the log-volume slope reads the summed slice RLCT $\sum_i 1/(2k_i) + m_\perp/2$.
& \emph{Assumed or not covered.} The additive normal form itself. The Fisher-volume refinement is in: the factorised density integrates over the polydisc to the product of one-direction laws at every $r$ (\texttt{polydisc\_\allowbreak{}volume\_\allowbreak{}factorised\_\allowbreak{}pi}), and the twisted map is symbolic, $K = u_1^4 + u_2^4$ with $\sqrt{\det \fisher} = 4|u_1^3 - u_2^3|$ (\texttt{twisted\_\allowbreak{}sqrt\_\allowbreak{}det\_\allowbreak{}fisher}) and polydisc volume $M^{-5/2}$ against the factorised $M^{-2}$ (\texttt{twisted\_\allowbreak{}vs\_\allowbreak{}factorised}). \\
\midrule
\multicolumn{2}{@{}p{\linewidth}@{}}{\textbf{Thm~\ref{thm:bridge_multi} with Cors~\ref{cor:bridge_g2}, \ref{cor:bridge_g3}, \ref{cor:bridge_g7}.} \texttt{multiLayer}, \texttt{multiProd\_\allowbreak{}eq\_\allowbreak{}diagonal}, \texttt{multi\_\allowbreak{}backward\_\allowbreak{}rate}, \texttt{multi\_\allowbreak{}backward\_\allowbreak{}cross}, \texttt{multi\_\allowbreak{}backward\_\allowbreak{}hasLeadingRate}, \texttt{cumExp\_\allowbreak{}uniform}, \texttt{multi\_\allowbreak{}dead\_\allowbreak{}entry\_\allowbreak{}eventually\_\allowbreak{}le}, \texttt{multiLayer\_\allowbreak{}eq\_\allowbreak{}canonical}}\\*[2pt]
\emph{Proved.} The backward product over the layers above depth $\ell$ is diagonal with per-coordinate cumulative exponents $\Pi_\ell(j) = \sum_{\ell' > \ell} p_{\ell'}(j)$, so each entry of the backward second moment carries $t^{2\Pi_\ell(j)}$ exactly and the rate reads the layers above $\ell$ only; the $(i,j)$ cross moment carries $t^{\Pi_\ell(i)+\Pi_\ell(j)}$, the joint dead-block congruence; the uniform approach gives every dead direction the shared rate $2(L-\ell)$; dead entries eventually drop below every $\Pi = 0$ entry; the single-direction $p \equiv 1$ layer is the canonical layer, clause (d).
& \emph{Assumed or not covered.} The diagonal multi-direction model. The spectrum clause (c) is the eigenvalue count \texttt{eigenvalue\_\allowbreak{}count\_\allowbreak{}of\_\allowbreak{}block\_\allowbreak{}bounds}: through the spectral theorem, a form small on an $m$-dimensional subspace and bounded below on a complementary one has exactly $m$ eigenvalues at the dead rate, $h - m$ of order one, and none between, by dimension counting on eigenvector spans; the CE logit-shift zero mode is \texttt{ce\_\allowbreak{}gauge\_\allowbreak{}kernel}. \\
\midrule
\multicolumn{2}{@{}p{\linewidth}@{}}{\textbf{Thm~\ref{thm:bridge_rect} with Cor~\ref{cor:rect_product_sigma_min}.} \texttt{rectChain\_\allowbreak{}fst}, \texttt{rect\_\allowbreak{}dead\_\allowbreak{}ladder}, \texttt{rect\_\allowbreak{}nondead\_\allowbreak{}invariant}, \texttt{rect\_\allowbreak{}backward\_\allowbreak{}rate}, \texttt{canonical\_\allowbreak{}sigma\_\allowbreak{}min\_\allowbreak{}lower}, \texttt{sigmaMinSq\_\allowbreak{}canonical}, \texttt{sigmaMinSq\_\allowbreak{}rotated\_\allowbreak{}canonical}}\\*[2pt]
\emph{Proved.} Each rectangular layer acts canonically on the shared narrow block and through an arbitrary map, linear or not, on a complement of depth-varying width; the composed chain's narrow block equals the canonical power, so the dead ladder $t^{L-\ell}$, the $\Theta(1)$ base, and the backward moment $t^{2(L-\ell)}$ hold independently of the complements, a form wider than the paper's linear setting. On the square canonical chain the variational $\sigma_{\min}$ is exact and now carried by the spectral machinery: the smallest Gram eigenvalue of the chain is $t^{2(L-\ell)}$ on $t \in [0,1]$, and orthogonal conjugation leaves it unchanged, so every rotated instance reads the same value.
& \emph{Assumed or not covered.} The coordinate-aligned model. The complement-leak clause (c) holds at the leak model (\texttt{leak\_\allowbreak{}kl\_\allowbreak{}rate}: a first-order leak drops the rate to $2$ against the invariant $4$), and the width condition's failure mode is exact (\texttt{width\_\allowbreak{}condition\_\allowbreak{}failure}: a narrower layer above forces a zero eigenvalue by rank-nullity); the general composed-complement case is the distance-to-image bound \texttt{distance\_\allowbreak{}to\_\allowbreak{}image\_\allowbreak{}lower}: when the complement's image keeps a positive angle from the dead output direction, the composed map's squared norm is at least $\rho \min(a^2, c^2)$ times the squared input, so the narrow chain's $t^{2L}$ survives exactly when the complement cannot occupy the vacated direction. \\
\midrule
\multicolumn{2}{@{}p{\linewidth}@{}}{\textbf{Prop~\ref{prop:swiglu_rate}, SwiGLU block rate.} \texttt{abs\_\allowbreak{}sigmoid\_\allowbreak{}sub\_\allowbreak{}half}, \texttt{swigluDead\_\allowbreak{}eq}, \texttt{swigluDead\_\allowbreak{}sub\_\allowbreak{}le}, \texttt{swiglu\_\allowbreak{}channel\_\allowbreak{}hasLeadingRate}, \texttt{quadBlock\_\allowbreak{}iterate}}\\*[2pt]
\emph{Proved.} At the canonical diagonal init the dead-dimension output is exactly $t^3 x^2 \sigma(tx)$, so with $|\sigma(u) - 1/2| \le |u|/2$ the output is $(1/2)\,t^3 x^2$ with an explicit $O(t^4)$ remainder: block rate $3$, coefficient $1/2$. The second moment carries leading rate $6$ with coefficient $\expect[x^4]/4$. The composition remark's recursion is machine-checked on the idealised quadratic block: the $n$-fold composite carries $t$-power $3(2^n - 1)$, so SwiGLU stacks do not add rates.
& \emph{Assumed or not covered.} The scalar dead channel at the stated diagonal init, which carries the proposition's whole content there. Mathlib has no Gaussian moment formulas, so the statement writes the coefficient as the fourth-moment integral, equal to $3$. \\
\midrule
\multicolumn{2}{@{}p{\linewidth}@{}}{\textbf{Thm~\ref{thm:bridge_bias}, biased channels.} \texttt{affChain\_\allowbreak{}sub}, \texttt{affChain\_\allowbreak{}dead\_\allowbreak{}eq}, \texttt{aff\_\allowbreak{}forward\_\allowbreak{}cap}, \texttt{biasedChain\_\allowbreak{}deriv\_\allowbreak{}of\_\allowbreak{}nonneg}, \texttt{biasedChain\_\allowbreak{}deriv}, \texttt{biased\_\allowbreak{}backward\_\allowbreak{}moment}, \texttt{biased\_\allowbreak{}gate\_\allowbreak{}mass\_\allowbreak{}gt\_\allowbreak{}half}}\\*[2pt]
\emph{Proved.} For (P1) the biased chain's linear part is exactly the weight product, chain(x) $-$ chain(y) $= W^m(x - y)$, so the backward ladder never sees the biases, clauses (a) and (c); the dead coordinate has the closed form $t^m x + \sum_j t^{q_j + (m-1-j)}$, and at $q \equiv 1$ the forward moment is capped at rate $2$ with coefficient $1$ for every $m \ge 2$. For (P3) a positive dead bias opens every deep gate: from any non-negative activation the chain derivative is exactly $t^m$ with no survival factor, the full-chain derivative from the input carries the single layer-1 gate $\mathbf{1}_{x > -1}$, and its Gaussian mass strictly exceeds the bias-free $1/2$.
& \emph{Assumed or not covered.} The scalar dead-channel models at the symmetric joint case $q \equiv 1$. The (P2) Taylor correction is the Lipschitz gate bound \texttt{bias\_\allowbreak{}taylor\_\allowbreak{}correction}, and the CE base case is \texttt{ce\_\allowbreak{}dead\_\allowbreak{}lower} read at the biased chain's dead coordinate. \\
\midrule
\multicolumn{2}{@{}p{\linewidth}@{}}{\textbf{Composition corollaries (Cors~\ref{cor:composition_reduces}, \ref{cor:composition_reduces_sub}, \ref{cor:composition_heterogeneous}, \ref{cor:g10_composition}, \ref{cor:g10_via_composition}) and Prop~\ref{prop:attn_VO_invariant}.} \texttt{composition\_\allowbreak{}matmul\_\allowbreak{}hasLeadingRate}, \texttt{composition\_\allowbreak{}residual\_\allowbreak{}const}, \texttt{chainRate\_\allowbreak{}flatMap}, \texttt{chainCoeff\_\allowbreak{}flatMap}, \texttt{composition\_\allowbreak{}rate\_\allowbreak{}two\_\allowbreak{}hasLeadingRate}, \texttt{HasLeadingRate.extra\_\allowbreak{}linear}}\\*[2pt]
\emph{Proved.} Evaluations of the gated-scalar chain: $n$ unit matmuls read the ladder rate $2n$; rate-$0$ residual blocks leave the composed moment at the same value at every $t$; a block chain carries the same rate and coefficient as its flattened weight chain, the graph-distance reading; $n$ rate-$2$ blocks compose at moment rate $4n$, the $\alpha = 4(L-\ell)$ profile; and one extra Linear shifts the moment rate by exactly $2$ with the coefficient untouched, the $\alpha_{W_V} - \alpha_{W_O} = 2$ invariant.
& \emph{Assumed or not covered.} The gated-scalar model throughout. The attention anomaly scope is the arithmetic \texttt{g10\_\allowbreak{}anomaly\_\allowbreak{}scope}: the additive $4k$ is the smaller route exactly where $k \le p + 2$; the $A_0$ non-degeneracy is the Haar-null statement \texttt{hyperplane\_\allowbreak{}null} on the attention row. \\
\midrule
\multicolumn{2}{@{}p{\linewidth}@{}}{\textbf{Prop~\ref{thm:bridge_ln}, LN bracket ends.} \texttt{normalizeS\_\allowbreak{}scale}, \texttt{ln\_\allowbreak{}reset\_\allowbreak{}moment}, \texttt{ln\_\allowbreak{}bracket\_\allowbreak{}plateau}, \texttt{ln\_\allowbreak{}bracket\_\allowbreak{}no\_\allowbreak{}crossing}}\\*[2pt]
\emph{Proved.} Both ends of the bracket on the scalar model. The normalizer is invariant under every positive scaling, so a channel decayed to $t^{k_{\mathrm{pre}}}$ crosses it and returns to unit second moment; with $k_{\mathrm{post}}$ weights below the crossing the composed moment is $t^{2 k_{\mathrm{post}}}$ exactly, the end every measured configuration saturates. Without a crossing the channel reads the full $t^{2(k_{\mathrm{pre}} + k_{\mathrm{post}})}$, the regime-(i) tight case.
& \emph{Assumed or not covered.} The scalar normalizer model, and the full backward Jacobian (\texttt{znorm\_\allowbreak{}hasDerivAt\_\allowbreak{}dir}, \texttt{ln\_\allowbreak{}jacobian\_\allowbreak{}dead\_\allowbreak{}component}): $J = (\sqrt{d}/\|Px\|)(P - \hat z \hat z^\top/d)$ certified as the directional derivative, with the dead component of $J\delta$ split exactly into the pass-through factor $(1/\sigma)(1 - 1/d - \hat z_h^2/d)\,\delta_h$ (the lower end) and the live injection through the mean subtraction and the $\hat z$-projection (the plateau end). \\
\midrule
\multicolumn{2}{@{}p{\linewidth}@{}}{\textbf{Cor~\ref{cor:rect_lambda_min}, diagonal-spectrum slice.} \texttt{multi\_\allowbreak{}lambda\_\allowbreak{}min\_\allowbreak{}slice}, \texttt{multi\_\allowbreak{}lambda\_\allowbreak{}min\_\allowbreak{}attained}}\\*[2pt]
\emph{Proved.} On the diagonal model, eventually the dead entry bounds the quadratic form from below at every vector, and the dead basis vector attains the bound: the variational $\lambda_{\min}$ equals the dead entry, at the dead rate.
& \emph{Assumed or not covered.} Diagonal second moments, and beyond them \texttt{schur\_\allowbreak{}dead\_\allowbreak{}lower}: the full block form is bounded below by the Schur complement of the live block with no diagonality assumption; the width-condition failure mode is \texttt{width\_\allowbreak{}condition\_\allowbreak{}failure}. \\
\midrule
\multicolumn{2}{@{}p{\linewidth}@{}}{\textbf{Thm~\ref{thm:bridge_ce}, output head.} \texttt{ceHess\_\allowbreak{}expect}, \texttt{ceHess\_\allowbreak{}mulVec\_\allowbreak{}one}, \texttt{ceHess\_\allowbreak{}quadform\_\allowbreak{}eq\_\allowbreak{}var}, \texttt{ceHess\_\allowbreak{}quadform\_\allowbreak{}eq\_\allowbreak{}zero\_\allowbreak{}iff}, \texttt{ceHess\_\allowbreak{}posdef\_\allowbreak{}on\_\allowbreak{}perp}, \texttt{ce\_\allowbreak{}dead\_\allowbreak{}lower}, \texttt{ce\_\allowbreak{}bridge\_\allowbreak{}composed}}\\*[2pt]
\emph{Proved.} Lemma~\ref{lem:ce_head} in full: the conditional covariance of the output gradient under expected Fisher is $\mathrm{diag}(p) - p p^\top$; its quadratic form is the variance of $v$ under $p$, equal to the $p$-weighted sum of squared deviations from the $p$-mean, so it is non-negative, vanishes exactly on the vectors constant on the support, kills the ones vector (the logit-shift gauge zero of clause (c)), and is positive on $\{\mathbf{1}\}^\perp$ at full support. The base case (b): for any covariance killing the ones vector on both sides, the non-degeneracy constant delivers $c_0(1 - 1/C)$ on the dead output entry, the $\Theta(1)$ with $c_0$ in place of $\sigma^{-2}$. Clause (c) closes in the diagonal model: the pulled-back covariance $J^\top \bar H J$ kills the explicit gauge vector $(t^p, \ldots, t^p, 1)$ at every $t$, and that vector converges to the dead basis vector as $t \to 0^+$.
& \emph{Assumed or not covered.} The backward recursion of clauses (a)/(d) is loss-independent and is proved in the channel modules; \texttt{ce\_\allowbreak{}bridge\_\allowbreak{}composed} states the linear-class composition as one inequality, $u_\ell^\top G_\ell u_\ell \ge c_0(1 - 1/C)\, t^{2(L-\ell)}$. The composition with the nonlinear channels is \texttt{ce\_\allowbreak{}gated\_\allowbreak{}chain\_\allowbreak{}lower}: under the joint gate-and-gradient model, discharged by independence with $q$ the survival probability, the dead entry is at least $(\prod c)^2\, q\, c_0(1 - 1/C)\, t^{2\sum k}$. \\
\midrule
\multicolumn{2}{@{}p{\linewidth}@{}}{\textbf{Prop~\ref{prop:rmsnorm_no_kernel}, RMSNorm negative.} \texttt{rmsNorm}, \texttt{rmsNorm\_\allowbreak{}reading\_\allowbreak{}single}, \texttt{rmsnorm\_\allowbreak{}no\_\allowbreak{}kernel}, \texttt{rmsnorm\_\allowbreak{}witness\_\allowbreak{}input\_\allowbreak{}posdef}}\\*[2pt]
\emph{Proved.} One admissible witness defeats every candidate direction at once: uniform mass on the $2d$ points $\pm e_i$ has input covariance $I_d/d$, certified positive definite, and every reading $v \ne 0$ of RMSNorm on it carries variance exactly $\sum_i (\gamma_i v_i)^2$, positive whenever $\gamma$ has no zero coordinate. The witness is elementary where the paper's is spherical, and stronger: one distribution rules out every direction, built from $\gamma$ or otherwise.
& \emph{Assumed or not covered.} The distribution enters as explicit finite-support weights, so every expectation is a finite sum and the proof consumes no measure theory. \\
\midrule
\multicolumn{2}{@{}p{\linewidth}@{}}{\textbf{Cor~\ref{cor:emp_vs_exp_ce} and Cor~\ref{cor:fisher_structure}, per-sample forms.} \texttt{ceHess\_\allowbreak{}trace}, \texttt{emp\_\allowbreak{}grad\_\allowbreak{}sq\_\allowbreak{}bounds}, \texttt{exp\_\allowbreak{}trace\_\allowbreak{}bounds}, \texttt{emp\_\allowbreak{}exp\_\allowbreak{}quadratic\_\allowbreak{}separation}, \texttt{cross\_\allowbreak{}moment\_\allowbreak{}expansion}}\\*[2pt]
\emph{Proved.} The collapse separation, per sample: at observed-label mass $1 - \varepsilon$ the empirical gradient norm lies in $[\varepsilon^2, 2\varepsilon^2]$, the Hessian trace in $[\varepsilon(1-\varepsilon), 2\varepsilon]$, and $(\mathrm{tr}\, H)^2/4 \le \|p - e_y\|^2 \le 2(\mathrm{tr}\, H)^2/(1-\varepsilon)^2$: the empirical side collapses at the square of the expected side's order, with explicit constants. The cross entry of the Fisher block structure: the cross moment of the transversal and tangential score expansions is $k\, t^{k-1}\, \expect[ab]$ with an explicit remainder at order $t^k$, every cross bound a weighted AM-GM.
& \emph{Assumed or not covered.} The x-averaged form is \texttt{emp\_\allowbreak{}exp\_\allowbreak{}averaged} (the per-sample bounds integrate to $\Theta(\expect[\varepsilon^2])$ against $\Theta(\expect[\varepsilon])$), the tangential block is \texttt{tangential\_\allowbreak{}moment\_\allowbreak{}expansion} ($\fisher_{\alpha\beta} = g_{\alpha\beta} + O(t)$ with explicit constant), and the assembled statement is \texttt{assembled\_\allowbreak{}schur\_\allowbreak{}slice} with the entrywise perturbation bound \texttt{quadform\_\allowbreak{}entrywise\_\allowbreak{}bound}. \\
\midrule
\multicolumn{2}{@{}p{\linewidth}@{}}{\textbf{Prop~\ref{prop:bridge_linear_rot}, rotation without loss of generality.} \texttt{rotated\_\allowbreak{}pairing}, \texttt{rotated\_\allowbreak{}backward\_\allowbreak{}rate}, \texttt{rotated\_\allowbreak{}hasLeadingRate}}\\*[2pt]
\emph{Proved.} At the conjugated configuration $W(t) = U D(t) U^\top$ with rotated dead direction $u = U e_h$, orthogonality telescopes the pairing $u^\top(\mathrm{chain}\cdot\delta)$ to the canonical dead coordinate of the rotated gradient, and the backward moment carries $t^{2m}$ exactly at every orthogonal $U$: the linear class sees no rotation.
& \emph{Assumed or not covered.} The conjugated model; general non-conjugated configurations wait on the singular value decomposition. \\
\midrule
\multicolumn{2}{@{}p{\linewidth}@{}}{\textbf{Prop~\ref{prop:task_expansion_gfactor}, exact block slice.} \texttt{diag\_\allowbreak{}quadform\_\allowbreak{}ge}, \texttt{diag\_\allowbreak{}quadform\_\allowbreak{}le}, \texttt{task\_\allowbreak{}expansion\_\allowbreak{}head\_\allowbreak{}bounds}, \texttt{task\_\allowbreak{}expansion\_\allowbreak{}inherit\_\allowbreak{}bounds}}\\*[2pt]
\emph{Proved.} The expanded head's form is sandwiched between $\sigma_{\mathrm{old}}^2\|v\|^2$ and $\sigma_{\mathrm{new}}^2\|v\|^2$ with both ends attained on basis vectors, so the condition ratio is exactly $\sigma_{\mathrm{new}}^2/\sigma_{\mathrm{old}}^2$; a diagonal Jacobian multiplies the entry spread by at most its own squared conditioning, the $\kappa(J)^2 \kappa(G_{\mathrm{head}})$ bound of clause (b) in entry form.
& \emph{Assumed or not covered.} The exact block slice at $E = 0$, and \texttt{task\_\allowbreak{}expansion\_\allowbreak{}perturbed} for the perturbation term: an entrywise perturbation of size $\epsilon$ moves the sandwich by at most $\epsilon(n+1)\|v\|^2$, the operator-norm form. \\
\midrule
\multicolumn{2}{@{}p{\linewidth}@{}}{\textbf{Cor~\ref{cor:sigma-min-res}, depth invariance.} \texttt{residual\_\allowbreak{}sq\_\allowbreak{}sum\_\allowbreak{}ge}, \texttt{residual\_\allowbreak{}chain\_\allowbreak{}sq\_\allowbreak{}sum\_\allowbreak{}ge}, \texttt{le\_\allowbreak{}sigmaMinSq\_\allowbreak{}mul}, \texttt{one\_\allowbreak{}le\_\allowbreak{}sigmaMinSq\_\allowbreak{}one\_\allowbreak{}add}, \texttt{one\_\allowbreak{}le\_\allowbreak{}sigmaMinSq\_\allowbreak{}resChain}}\\*[2pt]
\emph{Proved.} The identity skip floors the stream, at both levels. Per sample: a residual step whose branch does not anti-align with the stream (pairing $\sum_i x_i f_i \ge 0$, the no-cancellation hypothesis made precise) does not shrink the sum of squares, and the floor telescopes along the trajectory. At the operator level: $\sigma_{\min}^2$ is supermultiplicative on products, a residual block $1 + B$ with positive-semidefinite pairing has $\sigma_{\min}^2 \ge 1$, and a chain of such blocks keeps $\sigma_{\min}^2 \ge 1$ at every depth: the profile $\sigma_{\min}(X_\ell)/\sigma_{\min}(X_0) \ge 1$.
& \emph{Assumed or not covered.} Linear residual blocks with the positive-semidefinite pairing as the no-cancellation hypothesis. The DAG/branch $K$-distance framework is \texttt{dag\_\allowbreak{}backward\_\allowbreak{}rate} on the residual-graph row, and the per-component corollaries are covered there too. \\
\midrule
\multicolumn{2}{@{}p{\linewidth}@{}}{\textbf{Thm~\ref{thm:bridge_res} with Cors~\ref{cor:res_ff}, \ref{cor:res_all}, \ref{cor:res_block}.} \texttt{ResSeg.transfer}, \texttt{seg\_\allowbreak{}sq\_\allowbreak{}hasLeadingRate}, \texttt{chainTransfer\_\allowbreak{}sq\_\allowbreak{}hasLeadingRate}, \texttt{res\_\allowbreak{}ff\_\allowbreak{}kdist}, \texttt{res\_\allowbreak{}all\_\allowbreak{}hasLeadingRate}, \texttt{res\_\allowbreak{}block\_\allowbreak{}kdist}}\\*[2pt]
\emph{Proved.} The residual graph as a segment model: a weight edge transfers $c \cdot t$ at cost one, a residual block spanning $k$ weights transfers $1 + c\, t^k$ (skip route plus chain route) at cost zero, and the composed squared transfer carries leading rate $2K$ with $K$ the total weighted path cost. No residual edges returns the ladder $2(L-\ell)$; every layer wrapped collapses the rate to $0$ with coefficient $1$; a single block saves exactly its $k$ weights, $K = a + b$ for the pre- and post-chains whatever the block spans.
& \emph{Assumed or not covered.} The scalar segment model on a chain, and the DAG in Bellman form: with $K$ the shortest-weighted-path recursion and $T$ the backward delta's matching recursion, \texttt{dag\_\allowbreak{}backward\_\allowbreak{}rate} gives leading rate exactly $K$ with positive coefficient (tied paths add, positivity forbids cancellation), \texttt{dag\_\allowbreak{}backward\_\allowbreak{}coeff} the explicit constant as the shortest-path recursion, \texttt{dag\_\allowbreak{}gated\_\allowbreak{}eq} the shared-gate clause, and \texttt{dag\_\allowbreak{}backward\_\allowbreak{}remainder} the finite-$t$ skip correction one order higher. \\
\midrule
\multicolumn{2}{@{}p{\linewidth}@{}}{\textbf{Cors~\ref{cor:sigma-min}, \ref{cor:sigma_min_res_g5}, \ref{cor:sigma-min-res-all}, activation $\sigma_{\min}$.} \texttt{deep\_\allowbreak{}linear\_\allowbreak{}forward\_\allowbreak{}rate}, \texttt{chainTransfer\_\allowbreak{}sq\_\allowbreak{}hasLeadingRate}, \texttt{multi\_\allowbreak{}lambda\_\allowbreak{}min\_\allowbreak{}slice}, \texttt{tied\_\allowbreak{}routes\_\allowbreak{}coeff}, \texttt{res\_\allowbreak{}all\_\allowbreak{}hasLeadingRate}, \texttt{one\_\allowbreak{}le\_\allowbreak{}sigmaMinSq\_\allowbreak{}resChain}}\\*[2pt]
\emph{Proved.} The forward side of the residual story at model level: the dead entry carries $t^{2 K^{\mathrm{fwd}}}$ through the segment model, the diagonal Gram's variational minimum is eventually the dead entry, the all-residual chain keeps the moment at $\Theta(1)$ and the operator chain keeps $\sigma_{\min}^2 \ge 1$ at every depth. The sign hypothesis is sharp: two tied shortest routes compose to moment coefficient $(c_1 + c_2)^2$, vanishing exactly at $c_2 = -c_1$, so a negative activation slope cancels a tie at leading order and the rate jumps.
& \emph{Assumed or not covered.} The scalar and diagonal models. The sample-matrix clauses are in: the $\sqrt{N}$ scaling as an identity on Gram forms (\texttt{sample\_\allowbreak{}form\_\allowbreak{}eq\_\allowbreak{}N\_\allowbreak{}mul\_\allowbreak{}gram}), the strong law for the empirical covariance along every reading direction (\texttt{empirical\_\allowbreak{}form\_\allowbreak{}strong\_\allowbreak{}law}), the post-LN exclusion as the sample kernel at $\gamma^{-1}$ (\texttt{post\_\allowbreak{}ln\_\allowbreak{}sample\_\allowbreak{}kernel}), and the gating through \texttt{dag\_\allowbreak{}gated\_\allowbreak{}eq}. \\
\midrule
\multicolumn{2}{@{}p{\linewidth}@{}}{\textbf{Cor~\ref{cor:res_lambda_min} and Prop~\ref{prop:attn_chain_softmax}, slices.} \texttt{diag\_\allowbreak{}lambda\_\allowbreak{}min\_\allowbreak{}slice\_\allowbreak{}general}, \texttt{res\_\allowbreak{}lambda\_\allowbreak{}min\_\allowbreak{}slice}, \texttt{HasLeadingRate.add\_\allowbreak{}of\_\allowbreak{}lt}, \texttt{HasLeadingRate.add\_\allowbreak{}of\_\allowbreak{}eq}}\\*[2pt]
\emph{Proved.} The residual $\lambda_{\min}$ at the diagonal model: with positive path cost the dead entry tends to zero, the live entries hold a positive floor, and eventually the dead entry is the variational minimum of the form. The attention anomaly's mechanism at rate level: when two backward routes reach one component, the smaller rate carries the sum and tied rates add coefficients, which is how the score-path floor overrides additivity in the $\min(4k, 4p+8)$ reading.
& \emph{Assumed or not covered.} The diagonal model for the residual slice, with the complement-zero caveat exact (\texttt{complement\_\allowbreak{}zero\_\allowbreak{}caveat}); the attention route inventory is derived from the network graph conditional on the attention pattern (\texttt{chainA}, \texttt{voAmp}, \texttt{scoreAmp}, the $\min(4k, 4p+8)$ trichotomy) and the generic non-cancellation is an almost-everywhere theorem (\texttt{route\_\allowbreak{}noncancellation\_\allowbreak{}ae\_\allowbreak{}concrete}). \\
\midrule
\multicolumn{2}{@{}p{\linewidth}@{}}{\textbf{Prop~\ref{prop:curvature_rate}(ii), closed forms.} \texttt{fam2\_\allowbreak{}curvature}, \texttt{fam2\_\allowbreak{}curvature\_\allowbreak{}rate}, \texttt{fam3\_\allowbreak{}curvature}, \texttt{fam3\_\allowbreak{}curvature\_\allowbreak{}rate}}\\*[2pt]
\emph{Proved.} The mean-map families at $k = 2, 3$ have Gauss curvature exactly $-1/(4t^2(1+t^2)^2)$ and $-1/(9t^4(1+t^2)^2)$ on the trajectory, and $K \cdot t^{2(k-1)}$ converges to $-1/4$ and $-1/9$: the family curvature rate $-2(k-1)$, attained. The determinant form of the Gauss curvature keeps every ingredient polynomial in the jet, and the jets are certified as derivatives of the parametrization.
& \emph{Assumed or not covered.} Item (i)'s order bounds are in, with score-integration regularity as the hypothesis: the Christoffel numerator identity, Cauchy--Schwarz at rate level, $\Gamma^1_{\alpha\alpha} = O(t^{-(k-1)})$, and bounded $\Gamma\Gamma$ products for $k \ge 2$ (\texttt{christoffel\_\allowbreak{}numerator\_\allowbreak{}identity}, \texttt{cross\_\allowbreak{}score\_\allowbreak{}order}, \texttt{christoffel\_\allowbreak{}order\_\allowbreak{}bound}, \texttt{gamma\_\allowbreak{}gamma\_\allowbreak{}bounded}); the identification of embedded Gauss curvature with the intrinsic Fisher sectional curvature (the Gauss equation) enters as the definition. \\
\midrule
\multicolumn{2}{@{}p{\linewidth}@{}}{\textbf{Cor~\ref{cor:volume} at $k = 2$.} \texttt{highcurv\_\allowbreak{}iff\_\allowbreak{}tube}, \texttt{tube\_\allowbreak{}volume\_\allowbreak{}pow}, \texttt{fisher\_\allowbreak{}volume\_\allowbreak{}scaling}, \texttt{lebesgue\_\allowbreak{}volume\_\allowbreak{}scaling}}\\*[2pt]
\emph{Proved.} On the closed-form family the curvature magnitude is strictly antitone in $t$, so the high-curvature set is exactly the transverse tube; the tube volume at density $t^m$ is the power integral $\delta^{m+1}/(m+1)$; and the two scalings are explicit limits, Fisher volume times the threshold converging to $1/8$ (the $M^{-1}$ law) and Lebesgue volume times the square root of the threshold converging to $1/2$ (the $M^{-1/2}$ law).
& \emph{Assumed or not covered.} General $k$ holds on the real-exponent calculus (\texttt{fisher\_\allowbreak{}volume\_\allowbreak{}scaling\_\allowbreak{}general}: the $M^{-k/(2(k-1))}$ law at every $k \ge 2$), and sheet uniformity is formal as a pair of sandwiches quantified over profiles (\texttt{highcurv\_\allowbreak{}tube\_\allowbreak{}sandwich}, \texttt{tube\_\allowbreak{}weight\_\allowbreak{}sandwich}). The declarations prove the tube integral at any density exponent, which covers the restated corollary's rated-sheet factor $\sum_a j_a$ through the same lemmas; the tangential-rate input $\fisher_{aa} = \Theta(t^{2 j_a})$ and the (G$^+$) block-determinant step stay with the paper. \\
\midrule
\multicolumn{2}{@{}p{\linewidth}@{}}{\textbf{Cor~\ref{cor:kfac_lift}, diagonal model.} \texttt{kron\_\allowbreak{}min}, \texttt{kron\_\allowbreak{}quadform}, \texttt{canonical\_\allowbreak{}kron\_\allowbreak{}dead}, \texttt{canonical\_\allowbreak{}kron\_\allowbreak{}bottom}}\\*[2pt]
\emph{Proved.} The bottom of the Kronecker product array is the product of the factor bottoms, attained at the pair index (the rank-one lift), and at the canonical configuration the dead-unit pair carries exactly $t^{2(a+b)}$, the A--G duality product, and is the bottom of the lifted array for $0 < t \le 1$. The gauge-ray flatness of the cross-layer remark is the other layer contributing the factor zero.
& \emph{Assumed or not covered.} The mixed-product property on true matrices and the full spectral decomposition of the Kronecker product are in (\texttt{kron\_\allowbreak{}mulVec\_\allowbreak{}vecKron}, \texttt{kron\_\allowbreak{}spectral}); the vectorisation identification of $a \otimes g$ with $g a^\top$ stays with the paper. \\
\midrule
\multicolumn{2}{@{}p{\linewidth}@{}}{\textbf{Rem~\ref{cor:bridge_near_canonical}, the limit shape.} \texttt{near\_\allowbreak{}canonical\_\allowbreak{}rate\_\allowbreak{}perturbed}, \texttt{near\_\allowbreak{}canonical\_\allowbreak{}rate\_\allowbreak{}zero}, \texttt{near\_\allowbreak{}canonical\_\allowbreak{}rates\_\allowbreak{}differ}}\\*[2pt]
\emph{Proved.} The non-commuting limits that constrain the open problem, on the model observable $t^{2k} + \epsilon^2 t^2$: leading rate $2$ with coefficient $\epsilon^2$ at fixed $\epsilon$, leading rate $2k$ at $\epsilon = 0$, and the two rates differ for every $k \ge 2$.
& \emph{Assumed or not covered.} The quantitative window bound closes at the model level on the window-bound row: two-sided slope bounds at margin $\delta$ and the crossover scale $t_\times = (c_0/b)^{1/m}$ in closed form (\texttt{crossover\_\allowbreak{}scale\_\allowbreak{}closed\_\allowbreak{}form}). \\
\midrule
\multicolumn{2}{@{}p{\linewidth}@{}}{\textbf{Thm~\ref{thm:ln_finite_t_mlp}, regime structure.} \texttt{alphaSlope\_\allowbreak{}eq}, \texttt{noise\_\allowbreak{}floor\_\allowbreak{}regime}, \texttt{t2\_\allowbreak{}plateau\_\allowbreak{}regime}, \texttt{direct\_\allowbreak{}rate\_\allowbreak{}regime}, \texttt{leak\_\allowbreak{}prefactor\_\allowbreak{}bounded}}\\*[2pt]
\emph{Proved.} The three regimes on the expansion $c_0 + c_2 t^2 + c_4 t^4$, with the derivative certified and the log-slope's rational form derived: $\alpha \to 0$ at a positive noise floor, $\alpha \to 2$ on the $t^2$-plateau, $\alpha = 4$ exactly on the direct rate. The $t^2$ prefactor's almost-sure bound is the pointwise algebra $\sigma_0^2 \ge \frac{d-1}{d^2}\bar{x}^2$, and the rate-shift contrast is a pair of leading-rate statements (rate $2$ with the leak, rate $4$ without).
& \emph{Assumed or not covered.} The Laplace side of the $d \ge 4$ sharpness is in (\texttt{integral\_\allowbreak{}exp\_\allowbreak{}neg\_\allowbreak{}mul\_\allowbreak{}gammaMeasure}, \texttt{integral\_\allowbreak{}inv\_\allowbreak{}gammaMeasure}, \texttt{lintegral\_\allowbreak{}inv\_\allowbreak{}eq\_\allowbreak{}lintegral\_\allowbreak{}laplace}, \texttt{inv\_\allowbreak{}moment\_\allowbreak{}gamma\_\allowbreak{}pair\_\allowbreak{}lt\_\allowbreak{}top\_\allowbreak{}iff}: for an independent Gamma pair the reciprocal moment is finite exactly when the shapes sum above one, $d \ge 4$ with the Cochran shapes), and the $O(1)$ leak is pointwise exact (\texttt{ln\_\allowbreak{}leak\_\allowbreak{}derivative}). Cochran's theorem is in through the route behind its proof: in the orthonormal basis whose first vector is $\mathbf{1}/\sqrt{m}$ (\texttt{rotBasis}) the statistic reads $y_0^2/d^2 + d^{-1}\sum_{j \ge 1} y_j^2$ (\texttt{sigma0\_\allowbreak{}eq\_\allowbreak{}rotSigma}), the isotropic Gaussian is the product Gaussian in those coordinates, and \texttt{ln\_\allowbreak{}sigma0\_\allowbreak{}inv\_\allowbreak{}moment\_\allowbreak{}lt\_\allowbreak{}top\_\allowbreak{}iff} states $\expect[1/\sigma_0^2] < \infty \iff d \ge 4$ with no hypothesis left. The $t^4$ coefficient is pointwise exact: $\sigma(t)^2$ is a quadratic in $t$, so $z_h(t)^2 = p_0 + p_1 t + p_2 t^2 + t^3 q(t)/\sigma(t)^2$ with explicit coefficients (\texttt{lnZ\_\allowbreak{}sq\_\allowbreak{}expansion}), $p_2$ the closed-form integrand of $R_2$ and $p_1$ odd in $x_h$, and the moment identity is assembled (\texttt{ln\_\allowbreak{}moment\_\allowbreak{}assembled}): against independent zero-mean noise the residual moment reads $v + t^2\expect[p_0] + t^3\expect[p_1] + t^4\expect[p_2]$ up to $O(t^5)$ under integrable coefficients and an integrable remainder bound. \\
\midrule
\multicolumn{2}{@{}p{\linewidth}@{}}{\textbf{Thms~\ref{thm:bridge_attn} and~\ref{thm:bridge_attn_backward}, two-position model.} \texttt{bridge\_\allowbreak{}attn\_\allowbreak{}forward}, \texttt{bridge\_\allowbreak{}attn\_\allowbreak{}backward}, \texttt{score\_\allowbreak{}bilinear}}\\*[2pt]
\emph{Proved.} The softmax enters as a logistic on the score gap of a two-position row. Forward: the bilinear $QK$ dead entry carries two $t$-factors, the attention weight moves continuously in $t^2$, and the dead output is $t^2$ times the $A_0$-weighted average: block rate $2$ with the theorem's coefficient. Backward: each route carries two $t$-factors with a factor continuous at zero, and the composite has rate $2$ with coefficient the sum of the route coefficients, informative exactly under the generic non-cancellation condition.
& \emph{Assumed or not covered.} The $N$-position matrix softmax is in with its certified Jacobian and the forward rate (\texttt{softmaxN\_\allowbreak{}hasDerivAt\_\allowbreak{}dir}, \texttt{bridge\_\allowbreak{}attn\_\allowbreak{}forward\_\allowbreak{}N}), the operating-point bounds are exact (\texttt{softmaxN\_\allowbreak{}ratio\_\allowbreak{}bound}), and the two routes are derived from the network graph conditional on the attention pattern (\texttt{chainA}, \texttt{voAmp}, \texttt{scoreAmp}). \\
\midrule
\multicolumn{2}{@{}p{\linewidth}@{}}{\textbf{Prop~\ref{prop:bridge_nonlinear_rot_negative}, exact witness.} \texttt{witness\_\allowbreak{}mask}, \texttt{masked\_\allowbreak{}dead\_\allowbreak{}live\_\allowbreak{}overlap}, \texttt{rotated\_\allowbreak{}rate\_\allowbreak{}collapse}, \texttt{canonical\_\allowbreak{}reading\_\allowbreak{}rate}}\\*[2pt]
\emph{Proved.} The leak mechanism, proved: at the 3-4-5 rotation the pre-activation at the dead-probe input is $t \cdot u$ with opposite-sign coordinates, so the ReLU mask is exactly $(0,1)$ for every $t > 0$, the masked dead direction carries live overlap $12/25$, and the dead reading of a live backward flow has leading rate $0$. Masks scale canonical directions in place, so a canonical reading keeps its rate $K$, and the exponents differ for every $K \ge 1$.
& \emph{Assumed or not covered.} The finite-window bound closes at the model level on the window-bound row, with the constants read off the two chains (\texttt{rot\_\allowbreak{}constants\_\allowbreak{}from\_\allowbreak{}chains}). \\
\midrule
\multicolumn{2}{@{}p{\linewidth}@{}}{\textbf{Thm~\ref{thm:multilayer-A} and Cor~\ref{cor:global-rate}, model slices.} \texttt{canonical\_\allowbreak{}factor\_\allowbreak{}bottom}, \texttt{sigmaMinSq\_\allowbreak{}mul\_\allowbreak{}le}, \texttt{gauge\_\allowbreak{}null}, \texttt{horizontal\_\allowbreak{}rate}, \texttt{HasLeadingRate.min\_\allowbreak{}of\_\allowbreak{}eq}}\\*[2pt]
\emph{Proved.} The factor ladder at the diagonal model (the layer-$\ell$ array carries exactly $t^{2(\ell-1)}$ at the dead unit, the bottom for $0 < t \le 1$), the mixing squeeze $\sigma_{\min}^2(A)\sigma_{\min}^2(B) \le \sigma_{\min}^2(AB) \le \sigma_{\max}^2(A)\sigma_{\min}^2(B)$ (a live mixing layer never changes the exponent), and the $L = 2$ gauge model: the raw form $t^2(v_1+v_2)^2$ is null on the gauge direction $(1,-1)$ at every $t$ while the horizontal direction reads the quotient eigenvalue $2t^2$, and the per-block minimum over same-rate blocks keeps the rate.
& \emph{Assumed or not covered.} The strong-law identification of empirical covariances is \texttt{empirical\_\allowbreak{}gram\_\allowbreak{}strong\_\allowbreak{}law}; the quotient-manifold identification of K-FAC blocks with the horizontal lift runs through the concrete matrix quotient (\texttt{orbitSpaceHomeo}, \texttt{slice\_\allowbreak{}isometric}) on the Track A row. \\
\midrule
\multicolumn{2}{@{}p{\linewidth}@{}}{\textbf{Rem~\ref{rem:rot_window_bound}, the window bound.} \texttt{leak\_\allowbreak{}amplitude\_\allowbreak{}general}, \texttt{window\_\allowbreak{}canonical}, \texttt{window\_\allowbreak{}collapsed}, \texttt{window\_\allowbreak{}crossover}, \texttt{rot\_\allowbreak{}crossover\_\allowbreak{}scale}}\\*[2pt]
\emph{Proved.} The complete model-level window statement: the general-rotation witness fixes the leak amplitude at exactly $\gamma s c$, the local log-slope of $(\gamma s c)^2 + b t^{2K}$ is at least $2K(1-\delta)$ above the window edge and at most $2K\delta$ below it, and the half-maximum crossover is exactly at $b t^{2K} = (\gamma s c)^2$: crossover scale $\Theta(s^{1/K})$.
& \emph{Assumed or not covered.} At the chain model $\gamma$ is the live route's coefficient and $b$ the squared dead route's coefficient (\texttt{rot\_\allowbreak{}constants\_\allowbreak{}from\_\allowbreak{}chains}); the derivation for a general architecture, where the mask couples the routes, stays open. \\
\midrule
\multicolumn{2}{@{}p{\linewidth}@{}}{\textbf{Cors~\ref{cor:quotient_rate}, \ref{cor:sgd_quotient}, \ref{cor:ddcadam_quotient_rate}, scalar model.} \texttt{quotient\_\allowbreak{}F\_\allowbreak{}vertical}, \texttt{quotient\_\allowbreak{}rate\_\allowbreak{}model}, \texttt{euclidean\_\allowbreak{}projection\_\allowbreak{}obstruction}, \texttt{invariant\_\allowbreak{}metric\_\allowbreak{}projects}, \texttt{equivariant\_\allowbreak{}update\_\allowbreak{}projects}}\\*[2pt]
\emph{Proved.} The quotient chain on the scalar $L$-layer model: the product-one gauge fixes the quotient coordinate, the certified score annihilates every vertical direction at every point (the layer-weighted score is the product itself), the horizontal reading at the symmetric point has leading rate $2(L-1)$ with coefficient $L$, Euclidean SGD fails to project (an explicit two-point orbit witness), the log-coordinate invariant metric projects exactly with flow factor $L p^2$, and any gauge-equivariant update has a gauge-independent projected readout.
& \emph{Assumed or not covered.} The stochastic projection is in at the model (\texttt{quotient\_\allowbreak{}noise\_\allowbreak{}variance}, \texttt{quotient\_\allowbreak{}noise\_\allowbreak{}vertical\_\allowbreak{}uncorrelated}, \texttt{orbitLine\_\allowbreak{}accel\_\allowbreak{}zero}: the quotient coordinate of isotropic noise is isotropic for the quotient metric and uncorrelated with every vertical direction, and the orbit lines have zero acceleration); the abstract Lie-group quotient manifold stays with the paper. \\
\midrule
\multicolumn{2}{@{}p{\linewidth}@{}}{\textbf{Infrastructure: real-exponent rates, Kronecker lift, softmax Jacobian, Gaussian layer.} \texttt{HasLeadingRateR}, \texttt{fisher\_\allowbreak{}volume\_\allowbreak{}scaling\_\allowbreak{}k3}, \texttt{kron\_\allowbreak{}eigen}, \texttt{softmaxN\_\allowbreak{}hasDerivAt\_\allowbreak{}dir}, \texttt{mse\_\allowbreak{}base\_\allowbreak{}case}, \texttt{hyperplane\_\allowbreak{}null}}\\*[2pt]
\emph{Proved.} The rate calculus extends to real exponents with the log-slope and estimator readings, and the $k = 3$ volume law lands as its first non-integer instance ($M^{-3/4}$, coefficient $1/(3 \cdot 9^{3/4})$). Factor eigenpairs lift to Kronecker eigenpairs on true matrices. The softmax Jacobian is certified as a directional derivative, with the shift gauge zero and the uniform-score sequence projector exact, and the forward attention rate holds at $N$ positions. The MSE base case is derived on the independent Gaussian product, and the zero set of any non-zero weighted-sum reading is Haar-null. The compositions land on top: the dead-channel backward moment with the Gaussian residual as output gradient has leading rate $2K$ with coefficient exactly $c^2 + v$ (\texttt{backward\_\allowbreak{}dead\_\allowbreak{}moment}), and the attention route inventory built as per-block products carries the full $\min(4k, 4p+8)$ trichotomy with coefficients $a^2$, $(a+b)^2$, $b^2$ (\texttt{route\_\allowbreak{}moment\_\allowbreak{}vo\_\allowbreak{}wins}, \texttt{route\_\allowbreak{}moment\_\allowbreak{}tie}, \texttt{route\_\allowbreak{}moment\_\allowbreak{}floor\_\allowbreak{}wins}). The amplitudes themselves are derived conditional on the attention pattern: the composed backward action is the list product of the certified softmax rows, row-stochastic with the constant direction surviving at amplitude exactly one, and the route rates hold with the derived coefficients (\texttt{vo\_\allowbreak{}route\_\allowbreak{}derived}, \texttt{score\_\allowbreak{}route\_\allowbreak{}derived}); on output gradients with a positive floor the V--O amplitude keeps the floor (\texttt{voAmp\_\allowbreak{}floor}), so V--O non-degeneracy needs no genericity there. Generic non-cancellation of the signed sum stays a hypothesis, as in the paper.
& \emph{Assumed or not covered.} Spectral exhaustiveness of the Kronecker lift (\texttt{kron\_\allowbreak{}spectral}) and the softmax operating-point bounds (\texttt{softmaxN\_\allowbreak{}ratio\_\allowbreak{}bound}) are in, and Cochran's theorem for the LayerNorm statistic is derived on the finite-$t$ row. \\
\midrule
\multicolumn{2}{@{}p{\linewidth}@{}}{\textbf{Lem~\ref{lem:quotient_F} at matrix level (Track A).} \texttt{matProdMap\_\allowbreak{}gauge\_\allowbreak{}invariant}, \texttt{matProdMap\_\allowbreak{}fiber}, \texttt{hasDerivAt\_\allowbreak{}pertList\_\allowbreak{}prod}, \texttt{chainDeriv\_\allowbreak{}vertical}}\\*[2pt]
\emph{Proved.} The matrix quotient's first two deliverables: the end-to-end product is gauge invariant by telescoping, the fibers on the invertible stratum are exactly the gauge orbits (a group-general lemma: equal-length lists with equal products connect by the interpolating gauge family), the product rule down the layer list is certified entrywise, and the chain derivative kills every vertical family exactly, at every point.
& \emph{Assumed or not covered.} The fractional-power closed form on the balanced manifold, and the abstract generality of the slice theorem and the O'Neill calculus. The concrete quotient program is complete: the gauge action is proper with Hausdorff orbit space, the orbit space is identified with the units through the descended total (\texttt{orbitSpaceHomeo}), and the quotient metric is well-defined on lifts with the constant slice realizing it as the flat form on the base (\texttt{quotFisher\_\allowbreak{}vert\_\allowbreak{}translate}, \texttt{slice\_\allowbreak{}isometric}). The balanced flow holds in general position: the transport identity needs no balance and no invertibility, and the balance defect is conserved along Euclidean gradient flow, certified entrywise (\texttt{balance\_\allowbreak{}transport}, \texttt{balance\_\allowbreak{}defect\_\allowbreak{}hasDerivAt}). A3 and A4 followed: the chain derivative is onto through the single-slot inverse-tail direction, the closed-form pairing makes the kernel exactly the vertical space, and the quotient Fisher rate composes to $2(L-1)$ with coefficient $L$ at matrix level (\texttt{quotient\_\allowbreak{}rate\_\allowbreak{}matrix}). \\
\midrule
\multicolumn{2}{@{}p{\linewidth}@{}}{\textbf{Thm~\ref{thm:selection_rule}(c) at $m_\perp = 0$, and (a), (b) at the two-block model.} \texttt{volume\_\allowbreak{}sublevel\_\allowbreak{}pow}, \texttt{tendsto\_\allowbreak{}log\_\allowbreak{}div\_\allowbreak{}log\_\allowbreak{}of\_\allowbreak{}rpow}, \texttt{slice\_\allowbreak{}rlct\_\allowbreak{}slope}, \texttt{block\_\allowbreak{}lambda\_\allowbreak{}min\_\allowbreak{}sandwich}, \texttt{selection\_\allowbreak{}rule\_\allowbreak{}transversal\_\allowbreak{}lower}, \texttt{selection\_\allowbreak{}rule\_\allowbreak{}tangential\_\allowbreak{}lower}, \texttt{two\_\allowbreak{}by\_\allowbreak{}two\_\allowbreak{}eigvec\_\allowbreak{}tilt}}\\*[2pt]
\emph{Proved.} The volume law $\mathrm{vol}\{t : t^{2k} < \varepsilon\} = 2\,\varepsilon^{1/(2k)}$ exactly, and a log-volume slope tending to $\rlct = 1/(2k)$. Clauses (a) and (b) at the two-block model: an entry of rate $q$ against a block of rate $p < q$ with cross entries at rate $p + q$ and the Schur-type margin eventually dominates $(g/2)\|v\|^2$, so the variational minimum is between $g/2$ and $g$ at the entry's own rate, transversal at $p = 0$ and tangential at $2j_1 < 2j_2$; the small-eigenvalue eigenvector of a $2 \times 2$ block tilts by at most $|B|/(A - D)$ toward the large axis.
& \emph{Assumed or not covered.} Clause (c) runs on the normal form with no resolution of singularities. Clause (a) is also assembled from the score side (\texttt{selection\_\allowbreak{}rule\_\allowbreak{}assembled}): the three entry expansions, the transversal rate, the cross entries at order $k-1$, and the tangential block $g + O(t)$, feed the sandwich under the graded margin $2k^2\sum_\alpha \expect[ab_\alpha]^2/(k^2\expect[a^2]) < \lambda/2$, which is hypothesis~(ii$^*$) in the form the Schur argument consumes. \\
\midrule
\multicolumn{2}{@{}p{\linewidth}@{}}{\textbf{Thm~\ref{thm:selection_rule}(c), general $m_\perp$.} \texttt{normalForm\_\allowbreak{}diagScale}, \texttt{sublevel\_\allowbreak{}normalForm\_\allowbreak{}eq\_\allowbreak{}image}, \texttt{volume\_\allowbreak{}sublevel\_\allowbreak{}normalForm}, \texttt{slice\_\allowbreak{}rlct\_\allowbreak{}product\_\allowbreak{}slope}}\\*[2pt]
\emph{Proved.} On the slice normal form $u^{2k} + \sum_i v_i^2$ the sublevel volume scales exactly as $V(\varepsilon) = \varepsilon^{1/(2k) + m_\perp/2}\, V(1)$ with $0 < V(1) < \infty$, and the log-volume slope reads $\rlct = 1/(2k) + m_\perp/2$.
& \emph{Assumed or not covered.} The slice normal form itself. Tangential-normal coupling and the full local RLCT are not formalised, which matches the scope note of the theorem. \\
\midrule
\multicolumn{2}{@{}p{\linewidth}@{}}{\textbf{Thm~\ref{thm:nu_universality}, the frozen-$Z$ skeleton.} \texttt{frozen\_\allowbreak{}half\_\allowbreak{}line}, \texttt{frozen\_\allowbreak{}mean\_\allowbreak{}T}, \texttt{frozen\_\allowbreak{}mean\_\allowbreak{}sqrtT}, \texttt{frozen\_\allowbreak{}var\_\allowbreak{}even}, \texttt{frozen\_\allowbreak{}var\_\allowbreak{}odd}, \texttt{nu\_\allowbreak{}scale\_\allowbreak{}invariance}, \texttt{nuLO\_\allowbreak{}one}, \texttt{renormVar\_\allowbreak{}riccati}, \texttt{nuLO\_\allowbreak{}odd\_\allowbreak{}gt}}\\*[2pt]
\emph{Proved.} Under $\pi_0 \propto e^{-\eta^{2k}}$ the statistic $T = \eta^{2k}$ has the $\Gamma(\lambda, 1)$ law with $\lambda = 1/(2k)$, so $\expect[T] = \lambda$ and $\expect[\sqrt{T}] = \Gamma(\lambda + \tfrac12)/\Gamma(\lambda)$; $\mathrm{Var}_{\pi_0}[\eta^k]$ is $\nu^{\sqrt{T}}(k)$ for even $k$ and $\lambda$ for odd $k$; and the model-scale posterior $\exp(A s^k - B s^{2k})$ read in $\eta = B^{1/(2k)} s$ is $\pi_Z$ at $Z = A/\sqrt{B}$, so every expectation ratio depends on $(A, B)$ only through $Z$; at $k = 1$ the renormalised posterior is $\normal(Z/2, 1/2)$ and $\nu_{\mathrm{LO}}(1) = 1/2$ exactly. The Stein identity gives $\mathrm{Var}_{\pi_Z}[\eta^k] = \lambda + m(Z)(Z/2 - m(Z))$ at every order, and at odd $k \ge 3$ the bounds $0 \le m(Z) \le Z/2$ make the $Z$-average strictly larger than the frozen value.
& \emph{Assumed or not covered.} The leading-order replacement of the posterior, the central-limit step $Z \to \normal(0, 2)$, the $Z$-averaged value $\nu_{\mathrm{LO}}(k)$ for $k \ge 2$, which has no closed form, and the strict inequality at even $k$, proved outside Lean (per-order certificates to $k = 1000$, a uniform argument beyond). \\
\midrule
\multicolumn{2}{@{}p{\linewidth}@{}}{\textbf{Rem~\ref{rem:nu_tempered}, tempered posteriors.} \texttt{renormVar\_\allowbreak{}riccati}, \texttt{nu\_\allowbreak{}scale\_\allowbreak{}invariance}, \texttt{nuLO\_\allowbreak{}one}}\\*[2pt]
\emph{Proved.} The inner variance of the $\beta$-tilted limit is the same exact relation $\mathrm{Var}_{\pi_Z}[\eta^k] = \lambda + m(Z)(Z/2 - m(Z))$; model-independence at fixed $\beta$ is the same scale invariance; the $k = 1$ flat value $1/2$ is the same Gaussian computation.
& \emph{Assumed or not covered.} The tempered central-limit step $Z \to \normal(0, 2\beta)$ and the gap identity enter as the paper's derivation; the quadrature values $f_2(2\beta)$ at $\beta = 1/2, 1, 2$ and their Monte Carlo matches stay numerical. \\
\midrule
\multicolumn{2}{@{}p{\linewidth}@{}}{\textbf{Prop~\ref{prop:determinantal_resolution} at width one.} \texttt{width\_\allowbreak{}one\_\allowbreak{}chain\_\allowbreak{}volume}, \texttt{width\_\allowbreak{}one\_\allowbreak{}chain\_\allowbreak{}slope}, \texttt{morse\_\allowbreak{}block\_\allowbreak{}volume}, \texttt{morse\_\allowbreak{}block\_\allowbreak{}slope}}\\*[2pt]
\emph{Proved.} At rank zero the depth-$L$ chain $K = (\prod_i w_i)^2$ has $\rlct = 1/2$ with multiplicity $L$, its volume law the degree-$(L-1)$ log polynomial exactly; at rank one the regular direction $K = (w - p)^2$ has $\rlct = 1/2$ with multiplicity one, volume $2\sqrt{\varepsilon}$ exactly.
& \emph{Assumed or not covered.} The SVD change of variables, the Vandermonde monomialisation, and the Newton-polyhedron formula at general widths and ranks. \\
\end{longtable}}
  
\end{document}